%% file: main_thesis.tex
\documentclass[12pt,a4paper,twoside]{report}

\usepackage{utils/packages}
\usepackage{utils/commands}
\usepackage{eso-pic}
\newif\ifformal
\formalfalse

\title{\titlethesis}
\author{Imad Aouali}
\begin{document}

\includepdf[pages=-]{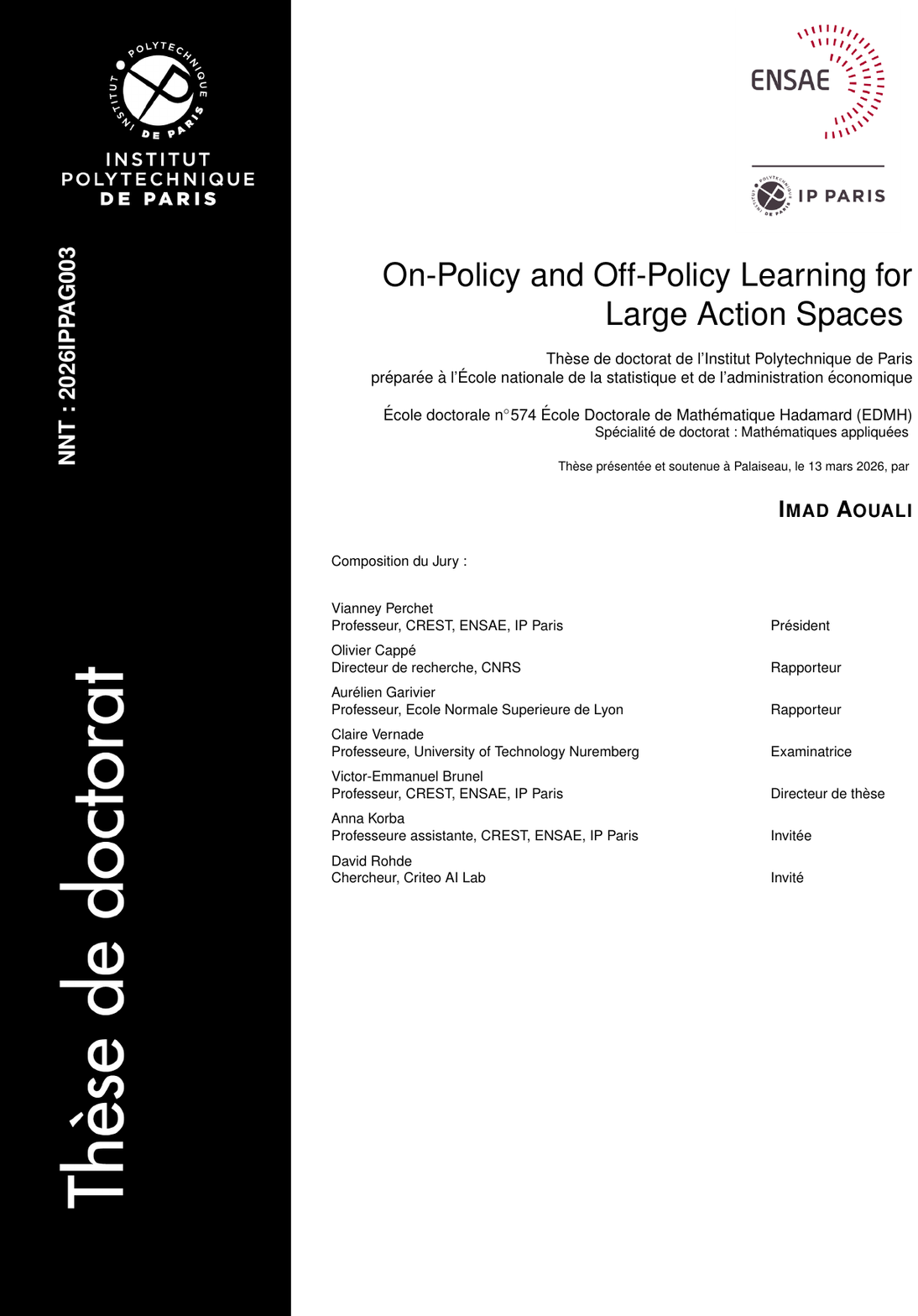}

\newpage

\chapter*{Acknowledgements}

First and foremost, I wish to express my sincere and profound gratitude to my PhD supervisors, Victor-Emmanuel Brunel, Anna Korba, and David Rohde. It has been an immense privilege to learn from and work with them over these years. They shaped my research and personal growth in ways that will stay with me far beyond this thesis.

Victor brought rigor, kindness, and openness to every discussion, profoundly shaping how I approach research and problem-solving. Anna’s brilliance, drive, and compassionate mentorship were central to the success of this PhD; her rare ability to combine deep technical insight with empathy and encouragement made her guidance invaluable. David was the best manager I could have hoped for, whose trust, patience, and human approach made all the difference when navigating both professional and personal challenges.

I am profoundly grateful to the members of my thesis jury for the time, care, and expertise they devoted to evaluating this work. I would first like to express my deepest appreciation to Olivier Cappé and Aurélien Garivier for accepting the demanding role of rapporteurs, and for the considerable time and attention they dedicated to reading this manuscript in depth. I am sincerely grateful for their careful assessment, thoughtful comments, and constructive feedback. I would also like to warmly thank Vianney Perchet for the honor of presiding over the jury, and for his invaluable support. Finally, I am deeply grateful to Claire Vernade for serving as examinatrice, and for her generosity, encouragement, and support. It was a true privilege to have such distinguished researchers on my jury, and I deeply appreciate their scientific perspective, insightful remarks, and kindness.

I would also like to thank Criteo, CAIL, and the Performance Science team, as well as CREST, ENSAE, Institut Polytechnique de Paris. Both the company and the laboratory provided outstanding scientific and institutional support throughout this thesis.

I am deeply grateful to several mentors: to Branislav Kveton, whose insights as my first major co-author shaped my approach to research and publication; to Florian Strub, my DeepMind scholarship mentor, whose guidance inspired me to pursue a PhD; and to Flavian Vasile, and Michal Valko, for their invaluable support, both seen and unseen.

My heartfelt thanks also go to all my co-authors, whose contributions have greatly enriched this work: A. Gilotte, B. Heymann, N. Nguyen, A. György, P. Alquier, N. Chopin, S. Katariya, AAS. Hammou, S. Ivanov, A. Benhalloum, M. Bompaire, M. Vono, M. Gartrell, V. Zaytsev, D. Legrand, and O. Jeunen. Special recognition goes to O. Sakhi, M. Cherifa and A.B. Yahmed who were exceptional companions throughout this journey.

Finally, to my family and friends: my deepest thanks to my mother and siblings for their unwavering love and support. This thesis is dedicated to my late father, whose dream was for me to pursue a PhD. To Basma, Hakim, Achraf, Ayman, Ismail, Tayeb, Anas, Nicolas, Youssef, Kini, Song, Issam, Charif, Yassine, Oussama, Abdellah, Ali, Hicham: thank you for your friendship and presence throughout this journey. 

\chapter*{Abstract} 

Many interactive systems (e.g., recommender systems) can be modeled as contextual bandits. This framework captures the core challenge of decision-making under uncertainty: selecting actions based on context while learning from partial, noisy feedback. Learning in this setting follows two paradigms: \textit{on-policy learning}, in which agents collect data and update their policy simultaneously in real time, and \textit{off-policy learning}, in which the agent's policy is learned offline from static logs collected under a different policy. Standard algorithms for both paradigms struggle to scale to large action spaces, facing either computational intractability or statistical inefficiency. This thesis develops principled and practical methods to make contextual bandit algorithms tractable in large action spaces, advancing both paradigms through novel algorithmic and theoretical contributions.

For on-policy learning, we introduce structured Bayesian models that enable efficient exploration via information sharing. Our first contribution, mixed-effect Thompson sampling (\texttt{meTS}) (\cref{chap:meTS}), couples action parameters through shared latent effects. This reduces Bayesian regret to $\tilde{\mathcal{O}}(\sqrt{TdK_{\mathrm{eff}}})$, where $K_{\mathrm{eff}}$ is an effective number of actions. When the number of shared effects is much smaller than the number of actions ($L \ll K$), we have $K_{\mathrm{eff}} \ll K$, yielding significant regret reduction. Moreover, \texttt{meTS} achieves dramatic improvements in both memory complexity (from $\mathcal{O}(K^2d^2)$ to $\mathcal{O}((L^2+K)d^2)$) and runtime complexity (from $\mathcal{O}(K^3d^3)$ to $\mathcal{O}((L^3+K)d^3)$). We extend this framework to diffusion Thompson sampling (\texttt{dTS}) (\cref{chap:dTS}), which leverages deep generative models to capture complex action distributions. \texttt{dTS} further improves memory complexity to $\mathcal{O}((L+K)d^2)$ and runtime complexity to $\mathcal{O}((L+K)d^3)$, where $L$ denotes the number of layers in the diffusion model. Both methods perform well empirically as analyzed without additional hyperparameter tuning, making them highly practical.

For off-policy learning, we address fundamental bottlenecks through three complementary approaches. First, in \cref{chap:sDM}, we develop the structured direct method (\texttt{sDM}), which models action parameters using a shared latent structure. We prove that \texttt{sDM} achieves $\mathcal{O}(1/\sqrt{n})$ convergence in Bayesian suboptimality without requiring the restrictive full logging support assumption. \texttt{sDM} performs well in practice, and the performance gap between \texttt{sDM} and standard direct methods widens as the action space grows. Second, in \cref{chap:las}, we challenge the conventional wisdom that better reward estimation yields better policies. We demonstrate that optimization intractability, rather than estimation accuracy, becomes the primary bottleneck in large action spaces, and advocate for policy-weighted log-likelihood objectives that prioritize optimization tractability; these consistently outperform sophisticated estimators on datasets with up to one million actions. Third, in \cref{chap:ES}, we address importance sampling variance by combining variance-reducing regularization with principled pessimism. Our exponential smoothing estimators and unified PAC-Bayesian analysis yield tractable learning objectives amenable to stochastic optimization, providing concentration bounds and superior empirical performance.

We validate these theoretical and algorithmic advances through extensive experiments on synthetic and real-world datasets. By developing scalable algorithms for both learning paradigms, this thesis enables the deployment of contextual bandits in modern applications where action spaces routinely exceed thousands or millions of actions.

\newpage
{
\hypersetup{linkcolor=black}
\tableofcontents
\newpage
}

\chapter*{Notation} 
\input{contents/notation}

\newpage

\input{contents/resume}

\chapter{Overview}\label{chap:overview}
\minitoc
\input{contents/overview/context}

\input{contents/overview/background}

\input{contents/overview/contributions}
\input{contents/overview/related_work}


\part{On-Policy Learning in Large Action Spaces}
\label{part:online}

\input{contents/part1/introduction}

\chapter{Scaling Thompson Sampling with Mixed Effects}\label{chap:meTS}
\minitoc
\input{contents/mixed_effect_ts/main_text}

\chapter{Scaling Thompson Sampling with Diffusion Models}\label{chap:dTS}
\minitoc

\input{contents/dts/main_text}


\part{Off-Policy Learning in Large Action Spaces}
\label{part:offline}

\input{contents/part2/introduction}

\chapter{Scaling Direct Methods with Latent Parameters}\label{chap:sDM}
\minitoc

\input{contents/sdm/main_text}

\chapter{Optimization Matters More than Esimation}\label{chap:las}
\minitoc

\input{contents/las/main_text}

\chapter{Principled Pessimism for Exponential Smoothing and Beyond}\label{chap:ES}
\minitoc

\input{contents/exp_smoothing/main_text}

\input{contents/conclusion}

\appendix

\chapter{Supplementary Materials for \cref{chap:meTS}}
\minitoc
\input{contents/mixed_effect_ts/appendix}

\chapter{Supplementary Materials for \cref{chap:dTS}}
\minitoc
\input{contents/dts/appendix}

\chapter{Supplementary Materials for \cref{chap:sDM}}
\minitoc
\input{contents/sdm/appendix}

\chapter{Supplementary Materials for \cref{chap:las}}
\minitoc
\input{contents/las/appendix}

\chapter{Supplementary Materials for \cref{chap:ES}}
\minitoc
\input{contents/exp_smoothing/appendix}

\bibliography{bib} 

\clearpage
\thispagestyle{empty}
\pagestyle{empty}

\AddToShipoutPictureBG*{%
  \AtPageLowerLeft{%
    \includegraphics[
      page=1,
      width=\paperwidth,
      height=\paperheight
    ]{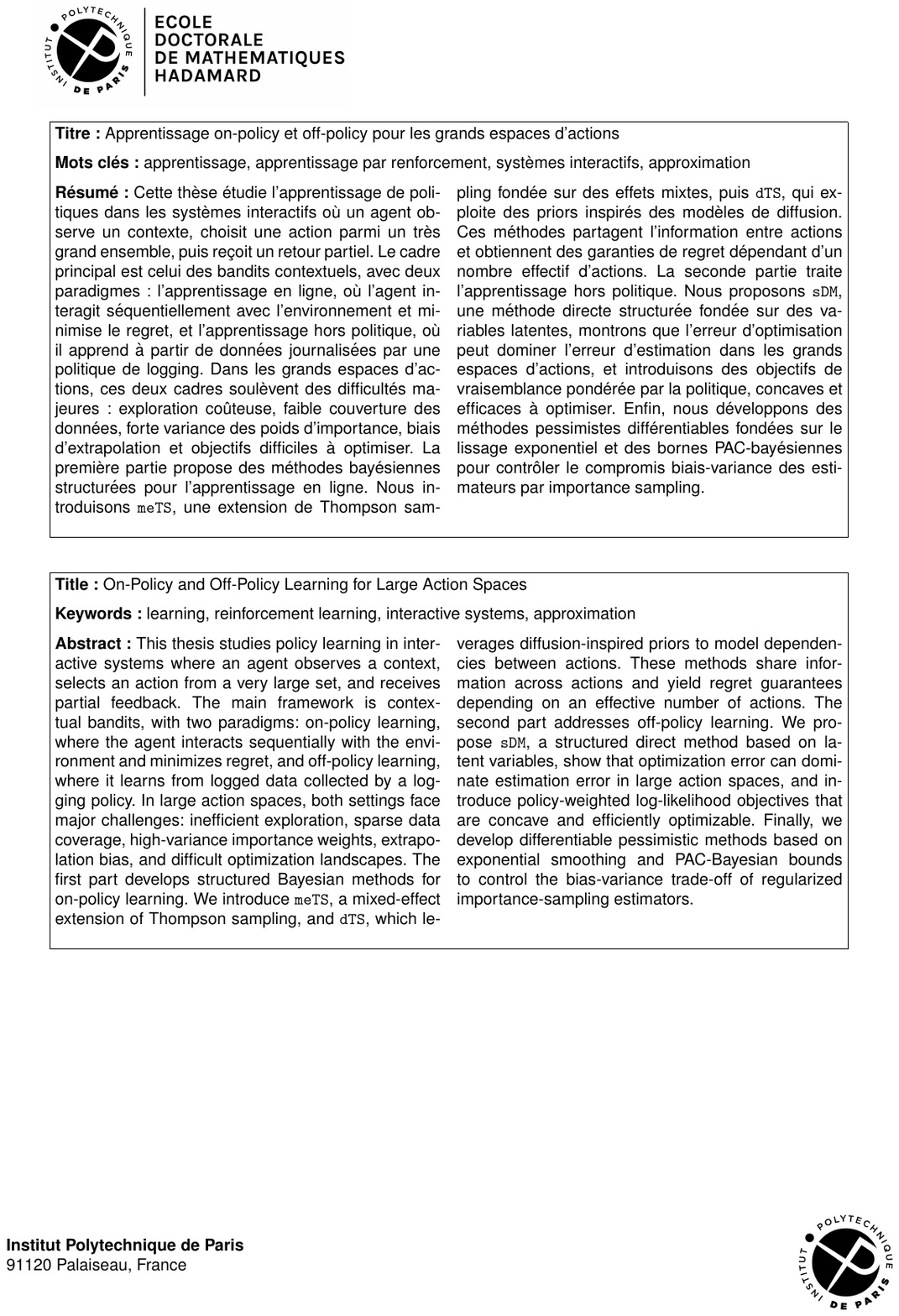}%
  }%
}
\null

\end{document}

%% file: contents/notation.tex
\section*{Notation}

\subsection*{General Mathematical Notation}
\begin{tabular}{@{}p{3.5cm}p{12cm}@{}}
\toprule
\textbf{Symbol} & \textbf{Definition} \\
\midrule
$[n]$ & Set of first $n$ positive integers: $\{1, 2, \ldots, n\}$ \\
$\mathbb{R}^d$ & $d$-dimensional real vector space \\
$I_d$ & Identity matrix of dimension $d \times d$ \\
$\Delta(\mathcal{A})$ & Probability simplex over action set $\mathcal{A}$ \\
$\mathcal{O}(\cdot)$ & Big-O notation for upper bounds \\
$\otimes$ & Kronecker product \\
\bottomrule
\end{tabular}

\noindent \textbf{Probability conventions.} Random variables are denoted with capital letters, and their realizations with the respective lowercase letters, except for Greek letters. With a slight abuse of notation, for random variables $X, Y$, the distribution (or density) of $X \mid Y=y$ evaluated at $x$ is denoted by $p(x \mid y)$.

\subsection*{Norms and Inner Products}
\begin{tabular}{@{}p{3.5cm}p{12cm}@{}}
\toprule
\textbf{Symbol} & \textbf{Definition} \\
\midrule
$\|\cdot\|$ & Euclidean norm (unless specified otherwise) \\
$\|a\|_{\Sigma}$ & Weighted norm: $\sqrt{a^\top \Sigma a}$ for $a \in \mathbb{R}^d$, $\Sigma \succ 0$ \\
\bottomrule
\end{tabular}

\subsection*{Contextual Bandit Framework}
\begin{tabular}{@{}p{3.5cm}p{12cm}@{}}
\toprule
\textbf{Symbol} & \textbf{Definition} \\
\midrule
$\mathcal{X} \subset \mathbb{R}^d$ & Context space ($d$-dimensional) \\
$\mathcal{A} = [K]$ & Finite action set with $K$ actions \\
$T$ & Number of interaction rounds \\
$n$ & Number of samples in logged dataset \\
\bottomrule
\end{tabular}

\subsection*{Policies and Value Functions}
\begin{tabular}{@{}p{3.5cm}p{12cm}@{}}
\toprule
\textbf{Symbol} & \textbf{Definition} \\
\midrule
$\pi : \mathcal{X} \to \Delta(\mathcal{A})$ & Stochastic policy mapping contexts to action distributions \\
$\pi(\cdot \mid x)$ & Probability distribution over actions given context $x$ \\
$\pi_t$ & Policy at round $t$ (on-policy setting) \\
$\pi_0$ & Logging policy (off-policy setting) \\
$\pi_*$ & Optimal policy (off-policy setting) \\
$\hat{\pi}$ & Learned policy (off-policy setting) \\
$V(\pi)$ & Value (expected reward) of policy $\pi$ \\
$\hat{V}(\pi)$ & Estimated value of policy $\pi$ \\
\bottomrule
\end{tabular}

\subsection*{Environment Components}
\begin{tabular}{@{}p{3.5cm}p{12cm}@{}}
\toprule
\textbf{Symbol} & \textbf{Definition} \\
\midrule
$\nu$ & Context distribution over $\mathcal{X}$ \\
$p(\cdot \mid x, a)$ & Conditional reward distribution given context $x$ and action $a$ \\
$r(x, a)$ & Expected reward function: $\mathbb{E}_{R \sim p(\cdot \mid x,a)}[R]$ \\
$\hat{r}(x, a)$ & Estimated reward function \\
$r(x, a; \theta)$ & Parametric reward model \\
$r(x, a; \hat{\theta})$ & Estimated reward function in parametric case\\
\bottomrule
\end{tabular}

\subsection*{Random Variables and Data}
\begin{tabular}{@{}p{3.5cm}p{12cm}@{}}
\toprule
\textbf{Symbol} & \textbf{Definition} \\
\midrule
$X_t$ & Context observed at round $t$ \\
$A_t$ & Action taken at round $t$ \\
$R_t$ & Reward received at round $t$ \\
$A_{t,*}$ & Optimal action at round $t$: $\arg\max_{a \in \mathcal{A}} r(X_t, a)$ \\
$H_t$ & History of interactions: $\{(X_\ell, A_\ell, R_\ell)\}_{\ell < t}$ \\
$\mathcal{D}_n$ & Logged dataset: $\{(X_i, A_i, R_i)\}_{i=1}^n$ \\
\bottomrule
\end{tabular}

\subsection*{Performance Metrics}
\begin{tabular}{@{}p{6.5cm}p{9cm}@{}}
\toprule
\textbf{Symbol} & \textbf{Definition} \\
\midrule
$\mathcal{R}(T) = \sum_{t=1}^T r(X_t, A_{t,*}) - r(X_t, A_t)$ & Cumulative regret \\
$\mathcal{BR}(T) = \mathbb{E}[\mathcal{R}(T)]$ & Bayesian cumulative regret \\
$\textsc{so}(\hat{\pi}) = V(\pi_*) - V(\hat{\pi})$ & Suboptimality gap of policy $\hat{\pi}$ \\
$\textsc{Bso}(\hat{\pi}) = \mathbb{E}[\textsc{so}(\hat{\pi})]$  & Bayesian suboptimality gap of policy $\hat{\pi}$\\
\bottomrule
\end{tabular}

\subsection*{Matrix Operations and Concatenation}
\begin{tabular}{@{}p{3.5cm}p{12cm}@{}}
\toprule
\textbf{Symbol} & \textbf{Definition} \\
\midrule
$[a_1, a_2, \ldots, a_n]$ & Horizontal concatenation of vectors into $d \times n$ matrix \\
$(a_i)_{i \in [n]}$ & Vertical concatenation: $(a_1^\top, \ldots, a_n^\top)^\top \in \mathbb{R}^{nd}$ \\
$\operatorname{Vec}(\cdot)$ & Vectorization operator \\
$\mathrm{diag}((\Alpha_i)_{i \in [n]})$ & Block diagonal matrix with blocks $\Alpha_1, \ldots, \Alpha_n$ \\
$(\Alpha_i)_{i \in [n]}$ & Vertical concatenation of matrices into $nd \times d$ matrix \\
$(\Alpha_{i,j})_{(i,j) \in [n] \times [m]}$ & Block matrix where $\Alpha_{i,j}$ is the $(i,j)$-th block \\
\bottomrule
\end{tabular}

\subsection*{Eigenvalues}
\begin{tabular}{@{}p{3.5cm}p{12cm}@{}}
\toprule
\textbf{Symbol} & \textbf{Definition} \\
\midrule
$\lambda_1(\Alpha)$ & Maximum eigenvalue of matrix $\Alpha$ \\
$\lambda_d(\Alpha)$ & Minimum eigenvalue of matrix $\Alpha$ \\
\bottomrule
\end{tabular}

%% file: contents/resume.tex
\chapter*{Résumé substantiel en français}
\addcontentsline{toc}{chapter}{Résumé substantiel en français}

Cette thèse étudie l'apprentissage séquentiel et contrefactuel dans des systèmes interactifs où l'espace des décisions est très grand. Ces systèmes sont aujourd'hui omniprésents : moteurs de recommandation, publicité computationnelle, places de marché, systèmes de tarification, robotique ou encore allocation de ressources. Leur fonctionnement repose sur une boucle d'interaction simple mais difficile à optimiser : à chaque étape, le système observe un contexte, choisit une action parmi un très grand nombre de possibilités, puis reçoit une récompense partielle et bruitée qui dépend conjointement du contexte et de l'action choisie. Dans un système de recommandation, le contexte peut représenter l'historique et les préférences d'un utilisateur, l'action correspond à l'item recommandé dans un catalogue contenant potentiellement des millions d'items, et la récompense mesure l'engagement de l'utilisateur, par exemple un clic ou un temps de visionnage. En publicité en ligne, l'action peut combiner le choix d'une annonce et d'un prix d'enchère, tandis que la récompense dépend d'événements successifs tels que le gain de l'enchère, le clic ou la conversion.

Le cadre mathématique central de cette thèse est celui des bandits contextuels. On considère un espace de contextes $\mathcal{X}\subset\mathbb{R}^d$, un ensemble fini d'actions $\mathcal{A}=[K]$, une distribution inconnue de contextes $\nu$, et des distributions conditionnelles de récompenses $p(\cdot\mid x,a)$. La fonction de récompense moyenne est définie par
\[
r(x,a)=\mathbb{E}[R\mid X=x,A=a].
\]
À chaque tour $t$, un contexte $X_t\sim\nu$ est observé, l'agent choisit une action $A_t$ selon une politique $\pi_t(\cdot\mid X_t)$, puis reçoit une récompense $R_t\sim p(\cdot\mid X_t,A_t)$. Ce formalisme capture l'essentiel de l'apprentissage interactif : l'agent ne voit que la récompense de l'action qu'il a effectivement choisie, et doit donc apprendre à partir d'un retour partiel. La difficulté principale analysée dans cette thèse est le passage à l'échelle lorsque $K$ est très grand.

La thèse traite deux paradigmes complémentaires. Le premier est l'apprentissage en ligne, ou \emph{on-policy learning}, dans lequel l'agent interagit séquentiellement avec l'environnement et met à jour sa politique au fil des observations. Sa performance est mesurée par le regret cumulé,
\[
\mathcal{R}(T)=\sum_{t=1}^T \left(r(X_t,A_{t,*})-r(X_t,A_t)\right),
\]
où $A_{t,*}$ désigne l'action optimale dans le contexte $X_t$. L'enjeu fondamental est le compromis exploration-exploitation : l'agent doit explorer les actions incertaines pour apprendre, tout en exploitant les actions déjà estimées comme performantes afin de limiter la perte de récompense. Le second paradigme est l'apprentissage hors politique, ou \emph{off-policy learning}, dans lequel l'agent ne peut plus interagir avec l'environnement et doit apprendre à partir d'un jeu de données journalisé
\[
\mathcal{D}_n=\{(X_i,A_i,R_i)\}_{i=1}^n
\]
collecté par une politique de logging $\pi_0$. L'objectif est alors d'apprendre une politique $\hat{\pi}$ de grande valeur
\[
V(\pi)=\mathbb{E}_{X\sim\nu}\mathbb{E}_{A\sim\pi(\cdot\mid X)}[r(X,A)],
\]
et la performance est mesurée par l'écart de sous-optimalité $V(\pi_*)-V(\hat{\pi})$. Dans ce second cadre, l'apprentissage exige un raisonnement contrefactuel : il faut estimer ce qui se serait passé si une autre action avait été choisie, alors même que les données ne contiennent que les actions effectivement sélectionnées par $\pi_0$.

Dans les deux paradigmes, la grande taille de l'espace d'actions amplifie les difficultés statistiques, computationnelles et d'optimisation. En ligne, explorer indépendamment des milliers ou millions d'actions devient prohibitif : chaque action reçoit peu d'observations, ce qui ralentit considérablement l'apprentissage et augmente le regret. Hors politique, la couverture des données se dégrade lorsque $K$ croît : de nombreuses actions sont rarement ou jamais observées dans certaines régions de l'espace des contextes. Les méthodes fondées sur un modèle de récompense souffrent alors d'un fort biais d'extrapolation, tandis que les méthodes par pondération inverse des propensions peuvent avoir une variance très élevée lorsque $\pi_0(a\mid x)$ est faible. Cette thèse montre également qu'un autre obstacle, souvent sous-estimé, devient dominant dans les grands espaces d'actions : l'optimisation des objectifs hors politique standards peut devenir intrinsèquement difficile, indépendamment de la qualité statistique de l'estimateur de valeur.

La première partie de la thèse est consacrée à l'apprentissage en ligne dans les grands espaces d'actions. Les méthodes classiques telles que \emph{Upper Confidence Bound} et \emph{Thompson Sampling} reposent souvent sur des modèles disjoints, où chaque action $a$ possède son propre paramètre $\theta_a$ et où la récompense est modélisée sous la forme $r(x,a)=\phi(x)^\top\theta_a$. Ces modèles sont attractifs en pratique, notamment dans les systèmes de recommandation, car ils évitent de construire manuellement des caractéristiques conjointes contexte-action. Cependant, leur faiblesse est statistique : apprendre séparément un paramètre pour chaque action nécessite beaucoup de données par action, ce qui est incompatible avec de très grands catalogues. La contribution principale de cette partie consiste à conserver la flexibilité des modèles disjoints tout en introduisant des structures bayésiennes capables de partager l'information entre actions.

Le premier algorithme proposé est \emph{mixed-effect Thompson Sampling}, ou \texttt{meTS}. Il repose sur un modèle bayésien hiérarchique où les paramètres d'actions $\theta_a$ sont couplés par des effets latents partagés $\Psi=(\psi_\ell)_{\ell\in[L]}$. Ces effets peuvent représenter, par exemple, des catégories ou des facteurs communs entre items. Le modèle suppose que les paramètres d'actions sont conditionnellement indépendants sachant les effets latents, mais qu'ils partagent de l'information à travers ces effets. À chaque tour, \texttt{meTS} échantillonne d'abord les effets latents depuis leur postérieur, puis échantillonne les paramètres d'actions conditionnellement à ces effets, avant de choisir l'action maximisant la récompense échantillonnée. Cette procédure conserve le principe de Thompson Sampling tout en rendant l'exploration statistiquement plus efficace.

Dans le cas linéaire-gaussien, la thèse dérive des mises à jour exactes en forme fermée, et propose des approximations de Laplace tractables pour les modèles linéaires généralisés. L'analyse théorique établit une borne de regret bayésien de l'ordre
\[
\widetilde{\mathcal{O}}\!\left(\sqrt{T d K_{\mathrm{eff}}(\sigma_0^2+\sigma_\Psi^2)}\right),
\]
où $K_{\mathrm{eff}}$ est un nombre effectif d'actions. Lorsque la structure latente est informative et que $L\ll K$, on a $K_{\mathrm{eff}}\ll K$, ce qui conduit à une amélioration multiplicative par rapport à Thompson Sampling standard. Sur le plan computationnel, la factorisation conditionnelle réduit fortement les coûts mémoire et temps par rapport à une modélisation bayésienne dense de toutes les actions. Les expériences montrent que les gains de \texttt{meTS} augmentent avec la taille de l'espace d'actions, confirmant que le partage d'information est essentiel pour l'exploration à grande échelle.

La seconde contribution en ligne est \emph{diffusion Thompson Sampling}, ou \texttt{dTS}. Cette méthode généralise \texttt{meTS} en remplaçant la hiérarchie à un niveau par une hiérarchie profonde inspirée des modèles de diffusion. Les paramètres d'actions sont générés au terme d'une chaîne de variables latentes reliées par des transformations non linéaires pré-entraînées. Cette structure permet de représenter des dépendances complexes entre actions, bien au-delà des effets linéaires ou catégoriels. Le défi technique est que le postérieur exact devient intraitable à cause des non-linéarités des fonctions de lien et du modèle de récompense. La thèse introduit donc une procédure d'inférence en ligne fondée sur des mises à jour de type gaussien, où les précisions postérieures combinent précision a priori et précision issue des données, et où les moyennes sont obtenues par combinaison pondérée entre les prédictions du prior et les estimations de maximum de vraisemblance.

L'intérêt de \texttt{dTS} est double. D'une part, l'algorithme exploite des priors riches appris hors ligne, par exemple à partir de représentations d'items, pour accélérer l'exploration en ligne. D'autre part, il conserve une structure de diffusion dans le postérieur, plutôt que de l'approximer par une simple gaussienne globale. Dans le cadre linéaire-gaussien, la thèse établit une borne de regret bayésien de l'ordre
\[
\widetilde{\mathcal{O}}\!\left(\sqrt{T d K_{\mathrm{eff}}\sum_{\ell=1}^{L+1}\sigma_\ell^2}\right),
\]
et montre que la complexité peut être rendue linéaire en $L+K$. Empiriquement, \texttt{dTS} améliore les performances des méthodes de référence, y compris lorsque les priors de diffusion sont imparfaits ou appris à partir de données limitées. Cette contribution montre que les modèles génératifs profonds peuvent être utilisés non seulement pour représenter des actions, mais aussi pour structurer l'incertitude nécessaire à l'exploration.

La seconde partie de la thèse porte sur l'apprentissage hors politique dans les grands espaces d'actions. Les méthodes classiques se divisent principalement en méthodes directes, qui apprennent un modèle de récompense $\hat r(x,a)$, et méthodes par \emph{importance sampling}, qui estiment directement la valeur d'une politique au moyen du ratio $\pi(a\mid x)/\pi_0(a\mid x)$. Les méthodes directes sont sensibles au biais de modèle et à la rareté des observations par action. Les méthodes IPS sont non biaisées sous des hypothèses de support appropriées, mais leur variance peut exploser lorsque la politique cible attribue de la masse à des actions peu probables sous la politique de logging. En outre, la thèse montre que les objectifs IPS induisent souvent des paysages non concaves, plats et riches en maxima locaux lorsque l'espace d'actions est grand.

La première contribution hors politique est la \emph{structured Direct Method}, ou \texttt{sDM}. Elle transpose au cadre offline l'idée de structuration bayésienne introduite dans \texttt{meTS}. Au lieu d'estimer indépendamment un paramètre par action, \texttt{sDM} couple les paramètres $\theta_a$ à travers un vecteur latent partagé $\psi$. Après observation du jeu de données journalisé, l'algorithme calcule le postérieur de $\psi$, puis les postérieurs conditionnels des paramètres d'actions. La récompense estimée pour chaque action est obtenue en intégrant l'incertitude postérieure, et la politique apprise agit ensuite gloutonnement par rapport à cette récompense moyenne postérieure.

L'analyse introduit une notion de sous-optimalité bayésienne adaptée au cadre hors politique. Elle montre que \texttt{sDM} atteint une convergence en $\mathcal{O}(1/\sqrt n)$ sans imposer l'hypothèse de support uniforme complet $\pi_0(a\mid x)\geq\gamma>0$ pour toutes les actions. La borne dépend plutôt de l'alignement entre la politique de logging et la politique optimale : plus les actions optimales sont couvertes par les données, plus l'apprentissage est efficace. Cette analyse révèle également un phénomène important : sous le critère bayésien considéré, les politiques gloutonnes sont optimales et peuvent surpasser les politiques pessimistes, contrairement au cadre fréquentiste où le pessimisme est souvent nécessaire pour se protéger contre les pires cas. Les expériences confirment que \texttt{sDM} améliore les méthodes directes standards, avec des gains croissants lorsque $K$ augmente.

La contribution suivante remet en question une hypothèse centrale de l'apprentissage hors politique : l'idée selon laquelle l'amélioration des estimateurs de valeur suffit à améliorer l'apprentissage de politiques. La thèse montre que, dans les grands espaces d'actions, l'erreur d'optimisation peut dominer l'erreur d'estimation. Même un estimateur statistiquement sophistiqué peut conduire à une mauvaise politique si l'objectif qu'il induit est difficile à optimiser. L'analyse des paysages d'optimisation montre que les objectifs fondés sur des estimateurs peuvent présenter des plateaux où les méthodes de gradient restent bloquées pendant $\mathcal{O}(K)$ itérations, ainsi qu'un nombre exponentiel de maxima locaux en fonction de $K$.

Pour comprendre ces échecs, la thèse analyse les politiques oracle associées à différents estimateurs, c'est-à-dire les politiques qui maximiseraient ces estimateurs avec une quantité infinie de données. Cette analyse met en évidence que chaque estimateur impose un biais inductif spécifique : IPS favorise les politiques proches du support de logging, tandis que des méthodes clusterisées opèrent au niveau de groupes d'actions. Ces observations motivent des paramétrisations de politiques adaptées à l'objectif, qui réduisent l'espace de recherche effectif de $K$ à une taille beaucoup plus petite, comme la taille du support de logging ou le nombre de clusters.

La conclusion principale de cette partie est toutefois plus radicale : il peut être préférable d'abandonner l'estimation explicite de valeur pour optimiser directement des objectifs de vraisemblance pondérée par la politique. La thèse introduit les objectifs \emph{policy-weighted log-likelihood}, de la forme
\[
\hat U_g(\pi)=\frac1n\sum_{i=1}^n g(R_i,\pi_0(A_i\mid X_i))\log\pi(A_i\mid X_i),
\]
où $g$ est une fonction de pondération positive. Ces objectifs ne sont pas des estimateurs de valeur, mais ils possèdent des paysages d'optimisation bien plus favorables : pour des politiques softmax linéaires, ils sont concaves, et deviennent fortement concaves avec régularisation $\ell_2$. Ils admettent donc un optimum global unique accessible par optimisation stochastique standard. Les expériences à très grande échelle, incluant des espaces allant jusqu'à un million d'actions, montrent que ces objectifs simples et stables surpassent des méthodes hors politique fondées sur des estimateurs de valeur plus complexes. Cette contribution établit que, pour l'apprentissage de politiques dans les grands espaces d'actions, l'optimisabilité de l'objectif est un critère aussi fondamental que sa précision statistique.

La dernière contribution principale de la thèse améliore les méthodes IPS régularisées en introduisant un pessimisme praticable et différentiable. Les ratios d'importance peuvent être très grands, ce qui augmente fortement la variance. Pour y remédier, la thèse étudie des estimateurs par \emph{exponential smoothing}, notamment
\[
\hat V^\alpha(\pi)=\frac1n\sum_{i=1}^n
\frac{\pi(A_i\mid X_i)}{\pi_0(A_i\mid X_i)^\alpha}R_i,
\]
et
\[
\tilde V^\beta(\pi)=\frac1n\sum_{i=1}^n
\left(\frac{\pi(A_i\mid X_i)}{\pi_0(A_i\mid X_i)}\right)^\beta R_i.
\]
Ces estimateurs interpolent entre absence de régularisation et forte réduction de variance, tout en restant différentiables et donc compatibles avec l'optimisation par gradient.

Pour apprendre de manière sûre avec ces estimateurs biaisés mais moins variables, la thèse dérive une borne PAC-bayésienne bilatérale contrôlant l'écart entre la valeur vraie et l'estimateur régularisé. Cette borne décompose l'erreur en plusieurs termes interprétables : une divergence entre la politique apprise et la politique de logging, un biais dû à la régularisation des poids, et une variance résiduelle. Elle conduit à un objectif pessimiste qui maximise une borne inférieure empirique de la valeur :
\[
\hat V_n^\alpha(\pi_\theta)
-\text{pénalités de divergence}
-\text{biais de régularisation}
-\text{variance résiduelle}.
\]
L'intérêt crucial de cette formulation est que tous les termes sont empiriques et différentiables. Contrairement à de nombreuses approches pessimistes dont les constantes théoriques sont inexploitables en pratique, cet objectif peut être optimisé à grande échelle par ascension de gradient stochastique. La thèse propose également un cadre PAC-bayésien unifié couvrant plusieurs familles de régularisation des poids d'importance, comme le clipping, l'exponential smoothing et l'implicit exploration. Ce cadre permet une comparaison cohérente de différentes formes de pessimisme et fournit des objectifs pratiques pour l'apprentissage offline sécurisé.

Enfin, la thèse présente plusieurs contributions additionnelles liées aux thèmes principaux. Dans le cadre en ligne, les idées de structuration bayésienne sont étendues au problème de \emph{best-arm identification} à budget fixé. L'algorithme \texttt{PI-BAI} utilise l'information a priori pour répartir efficacement le budget d'exploration dans des bandits structurés. L'analyse fournit des garanties bayésiennes dépendant du prior sur la probabilité d'erreur, et montre que des allocations non adaptatives bien informées peuvent surpasser des stratégies adaptatives classiques. Dans le cadre hors politique, la thèse contribue également au développement du \emph{logarithmic smoothing}, un estimateur pessimiste de la forme
\[
\hat V_{\mathrm{LS}}^\lambda(\pi)=
\frac1{n\lambda}\sum_{i=1}^n
\log\left(1+\lambda
\frac{\pi(A_i\mid X_i)}{\pi_0(A_i\mid X_i)}R_i\right).
\]
Cet estimateur agit comme une alternative douce et différentiable au clipping, bénéficie de garanties de concentration serrées, et permet d'obtenir des bornes de sous-optimalité plus fines que les approches précédentes. Enfin, cette thèse CIFRE maintient un lien constant avec les systèmes de recommandation industriels à grande échelle. Les applications développées autour de la recommandation optimisant la récompense, de l'évaluation offline, de la simulation contrefactuelle et des modèles de recommandation par ardoise ont fourni à la fois un terrain expérimental et une source de questions théoriques.

Dans son ensemble, cette thèse défend une idée centrale : pour apprendre efficacement dans de grands espaces d'actions, il ne suffit pas d'appliquer directement les algorithmes classiques de bandits contextuels ou d'apprentissage hors politique. Il faut exploiter la structure entre actions, contrôler explicitement l'incertitude et la couverture des données, et concevoir des objectifs dont le paysage d'optimisation reste favorable à grande échelle. Les contributions proposées répondent à ces exigences selon deux axes complémentaires. En apprentissage en ligne, des modèles bayésiens hiérarchiques et diffusionnels permettent de partager l'information entre actions et de réduire le regret. En apprentissage hors politique, des méthodes directes structurées, des objectifs de vraisemblance pondérée et des principes de pessimisme différentiable rendent l'apprentissage statistiquement robuste et computationnellement réalisable. Ces résultats contribuent à rapprocher la théorie des bandits contextuels des contraintes réelles des systèmes interactifs modernes, où les décisions doivent être prises parmi des catalogues massifs, à partir de signaux partiels, bruités et parfois fortement biaisés.

%% file: contents/overview/context.tex
\section{Context and Scope}

Interactive machine learning systems are a cornerstone of modern technology, optimizing decision-making in applications ranging from recommender systems and financial markets to robotics. These systems operate in a sequential loop: they process contextual information, select an action from a wide range of possibilities, and receive feedback that depends on both the context and the chosen action. A fundamental challenge in designing these systems is the scale of the decision space; in many real-world settings, the number of potential actions can be huge.

Consider \emph{recommender systems}, where streaming platforms or e-commerce sites must select an item to present to a user. The \emph{context} comprises rich data, such as user preferences and history. The \emph{action} is the selection of a specific item from a catalog containing thousands or millions of options. The system's objective is to learn a policy that maximizes user engagement (the \emph{reward}), measured by metrics such as watch time or clicks.

Similarly, in \emph{computational advertising}, a platform selects which ad to display via real-time bidding. The context includes user attributes and page details. The action is composite: selecting an ad from a large inventory and determining a bid price. The feedback (reward) arrives in stages, from winning the auction to subsequent user clicks or conversions. The system must maximize advertiser value while adhering to budget constraints.

We model these interactive systems using the \emph{contextual bandit} framework. This framework captures the essential characteristics of interactive learning while maintaining the tractability required for theoretical analysis and practical implementation.

\subsection{Contextual Bandits}

\begin{figure}
  \centering
  \begin{subfigure}[b]{0.48\textwidth}
    \centering
    \includegraphics[width=\linewidth]{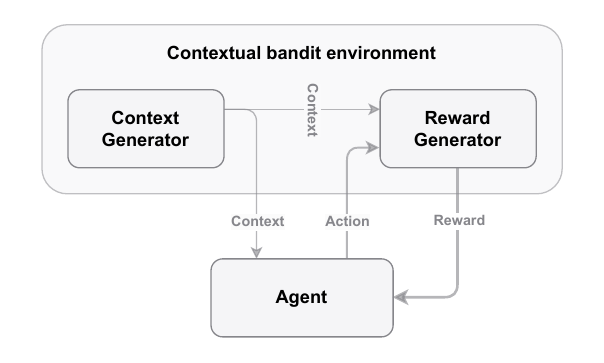}
    \caption{Single interaction loop.}
    \label{fig:cb-diagram}
  \end{subfigure}
  \hfill
  \begin{subfigure}[b]{0.48\textwidth}
    \centering
    \includegraphics[width=\linewidth]{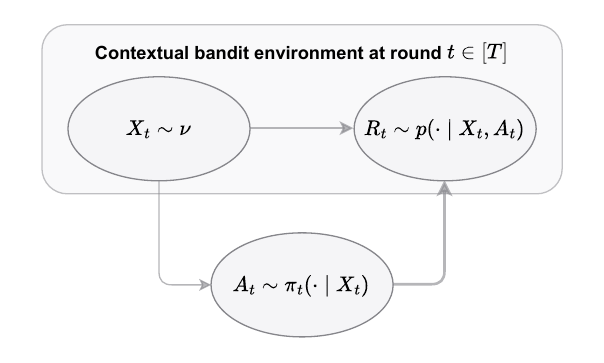}
    \caption{Graphical representation in round $t \in [T]$.}
    \label{fig:cb-graphical_representation}
  \end{subfigure}
  \caption{Contextual bandit framework.}
  \label{fig:contextual-bandit}
\end{figure}

\cref{fig:cb-diagram} visualizes the interaction loop. The environment consists of a \emph{Context Generator} and a \emph{Reward Generator}, both of which are \emph{assumed to be fixed but unknown to the agent}. At each round, the environment emits a context. The agent observes this context and selects an action from a finite set\footnote{The action space can technically be infinite, but this thesis focuses on large finite action spaces.}. Finally, the environment generates a scalar reward based on the context-action pair. By repeating this loop, the agent accumulates experience to refine its policy.

Formally, let $\mathcal{X} \subset \mathbb{R}^{d}$ denote the context space and $\mathcal{A} = [K]$ the finite action set. A \emph{stochastic policy} $\pi : \mathcal{X} \to \Delta(\mathcal{A})$ maps each context $x \in \mathcal{X}$ to a probability distribution $\pi(\cdot \mid x)$ over actions. The environment is specified by:
\begin{itemize}
  \item A context distribution $\nu$ over $\mathcal{X}$;
  \item A family of conditional reward distributions $\{p(\cdot \mid x,a)\}_{(x,a) \in \mathcal{X} \times \mathcal{A}}$.
\end{itemize}
The expected reward function for any context-action pair $(x,a)$ is defined as:
\begin{align}\label{eq:reward_function}
  r(x,a) = \mathbb{E}_{R \sim p(\cdot \mid x,a)}[R].
\end{align}

The interaction unfolds over $T$ rounds. In each round $t \in [T]$:
\begin{enumerate}
  \item The environment draws a context $X_t \sim \nu$ and reveals it to the agent.
  \item The agent selects an action $A_t \sim \pi_t(\cdot \mid X_t)$ according to its current policy $\pi_t$.
  \item The environment samples a reward $R_t \sim p(\cdot \mid X_t, A_t)$ and returns it to the agent.
\end{enumerate}
A graphical representation of this interaction in round $t \in [T]$ is visualized in \cref{fig:cb-graphical_representation}.

\subsection{Learning Paradigms}\label{subsec:learning_paradigms}
We address learning in this framework via two complementary paradigms: \emph{on-policy} (online) and \emph{off-policy} (offline). With a slight abuse of terminology, we use \emph{learning} loosely to encompass the full agent behavior, including both reward estimation and action selection.

\begin{description}
    \item[On-Policy (Online) Learning] \mbox{}\par 
In this setting, the agent \emph{updates} its policy $\pi_t$ sequentially. Let $H_t=\{(X_\ell,A_\ell,R_\ell)\}_{\ell<t}$ denote the history available at the start of round $t$. The agent uses $H_t$ to construct the policy $\pi_t$. Following the interaction, the agent augments the history with the new observation $(X_t, A_t, R_t)$ to form $H_{t+1}$ and repeats the process.

Performance is measured by the \emph{cumulative regret}:
\begin{align}\label{eq:regret}
  \mathcal{R}(T) = \sum_{t=1}^{T} \left( r(X_t,A_{t,*}) - r(X_t,A_t) \right),
\end{align}
where $A_{t,*} = \arg\max_{a\in\mathcal{A}} r(X_t,a)$ is the optimal action in round $t$. While minimizing regret is equivalent to maximizing cumulative reward, the literature prioritizes regret as it normalizes performance against the optimal oracle, facilitating theoretical comparisons across environments.

The agent faces the \emph{exploration-exploitation dilemma}: it must balance exploration of poorly understood actions with exploitation of actions believed to yield high rewards. Feedback is partial (only the reward for the chosen action is observed) and noisy, and exploration may be constrained by safety or budget requirements.

\vspace{0.5em}
\begin{remark}[Beyond regret minimization]
    While this thesis focuses on regret minimization, other objectives exist, most notably \emph{Best-Arm Identification (BAI)}. \emph{BAI} aims to identify the optimal action rather than maximize cumulative reward. This is important for applications like A/B testing and clinical trials. Although we focus on regret, our core modeling contributions for scaling Thompson sampling extend to the BAI setting, as demonstrated in our related work \citep{nguyen2025prior} (not included in this manuscript).
\end{remark}\vspace{0.5em}

    \item[Off-Policy (Offline) Learning] \mbox{}\par 
In this setting, the agent learns a policy $\hat{\pi}$ from a static logged dataset $\mathcal{D}_n = \{(X_i, A_i, R_i)\}_{i=1}^{n}$ collected by a logging policy $\pi_0$ as $X_i \sim \nu$, $A_i \sim \pi_0(\cdot \mid X_i)$ and $R_i \sim p(\cdot \mid X_i, A_i)$. No additional interactions with the environment are allowed. The objective is to find a policy maximizing the expected value:
\begin{align}
    V(\pi) = \mathbb{E}_{X \sim \nu} \mathbb{E}_{A \sim \pi(\cdot \mid X)}[r(X, A)].
    \label{eq:value}
\end{align}
Performance is measured by the \emph{suboptimality gap} of the learned policy $\hat{\pi}$:
\begin{align}\label{eq:suboptimality}
   \textsc{so}(\hat{\pi}) = V(\pi_*) - V(\hat{\pi}),
\end{align}
where $\pi_* = \argmax_{\pi \in \Pi} V(\pi)$ is the \emph{unknown} optimal policy in a class of policies $\Pi$.

Since the agent learns solely from data generated by $\pi_0$, it must perform \emph{counterfactual reasoning}. This introduces several challenges. First, \emph{support mismatch} occurs when specific actions are rarely or never selected by $\pi_0$ in certain regions of the context space; consequently, the dataset provides little to no information about the rewards in these regions. Second, \emph{re-weighting instability} arises since standard techniques, such as \emph{inverse propensity scoring} (importance sampling), re-weight logged samples using the density ratio $\pi(a \mid x) / \pi_0(a \mid x)$. When $\pi_0(a \mid x)$ is small, this ratio can explode, leading to high-variance estimates. Third, \emph{high bias} can affect methods that rely on parametric models to estimate the reward. Extrapolating rewards for unobserved context-action pairs introduces errors if the model is misspecified.

\end{description}

The difficulties in both paradigms are amplified when the action space is large ($K$ in the thousands or millions). In on-policy settings, independent exploration of every action becomes infeasible, yielding high regret. In off-policy settings, data sparsity worsens: reward models risk significant extrapolation error, and importance weights suffer from extreme variance as the probability of observing any specific action vanishes. Developing methods that scale gracefully (both statistically and computationally) with the size of the action space is the central theme of this thesis.

%% file: contents/overview/background.tex
\section{Background}\label{sec:background}

Most on-policy and off-policy learning algorithms\footnote{Recall that we use \emph{learning} loosely to encompass both reward estimation and action selection (\cref{subsec:learning_paradigms}).} are fundamentally constructed from two core components: \emph{reward estimation}, which approximates the expected reward function $r(x, a)$ for any context-action pair from data, and \emph{decision-making}, which leverages these estimates and their associated uncertainty to select actions.

\subsection{Reward Estimation}

The central task in reward estimation is to learn an approximate function $\hat{r}(x, a)$ that estimates the true expected reward $r(x, a)$ defined in \cref{eq:reward_function}. This function is trained on a dataset, denoted by $\textsc{Data}$, whose structure depends on the learning paradigm:

\textbf{On-policy data.} The agent collects data sequentially. The dataset at round $t$ is the history of interaction up to round $t$:
\begin{align}\label{eq:data_online}
    \textsc{Data} = H_t = \{(X_i, A_i, R_i)\}_{i=1}^{t-1}.
\end{align}

\textbf{Off-policy data.} The agent learns from a static dataset logged by a logging policy $\pi_0$:
\begin{align}\label{eq:data_offline}
    \textsc{Data} = \mathcal{D}_n = \{(X_i, A_i, R_i)\}_{i=1}^n, \quad \text{with } A_i \sim \pi_0(\cdot \mid X_i).
\end{align}

We denote the learned model as $\hat{r}(x, a) = r(x, a; \hat{\theta})$, where $\hat{\theta}$ are parameters obtained via a statistical objective. Common approaches include:

\begin{description}
    \item[Maximum Likelihood Estimation (MLE)]\mbox{}\par 
    MLE seeks parameters $\theta$ that maximize the probability of observing the collected rewards. The objective is:
    $$
    \hat{\theta}_{\text{MLE}} = \arg\max_{\theta} \sum_{(X_i, A_i, R_i) \in \textsc{Data}} \log p(R_i \mid X_i, A_i; \theta).
    $$
    For Gaussian rewards $p(\cdot \mid x, a; \theta) = \mathcal{N}(r(x, a; \theta), \sigma^2)$, this simplifies to minimizing the sum of squared errors (\emph{ordinary least squares}). For Bernoulli rewards, it corresponds to minimizing binary cross-entropy (e.g., logistic regression).

    \item[Maximum A Posteriori (MAP)]\mbox{}\par 
    MAP estimation incorporates a prior distribution $p_0(\theta)$ to regularize the objective:
    $$
    \hat{\theta}_{\text{MAP}} = \arg\max_{\theta} \left( \sum_{(X_i, A_i, R_i) \in \textsc{Data}} \log p(R_i \mid X_i, A_i; \theta) + \log p_0(\theta) \right).
    $$
    For example, combining a Gaussian likelihood with a zero-mean Gaussian prior $p_0(\cdot) = \mathcal{N}(0, \lambda I_d)$ is equivalent to \emph{ridge regression} ($\ell_2$ regularization).

    \item[Full Bayesian Inference]\mbox{}\par 
    Rather than a point estimate, Bayesian inference characterizes uncertainty by computing the full posterior distribution $p(\theta \mid \textsc{Data})$:
    $$
    p(\theta \mid \textsc{Data}) \propto p(\theta) \prod_{(X_i, A_i, R_i) \in \textsc{Data}} p(R_i \mid X_i, A_i; \theta).
    $$
    This approach is powerful for guiding exploration. A common implementation assumes conjugate Gaussian distributions: $p_0(\cdot) = \mathcal{N}(\mu_0, \Sigma_0)$ and $p(\cdot \mid x, a; \theta) = \mathcal{N}(r(x, a; \theta), \sigma^2)$. When $r(x, a; \theta)$ is linear in $\theta$ (\emph{Bayesian linear regression}), the posterior is Gaussian and analytically tractable. For non-linear models, posterior approximation methods are required.
\end{description}

The functional form of the reward model, $r(x, a; \theta)$, is an important design choice that dictates the balance between computational tractability, data efficiency, and expressive power. While numerous function classes exist, linear models remain a cornerstone of the field due to their simplicity and strong theoretical guarantees. Even within this linear family, there is an important distinction between \emph{joint} and \emph{disjoint} formulations.

The \emph{joint linear model} defines $r(x, a; \theta) = \phi(x,a)^\top \theta$, sharing a single parameter $\theta \in \mathbb{R}^d$ across all actions. While data-efficient, this approach relies on designing a feature map $\phi(x,a)$ capable of capturing complex context-action interactions. Designing $\phi(x, a)$ is very hard in practice, and inadequate feature engineering leads to poor performance. This is why in practice, it is often more common to adopt a \emph{disjoint linear model} that learns an independent parameter $\theta_a \in \mathbb{R}^{d}$ for each action, yielding $r(x, a; \theta) = \phi(x)^\top \theta_a$. This is common in recommender systems, where predictions are inner products of user and item embeddings. This formulation is robust and avoids complex feature engineering. However, its primary drawback is \emph{poor statistical scalability}: while the computational overhead of maintaining $K$ independent embeddings can be addressed with sufficient compute, the statistical challenge remains. This is because learning each embedding independently requires substantial data per action, which becomes prohibitive as $K$ grows. 

Other non-linear models exist for capturing complex reward functions. Generalized linear models (GLMs), for instance, employ a link function to model non-Gaussian rewards (e.g., logistic regression for binary outcomes: $p(R=1\mid x,a) = \sigma(\phi(x)^\top\theta_a)$). These models align closely with the linear setting, and the algorithms proposed in this thesis are suitable for them. A significant portion of this thesis focuses on scaling these disjoint reward models to large action spaces.

\begin{remark}[Scope of parametric reward modeling]
    The reward estimation framework presented above assumes parametric reward models $r(x, a; \theta)$. This assumption underlies \cref{chap:meTS,chap:dTS,chap:sDM}, where we develop structured parametric models that share information across actions to improve statistical efficiency. The remaining two chapters (\cref{chap:las,chap:ES}) take a slightly different approach: rather than explicitly estimating rewards, they optimize policies directly using inverse propensity scoring and policy-weighted objectives, thereby making them agnostic to the choice of reward model.
\end{remark}

\subsection{Decision-Making}

We now turn to the second component: decision-making. This stage (often) relies on the reward model $\hat{r}$ derived using the estimation techniques discussed previously. While the difference in reward estimation between on-policy and off-policy settings is primarily driven by how data is accumulated, the principles guiding action selection in these two paradigms are fundamentally distinct.

\subsubsection*{Decision-Making in On-Policy Learning}
Recall that in the on-policy setting, the agent must balance exploration and exploitation to minimize regret. The two dominant paradigms are \emph{Upper Confidence Bound (UCB)} and \emph{Thompson Sampling (TS)}.

\begin{description}
    \item[Upper Confidence Bound (UCB)]\mbox{}\par 
    UCB drives exploration via a bonus term added to the reward estimate, selecting:
    \begin{align*}
    A_t = \arg\max_{a \in \mathcal{A}} \left( \hat{r}(X_t,a) + \text{bonus}_t(X_t,a) \right).
    \end{align*}
    For linear models (e.g., \texttt{LinUCB}), the bonus scales with $\sqrt{\phi(x)^\top V_{t, a}^{-1}\phi(x)}$, where $V_{t, a}$ is the design matrix for action $a$, encouraging the selection of less-certain actions.

    \item[Thompson Sampling (TS)]\mbox{}\par
    TS (or posterior sampling) implements randomized exploration. The agent samples a reward function $\tilde{r}_t$ from the posterior and acts greedily with respect to it:
    \begin{align*}
    A_t = \arg\max_{a \in \mathcal{A}} \tilde{r}_t(X_t, a), \quad \text{where } \tilde{r}_t(x, a) = r(x, a; \theta_t), \; \theta_t \sim p(\theta \mid H_t).
    \end{align*}
    Exploration is implicit: high posterior uncertainty yields diverse samples $\theta_t$, leading to varied actions. As data accumulates, the posterior contracts, and behavior naturally becomes exploitative\footnote{While we describe sampling parameters $\theta_t$, the general principle involves sampling from the posterior predictive distribution of rewards.}.
\end{description}
In large action spaces, standard UCB and TS struggle. Treating actions independently gathers information too slowly, leading to prohibitive regret. This necessitates structured models that share information across actions.

\subsubsection*{Decision-Making in Off-Policy Learning}

Recall that the off-policy setting requires identifying an optimal policy from a static dataset collected under a distinct logging policy. The two dominant philosophies for tackling this are \emph{Greedy Policies} and \emph{Pessimistic Policies}.

\begin{description}
\item[Greedy Policies]\mbox{}\par
Greedy methods select the policy $\hat{\pi}_{\textsc{g}}$ that maximizes a point estimate of value, $\hat{V}(\pi)$:
\begin{align}
    \hat{\pi}_{\textsc{g}} = \arg\max_{\pi \in \Pi} \hat{V}(\pi).
    \label{eq:greedy_policy}
\end{align}
We distinguish between two primary classes of estimators. The \emph{direct method (DM)} relies on the reward model $\hat{r}$ derived in the previous section. In contrast, \emph{inverse propensity scoring (IPS)} bypasses reward modeling to estimate the policy value directly using importance weighting\footnote{Importance weighting allows IPS to be unbiased under the \emph{common support} assumption (i.e., $\pi_0(a|x)=0 \implies \pi(a|x)=0$)}:
\begin{align*}
    \hat{V}_{\textsc{dm}}(\pi) &= \frac{1}{n} \sum_{i=1}^n \sum_{a \in \mathcal{A}} \pi(a \mid X_i) \, \hat{r}(X_i, a), \\
    \hat{V}_{\textsc{ips}}(\pi) &= \frac{1}{n} \sum_{i=1}^n \frac{\pi(A_i \mid X_i)}{\pi_0(A_i \mid X_i)} \, R_i.
\end{align*}
The optimization procedures for these objectives differ significantly. The policy maximizing $\hat{V}_{\textsc{dm}}(\pi)$ is simply the one that acts greedily with respect to the learned reward model, 
\begin{align}\label{eq:greedy_dm}
    \hat{\pi}_{\textsc{g}}^{\textsc{dm}}(x) = \arg\max_{a \in \mathcal{A}} \hat{r}(x, a)\,.
\end{align}
For IPS, the optimization is typically performed numerically over a class of parameterized policies $\pi_\theta$:
\begin{align}\label{eq:greedy_ips}
    \hat{\pi}_{\textsc{g}}^{\textsc{ips}} = \arg\max_{\theta \in \real^d}  \hat{V}_{\textsc{ips}}(\pi_\theta)\,.
\end{align}

Greedy policies are effective when the underlying estimator is accurate. Specifically, when the reward model $\hat{r}$ is well-specified (for DM), or the importance weights have low variance (for IPS). However, the $\arg\max$ operator can amplify estimation errors, leading to a policy that over-exploits optimistic model inaccuracies or high-variance weight estimates.

\item[Pessimistic Policies]\mbox{}\par
Pessimistic methods mitigate error amplification by penalizing the objective with a quantified uncertainty term:
\begin{align}
    \hat{\pi}_{\textsc{p}} = \arg\max_{\pi \in \Pi} \left[ \hat{V}(\pi) - \operatorname{pen}(\pi) \right].
    \label{eq:pess_policy_template}
\end{align}
This principle applies to both DM and IPS as:
\begin{align}\label{eq:dm_pess}
    &\texttt{Pess-DM:}   & \hat{\pi}_{\textsc{p}}^{\textsc{dm}}(x) = \arg\max_{a \in \mathcal{A}} \Bigl[ \hat{r}(x, a) - \beta \, \hat{\sigma}_r(x, a) \Bigr]\,,
\end{align}
and 
\begin{align}\label{eq:ips_pess}
    &\texttt{Pess-IPS:}  & \hat{\pi}_{\textsc{p}}^{\textsc{ips}} = \arg\max_{\theta} \Bigl[\hat{V}_{\textsc{ips}}(\pi_\theta) - \beta \, \hat{\sigma}_{\textsc{ips}}(\pi_\theta)\Bigr].
\end{align}

Here, $\hat{\sigma}_r$ captures the uncertainty of the reward model, while $\hat{\sigma}_{\textsc{ips}}$ captures the uncertainty of the IPS estimator itself (e.g., its variance). The penalty term $\operatorname{pen}(\pi)$ prevents the maximization operator from selecting overestimated policies. It also regulates \emph{distributional shift} by penalizing policies that place probability mass on context-action pairs with low coverage under $\pi_0$ (support mismatch).
\end{description}

Standard methods, whether relying on DM or IPS, and whether adopting greedy or pessimistic policies, face severe limitations in large action spaces. For DM, the prevailing practice of modeling action parameters independently prevents information sharing, making learning statistically inefficient. For IPS, the well-recognized issue is \emph{variance}: importance weights can be large. However, we also demonstrate in this thesis that \emph{optimization} can be an even greater bottleneck in large action spaces. This is because standard IPS-based objectives (whether \cref{eq:greedy_ips} or \cref{eq:ips_pess}) induce highly non-concave landscapes with flat plateaus that trap gradient-based optimizers. Finally, for pessimistic methods, existing formulations often rely on intractable bounds that are incompatible with modern stochastic optimization techniques. This thesis addresses these specific pathologies: we introduce structured models for DM to enforce information sharing; we propose new policy-weighted log-likelihood objectives that yield superior optimization landscapes compared with IPS-based objectives; and we develop variance-reduced, tractable pessimistic objectives that are amenable to stochastic optimization at scale.

%% file: contents/overview/contributions.tex
\section{Contributions}\label{sec:contributions}
\subsection*{Part I: On-Policy Learning in Large Action Spaces}

In the \textit{on-policy setting}, exploration strategies that adopt disjoint reward models\footnote{Recall that disjoint models offer greater flexibility and are widely used in industrial settings such as large-scale recommender systems that use a separate embedding for each item, whereas joint reward models require careful feature engineering and are not, to the best of our knowledge, widely deployed in practical recommendation systems.} and learn each action parameter independently gather information slowly, resulting in high regret and failure to converge to an optimal policy within practical time horizons. \cref{part:online} addresses this challenge by scaling Thompson Sampling to large action spaces while retaining the disjoint reward model parameterization. Our primary contribution is the introduction of structured Bayesian models with informative priors that share statistical strength across actions, enabling efficient exploration without sacrificing the robustness and flexibility of disjoint reward models.

\begin{description}
\item[(\cref{chap:meTS}) Scaling Thompson Sampling with Mixed-Effects] \mbox{}\par
We propose a hierarchical Bayesian framework that couples action parameters through $L$ shared latent effect parameters $\Psi = (\psi_\ell)_{\ell \in [L]} \in \mathbb{R}^{dL}$, where each effect $\psi_\ell$ can represent, for example, a category of items:
\begin{align*}
      \Psi &\sim q_0\,, \\
      \theta_a \mid \Psi &\sim p_{0,a}(\cdot \mid \Psi)\,, & \forall a \in [K]\,, \\
      R_t \mid \theta, \Psi, X_t, A_t &\sim p(\cdot \mid X_t; \theta_{A_t})\,, & \forall t \in [T]\,.
    \end{align*}
Upon this model, we build \emph{mixed-effect Thompson sampling} (\texttt{meTS}). \texttt{meTS} maintains a posterior over effects $q_t(\Psi) = p(\Psi \mid H_t)$ and $K$ conditional posteriors over actions $p_{t,a}(\theta_a \mid \Psi) = p(\theta_a \mid \Psi, H_t)$. In round $t \in [T]$, parameters are sampled hierarchically:
    \begin{align*}
        \Psi_t &\sim q_t(\cdot)\,, \\ \theta_{t,a} &\sim p_{t,a}(\cdot \mid \Psi_t)\,, & \forall a \in [K]\,.
    \end{align*}
Actions are then selected via the standard TS rule: $A_t = \arg\max_{a \in [K]} r(X_t; \theta_{t, a})$. We derive exact closed-form updates for linear models and tractable Laplace approximations for generalized linear models.
    
Theoretically, we establish a Bayesian regret bound in the linear-Gaussian case:
    \begin{align*}
    \mathcal{BR}(T) = \tilde{\mathcal{O}}\left(\sqrt{T d K_{\text{eff}} (\sigma_0^2 + \sigma_\Psi^2)}\right)\,,
    \end{align*}
where $\sigma_0^2$ and $\sigma_\Psi^2$ are the prior variances of the action and effect parameters, respectively, and $K_{\mathrm{eff}}$ is the \emph{effective number of actions}. When $L \ll K$, we have $K_{\mathrm{eff}} \ll K$, yielding a multiplicative Bayesian regret improvement of $\sqrt{K / K_{\mathrm{eff}}}$ over standard TS. Computationally, \texttt{meTS} exploits the conditional independence of action parameters given the latent effects, reducing memory complexity from $\mathcal{O}(K^2 d^2)$ to $\mathcal{O}((L^2 + K) d^2)$ and runtime from $\mathcal{O}(K^3 d^3)$ to $\mathcal{O}((L^3 + K) d^3)$. Empirically, \texttt{meTS} consistently outperforms baselines, with gains increasing with $K$.
    
\textbf{AISTATS 2023 (Poster) - \citet{aouali2022mixed}:} 
    \begin{itemize}
    \item I. Aouali, B. Kveton, and S. Katariya. Mixed-effect Thompson sampling. In \textit{International
Conference on Artificial Intelligence and Statistics}, pages 2087–2115. PMLR, 2023b.
    \end{itemize}

\item[(\cref{chap:dTS}) Scaling Thompson Sampling with Diffusion Models] \mbox{}\par
This chapter extends the hierarchical framework of \texttt{meTS} by introducing \emph{diffusion Thompson Sampling} (\texttt{dTS}), which replaces the single-layer prior with a deep hierarchy of latent variables governed by a \emph{diffusion model}:
\begin{align*}
    \psi_L &\sim \mathcal{N}(0, \Sigma_{L+1})\,, \\
    \psi_{\ell-1} \mid \psi_\ell &\sim \mathcal{N}(f_\ell(\psi_\ell), \Sigma_\ell)\,, & \forall \ell \in [L] \setminus \{1\}\,, \\
    \theta_a \mid \psi_1 &\sim \mathcal{N}(f_1(\psi_1), \Sigma_1)\,, & \forall a \in [K]\,, \\
    R_t \mid \theta, (\psi_\ell)_{\ell \in [L]}, X_t, A_t &\sim p(\cdot \mid X_t; \theta_{A_t})\,, & \forall t \in [T]\,.
\end{align*}
The link functions $f_\ell$ are pre-trained non-linear transformations (e.g., neural networks), enabling rich representations of inter-action structure. As in \texttt{meTS}, exploration proceeds by sampling parameters top-down through the hierarchy and selecting the reward-maximizing action.

The key technical challenge is that the exact posterior is intractable due to non-linearities in both the reward and link functions. To enable fast online updates without expensive MCMC, we derive a tractable inference procedure based on Gaussian-like updates\footnote{By Gaussian-like updates, we mean that the posterior precision is the sum of prior and evidence precisions, and the posterior mean is the precision-weighted combination of prior mean and maximum likelihood estimate.}. The resulting posterior preserves the diffusion structure: it remains a hierarchy of conditional Gaussians, but with \emph{fine-tuned link functions and precisions}:
\begin{align*}
    \bar{\Sigma}_{t,\ell-1}^{-1} &=
    \underbrace{\Sigma_\ell^{-1}}_{\text{prior precision}}
    +
    \underbrace{\bar{G}_{t,\ell-1}}_{\text{data precision}},
    \\
    \hat{f}_{t,\ell}(\psi_\ell) &=
    \bar{\Sigma}_{t,\ell-1}\!\Big(
        \underbrace{\Sigma_\ell^{-1} f_\ell(\psi_\ell)}_{\text{prior contribution}}
        +
        \underbrace{\bar{B}_{t,\ell-1}}_{\text{data contribution}}
    \Big)\,.
\end{align*}
Here, $\bar{G}_{t,\ell-1}$ and $\bar{B}_{t,\ell-1}$ are sufficient statistics propagated upward through the hierarchy. As data accumulates, covariances contract and means shift from prior toward MLE. Crucially, this formulation preserves the expressiveness of diffusion models since the posterior is not a single Gaussian but a \emph{posterior diffusion model} that retains the generative structure of the prior.

Theoretically, we analyze \texttt{dTS} in the fully linear-Gaussian setting to gain analytical insight, deriving a Bayes regret bound:
$$\mathcal{BR}(T) = \tilde{\mathcal{O}}\left(\sqrt{T d K_{\text{eff}} \sum_{\ell=1}^{L+1} \sigma_{\ell}^2 }\right),$$
where $\Sigma_\ell = \sigma_\ell^2 I_d$ and $K_{\mathrm{eff}} \ll K$ is the effective number of actions. Computationally, \texttt{dTS} exploits hierarchical conditional independence to reduce memory and time complexity further from $\mathcal{O}(K^2 d^2)$ and $\mathcal{O}(K^3 d^3)$ to $\mathcal{O}((L + K) d^2)$ and $\mathcal{O}((L + K) d^3)$, respectively: linear scaling in $L$ that improves upon \texttt{meTS}. Empirically, \texttt{dTS} consistently outperforms baselines by leveraging pre-trained diffusion priors, even when these priors are imperfect or trained on limited data.

\textbf{NeurIPS 2025 (Poster) -
 \citep{aouali2025diffusion,aouali2023linear}:}
\begin{itemize}
    \item I. Aouali. Diffusion models meet contextual bandits. In \textit{The Thirty-ninth Annual Conference on Neural Information Processing Systems}, 2025.
    \item I. Aouali. Linear diffusion models meet contextual bandits with large action spaces. In
\textit{NeurIPS 2023 Foundation Models for Decision Making Workshop}, 2023.
\end{itemize}

\end{description}

\subsection*{Part II: Off-Policy Learning in Large Action Spaces}

In the \textit{off-policy setting}, both DM and IPS face severe limitations as the action space grows. DM suffers from high model bias and statistical inefficiency due to sparse data coverage. IPS exhibits high variance and potential bias due to insufficient support; moreover, as we demonstrate in this thesis, optimizing IPS-based objectives becomes intractable in large action spaces. Part II addresses these failure modes through three complementary approaches.

\begin{description}
\item[(\cref{chap:sDM}) Scaling Direct Methods with Latent Parameters] \mbox{}\par

Standard DMs estimate independent $d$-dimensional parameters for each action, which becomes statistically inefficient when actions are rarely observed. We introduce the \emph{structured direct method} (\texttt{sDM}), which couples action parameters through a shared latent vector $\psi$ (analogous to \cref{chap:meTS}):
\begin{align*}
  \psi &\sim q\,, \\
  \theta_a \mid \psi &\sim p_a(\cdot; f_a(\psi)), \\
  R \mid X,A,\theta, \psi &\sim p(\cdot \mid X;\theta_A).
\end{align*}

\texttt{sDM} computes the posterior over latent effects $p(\psi \mid \mathcal{D}_n)$ and conditional posteriors $p(\theta_a \mid \psi, \mathcal{D}_n)$. The marginal posterior $p(\theta_a \mid \mathcal{D}_n)$ is obtained by integrating out $\psi$, yielding the reward estimate $\hat{r}(x, a) = \mathbb{E}[r(x, a; \theta) \mid \mathcal{D}_n]$, which is used in a greedy policy as:
$$  \hat{\pi}_{\textsc{g}}(a \mid x) = \mathds{1}{\{a = \arg\max_{b \in \mathcal{A}} \hat{r}(x, b)\}}\,.$$
To analyze performance, we introduce \emph{Bayesian suboptimality} (BSO) and prove that \texttt{sDM} achieves $\mathcal{O}(1/\sqrt{n})$ convergence. The result avoids the restrictive full support assumption, which requires $\pi_0(a \mid x) \geq \gamma > 0$ for all actions. Instead, the bound depends on the alignment between the logging policy $\pi_0$ and the optimal policy $\pi_*$: performance improves smoothly as coverage of optimal actions increases. We also prove that under BSO, greedy policies are optimal and outperform pessimistic ones. This contrasts with the frequentist setting, where pessimism hedges against worst-case scenarios and is generally preferred. Experiments on synthetic and real-world data confirm that \texttt{sDM} outperforms existing methods, with gains increasing with $K$.

\textbf{AISTATS 2025 (Poster) - \citet{aouali2024bayesian}:} 
\begin{itemize}
    \item I. Aouali, V.-E. Brunel, D. Rohde, and A. Korba. Bayesian off-policy evaluation and
learning for large action spaces. In \textit{International Conference on Artificial Intelligence
and Statistics}, pages 136–144. PMLR, 2025.
\end{itemize}

\item[(\cref{chap:las}) Optimization Matters More Than Estimation] \mbox{}\par
This chapter challenges a common paradigm in off-policy learning. The field has traditionally focused on developing sophisticated value estimators with improved statistical properties, assuming that maximizing a more accurate estimator yields a better policy. We demonstrate that this emphasis is misplaced in large action spaces, where \emph{optimization error} dominates estimation error, making even advanced estimators ineffective for policy learning.

Our key insight is that estimator-based objectives, despite their statistical appeal, induce highly non-concave landscapes when paired with standard policy classes. We show that gradient-based optimization can remain trapped in suboptimal plateaus for $\mathcal{O}(K)$ iterations, and that the landscape contains exponentially many local maxima in $K$. These pathologies make global optimization intractable for large $K$.

To characterize these failures, we analyze the \emph{oracle policies} of various estimators: the policies that maximize the estimators with infinite data. This analysis reveals that each estimator induces a distinct inductive bias. For instance, standard IPS searches within the logging policy's support, while cluster-based methods such as \texttt{MIPS} \citep{saito2022off} operate at the cluster level. These insights motivate \emph{objective-aware policy parametrizations}: by aligning the policy parametrization with the estimator's bias, we reduce the effective search space from $K$ to the significantly smaller logging support size $k_0$ or cluster count $C$, partially alleviating the optimization challenges.

Ultimately, we advocate for a fundamental shift: abandoning value estimation in favor of \emph{policy-weighted log-likelihood (PWLL)} objectives:
\begin{align*}
    \hat{U}_g(\pi) = \frac{1}{n} \sum_{i=1}^n g(R_i, \pi_0(A_i \mid X_i)) \log \pi(A_i \mid X_i)\,,
\end{align*}
where $g$ is a positive weighting function. Although PWLL objectives are not value estimators, we prove they are concave (and strongly concave with $\ell_2$ regularization) for linear softmax policies, while achieving oracle policies comparable to estimator-based objectives. This guarantees efficient convergence to a unique global maximum, while eliminating the optimization pathologies.

Large-scale experiments on datasets with up to one million actions validate this approach. Simple PWLL methods consistently outperform state-of-the-art estimator-based objectives, with the performance gap widening as action spaces grow. Moreover, PWLL objectives exhibit remarkable robustness to optimization hyperparameters, whereas estimator-based methods require careful tuning and often fail under minor configuration changes.

\textbf{CONSEQUENCES, RecSys 2025 (Poster) - Submitted to ICLR 2026 - \citep{aouali2025off}:}
\begin{itemize}
    \item I. Aouali and O. Sakhi. Off-policy learning in large action spaces: Optimization matters
more than estimation. \textit{Under review at ICLR}, 2026.
\end{itemize}

\item[(\cref{chap:ES}) Principled Pessimism for Exponential Smoothing and Beyond] \mbox{}\par

While \texttt{sDM} and PWLL offer alternative paradigms, this chapter improves the widely used family of IPS-based methods by combining variance-reducing regularization with principled pessimism for safe policy learning.

The variance of IPS scales with importance weights, which can explode in large action spaces. To control this, we introduce differentiable \emph{exponential smoothing (ES)} estimators that regularize these weights:
\begin{align*}
    \texttt{IPS-}\alpha:  \quad \hat{V}^\alpha(\pi) &= \frac{1}{n} \sum_{i=1}^n \frac{\pi(A_i \mid X_i)}{\pi_0(A_i \mid X_i)^\alpha} R_i\,, \\
    \texttt{IPS-}\beta:  \quad \tilde{V}^\beta(\pi) &= \frac{1}{n} \sum_{i=1}^n \left(\frac{\pi(A_i \mid X_i)}{\pi_0(A_i \mid X_i)}\right)^\beta R_i\,.
\end{align*}
These estimators smoothly trade variance for bias while preserving differentiability.

To learn safely with these regularized estimators, we derive a \emph{two-sided PAC-Bayes generalization bound} that will be used in a pessimistic objective:
\begin{align*}
 \left| V(\pi_{\theta}) - \hat{V}_n^\alpha(\pi_{\theta}) \right|   \leq \sqrt{ \frac{{\textsc{kl}}_{1}(\theta, \theta_0)}{2n} } + \underbrace{B_n^\alpha(\pi_{\theta})}_{\text{Regularization bias}} &+ \frac{{\textsc{kl}}_{2}(\theta, \theta_0)}{n \lambda } + \frac{\lambda}{2}\underbrace{\operatorname{Var}_n^\alpha(\pi_{\theta})}_{\text{Remaining variance}}\,,  
\end{align*}
where ${\textsc{kl}}_{1,2}$ measure divergence between the current policy $\pi_\theta$ and the logging policy $\pi_{\theta_0}$, $B_n^\alpha$ captures the regularization bias, and $\operatorname{Var}_n^\alpha$ captures the remaining variance. The exact expressions of these quantities are given in \cref{chap:ES}. The pessimistic learning objective maximizes the lower bound as:
\begin{align*}
    \hat{\pi} = \arg\max_{\pi_\theta} \left[ \hat{V}_n^\alpha(\pi_\theta) - \sqrt{ \frac{{\textsc{kl}}_{1}(\theta, \theta_0)}{2n} } - B_n^\alpha(\pi_{\theta}) - \frac{{\textsc{kl}}_{2}(\theta, \theta_0)}{n \lambda } - \frac{\lambda}{2}\operatorname{Var}_n^\alpha(\pi_{\theta}) \right]\,.
\end{align*}
This objective penalizes policies with high bias or variance, steering optimization toward reliable regions. What is important for scalability is that all terms in the objective are empirical and differentiable, enabling end-to-end optimization via standard stochastic gradient ascent. This contrasts with prior pessimistic objectives that relied on intractable theoretical constants or were incompatible with stochastic optimization.

We further present a \emph{unified PAC-Bayes framework} that generalizes this approach to all importance-weight regularizers in the literature (clipping, ES, implicit exploration), enabling fair comparison through a universal set of practical pessimistic objectives. This work also laid the foundation for \emph{logarithmic smoothing} (see Additional Contributions below), which refines the analysis to achieve significantly tighter bounds and sharp suboptimality guarantees.

\textbf{ICML 2023 (Oral) - UAI 2024 (Poster) - \citep{aouali23a,aouali2024unified}}

\begin{itemize}
    \item I. Aouali, V.-E. Brunel, D. Rohde, and A. Korba. Exponential smoothing for off-policy
learning. In \textit{Proceedings of the 40th International Conference on Machine Learning},
pages 984–1017. PMLR, 2023a.
    \item I. Aouali, V.-E. Brunel, D. Rohde, and A. Korba. Unified PAC-Bayesian study of pessimism
for offline policy learning with regularized importance sampling. In \textit{Uncertainty in
Artificial Intelligence}, pages 88–109. PMLR, 2024.
\end{itemize}

\end{description}

\subsection*{Additional Contributions}
This section outlines additional research conducted during this thesis. While these contributions are not included in the main manuscript, they correspond to published works involving equal or significant contributions from the author.

\begin{description}

\item[On-Policy: Extension to Best-Arm Identification]\mbox{}\par

We extend hierarchical and structured modeling from regret minimization to fixed-budget best-arm identification (BAI), introducing \emph{prior-informed best-arm identification} (\texttt{PI-BAI}): a non-adaptive algorithm that leverages prior knowledge for efficient budget allocation.

We provide a fully Bayesian analysis for structured settings (e.g., linear and hierarchical bandits), departing from classical frequentist approaches. This yields the first prior-dependent guarantees on Bayesian error probability in fixed-budget BAI. \texttt{PI-BAI} is robust to prior misspecification and consistently outperforms baselines, including adaptive strategies, challenging the prevailing assumption that adaptivity is essential for fixed-budget exploration.

\textbf{AISTATS 2025 (Poster) - \citep{nguyen2025prior}}:
\begin{itemize}
    \item N. Nguyen, I. Aouali, A. György, and C. Vernade. Prior-dependent allocations for Bayesian fixed-budget best-arm identification in structured bandits. In \textit{Proceedings of the 28th International Conference on Artificial Intelligence and Statistics (AISTATS)}, 2025.
\end{itemize}

\item[Off-Policy: Extension to Logarithmic Smoothing] \mbox{}\par
We refine the theoretical framework for regularized IPS estimators from \cref{chap:ES}. While the bounds developed there were useful for learning in practice, they can be loose for suboptimality guarantees in certain cases. To address this, we derive a general high-order moment concentration bound for regularized estimators and identify the estimator that minimizes this bound. This analysis yields a novel pessimistic estimator, \emph{logarithmic smoothing (LS)}:
$$
\hat{V}_{\textsc{ls}}^\lambda(\pi) = \frac{1}{n\lambda} \sum_{i=1}^n \log\left(1 + \lambda \frac{\pi(A_i \mid X_i)}{\pi_0(A_i \mid X_i)} R_i \right).
$$
Similar to ES, LS acts as a soft, differentiable alternative to clipping, concentrates at a sub-Gaussian rate, and achieves finite variance without requiring bounded importance weights. The resulting high-probability risk bound is provably tighter than state-of-the-art alternatives, enabling sharp suboptimality guarantees.

\textbf{NeurIPS 2024 (Spotlight) - \citep{Sakhi2024LS}}:
\begin{itemize}
    \item O. Sakhi, I. Aouali, P. Alquier, and N. Chopin. Logarithmic Smoothing for Pessimistic Off-Policy Evaluation, Selection and Learning. In \textit{Advances in Neural Information Processing Systems (NeurIPS)}, 2024.
\end{itemize}

\item[Off-Policy: Applications to Large-Scale Recommender Systems] \mbox{}\par

This CIFRE thesis maintained a continuous feedback loop between theory and practice. Our work on large-scale industrial recommender systems served a dual purpose: it provided a testing ground for the off-policy methods developed in this thesis, while the real-world challenges encountered in these systems directly motivated the theoretical questions addressed in the main chapters. This resulted in several workshop publications and tutorials shared with the community \citep{gilotte2025offline,aouali2022reward,aouali2022offline,aouali2022probabilistic,aouali2021combining}.
\begin{itemize}
    \item A. Gilotte, O. Sakhi, I. Aouali, and B. Heymann. Offline contextual bandit with counterfactual sample identification. \textit{arXiv preprint arXiv:2509.10520}, 2025.
    \item I. Aouali, A. Benhalloum, M. Bompaire, A. Ait Sidi Hammou, S. Ivanov, B. Heymann,
D. Rohde, O. Sakhi, F. Vasile, and M. Vono. Reward optimizing recommendation using
deep learning and fast maximum inner product search. In \textit{proceedings of the 28th ACM
SIGKDD conference on knowledge discovery and data mining}, pages 4772–4773, 2022a..

    \item I. Aouali, A. Benhalloum, M. Bompaire, B. Heymann, O. Jeunen, D. Rohde, O. Sakhi,
and F. Vasile. Offline evaluation of reward-optimizing recommender systems: The case
of simulation. \textit{arXiv preprint arXiv:2209.08642}, 2022b.
    \item I. Aouali, A. A. S. Hammou, O. Sakhi, D. Rohde, and F. Vasile. Probabilistic rank and
reward: A scalable model for slate recommendation. \textit{arXiv preprint arXiv:2208.06263},
2022c.
    \item I. Aouali, S. Ivanov, M. Gartrell, D. Rohde, F. Vasile, V. Zaytsev, and D. Legrand.
Combining reward and rank signals for slate recommendation. \textit{arXiv preprint
arXiv:2107.12455}, 2021.
    
\end{itemize}
\end{description}

%% file: contents/overview/related_work.tex
\section{Related Work}

\subsection{On-Policy Learning in Contextual Bandits}

In the on-policy (online) setting \citep{mab_book,lattimore19bandit,bubeck2012regret,li10contextual,chu11contextual}, the agent must balance choosing actions that maximize current reward estimates (\emph{exploitation}) with exploring other actions to improve these estimates (\emph{exploration}). This trade-off is often addressed using upper confidence bounds (UCBs) \citep{auer02finitetime} or Thompson sampling (TS) \citep{thompson33likelihood}.

\emph{Upper confidence bound (UCB)} algorithms handle the exploration-exploitation trade-off by constructing high-probability confidence intervals around reward estimates and selecting the action with the largest upper bound \citep{auer02finitetime,chu11contextual,abbasi2011improved,dani2008stochastic}. The intuition is \emph{optimism under uncertainty}: poorly explored actions have wide confidence intervals and thus large upper bounds, which encourages exploration. Despite strong theoretical guarantees, UCB methods are often less practical than TS due to sensitivity to confidence-parameter tuning and lack of inherent randomization. Nevertheless, UCB remains a cornerstone of bandit theory and continues to inspire new exploration strategies. \cref{part:online} focuses on TS, but the hierarchical principles developed there naturally extend to UCB-based exploration.

\emph{Thompson sampling (TS)} operates in a Bayesian framework: a prior and likelihood are specified, the agent samples rewards from the posterior at each round, and then chooses the action with the highest sampled reward. TS is randomized by construction, easy to implement, and exhibits strong empirical performance in both simulated and real-world problems \citep{russo14learning,chapelle11empirical,russo2018tutorial}. It also enjoys strong theoretical guarantees, including optimal or near-optimal regret in a variety of models \citep{kaufmann2012thompson,pmlr-v31-agrawal13a,korda2013thompson,russo14learning,agrawal2017near,abeille17linear,russo16information,lu19informationtheoretic}. \cref{part:online} advances TS by integrating informative hierarchical priors that enable efficient learning in large action spaces.

\emph{Hierarchical Bayesian bandits} \citep{bastani19meta,kveton21metathompson,basu21noregrets,simchowitz21bayesian,wan21metadatabased,hong22hierarchical,peleg22metalearning,wan22towards,tomkins2021intelligentpooling,urteaga18variational}
apply TS to simple graphical models in which action parameters are typically drawn from Gaussian distributions centered at a small number of latent parameters. These works primarily address meta- and multi-task learning in multi-armed bandits, transferring information across tasks or arms. Our mixed-effect Thompson sampling (\cref{chap:meTS}) extends this line of work by introducing a hierarchical structure with multiple latent effect parameters in the contextual bandit setting. It also provides Bayes regret bounds and computational guarantees in the large action space regime. Our diffusion Thompson sampling (\cref{chap:dTS}) further generalizes these approaches by replacing simple Gaussian hierarchies with deep, non-linear diffusion models that capture complex inter-action dependencies through flexible link functions $f_\ell$.

\emph{Approximate Thompson sampling} is a central challenge in Bayesian bandits because most posteriors are intractable and require approximate inference. Prior work \citep{riquelme2018deep,chapelle11empirical,kveton2020randomized} highlights the strong empirical performance of approximate TS in complex models. For mixed-effect TS (\cref{chap:meTS}), we exploit the Gaussian structure of the hierarchy. We first apply a Laplace-like approximation to the reward likelihood at the action level, obtaining a Gaussian \emph{pseudo-observation} on each parameter $\theta_a$. Because both the priors on latent effects and on action parameters are Gaussian and the hierarchy is linear, these pseudo-observations can then be propagated exactly in closed form through the hierarchy, yielding an approximate posterior that preserves the original mixed-effect structure. For diffusion TS (\cref{chap:dTS}), the prior hierarchy is defined by non-linear link functions $f_\ell$, so Gaussian propagation is no longer exact. To retain a hierarchical diffusion model, we make an additional approximation: at each update we locally linearize the link-function updates. Combined with the Laplace-like approximation on the likelihood, this yields a chain of conditional Gaussians with updated means and precisions, i.e., a posterior diffusion model that preserves the prior hierarchy while remaining computationally tractable.

\emph{Bandits with underlying structure} are closely related to our setting, where we assume structured relationships among actions. In latent bandits \citep{maillard14latent,hong20latent}, a single latent variable indexes multiple candidate models. In structured finite-armed bandits \citep{lattimore14bounded,gupta18unified}, each action is linked to a known mean function parameterized by a common latent parameter that is learned online. TS has also been applied to more complex structures such as graphical and combinatorial bandits \citep{gopalan14thompson,yu20graphical}. However, these methods do not simultaneously guarantee computational and statistical efficiency in large action spaces. Meta- and multi-task learning with UCB-style methods also has a long history \citep{azar13sequential,gentile14online,deshmukh17multitask,cella20metalearning,hu2021near,cella2022multi,yang2020impact}, but these works typically adopt a frequentist perspective, analyze stronger notions of regret, and often yield conservative algorithms. In contrast, our mixed-effect and diffusion TS algorithms are Bayesian, come with Bayes regret guarantees, and are explicitly designed to exploit pre-learned structure to achieve both statistical efficiency and scalable online inference.

\emph{Large action spaces.} Our work directly addresses the challenge of learning with per-action parameters $\theta_a$ (disjoint reward models) rather than a single shared parameter $\theta$ (joint reward models). The disjoint formulation, while more expressive and widely used in practice, faces severe scalability issues that we address through hierarchical structure. Our analysis shows that both mixed-effect TS (\cref{chap:meTS}) and diffusion TS (\cref{chap:dTS}) achieve regret bounds that scale with an effective number of actions $K_{\text{eff}} \ll K$. The expression of $K_{\text{eff}}$ depends however on $K$. Some  prior works \citep{foster2020adapting,xu2020upper,zhu2022contextual} propose bandit algorithms whose regret is independent of $K$. However, their setting differs substantially from ours: they assume a reward function $r(x, a) = \phi(x, a)^\top \theta$ with a single shared parameter $\theta \in \mathbb{R}^d$ and a known mapping $\phi$, whereas we consider $r(x, a) = \phi(x)^\top \theta_a$ (or simply $r(x, a) = x^\top \theta_a$) with $K$ separate $d$-dimensional action parameters. The dependence on $K$ in our setting reflects the inherent complexity of learning individual action parameters, which is the price paid for expressiveness. Obtaining a rich, known mapping $\phi$ that captures complex context-action dependencies can be challenging in practice, whereas our setting mirrors common scenarios such as recommender systems where each product has its own embedding learned from data. Note that both algorithms can be applied to the joint reward-model case; in that setting, our analysis would yield a $K$-independent regret bound.

\subsection{Off-Policy Learning in Contextual Bandits}

The challenges of large action spaces extend beyond on-policy learning to the equally important off-policy setting \citep{li2011unbiased,bottou2013counterfactual,swaminathan2015batch}, where decisions must be made using historical data collected under different policies. This section surveys the off-policy contextual bandit literature and positions the methods developed in \cref{part:offline}. 

Off-policy learning fundamentally relies on \emph{off-policy evaluation}, which estimates the value of a target policy $\pi$ using data collected under a logging policy $\pi_0$. Given logged data $\mathcal{D}_n = \{(X_i, A_i, R_i)\}_{i=1}^n$ with $A_i \sim \pi_0(\cdot \mid X_i)$, off-policy evaluation seeks to estimate $V(\pi)$ without deploying $\pi$. Off-policy learning then optimizes over a policy class using this estimated value. Consequently, the prevailing paradigm is \emph{estimator-centric}: first design an estimator $\hat{V}(\pi)$ with good statistical properties, then maximize it. As we show in \cref{chap:las}, this estimate-then-optimize approach breaks down in large action spaces because optimization error, rather than estimation error, becomes the bottleneck.

\emph{Inverse propensity scoring (IPS)} \citep{horvitz1952generalization,dudik2012sample} corrects for distribution shift via importance weights:
\[
\hat{V}_{\textsc{ips}}(\pi)
  = \frac{1}{n} \sum_{i=1}^n
      \frac{\pi(A_i \mid X_i)}{\pi_0(A_i \mid X_i)} R_i.
\]
While unbiased under the \emph{common support} assumption, IPS suffers from extremely high variance. It can also incur substantial bias when the logging policy has deficient support \citep{sachdeva2020off}, especially in large action spaces where the logging policy can only cover a small fraction of the actions. To mitigate variance, numerous importance-weight regularization techniques have been proposed, such as weight clipping \citep{ionides2008truncated,bottou2013counterfactual} and others \citep{su2020doubly,metelli2021subgaussian,gabbianelli2023importance,swaminathan2015self,gilotte2018offline}. \cref{chap:ES} introduces differentiable exponential-smoothing (ES) estimators that smoothly trade bias for variance and enable gradient-based optimization.

\emph{Direct methods (DM)} \citep{jeunen2021pessimistic,aouali2024bayesian} avoid importance weighting by modeling the expected reward for any context–action pair and evaluating policies using:
\[
\hat{V}_{\textsc{dm}}(\pi)
  = \frac{1}{n} \sum_{i=1}^n
      \sum_{a \in \mathcal{A}} \pi(a \mid X_i)\,\hat{r}(X_i, a).
\]
DM is particularly attractive in large-scale recommender systems where IPS struggles due to its high variance \citep{sakhi2020blob,jeunen2021pessimistic,aouali2022probabilistic}. However, standard implementations typically rely on a \emph{disjoint model} that estimates one parameter vector $\theta_a \in \real^d$ per action. In large action spaces with sparse logging, this leads to severe statistical inefficiency: many actions are rarely observed and thus poorly estimated.  \cref{chap:sDM} addresses this via the structured direct method (\texttt{sDM}), which leverages hierarchical Bayesian modeling to share statistical strength across actions.

\emph{Direct methods (DM)} \citep{jeunen2021pessimistic,aouali2024bayesian} build a model of the expected reward for any context–action pairs and evaluate policies using
\[
\hat{V}_{\textsc{dm}}(\pi)
  = \frac{1}{n} \sum_{i=1}^n
      \sum_{a \in \mathcal{A}} \pi(a \mid X_i)\,\hat{r}(X_i, a).
\]
DM is particularly attractive in large-scale recommender systems, where IPS struggles \citep{sakhi2020blob,jeunen2021pessimistic,aouali2022probabilistic}. Standard implementations, however, typically use a disjoint model that estimates one parameter vector $\theta_a$ per action. In large action spaces with sparse logging, this leads to severe statistical inefficiency: many actions are rarely observed and thus poorly estimated. \cref{chap:sDM} introduces the structured direct method (\texttt{sDM}), which addresses this limitation via hierarchical Bayesian modeling.

\emph{Doubly robust (DR) estimators} \citep{robins1995semiparametric,bang2005doubly,dudik2011doubly,dudik14doubly,farajtabar2018more} combine DM and IPS to achieve robustness:
\[
\hat{V}_{\textsc{dr}}(\pi)
  = \hat{V}_{\textsc{dm}}(\pi)
    + \frac{1}{n} \sum_{i=1}^n
      \frac{\pi(A_i \mid X_i)}{\pi_0(A_i \mid X_i)}
      \bigl(R_i - \hat{r}(X_i, A_i)\bigr).
\]
DR has become a default choice in many off-policy evaluation studies \citep{dudik2011doubly,dudik14doubly,farajtabar2018more,su2020doubly}. Both of our contributions in \cref{chap:ES} (regularized importance weights) and \cref{chap:sDM} (structured reward models) can be used to enhance the components of DR.

\emph{Large-scale IPS variants.} Importance-weight regularization alone is often insufficient when the action space is very large. Structural assumptions can dramatically reduce the variance. For instance, marginalized IPS (MIPS) \citep{saito2022off} clusters actions via a mapping $h(a)$ and works with cluster-level importance weights:
\[
\hat{V}_{\textsc{mips}}(\pi)
  = \frac{1}{n} \sum_{i=1}^n
      \frac{\pi(h(A_i) \mid X_i)}{\pi_0(h(A_i) \mid X_i)} R_i.
\]
This reduces variance by operating over a smaller cluster space instead of the full action set.  This dimensionality reduction principle has inspired numerous extensions \citep{peng2023offline,sachdeva2023off,cief2024learning,taufiq2024marginal,saito2023off}. Our \texttt{sDM} (\cref{chap:sDM}) provides a complementary structural approach designed for DM instead of IPS; it can be viewed as a Bayesian latent-structure counterpart to MIPS, replacing hard clustering with soft probabilistic coupling.

\emph{Pessimistic off-policy learning.} Maximizing a point estimator (IPS, DM, or DR) can be unsafe when the estimator deviates from the true value. Pessimistic approaches instead construct lower confidence bounds on $V(\pi)$ and optimize those, following the principle of \emph{pessimism in the face of uncertainty} \citep{jin2021pessimism}. 
Asymptotic and finite-sample lower bounds have been developed for various estimators \citep{bottou2013counterfactual,kuzborskij2021confident,gabbianelli2023importance}, 
providing worst-case guarantees on policy performance. Many pessimistic learning methods are directly motivated by such bounds \citep{swaminathan2015batch,london2019bayesian,kuzborskij2021confident,aouali23a,wang2023oracle}. For example, \citet{swaminathan2015batch} combine empirical-Bernstein inequalities with clipped IPS, leading to variance-penalized learning objectives. 

Recently, the PAC-Bayesian paradigm \citep{mcallester1998some,catoni2007pac,alquier2021user} has been increasingly applied to off-policy learning, offering a flexible toolkit for deriving data-dependent generalization bounds. \citet{london2019bayesian} introduced a scalable PAC-Bayesian perspective, which has been further developed by \citet{flynn,sakhi2022pac,aouali23a,aouali2024unified,gabbianelli2023importance} to yield tight, directly optimizable bounds. \cref{chap:ES} advances this direction by deriving a two-sided PAC-Bayes bound for exponentially smoothed IPS estimators, resulting in a fully differentiable pessimistic learning objective that jointly controls reward, bias, variance, and divergence from the logging policy. Moreover, this framework generalizes to encompass other estimators in the literature, establishing a unified set of principles for pessimistic learning. While our subsequent work on logarithmic smoothing (LS) \citep{Sakhi2024LS} further tightens these guarantees, \cref{chap:ES} lays the foundational theoretical groundwork on which LS builds.

\emph{Optimization-centric learning.} All methods above follow an estimator-centric philosophy: find a good value estimator and maximize it. In large action spaces, these estimator-based objectives typically induce highly non-concave landscapes with flat plateaus and many local maxima (\cref{chap:las}), ensuring that optimization error dominates estimation error.  \cref{chap:las} proposes a paradigm shift toward \emph{optimization-centric} off-policy learning. Rather than insisting on high-quality estimators of the value, we advocate for \emph{policy-weighted log-likelihood} objectives whose optimization landscape is benign, ensuring convergence to effective policies even in massive action spaces.

%% file: contents/part1/introduction.tex
\chapter{Introduction to \cref{part:online}}

This first part of the thesis addresses the following fundamental question: 
\begin{center}
    \emph{How can we design exploration-exploitation algorithms that remain both statistically efficient and computationally feasible when the number of actions is large?}
\end{center}

\section{Setting and Background}\label{sec:online_background}

In this part, we consider the on-policy (online) contextual bandit setting where an agent interacts with an environment over $T$ rounds. At each round $t \in [T]$:
\begin{enumerate}
    \item The agent observes a context $X_t \in \mathcal{X} \subseteq \mathbb{R}^d$ drawn from a distribution $\nu$;
    \item The agent selects an action $A_t \in \cA = [K]$ based on the history $H_t = \{(X_s, A_s, R_s)\}_{s=1}^{t-1}$;
    \item The agent receives a stochastic reward $R_t \sim p(\cdot \mid X_t; \theta_{*, A_t})$.
\end{enumerate}

Each action $a \in [K]$ is associated with an unknown true parameter $\theta_{*, a} \in \mathbb{R}^d$. The true expected reward is given by $r(x, a; \theta_*) = r(x; \theta_{*, a})$, where $\theta_* = (\theta_{*, a})_{a \in [K]} \in \mathbb{R}^{Kd}$ denotes the concatenation of all true action parameters. 

Throughout this part, we assume the reward distribution is a generalized linear model (GLM) \citep{mccullagh89generalized}. For any context $x \in \mathcal{X}$ and action $a \in \mathcal{A}$, $p(\cdot \mid x; \theta_{*, a})$ is an exponential-family distribution with mean $g(\phi(x)^\top  \theta_{*, a})$, where $g$ is the mean function. This formulation recovers linear bandits \citep{auer02using} when $p(\cdot \mid x; \theta_{*, a}) = \mathcal{N}(\cdot; \phi(x)^\top \theta_{*, a}, \sigma^2)$ (with identity link $g(u)=u$), and logistic bandits \citep{filippi10parametric} when $p(\cdot \mid x; \theta_{*, a}) = \mathrm{Ber}(g(\phi(x)^\top \theta_{*, a}))$ with the sigmoid link $g(u) = (1+\exp(-u))^{-1}$.

We adopt a Bayesian perspective where the unknown true parameters $\theta_*$ are assumed to be drawn from a prior distribution $p_0$. Our objective is to minimize the \emph{Bayes regret}:
\begin{align}\label{eq:optimality}
\mathcal{BR}(T)  =  \mathbb{E}\left[\sum_{t=1}^T \Big( r(X_t, A_{t,*}; \theta_*) - r(X_t, A_t; \theta_*) \Big)\right],
\end{align}
where $A_{t,*} = \arg\max_{a \in [K]} r(X_t, a; \theta_*)$ is the optimal action at round $t$. The expectation is taken over the prior $p_0$, the stochastic rewards, contexts, and the agent's policy.

\subsection{Scalability Challenges}

Standard \emph{Thompson sampling (TS)} maintains a posterior distribution $p(\theta \mid H_t)$ over the parameter space and sampling $\theta_t \sim p(\cdot \mid H_t)$ at each round to select the action $A_t = \arg\max_{a \in [K]} r(X_t, a; \theta_t)$. However, when the number of actions $K$ is large, there is trade-off between statistical and computational efficiency:

\textbf{Statistical inefficiency of disjoint priors.} A common simplification is to learn each action's parameter $\theta_a$ independently using a factorized prior $p_0(\theta) = \prod_{a=1}^K p_{0,a}(\theta_a)$. While this makes posterior updates computationally cheap, it prevents information sharing. The agent must learn about every action from scratch, which is prohibitive in large action spaces.

\textbf{Computational intractability of joint priors.} Conversely, modeling dependencies via a full joint posterior over $\mathbb{R}^{dK}$ allows for information sharing but is computationally intractable. Storing the covariance requires $\mathcal{O}(K^2 d^2)$ memory, and a single update requires $\mathcal{O}(K^3 d^3)$ time, making it intractable for online interaction.

\section{Hierarchical Models}

To address this dilemma, we propose a general hierarchical Bayesian framework where action parameters are coupled through a set of \emph{latent parameters} $\Psi = (\psi_{\ell})_{\ell \in [L]} \in \mathbb{R}^{L d}$, with $L \ll K$. The generative process is defined as:
\begin{align}\label{eq:general_model}
\Psi &\sim q_0(\cdot) && \text{(Prior over latent structure)}, \\
\theta_a \mid \Psi &\sim p_{0,a}(\cdot \mid \Psi)\,, & \forall a \in [K] \qquad & \text{(Conditional prior per action)}, \\
R_t \mid \theta, \Psi, X_t, A_t &\sim p(\cdot \mid X_t; \theta_{A_t})\,, & \forall t \in [T] \qquad & \text{(Reward observation)}.
\end{align}
Here, $q_0$ encodes global uncertainty, while $p_{0,i}$ specifies how individual actions deviate from the shared structure. The resulting marginal prior $p_0(\theta)$ naturally couples all action parameters. The corresponding  posterior preserves this hierarchy:
\begin{align}
p(\theta, \Psi \mid H_t) = p(\Psi \mid H_t)\, \prod_{a=1}^K p(\theta_a \mid \Psi, H_{t,a}).
\end{align}
The latent posterior $p(\Psi \mid H_t)$ aggregates evidence from all actions, enabling \emph{global information sharing}, while the action posteriors $p(\theta_a \mid \Psi, H_{t,a})$ allow for efficient sampling of action parameters $\theta_a$ independently given $\Psi$. Thompson sampling then proceeds by first sampling the global structure $\Psi_t$, then sampling action parameters $\theta_{t,a}$ conditioned on $\Psi_t$.

\section{Roadmap of \cref{part:online}}

The following chapters present two concrete instantiations of this hierarchical framework.

\textbf{\cref{chap:meTS}: Mixed-Effects Thompson Sampling.}
We begin by investigating a linear instantiation of the hierarchy where each action parameter is modeled as a linear combination of $L$ shared effects, $\theta_a \mid \Psi \sim \cN(\sum_{\ell=1}^L b_{a,\ell} \psi_{\ell}, \Sigma_{0, a})$, with known weights $b_{a, \ell}$. We derive \emph{mixed-effect Thompson sampling} (\texttt{meTS}), an algorithm that exploits the conjugacy of linear-Gaussian models to perform exact, closed-form posterior updates. For non-linear GLM rewards, we introduce a tractable Laplace approximation that we propagate through the hierarchy. We provide theoretical guarantees showing that \texttt{meTS} achieves a Bayes regret bound scaling with the effective number of actions $K_{\textsc{eff}} \ll K$.

\textbf{\cref{chap:dTS}: Diffusion Thompson Sampling.}
We then extend the framework to support deep, non-linear hierarchical structures using \emph{diffusion models}. In this setting, the priors form a Markov chain of latent variables $\psi_L \to \dots \to \psi_1 \to \theta_a$, connected by potentially non-linear link functions $f_\ell$ (e.g., neural networks) learned from offline data. We propose \emph{diffusion Thompson sampling} (\texttt{dTS}) and develop a posterior approximation that updates the link functions and covariances to match observed data while preserving the generative diffusion. This chapter demonstrates how to leverage powerful generative priors for exploration while remaining computationally feasible for online deployment.

%% file: contents/mixed_effect_ts/main_text.tex
This chapter begins with the fundamental observation that the expected rewards of actions in real-world problems are often correlated. To model this phenomenon, we study a structured mixed-effect bandit environment in which each \emph{action parameter} depends on one or more \emph{effect parameters} that are shared across actions. Therefore, taking an action teaches the agent about its effect parameters, thereby informing it about other actions that share the same effect parameters. We present three motivating examples for this.

\textbf{Movie recommendation.} Here, we want to recommend a movie to a user with the highest expected rating. User $j$ and movie $a$ are represented by vectors $x_j$ (context) and $\theta_a$ (action parameter), respectively. The expected rating that user $j$ gives to movie $a$ is $x_j^\top \theta_a$. We assume that the vector $x_j$ is observed. Then the most natural idea is to learn all $\theta_a$ individually using standard bandit methods \citep{li10contextual,chu11contextual}. This is statistically inefficient when the number of movies is high. Fortunately, the movies could be organized into $L$ categories and such information can be leveraged to explore efficiently. We present three approaches \textbf{(A)}, \textbf{(B)} and \textbf{(C)} that do this next.

\begin{enumerate*}[label=\textbf{(\Alph*)}]
  \item For each category $\ell \in [L]$, a parameter $\psi_\ell$ is learned online using all interactions with the movies in category $\ell$. The parameter $\psi_\ell$ represents all the movies in category $\ell$ and is used instead of their individual $\theta_a$. Therefore, this approach has a high bias, as all movies in the same category are assumed to have the same expected rating. This issue can be addressed by a better model.
  \item We model each movie parameter $\theta_a$ as a random variable centered in its category parameter $\psi_\ell$. Now movies in the same category no longer have the same expected rating due to the additional uncertainty. Both the category parameters $\psi_\ell$ and movie parameters $\theta_a$ are learned online. The former is learned using all interactions with the movies in category $\ell$, while the latter is learned using all interactions with movie $a$ conditioned on $\psi_\ell$. The category parameter $\psi_\ell$ is learned using more data, which helps to learn $\theta_a$ more efficiently. This is a special case of our setting. 
  \item The shortcoming of \textbf{(B)} is that each movie belongs to a \emph{single category}, which is unrealistic. To address this issue, we allow movies to belong to \emph{multiple categories} and then proceed as in \textbf{(B)}. To connect with our terminology, the categories $\ell \in [L]$ denote effects; their parameters $\psi_\ell$ are the effect parameters; and the movie parameters $\theta_a$ are the action parameters.
\end{enumerate*}

\textbf{Ad placement:} Here, the agent selects a list (or \emph{slate}) of $M$ items from a catalog of $L$ items with the objective of maximizing the click-through-rate. We assume that the agent receives only binary bandit feedback indicating whether the user clicked \emph{one of the items in the slate} \citep{ijcai2019-308,RejwanM20}. Again, user $j$ and slate $a$ are represented by $x_j$ (context) and $\theta_a$ (action parameter), respectively. The corresponding click-through-rate is $g(x_j^\top \theta_a)$, where $g$ is the sigmoid function. The set of slates (of size $K \approx L^M$) is exponentially large, which makes learning $\theta_a$ individually difficult. Fortunately, the slates are related through a much smaller set of items (of size $L$). Therefore, slates containing common items can teach the agent about one another, enabling efficient exploration.

Efficient exploration is achieved by decomposing slate $a$'s parameter as $\theta_a = \sum_{\ell \in [L]} b_{a, \ell} \psi_\ell + \epsilon_a$. Here $\psi_\ell \in \real^d$ is the parameter of item $\ell$ and $b_{a, \ell} \in \real$ is a mixing weight that captures position biases. That is, $b_{a, \ell} = 0$ if item $\ell$ is not in slate $a$, and $b_{a, \ell}$ is high if item $\ell$ is ranked high in slate $a$. This captures the fact that the probability of a click on an item is influenced by its position on the slate, and this bias can be estimated offline. Finally, $\epsilon_a$ is a random noise that can incorporate uncertainty due to \emph{model misspecification}, for instance due to an estimation error of $b_{a, \ell}$. The benefit of this decomposition is that the parameter of item $\ell$, $\psi_\ell$, is learned using all interactions with the slates with item $\ell$. The slate parameter $\theta_a$ is learned using all interactions with slate $a$ conditioned on $\psi_\ell$. This is more statistically efficient than learning $\theta_a$ individually, which only uses the interactions with slate $a$.

\textbf{Drug design:} Here, the goal is to find the optimal drug design in clinical trials \citep{pmlr-v85-durand18a}. Subject $j$ and drug $a$ are represented by vectors $x_j$ and $\theta_a$, respectively, and the expected efficacy of drug $a$ for subject $j$ is $x_j ^\top \theta_a$. Again, the most natural idea is to learn all drug parameters $\theta_a$ individually. This leads to statistical inefficiency when the number of candidate drugs is high. Fortunately, we can leverage the fact that drug candidates in the same trial often share components to explore efficiently. Precisely, a drug is a combination of multiple components, each with a specific dosage. Each component $\ell$ is represented by a parameter $\psi_\ell$, and the drug parameter $\theta_a$ is a \emph{known} combination of the component parameters $\psi_\ell$ weighted by their dosage. That is, $\theta_a = \sum_{\ell \in [L]} b_{a,\ell} \psi_\ell + \epsilon_a$, where $b_{a, \ell}$ is the dosage of component $\ell$ in drug $a$ and $\epsilon_a$ is a random noise to incorporate uncertainty due to model misspecification. The efficacy of each component has an \emph{effect} on the overall efficacy of the drug and is boosted by the dosage.

In all examples, we assume an underlying structure among the actions, that they are affected by multiple effects. In some problems, it is known how the effect arises. For instance, in the drug design, the actions are the drugs and the effects are their components. The mixing weight that relates an action (drug) to an effect (component) is the dosage of that component in the drug. In other problems, it may not be apparent how the effect arises and this has to be learned. We discuss this in detail in \cref{subsec:creating_structure}.

We make the following contributions. \begin{enumerate*}[label=\textbf{\arabic*)}]
  \item We formalize a general mixed-effect bandit framework represented by a two-level graphical model where each action is associated with a $d$-dimensional parameter that depends on \emph{one or multiple} effect parameters.
  \item We design mixed-effect Thompson sampling (\meTS), which leverages this structure to be both statistically and computationally tractable. We show that closed-form posteriors can be derived for Gaussian instances and efficient approximations exist in more general cases.
  \item We prove that the Bayes regret of \meTS is bounded by a sum of two terms: one is associated with learning the action parameters and the other quantifies the cost of learning the effect parameters. Both terms reflect the structure of the environment and the quality of priors.
  \item We show empirically that \meTS and its variants perform extremely well, and are computationally efficient in both synthetic and real-world problems.
\end{enumerate*}

\section{Setting}
\label{sec:mets-setting}

We consider the contextual bandit setting in \cref{sec:online_background}. Each action $a \in \cA = [K]$ is associated with an \emph{unknown $d$-dimensional action parameter} $\theta_a \in \real^d$. The correlations between the action parameters arise because they are derived from $L$ shared \emph{unknown $d$-dimensional effect parameters}, $\psi_{\ell} \in \real^d$ for $\ell \in [L]$. Specifically, we assume that the action parameter $\theta_a$ is sampled from the \emph{action prior distribution} $p_{0, a}$ as $ \theta_a \mid  \Psi \sim p_{0, a}(\cdot \mid \Psi)$, where $\Psi = (\psi_{\ell})_{\ell \in [L]} \in \real^{Ld}$ is a concatenation of the effect parameters. The distribution $p_{0, a}$ can capture sparsity, when $\theta_a$ depends only on a subset of $\Psi$; and also incorporate uncertainty due to model misspecification, when $\theta_a$ is not a deterministic function of $\Psi$. Finally, the effect parameters $\Psi$ are sampled from a \emph{joint effect prior} $q_0$, which is known by the agent and represents its initial uncertainty about $\Psi$. In summary, all variables in our environment are generated as 
\begin{align}\label{eq:model}
  \Psi &\sim q_0\,, \\
  \theta_a \mid \Psi &\sim p_{0, a}(\cdot \mid \Psi)\,, &  \forall a \in \cA\,,\nonumber \\
  R_t \mid X_t, A_t, \theta, \Psi &\sim p(\cdot \mid X_t; \theta_{A_t})\,, &  \forall t \in [T]\,, \nonumber
\end{align}
where $p(\cdot \mid x; \theta_a)$ is the \emph{reward distribution} of action $a$ in context $x$, \textit{which only depends on parameter $\theta_a$ and the context $x$}. The terminology of effect parameters arises from the fact that $\psi_{\ell}$ affect the model parameters $\theta_a$, which in turn define $R_t$. The effects are mixed through the action prior $p_{0, a}$ and hence the name \emph{mixed-effect}.

\begin{figure}
  \centering
  \includegraphics[width=0.5\linewidth]{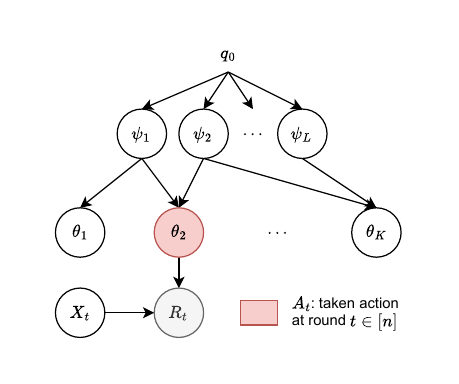}
  \caption{Example of a graphical model induced by \cref{eq:model}.}
  \label{fig:setting}
\end{figure}

Our setting can be viewed as a two-level graphical model, where $\psi_{1}, \ldots, \psi_{L}$ are parent nodes and $\theta_{1}, \ldots, \theta_{K}$ are child nodes (\cref{fig:setting}). The \emph{structure} is represented by missing arrows from parent (effect parameters) to child (action parameters) nodes. A missing arrow from parent $\psi_{\ell}$ to child $\theta_a$ means that action $a$ is independent of the $\ell$-th effect.

Our model can capture all examples provided in the introduction of this chapter. For instance, in movie recommendation, the categories $\ell \in [L]$ and movies $a \in \cA$ would be represented by the effect parameters $\psi_{\ell}$ and action parameters $\theta_a$, respectively. The weight $b_{a, \ell}$ is the relevance of movie $a$ to category $\ell$.

\textbf{Linearity in effects.} A simple yet powerful assumption is that the action prior $p_{0, a}$ is parametrized by a weighted sum of effect parameters
\begin{align*}
    \theta_a &\mid \Psi \sim p_{0, a}\Big(\cdot \Big| \sum_{\ell =1}^L b_{a, \ell} \psi_{\ell}\Big)\,, & \forall a \in \cA\,,
\end{align*} 
where $b_a = (b_{a, \ell})_{\ell \in [L]} \in \mathbb{R}^L$ are $L$ \emph{known mixing weights} for action $a$. The effect $\ell$ on action $a$ is determined by $b_{a, \ell}$. As an example, $b_{a, \ell}=0$ when action $a$ is independent of effect $\ell$. This is an important special case of our setting, since additive models are widely used in both theory and practice, as they often yield closed-form posteriors that are computationally tractable. Next we present two instances of this setting, where $p_{0, a}$ is a multivariate Gaussian with mean $\sum_{\ell =1}^L b_{a, \ell} \psi_{\ell}$ and covariance $\Sigma_{0, a}$.

\subsection{Mixed-Effect Linear Bandit}
\label{subsec:contextual_gaussian_bandits}

A natural joint effect prior $q_0$ for $d$-dimensional effect parameters $\psi_{\ell}$ is a multivariate Gaussian with mean $\mu_{\Psi} \in \real^{Ld}$ and covariance $\Sigma_{\Psi} \in \real^{Ld \times Ld}$. The action prior $p_{0, a}$ is a Gaussian with mean $\sum_{\ell =1}^L b_{a, \ell} \psi_{\ell} \in \real^d$ and covariance $\Sigma_{0, a} \in \real^{d \times d}$:
\begin{align}\label{eq:contextual_gaussian_model}
    \Psi & \sim \cN(\mu_{\Psi}, \Sigma_{\Psi})\,, \\ 
    \theta_a \mid  \Psi& \sim \cN\Big( \sum_{\ell =1}^L b_{a, \ell} \psi_{\ell} , \,  \Sigma_{0, a}\Big)\,, & \forall a \in \cA\,,\nonumber\\ 
    R_t \mid X_t, A_t, \theta, \Psi & \sim \cN(X_t^\top \theta_{A_t} ,  \sigma^2)\,, & \forall t \in [T]\,, \nonumber
\end{align}
where $\sigma^2>0$ is the variance of the observation noise.

\subsection{Mixed-Effect Generalized Linear Bandit}
\label{subsec:contextual_gl_bandits}

Here the effect and action parameters are generated as in \cref{eq:contextual_gaussian_model} but the reward $R_t$ is sampled from a \emph{generalized linear model (GLM)} \citep{mccullagh89generalized}, which is non-linear. In particular, $p(\cdot \mid X_t; \theta_a)$ is an exponential-family distribution with mean $g(X_t^\top \theta_a)$ and the whole model is
\begin{align}\label{eq:contextual_bernoulli_model}
    \Psi & \sim \cN(\mu_{\Psi}, \Sigma_{\Psi})\,,\\ 
    \theta_a \mid  \Psi& \sim \cN\Big( \sum_{\ell =1}^L b_{a, \ell} \psi_{\ell} , \,  \Sigma_{0, a}\Big)\,, & \forall a \in \cA\,,\nonumber\\ 
R_t \mid X_t, A_t, \theta, \Psi &\sim p(\cdot \mid X_t; \theta_{A_t})\,, &  \forall t \in [T]\,.\nonumber
\end{align} 
Let $ \mathrm{Ber}(p)$ be a Bernoulli distribution with mean $p$. One particular choice of a GLM is $g(u) = 1 / (1 + \exp(-u))$ and $p(\cdot \mid X_t; \theta) = \mathrm{Ber}(g(X_t^\top \theta))$, which corresponds to a logistic bandit \citep{filippi10parametric}.

\begin{remark}
    Note that in both settings, we use $x^\top \theta$ instead of $\phi(x)^\top \theta$ for some feature-map $\phi$, but this is just for ease of exposition, and everything generalizes smoothly to when using $\phi$.
\end{remark}

\subsection{Structure Learning}
\label{subsec:creating_structure}

The structures in \cref{eq:contextual_gaussian_model,eq:contextual_bernoulli_model} may be intrinsic in some problems, such as drug design. When this is not the case, we propose the following approach to learning a \emph{proxy structure}. For any $a \in \cA$, let $\hat{\theta}_a$ represent an offline estimate of action parameter $\theta_a$ (e.g., learned offline using interactions from previous bandit tasks). To learn, we fit a Gaussian mixture model (GMM) \citep{reynolds2009gaussian} with $L$ clusters to $\hat{\theta}_a$. Each cluster $\ell \in [L]$ is represented by its center $\mu_{\psi_\ell} \in \real^d$ and covariance $\Sigma_{\psi_\ell} \in \real^{d \times d}$. These correspond to the mean of the effect parameter $\psi_\ell$ and its uncertainty. The GMM also outputs the probability that $\hat{\theta}_a$ belongs to cluster $\ell$, for all combinations of $a \in \cA$ and $\ell \in [L]$. This probability is the mixing weight $b_{a, \ell}$.

The proposed procedure is general and adaptable to a wide range of use cases. The primary challenge lies in deriving the offline estimates $\hat{\theta}_a$. A straightforward approach involves learning these parameters from historical data collected in previous bandit tasks. Broadly, this can be formulated as an offline representation-learning problem \citep{tripuraneni2021provable}, for which numerous techniques exist. For instance, in our MovieLens experiments (\cref{sec:mets-movielens experiments}), we employ a low-rank factorization of the rating matrix to obtain these estimates. A key strength of our approach is its flexibility; it integrates seamlessly with standard offline learning tools, thereby taking a step toward bridging the gap between offline and online learning.

\section{Algorithm}
\label{sec:mets-algorithm}

\begin{algorithm}[t]
\caption{\meTS: \textbf{M}ixed-\textbf{E}ffect \textbf{T}hompson \textbf{S}ampling.}
\label{alg:ts}
\textbf{Input:} Joint effect prior $q_0$, action priors $p_{0, \cdot}$ \\
Initialize $q_1 \gets q_0$ and $p_{1, \cdot} \gets p_{0, \cdot}$ \\
\For{$t=1, \dots, T$}
{Sample $\Psi_t \sim q_t$ \\
\For{$a=1, \dots, K$}{Sample $\theta_{t, a} \sim p_{t, a} (\cdot \mid  \Psi_t)$}
$\theta_t \gets (\theta_{t, a})_{a \in \cA}$ \\
$A_t \gets \argmax_{ a \in \cA} r(X_t, a; \theta_t)$ \\
Receive reward $R_t \sim p(\cdot \mid X_t; \theta_{*, A_t})$ \\
Compute new posteriors $q_{t+1}$ and $p_{t+1, \cdot}$
}
\end{algorithm}

We propose a Thompson sampling algorithm \citep{thompson33likelihood, russo14learning, ts_scott}, which is a natural Bayesian solution to our problem. The algorithm is based on hierarchical sampling \citep{lindley72bayes}, which reflects the structure in our model. Before we present it, we need to introduce additional notation. We denote by $H_t = (X_i, A_i, R_i)_{i \in [t-1]}$ the \emph{history} of all interactions of the agent up to round $t$, by $S_{t, a} = \{i \in [t - 1]: A_i = a\}$ the rounds where the agent takes action $a$ up to round $t$, and by $H_{t, a} = (X_i, A_i, R_i)_{i \in S_{t, a}}$ the corresponding history.

Our algorithm \meTS is presented in \cref{alg:ts}. Since effect parameters are shared across actions, their posteriors exhibit dependencies. To handle this, we maintain two types of posterior densities:
\begin{itemize}
  \item A \emph{joint effect posterior} $q_t(\Psi) = p(\Psi \mid H_t)$ for all effect parameters $\Psi$ in round $t$;
  \item An \emph{action posterior} $p_{t, a}(\theta \mid \Psi) = p(\theta_a \mid H_{t, a}, \Psi)$ for each action $a \in \cA$, conditioned on the effect parameters.
\end{itemize}

\meTS employs hierarchical sampling in each round $t$:
\begin{enumerate}
  \item Sample effect parameters: $\Psi_t \sim q_t(\cdot)$
  \item Sample action parameters: $\theta_{t, a} \sim p_{t, a}(\cdot \mid \Psi_t)$ for each $a \in \cA$
  \item Select action: $A_t = \argmax_{ a \in \cA} r(X_t, a; \theta_t)$ where $\theta_t = (\theta_{t, a})_{a \in \cA}$.
\end{enumerate}

This hierarchical sampling scheme is equivalent to sampling from the exact marginal posterior $p(\theta_a \mid H_t)$. To see this, observe that marginalizing over $\Psi$ yields:
\begin{align}\label{eq:sampling_equivalence}
  p(\theta_a \mid H_t) &= \int_{\Psi} p(\theta_a, \Psi \mid H_t) \, \mathrm{d}\Psi \,,\nonumber\\
  &= \int_{\Psi} p(\theta_a \mid \Psi, H_t) \, p(\Psi \mid H_t) \, \mathrm{d}\Psi\,,\nonumber\\
  &= \int_{\Psi} p_{t, a}(\theta_a \mid \Psi) \, q_t(\Psi) \, \mathrm{d}\Psi\,.
\end{align}

\subsection{Posterior Derivations}

The posteriors are computed as follows. We first express the joint effect posterior $ q_t$ as
\begin{align}\label{eq:qt_derviation}
  q_t(\Psi) 
  & \propto \prod_{a =1}^K \int_{\theta_a} \LL_{t, a}(\theta_a) p_{0, a}\left(\theta_a \mid \Psi \right) \dif \theta_a \ q_0(\Psi)\,, 
\end{align}
where $\LL_{t, a}(\theta_a)= \prod_{(x, a, r) \in H_{t, a}} p(r \mid x; \theta_a)$ is the likelihood of all observations of action $a$ up to round $t$ given $\theta_a$. Next, for any action $a \in \cA$, the action posterior $p_{t, a}$ is expressed as 
\begin{align}\label{eq:pti_derviation}
    p_{t, a}(\theta_a \mid \Psi) 
    &\propto \LL_{t, a}(\theta_a) p_{0, a}(\theta_a \mid \Psi)\,.
\end{align} 
$p_{t, a}$ is similarly sparse to $p_{0, a}$. Specifically, in any round $t$, $p_{t, a}$ and $p_{0, a}$ are parameterized by the same subset of effect parameters $\Psi$, since $\LL_{t, a}(\theta_a)$ does not depend on $\Psi$.

The joint effect posterior $ q_t$ and action posteriors $p_{t, a}$ have closed forms in Gaussian models, which allows efficient sampling and theoretical analysis. Beyond these, MCMC and variational inference can be used to approximate $ q_t$ and $p_{t, a}$. Next we derive closed-form posteriors for the mixed-effect model with linear rewards in \cref{eq:contextual_gaussian_model} and provide an efficient approximation for the mixed-effect model with non-linear rewards in \cref{eq:contextual_bernoulli_model}.

\subsection{Mixed-Effect Linear Bandit}
\label{sec:mets-linear bandit posterior}

Let
\begin{align}
    &G_{t, a} = \sigma^{-2} \sum_{i \in S_{t, a}} X_i X_i^\top\,, & B_{t, a} = \sigma^{-2} \sum_{i \in S_{t, a}} R_i X_i\,.
\end{align}
be the outer product of contexts corresponding to action $a$ up to round $t$, and their sum weighted by rewards, respectively. Both are scaled by the observation noise variance $\sigma^2$. Using these quantities, the effect posterior is defined as follows.

\begin{proposition}\label{thm:pt_gaussian}
For any round $t \in [T]$, the joint effect posterior is a multivariate Gaussian $ q_t = \cN(\bar{\mu}_t, \bar{\Sigma}_t)$, where 
\begin{align}    \label{eq:linear effect posterior}
    &\bar{\Sigma}_t^{-1}= \Sigma_{\Psi}^{-1} +  \sum_{a \in \cA} b_a b_a^\top \otimes \left( \Sigma_{0, a}^{-1} - \Sigma_{0, a}^{-1} (G_{t, a} + \Sigma_{0, a}^{-1})^{-1}\Sigma_{0, a}^{-1} \right)\,,\\
    \bar{\mu}_t &= \bar{\Sigma}_t \Big(\Sigma_{\Psi}^{-1}  \mu_{\Psi} +\sum_{a \in \cA} b_a \otimes ( \Sigma_{0, a}^{-1} (G_{t, a} + \Sigma_{0, a}^{-1})^{-1} B_{t, a})\Big)\,.\nonumber
\end{align}
\end{proposition}

The effect posterior is additive in individual actions and each action contributes to the effect posterior mean and covariance proportionally to $b_{a, \ell}$, which is the mixture weight for $\theta_a$ in \cref{eq:contextual_gaussian_model}. \cref{thm:pt_gaussian} is proved in \cref{app:effect posterior-derivation}.

Now we present the action posterior.
\begin{proposition}\label{thm:pti_gaussian}
For any round $t \in [T]$, action $a \in \cA$, and effect parameters $\Psi_t$, the action posterior is a multivariate Gaussian $p_{t, a}(\cdot \mid \Psi_t) = \cN(\cdot;\tilde{\mu}_{t, a}, \tilde{\Sigma}_{t, a})$, where
\begin{align}  \label{eq:linear action posterior}
   \tilde{\Sigma}_{t, a}^{-1}  &= \Sigma_{0, a}^{-1} + G_{t, a}\,, \\
    \tilde{\mu}_{t, a} &= \tilde{\Sigma}_{t, a} \Big( \Sigma_{0, a}^{-1} \sum_{\ell =1}^L b_{a, \ell} \psi_{t, \ell} + B_{t, a}  \Big)\,.\nonumber
\end{align}
\end{proposition}
The action posterior in \cref{eq:linear action posterior} is a standard multivariate Gaussian posterior whose prior depends on $\Psi_t$, which is sampled by \meTS. \cref{thm:pti_gaussian} is proved in \cref{app:conditional-posterior-derivation}.

\subsection{Mixed-Effect Generalized Linear Bandit}
\label{sec:mets-glb bandit posterior}
Closed-form posteriors are unavailable in this setting, so approximations are required. We use a Laplace-style scheme that approximates the \emph{likelihood} $\LL_{t,a}(\cdot)$ by a Gaussian, rather than applying Laplace to the full posterior. This choice preserves a Gaussian form that can be propagated analytically through the hierarchical updates.

Let $\mu^\textsc{lap}_{t,a}$ denote the MLE (see remark below for a discussion about the computation of the MLE in practice), and let $G^\textsc{lap}_{t,a}$ be the Hessian\footnote{Note that we are assuming a generalized linear model where the log-likelihood of the data associated with action \(a\) can be written as $\log \LL_{t,a}(\theta_a) = \sum_{i \in S_{t,a}}
  \big[
    R_i X_i^\top \theta_a - A(X_i^\top \theta_a) + C(R_i)
  \big]$
where \(C\) is a real-valued function and \(A\) is twice continuously differentiable, with derivative \(\dot{A} = g\) representing the mean function.} of $- \log \LL_{t,a}(\cdot)$:
$$
\mu^\textsc{lap}_{t,a} = \argmax_{\theta_a} \log \LL_{t,a}(\theta_a), \qquad
G^\textsc{lap}_{t,a} = \sum_{i \in S_{t,a}} \dot g(X_i^\top \mu^\textsc{lap}_{t,a})\, X_i X_i^\top.
$$

We then approximate the \emph{likelihood} (not the posterior) by
\begin{equation}
\LL_{t,a}(\theta_a)
\;\propto\;
\exp\!\left(
-\tfrac12(\theta_a-\mu^\textsc{lap}_{t,a})^\top G^\textsc{lap}_{t,a}(\theta_a-\mu^\textsc{lap}_{t,a})
\right),
\label{eq:laplace_approximation_rewrite}
\end{equation}

Substituting \cref{eq:laplace_approximation_rewrite} into \cref{eq:qt_derviation} yields
$$
q_t(\cdot) \approx \cN(\cdot;\bar\mu_t, \bar\Sigma_t),
$$
where $\bar\mu_t$ and $\bar\Sigma_t$ are computed as in \cref{thm:pt_gaussian}, except for the replacements
$$
G_{t,a} \gets G^\textsc{lap}_{t,a},
\qquad
B_{t,a} \gets G^\textsc{lap}_{t,a} \mu^\textsc{lap}_{t,a}.
$$

Similarly, substituting \cref{eq:laplace_approximation_rewrite} into \cref{eq:pti_derviation} gives
$$
p_{t,a}(\cdot \mid \Psi) \approx \cN(\cdot;\tilde{\mu}_{t,a}, \tilde{\Sigma}_{t,a}),
$$
with
$$
G_{t,a} \gets G^\textsc{lap}_{t,a},
\qquad
B_{t,a} \gets G^\textsc{lap}_{t,a} \mu^\textsc{lap}_{t,a}.
$$

Although these expressions follow mechanically from substituting the Gaussian likelihood approximation, the intuition is straightforward. The replacement $G_{t,a} \gets G^\textsc{lap}_{t,a}$ reflects the curvature induced by the nonlinear mean function $g$, while $B_{t,a} \gets G^\textsc{lap}_{t,a} \mu^\textsc{lap}_{t,a}$ mirrors the linear-Gaussian case, where the MLE $\hat\theta^{\textsc{mle}}_{t,a}$ is characterized by the normal equations $G_{t,a}\,\hat\theta^{\textsc{mle}}_{t,a} = B_{t,a}$, and its generalized-linear counterpart is $\mu^\textsc{lap}_{t,a}$.

\begin{remark}
The MLE $\mu^{\textsc{lap}}_{t,a} = \argmax_{\theta_a\in\mathbb{R}^d} \log \LL_{t,a}(\theta_a)$
may be ill-posed. In practice, we use a small $\ell_2$-regularized
estimator: $\mu^{\textsc{lap}}_{t,a} \in \argmax_{\theta_a \in \mathbb{R}^d}
\log \LL_{t,a}(\theta_a) - \frac{\lambda}{2}\|\theta_a\|_2^2$, where $\lambda>0$ to fix this.
\end{remark}

\subsection{Computational Complexity}
\label{sec:mets-alternative algorithm designs}

The benefit of modeling the effect parameters is not immediately clear. Thus, it is tempting to marginalize them out, and only maintain a single joint posterior of all action parameters $\theta \in \real^{Kd}$. Posterior updates in this case would be complex and computationally inefficient when $K \gg L$, which is common in practice.

The main advantage of \meTS is that the sampling of effect parameters $\Psi_t \sim  q_t$ allows us to use the conditional independence of actions given $\Psi$, and model $\theta_a \mid  H_{t, a}, \Psi_t$ independently. This is more computationally efficient than modeling the joint $\theta \mid H_t$ when $K \gg L$. To see this, suppose that all posteriors are multivariate Gaussians (\cref{sec:mets-linear bandit posterior}). In this case, $\theta \mid H_t$ requires $\mathcal{O}(K^2 d^2)$ space, due to storing a $Kd \times Kd$ covariance matrix; while \meTS requires only $\mathcal{O}((L^2 + K) d^2)$ space, due to storing the covariances of $ q_t$ and $p_{t, a}$. Since the sampling relies on covariance inverses, the time complexity also improves. For the joint posterior, it is $\mathcal{O}(K^3 d^3)$, while it is only $\mathcal{O}((L^3 + K) d^3)$ for \meTS.

One can also marginalize out the effect parameters $\Psi$ and have $K$ separate posteriors, one for each action parameter $\theta_a$. While this improves computational efficiency, it does not model that the actions are correlated, since $\theta_a \mid H_{t, a}$ is modeled instead of $\theta_a \mid H_{t}$. This leads to a statistical inefficiency due to the loss of information as the histories of other actions $H_{t, a'}$ are discarded. We validate this through theory (\cref{sec:mets-benefits_structure}) and experiments (\cref{sec:mets-experiments}).

\section{Analysis}
\label{sec:mets-analysis}

This section is organized as follows. First, we state our regret bound. Then, we discuss how it captures the structure of our problem. We use $\tilde{\mathcal{O}}$ for the big O notation up to polylogarithmic factors.

\subsection{Main Result}
\label{sub:main_result}
We analyze \meTS in the linear setting in \cref{subsec:contextual_gaussian_bandits}. 
Throughout, we assume that the \emph{true} action parameters and rewards are generated according to the same hierarchical model used by \meTS (\cref{eq:contextual_gaussian_model}), i.e., we operate in the fully well-specified setting. For ease of exposition, we further assume the existence of constants $\sigma_{0}, \sigma_{\Psi}, \kappa_{x} > 0$ such that
\[
\Sigma_{0,a} = \sigma_{0}^2 I_d \quad \text{for all } a \in \cA, 
\qquad 
\Sigma_{\Psi} = \sigma_{\Psi}^2 I_{Ld},
\qquad 
\|X_t\|_2^2 \le \kappa_x \quad \text{for all } t \in [T].
\]
The bound on $\|X_t\|_2$ is standard, and we relax the other two assumptions in \cref{proof:regret_proof}.

\begin{theorem}
\label{thm:regret} For any $\delta \in (0, 1)$, the Bayes regret of \emph{\meTS} in the mixed-effect model in \cref{subsec:contextual_gaussian_bandits} is bounded as
\begin{align}\label{eq:regret_terms}
  \mathcal{B}\mathcal{R}(T)
  \leq \sqrt{2 T 
  \left( \mathcal{R}^{\textsc{a}}(T) + \mathcal{R}^{\textsc{e}}(T) \right) \log(1 / \delta)} +
  c T\delta\,,
\end{align}
where $c = \sqrt{\frac{2}{\pi} \kappa_x(\sigma_0^2 + \kappa_b \sigma_\Psi^2 )}K\,, \ \kappa_b = \max_{a \in \cA}\normw{b_a}{2}^2\,,$
\begin{align*}
& \mathcal{R}^{\textsc{a}}(T) = dK c_\textsc{a} \log\big(1 + \frac{T\kappa_x\sigma_0^2}{d \sigma^2} \big)\,, \, c_\textsc{a} = \frac{ \kappa_x\sigma^2_{0}}{\log\big(1 + \frac{\kappa_x\sigma_0^2}{\sigma^{2}}\big)}\,,\\ 
&\mathcal{R}^{\textsc{e}}(T) = dL c_\textsc{e} \log\big(1 +  \frac{K \kappa_b \sigma^2_{\Psi} }{\sigma^2_{0} + \frac{\sigma^2}{T \kappa_x}}\big)\,,  \, c_\textsc{e}= \frac{\kappa_x \kappa_b \sigma_\Psi^2 \big(1 + \frac{\kappa_x\sigma_0^2}{\sigma^{2}}  \big)}{\log\big(1 + \frac{\kappa_x \kappa_b \sigma_\Psi^2}{\sigma^{2}}\big)}\,.&
\end{align*}
\end{theorem}
The second term in \cref{eq:regret_terms} is constant for $\delta = 1 / T$, in which case the above bound is $\tilde{\mathcal{O}}(\sqrt{T})$. The main quantities of interest are $\mathcal{R}^{\textsc{a}}(T)$ and $\mathcal{R}^{\textsc{e}}(T)$, and they have natural interpretations. $\mathcal{R}^{\textsc{a}}(T)$ corresponds to the action regression problem: with $K$ parameters of dimension $d$, prior width $\sigma_{0}$, maximum context length $\sqrt{\kappa_x}$, and $T$ observations with noise $\sigma$. The dependence of $\mathcal{R}^{\textsc{a}}(T)$ on these quantities is identical to a corresponding linear bandit \citep{lu19informationtheoretic}. On the other hand, $\mathcal{R}^{\textsc{e}}(T)$ corresponds to the effect regression problem: with $L$ parameters of dimension $d$, prior width $\sigma_\Psi$, maximum mixing-weight length $\sqrt{\kappa_b}$, and $K$ actions that can be viewed as observations with noise $\sigma_{0}$ (\cref{sec:mets-linear bandit posterior}). The dependence of $\mathcal{R}^{\textsc{e}}(T)$ on these quantities mimics those in $\mathcal{R}^{\textsc{a}}(T)$.

To simplify exposition, let $\kappa_x = \kappa_b = \sigma = 1$. Then
\begin{align}\label{eq:simplified_regret}
  \mathcal{B}\mathcal{R}(T)
  = \tilde{\mathcal{O}}\!\left(\sqrt{T d \big(K \sigma_0^2 + L \sigma_\Psi^2 (1+\sigma_0^2)\big)}\right)\,.
\end{align}
This can be re-written as $\mathcal{BR}(T)
  = \tilde{\mathcal{O}}\!\left(\sqrt{T d K_{\mathrm{eff}} (\sigma_0^2 + \sigma_\Psi^2)}\right)$,
where $ K_{\mathrm{eff}} = \frac{K \sigma_0^2 + L \sigma_\Psi^2 (1+\sigma_0^2)}{\sigma_0^2 + \sigma_\Psi^2}$ is the \emph{effective number of actions}. When $L \ll K$ and $\sigma_0^2 \ll \sigma_\Psi^2$, we have $K_{\mathrm{eff}} \ll K$, yielding significant regret reduction over standard Thompson Sampling.

The dependence on $\sigma_0^2$ and $\sigma_\Psi^2$ is natural: since Bayesian regret measures performance under the prior, smaller prior variances correspond to more informative beliefs about the true parameters, which makes learning easier and reduces regret. Conversely, larger variances reflect greater prior uncertainty and increase the difficulty of identifying the optimal action. The scaling with $K$, $L$, and $d$ is also intuitive: fewer parameters to estimate lead to lower regret. These trends are consistent with our empirical observations in \cref{app:experiments}.

\subsection{Benefits of Structure}
\label{sec:mets-benefits_structure}
Note that we do not provide a matching lower bound. To argue that our upper bound reflects the intrinsic structure of the problem, we compare \meTS to agents that either have access to more information or exploit less structure. We start with the former. Consider \meTS with known effect parameters $\Psi$. Setting $\sigma_\Psi = 0$ in \cref{eq:simplified_regret} yields the reduced regret
\[
\mathcal{B}\mathcal{R}(T) = \tilde{\mathcal{O}}(\sqrt{T d K \sigma_0^2}),
\]
which no longer depends on $L$. Likewise, consider \meTS under a perfectly specified linear model, in which $\theta_a = \sum_{\ell \in [L]} b_{a,\ell}\psi_\ell$ for all $a \in \cA$. This corresponds to $\sigma_0 = 0$, giving
\[
\mathcal{B}\mathcal{R}(T) = \tilde{\mathcal{O}}(\sqrt{T d L \sigma_\Psi^2}),
\]
which is independent of $K$. In particular, the $K$-dependence in our regret bound arises precisely from modeling the variability of action parameters around the effect parameters via $\Sigma_{0,a}$. Without it, the regret of \alg is independent of $K$

We now turn to an agent that neither knows $\Psi$ nor models it explicitly. This agent learns only $\theta$ (\cref{sec:mets-alternative algorithm designs}) by marginalizing out $\Psi$ in \cref{eq:contextual_gaussian_model}:
\begin{align*}
 \theta_a &\sim \cN\Big( \sum_{\ell=1}^L b_{a,\ell} \mu_{\psi_\ell}, \; \breve{\Sigma}_{0,a} \Big), \qquad \forall a \in \cA,
\end{align*}
where $\breve{\Sigma}_{0,a} = (\sigma_0^2 + \|b_a\|_2^2 \sigma_\Psi^2) I_d
$ is the marginal prior covariance and $(\mu_{\psi_\ell})_{\ell \in [L]}$ is the prior mean of the effects, so that $\mu_\Psi = (\mu_{\psi_\ell})_{\ell \in [L]}$ (\cref{subsec:contextual_gaussian_bandits}).  
Importantly, marginalizing out $\Psi$ and treating each action parameter independently discards action correlations, even though the true generative model induces correlations via the shared effects. This agent therefore uses a less structured and less informative prior than \meTS.

Using the definition of $\breve{\Sigma}_{0,a}$ and $\kappa_b = \max_{a \in \cA} \|b_a\|_2^2 = 1$, the regret of this agent scales as in \cref{eq:simplified_regret} with $\sigma_\Psi = 0$, except that the maximum prior variance $\sigma_0^2$ is replaced by $\sigma_0^2 + \sigma_\Psi^2$. Hence,
\[
\mathcal{B}\mathcal{R}(T) = \tilde{\mathcal{O}}\big(\sqrt{T d K (\sigma_0^2 + \sigma_\Psi^2)}\big).
\]

When $K > L$ (up to constants), this regret can be substantially larger than the bound for \meTS in \cref{eq:simplified_regret}. The improvement is on the order of $\sqrt{K/L}$ in regimes where the effects are far more uncertain than the actions, i.e., $\sigma_\Psi \gg \sigma_0$. For example, in our ad-placement setting, $L$ is the number of catalog items, while $K \approx L^M$ is the number of slates of size $M$. Thus $K/L \approx L^{M-1}$, with typical scales such as $L \approx 10^6$ and $M \approx 10$. Our empirical results in \cref{sec:mets-synthetic experiments,app:experiments} support this: \meTS significantly outperforms classical methods when the effect parameters are more uncertain than the action parameters.

\section{Experiments}
\label{sec:mets-experiments}

We evaluate \meTS on both synthetic and real-world problems. In each plot, we report the average values and their standard errors. Additional experiments are conducted in \cref{app:experiments}. The code is provided in this \href{https://github.com/imadaouali/Mixed-Effect-Thompson-Sampling}{Github repository}.

\subsection{Synthetic Experiments}
\label{sec:mets-synthetic experiments}

We start with two synthetic problems: the linear and logistic bandit settings in \cref{eq:contextual_gaussian_model,eq:contextual_bernoulli_model}, respectively. The effect prior is parameterized by $\mu_\Psi=\mathbf{0}_{Ld}$ and $\Sigma_\Psi=3I_{Ld}$, the action covariance is $\Sigma_{0, a} = I_d$ for all $a \in \cA$, and the observation noise is $\sigma=1$. We use this setting since modeling of the effect parameters is the most beneficial when they are more uncertain than the action ones (\cref{sec:mets-benefits_structure}). The context $X_t$ is sampled uniformly from $[-1, 1]^d$. We run $50$ simulations and sample the mixing weights $b_{a, \ell}$ from $[-1,1]$ in each run.

We consider the following baselines. For the linear setting, we compare \alglin (\cref{sec:mets-linear bandit posterior}), \linucb \citep{abbasi2011improved}, \lints \citep{agrawal13thompson} and \hierts \citep{hong22hierarchical}. For the logistic setting, we compare \algglm (\cref{sec:mets-glb bandit posterior}), \alglin (\cref{sec:mets-linear bandit posterior}), \ucbglm \citep{li17provably}, \glmts \citep{chapelle11empirical} and \hierts \citep{hong22hierarchical}. \glmucb \citep{filippi10parametric} is excluded because it exhibits very high regret. We also include variational mean-field approximations of \meTS (\alglinfa and \algglmfa), where the full Gaussian effect posterior $q_t$ is approximated as $q_t(\Psi) \approx \prod_{\ell=1}^L q_{t,\ell}(\psi_\ell)$. This factorization enables sampling each $\psi_\ell \in \real^d$ independently and replaces operations on a full $Ld \times Ld$ covariance with blockwise updates. This improves the time and space complexities of \meTS by $L^2$ and $L$, respectively.

All baselines but \hierts ignore the structure. \hierts incorporates the structure similarly to \alglin but only has a single effect parameter with prior $\cN(\mathbf{0}_d,3I_{d})$, with the same mean and covariance as the effect parameters of \meTS. To compare fairly with \lints and \glmts, their marginal prior mean and covariance are chosen as $\mathbf{0}_d$ and $\breve{\Sigma}_{0, a} = \Sigma_{0, a} + \Gamma_a \Sigma_\Psi \Gamma_a^\top$, where $\Gamma_a = b_a^\top \otimes I_d$. This is to account for the uncertainty of the effect parameters despite marginalizing them out.

\begin{figure}[H]
  \centering
  \includegraphics[width=\linewidth]{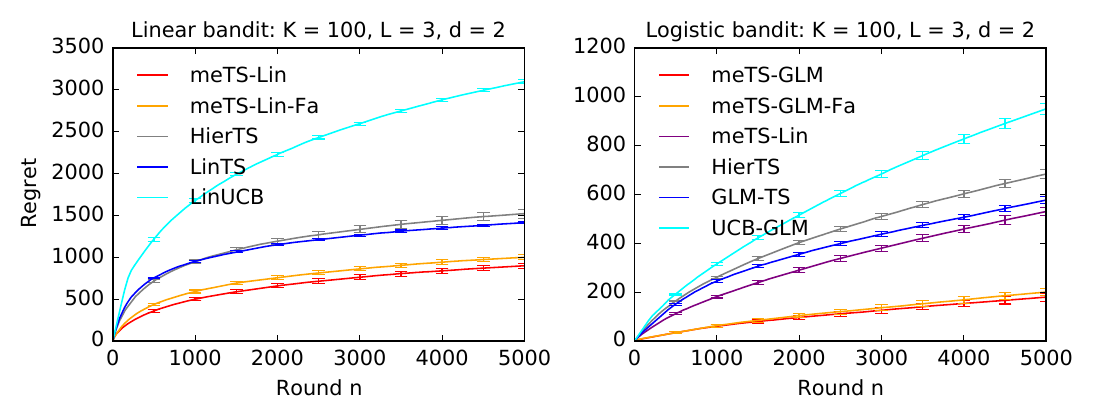}
  \caption{Evaluation on synthetic problems.}
  \label{fig:synthetic_regret}
\end{figure}

In \cref{fig:synthetic_regret}, we plot the regret in both problems for $T=5000, K=100, L=3$, and $d=2$ (higher values of $K$ up to $100,000$ are tested in our additional experiment in \cref{fig:varying_k} below). \meTS and its factored variant outperform all baselines that ignore the structure or incorporate it partially. Moreover, \algglm outperforms \alglin in the logistic bandit, which shows the benefit of the approximation in \cref{sec:mets-glb bandit posterior}. This attests to the generality and flexibility of \meTS and the posterior derivations in \cref{sec:mets-algorithm}. We also show in \cref{app:synthetic} that a higher $K, L$, or $d$ leads to a higher regret due to learning more parameters, which is captured by our regret bounds.

In \cref{fig:varying_k}, we examine how the final cumulative regret scales with the number of actions $K$ in the linear bandit setting, fixing $L=3$ and $d=2$. As $K$ increases from $100$ to $100,000$, \alglin consistently achieves substantially lower regret than \lints, with the gap widening as $K$ grows. This demonstrates that \alglin scales more favorably with the action space size by leveraging the shared effect structure, which aligns with our theoretical analysis. When $K \gg L$, this structural advantage becomes increasingly pronounced.

\begin{figure}[H]
  \centering
  \includegraphics[width=0.7\linewidth]{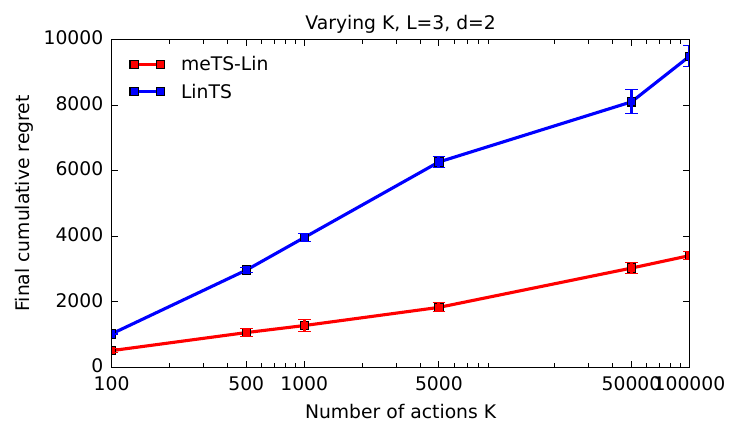}
  \caption{Final cumulative regret as a function of the number of actions $K$ in the linear bandit setting with $L=3$ and $d=2$.}
  \label{fig:varying_k}
\end{figure}

\subsection{MovieLens Experiments}
\label{sec:mets-movielens experiments}

We study the problem of movie recommendation using the MovieLens 1M dataset \citep{movielens}. This dataset contains one million ratings given by $6040$ users to $3952$ movies. We apply low-rank factorization to the rating matrix to obtain $5$-dimensional representations: $x_j \in \real^5$ for user $j \in [6040]$ and $\theta_a \in \real^5$ for movie $a \in [3952]$. We use the movies as actions and the context $X_t$ is sampled uniformly from user vectors $x_j$. We consider both linear and logistic rewards. Given a user $x_j$, the linear reward for movie $\theta_a$ is sampled from $\cN(x_j^\top \theta_a, \sigma^2)$ while the logistic reward is sampled from ${\rm Ber}(g(x_j^\top \theta_a))$, where $g$ is the sigmoid function. We run $50$ simulations with $K=100$ randomly sampled movies in each run. We compare \meTS to most baselines in \cref{sec:mets-synthetic experiments}. We do not include \ucbglm and \glmucb because their regret is very high. In \lints and \glmts, the prior mean of action $a$ is $\mu$ and its covariance is $\breve{\Sigma}_{0} = \mathrm{diag}(v) \in \real^{d \times d}$, where $\mu \in \real^d$ and $v \in \real^d$ are the mean and variance of the movie vectors along all dimensions, respectively.

The mixed-effect structure in \cref{eq:contextual_gaussian_model,eq:contextual_bernoulli_model} is not available in this problem. Therefore, we use the approach in \cref{subsec:creating_structure} to learn it. More precisely, we cluster the movies into $L=5$ mixture components by training a GMM on the offline action vectors $\theta_a$ (\cref{subsec:creating_structure}). Each cluster center corresponds to an effect parameter mean $\mu_{\psi_{\ell}} \in \real^d$ and the mixing weight $b_{a, \ell}$ is the probability that movie $a$ belongs to cluster $\ell$, as given by the GMM. We set the effect prior covariance as $\Sigma_\Psi = 0.75 \, \mathrm{diag}((\breve{\Sigma}_{0})_{\ell \in [L]}) \in \real^{Ld \times Ld}$ and the prior covariance of action $a$ as $\Sigma_{0, a} = 0.25 \, \breve{\Sigma}_{0} \in \real^{d \times d}$, where $\breve{\Sigma}_{0}$ is the same as in both \lints and \glmts. This means that the marginal covariance of action $a$ in \meTS is $0.25 \, \breve{\Sigma}_{0} + 0.75 \, \Gamma_a \Sigma_\Psi \Gamma_a^\top,$ where $\Gamma_a = b_a^\top \otimes I_d$. Therefore, it is on the same order as $\breve{\Sigma}_{0, a}$ when $\normw{b_a}{2}^2 \approx 1$, and \meTS is parameterized comparably to \texttt{LinTS} and \texttt{GLM-TS}. At the same time, we also model that the effect parameters are more uncertain than the action ones, since $\Sigma_{0, a} = 0.25 \, \breve{\Sigma}_{0}$ while $\Sigma_\Psi = 0.75 \, \mathrm{diag}((\breve{\Sigma}_{0})_{\ell \in [L]})$. 

\begin{figure}[H]
\includegraphics[width=\linewidth]{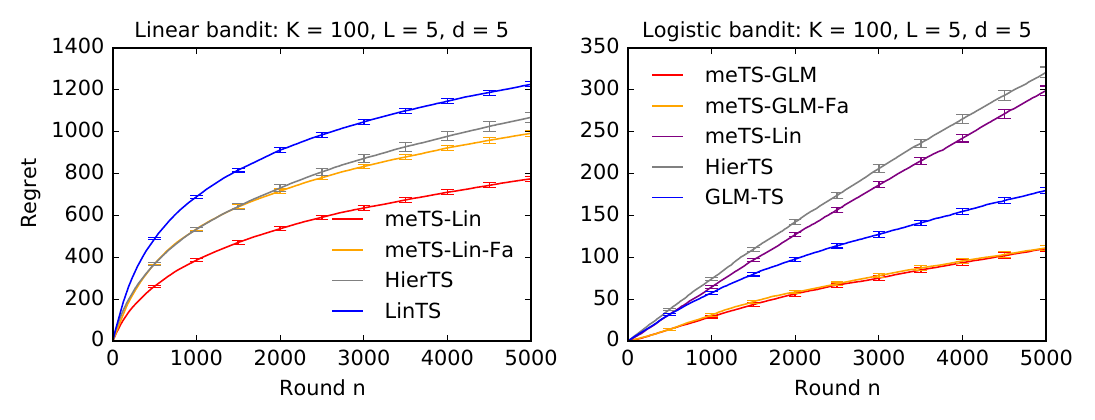}
\caption{Evaluation on MovieLens problems.}
\label{fig:movielens_regret}
\end{figure}

In \cref{fig:movielens_regret}, we plot the regret for $T=5000$ rounds. We observe that \meTS has the lowest regret, even if the true rewards are not generated from a mixed-effect model. This shows the robustness of \meTS to model misspecification, which we further validate in \cref{app:robustness_model_misspecification}. It also highlights the flexibility of our framework, where a proxy structure is learned from offline data.

\section{Conclusion}
\label{sec:mets-conclusions}

In this chapter, we introduced a mixed-effect bandit framework based on a two-level graphical model in which each action may depend on multiple underlying effects. This structure enables more efficient exploration, and we designed \meTS to leverage it effectively. When implemented as analyzed, \meTS performs strongly on both synthetic and real-world benchmarks. Although our presentation focused on the concrete models in \cref{eq:contextual_gaussian_model,eq:contextual_bernoulli_model}, the underlying algorithmic ideas extend seamlessly to the general mixed-effect model in \cref{eq:model}. The methodological and theoretical tools developed here lay the groundwork for richer formulations, one of which we explore in detail in the next chapter.

Our work has several limitations. First, the regret analysis assumes a well-specified prior: the true parameters must be generated from the same hierarchical model used by \meTS. While some experiments suggest robustness to misspecification, formal guarantees under prior mismatch remain open. Second, closed-form posteriors are available only for linear-Gaussian rewards; generalized linear models require Laplace approximations (\cref{sec:mets-glb bandit posterior}), which are not analyzed theoretically. Third, the mixed-effect structure must be known or learned offline, adding an additional modeling step. Finally, \meTS models only two-level hierarchies; deeper latent structures, which may better capture complex action correlations, require the diffusion-based approach developed in \cref{chap:dTS}.

%% file: contents/dts/main_text.tex
In the previous chapter, we explored how action correlations can be captured using mixed-effects models, in which actions share a set of effect parameters. This approach proved effective when the underlying structure, such as categories in movie recommendation or components in drug design, can be learned through clustering. However, real-world action correlations might exhibit more complex patterns. Thus, this chapter presents an alternative approach inspired by the remarkable success of diffusion models in approximating complex distributions \citep{sohl2015deep,ho2020denoising,dhariwal2021diffusion,rombach2022high}. Rather than explicitly modeling shared effects, we leverage pre-trained diffusion models to capture the rich structure of action parameters and use them as priors in contextual Thompson sampling. 

We make the following contributions. \begin{enumerate*}[label=\textbf{\arabic*)}]
  \item We introduce a framework for contextual bandits with \textit{diffusion-derived priors} and develop diffusion Thompson sampling (\dTS), which is both statistically efficient and computationally tractable. \dTS enables fast posterior updates and sampling via an efficient approximation inspired by exact Gaussian posteriors.
  \item Beyond applying pre-trained diffusion models to contextual bandits, a key contribution is enabling efficient \emph{posterior computation} and \emph{sampling} for a $d$-dimensional parameter $\theta \mid \mathcal{D}$ under a diffusion model prior, without updating the diffusion model parameters (i.e., without backpropagating through the neural network). This is relevant not only to bandits and RL but also to broader applications \citep{chung2022diffusion}. Our approximations are motivated by exact closed-form solutions available when the diffusion model is fully linear; these solutions form the basis for our nonlinear approximations, which achieve strong empirical performance while avoiding the computational burden of standard approximate posterior sampling techniques.
\end{enumerate*}

\section{Setting}
\label{sec:dts_setting}
We consider the contextual bandit setting in \cref{sec:online_background}. Then, we define the prior distribution using a diffusion model, with a set of $L$ consecutive \emph{unknown latent parameters} $\psi_{\ell} \in \real^d$ for $\ell \in [L]$. Precisely, the action parameter $\theta_a$ depends on the $1$-st latent parameter $\psi_{L}$ as $\theta_a \mid  \psi_{1} \sim \cN(f_1(\psi_{1}), \Sigma_1)$, where the \emph{link function} $f_1$ and covariance $\Sigma_1$ are \emph{known}. Also, the $\ell-1$-th latent parameter $\psi_{\ell-1}$ depends on the $\ell$-th latent parameter $\psi_{\ell}$ as $\psi_{\ell-1} \mid \psi_{\ell} \sim \cN(f_\ell(\psi_{\ell}), \Sigma_\ell)$, where $f_\ell$ and $\Sigma_\ell$ are \emph{known}. Finally, the $L$-th latent parameter $\psi_{L}$ is sampled as $\psi_{L} \sim \cN(0, \Sigma_{L+1})$, where $\Sigma_{L+1}$ is \emph{known}. We summarize this model in \cref{eq:dts_model} below:
\begin{align}\label{eq:dts_model}
  \psi_{L} & \sim \cN(0, \Sigma_{L+1})\,, \\
  \psi_{\ell-1} \mid \psi_{\ell} & \sim \cN(f_\ell(\psi_{\ell}), \Sigma_\ell)\,, & \forall \ell \in [L]/\{1\} \,, \nonumber\\
  \theta_a \mid \psi_{1} & \sim \cN(f_1(\psi_{1}), \Sigma_1)\,, & \forall a \in \cA \,, \nonumber\\
  R_t \mid \theta, (\psi_\ell)_{\ell \in [L]}, X_t, A_t &\sim p(\cdot \mid X_t; \theta_{A_t})\,, &  \forall t \in [T]\,. \nonumber
\end{align}
In practice, this model can be built by pre-training a diffusion model on offline estimates of the action parameters $\theta_a$.

\begin{remark}[\textbf{Joint models}]\label{remark}
Our algorithm and analysis also apply to the case where all actions share a single unknown parameter $\theta \in \real^d$.  
Let $\phi: \cX \times [K] \to \real^d$ be a known feature map, and assume the reward distribution mean is 
$g \left(\phi(x,a)^\top\theta\right)$.
Then, the diffusion prior in \cref{eq:dts_model} specializes by replacing the per-action parameters $(\theta_a)_{a \in [K]}$ with a single shared parameter $\theta$:
\begin{align}\label{eq:dts_model-shared-theta}
  \psi_L &\sim \cN(0,\Sigma_{L+1}), \\
  \psi_{\ell-1}\mid\psi_\ell &\sim \cN \left(f_\ell(\psi_\ell),\Sigma_\ell\right),
  && \forall\,\ell\in[L]\setminus\{1\},\nonumber\\
  \theta \mid \psi_1 &\sim \cN \left(f_1(\psi_1),\Sigma_1\right),\nonumber\\
  R_t \mid \theta, (\psi_\ell)_{\ell \in [L]}, X_t, A_t &\sim 
  p \left(\,\cdot\,\middle|\,\phi(X_t,A_t)^\top \theta\right),
  && \forall\,t\in[T].\nonumber
\end{align}
This formulation is useful when a shared feature map $\phi$ is available. 
In that case, the diffusion model can be pre-trained on parameters $\{\theta_s\}_{s=1}^S$ from previous tasks, and \dTS\ can then be applied to a new task $S{+}1$ using the pre-trained prior. 
To avoid clutter, our main exposition focuses on the model in \cref{eq:dts_model}, but all theoretical results and algorithmic components extend naturally to this shared-parameter case, which we also include in some experiments (explicitly noted when applicable).
\end{remark}

\section{Algorithm}
\label{sec:dts_algorithm}
We design a Thompson sampling algorithm that samples the latent and action parameters hierarchically \citep{lindley72bayes}. Let $H_t = (X_i, A_i, R_i)_{i\in [t-1]}$ denote the history of all interactions up to round $t$, and let $H_{t,a} = (X_i, A_i, R_i)_{\{i\in [t-1]; A_i=a\}}$ be the history of interactions \emph{with action $a$} up to round $t$. To motivate our algorithm, we decompose the posterior density $p(\theta_a \mid H_t)$ recursively as
\begin{align}\label{eq:dts_sampling_equivalence}
  p(\theta_a \mid H_t)
  = \int_{\psi_{1:L}}
    p(\psi_L \mid H_t)
    \prod_{\ell=2}^{L}
    p(\psi_{\ell-1} \mid \psi_\ell, H_t)
    p(\theta_a \mid \psi_1, H_{t,a})
    \, \dif \psi_{1:L}.
\end{align}

\textbf{Hierarchical sampling.} This decomposition induces the following sampling procedure. 
First, draw a sample $\psi_{t,L}$ according to the posterior density $p(\psi_L \mid H_t)$. 
Then, for each $\ell \in [L]\setminus\{1\}$, draw $\psi_{t,\ell-1}$ from the conditional posterior $p(\psi_{\ell-1} \mid \psi_{t,\ell}, H_t)$. Finally, given $\psi_{t,1}$, draw each action parameter independently from $p(\theta_{a} \mid \psi_{t,1}, H_{t,a})$ (the $\theta_a$ are conditionally independent given $\psi_1$). 
This defines \cref{alg:dts_ts}, \textbf{d}iffusion \textbf{T}hompson \textbf{S}ampling (\dTS).

\textbf{Posterior components via recursion.} To implement \dTS, we provide a recursive scheme to express the required posteriors using known quantities. These expressions may not always admit closed forms and often require approximation. The conditional action-posterior can be written as
\begin{align}\label{eq:dts_action_posterior}
  p(\theta_a \mid \psi_1, H_{t,a})
  \propto
  \prod_{i \in S_{t,a}}
  p(R_i \mid X_i; \theta_a)
  \, \cN(\theta_a; f_1(\psi_1), \Sigma_1),
\end{align}
where $S_{t,a} = \{\ell \in [t-1] : A_\ell = a\}$ is the set of rounds in which action $a$ was selected. 

Now, we characterize the conditional latent-posteriors. Before we do so, we make the following notation clarification. With slight abuse of notation, $p(H_t \mid \psi_\ell)$ denotes the likelihood of the observations up to round $t$ given $\psi_\ell$: 
\begin{align*}
   p(H_t \mid \psi_\ell) = p((R_i)_{i<t} \mid (X_i)_{i<t}, (A_i)_{i<t}, \psi_\ell)
\end{align*}
With this notation in mind, for any $\ell \in [L]\setminus\{1\}$, the conditional latent-posterior is
\begin{align*}
  p(\psi_{\ell-1} \mid \psi_\ell, H_t)
  \propto
  p(H_t \mid \psi_{\ell-1})
  \, \cN(\psi_{\ell-1}; f_\ell(\psi_\ell), \Sigma_\ell),
\end{align*}
and the top-layer posterior is
\begin{align*}
  p(\psi_L \mid H_t)
  \propto
  p(H_t \mid \psi_L)
  \, \cN(\psi_L; 0, \Sigma_{L+1}).
\end{align*}
All terms above are known except the likelihoods $p(H_t \mid \psi_{\ell})$, which are computed recursively.  
The recursion starts with
\begin{align}\label{eq:dts_base_rec}
  p(H_t \mid \psi_1)
  = \prod_{a=1}^K
    \int_{\theta_a}
    \Bigg[
      \prod_{i \in S_{t,a}}
      p(R_i \mid X_i; \theta_a)
    \Bigg]
    \cN(\theta_a; f_1(\psi_1), \Sigma_1)
    \, \dif \theta_a,
\end{align}
and for $\ell \in [L]\setminus\{1\}$, proceeds as
\begin{align}\label{eq:dts_rec}
 p(H_t \mid \psi_{\ell})
  = \int_{\psi_{\ell-1}}
    p(H_t \mid \psi_{\ell-1})
    \, \cN(\psi_{\ell-1}; f_\ell(\psi_\ell), \Sigma_\ell)
    \, \dif \psi_{\ell-1}.
\end{align}

\begin{algorithm}
\caption{\dTS : \textbf{d}iffusion  \textbf{T}hompson \textbf{S}ampling}
\label{alg:dts_ts}
\textbf{Input:} Prior components $\{f_\ell, \Sigma_\ell\}_{\ell=1}^{L+1}$ and reward model $p$. \\
\For{$t = 1, \dots, T$}{
    Draw $\psi_{t,L}$ according to the posterior density $p(\psi_L \mid H_t)$ \\
    \For{$\ell = L, \ldots, 2$}{
        Draw $\psi_{t,\ell-1}$ according to $p(\psi_{\ell-1} \mid \psi_{t,\ell}, H_t)$
    }
    \For{$a = 1, \ldots, K$}{
        Draw $\theta_{t,a}$ according to $p(\theta_a \mid \psi_{t,1}, H_{t,a})$
    }
    Select action $A_t = \argmax_{a \in [K]} r(X_t, a; \theta_t)$, where $\theta_t = (\theta_{t,a})_{a \in [K]}$ \\
    Observe reward $R_t \sim p(\cdot \mid X_t; \theta_{*, A_t})$ and update the posteriors.
}
\end{algorithm}

All posterior expressions above use known quantities \((f_\ell, \Sigma_\ell, p(r \mid x; \theta))\). However, these expressions typically need to be approximated, except when the link functions \(f_\ell\) are linear and the reward distribution \(p(\cdot \mid x; \theta)\) is linear-Gaussian, where closed-form solutions can be obtained with careful derivations. These approximations are not trivial, and prior studies often rely on computationally intensive approximate sampling algorithms. In the following sections, we explain how we derive our efficient approximations which are motivated by the closed-form solutions of linear instances.

\subsection{Posterior Approximation}
\label{subsec:dts_dts_nonlin_posterior_derivation}
The reward distribution is parameterized as a generalized linear model (GLM) \citep{mccullagh89generalized}, 
which allows for non-linear rewards. In addition, the diffusion model itself is highly non-linear due to the link functions \(f_\ell\).  
These two sources of non-linearity make the posterior intractable, so we apply two layers of approximation:  
(i) a likelihood approximation to linearize the reward model, and  
(ii) a diffusion approximation to handle the non-linear hierarchy induced by the diffusion model prior.

\textbf{(i) Likelihood approximation.}
We use an approach similar to the Laplace approximation, but instead of approximating the entire posterior, 
we approximate only the likelihood by a Gaussian.  
Precisely, the reward distribution \(p(\cdot \mid x; \theta_a)\) belongs to the exponential family with mean function \(g\). Thus
\begin{align}\label{eq:dts_likelihood_approxximation}
    \prod_{i \in S_{t,a}} p(R_i \mid X_i; \theta_a)
    \;\approx\;
    \cN \big(\theta_a; \hat{B}_{t,a}, \hat{G}_{t,a}^{-1}\big),
\end{align}
where \(\hat{B}_{t,a}\) is the maximum likelihood estimate and \(\hat{G}_{t,a}\) is the Hessian of the negative log-likelihood:
\begin{align}\label{eq:dts_glm_param}
    \hat{B}_{t,a} &= \argmax_{\theta_a \in \real^d} 
    \sum_{i \in S_{t,a}} \log p(R_i \mid X_i; \theta_a), 
    \qquad
    \hat{G}_{t,a} = \sum_{i \in S_{t,a}} \dot{g} \big(X_i^\top \hat{B}_{t,a}\big) X_i X_i^\top,
\end{align}
and \(S_{t,a} = \{\ell \in [t-1] : A_\ell = a\}\) is the set of rounds in which action \(a\) was selected. Of course, $\hat{G}_{t,a}$ might not be invertivle and thus we replace it by $\hat{G}_{t,a} + 10^{-3} I_d$ in practice. Unlike Laplace, which fits a global Gaussian to the full posterior, this step linearizes only the likelihood, thereby preserving the hierarchical diffusion structure of the prior.

\textbf{(ii) Diffusion approximation.}
Plugging the Gaussian likelihood approximation \eqref{eq:dts_likelihood_approxximation} into the posterior expressions 
\(p(\theta_a \mid \psi_1, H_{t,a})\) and \(p(\psi_{\ell-1} \mid \psi_\ell, H_t)\) 
removes the non-linearity of the reward model.  
However, the diffusion hierarchy remains non-linear through \(f_\ell\).  
To handle this, we build on the closed-form posteriors of the \emph{linear diffusion case} 
(where \(f_\ell(\psi_\ell)=\W_\ell \psi_\ell\); see \cref{sec:dts_posterior_proofs}) 
and generalize them by replacing the linear terms \(\W_\ell \psi_\ell\) 
with their non-linear counterparts \(f_\ell(\psi_\ell)\).  This substitution yields a \emph{posterior diffusion model} that retains the same hierarchical form as the prior but with data-dependent means and covariances for the conditional Gaussians. Details on how we transition from the linear to the general non-linear setting are provided in \cref{sec:dts_posterior_proofs,app:non_linear_explanation}. The resulting approximate posteriors admit the following closed-form expressions.

\textbf{Approximate action posterior.}
We approximate the conditional action posterior as
\[
p(\theta_a \mid \psi_1, H_{t,a}) 
\;\approx\; 
\cN \big(\theta_a; \hat{\mu}_{t,a}, \hat{\Sigma}_{t,a}\big),
\]
where
\begin{align}\label{eq:dts_non_linear_action_posterior}
    \hat{\Sigma}_{t,a}^{-1} &= 
    \underbrace{\Sigma_1^{-1}}_{\text{prior precision}}
    \;+\;
    \underbrace{\hat{G}_{t,a}}_{\text{data precision}},
    &\qquad
    \hat{\mu}_{t,a} &= 
    \hat{\Sigma}_{t,a} \Big(
        \underbrace{\Sigma_1^{-1} f_1(\psi_1)}_{\text{prior contribution}}
        \;+\;
        \underbrace{\hat{G}_{t,a}\hat{B}_{t,a}}_{\text{data contribution}}
    \Big).
\end{align}
This posterior update has a clear interpretation. The posterior precision $\hat{\Sigma}_{t,a}^{-1}$ is the sum of the prior precision and the \textit{data precision}.  
The posterior mean $\hat{\mu}_{t,a}$ is the precision-weighted average of the prior mean and the MLE \(\hat{B}_{t,a}\).  
As more data are observed, the covariance shrinks and the mean moves from the prior mean \(f_1(\psi_1)\) toward the MLE \(\hat{B}_{t,a}\).  
When no data are available (\(\hat{G}_{t,a}=0\)), the posterior reduces to the prior \(\cN(f_1(\psi_1), \Sigma_1)\);  
in the limit of infinite data (\(\hat{G}_{t,a} \to \infty\)), the posterior collapses to the MLE \(\hat{B}_{t,a}\), with  
\(\hat{\mu}_{t,a} \to \hat{B}_{t,a}\) and \(\hat{\Sigma}_{t,a} \to 0\).

\textbf{Approximate latent posteriors.}
For each $\ell \in [L+1]\setminus\{1\}$, we approximate the latent posterior as
\[
p(\psi_{\ell-1} \mid \psi_\ell, H_t)
\;\approx\;
\cN \big(\psi_{\ell-1}; \bar{\mu}_{t,\ell-1}, \bar{\Sigma}_{t,\ell-1}\big),
\]
with
\begin{align}\label{eq:dts_non_linear_hyper_posteriors}
    \bar{\Sigma}_{t,\ell-1}^{-1} =
    \underbrace{\Sigma_\ell^{-1}}_{\text{prior precision}}
    +
    \underbrace{\bar{G}_{t,\ell-1}}_{\text{data precision}},
    \;
    \bar{\mu}_{t,\ell-1} =
    \bar{\Sigma}_{t,\ell-1} \Big(
        \underbrace{\Sigma_\ell^{-1} f_\ell(\psi_\ell)}_{\text{prior contribution}}
        +
        \underbrace{\bar{B}_{t,\ell-1}}_{\text{data contribution}}
    \Big)\,,
\end{align}
where, by convention, $f_{L+1}(\psi_{L+1}) = 0$ since the top layer $\psi_L$ has no parent $\psi_{L+1}$. The quantities $\bar{G}_{t,\ell}$ and $\bar{B}_{t,\ell}$ are computed recursively.  
The base recursion is
\begin{align}\label{eq:dts_non_linear_basis_gaussian_recursive}
    \bar{G}_{t,1} &= \sum_{a=1}^K 
    \big(\Sigma_1^{-1} - \Sigma_1^{-1} \hat{\Sigma}_{t,a} \Sigma_1^{-1}\big),
    &\qquad
    \bar{B}_{t,1} &= \Sigma_1^{-1} 
    \sum_{a=1}^K \hat{\Sigma}_{t,a} \hat{G}_{t,a} \hat{B}_{t,a},
\end{align}
and for each $\ell \in [L]\setminus\{1\}$,
\begin{align}\label{eq:dts_non_linear_step_gaussian_recursive}
    \bar{G}_{t,\ell} &= \Sigma_\ell^{-1} - \Sigma_\ell^{-1}\bar{\Sigma}_{t,\ell-1}\Sigma_\ell^{-1},
    &\qquad
    \bar{B}_{t,\ell} &= \Sigma_\ell^{-1}\bar{\Sigma}_{t,\ell-1}\bar{B}_{t,\ell-1}.
\end{align}

The latent posterior update in \cref{eq:dts_non_linear_hyper_posteriors} has the same structure as the action posterior.  
The posterior precision $\bar{\Sigma}_{t,\ell-1}^{-1}$ is the sum of the prior and data precisions ,  
and the posterior mean is their precision-weighted combination.  
The data terms $\bar{G}_{t,\ell-1}$ and $\bar{B}_{t,\ell-1}$ are computed recursively 
(\cref{eq:dts_non_linear_basis_gaussian_recursive,eq:dts_non_linear_step_gaussian_recursive}),  
so information collected at the action level propagates upward through the hierarchy.

\textbf{Interpretation.}
The resulting approximate posterior remains a diffusion model whose conditional Gaussians have updated, data-dependent means and covariances.  
The latent-posterior means can be viewed as \emph{refined link functions}:
\[
\hat{f}_{t,\ell}(\psi_\ell) = 
\bar{\mu}_{t,\ell-1}
=
\bar{\Sigma}_{t,\ell-1} \left(\Sigma_\ell^{-1} f_\ell(\psi_\ell) + \bar{B}_{t,\ell-1}\right),
\]
and $\bar{\Sigma}_{t,\ell}$ represents their updated uncertainty.  
Both are updated with data: covariances contract as uncertainty decreases, and means move from the prior toward the MLE.  
Unlike a full Laplace approximation, this formulation preserves the expressiveness of the posterior rather than replacing it globally with a single Gaussian, while also avoiding the heavy computation required by other approximate inference methods.

\subsection{Extension to Joint Reward Models}\label{remark_section}

For the shared-parameter model in \cref{remark}, \dTS's posterior approximations are similar. The action posterior is $p(\theta \mid \psi_1, H_t)
\approx 
\cN(\hat{\mu}_t, \hat{\Sigma}_t)
$, where 
\begin{align}
&\hat{\Sigma}_t^{-1} = \Sigma_1^{-1} + \hat{G}_t\,, &
\hat{\mu}_t = \hat{\Sigma}_t \big(\Sigma_1^{-1} f_1(\psi_1) + \hat{G}_t \hat{B}_t\big).
\end{align}
where
\begin{align*}
&\hat{B}_t = \argmax_{\theta \in \mathbb{R}^d} 
\sum_{i<t} \log p \left(R_i \mid \phi(X_i,A_i)^\top \theta\right), &
\hat{G}_t = \sum_{i<t} \dot{g} \big(\phi(X_i,A_i)^\top \hat{B}_t\big)
\, \phi(X_i,A_i)\phi(X_i,A_i)^\top.
\end{align*}
Similarly, for $\ell \in [L+1]\setminus\{1\}$, the latent posterior is $p(\psi_{\ell-1} \mid \psi_\ell, H_t)
\approx
\cN(\bar{\mu}_{t,\ell-1}, \bar{\Sigma}_{t,\ell-1})$, where
\begin{align}
\bar{\Sigma}_{t,\ell-1}^{-1}
&= \Sigma_\ell^{-1} + \bar{G}_{t,\ell-1},
\qquad
\bar{\mu}_{t,\ell-1}
= \bar{\Sigma}_{t,\ell-1} \big(\Sigma_\ell^{-1} f_\ell(\psi_\ell) + \bar{B}_{t,\ell-1}\big)\,,
\end{align}
where, by convention, $f_{L+1}(\psi_{L+1}) = 0$ and the quantities $\bar{G}_{t,\ell}$ and $\bar{B}_{t,\ell}$ are computed recursively as
\begin{align}
\text{Base case:}& &\bar{G}_{t,1} = \Sigma_1^{-1} - \Sigma_1^{-1}\hat{\Sigma}_t\Sigma_1^{-1},
\qquad
\bar{B}_{t,1} = \Sigma_1^{-1}\hat{\Sigma}_t\hat{G}_t\hat{B}_t.\\
\text{Recursive case:}& &\bar{G}_{t,\ell} = \Sigma_\ell^{-1} - \Sigma_\ell^{-1}\bar{\Sigma}_{t,\ell-1}\Sigma_\ell^{-1},
\qquad
\bar{B}_{t,\ell} = \Sigma_\ell^{-1}\bar{\Sigma}_{t,\ell-1}\bar{B}_{t,\ell-1}.
\end{align}
Again, this shared-parameter variant of \dTS is presented for completeness and to illustrate the generality of our posterior derivations; the main focus of the chapter remains on the per-action disjoint formulation in \cref{eq:dts_model}. Unless stated otherwise, all theoretical results and experiments use the main version of \dTS described in \cref{alg:dts_ts}.

\section{Analysis}\label{sec:dts_infomal}
In this section, we present an informal Bayes regret analysis of \dTS to build intuition around \dTS's Bayesian regret scaling with problem parameters $d$, $K$, $L$, etc. This analysis is informal for two reasons. First, we analyze a simplified linear-Gaussian setting rather than the general nonlinear case on which we focus in this chapter: the reward distribution is linear-Gaussian and each link function $f_\ell(\psi_\ell) = \W_\ell \psi_\ell$ is a known linear mapping, inducing a hierarchy of $L$ linear-Gaussian layers from the latent root to the action parameters. 

Second, we assume the model is well-specified (similar to \cref{chap:las}): the true action parameters are generated according to the diffusion prior used by \dTS. Under these assumptions, the posterior becomes exact, enabling an analysis analogous to that used in \cref{chap:meTS}. However, our recursive hierarchical structure introduces technical differences: posteriors must be derived inductively using total covariance decompositions, and regret bounds require tracking information flow across all latent layers. We emphasize that this regret bound does not extend to the general nonlinear case studied in experiments; it is included here solely to provide theoretical intuition under simplifying assumptions. Formal statements and derivations are provided in \cref{sec:dts_analysis,sec:dts_regret_proof}.

\textbf{Bayes regret bound.} The bound of \dTS in this case is 
\begin{align*}
    \mathcal{B}\mathcal{R}(T) =\tilde{\mathcal{O}}\Big(\sqrt{T  (dK \sigma_1^2 + d \sum_{\ell=1}^L \sigma_{\ell+1}^2\sigma_{\textsc{max}}^{2 \ell} )}\Big)\,,
\end{align*}
where $\sigma_{\textsc{max}}^2 = \max_{\ell \in [L+1]} 1 + \frac{\sigma_\ell^2}{\sigma^2}$. This can be re-written as $ \mathcal{BR}(T) = \tilde{\mathcal{O}}(\sqrt{T d K_{\mathrm{eff}} \sum_{\ell=1}^{L+1} \sigma_{\ell}^2})$, where $K_{\mathrm{eff}} = \frac{K \sigma_1^2 + \sum_{\ell=1}^{L} \sigma_{\ell+1}^2 \sigma_{\textsc{max}}^{2\ell}}{\sum_{\ell=1}^{L+1} \sigma_\ell^2}$ is the \emph{effective number of actions}. This dependence on the horizon $T$ aligns with prior Bayes regret bounds scaling with $T$. However, the bound comprises $L+1$ main terms. First, one relates to action parameters learning, conforming to a standard form \citep{lu19informationtheoretic}, while the $L$ remaining terms are associated with learning each of the latent parameters.

\textbf{Sparsity refinement.} If each mixing matrix exhibits column sparsity, that, $\W_\ell=(\bar \W_\ell,0_{d,d-d_\ell})$ with $d_\ell \ll  d$ active columns, then the bound becomes
\begin{align*}
    \mathcal{B}\mathcal{R}(T) = \tilde{\mathcal{O}}\Big(\sqrt{T  (dK \sigma_1^2 + \sum_{\ell=1}^L d_\ell \sigma_{\ell+1}^2\sigma_{\textsc{max}}^{2 \ell} )}\Big)\,.
\end{align*}
Hence, informative, \emph{sparse} priors can cut the cost of learning deep latent chains down from $d$ to $d_\ell$. As in \cref{chap:meTS}, a less informative prior (such as high variance) leads to a more challenging problem and thus a higher bound. Therefore, smaller values of $K$, $L$, $d$, $d_\ell$ translate to fewer parameters to learn, leading to lower regret. The regret also decreases when the initial variances $\sigma_\ell^2$ decrease. These dependencies are common in Bayesian analysis, and empirical results match them. 

\textbf{Dependence on $K$.}
The reader may question why our bound depends on $K$. This dependence arises from two modeling choices. First, we study the \emph{disjoint} (per-action) setting $r(x, a; \theta) = x^\top \theta_a$, where $\theta = (\theta_a)_{a \in [K]} \in \mathbb{R}^{dK}$, requiring the learning of $Kd$ parameters. Second, we model the relationship between $\theta_a$ and $\psi_1$ stochastically as $\mathcal{N}(W_1 \psi_1, \sigma_1^2 I_d)$ to accommodate potential nonlinearity. While this choice confers robustness to model misspecification, it introduces additional uncertainty and requires learning both the action parameters $\theta_a$ and the latent parameters $\psi_\ell$, resulting in a bound that depends on both $K$ and $L$.

Despite this dependence, \dTS enjoys two key advantages. First, the regret scales with $K\sigma_1^2$ rather than $K\sum_{\ell}\sigma_\ell^2$, which is particularly beneficial when $\sigma_1$ is small, as is often the case with diffusion model priors. Second, thanks to informative priors, our bound has significantly smaller constants compared to both the Bayesian and frequentist regret bounds for \texttt{LinTS}. We demonstrate this empirically in \cref{app:bound_comparison} and provide a theoretical comparison in \cref{subsec:dts_dts_discussion}. Both analyses confirm that \dTS's advantage over \texttt{LinTS} increases as the action space grows.

\textbf{Can regret be independent of $K$?} Prior works \citep{foster2020adapting,xu2020upper,zhu2022contextual} have proposed bandit algorithms whose regret does not scale with $K$. However, these results apply to the \emph{shared-parameter} setting $r(x, a; \theta) = \phi(x, a)^\top \theta$, where only a single $d$-dimensional parameter must be learned, but this formulation requires access to a suitable feature map $\phi$. \dTS is compatible with this setting (\cref{remark_section}), in which case its regret would indeed be independent of $K$. Alternatively, even in the disjoint per-action case considered in this chapter, setting $\sigma_1 = 0$ would yield a $K$-independent regret bound. However, we believe this assumption is unrealistic in practice and would compromise the robustness of \dTS to model misspecification.

\subsection{Benefits} \label{subsec:dts_dts_discussion}

\textbf{Computational benefits.} Action correlations prompt an intuitive approach: marginalize all latent parameters and maintain a joint posterior of $(\theta_{a})_{a \in [K]} \mid H_t$. Unfortunately, this is computationally inefficient for large action spaces. To illustrate, suppose that all posteriors are multivariate Gaussians. Then maintaining the joint posterior $(\theta_{a})_{a \in [K]} \mid H_t$ necessitates converting and storing its $dK \times dK$-dimensional covariance matrix, leading to $\mathcal{O}(K^3 d^3)$ and $\mathcal{O}(K^2 d^2)$ time and space complexities. In contrast, the time and space complexities of \dTS are $\mathcal{O}\big(\big( L + K\big) d^3\big)$ and $\mathcal{O}\big(\big( L + K\big) d^2\big)$. This is because \dTS requires converting and storing $L+K$ covariance matrices, each being $d \times d$-dimensional. The improvement is huge when $K \gg L$, which is common in practice. Certainly, a more straightforward way to enhance computational efficiency is to discard latent parameters and maintain $K$ individual posteriors, each relating to an action parameter $\theta_{a} \in \mathbb{R}^{d}$ (\lints). This improves time and space complexity to $\mathcal{O}\big( K d^3\big)$ and $\mathcal{O}\big( K d^2\big)$. However, \lints maintains independent posteriors and fails to capture the correlations among actions; it only models $\theta_{a} \mid H_{t, a}$ rather than $\theta_{a} \mid H_{t}$ as done by \dTS. Consequently, \lints incurs higher regret due to the information loss caused by unused interactions of similar actions. Our regret bound and empirical results reflect this aspect.

\textbf{Statistical benefits.} We argue that our bound reflects the overall structure of the problem by comparing \dTS to algorithms that only partially use the structure or do not use it at all as follows. Precisely, when the link functions are linear, we can transform the diffusion prior into a Bayesian linear model (\lints) by marginalizing out the latent parameters; in which case the prior on action parameters becomes $\theta_{a} \sim \cN( 0 , \,  \Sigma)$, with the $\theta_{a}$ being not necessarily independent, and $\Sigma$ is the marginal initial covariance of action parameters and it writes $\Sigma = \sigma_1^{2} I_d +  \sum_{\ell=1}^L \sigma_{\ell+1}^2  \Beta_\ell \Beta_\ell^\top$ with $\Beta_\ell = \prod_{i=1}^\ell\W_i$. Then, it is tempting to directly apply \lints to solve our problem. This approach will induce higher regret because the additional uncertainty of the latent parameters is accounted for in $\Sigma$ despite integrating them. This causes the \emph{marginal} action uncertainty $\Sigma$ to be much higher than the \emph{conditional} action uncertainty $\sigma_1^2 I_d$, since we have $\Sigma = \sigma_1^{2} I_d +  \sum_{\ell=1}^L \sigma_{\ell+1}^2  \Beta_\ell \Beta_\ell^\top \succcurlyeq \sigma_1^2 I_d$. This discrepancy leads to higher regret, especially when $K$ is large. This is due to \lints needing to learn $K$ independent $d$-dimensional parameters, each with a considerably higher initial covariance $\Sigma$. This is also reflected by our regret bound. To simply comparisons, suppose that $\sigma \geq \max_{\ell \in [L+1]} \sigma_\ell$ so that $\sigma_{\textsc{max}}^2 \leq 2$. Then the regret bounds of \dTS (where we bound $\sigma_{\textsc{max}}^{2 \ell}$ by $2^\ell$) and \lints read
\begin{align*}
  &\dTS: \tilde{\mathcal{O}}\big(\sqrt{T  (dK \sigma_1^2 + \sum_{\ell=1}^L d_\ell \sigma_{\ell+1}^2 2^\ell)}\big)\,, \qquad \lints: \tilde{\mathcal{O}}\big(\sqrt{T d K (\sigma_1^2 + \sum_{\ell=1}^L \sigma_{\ell+1}^2 )}\big)\,.
\end{align*}
Then regret improvements are captured by the variances $\sigma_\ell$ and the sparsity dimensions $d_\ell$, and we proceed to illustrate this through the following scenarios. 

\textbf{(I) Decreasing variances.} Assume that $\sigma_\ell=2^\ell$ for any $\ell \in [L+1]$. Then, the regrets become
\begin{align*}
  &\dTS: \tilde{\mathcal{O}}\big(\sqrt{T  (dK + \sum_{\ell=1}^L d_\ell 4^\ell))}\big)\,, \qquad \lints: \tilde{\mathcal{O}}\big(\sqrt{T d K 2^L)}\big)
\end{align*}
Now to see the order of gain, assume the problem is high-dimensional ($d \gg 1$), and set $L = \log_2(d)$ and $d_\ell = \lfloor  \frac{d}{2^\ell} \rfloor$. Then the regret of \dTS becomes $\tilde{\mathcal{O}}\big(\sqrt{nd(K + L))}\big)$, and hence the multiplicative factor $2^L$ in \lints is removed and replaced with a smaller additive factor $L$.

\textbf{(II) Constant variances.} Assume that $\sigma_\ell=1$ for any $\ell \in [L+1]$. Then, the regrets become
\begin{align*}
  \dTS: \tilde{\mathcal{O}}\big(\sqrt{T  (dK + \sum_{\ell=1}^L d_\ell 2^\ell))}\big)\,, \qquad 
  \lints: \tilde{\mathcal{O}}\big(\sqrt{T d K L)}\big)
\end{align*}
Similarly, let $L = \log_2(d)$, and $d_\ell = \lfloor  \frac{d}{2^\ell} \rfloor$. Then \dTS's regret is $\tilde{\mathcal{O}}\big(\sqrt{T d (K + L)}\big)$. Thus the multiplicative factor $L$ in \lints is removed and replaced with the additive factor $L$. By comparing this to \textbf{(I)}, the gain with decreasing variances is greater than with constant ones. In general, diffusion models use decreasing variances \citep{ho2020denoising} and hence we expect great gains in practice. All observed improvements in this section could become even more pronounced when employing non-linear diffusion models. In our theory, we used linear diffusion models, and yet we can already discern substantial differences. Moreover, under non-linear diffusion \cref{eq:dts_model}, the latent parameters cannot be analytically marginalized, making \lints with exact marginalization inapplicable.

\section{Experiments}\label{sec:dts_experiments}
\paragraph{Experimental setup.}
We evaluate \dTS using both synthetic and MovieLens problems. In our experiments, we run 50 random simulations and plot the average regret with standard error. Our main contribution is to demonstrate that pretraining a diffusion model offline enables the construction of expressive and informative priors that substantially improve exploration efficiency in contextual bandits. We first evaluate \dTS in a setting where the prior matches the true generative process (\cref{sec:dts_well_specified} to isolate the benefit of informative priors), and then consider a misspecified regime (\cref{subsec:dts_dts_effect_pretraining} and \cref{app:exp}) where the prior is either trained on out-of-distribution data or intentionally perturbed. These experiments show that even when the prior is imperfect, \dTS maintains strong performance: highlighting its robustness and practical relevance.

\subsection{True Prior is a Diffusion Model}\label{sec:dts_well_specified}
Synthetic bandit problems are generated from the diffusion model in \cref{eq:dts_model} with both linear and non-linear rewards. Linear rewards follow $p(\cdot \mid x; \theta_{ a}) = \cN(x^\top \theta_{ a}, 1)$, while non-linear rewards are binary from $p(\cdot \mid x; \theta_{ a}) =\mathrm{Ber}(g(x^\top \theta_{ a}))$, with $g$ as the sigmoid function. Covariances are $\Sigma_\ell= I_d$, and contexts $X_t$ are uniformly drawn from $[-1, 1]^d$. We vary $d \in \{5, 20\}$, $L \in \{2, 4\}$, $K \in \{10^2, 10^4\}$, and set the horizon to $T=5000$, considering both linear and non-linear models.

\textbf{Linear diffusion.} We consider \cref{eq:dts_model} with $f_\ell(\psi)=\W_\ell \psi$, where $\W_\ell$ uniformly drawn from $[-1,1]^{d \times d}$. Sparsity is introduced by zeroing the last $d_\ell$ columns of $\W_\ell$ as $\W_\ell = (\bar{\W}_\ell, 0_{d, d-d_\ell})$. For $d=5$ and $L=2$, $(d_1,d_2)=(5,2)$; for $d=20$ and $L=4$, $(d_1,d_2,d_3,d_4)=(20, 10, 5,2)$.

\textbf{Non-linear diffusion.} We consider \cref{eq:dts_model} where $f_\ell$ are 2-layer neural networks with random weights in $[-1,1]$, \texttt{ReLU} activation, and hidden layers of size $h=20$ for $d=5$, and $h=60$ for $d=20$.

\begin{figure*}
  \centering
  \includegraphics[width=\linewidth]{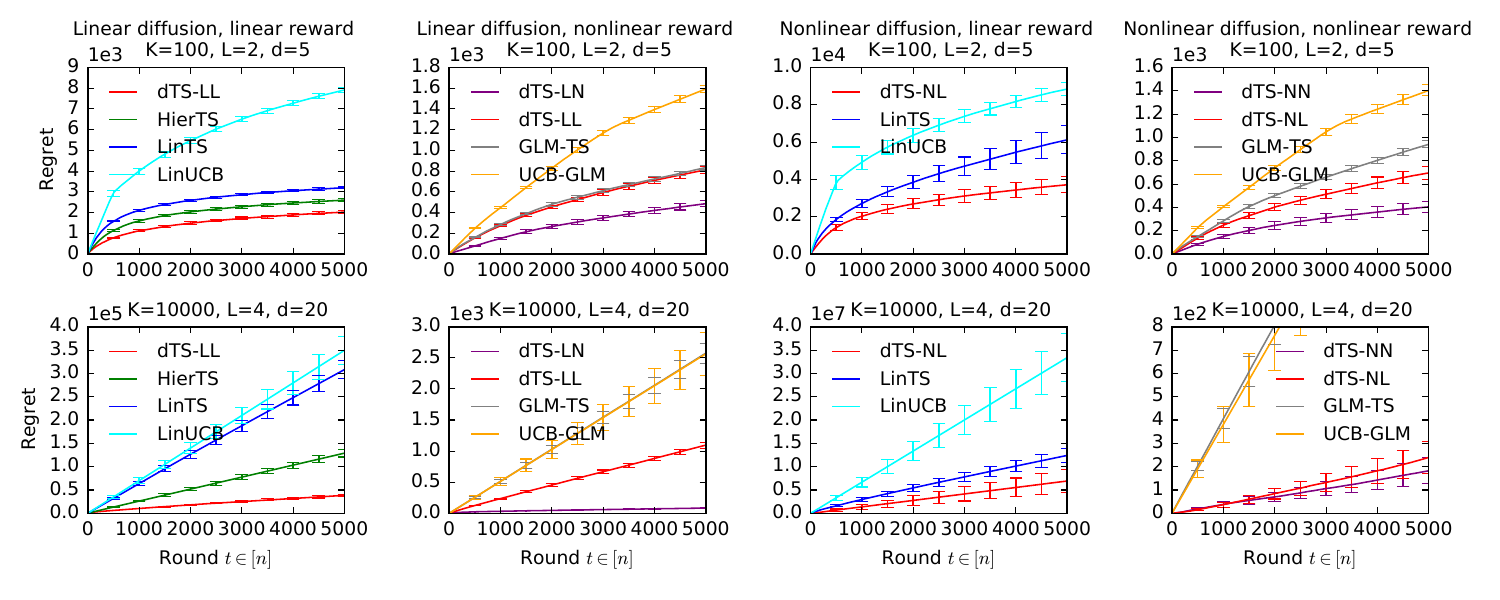}
  \caption{Regret of \dTS with varying diffusion and reward models and varying parameters $d$, $K$, $L$.} 
  \label{fig:dts_synthetic_regret}
\end{figure*}

\textbf{Baselines.} For linear rewards, we use \linucb \citep{abbasi2011improved}, \lints \citep{agrawal13thompson}, and \hierts \citep{hong22hierarchical}, marginalizing out all latent parameters except $\psi_{ L}$, which corresponds to \texttt{HierTS-1} in \cref{subsec:dts_dts_two_level_hierarchies}. For non-linear rewards, we include \ucbglm \citep{li17provably} and \glmts \citep{chapelle11empirical}. We exclude \glmucb \citep{filippi10parametric} due to high regret and \texttt{HierTS} as it’s designed for linear rewards. We name \dTS as \texttt{\dTS-dr}, where \texttt{d} refers to diffusion type (\texttt{L} for linear, \texttt{N} for non-linear) and \texttt{r} indicates reward type (\texttt{L} for linear, \texttt{N} for non-linear). For example, \texttt{\dTS-LL} signifies \dTS in linear diffusion with linear rewards.

\textbf{Results and interpretations.} Results are shown in \cref{fig:dts_synthetic_regret} and we make the following observations:

\textbf{1) \dTS demonstrates superior performance (\cref{fig:dts_synthetic_regret}).} \dTS consistently outperforms the baselines across all settings, including the four combinations of linear/non-linear diffusion and reward (columns in \cref{fig:dts_synthetic_regret}) and both bandit settings with varying $K$, $L$, and $d$ (rows in \cref{fig:dts_synthetic_regret}).

\textbf{2) Latent diffusion structure may be more important than the reward distribution.} When rewards are non-linear (second and fourth columns in \cref{fig:dts_synthetic_regret}), we include variants of \dTS that use the correct diffusion prior but the wrong reward distribution, applying linear-Gaussian instead of logistic-Bernoulli (\linalg in the second column and \texttt{\dTS-NL} in the fourth). Despite the reward misspecification, these variants outperform models using the correct reward distribution but ignoring the latent diffusion structure, such as \glmts and \ucbglm. This highlights the importance of accounting for latent structure, which can be more critical than an accurate reward distribution.

\begin{figure*}
     \centering
     \begin{subfigure}[b]{0.32\textwidth}
         \centering
         \includegraphics[width=\textwidth]{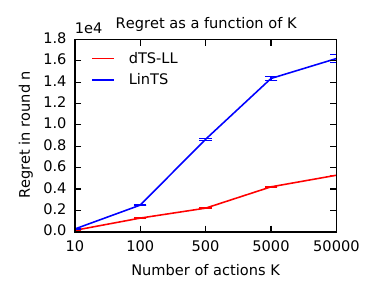}
         \caption{Perf. gap w.r.t. $K$.}\label{fig:dts_synthetic_las_regret}
     \end{subfigure}
     \hfill
     \begin{subfigure}[b]{0.32\textwidth}
         \centering
         \includegraphics[width=\textwidth]{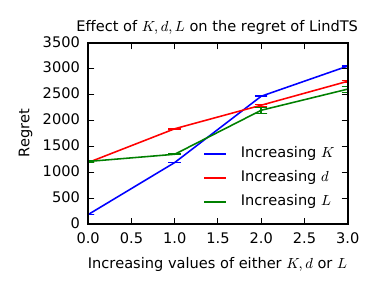}
         \caption{Scaling w.r.t. $K$, $d$, $L$.}\label{fig:dts_synthetic_varying}
     \end{subfigure}
     \hfill
     \begin{subfigure}[b]{0.32\textwidth}
         \centering
         \includegraphics[width=\textwidth]{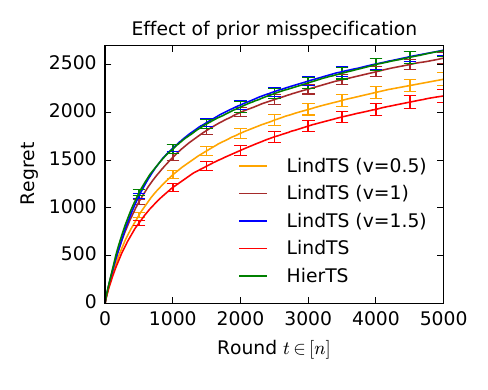}
         \caption{Prior misspecification.}  \label{fig:dts_prior_mis}
     \end{subfigure}
    \caption{Effect of various factors on \dTS's performance.}
\end{figure*}

\textbf{3) Performance gap between \dTS and \lints widens as $K$ increases (\cref{fig:dts_synthetic_las_regret}).} To show \dTS's improved scalability, we evaluate its performance with varying values of $K\in[10, 5 \times 10^4]$, in the linear diffusion and rewards setting. \cref{fig:dts_synthetic_las_regret} shows the final cumulative regret for varying $K$ values for both \texttt{\dTS-LL} and \texttt{\texttt{\texttt{LinTS}}}, revealing a widening performance gap as $K$ increases.

\textbf{4) Regret scaling with $K$, $d$ and $L$ matches our theory (\cref{fig:dts_synthetic_varying}).} We assess the effect of the number of actions $K$, context dimension $d$, and diffusion depth $L$ on \dTS's regret. Using the linear diffusion and rewards setting, for which we have derived a Bayes regret upper bound, we plot \linalg's regret across varying values of $K \in \set{10, 100, 500, 1000}$, $d \in \set{5, 10, 15, 20}$, and $L \in \set{2, 4, 5, 6}$ in \cref{fig:dts_synthetic_varying}. As predicted by our theory, the empirical regret increases with larger values of $K$, $d$, or $L$, as these make the learning problem more challenging, leading to higher regret.

\textbf{5) Diffusion prior misspecification (\cref{fig:dts_prior_mis}).} Here, \dTS's diffusion prior parameters differ from the true diffusion prior. In the linear diffusion and reward setting, we replace the true parameters $\W_\ell$ and $\Sigma_\ell$ with misspecified ones, $\W_\ell + \epsilon_1$ and $\Sigma_\ell + \epsilon_2$, where $\epsilon_1$ and $\epsilon_2$ are uniformly sampled from $[v, v+0.5]^{d \times d}$, with $v$ controlling the misspecification level. We vary $v \in \{0.5, 1, 1.5\}$ and assess \dTS's performance, comparing it to the well-specified \linalg and the strongest baseline in this fully-linear setting, \hierts. As shown in \cref{fig:dts_prior_mis}, \dTS's performance decreases with increasing misspecification but remains superior to the baseline, except at $v=1.5$, where their performances are comparable. Additional misspecification experiments are presented in \cref{subsec:dts_dts_effect_pretraining}, where the bandit environment is not sampled from a diffusion model.

\subsection{True Prior is Not a Diffusion Model}\label{subsec:dts_dts_effect_pretraining}
\begin{figure*}
     \centering
     \begin{subfigure}[b]{0.32\textwidth}
         \centering
         \includegraphics[width=\textwidth]{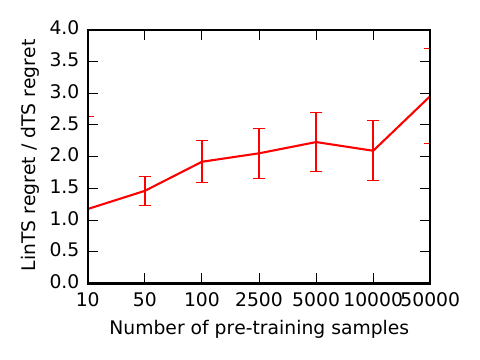}
         \caption{Ratio of \texttt{\texttt{LinTS}}/\dTS cumulative regret in the last round with varying pre-training sample size in $[10, 5\times 10^4]$. \textbf{Higher values mean a bigger performance gap}.}
         \label{fig:dts_effect_n}
     \end{subfigure}
     \hfill
          \begin{subfigure}[b]{0.32\textwidth}
         \centering
         \includegraphics[width=\textwidth]{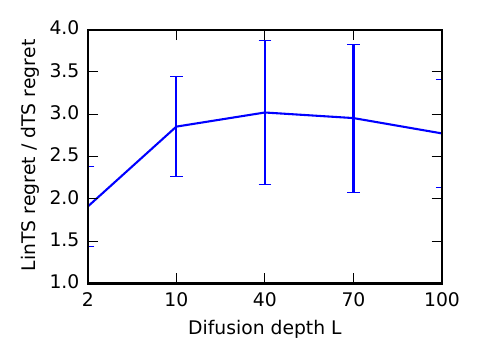}
         \caption{Ratio of \texttt{\texttt{LinTS}}/\dTS cumulative regret in the last round with varying diffusion depth $L$ in $[2, 100]$. \textbf{Higher values mean a bigger performance gap}.}
         \label{fig:dts_effect_L}
     \end{subfigure}
     \hfill
     \begin{subfigure}[b]{0.32\textwidth}
         \centering
         \includegraphics[width=\textwidth]{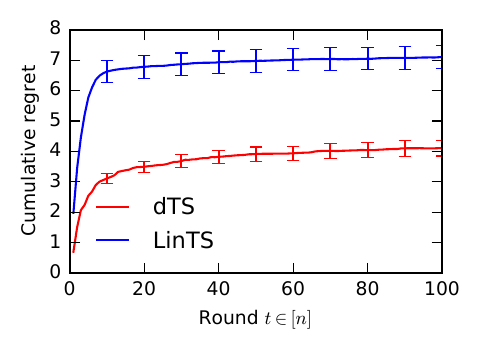}
         \caption{Regret of \dTS in \textbf{MovieLens}. The diffusion model with $L=40$ is pre-trained on embeddings obtained by low-rank factorization of MovieLens rating matrix.}\label{fig:dts_movielens}
     \end{subfigure}
    \caption{\textbf{(a) and (b):} Impact of pre-training sample size and diffusion depth $L$ for the \textbf{Swiss roll data}. \textbf{(c):} Regret of \dTS in \textbf{MovieLens}.}
        \label{fig:dts_three graphs}
\end{figure*}

\textbf{Swiss roll data.} Unlike previous experiments, the true action parameters are now sampled from the Swiss roll distribution (see \cref{fig:dts_swiss_roll_data} in \cref{app:swiss_roll_data}), rather than from a diffusion model. The diffusion model used by \dTS is pre-trained on samples from this distribution, with the offline pre-training procedure described in \cref{app:pretraining}. \cref{fig:dts_effect_n} shows that larger sample sizes increase the performance gap between \dTS and \texttt{\texttt{LinTS}}. More samples improve the estimation of the diffusion prior (see \cref{fig:dts_swiss_roll_data} in \cref{app:swiss_roll_data}), leading to better \dTS performance. Notably, comparable performance was achieved with as few as 10 samples, and \dTS outperformed \texttt{\texttt{LinTS}} by a factor of 1.5 with just 50 samples. While more samples may be required for more complex problems, \texttt{\texttt{LinTS}} would also struggle in such cases. Therefore, we expect these gains to be even more significant in more challenging settings. 

We studied the effect of the pre-trained diffusion model depth \(L\) and found that \(L \approx 40\) yields the best performance, with a drop beyond that point (\cref{fig:dts_effect_L}). While our theory doesn't apply directly here, as it assumes a linear diffusion model, it still offers some intuition on the decreased performance for \(L > 40\). The theorem shows \dTS's regret bound increases with \(L\) when the true distribution is a diffusion model. For small \(L\), the pre-trained model doesn't fully capture the true distribution, making the theorem inapplicable, but at \(L \approx 40\), the distribution is nearly captured, and further increases in \(L\) lead to higher regret, consistent with our theory.

\textbf{MovieLens data.} We also evaluate \dTS using the standard MovieLens \citep{movielens} setting. In this semi-synthetic experiment, a user is sampled from the rating matrix in each interaction round, and the reward is the rating the user gives to a movie (see \citet[Section 5]{clavier2023vits} for details about this setting). Here, the true distribution of action parameters is unknown and not a diffusion model. The diffusion model is pre-trained on offline estimates of action parameters obtained through low-rank factorization of the rating matrix. \cref{fig:dts_movielens} demonstrates that \dTS outperforms \texttt{LinTS} in this setting. Additional \textbf{CIFAR} ablations are provided in \cref{app:cifar} where similar strong improvements are observed.

\section{Conclusion}\label{sec:dts_conclusion}

We use a pre-trained diffusion model as a strong and flexible prior for \dTS. 
Diffusion pre-training leverages abundant offline data, which is then fine-tuned through online interactions via our tractable posterior approximation. 
This approximation enables efficient posterior sampling and updates while maintaining strong empirical performance. 
Moreover, \dTS admits a simple Bayesian regret bound in the linear–Gaussian setting.

Our work has several limitations. First, our Bayes regret analysis applies only to the linear-Gaussian setting with a well-specified prior; extending formal guarantees to nonlinear diffusion models remains open. Second, our posterior approximation, while motivated by exact solutions in the linear case, lacks theoretical justification for general nonlinear link functions: its strong empirical performance does not come with formal approximation error bounds. Finally, \dTS requires offline pre-training of the diffusion model, which assumes access to historical estimates of action parameters; in domains where such data is unavailable or expensive to obtain, the benefits of diffusion priors may not be realized.

%% file: contents/part2/introduction.tex
\chapter{Introduction to \cref{part:offline}}\label{intro:part2}

This second part of the thesis addresses the following fundamental question: 
\begin{center}
    \emph{How can we reliably learn high-performing policies from static logged data when the number of actions is large?}
\end{center}

\section{Setting and Background}\label{sec:offline_background}

In this part, we consider the off-policy (offline) setting where an agent is provided with a static logged dataset $\mathcal{D}_n = \{(X_i, A_i, R_i)\}_{i=1}^n$ collected by a logging policy $\pi_0$. The data collection process proceeds as follows: for each round $i \in [n]$:
\begin{enumerate}
    \item The environment draws a context $X_i \sim \nu$, where $\nu$ is a distribution with support $\mathcal{X}$ forming a compact subset of $\mathbb{R}^{d}$;
    \item The logging policy selects an action $A_i \sim \pi_0(\cdot \mid X_i)$ from the action set $\mathcal{A} = [K]$;
    \item The environment generates a stochastic reward $R_i \sim p(\cdot \mid X_i, A_i)$, where $R_i \in [0, 1]$.
\end{enumerate}

Unlike the on-policy setting of \cref{part:online}, no further interaction with the environment is permitted. The objective is to learn a new policy $\hat{\pi}$ from this static dataset that maximizes the true (but unknown) expected value:
\begin{align}\label{eq:offline_value}
    V(\pi) = \mathbb{E}_{X \sim \nu}\left[\mathbb{E}_{A \sim \pi(\cdot \mid X)}[r(X, A)]\right],
\end{align}
where $r(x, a) = \mathbb{E}_{R \sim p(\cdot \mid x, a)}[R]$ is the expected reward function. Performance is measured by the \emph{suboptimality gap} of the learned policy:
\begin{align}\label{eq:offline_subopt}
    \textsc{so}(\hat{\pi}) = V(\pi_*) - V(\hat{\pi}),
\end{align}
where $\pi_* = \arg\max_{\pi \in \Pi} V(\pi)$ is the optimal policy within a class $\Pi$.

Since $V(\pi)$ cannot be computed directly, off-policy learning algorithms often rely on an empirical estimate $\hat{V}(\pi)$ constructed from $\mathcal{D}_n$. This estimation task is known as \emph{off-policy evaluation (OPE)} in the literature. The two dominant estimation approaches are: direct method (DM) and inverse propensity scoring (IPS). DM employ a learned reward model $\hat{r}(x, a)$ to estimate the value as:
\begin{align}
    \hat{V}_{\textsc{dm}}(\pi) = \frac{1}{n} \sum_{i=1}^{n} \sum_{a \in \mathcal{A}} \pi(a \mid X_i) \, \hat{r}(X_i, a)\,.
\end{align}
IPS re-weights observed rewards using importance sampling as:
\begin{align}
    \hat{V}_{\textsc{ips}}(\pi) = \frac{1}{n} \sum_{i=1}^{n} \frac{\pi(A_i \mid X_i)}{\pi_0(A_i \mid X_i)} R_i.
\end{align}
Given an estimator $\hat{V}(\pi)$, the agent must then select a policy. This step distinguishes between \emph{greedy policies}, which directly maximize $\hat{\pi} = \arg\max_{\pi \in \Pi} \hat{V}(\pi)$, and \emph{pessimistic policies}, which incorporate an uncertainty penalty $\hat{\pi} = \arg\max_{\pi \in \Pi} [\hat{V}(\pi) - \operatorname{pen}(\pi)]$.

\subsection{Scalability Challenges}

When the number of actions $K$ is large, both estimation paradigms and their associated optimization procedures encounter fundamental obstacles:

\textbf{Statistical inefficiency of DM.} Standard DM approaches model each action's reward function independently. As $K$ grows, the data available per action diminishes, leading to poorly estimated reward functions and lower performance.

\textbf{High variance of importance sampling.} IPS's variance grows with the importance weights $\pi(a|x)/\pi_0(a|x)$. These weights explode in large action spaces, producing estimates too noisy for reliable optimization.

\textbf{Intractable optimization landscapes.} Beyond estimation challenges, the optimization problem $\arg\max_{\pi \in \Pi} \hat{V}(\pi)$ itself becomes computationally intractable in large action spaces. IPS-based objectives induce highly non-concave landscapes with exponentially many local maxima and flat plateaus that trap gradient-based optimizers. As we show in \cref{chap:las}, this optimization bottleneck often dominates estimation error, making even statistically superior estimators ineffective in practice.

\section{Methodological Approaches}

To address these challenges, the methods developed in this part pursue three complementary strategies: \emph{structured reward modeling} for sample-efficient DM, \emph{surrogate objectives} that prioritize optimization tractability over estimation accuracy, and \emph{principled regularization and pessimism} for importance-weighted estimators.

\textbf{Structured Bayesian models.} Drawing inspiration from the hierarchical framework of \cref{part:online}, we introduce latent structure into reward modeling. Action parameters are coupled through shared latent variables $\psi$:
\begin{align}\label{eq:offline_structured}
    \psi &\sim q(\cdot), \\
    \theta_a \mid \psi &\sim p_a(\cdot; f_a(\psi)), \quad \forall a \in \mathcal{A}, \nonumber \\
    R \mid X, A, \theta, \psi &\sim p(\cdot \mid X; \theta_A). \nonumber
\end{align}
This formulation enables information sharing across actions: observations from frequently selected actions inform the posterior over $\psi$, which in turn improves reward estimates for rarely observed actions.

\textbf{Optimization-aware objectives.} Rather than designing sophisticated value estimators and then optimizing them, we propose objectives designed primarily for favorable optimization landscapes. The \emph{policy-weighted log-likelihood (PWLL)} family:
\begin{align}
    \hat{U}_g(\pi) = \frac{1}{n} \sum_{i=1}^n g(R_i, \pi_0(A_i \mid X_i)) \log \pi(A_i \mid X_i)\,,
\end{align}
where $g$ is a positive weighting function, yields concave objectives for linear-softmax policies $\pi$. This guarantees efficient convergence to a unique global maximum, bypassing the optimization pathologies of value estimation altogether.

\textbf{Regularized importance weighting.} For practitioners committed to IPS-based methods, we develop variance-controlled estimators through importance weight regularization. The exponential smoothing estimator:
\begin{align}
    \hat{V}^\alpha(\pi) = \frac{1}{n} \sum_{i=1}^n \frac{\pi(A_i \mid X_i)}{\pi_0(A_i \mid X_i)^\alpha} R_i\,, \quad \alpha \in [0, 1]\,,
\end{align}
smoothly trades variance for bias while preserving differentiability. Combined with pessimistic optimization and PAC-Bayes generalization bounds, this yields principled, tractable objectives  that are amenable to stochastic gradient ascent for safe off-policy learning.

\section{Roadmap of \cref{part:offline}}

The following chapters develop these methodological approaches.

\paragraph{\cref{chap:sDM}: Scaling Direct Methods with Latent Parameters.}
We begin by addressing the statistical inefficiency of DM through structured Bayesian modeling. Building on the hierarchical framework of \cref{part:online}, we introduce the \emph{structured direct method} (\texttt{sDM}), which couples action parameters through a shared latent vector. The posterior over these latent variables aggregates evidence across all actions, enabling effective generalization to actions with sparse data coverage. We analyze performance through \emph{Bayesian suboptimality} and prove that greedy policies paired with \texttt{sDM} achieve $\mathcal{O}(1/\sqrt{n})$ convergence under mild assumptions on the alignment between logging and optimal policies.

\paragraph{\cref{chap:las}: Optimization Matters More Than Estimation.}
We then challenge the conventional paradigm of off-policy learning. Through theoretical analysis and large-scale experiments, we demonstrate that \emph{optimization error} dominates estimation error in large action spaces. Specifically, we prove that for any IPS-based estimator, gradient ascent can remain trapped in suboptimal regions for $\mathcal{O}(K)$ iterations, and that the optimization landscape contains exponentially many local maxima in $K$. We then propose \emph{objective-aware policy parametrizations}: by aligning the policy class with the estimator's inductive bias, we can partially mitigate these optimization challenges. However, for a more complete solution, we propose \emph{policy-weighted log-likelihood (PWLL)} objectives as an alternative to IPS-based objectives. These objectives are provably concave for linear softmax policies, guaranteeing efficient convergence to a global optimum. Experiments on datasets with up to one million actions validate that PWLL consistently outperforms state-of-the-art estimator-based methods.
\paragraph{\cref{chap:ES}: Principled Pessimism for Exponential Smoothing and Beyond.}
Finally, for practitioners committed to importance weighting methods, we develop a theoretically grounded framework for variance control and pessimistic policy learning. We propose \emph{exponential smoothing} estimators that regularize importance weights, trading controlled bias for reduced variance. To leverage these regularized estimators for safe policy learning, we derive two-sided PAC-Bayes generalization bounds where all quantities are empirical and differentiable. The pessimistic learning objective maximizes the lower bound on policy value, penalizing policies with high bias or variance and steering optimization toward reliable regions. This also yields tractable objectives amenable to stochastic gradient optimization. We further present a \emph{unified PAC-Bayes framework} covering the major importance weight regularization techniques in the literature (clipping, exponential smoothing, implicit exploration), enabling principled comparison and demonstrating that the choice of pessimistic objective often matters more than the specific regularizer.

%% file: contents/sdm/main_text.tex
In this chapter, we address the statistical inefficiency of direct methods (DMs) in large action spaces: the first challenge highlighted in \cref{intro:part2}. Standard DMs estimate independent $d$-dimensional parameters for each action. This approach becomes statistically inefficient in large action spaces, where data are collected by a logging policy that explores only a small subset of available actions, leaving many actions rarely or never observed. To overcome this limitation, we analyze DMs through a Bayesian lens and propose making them sample-efficient by incorporating informative priors.

We make the following contributions.
\begin{enumerate*}[label=\textbf{\arabic*)}]
    \item We introduce the \emph{structured direct method} (\alg), a Bayesian approach that uses informative priors to share reward information across actions. By updating beliefs about similar actions based on observed data, \alg improves statistical efficiency without compromising computational scalability.
    \item To evaluate \alg, we propose Bayesian metrics that assess the average performance across problem instances sampled from the prior. This departs from the standard frequentist focus on worst-case scenarios. These metrics formally quantify the benefits of informative priors.
    \item Our theoretical analysis of Bayesian suboptimality (BSO) reveals two key insights: (1) performance degrades gracefully even without the standard assumption of full logging support, and (2) greedy policies are provably optimal under the BSO metric, standing in contrast to the pessimistic policies typically favored in frequentist settings.
    \item We empirically validate \alg and our theoretical findings using both synthetic and real-world datasets.
\end{enumerate*}

\section{Setting}
\label{sec:sdm-setting}

We consider the setting described in \cref{sec:offline_background}. The only additional assumption is the existence of \emph{unknown true parameters} $\theta_{*, a} \in \mathbb{R}^d$ for each action $a$, such that rewards are distributed as $R_i \sim p(\cdot \mid X_i; \theta_{*, A_i})$. Let $\theta_* = (\theta_{*, a})_{a \in \mathcal{A}} \in \mathbb{R}^{dK}$ denote the concatenation of all action parameters. The \emph{reward function} $r(x, a; \theta_*) = \mathbb{E}_{R \sim p(\cdot \mid x; \theta_{*, a})}[R]$ gives the expected reward of action $a$ in context $x$. The goal is to find a policy $\pi \in \Pi$ that maximizes:
$$V(\pi; \theta_*) = \mathbb{E}_{X \sim \nu} \mathbb{E}_{A \sim \pi(\cdot \mid X)}[r(X, A; \theta_*)]\,.$$

This chapter focuses on DM that estimates the value $V(\pi; \theta_*)$ as:
\begin{align}\label{eq:sdm-dm_policy_value}
    \hat{V}_{\textsc{dm}}(\pi) &= \frac{1}{n} \sum_{i \in [n]} \sum_{a \in \cA} \pi(a \mid  X_i) \hat{r}(X_i, a)\,,
\end{align}
where $\hat{r}(x, a)$ is an estimation of $r(x, a; \theta_*)$. DM estimators may exhibit modeling bias, but they generally have lower variance than IPS \citep{saito2022off}. Another advantage of DM is its practical utility without assuming access to the logging policy $\pi_0$ \citep{jeunen2021pessimistic,aouali2022probabilistic,hong2023multi}. Also, DMs can be incorporated into a Bayesian framework, where informative priors can be used to enhance statistical efficiency. This allows for the development of scalable methods suitable for large action spaces, as shown in our work.

\section{Structured DM}\label{sec:sdm-model}
\subsection{Structured Priors}
\textbf{Pitfalls of non-structured priors.} Before presenting \alg, we first describe the pitfalls of using the following widely used standard prior,
\begin{align}\label{eq:sdm-basic_model}
  \theta_a &\sim \cN(\mu_{a}, \Sigma_{a})\,, & \forall a \in \cA\,,\\
    R \mid \theta, X, A &\sim \cN(\phi(X)^\top \theta_A , \sigma^2)\,,\nonumber
\end{align}
where \( \phi(x) \) provides a \( d \)-dimensional representation of the context \( x \in \mathcal{X} \), and \( \mathcal{N}(\mu_{a}, \Sigma_{a}) \) represents the prior density of \( \theta_a \), with \( \sigma^2 \) being the reward noise variance. Under this prior, each action \( a \) has an associated parameter \( \theta_a \). Given the prior in \cref{eq:sdm-basic_model}, the posterior distribution of an action parameter follows a multivariate Gaussian: \( \theta_a \mid \cD_n \sim \mathcal{N}(\hat{\mu}_a, \hat{\Sigma}_a) \), where 
\begin{align*}
    &\hat{\Sigma}_{a}^{-1} = \Sigma_{a}^{-1} + G_{a}\,, & \hat{\Sigma}_{a}^{-1} \hat{\mu}_{a} = \Sigma_{a}^{-1} \mu_a + B_a\,.
\end{align*}
with
\begin{align*}
    &G_{a} = \sigma^{-2} \sum_{i \in [n]} \mathds{1}_{\{A_i=a\}} \phi(X_i) \phi(X_i)^\top\,, & B_{a} = \sigma^{-2} \sum_{i \in [n]} \mathds{1}_{\{A_i=a\}} R_i \phi(X_i)
\end{align*}
Note that \( G_a \) and \( B_a \) only use the subset of samples \( \cD_n \) where action \( a \) was observed, meaning data from other actions \( b \neq a \) do not contribute to the posterior inference for action \( a \). This results in statistical inefficiency, especially if the logged data \( \cD_n \) doesn't cover all actions. In particular, the posterior for an unseen action \( a \), \( \theta_a \mid \cD_n \), would simply revert to the prior \( \mathcal{N}(\mu_a, \Sigma_a) \), since we would have \( G_a = 0_{d \times d} \) and \( B_a = 0_d \) in such case.

\textbf{Structured priors.} To address the above issue, we assume that action rewards correlate and embed this knowledge into the prior. While one could model these correlations by considering the joint posterior distribution of \( (\theta_a)_{a \in \mathcal{A}} \mid \cD_n \), this becomes computationally burdensome when the number of actions \( K \) is large. Instead, we introduce an \emph{unknown $d^\prime$-dimensional latent parameter} \( \psi \in \mathbb{R}^{d^\prime} \), sampled from a \emph{latent prior} \( q(\cdot) \), such as \( \psi \sim q(\cdot) \). The correlations between actions naturally arise because each action parameter \( \theta_a \) is derived from the same latent parameter \( \psi \).

Specifically, the action parameters \( \theta_a \) are conditionally independent given \( \psi \) and are sampled from a \emph{conditional prior} \( p_a \) as \( \theta_a \mid \psi \sim p_a(\cdot ; f_a(\psi)) \) for all \( a \in \mathcal{A} \). Here, \( p_a \) is parameterized by \( f_a(\psi) \), where \( f_a: \mathbb{R}^{d^\prime} \rightarrow \mathbb{R}^d \) is a known prior function that encodes the hierarchical relationship between action parameters \( \theta_a \) and the latent parameter \( \psi \). This structure allows for sparsity, meaning that \( \theta_a \) may depend only on a subset of \( \psi \)'s coordinates. Moreover, \( p_a \) accounts for model uncertainty, allowing for cases where \( \theta_a \) is not a deterministic function of \( \psi \), i.e., \( \theta_a \neq f_a(\psi) \).

\begin{figure}[H]
\begin{center}
\includegraphics[width=0.6\linewidth]{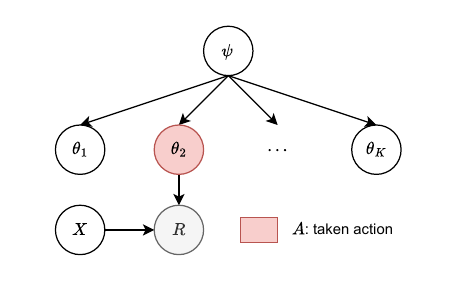}
\caption{Graph representation of the structured prior.}
\label{fig:sdm-sdm_graph}
\end{center}
\end{figure}

The reward distribution for action \( a \) in context \( x \) is given by \( p(\cdot \mid x; \theta_a) \), which depends only on \( x \) and \( \theta_a \). To summarize, the structured prior is defined below, and its graphical representation is given in \cref{fig:sdm-sdm_graph}.
\begin{align}\label{eq:sdm-model}
  \psi &\sim q(\cdot)\,, \\
  \theta_a \mid  \psi &\sim p_{a}(\cdot ; f_a(\psi))\,, &  \forall a \in \mathcal{A}\,, \nonumber \\
  R \mid \psi, \theta, X, A &\sim p(\cdot \mid  X; \theta_A)\,. \nonumber
\end{align}
To derive the posterior under this prior, we assume that: \emph{(i)} \( (X, A) \) is independent of \( \psi \), and given \( \psi \), \( (X, A) \) is independent of \( \theta \); and \emph{(ii)} given \( \psi \), the parameters \( \theta_a \) for all \( a \in \mathcal{A} \) are independent.

Now, we discuss how to perform off-policy learning under this general structured prior in \cref{eq:sdm-model}, before applying it to linear-Gaussian distributions in \cref{sec:sdm-application}.

\subsection{Off-Policy Learning}\label{subsec:sdm-ope}
Off-policy learning relies on an estimate of the value function \(V(\pi; \theta_*)\) obtained using the logged data \(\cD_n\). In DMs, the estimator \(\hat{V}_{\textsc{dm}}\) in \cref{eq:sdm-dm_policy_value} requires access to the learned reward \(\hat{r}(x, a) \approx r(x, a; \theta_*)\). In our Bayesian setting, this requires access to the action posterior \(\theta_{a} \mid \cD_n\) under the prior in \cref{eq:sdm-model} since the reward is then estimated as \(\hat{r}(x, a) = \condE{r(x, a; \theta)}{\cD_n}\) for any \((x, a) \in \cX \times \cA\), and this estimate is plugged into \(\hat{V}_{\textsc{dm}}\) in \cref{eq:sdm-dm_policy_value} to estimate \(V(\pi; \theta_*)\). Thus, we need to derive the posterior density of the action parameter \(\theta_a\), \(p(\theta_a \mid \cD_n)\), under the structured prior in \cref{eq:sdm-model}, which reads
\begin{align}\label{eq:sdm-action_posterior}
  p(\theta_a \mid \cD_n)= \int_{\psi} p(\theta_a \mid \psi, \cD_n) p(\psi \mid \cD_n)
 \dif \psi\,,
\end{align}
where $\psi \mid \cD_n$ is the latent posterior and $\theta_a \mid \psi, \cD_n$ is the \emph{conditional} action posterior. To compute $p(\theta_a \mid \cD_n)$, we first compute $p(\theta_a \mid \psi, \cD_n)$ and $ p(\psi \mid \cD_n)$ and then integrate out $\psi$ following \cref{eq:sdm-action_posterior}. First,
\begin{align}\label{eq:sdm-p_derviation}
 p(\theta_a \mid \psi, \cD_n) \propto \LL_{a}(\theta_a) p_a (\theta_a ; f_a(\psi))\,, 
\end{align}
with $\LL_{a}(\theta_a) = \prod_{(X, A, R) \in S_a} p(R| X; \theta_a)$ is the likelihood of observations of action $a$ ($S_a = (X_i, A_i, R_i)_{i \in [n], A_i=a}$ is the subset of $\cD_n$ where $A_i=a$). Similarly,
\begin{align}\label{eq:sdm-q_derviation}
\hspace{-0.1cm} p(\psi \mid \cD_n) & \propto \prod_{b \in \cA} \int_{\theta_b} \LL_{b}(\theta_b) p_b\left(\theta_b ; f_b(\psi) \right) \dif \theta_b \, q(\psi)\,, 
\end{align}
This allows us to further develop \cref{eq:sdm-action_posterior} as
\begin{align}\label{eq:sdm-action_posterior_summary}
  p(\theta_a \mid \cD_n)  \propto &  \int_{\psi}\LL_{a}(\theta_a) p_a (\theta_a ; f_a(\psi)) \prod_{b \in \cA} \int_{\theta_{b}} \LL_{b}(\theta_{b}) p_{b}\left(\theta_{b} ; f_{b}(\psi) \right) \dif \theta_{b}  \, q(\psi) \dif \psi\,.
\end{align}
All the quantities inside the integrals in \cref{eq:sdm-action_posterior_summary} are given (the parameters of $p_a$ and $q$) or tractable (the terms in $\LL_{a}$). Thus, if these integrals can be computed, then the posterior can be fully characterized in closed form, which we will do in \cref{sec:sdm-application} in the fully linear case. 
Otherwise, the posterior should be approximated.

Finally, we act greedy with respect to our estimator $\hat{V}_{\textsc{dm}}$ and define the learned policy as the one maximizing it: $\hat{\pi}_{\textsc{g}} = \argmax_{\pi \in \Pi} \hat{V}_{\textsc{dm}}(\pi)$. If the set of policies $\Pi$ contains deterministic policies, then 
\begin{align}\label{eq:sdm-learning_procedure_2}
     \hat{\pi}_{\textsc{g}}(a \mid x) = \mathds{1}{\{a = \argmax_{b \in \cA} \hat{r}(x, b)\}}\,.
\end{align}
In particular, we do not adopt the common pessimism approach \citep{jin2021pessimism}. In pessimism, one constructs confidence intervals of the reward estimate $\hat{r}(x, a)$ of the form $|r(x, a; \theta) - \hat{r}(x, a)| \leq u(x, a)$, and then defines the learned policy as $\hat{\pi}_{\textsc{p}}(a \mid x) = \mathds{1}{\{a = \argmax_{b \in \cA} \hat{r}(x, b) - u(x, b)\}}$. The advantage of one over another depends on the evaluation metric used. Our metric is the Bayesian suboptimality (BSO), defined in \cref{sec:sdm-analysis}. It assesses the average performance of algorithms across multiple problems rather than the worst-case. The Greedy policy is more suitable for BSO optimization than pessimism (demonstrated theoretically and empirically in \cref{subsec:sdm-greedy_pessimism,app:sdm-greedy_pessimism_experiments}).

\section{Linear-Gaussian Case}\label{sec:sdm-application}
In this section, we use linear functions $f_a$ combined with Gaussian distributions for the structured prior~\cref{eq:sdm-model}. Precisely, we assume that the latent prior $q(\cdot) = \cN(\cdot; \mu, \Sigma)$ is Gaussian with mean $\mu \in \real^{d^\prime}$ and covariance $\Sigma \in \real^{d^\prime \times d^\prime}$. Moreover, let $\W_a \in \real^{d \times d^\prime}$ be the \emph{mixing matrix} for action $a$, we define $f_a(v) = \W_a v$ for any $v \in \real^{d^\prime}$. We define the conditional prior $p_{a}(\cdot ; f_a(\psi)) = \cN(\cdot; \W_a \psi, \Sigma_a)$ is Gaussian with mean $f_a(\psi) = \W_a \psi \in \real^d$ and covariance $\Sigma_{a} \in \real^{d \times d}$. The reward distribution $p(\cdot \mid x; \theta_a)$ is also linear-Gaussian as $\cN(\cdot; \phi(x)^\top \theta_a, \sigma^2)$, where $\phi(\cdot)$ outputs
a $d$-dimensional representation of $x$ and $\sigma>0$ is the observation noise variance. The whole prior is 
\begin{align}\label{eq:sdm-contextual_gaussian_model}
    \psi & \sim \cN(\mu, \Sigma)\,, \\ 
    \theta_a \mid   \psi& \sim \cN\Big( \W_a \psi , \,  \Sigma_{a}\Big)\,, & \forall a \in \cA\,,\nonumber\\ 
    R \mid  \psi, \theta, X, A & \sim \cN(\phi(X)^\top \theta_A ,  \sigma^2)\,. \nonumber
\end{align}

\subsection{Applications}
\textbf{Mixed-effect modeling.} \cref{eq:sdm-contextual_gaussian_model} allows modeling that action parameters depend on a linear mixture of effect parameters (\cref{chap:meTS}). Precisely, let $J$ be the number of effects and assume that $d^\prime = dJ$ so that the latent parameter $\psi$ is the concatenation of $J$, $d$-dimensional effect parameters, $\psi_j \in \real^d$, such as $ \psi = (\psi_j)_{j \in [J]} \in \real^{dJ}$. Moreover, assume that for any $a \in \cA\,,$ $\W_a=w_a^\top \otimes I_d \in \real^{d \times dJ}$ where $w_a = (w_{a, j})_{j \in [J]} \in \mathbb{R}^{J}$ are the \emph{mixing weights} of action $a$. Then, $\W_a \psi = \sum_{j \in [J]} w_{a, j} \psi_j$ for any $a \in \cA$. Sparsity, i.e., when an action $a$ only depends on a subset of effects, is captured through the mixing weights $w_a$: $w_{a, j}=0$ when action $a$ is independent of the $j$-th effect parameter $\psi_j$ and $w_{a, j}\ne 0$ otherwise. Also, the level of dependence between action $a$ and effect $j$ is quantified by the absolute value of $w_{a, j}$. This mixed-effect model can be used in numerous applications (the reader can refer to the first paragraphs of \cref{chap:meTS} for examples).

\textbf{Low-rank modeling.} \cref{eq:sdm-contextual_gaussian_model} can also model the case where the dimension of the latent parameter $\psi$ is much smaller than that of the action parameters $\theta_a$, i.e., when $d^\prime \ll d$. Again, this is captured through the mixing matrices $\W_a$, when $\W_a$ is low-rank.

\subsection{Closed-Form Solutions for \alg}\label{subsec:sdm-steps_lin_hierarchy}
The conditional action posterior  is known in closed-form as 
$\theta_a \mid \psi, \cD_n \sim \cN(\tilde{\mu}_{a}, \tilde{\Sigma}_{a})$, with
\begin{align}\label{eq:sdm-conditional_posterior}
   &\tilde{\Sigma}_{a}^{-1}  = \Sigma_{a}^{-1} + G_{a}\,, &   \tilde{\Sigma}_{a}^{-1} \tilde{\mu}_{a} =  \Sigma_{a}^{-1} \W_a \psi + B_{a}\,,
\end{align}
where 
\begin{align*}
    &G_{a} = \sigma^{-2} \sum_{i \in [n]} \mathds{1}{\{A_i=a\}} \phi(X_i) \phi(X_i)^\top\,, & B_{a} = \sigma^{-2} \sum_{i \in [n]} \mathds{1}{\{A_i=a\}} R_i \phi(X_i)\,.
\end{align*}
This posterior has the standard form except that the prior mean $\W_a \psi$ now depends on the latent parameter $\psi$. Similarly, the effect posterior writes $\psi \mid \cD_n \sim \cN(\bar{\mu}, \bar{\Sigma})$, where
\begin{align}\label{eq:sdm-effect_posterior}
   &\bar{\Sigma}^{-1} =   \Sigma^{-1}   + \sum_{a \in \cA} \W_a^{\top} ( \Sigma_a^{-1} - \Sigma_a^{-1} \tilde{\Sigma}_{a} \Sigma_a^{-1} )\W_a\,, \nonumber\\ 
   &\bar{\Sigma}^{-1}\bar{\mu} = \Sigma^{-1} \mu +\sum_{a \in \cA}   \W_a^\top \Sigma_a^{-1} \tilde{\Sigma}_{a}  B_a.
\end{align}
The latent posterior precision \( \bar{\Sigma}^{-1} \) is the sum of the latent prior precision \( \Sigma^{-1} \) and the learned action precisions \( \Sigma_a^{-1} - \Sigma_a^{-1} \tilde{\Sigma}_a \Sigma_a^{-1} \), weighted by \( \W_a^\top \W_a \). The contribution of each action's learned precision to the latent precision is proportional to \( \W_a^\top \W_a \). This intuition similarly applies to interpreting \( \bar{\mu} \). Finally, from \cref{eq:sdm-action_posterior_summary}, the action posterior is \( \theta_a \mid \cD_n \sim \mathcal{N}(\hat{\mu}_a, \hat{\Sigma}_a) \), where
\begin{align}\label{eq:sdm-action_posterior_gaussian}
   &\hat{\Sigma}_a = \tilde{\Sigma}_{a}  +\tilde{\Sigma}_{a} \Sigma_{a}^{-1}  \W_a \bar{\Sigma} \W_a^\top \Sigma_{a}^{-1} \tilde{\Sigma}_{a}\,,   &\hat{\mu}_a = \tilde{\Sigma}_a \big( \Sigma_{a}^{-1} \W_a \bar{\mu} + B_{a} \big)  \,.
\end{align}
Finally, from \cref{eq:sdm-contextual_gaussian_model}, the reward function is \( r(x, a; \theta) = \phi(x)^\top \theta_a \). Thus, the estimated reward is 
\begin{align*}
    &\hat{r}(x, a) = \mathbb{E}[r(x, a; \theta) \mid \cD_n] = \phi(x)^\top \hat{\mu}_a\,, & \forall (x, a) \in \cX \times \cA\,.
\end{align*}
This can then be plugged in \cref{eq:sdm-learning_procedure_2} for decision-making, leading to 
\begin{align*}
    \hat{\pi}_{\textsc{g}}(a \mid x) = \mathds{1}{\{a = \argmax_{b \in \cA} \phi(x)^\top \hat{\mu}_b\}}\,.
\end{align*}
To see why this is more beneficial than the standard prior in \cref{eq:sdm-basic_model}, notice that the mean and covariance of the posterior of action $a$, $\hat{\mu}_a$ and $\hat{\Sigma}_a$, are now computed using the mean and covariance of the latent posterior, $\bar \mu $ and $\bar \Sigma$. But $\bar \mu $ and $\bar \Sigma$ are learned using the interactions with all the actions in $\cD_n$. Thus $\hat{\mu}_a$ and $\hat{\Sigma}_a$ are also learned using the interactions with all the actions in $\cD_n$, in contrast with the standard prior in \cref{eq:sdm-basic_model} where they were learned using only the interaction with action $a$. The additional computational cost of considering the structured prior in \cref{eq:sdm-contextual_gaussian_model} is small. The computational and space complexities are $\mathcal{O}(K((d^2 + {d^\prime}^2)(d + d^\prime)))$ and $\mathcal{O}(K d^2)$. For example, when $d^\prime = \mathcal{O}(d)$, these complexities become $\mathcal{O}(Kd^3)$ and $\mathcal{O}(Kd^2)$, respectively. This is exactly the cost of the standard prior in \cref{eq:sdm-basic_model}. In contrast, this strictly improves the computational efficiency of jointly modeling the action parameters, where the complexities are $\mathcal{O}(K^3d^3)$ and $\mathcal{O}(K^2d^2)$ since the joint posterior of $(\theta_a)_{a \in \cA} \mid \cD_n$ requires converting and storing a $dK \times dK$ covariance matrix.

\begin{remark}
\alg with linear-Gaussian hierarchies can be used even with data generated from non-linear rewards, and we empirically investigate its robustness to misspecification. We found that this model performs well even if the true rewards are not generated from a linear-Gaussian distribution.
\end{remark}

\section{Analysis}\label{sec:sdm-analysis}
\subsection{Bayesian Metrics}\label{subsec:sdm-bayes_suboptimality}
The performance of a learned policy \( \hat{\pi} \) is evaluated using suboptimality (SO): 
\begin{align*}
    \textsc{so}(\hat{\pi}; \theta_*) = V(\pi_*; \theta_*) - V(\hat{\pi}; \theta_*)\,,
\end{align*}
where \( \pi_* = \argmax_{\pi \in \Pi} V(\pi; \theta_*) \) is the optimal policy. This metric is well-suited when the environment is governed by a unique, fixed ground truth \( \theta_* \). It applies to any policy \( \hat{\pi} \), whether learned through frequentist approaches (e.g., MLE) or Bayesian ones (e.g., ours). However, when the environment is modeled as a random variable \( \theta_* \) sampled from some prior distribution, SO becomes less appropriate. Thus, drawing on recent developments in Bayesian analysis for online bandits through Bayes regret \citep{russo14learning}, we introduce a new metric for offline settings, termed \emph{Bayes suboptimality}, defined as:
\begin{align}\label{eq:sdm-bayes_suboptimality} 
  \textsc{Bso}(\hat{\pi})=   \mathbb{E}[V(\pi_*; \theta_*) - V(\hat{\pi}; \theta_*)]\,,
\end{align}
where the expectation is taken over all random variables: the logged data \(\cD_n\) and \(\theta_*\), which is treated as a random variable sampled from the prior. The BSO can be computed in two ways. One method involves taking the expectation under the prior \(\theta_*\), followed by taking an expectation under data generated from a fixed environment \(\theta_*\) as \(\cD_n \mid \theta_*\). The other method involves taking an expectation under the data \(\cD_n\), followed by taking an expectation under the posterior \(\theta_* \mid \cD_n\). The BSO is a reasonable metric for assessing the average performance of algorithms across multiple environments, due to the expectation over \(\theta_*\). It is also known that Bayes regret captures the benefits of using informative priors (\cref{chap:meTS}), and this is similarly achieved by the BSO.

\subsection{Theoretical Results}\label{subsec:sdm-main_result}
Our theory relies on the important well-specified assumption:
\begin{assumption}[Well-specified priors]\label{assum:well-specified}
Action parameters $\theta_{*, a}$ and rewards are drawn from \cref{eq:sdm-contextual_gaussian_model}. 
\end{assumption}
We also make simplifying assumptions for the sake of exposition.
\begin{assumption}[Diagonal covariances for simplicity]\label{assum:isotropic}
We assume $\Sigma_a = \sigma_0^2 I_d$, $\Sigma = \tau^2 I_{d'}$, $\normw{\phi(x)}{2} \leq 1$, and the matrices $\W_a$ are normalized such that $\lambda_1(\W_a \W_a^\top) = \lambda_d(\W_a \W_a^\top) = 1$.
\end{assumption}

This yields our bound on the BSO of \alg.
\begin{theorem}[Covariance-Dependent Bound]\label{thm:main_thm_1}
Let $\pi_*(x)$ be the optimal action for context $x$. Then the BSO of \emph{\alg} under the structured prior in \cref{eq:sdm-contextual_gaussian_model} satisfies
\begin{align}
\textsc{Bso}(\hat\pi_{\textsc{g}})
& \le
\alpha_n\,\E{}{\|\phi(X)\|_{\hat\Sigma_{\pi_*(X)}}}
+
\sqrt{\frac{(2\log(2K)+2)(\sigma_0^2 + \tau^2)}{n}},
\end{align}
where $\alpha_n
=
\sqrt{d + 2\sqrt{d\log(Kn)} + 2\log(Kn)}.$
\end{theorem}
Scaling of the bound in \cref{thm:main_thm_1} aligns with existing frequentist results \citep[Theorem 4.4]{jin2021pessimism}. The main differences lie in the constants and the fact that this rate is achieved using greedy policies in \cref{eq:sdm-learning_procedure_2}. This contrasts with the frequentist setting where pessimism is used \citep{jin2021pessimism} and known to be optimal \citep[Theorem 4.7]{jin2021pessimism}. In fact, greedy policies are optimal when BSO is used as the performance metric. Specifically, \(\textsc{Bso}(\hat{\pi}_{\textsc{g}}) \leq \textsc{Bso}(\pi)\) for any policy \(\pi\), including pessimistic ones. Therefore, in the Bayesian setting and when BSO is used as a performance metric, greedy policies should always be preferred to pessimistic ones. This fundamental difference is proven in \cref{subsec:sdm-greedy_pessimism} and it is of independent interest beyond this work. 

\cref{thm:main_thm_1} suggests that the BSO primarily depends on the posterior covariance of action $\pi_*(X)$ in the direction of the context $\phi(X)$. That is, when the uncertainty in the posterior distribution of the optimal action $\pi_*(X)$ is low on average across different contexts $X$ and logged data $\cD_n$, then the BSO bound is correspondingly small. In particular, the tightness of the bound depends on the degree to which the logged data covers the optimal actions on average.

\cref{thm:main_thm_1} can highlight the advantages of using \alg over the non-structured prior in \cref{eq:sdm-basic_model}. To see this, notice that the parameters of the non-structured prior in \cref{eq:sdm-basic_model}, $\mu_a$ and $\Sigma_a$, are obtained by marginalizing out $\psi$ in \cref{eq:sdm-contextual_gaussian_model}. In this case, $\mu_a^{\textsc{ns}} \gets \W_a\mu$ and $\Sigma_a^{\textsc{ns}} \gets \Sigma_a + \W_a\Sigma\W_a^\top$. The corresponding posterior covariance is $\hat{\Sigma}^{\textsc{ns}}_a = ((\Sigma_a + \W_a\Sigma\W_a^\top)^{-1} + G_{a})^{-1}$, and is generally larger than the covariance of \alg, $\hat{\Sigma}_a$ in \cref{eq:sdm-action_posterior_gaussian}. This is more pronounced when the number of actions $K$ is large and when the latent parameters are more uncertain than the action parameters. Thus, the BSO bound of \alg is smaller due to the reduced posterior uncertainty it exhibits. Also, note that even when $\pi_*(X)$ is unobserved in the logged data $\cD_n$, \alg's posterior covariance $\hat{\Sigma}_{\pi_*(X)}$ can remain small since we use interactions with all actions to compute it. This contrasts with standard non-structured priors in \cref{eq:sdm-basic_model}, where observing $\pi_*(X)$ is necessary; without such observations, the posterior covariance $\hat{\Sigma}_{\pi_*(X)}$ would simply be the prior covariance $\Sigma_{\pi_*(X)}$.

Next, we provide another bound on the BSO that scales as \( \mathcal{O}(1/\sqrt{n}) \). To simplify the exposition, we roughly present its scaling with \( n \) in \cref{thm:main_thm} and defer the complete general statement to \cref{subsec:sdm-explicit_bound}. We make the following additional assumptions:

\begin{assumption}
Let $G = \mathbb{E}_{X \sim \nu}[X X^\top]$ with $g = \lambda_d(G)$. We assume that $g > 0$.
\end{assumption}

\begin{assumption}[Context-independent logging policy]
$A$ is independent of $X$, i.e., $\pi_0(a \mid x) = \pi_0(a) = p_a$ for all $x$ and $a$. Equivalently, $(X_i)$ are i.i.d.\ $\sim \nu$ and independent of $(A_i)$, with $\mathbb{P}(A = a) = p_a$.
\end{assumption}

\begin{theorem}[Scaling with $n$]\label{thm:main_thm}
For $n$ large enough, the BSO of \emph{\alg} under the structured prior in \cref{eq:sdm-contextual_gaussian_model} scales as
$$
\textsc{Bso}(\hat{\pi}_{\textsc{g}})
=
\tilde{\mathcal O} \left(
 \sqrt{\mathbb E \left[\frac{d}{n\rho_X+1}\right]}
 + 
\sqrt{\frac{\log K}{n}}
\right),
$$
where $\rho_X = \pi_0(\pi_*(X))$.
\end{theorem}

The above bound becomes smaller or larger depending on how well the logging policy \( \pi_0 \) covers the optimal actions for each context \( x \).

\section{Experiments}\label{sec:sdm-experiments}
We evaluate \alg using both synthetic and real datasets. We use the average reward of the learned policy relative to the optimal policy as the evaluation metric. 

\subsection{Synthetic Problems}

\textbf{Setting.} We simulate synthetic data using the linear-Gaussian model in \cref{eq:sdm-contextual_gaussian_model} with \( \sigma = 1 \). The contexts \( X \) are sampled uniformly from \( [-1, 1]^d \), with \( d = 10 \). The matrices \( \W_a \) are sampled uniformly from \( [-1,1]^{d \times d^\prime} \), where we very $d'$ as \( d^\prime \in \{5, 10., 20\} \). We set \( \Sigma = 3I_{d^\prime} \) and \( \Sigma_a = I_d \), meaning the latent parameters are more uncertain than the action parameters. The latent mean \( \mu \) is randomly sampled from \( [-1, 1]^{d^\prime} \). The number of actions is varied as \( K \in\{100, 1000\} \), and we use a uniform logging policy to collect data. Additional experiments with different logging policies are presented in \cref{app:sdm-experiments}.

\textbf{Baselines.} First, we use \alg under prior in \cref{eq:sdm-contextual_gaussian_model}. Second, we examine \texttt{DM (Bayes)}, which uses the standard non-structured prior in \cref{eq:sdm-basic_model}, where parameters $\mu_a$ and $\Sigma_a$ are obtained by marginalizing out the latent parameters $\psi$ in \cref{eq:sdm-contextual_gaussian_model}. Thus \texttt{DM (Bayes)} is a standard Bayesian DM that does not capture arm reward correlations. We also include \texttt{DM (Freq)}, which estimates $\theta_{*, a}$ by the MLE. We include \texttt{\texttt{IPS}} \citep{horvitz1952generalization}, self-normalized \texttt{IPS} (\texttt{sn\texttt{IPS}}) \citep{swaminathan2015self}, and doubly robust (\texttt{DR}) \citep{dudik14doubly}, which we optimize to learn the optimal policy. M\texttt{IPS} \citep{saito2022off} and \texttt{PC} \citep{sachdeva2023off} are also included. Implementation details of baselines is provided in \cref{app:sdm-details_imp}.

\begin{figure}[H]
  \centering  \includegraphics[width=\linewidth]{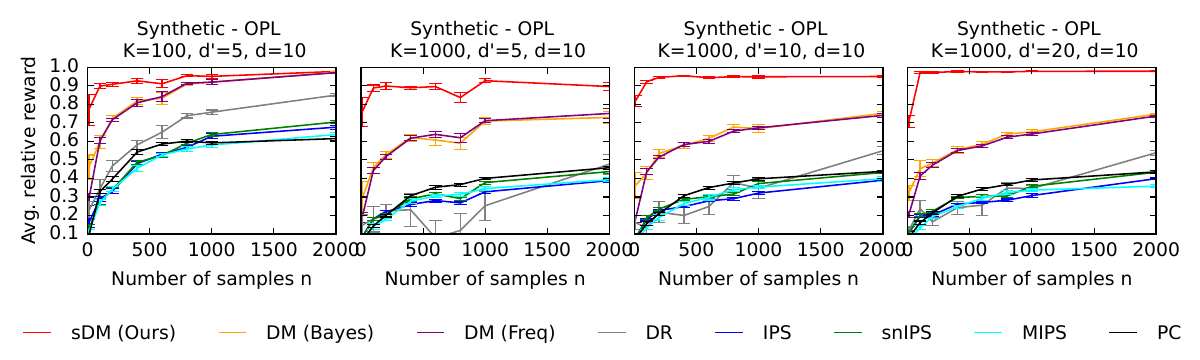}
  \caption{The average relative reward of the learned policy using one of the baselines on \textbf{synthetic problems} with varying $n$, $K$ and $d^\prime$.} 
  \label{fig:sdm-synthetic_linear}
\end{figure}

\textbf{Results.} In \cref{fig:sdm-synthetic_linear}, we plot the results and we observe that \alg consistently outperforms the baselines across all settings. This performance gap becomes even more significant when sample size \(n\) is small. These results highlight \alg's enhanced efficiency in using available logged data, making it particularly beneficial in data-limited situations and scalable to large action spaces.

\begin{figure}[H]
\begin{center}
\includegraphics[width=0.4\linewidth]{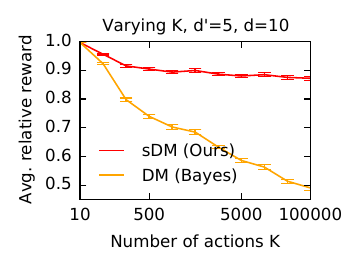}
\caption{\alg \emph{vs.} \texttt{DM (Bayes)} for varying $K$.}
\label{fig:sdm-las}
\end{center}
\end{figure}

\textbf{Scaling to large action spaces.} \texttt{sDM} achieves improved scalability compared to standard DM as it leverages data more efficiently. While it still learns a $d$-dim. parameter for each action $a$, it does so by considering interactions with all actions in the logged data $\cD_n$, instead of only using interactions with the specific action $a$. This is crucial, especially given that many actions may not even be observed in $\cD_n$. To show \texttt{sDM}'s improved scalability, we compare it to the most competitive baseline, \texttt{DM (Bayes)}, for varying $K \in [10,100000]$ with $n=1000$. The results in \cref{fig:sdm-las} reveal that the performance gap between \texttt{sDM} and \texttt{DM (Bayes)} becomes more significant when the number of actions $K$ increases. Hence, despite the necessity for \texttt{sDM} to learn distinct parameters for each action, accommodating practical scenarios like recommender systems where unique embeddings are learned for each product, it still enjoys good scalability.

\subsection{MovieLens Problems}
\textbf{Setting.} We use MovieLens 1M \citep{movielens}, which contains 1 million ratings representing the interactions between 6,040 users and 3,952 movies. To create a semi-synthetic environment, we first apply a low-rank factorization to the rating matrix, producing 5-dim. representations: \( x_u \in \mathbb{R}^5 \) for user \( u \in [6040] \) and \( \theta_a \in \mathbb{R}^5 \) for movie \( a \in [3952] \). Movies are treated as actions, and contexts \( X \) are sampled randomly from the user vectors. The reward for movie \( a \) and user \( u \) is modeled as \( \mathcal{N}(x_u^\top \theta_a, 1) \), serving as proxy for ratings. A uniform logging policy is used to collect data.

\textbf{Baselines.} We consider the same baselines as in synthetic data. A prior is not needed for \texttt{DM (Freq)}, \texttt{\texttt{IPS}}, \texttt{sn\texttt{IPS}}, and \texttt{DR}. However, for \texttt{DM (Bayes)}, a standard prior in \cref{eq:sdm-basic_model} is inferred from data, where we set $\mu_a$ to be the mean of movie vectors across all dimensions, and $\Sigma_a = {\rm diag}(v)$, where $v$ represents the variance of movie vectors across all dimensions. Unlike the synthetic experiments, the latent structure assumed by \alg is not inherently present in MovieLens. But we learn it by training a Gaussian Mixture Model (GMM) to cluster movies into $J=5$ mixture components. This gives rise to the mixed-effect structure described in \cref{sec:sdm-application}, which represents a specific instance of \alg with $d^\prime = dJ = 25$. \texttt{MIPS} also has access to movie clusters, while we use the knn smoothing implementation of \texttt{PC} (see \citep[Section 3]{sachdeva2023off}). Note that \texttt{DM (Bayes)}, \alg, \texttt{MIPS} and \texttt{PC} use the same subset of data (of size $1000$) to learn their priors/assumed structure and thus we compare them fairly. We conduct experiments with $K \in \{100, 1000\}$ randomly selected movies. 

\begin{figure}[H]\label{fig:sdm-movielens}
  \centering  \includegraphics[width=0.7\linewidth]{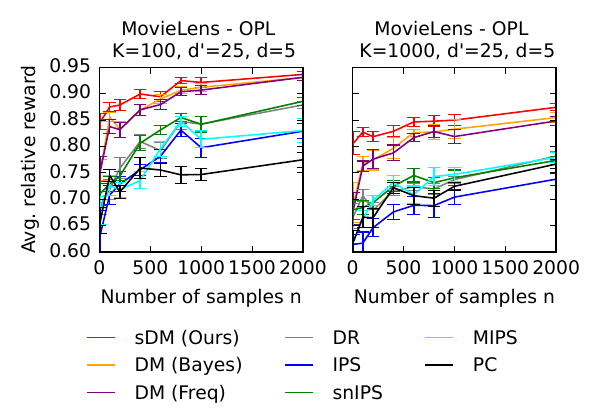}
  \caption{The average relative reward of the learned policy using one of the baselines on \textbf{MovieLens problems} with varying $n$, $K$ and $d^\prime$.} 
\end{figure}

\textbf{Results.} Results are in \cref{fig:sdm-movielens}. Even though the latent structure assumed by \alg is not inherently present in MovieLens, \alg still outperforms the baselines by learning it offline.

\section{Conclusion}\label{sec:sdm-conclusion}
We introduced \alg, a structured approach to off-policy learning that leverages latent structure among actions to enhance statistical efficiency while maintaining computational tractability, particularly in large action spaces with limited data coverage. Within a Bayesian framework, we proved that greedy policies outperform pessimistic ones under Bayesian suboptimality and established $\mathcal{O}(1/\sqrt{n})$ convergence without requiring restrictive full-support assumptions.

Our work has several limitations. First, our theoretical analysis assumes a well-specified prior; while we empirically observed robustness to misspecification, formal guarantees under prior mismatch remain an open question. Second, closed-form posterior updates are available only for linear-Gaussian hierarchies; extending to nonlinear reward models requires approximate inference, which may compromise computational efficiency or statistical accuracy. Third, the latent structure must be specified or learned pre-trained, adding an additional modeling task. Finally, extending \alg to handle nonlinear hierarchies, building on the diffusion-based approach of \cref{chap:dTS}, is a promising avenue for future work.

%% file: contents/las/main_text.tex
This chapter challenges the dominant paradigm in off-policy learning (explored in \cref{chap:ES}), which frames the problem as finding a policy $\hat{\pi} = \argmax_\pi \hat{V}(\pi)$ (or, with pessimism, $\hat{\pi} = \argmax_\pi [\hat{V}(\pi) - \operatorname{pen}(\pi)]$), where $\hat{V}$ is an IPS-based\footnote{Recall that IPS is an importance-weighting estimator of the policy value. We use \emph{IPS-based} to refer to any estimator derived from or inspired by importance weighting.} estimate of the true policy value $V(\pi)$. The rationale behind these objectives is that maximizing a more accurate value estimate yields a better policy. However, this \emph{estimator-centric view} neglects a crucial factor: the optimization landscape.

IPS-based objectives \citep{dudik2011doubly,dudik2012sample,dudik14doubly,wang2017optimal,farajtabar2018more,su2020doubly,metelli2021subgaussian,kuzborskij2021confident,saito2022off} are highly non-concave under common policy parameterizations \citep{surrogate}, prone to suboptimal local maxima and plateaus: issues that are exacerbated in large action spaces. Even sophisticated estimators designed to reduce variance fail to overcome this optimization barrier, as they often induce equally difficult landscapes.

We make the following contributions. \begin{enumerate*}[label=\textbf{\arabic*)}]
\item We show that \emph{objective-aware policy parametrization} can partially alleviate these difficulties by structuring the policy class to match the implicit biases of the estimator. Such parametrizations reduce the effective search space and can shorten optimization plateaus and local maxima. However, this strategy does not eliminate the fundamental non-concavity of IPS-based objectives, leaving optimization as the central bottleneck.
\item Motivated by this limitation, we advocate for an alternative approach based on \textit{policy-weighted log-likelihood (PWLL)} objectives. Unlike traditional estimators, PWLL optimizes an objective $\hat{U}(\pi)$ designed for ease of optimization rather than accuracy in estimating $V(\pi)$. Although PWLL objectives perform poorly as value estimators, their favorable concave landscape makes them significantly more effective for policy learning.
\item Through theoretical and empirical analysis, we demonstrate that this optimization-centric approach consistently enables simpler PWLL objectives to outperform complex, state-of-the-art IPS-based methods, particularly in large action spaces.
\end{enumerate*}

\textbf{Setting and organization.} This chapter considers the general setting of \cref{sec:offline_background} with $R \in [0, 1]$. The remainder is organized as follows. \cref{sec:las-ope_based} employs an asymptotic lens to analyze IPS-based objectives and derives objective-aware policy parametrizations that partially alleviate their optimization challenges. \cref{sec:las-PWLL_based} introduces PWLL objectives and establishes their favorable optimization properties. \cref{sec:las-analysis} presents large-scale experiments. We conclude in \cref{sec:las-conclusion}.

\section{Analysis of IPS-Based Objectives}
\label{sec:las-ope_based}
IPS-based objectives optimize an estimator $\hat{V}(\pi)$ of the policy value $V(\pi)$. To understand the policies to which these estimators converge, we study their \emph{oracle policies} $\pi^{\textsc{method}}_* = \argmax_{\pi} \mathbb{E}[\hat{V}^{\textsc{method}}(\pi)]$. Taking the expectation removes sampling fluctuations and isolates the inductive bias of each objective: different estimators yield different oracle policies, even with infinite data. Crucially, oracle policies admit closed-form expressions, enabling precise characterization of each estimator's implicit bias. This analysis motivates \emph{objective-aware parametrizations} that align the policy class with the estimator's bias to ease optimization: the first improvement we propose in this chapter.

\subsection{Standard IPS-Based Objectives}
The foundational
\texttt{IPS} estimator \citep{horvitz1952generalization} re-weights observed rewards by the ratio between the target policy $\pi$ and the logging policy $\pi_0$:
\begin{align}\label{eq:las-ips_value}
    \hat{V}_{\textsc{ips}}(\pi) = \frac{1}{n} \sum_{i=1}^n \frac{\pi(A_i\mid X_i)}{\pi_0(A_i\mid X_i)} R_i .
\end{align}
In expectation, \texttt{IPS} selects the best-rewarding action among those in the support of $\pi_0$:
\begin{align}\label{eq:las-ips_optimal_policy}
   \pi^{\texttt{IPS}}_*(a \mid x) = \mathbbm{1} \left[ a = \argmax_{a' \in \cA} r(x, a')  \mathbbm{1}[\pi_0(a' \mid x) > 0] \right].
\end{align}

\textbf{Clipped IPS (\texttt{cIPS}).} To mitigate the high variance of \texttt{IPS}, a widely used variant is \texttt{cIPS} \citep{bottou2013counterfactual} that clips small propensity scores at a threshold $\tau \in (0,1)$:
\begin{align}\label{eq:las-cips_value}
    \hat{V}_{\textsc{cips}}(\pi) = \frac{1}{n} \sum_{i=1}^n \frac{\pi(A_i\mid X_i)}{\max\{\pi_0(A_i\mid X_i), \tau\}} R_i .
\end{align}
This clipping introduces a bias. The oracle policy down-weights the rewards of rare actions, causing it to favor actions that were frequent under $\pi_0$, even if they are suboptimal:
\begin{align}\label{eq:las-cips_optimal_policy}
    \pi^{\texttt{\texttt{cIPS}}}_*(a \mid x) = \mathbbm{1}\Big[ a = \argmax_{a' \in \mathcal{A}} \frac{\pi_0(a' \mid x)}{\max\{\pi_0(a' \mid x), \tau\}}  r(x, a')   \Big] .
\end{align}

\textbf{Exponential smoothing (\texttt{ES}).}  
Instead of hard clipping, \texttt{ES} (\citet{aouali23a}, \cref{chap:ES}) smooths importance weights by raising propensities to a fractional power $\alpha \in (0,1)$:
\begin{align}\label{eq:las-es_value}
    \hat{V}_{\textsc{es}}(\pi) = \frac{1}{n} \sum_{i=1}^n \frac{\pi(A_i \mid X_i)}{\pi_0(A_i \mid X_i)^{\alpha}} R_i .
\end{align}
Its oracle policy balances reward maximization with preference for frequent actions:
\begin{align}\label{eq:las-es_optimal_policy}
    \pi^{\texttt{ES}}_*(a \mid x) = \mathbbm{1} \left[a = \argmax_{a' \in \cA} r(x,a') \pi_0(a' \mid x)^{1-\alpha} \right].
\end{align}
Another variant of \texttt{ES} regularizes the entire importance weight as $(\frac{\pi}{\pi_0})^\beta$ instead of only the denominator. In contrast to the deterministic policies derived from \texttt{IPS}, \texttt{cIPS}, and the \texttt{ES} formulation above, this approach yields a stochastic oracle policy: $\pi^{\texttt{ES}}_*(a \mid x) \propto r(x, a)^{1/(1 - \beta)} \pi_0(a \mid x)$. Other regularizations include logarithmic smoothing \citep{Sakhi2024LS}, implicit exploration \citep{gabbianelli2023importance}, harmonic correction \citep{metelli2021subgaussian}, shrinkage \citep{su2020doubly}. But we do not include as \texttt{ES} and \texttt{cIPS} are already representative of them.

\textbf{Doubly robust (\texttt{DR}).} The \texttt{DR} estimator incorporates a reward model $\hat{r}(x, a)$ to reduce variance and enable generalization to actions outside $\pi_0$'s support. A common clipped variant is:
\begin{align}\label{eq:las-dr_value} \hat{V}_{\textsc{dr}}(\pi) = \frac{1}{n} \sum_{i=1}^n \frac{\pi(A_i\mid X_i)}{\max\{\pi_0(A_i\mid X_i), \tau\}} \left(R_i - \hat{r}(X_i, A_i) \right) + \E{A \sim \pi(\cdot \mid X_i)}{\hat{r}(X_i, A)} . 
\end{align} 
Its oracle policy interpolates between the reward model prediction and an importance weighting correction for the reward model error:
\begin{align}\label{eq:las-dr_optimal_policy} \pi^{\texttt{DR}}_*(a \mid x) = \mathbbm{1}\Big[ a = & \argmax_{a' \in \mathcal{A}} \hat{r}(x, a') + \frac{\pi_0(a' \mid x) }{\max\{\pi_0(a'\mid x), \tau\}}\left(r(x, a') - \hat{r}(x, a') \right)\Big] . \end{align}

\subsection{Large-Scale IPS-Based Objectives}
In large action spaces, importance weights $\tfrac{\pi(a \mid x)}{\pi_0(a \mid x)}$ can become huge, leading to estimators with high variance. To mitigate this, modern methods compute marginalized importance weights over a lower-dimensional action representation, trading bias for reduced variance.

\textbf{Marginalized IPS (\texttt{MIPS}).} \texttt{MIPS} \citep{saito2022off} tackles large action spaces by clustering actions. It maps each action $a$ to a cluster $c$ via a function $h: \mathcal{A} \rightarrow \mathcal{C}$, where $ \mid \mathcal{C} \mid  \ll  \mid \mathcal{A} \mid $. Estimation is then performed at the cluster level:
\begin{align}\label{eq:las-mips_value}
\hat{V}_{\textsc{mips}}(\pi) = \frac{1}{n} \sum_{i=1}^n \frac{\pi(C_i\mid X_i)}{\pi_0(C_i\mid X_i)} R_i , \quad \text{where } C_i = h(A_i) \text{ and } \pi(c \mid x) = \sum_{a \in c} \pi(a \mid x) .
\end{align}
This cluster-level marginalization introduces bias: the oracle policy only selects the best \textit{cluster} based on its average reward under $\pi_0$, and cannot differentiate between actions within that cluster:
\begin{align}\label{eq:las-mips_optimal_policy}
\pi^{\texttt{MIPS}}_*(c \mid x) = \mathbb{I}\left[ c = \operatorname*{argmax}_{c' \in \mathcal{C}} \left\{ \frac{ \sum_{a \in c'} \pi_0(a\mid x) r(x, a)}{\sum_{a \in c'} \pi_0(a\mid x)} \right\} \right] .
\end{align}
Hence, \texttt{MIPS} offers no specific guidance for selecting an action within the optimal cluster; any action is considered equally valid. Consequently, one possible induced action-level oracle under uniform tie-breaking is: 
\begin{align*}
    \pi^{\texttt{MIPS}}_*(a \mid x) = \frac{\mathbb{I}\left[ h(a) = \operatorname*{argmax}_{c' \in \mathcal{C}} \left\{ \frac{ \sum_{a \in c'} \pi_0(a\mid x) r(x, a)}{\sum_{a \in c'} \pi_0(a\mid x)} \right\} \right]}{|h(a)|}\,.
\end{align*}
where $|h(a)|$ denotes the size of the cluster containing action $a$.

\textbf{Conjunct effect modeling (\texttt{OffCEM}).} Building on \texttt{MIPS}, \texttt{OffCEM} \citep{saito2023off} uses a reward model $\hat{r}$ to correct for the cluster-level aggregation bias, in a doubly robust fashion:
\begin{align}\label{eq:las-offcem_value}
\hat{V}_{\textsc{offcem}}(\pi) = \frac{1}{n} \sum_{i=1}^n \left(\frac{\pi(C_i\mid X_i)}{\pi_0(C_i\mid X_i)}\left(R_i - \hat{r}(X_i, A_i) \right) + \mathbb{E}_{A \sim \pi(\cdot \mid X_i)}[\hat{r}(X_i, A)]\right).
\end{align}
The resulting oracle policy selects the action that maximizes the model-predicted reward $\hat{r}$, plus a cluster-level correction term that accounts for model error:
\begin{align}\label{eq:las-offcem_optimal_policy}
\pi^{\texttt{OffCEM}}_*(a \mid x) = \mathbb{I}\left[ a = \operatorname*{argmax}_{a' \in \mathcal{A}} \left\{ \hat{r}(x, a') +\frac{\sum_{\bar{a} \in h(a')} \pi_0(\bar{a} \mid x)(r(x, \bar{a}) - \hat{r}(x, \bar{a})) }{\sum_{\bar{a} \in h(a')}\pi_0(\bar{a} \mid x) }\right\}\right].
\end{align}

\textbf{Two-stage decomposition (\texttt{POTEC}).} In this chapter, we see \texttt{POTEC} \citep{saito2025potec} as an \emph{optimization strategy of} \texttt{OffCEM} (rather than seeing it as a new estimator). It restricts the policy to a cluster-informed form,
\begin{align*}
    \pi(a \mid x) = \sum_{c \in \mathcal{C}} \pi^{\textsc{rm}}(a \mid x,c)\pi^{\textsc{cl}}(c\mid x) ,
\end{align*}
where $\pi^{\textsc{rm}}(a \mid x, c) = \mathbbm{1}[a = \argmax_{a' \in c} \hat{r}(x, a')]$ is fixed, model-based policy that deterministically selects the best action within each cluster. Learning is then simplified to finding the optimal cluster-level policy $\pi^{\textsc{cl}}$ that maximizes the \texttt{OffCEM} objective in \cref{eq:las-offcem_value}:
\begin{align}\label{eq:las-potec_value}
\hat{V}_{\textsc{potec}}(\pi^{\textsc{cl}}) 
&= \frac{1}{n} \sum_{i=1}^n \left(
\frac{\pi^{\textsc{cl}}(C_i\mid X_i)}{\pi_0(C_i\mid X_i)}\left(R_i - \hat{r}(X_i, A_i)\right) 
+ \sum_{c\in\mathcal C} \pi^{\textsc{cl}}(c\mid X_i) \hat r^*_c(X_i)
\right),
\end{align}
where $\hat r^*_c(x) = \max_{a\in c}\hat r(x,a)$ is the estimated reward of the best action in cluster $c$. This practical decomposition has the same optimal oracle policy as \texttt{OffCEM}: 
$$\pi^{\texttt{POTEC}}_* = \pi^{\texttt{OffCEM}}_*.$$

\textbf{Policy convolution (\texttt{PC}).}  
Moving beyond hard clustering, \texttt{PC} \citep{sachdeva2023off} leverages the assumption that actions close in an embedding space yield similar rewards. 
For each action $a$, it aggregates over its neighborhood of nearest neighbors $N_\epsilon(a) = \{a' : d(a,a') < \epsilon\}$, 
where $d$ is a pre-defined distance metric (e.g., $\ell_2$ distance between action embeddings): 
\begin{align}\label{eq:las-pc_policy_value}
    \hat{V}_{\textsc{pc}}(\pi) 
    = \frac{1}{n} \sum_{i=1}^n \frac{\pi(N_\epsilon(A_i)\mid X_i)}{\pi_0(N_\epsilon(A_i)\mid X_i)} R_i, 
    \quad \text{with } \pi(N_\epsilon(a)\mid x) = \sum_{a' \in N_\epsilon(a)} \pi(a' \mid x)\,.
\end{align}

The induced oracle policy is deterministic: it selects the action $a'$ that maximizes an aggregated neighborhood score. 
Each logged neighbor $\bar{a} \in N_\epsilon(a')$ contributes its reward $r(x,\bar{a})$, 
weighted by the conditional probability of observing $\bar{a}$ under the logging policy restricted to its neighborhood.
\begin{align}\label{eq:las-pc_optimal_policy}
    \pi^{\texttt{PC}}_*(a \mid x) = \mathbb{I}\left[ a = \operatorname*{argmax}_{a' \in \mathcal{A}} \left\{ \sum_{\bar{a} \in N_\epsilon(a')} \frac{\pi_0(\bar{a} \mid x) r(x, \bar{a})}{\pi_0(N_\epsilon(\bar{a}) \mid x)} \right\} \right].
\end{align}

Other recent IPS variants for large action spaces \citep{peng2023offline,cief2024learning,taufiq2024marginal} are often extensions of MIPS that relax its core assumptions. We focused on four methods (\texttt{MIPS}, \texttt{OffCEM}, \texttt{POTEC}, and \texttt{PC}), which we consider representative of this family. Since these variants largely share the same MIPS foundation and optimization procedure (with the notable exception of \texttt{POTEC}), we expect our findings to be generally applicable.

\subsection{Optimization Challenges}

The effectiveness of IPS-based estimators in off-policy learning is often limited by their challenging optimization landscape. These objectives become difficult to optimize when paired with standard, expressive policy classes such as the softmax. This section explores why this occurs and introduces \emph{objective-aware parametrization} as a strategy to mitigate, though not entirely solve, the problem.

To analyze the optimization process, we consider policies parametrized by a softmax function over an \emph{effective action space}\footnote{The effective action space can also depend on context $x$, i.e., $\mathcal{A}_{\mathrm{eff}}(x) \subseteq \mathcal{A}$. We omit this dependence for notational simplicity.} $\mathcal{A}_{\mathrm{eff}} \subseteq \mathcal{A}$, which is the set of actions that can be assigned non-zero probability. By default, $\mathcal{A}_{\mathrm{eff}} = \mathcal{A}$, but we explain below why restricting it to match the structure of the estimator's oracle policy can be beneficial. Specifically, the policy takes the form:
\begin{align}\label{eq:las-softmax}
  \pi_\theta(a \mid x) = \frac{\exp(s_\theta(x, a))}{\sum_{a' \in \mathcal{A}_{\mathrm{eff}}} \exp(s_\theta(x, a'))} \mathds{1}_{a \in \mathcal{A}_{\mathrm{eff}}}\,, \quad \forall a \in \mathcal{A}\,,
\end{align}
where $s_\theta(x, a)$ is a learnable score function. Common choices are linear softmax scores:
\begin{align}\label{eq:las-softmax_scores}
     \text{lightweight: } s_\theta(x, a) = \phi(x, a) ^\top \theta\,,  \qquad \qquad \qquad  \text{heavyweight: }  s_\theta(x, a) = \phi(x)^\top \theta_a \,,
\end{align}
which we call \emph{lightweight parametrization} (a single shared parameter vector $\theta$, corresponding to a joint reward model) and \emph{heavyweight parametrization} (separate parameters $\theta_a$ for each action, corresponding to a disjoint reward model).

The size of the effective action space, $K_{\mathrm{eff}} = |\mathcal{A}_{\mathrm{eff}}|$, is the critical factor governing optimization difficulty. The following propositions (proofs in \cref{app:las-additional_proofs}, adapted from \citet{surrogate,gravitational_pull}) reveal the severity of the problem.

First, gradient-based methods can become trapped in suboptimal regions for extended periods.
\begin{proposition}[Optimization plateaus]\label{prop:plateaus}
    For any IPS-based estimator $\hat{V}$ that is linear in $\pi$, even with a linear softmax policy, there exist problem instances where gradient ascent remains trapped in a suboptimal region for $\mathcal{O}(K_{\mathrm{eff}})$ iterations.
\end{proposition}

Second, the optimization landscape has numerous poor local maxima.
\begin{proposition}[Local maxima]\label{prop:local_maxima}
    Under similar conditions, the optimization landscape for IPS-based objectives can contain a number of local maxima that is exponential in $K_{\mathrm{eff}}$.
\end{proposition}

These results highlight that $K_{\mathrm{eff}}$ plays a central role in optimization difficulty. The standard choice of $\mathcal{A}_{\mathrm{eff}} = \mathcal{A}$, which sets $K_{\mathrm{eff}} = K$, leads to optimization failure in large action spaces where $K$ can reach millions: learning must navigate a landscape with potentially $\mathcal{O}(K)$-length plateaus and exponentially many local maxima.

Surprisingly, even sophisticated methods designed specifically for large action spaces often fall into this trap. At first glance, methods such as \texttt{MIPS}, \texttt{OffCEM}, and \texttt{PC} appear to operate in a smaller space because their objectives involve marginalized probabilities: $\pi(C_i \mid X_i)$ in \texttt{MIPS} and \texttt{OffCEM}, or $\pi(N_\epsilon(A_i) \mid X_i)$ in \texttt{PC}. However, these marginalized terms are defined as sums over an underlying action-level policy:
\begin{align*}
    \pi(C_i \mid X_i) = \sum_{a \in C_i} \pi(a \mid X_i)\,, \qquad \text{and} \qquad \pi(N_\epsilon(a)\mid x) = \sum_{a' \in N_\epsilon(a)} \pi(a' \mid x)\,.
\end{align*}
Then, if $\pi(a \mid x)$ is a softmax over $\mathcal{A}$, then $K_{\mathrm{eff}} = K$ and \cref{prop:plateaus,prop:local_maxima} apply with $K_{\mathrm{eff}} = K$ which is large. The only exception is \texttt{POTEC}, which fixes the intra-cluster policy $\pi^{\textsc{rm}}$ and only optimizes a cluster-level policy $\pi^{\textsc{cl}}$. This reduces the effective action space to $\mathcal{A}_{\mathrm{eff}} = \mathcal{C}$ with $K_{\mathrm{eff}} = |\mathcal{C}| \ll K$, directly mitigating the optimization pathologies.

\subsubsection{Design implications: objective-aware parametrization}

The choice of $K_{\mathrm{eff}}$ introduces a fundamental trade-off. A smaller effective action space simplifies the optimization landscape, but risks excluding the optimal action and reduces policy expressiveness. If $\mathcal{A}_{\mathrm{eff}}$ is chosen arbitrarily, it may degrade performance. The challenge is to find the \emph{sweet spot}: a parametrization constrained enough to be optimizable, yet expressive enough to contain the objective's maximizer.

This is precisely where our asymptotic analysis helps. The oracle policy $\pi_*^{\textsc{method}}$ reveals the minimal sufficient set of actions required to maximize each objective. By aligning the policy parametrization with this structure, we can reduce $K_{\mathrm{eff}}$ without sacrificing performance: the core principle of our proposed \emph{objective-aware parametrization}.

For instance, the oracle policies for \texttt{IPS}, \texttt{cIPS}, and \texttt{ES} are confined to the support of the logging policy, $S_0(x)$. This implies that $\mathcal{A}_{\mathrm{eff}} = S_0$ is sufficient, reducing $K_{\mathrm{eff}}$ from $K$ to $|S_0| \ll K$. Similarly, for \texttt{OffCEM} and \texttt{MIPS}, the cluster-level structure of their oracle policies suggests a two-stage decomposition similar to that of \texttt{POTEC}, reducing $K_{\mathrm{eff}}$ to $|\mathcal{C}|$. We summarize these observations as claims, validated empirically in \cref{sec:las-analysis}:

\begin{claim}\label{claim:ips}
For \texttt{IPS}, \texttt{cIPS}, and \texttt{ES}, restricting the policy support to $S_0$ reduces $K_{\mathrm{eff}}$ and yields superior learned policies.
\end{claim}

\begin{claim}\label{claim:potec}
For \texttt{OffCEM} and \texttt{MIPS}, a two-stage \texttt{POTEC}-style decomposition that optimizes at the cluster level outperforms action-level parametrization.
\end{claim}

While objective-aware parametrization mitigates the optimization pathologies of \cref{prop:plateaus,prop:local_maxima} by reducing $K_{\mathrm{eff}}$, it only treats the symptoms without curing the underlying non-concavity. In the next section, we propose a more fundamental shift: abandoning value estimation in favor of inherently tractable objectives.

\section{Analysis of PWLL objectives}
\label{sec:las-PWLL_based}
To overcome the optimization challenges of IPS-based objectives, we consider policy-weighted log-likelihood (PWLL) objectives. These methods trade accurate value estimation for a well-behaved, concave optimization landscape, leading to more robust and effective policy learning.

\textbf{General form.} Given a positive weighting function $g(r, p_0)$, the PWLL objective is:
\begin{align}\label{eq:las-pwll}
\hat{U}_{g}(\pi) = \frac{1}{n} \sum_{i=1}^n g(R_i, \pi_0(A_i\mid X_i)) \log \pi(A_i\mid X_i) .
\end{align}
The key motivation behind PWLL is to replace the linear dependence on the policy in IPS-based estimators, responsible for plateaus and local maxima in \cref{sec:las-ope_based}, with a concave transformation. Softmax policies are parametrized through scores $s_\theta(x,a)$, and the map $s \mapsto \log \mathrm{softmax}(s)$ is concave. Consequently, for common linear parametrizations in \cref{eq:las-softmax_scores}, the composition $\log \pi_\theta(a\mid x)$ is concave in $\theta$. This removes the optimization pathologies inherent to IPS-based objectives. \cref{prop:concave} (proof in \cref{app:las-additional_proofs}.) formalizes this advantage.

\begin{proposition}\label{prop:concave}
For linear softmax policies $\pi_\theta$, the PWLL objective $\hat{U}_g(\pi_\theta)$ is concave in $\theta$. With $\ell_2$ regularization, it is strongly concave.
\end{proposition}

\cref{prop:concave} makes PWLL appealing for stochastic optimization. In \cref{sec:las-optimization-convergence}, we show that under standard assumptions of bounded feature norms $\|\phi(x, a)\|$ and weights $g(R_i, \pi_0(A_i, X_i))$, these objectives satisfy the regularity conditions necessary to invoke established convergence theorems \citep{garrigos2023}. This allows us to derive problem-dependent convergence guarantees: stochastic gradient ascent attains a global $\mathcal{O}(1/\sqrt{T})$ rate in the general (concave) case (\cref{thm:convex-constant}), accelerating to a geometric rate under $\ell_2$-regularization (\cref{thm:strongly-convex-constant}).

Beyond optimization properties, PWLL also admits a simple statistical interpretation. $\hat{U}_g(\pi)$ in \cref{eq:las-pwll} is a weighted log-likelihood: the term $\log \pi(A_i \mid X_i)$ performs standard behavior cloning, while the weight $g(R_i,\pi_0(A_i\mid X_i))$ determines how \emph{desirable}\footnote{By how desirable an action is, we mean how strongly this action should influence the learned policy.} each logged sample is. This turns off-policy learning into a form of logging-aware and reward-weighted maximum-likelihood estimation. Different choices of $g$ encode different notions of desirability. For example, the weighting
\begin{align*}
    g(r,\pi_0(a\mid x))=\frac{r}{\max\{\pi_0(a\mid x),\tau\}}
\end{align*}
emphasizes samples with high reward while reducing the influence of actions that the logging policy selected very frequently. At the same time, the clipping at $\tau$ prevents extremely rare actions from receiving disproportionately large weights, ensuring that their contribution is attenuated once $\pi_0(a\mid x)$ falls below the threshold. In this view, desirable samples are those that provide strong reward evidence without allowing very small propensities to dominate the updates. Many other PWLL variants arise from different choices of $g$ (see below), each specifying a distinct prioritization scheme for the logged data, while all benefit from the concavity induced by the logarithmic term.

To illustrate the qualitative difference between PWLL and IPS-based objectives, we construct a simple offline bandit problem with $K=3$ actions and visualize the resulting optimization landscapes in a two-parameter policy space. Concretely, we consider a non-contextual setting with deterministic mean rewards $r = (0.9,\,0.7,\,0.2)$ and a logging policy $\pi_{0}$ whose support places almost all mass on action 3 ($\pi_{0}(1) = 0.002$, $\pi_{0}(2) = 0.003$, $\pi_{0}(3) = 0.995$). We generate a fixed dataset of $n=60$ logged samples $(A_i, R_i)$ by drawing actions $A_i \sim \pi_0$  and binary rewards from the corresponding Bernoulli distributions, $R_i \sim {\rm Bern}(r(A_i))$. To obtain a two-dimensional visualization, we parameterize the target policy using a softmax over three logits: $\pi_\theta(a) = e^{\theta_a}/\sum_{b \in [3]} e^{\theta_b}$, fixing the logit associated with action 3 as $\theta_3=1$, and letting the remaining two logits be free parameters $(\theta_1,\theta_2)$.

\begin{figure}
    \centering
    \begin{subfigure}[b]{0.22\linewidth}
        \centering
        \includegraphics[width=\linewidth]{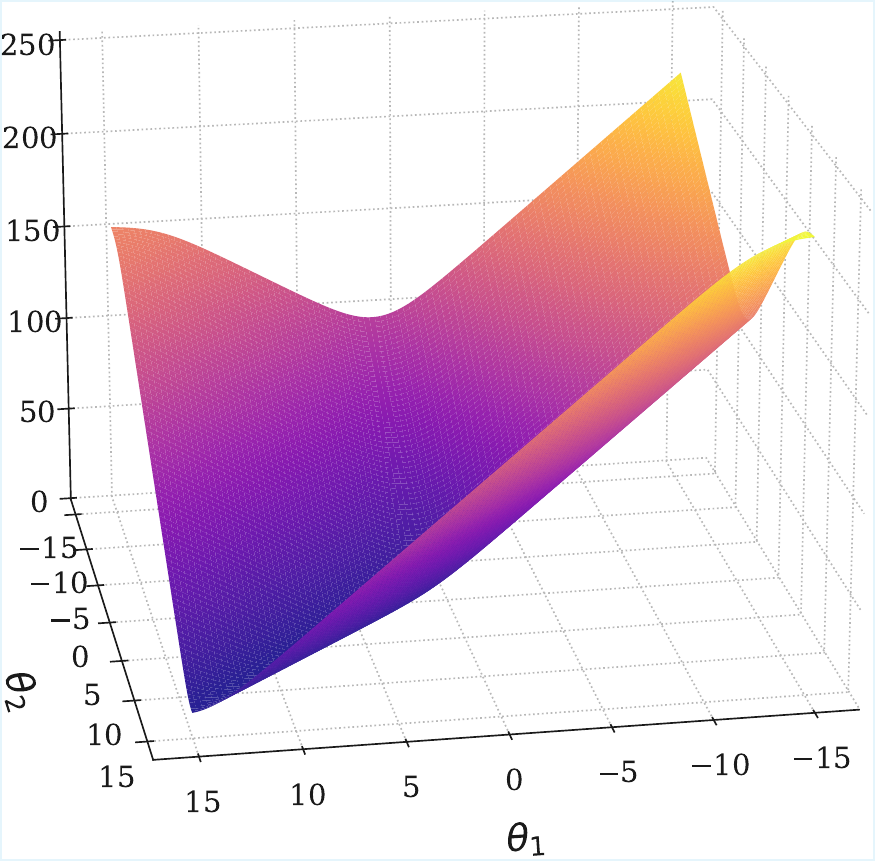}
        \caption{{\tiny PWLL (3D view)}}
        \label{fig:las-pwll_3d_wrap}
    \end{subfigure}
        \begin{subfigure}[b]{0.26\linewidth}
        \centering
        \includegraphics[width=\linewidth]{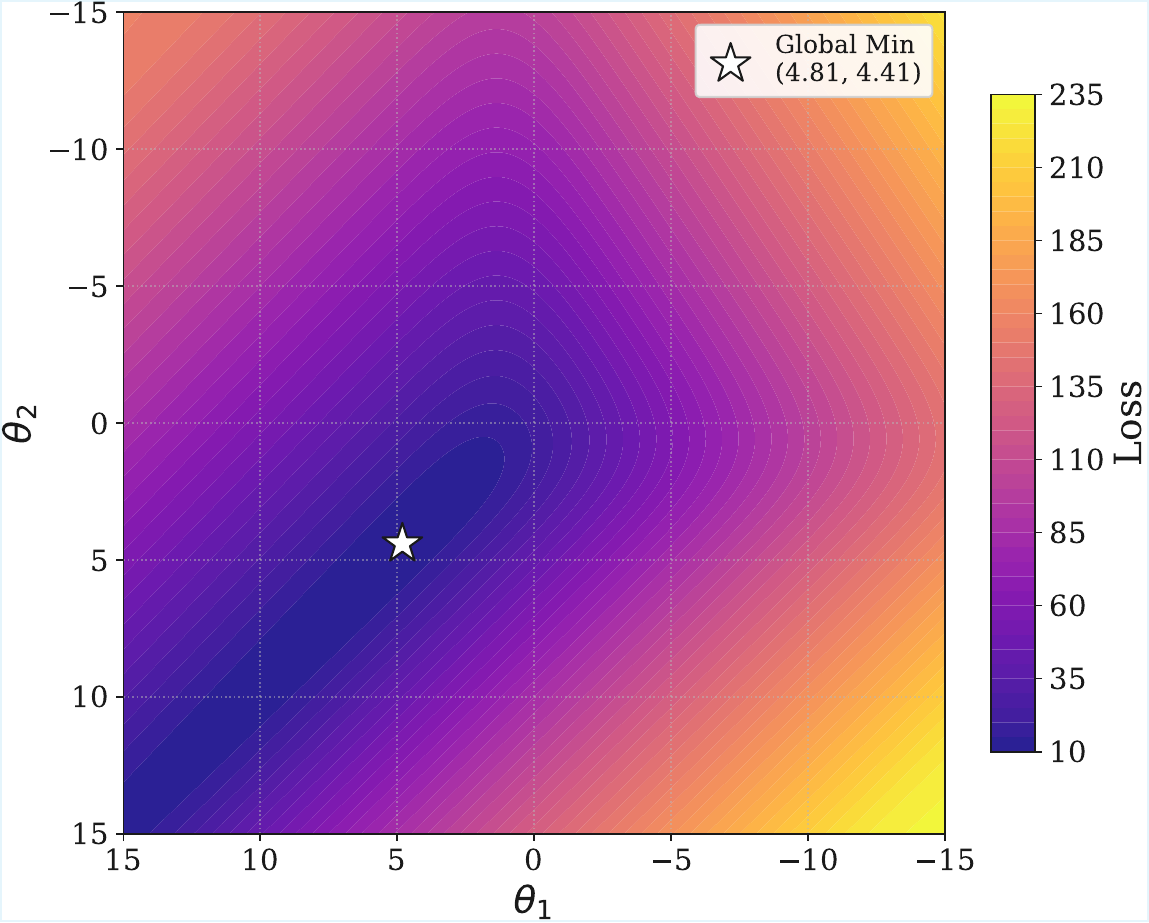}
        \caption{{\tiny PWLL (2D projection)}}
        \label{fig:las-pwll_2d_wrap}
    \end{subfigure}
    \hfill 
    \begin{subfigure}[b]{0.22\linewidth}
        \centering
        \includegraphics[width=\linewidth]{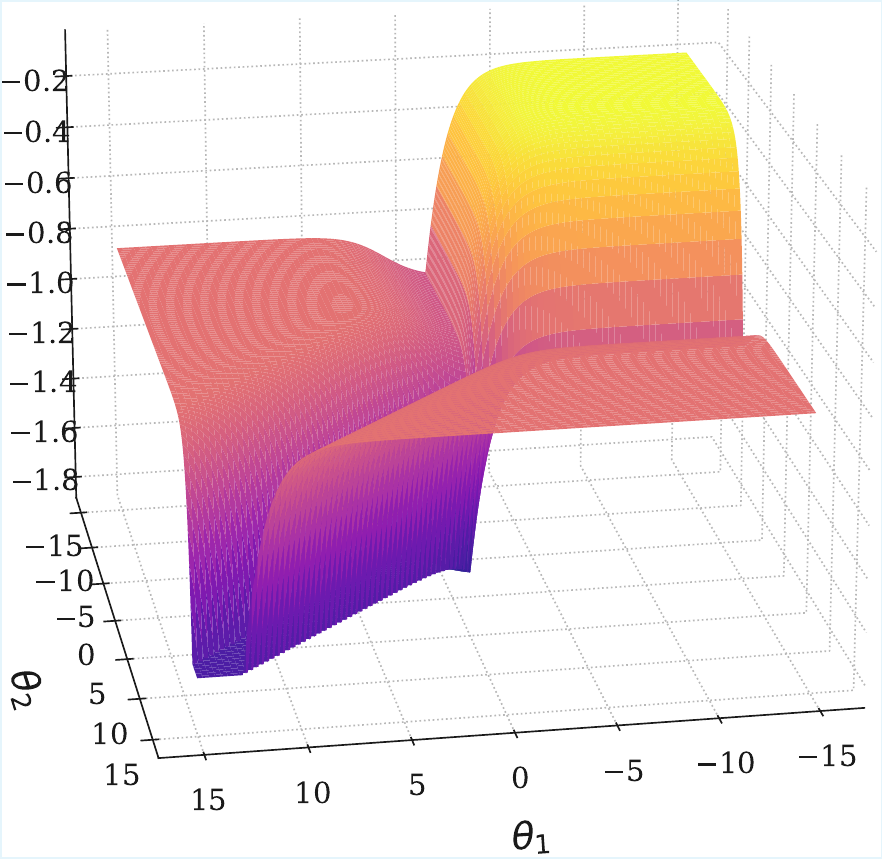}
        \caption{{\tiny IPS-based  (3D view)}}
        \label{fig:las-ope_3d_wrap}
    \end{subfigure}
    \begin{subfigure}[b]{0.26\linewidth}
        \centering
        \includegraphics[width=\linewidth]{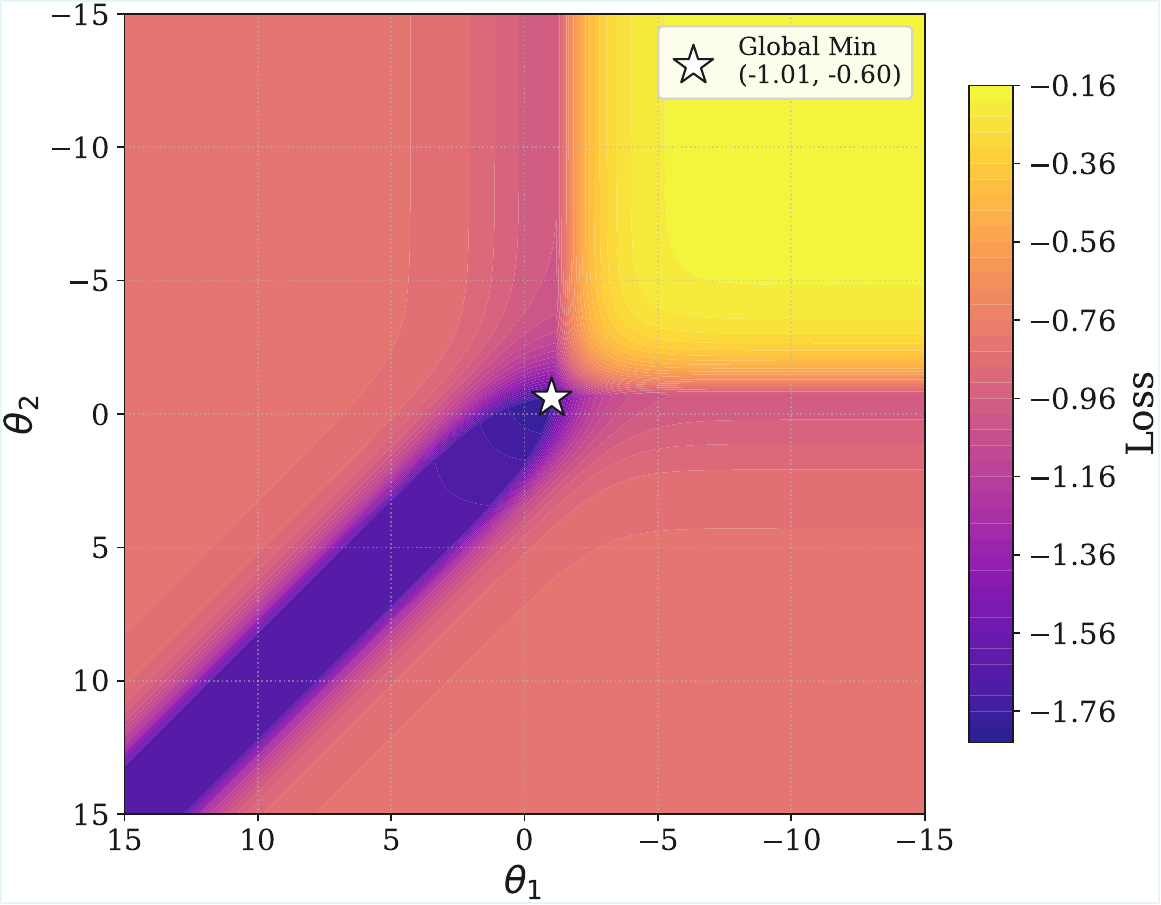}
        \caption{{\tiny IPS-based (2D projection)}}
        \label{fig:las-ope_2d_wrap}
    \end{subfigure}
\caption{Optimization landscapes on a toy example. PWLL (\texttt{cLPI}) vs IPS-based (\texttt{cIPS}).}    
\label{fig:las-toy_example}
\end{figure}

In \cref{fig:las-toy_example}, the PWLL landscape is concave with well-scaled gradients, and optimization trajectories converge reliably from roughly any initialization. In contrast, the IPS-based landscape consists of flat regions, separated by a narrow band of extremely steep curvature. This creates both vanishing and exploding gradients, severe ill-conditioning, and high sensitivity to initialization and learning rate. This aligns with the optimization pathologies in \cref{prop:plateaus,prop:local_maxima}.

\begin{remark}[Beyond linear-softmax policies]
The concavity guarantee of \cref{prop:concave} assumes linear-softmax policies. In many large-scale recommendation systems, a deep encoder is pre-trained and kept fixed, and only a final linear head is optimized for the downstream task; in this case, the policy is still linear-softmax in the trainable parameters, and PWLL objectives retain their concavity. When the full network is trained end-to-end, concavity no longer holds. Yet, PWLL's gradients $g(R_i, \pi_0(A_i|x_i))  \nabla_\theta \log \pi_\theta(A_i\mid X_i)$ match the structure of cross-entropy gradients, which are known to produce stable and well-scaled updates in deep architectures. Thus, even without formal guarantees, PWLL maintains substantially more benign optimization dynamics than IPS-based objectives.
\end{remark}
 
\textbf{Local policy improvement (\texttt{LPI}).} \citet{liang2022local} set $g(r, p_0) = r$, which optimizes the log-likelihood of actions weighted by their observed rewards:
\begin{align}\label{eq:las-lpi_value}
  \hat{U}_{\textsc{lpi}}(\pi) = \frac{1}{n} \sum_{i=1}^n R_i \log \pi(A_i\mid X_i)\,.
\end{align}
The oracle policy balances reward-seeking with imitation of the logging policy:
\begin{align}\label{eq:las-lpi_optimal_policy}
    \pi^{\texttt{LPI}}_*(a \mid x) \propto r(x, a) \pi_0(a \mid x)\,.
\end{align}

\textbf{Clipped LPI (\texttt{cLPI}).} uses importance-weight clipping, setting $g(r, p_0) = \tfrac{r}{\max(p_0, \tau)}$:
\begin{align}\label{eq:las-clpi_value}
\hat{U}_{\textsc{clpi}}(\pi)
    = \frac{1}{n} \sum_{i=1}^n
    \frac{R_i}{\max\{\pi_0(A_i \mid X_i), \tau\}}
     \log \pi(A_i \mid X_i)\,.
\end{align}
In a similar spirit to \texttt{cIPS}, its oracle policy corrects for action frequency under $\pi_0$, down-weighting the influence of rare actions due to the clipping:
\begin{align}\label{eq:las-clpi_optimal_policy}
\pi^{\texttt{cLPI}}_*(a \mid x)
\propto
r(x, a) \frac{\pi_0(a \mid x)}{\max\{\pi_0(a \mid x), \tau\}} .
\end{align}

\textbf{KL regularization (\texttt{RegKL}).} To further amplify the reward signal relative to the logging policy prior, \texttt{RegKL} uses an exponential weighting function $g(r, p_0) = \exp(r / \beta)$:
\begin{align}\label{eq:las-regkl_value}
    \hat{U}_{\textsc{regkl}}(\pi)
    = \frac{1}{n} \sum_{i=1}^n
\exp(R_i / \beta) \log \pi(A_i \mid X_i)\,.
\end{align}
The oracle policy is proportional to the logging policy, weighted by the exponentiated reward:
\begin{align}\label{eq:las-kl_optimal_policy}
    \pi^{\texttt{RegKL}}_*(a \mid x) \propto
    \mathbb{E}_{r \sim p(\cdot \mid x, a)}\bigl[\exp(r/\beta) \bigr]\pi_0(a \mid x) .
\end{align}
The temperature parameter $\beta$ smoothly interpolates between behavior cloning ($\beta \to \infty$) and greedy reward maximization ($\beta \to 0$).

Note that \texttt{BPR} \citep{rendle2012bpr} can be seen as an approximate PWLL objective, and we included it in our experiments. In fact, this general form of PWLL lends itself to numerous variations by modifying the weighting function $g$. For instance, one could introduce variants inspired by regularized IPS like exponential smooting (\cref{chap:ES}). While many such variants can be proposed for specific use cases, the central message of our work is that the well-behaved optimization landscape of the PWLL family is of greater practical importance than the estimation accuracy of IPS-based objectives. Thus, an exploration of these PWLL variants is beyond our scope. We contend that the foundational methods analyzed above, \texttt{LPI}, \texttt{cLPI}, and \texttt{RegKL}, along with the widely used \texttt{BPR} are sufficient to demonstrate the inherent advantages of PWLL objectives.

Finally, PWLL resembles reward- or advantage-weighted behavioral cloning objectives in RL \citep{nair2020awac,wang2020critic,peng2019advantage,peters2006reinforcement}. While those methods address multi-step MDPs and often focus on mitigating distributional shift and bootstrapping errors, we focus on offline contextual bandits with large action spaces: identifying objectives and parametrizations that remain optimizable as $K$ grows, rather than accurately estimating $V(\pi)$. PWLL is critic-free and uses logged rewards and propensities through a weighting function $g(R_i,\pi_0(A_i\mid X_i))$ that induces concave optimization landscapes for common policy classes. This yields substantial gains in large-$K$ bandits without the overhead of value-function estimation. PWLL’s optimization-centric perspective complements the usual KL-regularized or trust-region interpretations of these RL methods.

\section{Empirical Analysis}
\label{sec:las-analysis}
We conduct our empirical evaluation on three large-scale recommendation datasets: \texttt{MovieLens} ($K=60\text{k}$)~\citep{movielens}, \texttt{Twitch} ($K=200\text{k}$)~\citep{twitch}, and \texttt{GoodReads} ($K=1\text{M}$)~\citep{gr2}. These benchmarks feature action spaces with up to one million items, representing some of the largest settings studied in the offline policy learning literature. For all experiments, we employ the common softmax inner-product policies. We compare methods from both objective families. For IPS-based objectives, we include \texttt{IPS}, \texttt{ES}, \texttt{DR}, \texttt{MIPS}, \texttt{OffCEM}, \texttt{POTEC}, and \texttt{PC} in \cref{sec:las-ope_based}. For PWLL objectives, we evaluate \texttt{LPI}, \texttt{cLPI}, \texttt{RegKL}, and \texttt{BPR} in \cref{sec:las-PWLL_based}. All implementation details are provided in \cref{app:las-additional_experiments}.

\subsection{Optimization is the Main Bottleneck}
To test our central hypothesis that \emph{optimization challenges are a more significant barrier than estimation accuracy}, we evaluate how objectives perform under various optimization configurations. If an algorithm's success is highly dependent on specific hyperparameters like batch size or learning rate, it suggests a difficult, non-robust optimization landscape. This experiment directly probes the practical trainability of each method, a key aspect our paper argues is often overlooked.

\begin{figure}
  \centering
  \includegraphics[width=0.85\linewidth]{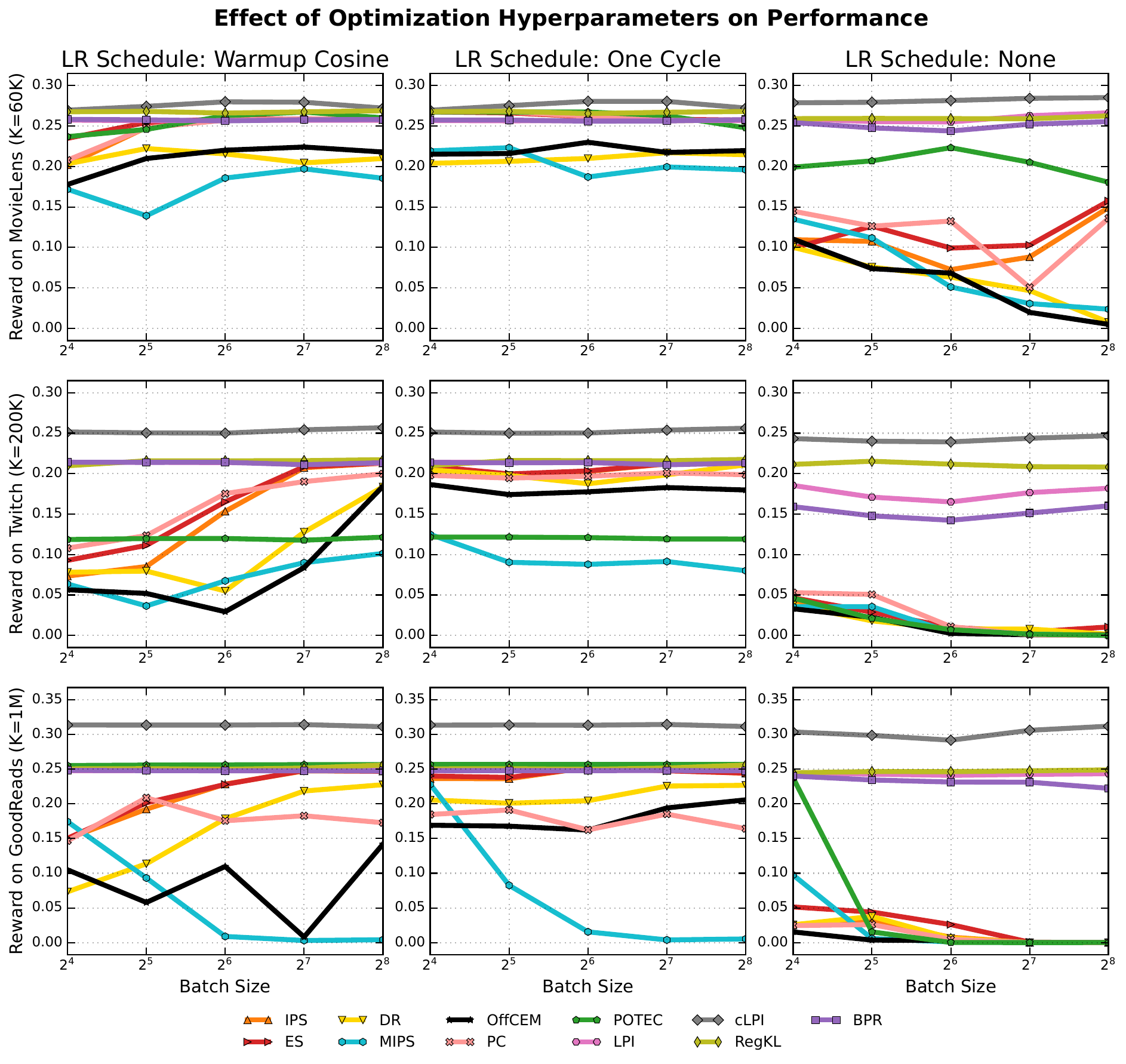}
  \caption{Effect of batch size and learning rate schedule on final validation reward using three large-scale datasets. IPS-based objectives are highly sensitive, while PWLL objectives are robust.}
  \label{fig:las-reward-vs-batch}
\end{figure}

The results strongly support our claim. As shown in \cref{fig:las-reward-vs-batch}, \emph{IPS-based objectives are highly sensitive} to batch size and learning rate schedule: minor changes can cause performance collapse, making them difficult to tune and train reliably. In contrast, \emph{PWLL objectives remain robust}, achieving consistently high reward across all configurations. This stability translates directly into better learned policies: \emph{PWLL objectives outperform IPS-based objectives on all datasets}. Even \texttt{POTEC}, a state-of-the-art method designed for large action spaces, is surpassed by the much simpler and easier-to-optimize \texttt{cLPI}.

One might assume that an objective designed for estimation fidelity, such as a low-MSE IPS-based estimator, would naturally yield a better policy. Our findings show this is not the case. The superiority of PWLL objectives, which are poor value estimators by design, provides compelling evidence against this estimator-centric view. This reinforces our main takeaway: in large action space settings, a tractable optimization landscape is a more critical feature for a learning objective than its statistical accuracy. For completeness, an experiment tracking the MSE of methods is given in \cref{app:las-additional_experiments}. 

The figure also supports \cref{claim:potec}. Indeed, there is a consistent performance gap between \texttt{POTEC} and \texttt{OffCEM}. Both methods are designed to maximize the same asymptotic objective as we show in \cref{sec:las-ope_based}; their statistical goals are identical. The divergence in performance, therefore, can be attributed entirely to their differing optimization strategies. \texttt{POTEC}'s use of a two-stage, cluster-level optimization proves far more effective than \texttt{OffCEM}'s naive, action-level parametrization.

\subsection{Objective-Aware Parametrization}

\begin{figure}
  \centering
  \includegraphics[width=0.9\linewidth]{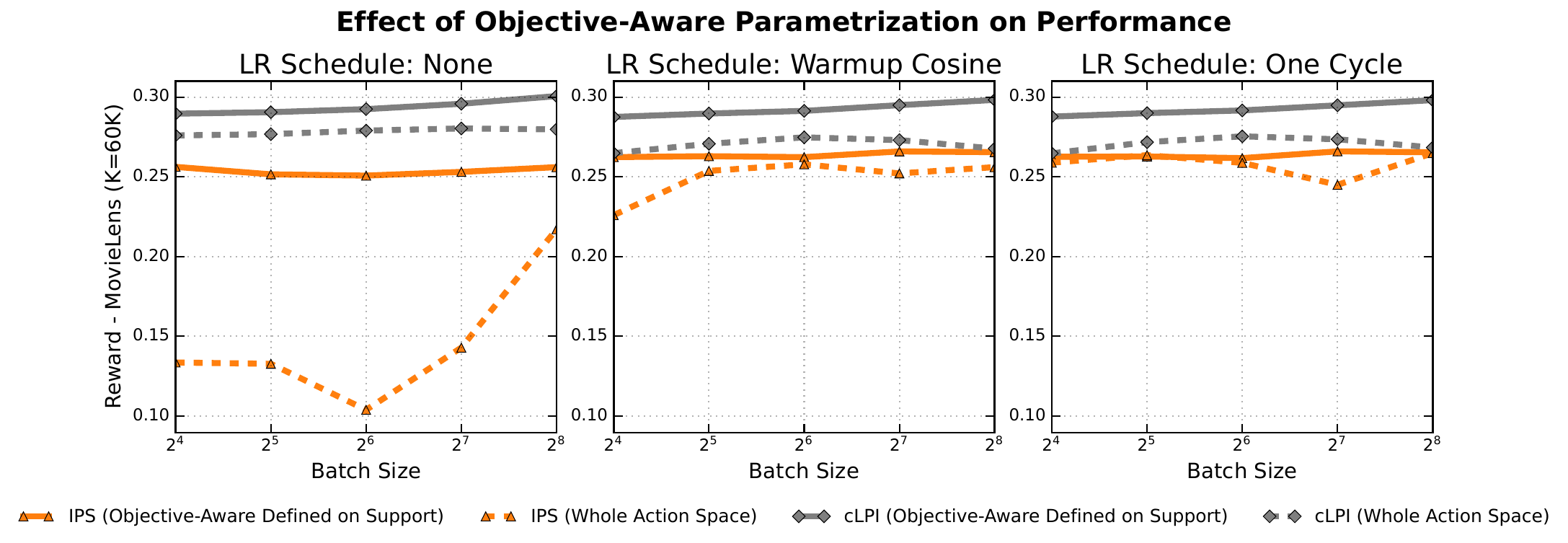}
  \caption{The effect of objective-aware parametrization for \texttt{IPS} and \texttt{cLPI} on \texttt{MovieLens}.}
  \label{fig:las-objective-aware}
\end{figure}

To empirically validate \cref{claim:ips}, we compare a naive, whole-action-space parametrization against our proposed objective-aware approach, which restricts the policy's effective action space to the logging policy support, $S_0$. As shown for the \texttt{IPS} objective in \cref{fig:las-objective-aware}, the naive approach is highly unstable, with performance collapsing under simple learning configurations. In contrast, the objective-aware version is very robust, achieving high reward consistently across all batch sizes and schedules. This benefit extends even to inherently stable PWLL objectives like \texttt{cLPI}, which achieve even better performance with the restricted support. This provides strong evidence for \cref{claim:ips}: aligning the policy structure with the objective's inductive bias simplifies the optimization landscape, leading to greater stability and superior learned policies. This finding holds across all datasets, with full results available in \cref{app:las-additional_experiments}.

\section{Conclusion}\label{sec:las-conclusion}

The dominant approach to off-policy learning focuses on developing sophisticated IPS-based estimators while neglecting a crucial factor: the optimization landscape. We demonstrated, both theoretically and empirically, that this landscape becomes prohibitively difficult to optimize in large action spaces, undermining the practical effectiveness of even state-of-the-art estimators.

Our analysis motivates two strategies. First, objective-aware policy parametrizations align the policy class with the estimator's inductive bias, reducing the effective search space. Second, PWLL objectives abandon value estimation entirely in favor of inherently concave optimization landscapes. Experiments confirm that this focus on optimization tractability yields more robust learning, reduced sensitivity to hyperparameters, and superior policies.

Our work has several limitations. First, PWLL objectives are not value estimators: they cannot be used for off-policy evaluation or policy selection (choosing the best policy from a finite candidate set) when accurate value estimates and their comparison are required. Second, the concavity guarantee of \cref{prop:concave} holds only for linear-softmax policies; when training deep networks end-to-end, PWLL retains favorable gradient structure but loses formal concavity guarantees, although IPS-based objectives face even more severe optimization challenges in this setting. Third, PWLL's oracle policies inherently depend on the logging policy (e.g., $\pi^{\texttt{LPI}}_* \propto r(x,a) \pi_0(a \mid x)$), which may be suboptimal when $\pi_0$ has poor coverage of high-reward actions; however, this limitation is shared by IPS-based methods, whose oracle policies similarly depend on $\pi_0$'s support.

%% file: contents/exp_smoothing/main_text.tex
Having explored structured direct methods in \cref{chap:sDM} and optimization-focused objectives in \cref{chap:las}, we now turn to inverse propensity scoring (IPS). Despite its current practical limitations in large action spaces, many practitioners remain committed to IPS-based methods for their unbiasedness and theoretical guarantees, which enable principled safe off-policy learning. In this chapter, we improve IPS through \emph{exponential smoothing}, a differentiable importance-weight regularization technique that enables a controlled bias-variance trade-off. Then, we adopt the pessimistic framework introduced in \cref{sec:offline_background}, deriving principled uncertainty penalties for our regularized estimators for safe policy learning. Prior work on pessimistic off-policy learning has derived objectives from generalization bounds \citep{swaminathan2015batch,london2019bayesian}, but these approaches suffer from critical limitations: (i) they provide only one-sided bounds that fail to control estimation error in absolute value, limiting their ability to certify estimator quality, (ii) the resulting bounds are intractable and incompatible with stochastic optimization, and (iii) the pessimistic objectives require careful hyperparameter tuning.

We address these limitations by deriving \emph{tractable two-sided} PAC-Bayes generalization bounds that can be optimized directly via stochastic gradient ascent. Unlike prior work \citep{sakhi2022pac}, our analysis applies to standard IPS without assuming bounded importance weights, requiring only bounded second moments. Our bounds reveal that the optimal importance-weight smoothing parameter $\alpha$ depends on the quality of the logging policy. Furthermore, our framework generalizes to a broad class of importance-weight regularization techniques, yielding unified pessimistic objectives that enable fair comparison across different importance-weight regularization techniques. We present this extension in the final two sections of this chapter.

The chapter is organized as follows. \cref{sec:es-setting} presents background on regularized IPS and pessimism. \cref{sec:es-ope} identifies the shortcomings of hard clipping and introduces our exponential smoothing estimators. \cref{sec:es-opl} leverages PAC-Bayes theory to derive two-sided generalization bounds within the pessimistic framework. \cref{sec:es-discussion} discusses implications of our results. \cref{sec:es-experiments} demonstrates favorable performance across diverse benchmarks. Finally, \cref{sec:es-extension,sec:other-experiments} extend the framework to other importance-weight regularizations and compare them under a unified pessimistic objective.

\section{Background}
\label{sec:es-setting}
We consider the off-policy setting in \cref{sec:offline_background}, where we have access to logged data $\mathcal{D}_n = \{(X_i, A_i, R_i)\}_{i=1}^n$ collected by a known logging policy $\pi_0$. As additional notation, we let $\mu_\pi$ be the joint distribution of $(X, A, R)$; 
\begin{align*}
    \mu_\pi(x, a, r) = \nu(x)\pi(a|x)p(r|x, a)\,, \quad \text{so that} \quad (X_i, A_i, R_i) \sim \mu_{\pi_0}\,.
\end{align*}
Our goal remains to find a policy $\hat{\pi} \in \Pi$ that maximizes the value $$V(\pi) = \mathbb{E}_{X \sim \nu, A \sim \pi(\cdot | X)}[r(X, A)]\,.$$ 

\subsection{Regularized IPS}

This chapter focuses on the IPS estimator \citep{horvitz1952generalization,dudik2012sample}, which estimates the value $V(\pi)$ by re-weighting the samples as
\begin{align}\label{eq:es-ips_policy_value}
    \hat{V}_{\textsc{ips}}(\pi) = \frac{1}{n} \sum_{i=1}^n R_i \, w(A_i | X_i),
\end{align}
where $w(a | x) = \pi(a | x)/\pi_0(a | x)$ are the \emph{importance weights}. While IPS provides an unbiased estimate of $V(\pi)$ when the common support condition holds (i.e., $\pi_0(a|x)=0$ implies $\pi(a|x)=0$), its variance grows with these importance weights \citep{swaminathan2017off}, which can be arbitrarily large when the target policy $\pi$ and logging policy $\pi_0$ differ significantly. To mitigate this variance issue, it is common to transform the importance weights using regularization functions that introduce controlled bias to reduce variance. A regularized IPS estimator takes the form:
\begin{align}\label{eq:reg_ips_policy_value}
    \hat{V}(\pi) = \frac{1}{n} \sum_{i=1}^n R_i \, \hat{w}(A_i | X_i),
\end{align}
where $\hat{w}(a | x) \leq w(a | x) $ are the regularized importance weights. A common importance-weight regularization approach is clipping where $\hat{w}(a | x) = \min(\frac{\pi(a | x)}{\pi_0(a | x)}, M)$, $M>0$.

\subsection{Pessimistic Objectives}

Within the pessimistic framework introduced in \cref{sec:offline_background}, we seek to maximize: $\hat{\pi} = \argmax_{\pi \in \Pi} [ \hat{V}(\pi) - \operatorname{pen}(\pi) ]$ where $\operatorname{pen}(\cdot)$ is a penalty term. The construction of this penalty has been approached in various ways, but generally relies on lower confidence bounds on the policy value:

\textbf{Evaluation bounds} \citep{metelli2021subgaussian} provide confidence intervals for a \emph{fixed} target policy $\pi$, showing that with probability at least $1-\delta$:
\begin{align}\label{eq:eval_bound}
    |V(\pi) - \hat{V}(\pi)| \leq f(\delta, \pi, \pi_0, n).
\end{align}
Essentially, \cref{eq:eval_bound} indicates that for a fixed policy $\pi \in \Pi$, the event $|V(\pi) - \hat{V}(\pi)| \leq f(\delta, \pi, \pi_0, n)$ holds with high probability. However, this event depends on the target policy $\pi$. Thus \cref{eq:eval_bound} is useful for evaluating a \emph{single target policy} when having access to \emph{multiple logged data sets $\cD_n$}. This poses a problem for off-policy learning, where we optimize over a potentially \emph{infinite space of policies} using a \emph{single logged data set $\cD_n$}. This is the fundamental theoretical limitation of using evaluation bounds similar to \cref{eq:eval_bound} in off-policy learning. While one can transform \cref{eq:eval_bound} into a generalization bound that holds uniformly over all $\pi \in \Pi$ via a union bound, this typically introduces intractable complexity terms, making the resulting pessimistic objectives, which maximize the lower confidence bound, equally intractable.

\textbf{One-sided generalization bounds} \citep{swaminathan2015batch,london2019bayesian,sakhi2022pac} address this limitation by providing bounds that hold simultaneously for all policies $\in \in \Pi$. For $\delta \in (0, 1)$, with probability at least $1-\delta$:
\begin{align}\label{eq:off-policy learning_one_sided}
    V(\pi) \geq \hat{V}(\pi) - g(\delta, \Pi, \pi, \pi_0, n), \quad \forall \pi \in \Pi,
\end{align}
where the function $g$ now depends on the policy space $\Pi$. This leads to the pessimistic objective:
\begin{align}\label{eq:objective}
    \hat{\pi} = \argmax_{\pi \in \Pi} \hat{V}(\pi) - g(\delta, \Pi, \pi, \pi_0, n).
\end{align}

However, one-sided bounds fail to attest to the quality of the estimator. To illustrate, consider a degenerate estimator $\hat{V}^{\textsc{poor}}(\pi) = 0$ for all $\pi \in \Pi$. Since $V(\pi) \in [0, 1]$, we trivially have a one-sided bound $V(\pi) \geq  \hat{V}^{\textsc{poor}}(\pi)$ with probability 1, yet this estimator is entirely uninformative about the true rewards.

\textbf{Two-sided generalization bounds} resolve this issue by controlling both the upper and lower deviations leading to
\begin{align}\label{eq:off-policy learning_two_sided}
    |V(\pi) - \hat{V}(\pi)| \leq g(\delta, \Pi, \pi, \pi_0, n), \quad \forall \pi \in \Pi.
\end{align}
These bounds ensure the estimator quality and enable oracle inequalities of the form $V(\hat{\pi}) \geq V(\pi_*) - 2g(\delta, \Pi, \pi_*, \pi_0, n)$, where $\hat{\pi}$ is learned using \cref{eq:objective} and $\pi_* = \argmax_{\pi \in \Pi} V(\pi)$ is the optimal policy. This shows the appeal of pessimism: the suboptimality gap depends on the bound evaluated at the optimal policy $\pi_*$, meaning the estimator only needs to be precise for near-optimal policies rather than uniformly across the policy class $\Pi$.

\textbf{Alternative approaches} include heuristics that simplify theoretical bounds for tractability \citep{swaminathan2015batch,london2019bayesian,wang2023oracle}, often penalizing by empirical variance or policy divergence while discarding complexity terms. Recent work on implicit pessimism \citep{gabbianelli2023importance,Sakhi2024LS} (published after the work in this chapter) shows that careful analysis of specific importance-weight regularizations can yield bounds where the penalty is policy-independent. In this case, maximizing the lower bound reduces to maximizing the estimator directly: the pessimism becomes implicit in the regularization itself.

In this work, we derive a tractable two-sided PAC-Bayesian generalization bound for our exponential smoothing estimator and then generalize it to other importance-weight regularizations.

\section{Exponential Smoothing}
\label{sec:es-ope} 
Importance-weight clipping \citep{swaminathan2015batch} yields the following commonly used estimators
\begin{align}\label{eq:es-clip_ips_policy_value}
 \texttt{IPS-min} \quad  \tilde{V}^{\textsc m}(\pi)&= \frac{1}{n} \sum_{i=1}^n R_i \min\big(w(A_i | X_i), M\big)\,,\nonumber\\
    \texttt{IPS-max} \quad \hat{V}^\tau(\pi)&= \frac{1}{n} \sum_{i=1}^n R_i \frac{\pi(A_i | X_i)}{\max(\pi_0(A_i | X_i), \tau)}\,.
\end{align}
Here \texttt{IPS-min} clips the weights while \texttt{IPS-max} only clips $\pi_0$ in the denominator since $\pi$ is always smaller than 1. For instance, $M \in \real^+$ in $\tilde{V}^{\textsc m}(\pi)$ trades the bias and variance of the estimator. When $M$ is large, the bias of $\tilde{V}^{\textsc m}(\pi)$ is small but its variance may be large. On the other hand, the variance goes to $0$ when $M \approx 0$ since in that case $\tilde{V}^{\textsc m}(\pi) \approx 0$ for any $\pi \in \Pi$. Similarly, $\tau \in [0, 1]$ trades the bias and variance of $\hat{V}^\tau(\pi)$ and can be seen as $\tau \approx \frac{1}{M}$.

This \emph{hard} clipping has some limitations. First, $\min(\cdot, M)$ leads to non-differentiable objectives that may require additional care in optimization \citep{papini2019optimistic}. Also, $\min(\cdot, M)$ is constant on $[M, \infty)$ leading to objectives with zero gradients for any policy $\pi$ that satisfies $w(A_i | X_i) > M$ for any $i \in [n]$. More importantly, hard clipping is sensitive to the choice of the clipping threshold $M$. In practice, tuning $M$ is challenging and may cause the learned policy to match the logging policy, leading to minimal improvements. To see this, consider the following illustrative example.

For simplicity, suppose that the problem is non-contextual, in which case the reward function $r$ only depends on the actions $a \in \cA$. It follows that policies do not depend on $x \in \cX$; they are now probability distributions $\pi(\cdot)$ over $\cA$. Also, assume that $\cA = [100]$ and that the reward received after taking action $a \in [100]$ is binary. That is, $R \sim {\rm Bern}(r(a))$ where $r(a) = 0.1 - 10^{-3}(a-1)$ is the expected reward of action $a$, and for any $p \in [0, 1],$ ${\rm Bern}(p)$ is the Bernoulli distribution with parameter $p$. This means that the best action is $1$ and the worst is $100$. Finally, the logging policy $\pi_0(\cdot)$ is $\epsilon$-greedy centered at action $50$. That is $\pi_0(50) = 1-\epsilon$, and for any $a \neq 50$, $\pi_0(a) = \frac{\epsilon}{99}$, with $\epsilon=0.05$. 

\begin{figure}[ht]
\begin{center}
\centerline{\includegraphics[width=0.7\columnwidth]{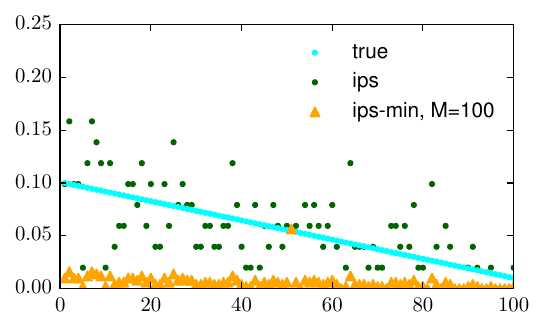}}
\caption{Effect of hard clipping on the estimation quality. The $x$-axis corresponds to actions $a \in [100]$. The $y$-axis is the estimated reward of each of the $100$ policies $\pi_a$ using either \texttt{IPS} or \texttt{IPS-min}. The cyan line is the true reward for each policy $\pi_a$.}
\label{fig:es-example}
\end{center}
\end{figure}

Now consider $100$ deterministic policies $\pi_a(\cdot)$ for $a \in [100]$ such that $\pi_a(\cdot)$ is the Dirac distribution centered at $a$. In \cref{fig:es-example}, we plot the estimated reward of the policies $\pi_a$ using either \texttt{IPS} in \cref{eq:es-ips_policy_value} or \texttt{IPS-min} in \cref{eq:es-clip_ips_policy_value}. We generate $n=50{\rm k}$ samples and set $M=100 = \mathcal{O}(\sqrt{n})$ as suggested by \citet{ionides2008truncated}. \emph{With this choice of $M$}, \texttt{IPS-min} underestimates the reward of all policies $\pi_a$ for $a \neq 100$ since their weights $\pi_a/\pi_0$ are either $0$ or $ 99/ \epsilon>M$. The estimated reward of \texttt{IPS-min} is maximized in $\pi_{50} \approx \pi_0$ only. Thus, if we optimize $\tilde{V}^{\textsc m}(\cdot)$ over Dirac policies, we will converge to the logging policy despite its bad performance.

Although the other variant of hard clipping, \texttt{IPS-max} in \cref{eq:es-clip_ips_policy_value}, is differentiable, it is still sensitive to $\tau$ and may induce high bias similar to \cref{fig:es-example}. This is due to some loss of information related to the preferences of the logging policy. Indeed, for two actions $a$ and $a^\prime$ such that $\pi_0(a  | X_i) \ll \pi_0(a^\prime  | X_i) < \tau$ for an observed context $X_i$, the propensity scores $\pi_0(a  | X_i)$ and $\pi_0(a^\prime  | X_i)$ will be clipped to the same value $\tau$. Thus the information that, for context $X_i$, action $a^\prime$ is preferred by the logging policy than action $a$ will be lost.

To mitigate this, we propose the following \emph{exponential smoothing} correction for IPS. Our estimators are defined as
\begin{align}\label{eq:es-exp_ips_policy_value}
\hspace{-0.3cm} \texttt{IPS-}\alpha:   \, \,    \hat{V}^\alpha(\pi)
  = \frac{1}{n} \sum_{i=1}^n R_i \hat{w}^\alpha(A_i | X_i)\,, \, \, \alpha \in [0, 1]\,, \nonumber\\
\hspace{-0.3cm} \texttt{IPS-}\beta: \, \,    \tilde{V}^\beta(\pi) 
= \frac{1}{n} \sum_{i=1}^n R_i \tilde{w}^\beta_{\pi}(A_i | X_i)\,, \, \, \beta \in [0, 1]\,,
\end{align}
where $\hat{w}^\alpha(a | x) = \frac{\pi(a | x)}{\pi_0(a | x)^\alpha}$ and $\tilde{w}^\beta_{\pi}(a | x) = \frac{\pi(a | x)^\beta}{\pi_0(a | x)^\beta}$. Here standard IPS is recovered for $\alpha=1$ and $\beta=1$. These estimators yield smooth, everywhere-differentiable objectives and avoid the flat regions induced by hard clipping; this improves optimization in practice. Also, in contrast with \texttt{IPS-max} in \cref{eq:es-clip_ips_policy_value}, $\hat{V}^\alpha(\pi)$ preserves the preferences of the logging policy. Precisely, for two actions $a$ and $a^\prime$ such that $\pi_0(a  | X_i) < \pi_0(a^\prime  | X_i)$ for an observed context $X_i$, we still have $\pi_0(a  | X_i)^\alpha < \pi_0(a^\prime  | X_i)^\alpha$ and the information that action $a^\prime$ is preferred by the logging policy than action $a$ is preserved.

While a similar correction to \texttt{IPS-}$\beta$ was proposed in \citet{korba2022adaptive}, its use in off-policy learning is novel. Also, \citet{su2020doubly,metelli2021subgaussian} regularized the importance weights $w$ as $ \frac{\lambda_1 w}{\lambda_1 + w^2}\,, \lambda_1>0$ and   $ \frac{w}{1-\lambda_2 + \lambda_2 w }\,, \lambda_2 \in [0, 1]$, respectively. Thus, the expression of both corrections is very different from ours. More importantly, these corrections entail different properties than ours. Roughly speaking, our correction allows us to \emph{simultaneously} \textbf{(1)} control a tuning parameter $\alpha \in [0, 1]$ that is in a bounded domain $[0, 1]$, \textbf{(2)} without constraining the resulting importance weights to be bounded, \textbf{(3)} and to obtain tractable PAC-Bayes generalization bounds as the correction $\frac{\pi}{\pi_0^\alpha}$ is linear in $\pi$; a technical requirement of PAC-Bayes analysis. In contrast, \citet{metelli2021subgaussian,su2020doubly} do not provide generalization guarantees; they focus on estimation accuracy (e.g., through mean squared error) and only propose heuristics for off-policy learning. Those heuristics are not based on theory, in contrast with ours which is directly derived from our generalization bound. Also, our approach has favorable empirical performance (\cref{app:other_corrections}).

Although \citet[Lemma 1]{korba2022adaptive} show that smoothing the importance weights similarly to \texttt{IPS-}$\beta$ in \cref{eq:es-exp_ips_policy_value} reduces the variance, it might still be unclear how $\alpha$ and $\beta$ trade the bias and variance of our estimators in off-policy learning. To see this, let $\alpha \in [0, 1]$, then we have
\begin{align}\label{eq:es-alpha_bias_variance} 
    |\mathbb{B}(\hat{V}^\alpha(\pi))|  & \leq \E{X \sim \nu, A \sim \pi(\cdot | X)}{1 - \pi_0(A | X)^{1-\alpha}} \,,\\
\mathbb{V}\left[\hat{V}^\alpha(\pi)\right] & \leq \frac{1}{n} \mathbb{E}_{X \sim \nu, A \sim \pi(\cdot | X)}\big[ \frac{\pi(A | X)}{\pi_0(A | X)^{2\alpha-1}} \big] \nonumber\,,
\end{align}
with $\mathbb{B}(\hat{V}^\alpha(\pi)) = \mathbb{E}[\hat{V}^\alpha(\pi)] - V(\pi)$  and $\mathbb{V}[\hat{V}^\alpha(\pi)] = \mathbb{E}[(\hat{V}^\alpha(\pi)-\mathbb{E}[\hat{V}^\alpha(\pi)])^2]$ are respectively the bias and the variance of $\hat{V}^\alpha(\pi)$. The bound of the bias in \cref{eq:es-alpha_bias_variance} is minimized in $\alpha=1$ (standard IPS); in which case it is equal to 0 (standard IPS is unbiased). In contrast, the bound of the variance is minimized in $\alpha =0$. Thus if the variance is small or $n$ is large enough such that $\mathbb{E}[ \pi(A | X)/\pi_0(A | X)^{2\alpha-1} ] /n \rightarrow 0$, then we set $\alpha  \rightarrow 1$. Otherwise, we set $\alpha  \rightarrow 0$. This shows that $\alpha$ trades the bias and variance of $\hat{V}^\alpha$. More details and a similar discussion for $ \tilde{V}^\beta(\pi)$ are deferred to \cref{proofs:es-ope}.

\section{PAC-Bayes Analysis for Off-Policy Learning}\label{sec:es-opl}

We now derive generalization bounds for our estimator. We opt for the PAC-Bayes framework for the following reasons. First, it is known to provide some of the tightest generalization bounds in challenging scenarios \citep{farid2021generalization}, for aggregated and randomized predictors \citep{alquier2021user}. Second, the bounds have a Kullback–Leibler (KL) divergence \citep{van2014renyi} term $D_{\mathrm{KL}}(\mathbb{Q} \| \mathbb{P})$ that depends on a \emph{fixed prior} $\mathbb{P}$ and a \emph{learning posterior} $\mathbb{Q}$ (see \cref{subsec:es-pac_bayes_framekwork} for a brief introduction). This quantity can be seen as a complexity measure, similarly to the covering number \citep{maurer2009empirical}. The difference is that complexity measures are uniform on the space of policies while the KL term in PAC-Bayes depends on the prior $\mathbb{P}$ and the posterior $\mathbb{Q}$. This allows getting sharper bounds when the former is well chosen. Third, the PAC-Bayes perspective fits very well with off-policy learning. In fact, a policy $\pi$ can be written as an aggregation of predictors under some distribution $\mathbb{Q}$. Thus the prior $\mathbb{P}$ can be associated with the logging policy $\pi_0$ that we want to improve upon while the posterior $\mathbb{Q}$ is related to the learning policy $\pi$. Fourth, \citet{london2019bayesian} showed that PAC-Bayes can lead to tractable and scalable objectives, an important consideration for this thesis.

\subsection{Elements of PAC-Bayes }\label{subsec:es-pac_bayes_framekwork}

Let $\mathcal{Z} = \mathcal{X} \times \mathcal{Y}$ be an instance space: e.g., $\mathcal{X}$ and $\mathcal{Y}$ are the input and output space in supervised learning. Let $\mathcal{H} = \set{h : \cX \rightarrow \mathcal{Y}}$ denote a hypothesis space of mappings from $\mathcal{X}$ to $\mathcal{Y}$ (predictors). Also, let $L : \mathcal{H} \times \mathcal{Z} \rightarrow \real$ be a loss function and assume access to data $\cD_n = (Z_i)_{i \in [n]}$ drawn from an unknown distribution $\mathbb{D}$. Let 
\[
\operatorname{Risk}(h)= \E{Z \sim \mathbb{D}}{L(h, Z)}
\]
be the risk of $h \in \mathcal{H}$ while 
\[
\widehat{\operatorname{Risk}}_n(h)= \frac{1}{n} \sum_{i=1}^n L(h, Z_i)
\]
is its empirical counterpart. Then the main focus in PAC-Bayes is to study the generalization capabilities of random hypotheses $\mathbb{Q}$ on $\mathcal{H}$ by controlling the gap between the expected risk under $\mathbb{Q}$, $\E{h \sim \mathbb{Q}}{\operatorname{Risk}(h)}$, and the expected empirical risk under $\mathbb{Q}$, $\E{h \sim \mathbb{Q}}{\widehat{\operatorname{Risk}}_n(h)}$. 

For example, assume that $L(h, Z) \in [0,1]$ for any $(h, Z) \in \mathcal{H} \times \mathcal{Z}$, let $\mathbb{P}$ be a \emph{fixed prior} distribution on $\mathcal{H}$ and let $\delta \in (0, 1)$. Then with probability at least $1-\delta$ over $\cD_n \sim \mathbb{D}^n$, the following inequality holds \emph{simultaneously for any posterior} distribution $\mathbb{Q}$ on $\mathcal{H}$:
\begin{align*} 
\E{h \sim \mathbb{Q}}{\operatorname{Risk}(h)} 
\leq \E{h \sim \mathbb{Q}}{\widehat{\operatorname{Risk}}_n(h)} 
+ \sqrt{ \frac{D_{\mathrm{KL}}(\mathbb{Q} \| \mathbb{P})+\log \tfrac{2\sqrt{n}}{\delta}}{2n}} \,.
\end{align*} 
This was originally proposed by \citet{mcallester1998some}, and the reader may refer to \citet{alquier2021user,guedj2019primer} for more elaborate introductions of PAC-Bayes theory.

\textbf{Connection to value functions.} The loss $L$ is often chosen as the negative reward, $L(h,Z) = -r(h,Z)$. In this case, minimizing the $\operatorname{Risk}(h)$ is equivalent to maximizing the value function. Thus, PAC-Bayes bounds on the risk directly translate into guarantees on the discrepancy between empirical and true value, providing a principled way to reason about generalization in off-policy learning.

\subsection{PAC-Bayes for Off-Policy Learning}\label{pac_bayes_for_opl}
Let $\mathcal{H}=\{h: \cX \rightarrow \cA\}$ be a hypothesis space of mappings from $\cX$ (contexts) to $\cA$ (actions). Given a policy $\pi$ and a context $x \in \cX$, the action distribution $\pi(\cdot|x)$ is induced by a distribution $\mathbb{Q}$ over $\mathcal{H}$ \citep{london2019bayesian} such as
\begin{align}\label{eq:es-pac_policies}
   &\pi(a |x) = \pi_{\mathbb{Q}}(a | x) = \E{h \sim \mathbb{Q}}{\mathds{1}_{\{h(x)=a\}}}\,.
\end{align}
This is not an assumption since any policy $\pi$ has this form when $\mathcal{H}$ is rich enough \citep[Theorem 2]{sakhi2022pac}. From \cref{eq:es-pac_policies}, we observe that policies can be seen as an aggregation $\E{h \sim \mathbb{Q}}{\cdot}$ (under some distribution $\mathbb{Q}$ on the pre-defined hypothesis space $\mathcal{H}$) of deterministic decision rules $\mathds{1}_{\{h(x)=a\}}$. This allows formulating off-policy learning as a PAC-Bayes problem. Before showing how this is achieved, we start by providing two practical policies of such form.

\textbf{Example 1 (softmax and mixed-logit policies).} We define the hypothesis space $\mathcal{H} = \set{h_{\theta, \gamma} \, ; \theta \in \real^{dK}, \gamma \in \real^K}$ of mappings $h_{\theta, \gamma}(x) = \argmax_{a \in \cA} \phi(x)^\top \theta_a + \gamma_a$. Here $\phi(x)$ outputs a $d$-dimensional representation of $x$, and $\gamma_a$ is a standard Gumbel perturbation, $\gamma_a \sim {\rm G}(0, 1)$ for any $a \in \cA$. Then
\begin{align}\label{eq:es-softmax_pac_bayes}
    \pi^{\textsc{sof}}_{\theta}(a | x) &= \frac{\exp(\phi(x)^\top \theta_a)}{\sum_{a^\prime \in \cA}\exp(\phi(x)^\top  \theta_{a^\prime})}\,,\nonumber\\
    &\stackrel{(i)}{=}  \E{\gamma \sim {\rm G}(0, 1)^K}{\mathds{1}_{\{ h_{\theta, \gamma}(x) = a \}}}\,,
\end{align}
where $(i)$ follows from the Gumbel-Max trick (GMT) \citep{luce2012individual,maddison2014sampling}. Thus a \texttt{softmax} policy $\pi^{\textsc{sof}}_{\theta}$ can be written as in \cref{eq:es-pac_policies}. Now we also consider random parameters $\theta \sim \cN(\mu, \sigma^2 I_{dK})$ with $\mu \in \real^{dK}$ and $\sigma>0$. Then, let $\mathbb{Q}= \cN(\mu, \sigma^2I_{dK}) \times {\rm G}(0, 1)^K$, it follows that $\pi_{\mathbb{Q}} = \pi^{\textsc{mixL}}_{\mu, \sigma}$ is a mixed-logit policy and it reads
\begin{align}\label{eq:es-logit_pac_bayes}
 \pi^{\textsc{mixL}}_{\mu, \sigma}&(a | x) =  \E{\theta \sim \cN(\mu, \sigma^2 I_d)}{\frac{\exp(\phi(x)^\top \theta_a)}{\sum_{a^\prime \in \cA}\exp(\phi(x)^\top  \theta_{a^\prime})}}\,,\nonumber\\
    &= \E{\theta \sim \cN(\mu, \sigma^2 I_d)\,, \gamma \sim {\rm G}(0, 1)^K}{\mathds{1}_{\{ h_{\theta, \gamma}(x) = a \}}}.
\end{align}

\textbf{Example 2 (Gaussian policies):} \citet{sakhi2022pac} removed the Gumbel noise $\gamma$ in \cref{eq:es-logit_pac_bayes} and consequently defined the hypothesis space as $\mathcal{H} = \set{h_{\theta} \, ; \theta \in \real^{dK}}$ of mappings $h_{\theta}(x) = \argmax_{a \in \cA} \phi(x)^\top \theta_a$ for any $x \in \cX$. Then, let $\mathbb{Q} = \cN(\mu, \sigma^2 I_{dK})$, it follows that $ \pi_{\mathbb{Q}} = \pi^{\textsc{gaus}}_{\mu, \sigma}$ reads 
\begin{align}\label{eq:es-gaussian_pac_bayes}
   \pi^{\textsc{gaus}}_{\mu, \sigma}(a | x) = \E{\theta \sim \cN(\mu, \sigma^2 I_d)}{\mathds{1}_{\{ h_{\theta}(x) = a \}}}\,.
\end{align}
To see why removing the Gumbel noise can be beneficial, the reader may refer to \cref{app:policies}.

After motivating the definition of policies in \cref{eq:es-pac_policies}, we are in a position to relate our estimators to the general PAC-Bayes framework in \cref{subsec:es-pac_bayes_framekwork}. One technical requirement of our proof is that the estimator should be linear in $\pi$. Thus we focus on $\hat{V}^\alpha(\cdot)$ since $\tilde{V}^\beta(\pi)$ is non-linear in $\pi$. Let $h \in \mathcal{H}$, $x \in \cX$, $a \in \cA$ and $r \in [0, 1]$, we define the objective $U_\alpha$ as 
\begin{align}\label{eq:es-our_loss}
   & U_\alpha(h, x, a, r) = \frac{\mathds{1}_{\{h(x)=a\}}}{\pi_0(a | x)^\alpha}r\,.
\end{align}
Using the definition in \cref{eq:es-pac_policies} and the linearity of the expectation, we have that $\hat{V}^\alpha(\cdot)$ in \cref{eq:es-exp_ips_policy_value} can be written as 
\begin{align*}
    \hat{V}^\alpha(\pi_{\mathbb{Q}}) = \E{h \sim \mathbb{Q}}{\frac{1}{n} \sum_{i=1}^n U_\alpha(h, X_i, A_i, R_i)}\,.
\end{align*}
Moreover, the expectation of $\hat{V}(\pi_{\mathbb{Q}})$ reads
\begin{align*}
    V^\alpha(\pi_{\mathbb{Q}}) = \mathbb{E}_{h \sim \mathbb{Q}} \E{(X, A, R) \sim \mu_{\pi_0}}{U_\alpha(h, X, A, R)}\,.
\end{align*}
Finally, the main quantity of interest, the value $V(\pi_{\mathbb{Q}})\,,$ can be expressed in terms of the objective with $\alpha=1\,,$ $U_1\,,$ as
\begin{align*}
    V(\pi_{\mathbb{Q}}) = \mathbb{E}_{h \sim \mathbb{Q}} \E{(X, A, R) \sim \mu_{\pi_0}}{U_1(h, X, A, R)}\,.
\end{align*}
Since $\hat{V}^\alpha(\pi_{\mathbb{Q}})$ is an unbiased estimator of $ V^\alpha(\pi_{\mathbb{Q}})$, PAC-Bayes can be used to bound $V^\alpha(\pi_{\mathbb{Q}}) - \hat{V}^\alpha(\pi_{\mathbb{Q}})$. This will allow bounding our quantity of interest $V(\pi_{\mathbb{Q}}) - \hat{V}^\alpha(\pi_{\mathbb{Q}})$.

\subsection{Main Result}\label{subsec:es-main_result}
To ease the exposition, we assume that the rewards are deterministic. Then, in logged data $\cD_n$, $R_i = r(X_i, A_i)$ for any $i \in [n]$. Note that the same result holds for stochastic rewards. We discuss our result and sketch its proof in \cref{sec:es-discussion}. The complete proof can be found in \cref{proof:main_thm_proof}.

\begin{theorem}\label{thm:es-main_result} Let $\lambda>0$,  $n \ge 1$, $\delta \in (0, 1)$, $\alpha \in [0, 1]$, and let $\mathbb{P}$ be a fixed prior on $\mathcal{H}$, then with probability at least $1-\delta$ over draws $\cD_n \sim \mu_{\pi_0}^n$, the following holds simultaneously for any posterior $\mathbb{Q}$ on $\mathcal{H}$ 
\begin{align*}
  |V(\pi_{\mathbb{Q}}) -\hat{V}^\alpha(\pi_{\mathbb{Q}})| \leq \sqrt{ \frac{{\textsc{kl}}_{1}(\pi_{\mathbb{Q}})}{2n} } + B_n^\alpha(\pi_{\mathbb{Q}})  +
\frac{{\textsc{kl}}_{2}(\pi_{\mathbb{Q}})}{n \lambda } + \frac{\lambda}{2}\operatorname{Var}_n^\alpha(\pi_{\mathbb{Q}})\,.
\end{align*}
where ${\textsc{kl}}_{1}(\pi_{\mathbb{Q}})  =D_{\mathrm{KL}}(\mathbb{Q} \| \mathbb{P})+\ln \frac{4\sqrt{n}}{\delta}\,,$ and
\begin{align*}
    &{\textsc{kl}}_{2}(\pi_{\mathbb{Q}})  =  D_{\mathrm{KL}}(\mathbb{Q} \| \mathbb{P})+\ln \frac{4}{\delta}\,,\qquad B_n^\alpha(\pi_{\mathbb{Q}}) = 1 - \frac{1}{n}\sum_{i=1}^{n} \E{A \sim \pi_{\mathbb{Q}}(\cdot | X_i)}{\pi_0^{1-\alpha}(A | X_i)}\,,\\
    &\operatorname{Var}_n^\alpha(\pi_{\mathbb{Q}}) = \frac{1}{n}\sum_{i=1}^n  \E{A \sim \pi_{0}(\cdot | X_i)}{\frac{\pi_{\mathbb{Q}}(A | X_i)}{\pi_0(A | X_i)^{2\alpha }}} + \frac{\pi_{\mathbb{Q}}(A_i | X_i)R_i^2}{\pi_0(A_i | X_i)^{2\alpha}}.
\end{align*}
\end{theorem}
We start by clarifying that the prior $\mathbb{P}$ can be any fixed distribution on $\mathcal{H}$. If we have access to $\mathbb{P}_0$ on $\mathcal{H}$ such that $\pi_0 = \pi_{\mathbb{P}_0}$, then it is natural to set $\mathbb{P} =  \mathbb{P}_0$. But this is just a choice and one may use priors that do not depend on $\pi_0$. Now we explain the main terms in our bound. First, the terms $ {\textsc{kl}}_{1}(\pi_{\mathbb{Q}})$ and  ${\textsc{kl}}_{2}(\pi_{\mathbb{Q}})$ contain the divergence $D_{\mathrm{KL}}(\mathbb{Q} \| \mathbb{P})$ which penalizes posteriors $\mathbb{Q}$ that differ a lot from the prior $\mathbb{P}$. Moreover, $B_n^\alpha(\pi_{\mathbb{Q}})$ is the bias conditioned on the contexts $(X_i)_{i \in [n]}\,$; $B_n^\alpha(\pi_{\mathbb{Q}})=0$ when $\alpha=1$ and $B_n^\alpha(\pi_{\mathbb{Q}})>0$ otherwise. Also, the first term in $\operatorname{Var}_n^\alpha(\pi_{\mathbb{Q}})$ \emph{resembles} the theoretical second moment of the regularized importance weights $\frac{\pi}{\pi^\alpha_0}$ (without the reward) when they are seen as random variables. Similarly, the second term in $\operatorname{Var}_n^\alpha(\pi_{\mathbb{Q}})$ \emph{resembles} the empirical second moment of $\frac{\pi}{\pi^\alpha_0}R$ (with the reward). Finally, if $\operatorname{Var}_n^\alpha(\pi_{\mathbb{Q}})$ is bounded, then we can set $\lambda = 1/\sqrt{n}$, in which case our bound scales as $\mathcal{O}(1/\sqrt{n} + B_n^\alpha(\pi_{\mathbb{Q}}))$. In practice, we set $\alpha \approx 1$ leading to $B_n^\alpha(\pi_{\mathbb{Q}}) \approx 0$ and the bound would scale as $\mathcal{O}(1/\sqrt{n})$.

One of the main strengths of our result is that it holds for standard IPS with $\alpha=1$ under the assumption that $\operatorname{Var}_n^1(\pi_{\mathbb{Q}})$ is bounded. This assumption is less restrictive than assuming that the importance weight as a random variable, $\pi_{\mathbb{Q}}(A | X)/\pi_0(A|X)$, is bounded, a required assumption for traditional concentration bounds. In contrast, $\operatorname{Var}_n^\alpha(\pi_{\mathbb{Q}})$ only involves the \emph{expectations} of the random variables $\pi_{\mathbb{Q}}(A | X_i) / \pi_0(A | X_i)^{2\alpha }$, and ratios of $\pi_0$ evaluated at observed contexts and actions and  $(X_i, A_i)_{i \in [n]}$, that have non-zero probabilities under $\pi_0$ by definition.

Our result holds for fixed $\lambda>0$ and $\alpha \in [0, 1]$. In \cref{app:thm_extension}, we extend this to any potentially data-dependent $\lambda \in (0, 1)$ and $\alpha \in (0, 1]$. The assumption that $R \in [0, 1]$ can be relaxed to $R \in [0, B]$ up to additional factors $B^2$ and $B$ in $\operatorname{Var}_n^\alpha(\pi_{\mathbb{Q}})$ and  ${\textsc{kl}}_{1}(\pi_{\mathbb{Q}})$, respectively. Finally, our bound is suitable for stochastic gradient ascent \citep{robbins1951stochastic} since data-dependent quantities are not inside a square root. This is important for scalability.

\textbf{Limitations.} Our bound in \cref{thm:es-main_result} has two main limitations. 
\textbf{(i)} Using it to directly derive a data-independent suboptimality gap bound is not straightforward. This difficulty arises because our bound involves empirical quantities such as $B_n^\alpha(\pi_{\mathbb{Q}})$ and $\operatorname{Var}_n^\alpha(\pi_{\mathbb{Q}})$, whose dependence on the logged data prevents expressing the gap purely as a function of $n$. However, obtaining data-independent suboptimality guarantees was not the goal of this chapter. Instead, our focus was on deriving tractable and theoretically grounded bounds for exponential smoothing, that also perform well in practice when used for pessimistic objectives. 
\textbf{(ii)} Our result provides symmetric deviation bounds that simultaneously control the upper and lower deviations of $\hat{V}^\alpha(\pi_{\mathbb{Q}})$ from $V(\pi_{\mathbb{Q}})$. Yet, recent work \citep{gabbianelli2023importance}, published after the paper corresponding to this chapter, indicates that the tails of regularized IPS estimators are inherently asymmetric. Consequently, tighter bounds may arise from developing asymmetric two-sided bounds that treat each deviation separately. We explored this direction in our follow-up work \citep{Sakhi2024LS}, where we derived some of the tightest bounds in the literature, with strong empirical performance.

\subsection{Adaptive and Data-Driven Tuning of $\alpha$}\label{subsec:es-data_dep_alpha}

\cref{thm:es-main_result} assumes that $\alpha$ is fixed (although we extend it for data-dependent $\alpha$ in \cref{app:thm_extension}). However, providing a procedure to tune $\alpha$ in an adaptive and data-dependent fashion is important in practice. Thus we propose to set 
\begin{align}\label{eq:es-data_dependent_alpha}
    \alpha_* = \argmin_{\alpha \in [0, 1]} B^\alpha_n(\pi_{\mathbb{Q}})  + \sqrt{\frac{2 {\textsc{kl}}_{2}(\pi_{\mathbb{Q}})\operatorname{Var}^\alpha_n(\pi_{\mathbb{Q}}) }{n}}\,,
\end{align}
where all the terms are defined in \cref{thm:es-main_result}. Roughly speaking, $\alpha_*$ establishes a bias-variance trade-off; it minimizes the sum of the bias term $B^\alpha_n(\pi_{\mathbb{Q}})$ and the square root of the second moment term $\operatorname{Var}^\alpha_n(\pi_{\mathbb{Q}})$, weighted by $\sqrt{\frac{2 {\textsc{kl}}_{2}(\pi_{\mathbb{Q}})}{n}}$. Here \cref{eq:es-data_dependent_alpha} is obtained by minimizing the bound in \cref{thm:es-main_result} with respect to both $\alpha$ and $\lambda$ as follows. First, we minimize the bound in \cref{thm:es-main_result} with respect to $\lambda$; the minimizer is $\lambda_* = \sqrt{\frac{2 {\textsc{kl}}_{2}(\pi_{\mathbb{Q}}) }{n\operatorname{Var}^\alpha_n(\pi_{\mathbb{Q}})}}$. Then, the bound in \cref{thm:es-main_result} evaluated at $\lambda = \lambda_*$ becomes
\begin{align}\label{eq:es-data_dependent_alpha_2}
    \sqrt{\frac{ {\textsc{kl}}_{1}(\pi_{\mathbb{Q}})}{2n}} +  B^\alpha_n(\pi_{\mathbb{Q}})  + \sqrt{\frac{2 {\textsc{kl}}_{2}(\pi_{\mathbb{Q}})\operatorname{Var}^\alpha_n(\pi_{\mathbb{Q}}) }{n}}\,.
\end{align}
Finally, $\alpha_*$ is defined as the minimizer of \cref{eq:es-data_dependent_alpha_2} with respect to $\alpha \in [0, 1]$, and $\sqrt{\frac{ {\textsc{kl}}_{1}(\pi_{\mathbb{Q}})}{2n}}$ does not appear in \cref{eq:es-data_dependent_alpha} as it does not depend on $\alpha$. Note that $\alpha_*$ depends on both logged data $\cD_n$ and the learning policy $\pi_{\mathbb{Q}}$. Thus it is adaptive; its value changes in each iteration during optimization.

\section{Discussion}\label{sec:es-discussion}
We start by interpreting and comparing our results to related work. Then, we present the technical challenges in \cref{subsec:es-tech_challenges}. After that, we sketch our proof in \cref{subsec:es-sketch}.

\subsection{Interpretation and Comparison to Related Work}\label{subsec:es-interpretation}

\cref{thm:es-main_result} gives insight into the number of samples needed so that the performance of $\hat{\pi}$ is close to that of the optimal policy $\pi_*$. To simplify the problem, we consider the Gaussian policies in \cref{eq:es-gaussian_pac_bayes} and assume that there exists $\mathbb{Q}_* = \cN(\mu_*, I_{dK})$ with $\mu_* \in \real^{dK}$ such that the optimal policy is $\pi_* = \pi_{\mathbb{Q}_*}$. Also, we let the prior $\mathbb{P} =  \cN(\mu_0, I_{dK})$ and assume that $\pi_0$ is uniform. This is possible since as we said before, the prior $\mathbb{P}$ does not have to depend on the logging policy $\pi_0$. Then we have that $D_{\mathrm{KL}}(\mathbb{Q}_* \| \mathbb{P}) = \norm{\mu_* - \mu_0}^2 / 2$, $ B_n^\alpha(\pi_{\mathbb{Q}_*}) = 1 - 1/K^{1-\alpha}$ and $\operatorname{Var}_n^\alpha(\pi_{\mathbb{Q}_*}) \leq 2K^{2 \alpha}$. The last inequality is not tight but it allows getting an easy-to-interpret term that does not depend on $n$. Now let $\epsilon> 2(1 - K^{\alpha-1})$ for $\alpha \in [1- \log 2 / \log K, 1]$. This condition on $\alpha$ ensures that $\epsilon \in [0, 1]$ and it is mild as $\alpha$ is often close to 1. Then, it holds with high probability that
\begin{align*}
 n \, \widetilde{>} \Big(\frac{ \norm{\mu_* - \mu_0}^2  + K^{2 \alpha}}{\epsilon -  2 (1 - K^{\alpha-1})}\Big)^2  \implies V(\hat{\pi}) \geq V(\pi_{\mathbb{Q}_*}) - \epsilon \,,
\end{align*}
where we omit constant and logarithmic terms in $\widetilde{>}$. This gives an intuition on the sample complexity for our procedure. In particular, fewer samples are needed in four cases. The first is when $\epsilon$ is large, which means that we afford to learn a policy whose performance is far from the optimal one. The second is when the prior $\mathbb{P}$ is close to $\mathbb{Q}_*$, that is when $\norm{\mu_* - \mu_0}$ is small. This highlights that the choice of the prior $\mathbb{P}$ is important. The third is when the second-moment term $K^{2 \alpha}$ is small. The fourth is when the bias $B_n^\alpha(\pi_{\mathbb{Q}_*})$ is small. In particular, when $\alpha=1$, the bias is 0. In contrast, the second-moment term is minimized in $\alpha=0$. This is where the choice of $\alpha$ matters. The proofs of these claims and more detail can be found in \cref{proof:practice_theory}.

Our chapter derives a \emph{tractable generalization bound} for an estimator other than clipped IPS in \cref{eq:es-clip_ips_policy_value}, which also holds for the standard IPS in \cref{eq:es-ips_policy_value}. The bounds in \citet{swaminathan2015batch,london2019bayesian,sakhi2022pac} have a multiplicative dependency on the clipping threshold ($M$ or $1/\tau$ in \cref{eq:es-clip_ips_policy_value}). Standard IPS is recovered when $M \rightarrow \infty$ (or $\tau = 0$) in which case their bounds are infinite. We successfully avoid any similar dependency on $\alpha$. Moreover, \citet{swaminathan2015batch,london2019bayesian} only used their generalization bounds to inspire pessimistic objectives. Although we directly optimize our theoretical bound (\cref{thm:es-main_result})
in our experiments, our analysis also inspires a pessimistic objective where we simultaneously penalize the $L_2$ distance, the variance and the bias. That is, we find $\mu \in \real^{dK}$ that maximizes
\begin{align}\label{eq:es-learning_principle}
\hspace{-0.2cm} \hat{V}^\alpha(\pi_{\mu}) - \lambda_1 \norm{\mu - \mu_0}^2  - \lambda_2  \operatorname{Var}_n^\alpha(\pi_{\mu}) - \lambda_3 B_n^\alpha(\pi_{\mu})\,.
\end{align}
Here $\lambda_1, \lambda_2$ and $\lambda_3$ are tunable hyper-parameters, $\pi_{\mu}$ can be the Gaussian policy in \cref{eq:es-gaussian_pac_bayes}, $\pi_{\mu} = \pi^{\textsc{gaus}}_{\mu, 1}$, with a fixed $\sigma=1$, and $\mu_0$ is the mean of the prior $\mathbb{P} = \cN(\mu_0, I_{dK})$. Existing works either penalize the $L_2$ distance or the variance. For completeness, we also show that this pessimistic objective should be preferred over existing ones in \cref{app:add_discussion}.

\subsection{Technical Challenges}\label{subsec:es-tech_challenges}

\citet{london2019bayesian,sakhi2022pac} derived PAC-Bayes generalization bounds for the estimator \texttt{IPS-max} in \cref{eq:es-clip_ips_policy_value}. Extending their analyses to our case is not straightforward. First, their estimator \texttt{IPS-max} is upper bounded by $1/\tau$, and thus they relied on traditional techniques for $[0,1]$-objectives \citep{alquier2021user}. In contrast, our objective in \cref{eq:es-our_loss} is not upper-bounded, and controlling it without assuming that the importance weights are bounded is challenging.

Moreover, their bounds have a multiplicative dependency on $1/\tau$, hence they explode as $\tau \rightarrow 0$. This makes them vacuous for small values of $\tau$ and inapplicable to the standard IPS estimator in \cref{eq:es-ips_policy_value} recovered for $\tau =0$. 
In contrast, our bound does not have a similar dependency on $\alpha$ and it is also valid for standard IPS recovered for $\alpha=1$. Moreover, we derive two-sided inequalities rather than one-sided ones for the important reasons that we priorly discussed.  
This requires carefully controlling in \emph{closed-form} the absolute value of the bias. Prior works only used that the bias is negative which was enough to obtain one-sided inequalities. 

Explaining other challenges requires stating a result that inspired our analysis: \citet{kuzborskij2019efron} derived PAC-Bayes generalization bounds for unbounded losses by only controlling their second moments. Recently, \citet{haddouche2022pac} proposed a similar result using Ville's inequality \citep{bercu2008exponential}. Adapting their theorem to our problem is given \cref{prop:direct_application}. We slightly adapt their proof to get a \emph{two-sided} inequality for a \emph{negative} loss. The proof is deferred to \cref{proof:direct_application}.

\begin{proposition}\label{prop:direct_application} Let $\lambda>0$, $n \geq 1$, $\delta \in(0,1)$, $\alpha \in [0, 1]$ and let $\mathbb{P}$ be a fixed prior on $\mathcal{H}$, then with probability at least $1-\delta$ over draws $\cD_n \sim \mu_{\pi_0}^n$, the following holds simultaneously for all posteriors, $\mathbb{Q}$, on $\mathcal{H}$
\begin{align}\label{eq:es-direct_application}
    |V^\alpha(\pi_{\mathbb{Q}}) -\hat{V}^\alpha(\pi_{\mathbb{Q}})|   \leq \frac{D_{\mathrm{KL}}(\mathbb{Q} \| \mathbb{P})+\log \frac{2}{\delta}}{\lambda n} + & \frac{\lambda}{2 n} \sum_{i=1}^n \frac{\pi_{\mathbb{Q}}(A_i | X_i)}{\pi_0^{2\alpha}(A_i | X_i)} R_i^2 \nonumber \\ & +\frac{\lambda}{2}\mathbb{E}_{(X, A, R) \sim \mu_{\pi_0}}\left[\frac{\pi_{\mathbb{Q}}(A | X)}{\pi_0^{2\alpha}(A | X)} R^2\right]\,,
\end{align}
\end{proposition}

There are two main issues with \cref{prop:direct_application}. First, the term $\mathbb{E}_{(X, A, R) \sim \mu_{\pi_0}}\big[\frac{\pi_{\mathbb{Q}}(A | X)}{\pi_0^{2\alpha}(A | X)} R^2\big]$ in \cref{eq:es-direct_application} is intractable. One could bound $R^2$ by $1$, but the resulting term will still be intractable due to the expectation over the unknown distribution of contexts $\nu$. Second, we need an upper bound of $|V(\pi_{\mathbb{Q}}) - \hat{V}^\alpha(\pi_{\mathbb{Q}})|$ while \cref{prop:direct_application} only provides one for $|V^\alpha(\pi_{\mathbb{Q}})-\hat{V}^\alpha(\pi_{\mathbb{Q}})|$. Thus it remains to quantify the approximation error $|V(\pi_{\mathbb{Q}})-V^\alpha(\pi_{\mathbb{Q}})|$. This will also require computing an expectation over $X \sim \nu$, which is intractable.

\subsection{Sketch of Proof for \Cref{thm:es-main_result}}\label{subsec:es-sketch}

We conclude by showing how the technical challenges above were solved. First, We decompose $V(\pi_{\mathbb{Q}})-\hat{V}^\alpha(\pi_{\mathbb{Q}})$ as
\begin{align*}
    V(\pi_{\mathbb{Q}})-\hat{V}^\alpha &(\pi_{\mathbb{Q}}) = I_1 + I_2 + I_3\,, \qquad \text{where}
\end{align*}
  \begin{align*}
    I_1 &= V(\pi_{\mathbb{Q}}) - \frac{1}{n}\sum_{i=1}^n V(\pi_{\mathbb{Q}} | X_i)\,,\\
     I_2 &=  \frac{1}{n} \sum_{i=1}^n V(\pi_{\mathbb{Q}} | X_i) - \frac{1}{n}\sum_{i=1}^n V^\alpha(\pi_{\mathbb{Q}} | X_i)\,,\\
     I_3 &= \frac{1}{n}\sum_{i=1}^n V^\alpha(\pi_{\mathbb{Q}} | X_i) - \hat{V}^\alpha(\pi_{\mathbb{Q}})\,,
\end{align*}
where 
\begin{align*}
&V(\pi_{\mathbb{Q}} | X_i) = \E{A \sim \pi_{\mathbb{Q}}(\cdot | X_i)}{r(X_i, A)}\,,
    & V^\alpha(\pi_{\mathbb{Q}} | X_i) = \mathbb{E}_{A \sim \pi_0(\cdot | X_i)}\Big[\frac{\pi_{\mathbb{Q}}(A | X_i)}{\pi_0(A | X_i)^\alpha}r(X_i, A)\Big]\,.
\end{align*}
$I_1$ is the estimation error of the empirical mean of the value using $n$ i.i.d. contexts $(X_i)_{i \in [n]}$. This term is introduced to avoid the intractable expectation over $X \sim \nu$. Moreover, $I_2$ is the bias term conditioned on the contexts $(X_i)_{i \in [n]}$ and we bound it in closed-form. Finally, $I_3$ is the estimation error of the value conditioned on the contexts $(X_i)_{i \in [n]}$. Again, this conditioning allows us to avoid the intractable expectation over $X \sim \nu$ and to consequently bound $|I_3|$ by tractable terms. First, \citet[Theorem~3.3]{alquier2021user} yields that with probability at least $1-\frac{\delta}{2}$, it holds for any $\mathbb{Q}$ on $\mathcal{H}$ that
\begin{align*}
    |I_1| \leq \sqrt{ \frac{D_{\mathrm{KL}}(\mathbb{Q} \| \mathbb{P})+\log \frac{4\sqrt{n}}{\delta}}{2n}} \,.
\end{align*}
Also, $|I_2|$ is bounded similarly to \cref{eq:es-alpha_bias_variance} as 
\begin{align*}
   |I_2| \leq \frac{1}{n} \sum_{i=1}^{n} \E{A \sim \pi_{\mathbb{Q}}(\cdot | X_i)}{1 - \pi_0^{1-\alpha}(A | X_i)}\,.
\end{align*}
Bounding $|I_3|$ is achieved by expressing it using martingale difference sequences $(f_i(A_i, h))_{i \in [n]}$ that we construct as follows. Let $(\mathcal{F}_i)_{i \in \{0\} \cup [n]}$ be a filtration adapted to $(S_i)_{i \in [n]}$ where $S_i = (A_\ell)_{\ell \in [i]}$ for any $i \in [n]$, we define
\begin{align*}
f_i\left(A_i, h\right) = \E{A \sim \pi_0(\cdot | X_i)}{\frac{\mathds{1}_{\{h(X_i)=A\}}r(X_i, A)}{\pi_0(A | X_i)^\alpha} } - \frac{\mathds{1}_{\{h(X_i)=A_i\}}R_i}{\pi_0(A_i | X_i)^\alpha}\,.
\end{align*}
Then we show that for any $h \in \mathcal{H}$,  $(f_i(A_i, h))_{i \in [n]}$ is a martingale difference sequence. After that, we apply \citet[Theorem 5]{haddouche2022pac} and obtain that with probability at least $1-\delta/2$, it holds for any $\mathbb{Q}$ on $\mathcal{H}$ that
\begin{align*}
    \left|\E{h \sim \mathbb{Q}}{M_n(h)}\right| &\leq \frac{D_{\mathrm{KL}}(\mathbb{Q} \| \mathbb{P})+\log \frac{4}{\delta}}{\lambda} +\frac{\lambda}{2} \E{h \sim \mathbb{Q}}{\operatorname{Var}_n(h)},\nonumber
\end{align*}
where $M_n(h)=\sum_{i=1}^n f_i\left(A_i, h\right)$ and $\operatorname{Var}_n(h)=\sum_{i=1}^n f_i\left(A_i, h\right)^2 + \mathbb{E}\left[f_i\left(A_i, h\right)^2 | \mathcal{F}_{i-1}\right]$. Then notice that $\E{h \sim \mathbb{Q}}{M_n(h)}$ can be expressed in terms of $I_3$ as 
\begin{align*}
    \E{h \sim \mathbb{Q}}{M_n(h)} &= \sum_{i=1}^n V^\alpha(\pi_{\mathbb{Q}} | X_i) - n\hat{V}^\alpha(\pi_{\mathbb{Q}}) = n I_3\,,
\end{align*}
Moreover, $\E{h \sim \mathbb{Q}}{\operatorname{Var}_n(h)}$ is bounded by
\begin{align*}
    \sum_{i=1}^n  \E{A \sim \pi_0(\cdot | X_i)}{\frac{\pi_{\mathbb{Q}}(A | X_i)}{\pi_0(A | X_i)^{2\alpha}}}+ \frac{\pi_{\mathbb{Q}}(A_i | X_i)}{\pi_0(A_i | X_i)^{2\alpha}}R_i^2\,.
\end{align*}
Thus with probability at least $1-\frac{\delta}{2}$, it holds for any $\mathbb{Q}$ that
\begin{align*}
   |I_3 | &\leq \frac{D_{\mathrm{KL}}(\mathbb{Q} \| \mathbb{P})+\log\frac{4}{\delta}}{n\lambda}+ \frac{\lambda}{2n} \sum_{i=1}^n\frac{\pi_{\mathbb{Q}}(A_i | X_i)}{\pi_0(A_i | X_i)^{2\alpha}}R_i^2 + \frac{\lambda}{2n}\sum_{i=1}^n  \E{A \sim \pi_0(\cdot | X_i)}{\frac{\pi_{\mathbb{Q}}(A | X_i)}{\pi_0(A | X_i)^{2\alpha}}}\,.
\end{align*}
Our result is obtained by bounding $|I_1| + |I_2| + |I_3|$. One shortcoming of our analysis is that $\operatorname{Var}_n^\alpha(\pi_{\mathbb{Q}})$ is not exactly and only resembles the sum of the theoretical and empirical second moments of our estimator. Precisely, the terms $\pi_{\mathbb{Q}}/\pi_0^{2\alpha}$ should be $\pi_{\mathbb{Q}}^2/\pi_0^{2\alpha}$. This problem arises due to our definition of the martingale difference sequences $(f_i(A_i, h))_{i \in [n]}$ in \cref{eq:es-our_loss}. Precisely, in our proof, we compute the square $f_i(A_i, h)^2$. However, the square of an indicator function is the indicator function itself. Thus applying the expectation afterwards, $\E{h \sim \mathbb{Q}}{f_i(A_i, h)^2}$, leads to $\pi_{\mathbb{Q}}$ appearing instead of $\pi_{\mathbb{Q}}^2$. This issue is inherent in the PAC-Bayes formulation and seminal works \citep{london2019bayesian,sakhi2022pac} would suffer the same issue. Solving this would be beneficial and we leave it to future work.

\section{Experiments for Exponential Smoothing}\label{sec:es-experiments}

We briefly present our experiments. More details and discussions can be found in \Cref{app:all_experiments}. We consider the standard supervised-to-bandit conversion \citep{agarwal2014taming} where we transform a supervised training set $\mathcal{S}^{\textsc{tr}}_{n}$ to a logged bandit data $\cD_n$ as described in \cref{alg:es-supervised_to_bandit} in \cref{app:setup}. Here the action space $\cA$ is the label set and the context space $\cX$ is the input space. Then, $\cD_n$ is used to train our policies. After that, we evaluate the value of the learned policies on the supervised test set $\mathcal{S}^{\textsc{ts}}_{n_{\textsc{ts}}}$ as described in \cref{alg:es-supervised_to_bandit_test} in \cref{app:setup}. Roughly speaking, the resulting value quantifies the ability of the learned policy to predict the true labels of the inputs in the test set. This is our performance metric; the higher the better. We use 4 image classification datasets \texttt{MNIST} \citep{lecun1998gradient}, \texttt{FashionMNIST} \citep{xiao2017fashion},  \texttt{EMNIST} \citep{cohen2017emnist} and \texttt{CIFAR100} \citep{krizhevsky2009learning}.

The logging policy is defined as $\pi_0 = \pi_{\eta_0 \cdot \mu_0}^{\textsc{sof}}$ in \cref{eq:es-softmax_pac_bayes}, where $\mu_0 = (\mu_{0,a})_{a \in \cA} \in \real^{dK}$ and $\eta_0 \in [0, 1]$ is the inverse-temperature parameter. The higher $\eta_0$, the better the performance of $\pi_0$. When $\eta_0=0$, $\pi_0$ is uniform. The parameters $\mu_0$ are learned using 5\% of the training set $\mathcal{S}^{\textsc{tr}}_{n}$. In our experiments, we consider both, Gaussian and mixed-logit policies, in \cref{eq:es-logit_pac_bayes} and \cref{eq:es-gaussian_pac_bayes}, for which we set the prior as $\mathbb{P} = \cN(\eta_0 \mu_0, I_{dK})$ and $\mathbb{P} = \cN(\eta_0 \mu_0, I_{dK}) \times {\rm G}(0, 1)^K$, respectively. Given that $\mu_0$ are learnt on $5\%$ of $\mathcal{S}^{\textsc{tr}}_{n}$, we train our policies on the remaining $95\%$ portion of $\mathcal{S}^{\textsc{tr}}_{n}$ to match our theory that requires the prior to not depend on training data. The policies are trained using Adam \citep{kingma2014adam} with a learning rate of $0.1$ for $20$ epochs.

\textbf{Main results.} We compare our bound to those in \citet{london2019bayesian,sakhi2022pac}; discarding the intractable bound in \citet{swaminathan2015batch} as it requires computing a covering number. Here we do not include the pessimistic objectives in \citet{swaminathan2015batch,london2019bayesian} since we directly optimize our bounds. But we make such a comparison in \cref{app:add_discussion} for completeness, showing the favorable performance of our bound and the newly proposed pessimistic objective in \cref{eq:es-learning_principle}. Also, we do not compare to \citet{su2020doubly,metelli2021subgaussian} since they do not provide generalization guarantees; they focus on estimation accuracy and only propose a heuristic for off-policy learning. However, we still show the favorable performance of our approach in off-policy learning compared to \citet{su2020doubly,metelli2021subgaussian} in \cref{app:other_corrections} for completeness.

Prior methods are not named. Thus we refer to them as \textbf{(Author, Policy)} where \textbf{Author} $\in$ \{\textbf{Ours, London et al., Sakhi et al. 1, Sakhi et al. 2\} and Policy $\in$ \{Gaussian, Mixed-Logit}\}. Here \textbf{Ours}, \textbf{London et al.}, \textbf{Sakhi et al. 1} and \textbf{Sakhi et al. 2} correspond to \cref{thm:es-main_result},  \citet[Theorem 1]{london2019bayesian}, \citet[Proposition 1]{sakhi2022pac}, and \citet[Proposition 3]{sakhi2022pac}, respectively. Since we have two classes of policies, each bound leads to two baselines. For example, \citet[Theorem 1]{london2019bayesian} leads to\textbf{ (London et al., Gaussian)} and \textbf{(London et al., Mixed-Logit)}. More details are provided in \cref{app:baselines}.

\begin{figure*}
  \centering  \includegraphics[width=\linewidth]{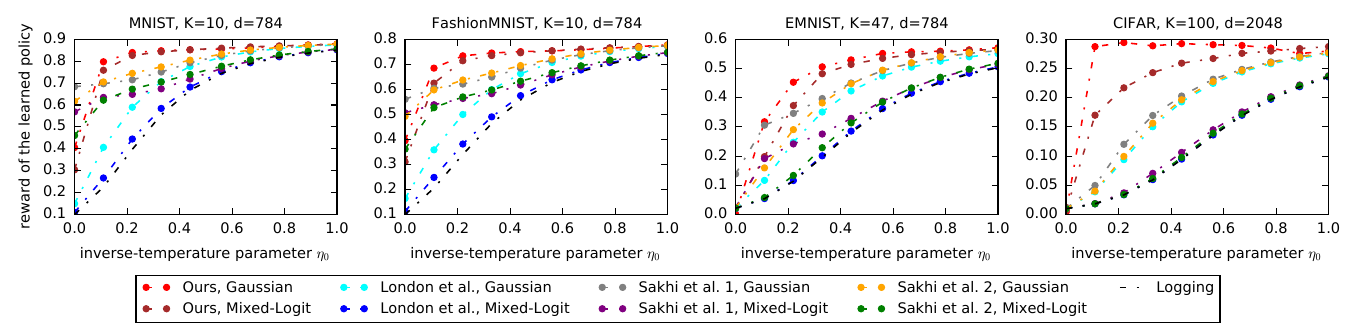}
  \caption{The reward of the learned policy using one of the baselines with varying quality of the logging policy $\eta_0 \in [0, 1]$.} 
  \label{fig:es-main_exp_results}
\end{figure*}

\begin{figure}[H]
\begin{center}
\centerline{\includegraphics[width=0.7\columnwidth]{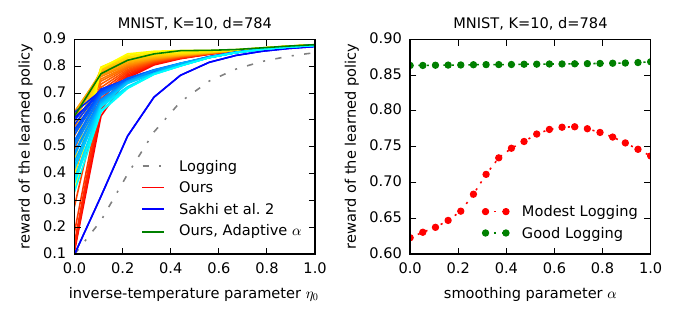}}
\caption{On the left-hand side is the reward of the learned policy with varying $\tau \in (0, 1)$, $\alpha \in (0, 1)$ and $\eta_0 \in [0, 1]$, and for an adaptive $\alpha$ using the procedure in \cref{subsec:es-data_dep_alpha} (green curve). The blue-to-cyan and red-to-yellow colors correspond to varying values of $\tau$ and $\alpha$, respectively. The lighter the color, the higher the value of $\tau$ or $\alpha$. The green curve corresponds to the reward of the learned policy with an adaptive and data-dependent $\alpha$ (\cref{subsec:es-data_dep_alpha}). On the right-hand side is the \emph{average} reward of the learned policies using our method across the modest and good logging groups, $\eta_0 \in [0, 0.5]$ (red) and $\eta_0 \in [0.5, 1]$ (green), respectively.}
\label{fig:es-varying_params}
\end{center}
\end{figure}

In \cref{fig:es-main_exp_results}, we report the value of the learned policies. Here we fix $\tau= 1/\sqrt[\leftroot{-2}\uproot{2}4]{n} \approx 0.06$ and $\alpha = 1-1/\sqrt[\leftroot{-2}\uproot{2}4]{n} \approx 0.94$ so that when $n$ is large enough, both $\hat{V}^\tau(\pi)$ and $ \hat{V}^\alpha(\pi)$ approach $\hat{V}^{\textsc{ips}}(\pi)$ \citep{ionides2008truncated}. This is because standard IPS should be preferred when $n \rightarrow \infty$. To have a fair comparison, we fixed $\alpha$ instead of tuning it in an adaptive fashion as described in \cref{subsec:es-data_dep_alpha}. However, we also provide the results with an adaptive $\alpha$ in \cref{fig:es-varying_params}. Let us start with interpreting \cref{fig:es-main_exp_results} (with fixed $\alpha$ and $\tau$). Overall, our method outperforms all the baselines. We also observe that Gaussian policies behave better than mixed-logit policies. However, this is less significant for our method where the performances of both Gaussian and mixed-logit policies are comparable. Moreover, our method reaches the maximum value even when the logging policy has an average performance. In contrast, the baselines only reach their best value when the logging policy is well-performing ($\eta_0 \approx 1$), in which case minor to no improvements are made. Finally, the baselines induce a better value when the logging policy is uniform ($\eta_0 = 0$). But our method has a better value when $\eta_0>0$, which is more common in practice. 

\paragraph{Larger action spaces.} The experiments above did not consider very large values of $K$. However, \cref{chap:las} evaluated IPS-based methods, including exponential smoothing and clippped IPS, on datasets with up to one million actions. In those experiments, exponential smoothing outperformed clipped IPS, though the improvements were modest compared to the gains observed here.

\textbf{Choice of hyperparameters.} Our choice of $\tau$ and $\alpha$ does not affect the above conclusions. In \cref{fig:es-varying_params} (left-hand side), we compare our method with the best baseline, \textbf{(Sakhi et al. 2)} with Gaussian policies, for $20$ evenly spaced values of $\tau \in (0, 1)$ and $\alpha \in (0, 1)$. We also include the results using the adaptive tuning procedure of $\alpha$ described in \cref{subsec:es-data_dep_alpha} (green curve). This procedure is reliable since the performance with an adaptive $\alpha$ (green curve) is comparable with the best possible choice of $\alpha$. Also, our method consistently outperforms the best baseline \textbf{(Sakhi et al. 2)} with the best value of $\tau$ when the logging policy is not uniform $(\eta_0 >0)$. Also, there is no very bad choice of $\alpha$, in contrast with $\tau = 10^{-5}$ (dark blue plot) which led to minimal improvement upon all logging policies. This might be due to the $1/\tau$ dependency in existing bounds. 

To see the effect of $\alpha$, we consider the following experiment. We split the logging policies into two groups. The first is called \emph{modest logging} which corresponds to logging policies $\pi_0$ whose $\eta_0$ is between $0$ and $0.5$. This group includes the uniform policy and other average-performing policies. The second is called \emph{good logging} and it includes the logging policies whose $\eta_0$ is between $0.5$ and $1$. Then, for each $\alpha$, we compute the average value of the learned policy, with that value of $\alpha$, across these two groups. This leads to the two red and green curves in \cref{fig:es-varying_params} (right-hand side). Overall, we observe that $\alpha \approx 0.7$ leads to the best performance across the modest logging group. Thus when the performance of the logging policy is bad or average, which is common in practice, importance-weight regularization can be critical. In contrast, when the performance of the logging policy is already good and $n$ is large enough, importance-weight regularization might not be needed and $\alpha \approx 1$ would also lead to good performance. This is one of the main strengths of our bound; it holds for the standard IPS recovered with $\alpha=1$. This result goes against the belief that clipped IPS should always be preferred to standard IPS. Here, our bound applied to standard IPS outperformed clipping by a large margin when the logging policy is relatively well-performing. Similar results for the other datasets are deferred to \cref{app:add_results}.

\section{Extension to Other Regularizations}
\label{sec:es-extension}

The experiments above demonstrated that exponential smoothing substantially outperforms clipping. However, we compared exponential smoothing with our pessimistic objective against clipping with pessimistic objectives specifically designed for it. This makes it difficult to isolate whether the gains stem from exponential smoothing as a regularization technique or from our pessimistic objective. Moreover, exponential smoothing and clipping are only two instances within a broader class of importance-weight regularizations.

While numerous methods have been proposed to stabilize IPS through importance-weight transformations \citep{bottou2013counterfactual, swaminathan2015batch, su2020doubly, metelli2021subgaussian}, most focus on estimation accuracy rather than learning performance. As highlighted in \cref{chap:las}, improved estimation does not necessarily yield improved policies, motivating a reassessment of importance-weight regularization specifically within the learning paradigm. Moreover, existing approaches followed a case-by-case basis: each regularization technique comes with its own theoretical analysis and corresponding pessimistic objective. This inconsistency makes it impossible to determine whether empirical improvements arise from the regularizer itself or from its specific objective formulation.

This reveals a critical gap: the absence of a unified framework providing principled pessimistic objectives across diverse importance-weight regularizations. We address this by developing a generic PAC-Bayesian generalization bound that applies uniformly to a broad family of regularizations, enabling fair comparison within a single theoretical framework.

Recall that the regularized IPS estimator has the form:
\begin{align}\label{eq:general_reg_ips}
    \hat{V}(\pi) = \frac{1}{n} \sum_{i=1}^n R_i \, \hat{w}(X_i, A_i),
\end{align}
where $\hat{w}(X, A)$ are the regularized importance weights. We further assume that $ \hat{w}(X, A)= g(\pi(A | X), \pi_0(A | X))$ for some function $g: [0, 1] \times [0, 1] \to \mathbb{R}^+$. Examples of $\hat{w}$ include clipping (\texttt{Clip}) \citep{london2019bayesian}, exponential smoothing (\texttt{ES}) \citep{aouali23a}, implicit exploration (\texttt{IX}) \citep{gabbianelli2023importance}, and harmonic (\texttt{Har}) \citep{metelli2021subgaussian}, defined as
\begin{align}\label{eq:regs}
    \texttt{Clip}: \qquad &\hat{w}(x, a) = \frac{\pi(a \mid x)}{\max(\pi_0(a \mid x), \tau)}\,, \, \tau \in [0, 1]\,,\\
    \texttt{ES}: \qquad &\hat{w}(x, a) = \frac{\pi(a \mid x)}{\pi_0(a \mid x)^\alpha}\,, \, \alpha \in [0, 1]\,,\nonumber\\
    \texttt{IX}: \qquad &\hat{w}(x, a) = \frac{\pi(a \mid x)}{\pi_0(a \mid x) + \gamma}\,, \, \gamma \in [0, 1]\,,\nonumber\\
    \texttt{Har}: \qquad &\hat{w}(x, a) = \frac{{w}(x, a)}{(1-\lambda){w}(x, a) +\lambda}\,, \, \lambda \in [0, 1]\,.\nonumber
\end{align}
\subsection{Generalization Bounds}\label{subsec:gbound}
PAC-Bayes theory (\cref{sec:es-opl}) allows bounding \(\left|\mathbb{E}_{\theta \sim \mathbb{Q}}[V(\pi_\theta) - \hat{V}(\pi_\theta)]\right|\), with
\begin{align*}
   & V(\pi_\theta) = \mathbb{E}_{X \sim \nu, A \sim \pi_{\theta}(\cdot | X)}[r(X, A)]\,, &
    \hat{V}(\pi_\theta) = \frac{1}{n} \sum_{i=1}^n \hat{w}_\theta(X_i, A_i)R_i\,,
\end{align*}
where we make the dependence of $\hat{w}_\theta$ on $\theta$ explicit to avoid confusion when taking the expectation $\mathbb{E}_{\theta \sim \mathbb{Q}}$. Below is our first general result that extends \cref{thm:es-main_result} to any regularization function $g$, instead of just exponential smoothing. Its proof follows exactly the same techniques we employed for \cref{thm:es-main_result}.

\begin{theorem}\label{thm:main_result}
Let \(\lambda > 0\), \(n \ge 1\), \(\delta \in (0, 1)\), and let \(\mathbb{P}\) be a fixed prior on \(\Theta\). The following inequality holds with probability at least \(1 - \delta\) for any distribution \(\mathbb{Q}\) on \(\Theta\):
\begin{align}\label{eq:app_main_inequality_maint}
    &\left|\mathbb{E}_{\theta \sim \mathbb{Q}}[V(\pi_\theta) - \hat{V}(\pi_\theta)]\right| \leq \sqrt{ \frac{{\textsc{kl}}_1(\pi_{\mathbb{Q}})}{2n} }  + \frac{{\textsc{kl}}_2(\pi_{\mathbb{Q}})}{n \lambda } + B_n(\mathbb{Q})  + \frac{\lambda}{2}\operatorname{Var}_n(\mathbb{Q})\,,
\end{align}
where \({\textsc{kl}}_1(\pi_{\mathbb{Q}}) = D_{\mathrm{KL}}(\mathbb{Q} \| \mathbb{P}) + \log \frac{4\sqrt{n}}{\delta}\), \({\textsc{kl}}_2(\pi_{\mathbb{Q}}) = D_{\mathrm{KL}}(\mathbb{Q} \| \mathbb{P}) + \log \frac{4}{\delta}\), and
\begin{align*}
    \operatorname{Var}_n(\mathbb{Q}) &= \frac{1}{n}\sum_{i=1}^n \mathbb{E}_{\theta \sim \mathbb{Q}}\big[\mathbb{E}_{A \sim \pi_0(\cdot | X_i)}[\hat{w}_\theta(X_i, A)^2] + \hat{w}_\theta(X_i, A_i)^2 R_i^2\big]\,,\\
  B_n(\mathbb{Q}) &= \frac{1}{n} \sum_{i=1}^{n} \sum_{A \in \cA} \mathbb{E}_{\theta \sim \mathbb{Q}}\big[|\pi_{\theta}(A | X_i) - \pi_0(A | X_i) \hat{w}_\theta(X_i, A)|\big]\,.
\end{align*}
\end{theorem}
The terms in the above bound have similar interpretations to those in \cref{thm:es-main_result}.

\textbf{Linear vs. non-linear regularization.} If \(\hat{w}(X, A)\) is linear in \(\pi_{\theta}(X, A)\) (i.e., $g$ linear in its first variable), then \(\hat{V}\) is also linear in \(\pi_\theta\), yielding
\begin{align*}
    \left|\mathbb{E}_{\theta \sim \mathbb{Q}}[V(\pi_{\theta}) - \hat{V}(\pi_{\theta})]\right| = \left|V(\pi_{\mathbb{Q}}) - \hat{V}(\pi_{\mathbb{Q}})\right|\,,
\end{align*}
where we define (similar to \cref{pac_bayes_for_opl})
\begin{align}\label{eq:pac_bayes_policy}
    \pi_{\mathbb{Q}} = \mathbb{E}_{\theta \sim \mathbb{Q}}[\pi_{\theta}]\,.
\end{align}
As seen in \cref{pac_bayes_for_opl}, this technique allows translating the bound in \cref{thm:main_result}, which controls \(\left|\mathbb{E}_{\theta \sim \mathbb{Q}}[V(\pi_{\theta}) - \hat{V}(\pi_{\theta})]\right|\), into a bound that controls \(|V(\pi_{\mathbb{Q}}) - \hat{V}(\pi_{\mathbb{Q}})|\), the quantity of interest in off-policy learning. The main requirement is to find linear importance-weight regularizations and policies that satisfy \cref{eq:pac_bayes_policy}. Fortunately, many importance-weight regularizations, such as \texttt{Clip}, \texttt{IX}, and \texttt{ES} in \cref{eq:regs}, are linear in \(\pi\), and several practical policies adhere to the formulation in \cref{eq:pac_bayes_policy}; see \cref{pac_bayes_for_opl} for an in-depth explanation of such policies, including softmax, and Gaussian policies.

In \cref{corr:lin_reg_main}, we specialize \cref{thm:main_result} to linear importance-weight regularizations of the form $\hat{w}_\theta(x, a) = \frac{\pi_\theta(a|x)}{h(\pi_0(a|x))}$, where $h(\pi_0(a|x)) \geq \pi_0(a|x)$ for all $(x,a) \in \mathcal{X} \times \mathcal{A}$. We additionally assume that the base policies $\pi_\theta$ are deterministic, i.e., $\pi_\theta(a \mid x) \in \{0, 1\}$, which implies $\pi_\theta(a | x)^2 = \pi_\theta(a | x)$. This assumption is only needed here and it is mild: the PAC-Bayes policies $\pi_{\mathbb{Q}}$ defined in \cref{eq:pac_bayes_policy} are mixtures of deterministic policies under $\mathbb{Q}$, and common policy classes such as softmax, mixed-logit, and Gaussian policies admit such representations (\cref{pac_bayes_for_opl}). Under these assumptions, \cref{thm:main_result} yields the following result.

\begin{corollary}\label{corr:lin_reg_main} Assume the regularized importance weights can be written as \(\hat{w}_\theta(x, a) = \frac{\pi_\theta(a|x)}{h(\pi_0(a|x))}\) with $h:[0,1]\to \mathbb{R}^+$ verifies $h(p) \geq p$ for any $p \in [0, 1]$. Moreover, for any distribution $\mathbb Q$ in the parameter space $\Theta$, we define $\pi_{\mathbb{Q}} = \mathbb{E}_{\theta \sim \mathbb{Q}}[\pi_{\theta}]$ where $\pi_{\theta}$ is binary. Then, let $\lambda>0$,  $n \ge 1$, $\delta \in (0, 1)$, and let $\mathbb{P}$ be a fixed prior on $\Theta$,
The following inequality holds with probability at least $1-\delta$ for any distribution $\mathbb{Q}$ on $\Theta$
\begin{align}
    \Big|V(&\pi_{\mathbb{Q}})-\hat{V}(\pi_{\mathbb{Q}})\Big|  \le \sqrt{ \frac{{\textsc{kl}}_{1}(\mathbb{Q})}{2n} } + B_n(\pi_{\mathbb{Q}})  +
\frac{{\textsc{kl}}_{2}(\mathbb{Q})}{n \lambda } + \frac{\lambda}{2}\operatorname{Var}_n(\pi_{\mathbb{Q}})\,,
\end{align}
where ${\textsc{kl}}_{1}(\mathbb{Q})$ and  ${\textsc{kl}}_{2}(\mathbb{Q})$ are defined in \cref{thm:main_result}, and
\begin{align*}
    &\operatorname{Var}_n(\pi_{\mathbb{Q}}) = \frac{1}{n}\sum_{i=1}^n  \E{A \sim \pi_0(\cdot | X_i)}{\frac{\pi_{\mathbb{Q}}(A |X_i)}{h(\pi_{0}(A |X_i))^2}} +\frac{\pi_{\mathbb{Q}}(A_i |X_i)}{h(\pi_{0}(A_i |X_i))^2} R_i^2\,,\\
   &B_n(\pi_{\mathbb{Q}}) = 1 - \frac{1}{n} \sum_{i=1}^{n} \sum_{A \in \cA}\pi_0(A | X_i)\frac{\pi_{\mathbb{Q}}(A |X_i)}{h(\pi_{0}(A |X_i))}\,.
\end{align*}
\end{corollary}
The main benefit of \cref{corr:lin_reg_main} compared to \cref{thm:main_result} is that it eliminates the need for the expectation \(\E{\theta \sim \mathbb{Q}}{\cdot}\), which is now embedded in the definition of policies in \cref{eq:pac_bayes_policy}. For example, \cref{corr:lin_reg_main} allows us to recover the main result of \texttt{ES} above \citet{aouali23a} when \(h(p) = p^\alpha\), \(\alpha \in [0, 1]\). Similarly, we can apply it to \texttt{IX} \citep{gabbianelli2023importance} by setting \(h(p) = p + \gamma\), \(\gamma \geq 0\), and to \texttt{Clip} \citep{london2019bayesian} by setting \(h(p) = \max(p, \tau)\), \(\tau \in [0, 1]\). However, if \(\hat{w}_\theta(x, a)\) is not linear in \(\pi_{\theta}(a|x)\), then this technique cannot be used, and the original expectation \(\mathbb{E}_{\theta \sim \mathbb{Q}}[\cdot]\) in \cref{thm:main_result} must be retained.

\subsection{Pessimistic Objectives}\label{subsec:lp}

\cref{thm:main_result} yields two pessimistic objectives.

\paragraph{Bound optimization.} The first approach directly maximizes the lower bound from \cref{thm:main_result}:
\begin{align}\label{eq:objective_pac_bayes}
   \argmax_{\mathbb{Q}} \E{\theta \sim \mathbb{Q}}{\hat{V}(\pi_{\theta})} - \sqrt{ \frac{{\textsc{kl}}_{1}(\mathbb{Q})}{2n} } - B_n(\mathbb{Q})  -
\frac{{\textsc{kl}}_{2}(\mathbb{Q})}{n \lambda } - \frac{\lambda}{2}\operatorname{Var}_n(\mathbb{Q})\,,
\end{align}
The main challenge is that the objective involves expectations under $\mathbb{Q}$. We address this using the \emph{local reparameterization trick} \citep{kingma2015variational}, which expresses gradients of expectations as expectations of gradients, estimated via Monte Carlo sampling. Specifically, we consider softmax policies $\pi^{\textsc{sof}}_{\theta}(a | x)$ from \cref{eq:es-softmax_pac_bayes} and set $\mathbb{Q} = \mathcal{N}(\mu, \sigma^2 I_{dK})$ with learnable parameters $\mu \in \mathbb{R}^{dK}$ and $\sigma > 0$. All terms in \cref{eq:objective_pac_bayes} take the form $\mathbb{E}_{\theta \sim \mathcal{N}(\mu, \sigma^2 I_{dK})}[f(\pi^{\textsc{sof}}_{\theta}(a | x))]$, which can be rewritten as:
\begin{align*}
  &\E{\theta \sim \mathcal{N}\left(\mu, \sigma^2 I_{dK}\right)}{f(\pi^{\textsc{sof}}_{\theta}(a | x))} \\
  &= \E{\epsilon \sim \mathcal{N}(0, \|\phi(x)\|_2^2 I_{K})}{f\left(\frac{\exp(\phi(x)^\top \mu_a + \sigma \epsilon_a)}{\sum_{a^\prime \in \cA}\exp(\phi(x)^\top  \mu_{a^\prime} + \sigma \epsilon_{a^\prime})}\right)}\,.
\end{align*}
This expectation is approximated by sampling $\epsilon_i \sim \mathcal{N}(0, \|\phi(x)\|_2^2 I_{K})$ and computing the empirical mean; gradients are estimated similarly. However, this approach can exhibit high variance when $K$ is large. For linear importance-weight regularizations, this can be mitigated by optimizing the bound in \cref{corr:lin_reg_main}. For the general case, we propose a practical alternative.

\paragraph{Heuristic optimization.} The second approach avoids the challenges of direct bound optimization at the cost of additional hyperparameters. Inspired by \cref{thm:main_result}, we maximize the estimated value penalized by bias, variance, and proximity to the logging policy:
\begin{align}\label{eq:learning_principle}
\hat{V}(\pi_{\theta}) - \lambda_1 \|\theta - \theta_0\|^2 - \lambda_2 \tilde{\operatorname{Var}}_n(\pi_{\theta}) - \lambda_3 \tilde{B}_n(\pi_{\theta})\,,
\end{align}
where $\tilde{\operatorname{Var}}_n(\pi_{\theta})$ and $\tilde{B}_n(\pi_{\theta})$ are the terms inside the expectations in $\operatorname{Var}_n(\mathbb{Q})$ and $B_n(\mathbb{Q})$, respectively, $\theta_0$ parameterizes the logging policy $\pi_0$, and $\lambda_1, \lambda_2, \lambda_3$ are hyperparameters.

Both objectives in \cref{eq:objective_pac_bayes,eq:learning_principle} are amenable to stochastic gradient optimization and are generic across importance-weight regularizations, enabling fair comparison. We empirically compare these objectives and evaluate different regularization techniques in \cref{sec:other-experiments}.

\section{Experiments for Other Regularizations}\label{sec:other-experiments}

We adopt the experimental setting of \cref{sec:es-experiments} and conduct two main experiments. In \cref{subsec:fixed_iw}, we fix the importance-weight regularization to \texttt{Clip} (\cref{eq:regs}) and compare our pessimistic objective against PAC-Bayesian objectives from the literature specifically designed for clipping. The goal is to demonstrate that our objective not only applies more broadly but also outperforms existing alternatives. In \cref{subsec:fixed_lp}, having validated our pessimistic objective, we fix it and compare across importance-weight regularizations. The goal is to determine whether any particular regularization technique yields superior off-policy learning performance.

\subsection{Varying Pessimistic Objectives, Fixed Regularization}\label{subsec:fixed_iw}

We examine the impact of different pessimistic objectives on learned policy performance, fixing the importance-weight regularization to \texttt{Clip}: $\hat{w}(x, a) = \frac{\pi(a|x)}{\max(\pi_0(a|x), \tau)}$ in \cref{eq:regs}, with $\tau = 1/\sqrt[4]{n}$ following \citet{ionides2008truncated}. For fair comparison, we consider PAC-Bayesian objectives from prior work where the theoretical bound is optimized directly. Specifically, we include \textbf{London et al.}\ \citep[Theorem 1]{london2019bayesian}, and two bounds from \citet{sakhi2022pac}: \textbf{Sakhi et al.\ 1} \citep[Proposition 1]{sakhi2022pac}, based on \citet{catoni2007pac}, and \textbf{Sakhi et al.\ 2} \citep[Proposition 3]{sakhi2022pac}, a Bernstein-type bound. Since these baselines use linear importance-weight regularization (\cref{subsec:gbound}), we compare against our bound in \cref{corr:lin_reg_main}. Following \citet{sakhi2022pac,aouali23a}, we optimize over Gaussian policies (\cref{eq:es-gaussian_pac_bayes}), which perform better in this setting. We also include the logging policy as a baseline.

\begin{figure}[H]
  \centering  \includegraphics[width=0.7\textwidth]{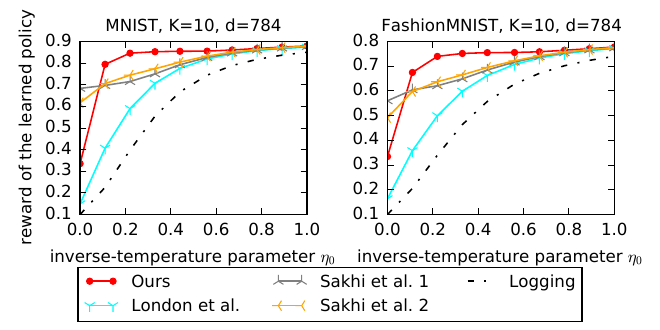}
  \caption{Performance of the learned policy with different PAC-Bayes pessimistic objectives (our \cref{corr:lin_reg_main} and those in \citet{london2019bayesian,sakhi2022pac}) using the \texttt{Clip} IPS estimator in \cref{eq:regs}
.} 
  \label{fig:sota}
\end{figure}

\cref{fig:sota} plots the reward of learned policies as a function of logging policy quality $\eta_0 \in [0, 1]$. Our objective outperforms all baselines across a wide range of logging policies. Thus, in addition to being generic across importance-weight regularizations, our approach proves more effective than objectives tailored specifically for \texttt{Clip}. This advantage holds when $\eta_0$ is not too close to zero: a realistic scenario where logging policies typically outperform uniform random selection. Note that all methods (including ours) improve upon the logging policy (dashed black lines).

\subsection{Varying Regularization, Fixed Pessimistic Objective}\label{subsec:fixed_lp}
Having demonstrated the favorable performance of our pessimistic objective, we now compare different importance-weight regularization techniques: \texttt{Clip}, \texttt{Har}, \texttt{IX}, and \texttt{ES} (\cref{eq:regs}). We evaluate both pessimistic objectives from \cref{subsec:lp}, optimizing over softmax policies. For bound optimization, we use \cref{thm:main_result} rather than \cref{corr:lin_reg_main}, since \texttt{Har} is non-linear in $\pi$. We set $\lambda$ to its optimal value $\lambda_*$ minimizing the bound. While our theory requires $\lambda$ to be fixed a priori (since $\lambda_*$ is data-dependent), we found this yields good empirical performance. For heuristic optimization (\cref{eq:learning_principle}), we set $\lambda_1 = \lambda_2 = \lambda_3 = 10^{-5}$.

\begin{figure}
  \centering  \includegraphics[width=\linewidth]{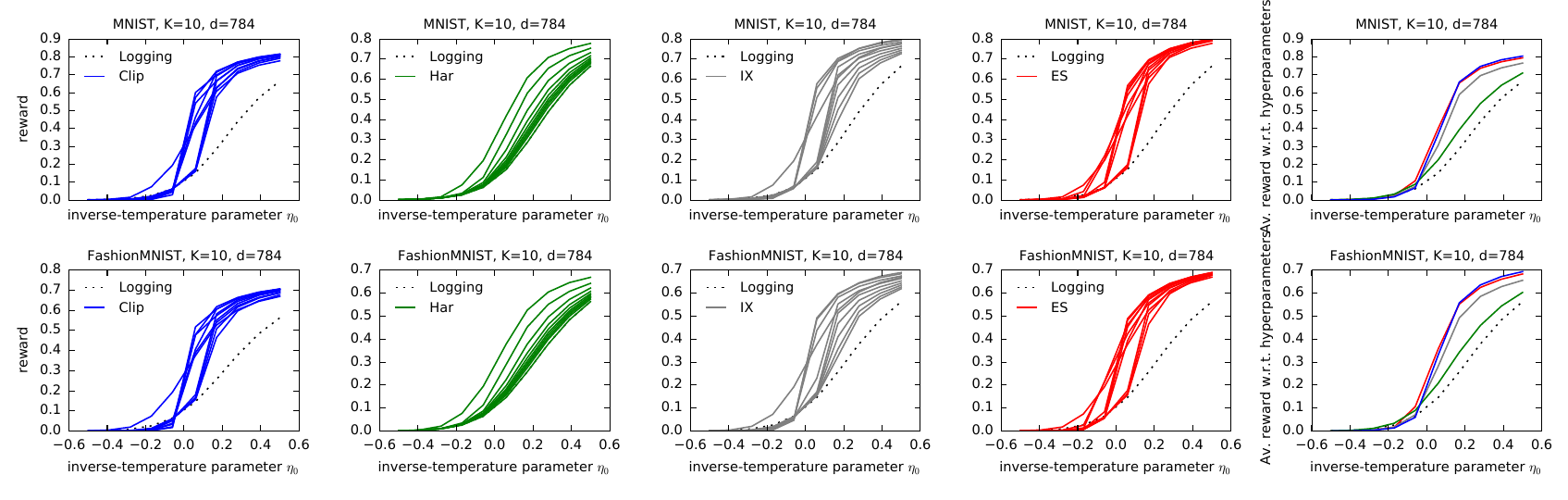}
  \caption{Performance of the policy learned by \textbf{Bound optimization} (i.e, \cref{eq:objective_pac_bayes}) for different importance-weight regularizations. The \(x\)-axis reflects the quality of the logging policy \(\eta_0 \in [-0.5, 0.5]\). In the first four columns, we plot the reward of the learned policy using a fixed importance-weight regularization technique (\texttt{Clip}, \texttt{Har}, \texttt{IX}, or \texttt{ES} as defined in \cref{eq:regs}) for various values of its hyperparameter within \([0,1]\). In the last column, we report the mean reward across these hyperparameter values.} 
  \label{fig:main_exp_results}
\end{figure}

\begin{figure}
  \centering  \includegraphics[width=\linewidth]{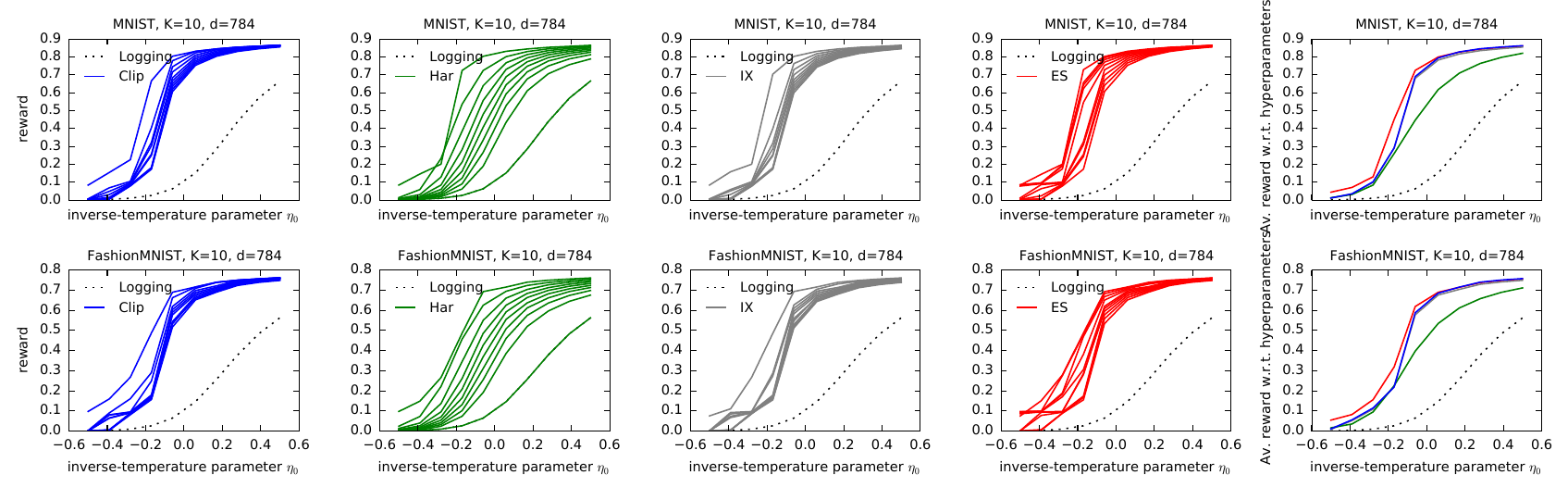}
  \caption{Performance of the policy learned by \textbf{Heuristic optimization} in \cref{eq:learning_principle} for different importance-weight regularizations. The \(x\)-axis reflects the quality of the logging policy \(\eta_0 \in [-0.5, 0.5]\). In the first four columns, we plot the reward of the learned policy using a fixed importance-weight regularization technique (\texttt{Clip}, \texttt{Har}, \texttt{IX}, or \texttt{ES} as defined in \cref{eq:regs}) for various values of its hyperparameter within \([0,1]\). In the last column, we report the mean reward across these hyperparameter values.
  } 
  \label{fig:main_exp_results2}
\end{figure}

\cref{fig:main_exp_results,fig:main_exp_results2} present learned policy rewards as a function of logging policy quality $\eta_0$, for bound optimization and heuristic optimization respectively. We vary $\eta_0 \in [-0.5, 0.5]$, including logging policies worse than uniform ($\eta_0 < 0$) to highlight settings requiring stronger regularization, though such scenarios are rarely encountered in practice. Rows correspond to \texttt{MNIST} and \texttt{FashionMNIST}. The first four columns show results for each regularization technique across hyperparameter values in $[0,1]$; the last column reports mean reward across hyperparameters for each regularization technique to assess sensitivity to hyperparameters.

\textbf{Bound optimization (\cref{fig:main_exp_results}).} All regularizations improve over the logging policy (all curves above the dashed baseline), with \texttt{Har} showing less improvement. \texttt{Clip}, \texttt{IX}, and \texttt{ES} achieve comparable performance despite regularizing importance weights differently. These results align with the generality of our bound and suggest that the choice of regularization has limited impact when optimizing the theoretical bound directly.

\paragraph{Heuristic optimization (\cref{fig:main_exp_results2}).} Heuristic optimization achieves better performance than bound optimization, likely due to practical limitations of Monte Carlo estimation in high dimensions (\cref{subsec:lp}). The far-right column reveals comparable average performance across regularizations, with two exceptions: \texttt{ES} outperforms the others while \texttt{Har} underperforms. This clarifies our results from \cref{sec:es-experiments}: the superior performance of exponential smoothing is more related to our pessimistic objective than the smooth regularization itself. Here, the smooth regularization adds some improvements compared to others, but the improvments are not significant compared to the improvments we get by simply changing the pessimistic learnin principle even with the standard clipping regularization(\cref{fig:sota}) 

\paragraph{Larger action spaces.} The experiments in this section did not consider very large values of $K$. However, \cref{chap:las} evaluated numerous IPS-based methods on datasets with up to one million actions. In those experiments, exponential smoothing outperformed other IPS-based methods, though the improvements were modest. Combined with the results above, this reinforces our conclusion: the choice of the objective has a larger impact on learning performance than the choice of importance-weight regularization, which primarily affects estimation accuracy.

\section{Conclusion}\label{conslution_es}

In this chapter, we investigated importance-weight regularization techniques within the pessimistic paradigm, with particular focus on exponential smoothing as a principled alternative to hard clipping. Our key contributions include: (i) tractable two-sided PAC-Bayes generalization bounds that, unlike prior work, apply to both regularized and standard IPS estimators and are amenable to stochastic gradient optimization; and (ii) the first unified framework for comparing diverse importance-weight regularizations under a common pessimistic objective. This work addresses fundamental theoretical limitations in existing approaches, including the reliance on one-sided inequalities and the misapplication of evaluation bounds in off-policy learning. Rather than using theoretical bounds merely as inspiration for heuristics, we directly optimize them, representing a first step toward making IPS-based pessimism more practical.

Our work has two primary limitations. First, the inclusion of empirical bias and variance terms in our bounds makes deriving data-independent suboptimality gaps challenging. Second, two-sided bounds for regularized IPS can be loose as they treat both tails symmetrically, whereas recent work indicates significant asymmetry between lower and upper tails. We address both limitations in \citet{Sakhi2024LS}, which investigates tail-specific bounds to achieve significantly tighter guarantees and sharp suboptimality results.

This chapter serves practitioners committed to IPS-based methods, whose appeal is well-founded: unbiasedness, theoretical guarantees, and the ability to derive principled pessimistic objectives for safe policy learning in high-stakes scenarios. However, from a purely empirical perspective, particularly in very large action spaces, we would favor the PWLL objectives introduced in \cref{chap:las}, which consistently outperform IPS-based methods. The reader can find a direct comparison of these approaches on datasets with up to one million actions in that chapter.

%% file: contents/conclusion.tex
\chapter{Conclusions and Future Work}

This thesis addressed, from both a practical and theoretical perspective, the core obstacle to deploying contextual bandits in modern applications: \emph{scalability to large action spaces while maintaining computational tractability}.

\textbf{On-Policy Learning (\cref{part:online}).}
We introduced structured Bayesian models that enable principled information sharing across actions and derived scalable exploration algorithms. \texttt{meTS} in \cref{chap:meTS} couples action parameters through shared latent effects, yielding regret and complexity that scale with an \emph{effective} number of actions rather than $K$. \texttt{dTS} in \cref{chap:dTS} further develops this idea by using a pre-trained diffusion model to encode richer structure. These algorithms perform well in practice in their theoretical form, without additional tweaks or hyperparameter tuning.

\textbf{Off-Policy Learning (\cref{part:offline}).}
We tackled both pillars: DM and IPS approaches. \texttt{sDM} in \cref{chap:sDM} extends the structured modeling of \cref{part:online} to the offline regime. We then showed in \cref{chap:las} that, in large action spaces, \emph{optimization} matters more than estimation: estimator-based objectives induce highly non-concave landscapes, whereas policy-weighted log-likelihoods produce concave objectives for common policy classes and win decisively at scale. Finally, we developed a principled pessimistic framework for regularized IPS in \cref{chap:ES}: smooth importance-weight regularization (exponential smoothing) paired with two-sided PAC-Bayes bounds, and a unified analysis that clarifies when regularization matters and how to compare it across methods. Our additional work on \emph{logarithmic smoothing} \citep{Sakhi2024LS} sharpens the concentration analysis further and yields tighter learning guarantees.

\textbf{Thesis message and practical implications.} Scaling to large action spaces causes methods that perform well in small settings (e.g., standard IPS) to fail at scale. This thesis advances three design principles to address this challenge: (i) encode \emph{structure} to shrink the effective action space; (ii) prioritize \emph{objectives with favorable optimization properties} over faithful but intractable estimators; and (iii) when relying on IPS with pessimism, couple \emph{differentiable} importance-weight corrections with theoretically grounded, \emph{data-driven} bounds amenable to stochastic gradient descent. Together, these principles yield algorithms that are statistically efficient, computationally tractable, and numerically stable.

\textbf{Future work.} This thesis opens several promising directions for future research. A key theoretical challenge is to establish \emph{robust guarantees under model misspecification}, extending the Bayesian analysis of \texttt{sDM}, \texttt{meTS}, and \texttt{dTS} beyond the well-specified setting. For on-policy learning, developing a comprehensive \emph{nonlinear diffusion theory for Thompson sampling} remains an open problem. In the off-policy setting, future work could investigate the extensions and applications of our methods to LLM and diffusion model fine-tuning, where the objective closely mirrors offline contextual bandit objectives. Moreover, integrating these approaches into \emph{large-scale recommender pipelines} requires efficient action retrieval, slate constraints, and systems-level optimization. Some of these aspects, such as coupling decision-making with approximate maximum inner product search, were explored in our applied studies (see Additional Contributions in \cref{sec:contributions}) but omitted from this manuscript. These practical directions have a tangible impact on the online advertising industry and beyond, and are worth pursuing.

%% file: contents/mixed_effect_ts/appendix.tex
\section{Preliminaries}\label{sub:preliminary_results}
In this section, we recall some basic properties of matrix operations.

\begin{enumerate}[label=(\alph*)]
    \item \textbf{The mixed-product property.} We have that $(\Alpha \otimes \Beta) (\mathrm{C} \otimes \mathrm{D}) = \Alpha \mathrm{C} \otimes \Beta \mathrm{D}$ for any matrices $\Alpha, \Beta, \mathrm{C}, \mathrm{D}$ such that the products $\Alpha \mathrm{C}$ and $\Beta \mathrm{D}$ exist. 
    \item \textbf{Transpose.} We have that $\left(\Alpha \otimes \Beta\right)^\top = \Alpha^\top \otimes \Beta^\top$ for any matrices $\Alpha, \Beta$.
    \item \textbf{Vectorization.} Let $\Alpha \in \real^{n \times m}, \Beta \in \real^{m \times p}$, then  $\operatorname {Vec}(\Alpha \Beta)=(I_{p} \otimes \Alpha)\operatorname {Vec} (\Beta)=(\Beta^\top \otimes I_{n})\operatorname {Vec} (\Alpha)$.
    \item For any matrix $\Alpha$, we have that $I_1 \otimes \Alpha = \Alpha$.
    \item For any positive semi-definite matrices $\Alpha$ and $\Beta$, we have that $ \lambda_1(\Alpha \otimes \Beta) = \lambda_1(\Alpha) \lambda_1(\Beta)$.
    \item For any matrix $\Alpha$ and any positive semi-definite matrix $\Beta$ such that the product $\Alpha^\top \Beta \Alpha$ exists, the following inequality holds  $\lambda_1(\Alpha^\top \Beta \Alpha) \leq \lambda_1(\Beta ) \lambda_1(\Alpha^\top \Alpha)$.
\end{enumerate}

\section{Posterior Derivations}
\label{app:posterior_derivations}

Here we provide the derivations of the effect posterior and action posteriors for the setting presented in \cref{subsec:contextual_gaussian_bandits}. Precisely, we present the proof for \cref{thm:pt_gaussian} in \cref{app:effect posterior-derivation} and the proof of \cref{thm:pti_gaussian} in \cref{app:conditional-posterior-derivation}.

\subsection{Effect Posterior Derivation}
\label{app:effect posterior-derivation}

\begin{proof}[Proof of \cref{thm:pt_gaussian} (derivation of $q_t$)]
First, from basic properties of matrix operations, we observe that the mean of the action parameter can be rewritten using Kronecker products. Specifically, $\sum_{\ell \in [L]} b_{a, \ell} \psi_{\ell} = \Gamma_a \Psi$, where $\Psi = (\psi_{\ell})_{\ell \in [L]} \in \real^{Ld}$ is the concatenated effect vector and $\Gamma_a = b_a^\top \otimes I_d \in \real^{d \times Ld}$. Thus, the model in \cref{eq:contextual_gaussian_model} (up to round $t \in [T]$) can be written as
\begin{align}
    \Psi & \sim \cN(\mu_{\Psi}, \Sigma_{\Psi}) \,, \nonumber \\
    \theta_{a} \mid \Psi &\sim \cN(\Gamma_a \Psi, \Sigma_{0, a})\,, & \forall a \in \cA\,, \nonumber \\
    R_i \mid X_i, A_i, \theta, \Psi & \sim \cN(X_i^\top \theta_{A_i}, \sigma^2)\,, & \forall i \in [t-1]\,.
\end{align}

Under this model, conditional on $(\theta_a)_{a\in\cA}$ and $(X_i,A_i)_{i<t}$, the rewards
$(R_i)_{i < t}$ are independent and each $R_i$ depends on $\Psi$ only through
$\theta_{A_i}$. Hence
\[
p((R_i)_{i<t} \mid (X_i,A_i)_{i<t},\Psi)
=\int_{\theta \in \real^{dK}} p((R_i)_{i<t} \mid (X_i,A_i)_{i<t},\theta)\,p(\theta\mid \Psi)\,d\theta.
\]
Moreover, since $p(\theta\mid \Psi)=\prod_{a\in\cA}p_{0, a}(\theta_a\mid \Psi)$ and
\[
p((R_i)_{i<t} \mid (X_i,A_i)_{i<t},\theta)=\prod_{a\in\cA}\prod_{i\in S_{t,a}}
\cN(R_i;X_i^\top \theta_a,\sigma^2)
=\prod_{a\in\cA}\LL_{t,a}(\theta_a),
\]
the integral factorizes across arms:
\[
p((R_i)_{i<t} \mid (X_i,A_i)_{i<t},\Psi)
=\prod_{a\in\cA}\int \LL_{t,a}(\theta_a)\,p_{0,a}(\theta_a\mid \Psi)\,d\theta_a.
\]

It follows that the joint effect posterior in round $t$ reads
\begin{align}
    q_t(\Psi)\ &\propto\ p((R_i)_{i<t} \mid (X_i,A_i)_{i<t},\Psi)\,q_0(\Psi)\,,\\
    &= \prod_{a \in \cA} \int_{\theta_a} \LL_{t, a}(\theta_a) p_{0, a}(\theta_a \mid \Psi) \dif \theta_a \, q_0(\Psi) \nonumber \\
    &= \prod_{a \in \cA} \underbrace{\int_{\theta_a} \LL_{t, a}(\theta_a) \cN \left(\theta_a ; \Gamma_a \Psi, \Sigma_{0, a}\right) \dif \theta_a}_{\mathcal{I}_a(\Psi)} \, \cN(\Psi;\mu_{\Psi}, \Sigma_{\Psi})\,,
\end{align}
where $\LL_{t, a}(\theta_a) = \prod_{i \in S_{t, a}} \cN(R_i; X_i^\top \theta_a, \sigma^2)$. 

We compute the integral term $\mathcal{I}_a(\Psi)$ using \cref{lemma:gaussian_posterior_per_arm_contextual}. Specifically, we obtain that $\mathcal{I}_a(\Psi)$ is proportional to a Gaussian density on $\Psi$, denoted $\cN(\Psi; \bar{\mu}_{t, a}, \bar{\Sigma}_{t, a})$, where

\begin{align*}
    \bar{\Sigma}_{t, a}^{-1} & = \Gamma_a^{\top} \left( \Sigma_{0, a}^{-1} - \Sigma_{0, a}^{-1} (G_{t, a} + \Sigma_{0, a}^{-1})^{-1}\Sigma_{0, a}^{-1} \right)\Gamma_a \,, \\
    \bar{\mu}_{t, a} & = \bar{\Sigma}_{t, a}\left( \Gamma_a^\top \Sigma_{0, a}^{-1} (G_{t, a} + \Sigma_{0, a}^{-1})^{-1} B_{t, a}\right) \,,
\end{align*}
and $G_{t,a}$ and $B_{t,a}$ are defined in \cref{sec:mets-linear bandit posterior}. Consequently, the effect posterior $q_t(\Psi)$ is proportional to the product of $K+1$ multivariate Gaussian distributions: the prior $\cN(\mu_{\Psi}, \Sigma_{\Psi})$ and the likelihood contributions $\cN(\bar{\mu}_{t, a}, \bar{\Sigma}_{t, a})$ for each $a \in \cA$. Since the product of Gaussians is Gaussian, $q_t = \cN(\bar{\mu}_t, \bar{\Sigma}_t)$, where the precision matrix is the sum of the individual precisions:
\begin{align*}
    \bar{\Sigma}_t^{-1} &= \Sigma_{\Psi}^{-1} + \sum_{a \in \cA} \bar{\Sigma}_{t, a}^{-1}
    = \Sigma_{\Psi}^{-1} + \sum_{a \in \cA}\Gamma_a^{\top} \left( \Sigma_{0, a}^{-1} - \Sigma_{0, a}^{-1} (G_{t, a} + \Sigma_{0, a}^{-1})^{-1}\Sigma_{0, a}^{-1} \right)\Gamma_a\,.
\end{align*}
Using that $\Gamma_a = b_a^\top \otimes I_d$, we rewrite the term inside the sum as:
\begin{align*}
   \Gamma_a^{\top} \left( \Sigma_{0, a}^{-1} - \Sigma_{0, a}^{-1} (G_{t, a} + \Sigma_{0, a}^{-1})^{-1}\Sigma_{0, a}^{-1} \right)\Gamma_a
    &= (b_a \otimes I_d) \left( \Sigma_{0, a}^{-1} - \Sigma_{0, a}^{-1} (G_{t, a} + \Sigma_{0, a}^{-1})^{-1}\Sigma_{0, a}^{-1} \right) (b_a^\top \otimes I_d) \\
    &= (b_a b_a^\top) \otimes \left( \Sigma_{0, a}^{-1} - \Sigma_{0, a}^{-1} (G_{t, a} + \Sigma_{0, a}^{-1})^{-1}\Sigma_{0, a}^{-1} \right)\,.
\end{align*}
Similarly, for the mean $\bar{\mu}_t$, we have:
\begin{align*}
    \bar{\mu}_t &= \bar{\Sigma}_t \left( \Sigma_{\Psi}^{-1} \mu_{\Psi} + \sum_{a \in \cA} \bar{\Sigma}_{t, a}^{-1}\bar{\mu}_{t, a} \right) \\
    &= \bar{\Sigma}_t \left( \Sigma_{\Psi}^{-1} \mu_{\Psi} + \sum_{a \in \cA} \Gamma_a^\top \Sigma_{0, a}^{-1} (G_{t, a} + \Sigma_{0, a}^{-1})^{-1} B_{t, a} \right)\,.
\end{align*}
Using the mixed (Kronecker) product property, 
$$\Gamma_a^\top \Sigma_{0, a}^{-1} (G_{t, a} + \Sigma_{0, a}^{-1})^{-1} B_{t, a} = b_a \otimes \left(\Sigma_{0, a}^{-1} (G_{t, a} + \Sigma_{0, a}^{-1})^{-1} B_{t, a}\right).$$
This recovers the expressions in \cref{thm:pt_gaussian}.
\end{proof}

To reduce clutter in the following lemma, we fix an action $a \in \cA$ and a round $t$. We drop the sub-indices $a$ and $t$, so that we have the following correspondences:
\begin{align*}
    \Gamma \gets \Gamma_a\,, \quad \ \Sigma_{0} \gets \Sigma_{0, a}\,, \quad N \gets N_{t, a}\,, \quad \theta \gets \theta_a\,, \quad (X_i, R_i)_{i \in [N]} \gets (X_i, R_i)_{i \in S_{t,a}}\,,
\end{align*}

\begin{lemma}[Gaussian posterior update]
\label{lemma:gaussian_posterior_per_arm_contextual}
Let $\Gamma \in \real^{d \times Ld}$, $\Sigma_0 \in \real^{d \times d}$, and $\sigma > 0$. Consider a dataset of $N$ observations $(X_i, R_i)_{i=1}^N$. Then,
\begin{align*}
    \int_\theta \left(\prod_{i = 1}^N \cN(R_i; X_i^\top \theta, \sigma^2)\right) \cN(\theta; \Gamma \Psi, \Sigma_0) \dif\theta \propto \cN\left(\Psi; \mu_N, \Sigma_N \right)\,,
\end{align*}
where
\begin{align*}
    \Sigma_N^{-1} & = \Gamma^{\top} \left( \Sigma_0^{-1} - \Sigma_0^{-1} \left(G_N + \Sigma_0^{-1}\right)^{-1} \Sigma_0^{-1} \right)\Gamma\,, \nonumber \\
    \Sigma_N^{-1} \mu_N & = \left( \Gamma^\top \Sigma_0^{-1} \left(G_N + \Sigma_0^{-1}\right)^{-1} B_N \right) \,.
\end{align*} 
with $G_N = \sigma^{-2} \sum_{i=1}^N X_i X_i^\top$ and $B_N = \sigma^{-2} \sum_{i= 1}^N R_i X_i$.
\end{lemma}

\begin{proof}
Let $v = \sigma^{-2}$ and $\Lambda_0 = \Sigma_0^{-1}$. We denote the integral in the lemma by $f(\Psi)$. Completing the square for $\theta$, we have:
\begin{align*}
    f(\Psi) & \propto \int_\theta \exp\left[
    - \frac{v}{2} \sum_{i= 1}^N (R_i - X_i^\top \theta)^2 -
    \frac{1}{2} (\theta - \Gamma \Psi)^\top \Lambda_0 (\theta - \Gamma \Psi)\right] \dif \theta \\
    & \propto \int_\theta \exp\Big[- \frac{1}{2}
    \Big(\theta^\top \underbrace{\left( v \sum_{i=1}^N X_i X_i^\top + \Lambda_0 \right)}_{V_N^{-1}} \theta - 2 \theta^\top \left(v \sum_{i= 1}^N R_i X_i + \Lambda_0 \Gamma \Psi\right) \\ & \hspace{8cm} + (\Gamma \Psi)^\top \Lambda_0 (\Gamma \Psi)\Big) \Big] \dif \theta \,.
\end{align*}
To reduce clutter, let 
\begin{align*}
    G_N = v \sum_{i=1}^N X_i X_i^\top\,, \quad V_N = \left(G_N + \Lambda_0\right)^{-1}\,, \quad U_N = V_N^{-1}\,,  \\ B_N = v \sum_{i = 1}^N R_i X_i \quad \text{ and } \quad \beta_N = V_N \left(B_N  + \Lambda_0 \Gamma \Psi\right)\,.
\end{align*}
We have that $U_N V_N = V_N U_N = I_{d}\,,$ and thus
 \begin{align*}
  f(\Psi) & \propto \int_\theta \exp\left[- \frac{1}{2}
  \left(\theta^\top U_N \theta - 2 \theta^\top U_N V_N\left(B_N + \Lambda_0 \Gamma \Psi\right) + (\Gamma \Psi)^\top \Lambda_0 (\Gamma \Psi)\right) \right] \dif \theta \,,\\
  & = \int_\theta \exp\left[- \frac{1}{2}
  \left(\theta^\top U_N \theta - 2 \theta^\top U_N \beta_N +  (\Gamma \Psi)^\top \Lambda_0 (\Gamma \Psi)\right) \right] \dif \theta \,,\\
  & = \int_\theta \exp\left[- \frac{1}{2}
  \left( (\theta - \beta_N)^\top U_N (\theta - \beta_N) - \beta_N^\top U_N \beta_N + (\Gamma \Psi)^\top \Lambda_0 (\Gamma \Psi)\right) \right] \dif \theta \,, \\
  & \propto  \exp\left[- \frac{1}{2}
  \left( - \beta_N^\top U_N \beta_N + (\Gamma \Psi)^\top \Lambda_0 (\Gamma \Psi)\right) \right]\,,\\
  & =  \exp\left[- \frac{1}{2}
  \left( -  \left(B_N + \Lambda_0 \Gamma \Psi\right)^\top  V_N \left(B_N + \Lambda_0 \Gamma \Psi\right) + (\Gamma \Psi)^\top \Lambda_0 (\Gamma \Psi)\right) \right] \,,\\
  & \propto  \exp\left[- \frac{1}{2}
  \left( \Psi^{\top} \Gamma^{\top} \left( \Lambda_0 - \Lambda_0 V_N \Lambda_0 \right) \Gamma \Psi - 2 \Psi^\top \left(\Gamma^\top \Lambda_0 V_N B_N\right) \right) \right] \,,\\
  & = \exp\left[- \frac{1}{2} \Psi^\top \Sigma_N^{-1} \Psi + \Psi^\top \Sigma_N^{-1} \mu_N \right]\,,
\end{align*}
where 
\begin{align}\label{eq:effect_posterior_formulas}
    \Sigma_N^{-1} & = \Gamma^{\top} \left( \Lambda_0 - \Lambda_0 V_N \Lambda_0 \right)\Gamma\,, \nonumber \\
    \Sigma_N^{-1} \mu_N & = \left( \Gamma^\top \Lambda_0 V_N B_N \right) \,.
\end{align} 
Plugging the expression of $V_N$ concludes the proof.
\end{proof}

\subsection{Action Posterior Derivation}\label{app:conditional-posterior-derivation}

\begin{proof}[Proof of \cref{thm:pti_gaussian} (Derivation of $p_{t, a}$)]
This proposition is a direct application of \cref{lemma:posterior_theta}; in which case we get that the posterior $p_{t, a}$ is a multivariate Gaussian distribution $\cN(\tilde{\mu}_{t, a}, \tilde{\Sigma}_{t, a})$, where
\begin{align*}
    \tilde{\Sigma}_{t, a}^{-1} &=  G_{t, a} + \Sigma_{0, a}^{-1}\,, \nonumber\\
    \tilde{\mu}_{t, a} &= \tilde{\Sigma}_{t, a} \left( B_{t, a} +  \Sigma_{0, a}^{-1} \sum_{\ell=1}^L b_{a, \ell} \psi_{t, \ell}\right)\,.
\end{align*}
\end{proof}

To reduce clutter, we consider a fixed action $a \in [K]$ and round $t \in [T]$, and drop subindexing by $t$ and $a$ in \cref{lemma:posterior_theta}. In summary, fix $a \in [K]$ and $t\in [T]$ such that we have the following correspondences:
\begin{align*}
    b_\ell \gets b_{a, \ell}\,, \quad \ \Sigma_{0} \gets \Sigma_{0, a}\,, \quad N \gets N_{t, a}\,, \quad \theta \gets \theta_{a}\,, \quad (X_i, R_i)_{i \in [N]} \gets (X_i, R_i)_{i \in S_{t, a}}\,.
\end{align*}

\begin{lemma}\label{lemma:posterior_theta}
Consider the following model
\begin{align*}
    \theta  \mid \Psi & \sim \cN\left(\sum_{\ell=1}^L b_\ell \psi_{\ell}, \Sigma_0\right)\,,\\
    R_i  \mid X_i, \theta & \sim \cN\left(X_i^\top \theta, \sigma^2\right)\,, & \forall i \in [N]\,.
\end{align*}
Let $H = \set{X_1, R_1, \ldots, X_N, R_N }$ then we have that $p(\theta \mid \Psi, H) = \cN\left(\theta; \tilde{\mu}_{N}, \tilde{\Sigma}_{N}\right),$
where 
\begin{align*}
\tilde{\Sigma}^{-1}_{N} &= \sigma^{-2}\sum_{i=1}^N X_i X_i^\top + \Sigma_{0}^{-1}\,,\\
   \tilde{\mu}_{N} &=   \tilde{\Sigma}_{N}\left( \sigma^{-2} \sum_{i=1}^N X_i R_i + \Sigma_{0}^{-1} \sum_{\ell=1}^L b_\ell \psi_\ell \right)\,.
\end{align*}

\end{lemma}

\begin{proof}
Let $v
  = \sigma^{-2}\,, \quad
  \Lambda_0
  = \Sigma_0^{-1}\,.$ Then the action posterior decomposes as
\begin{align*}
    &p(\theta \mid \Psi, H) \propto p((R_i)_{i \in [N]} \mid \Psi, \theta, (X_i)_{i \in [N]}) p(\theta \mid \Psi)\,,  \\
    &= p((R_i)_{i \in [N]} \mid \theta, (X_i)_{i \in [N]}) p(\theta \mid \Psi)\,,  \\
    & = \prod_{i=1}^N \cN(R_i; X_i^\top \theta, \sigma^2) \cN(\theta; \sum_{\ell=1}^L b_\ell \psi_\ell , \Sigma_{0})\,,\\
    & = \exp\Big[- \frac{1}{2}
  \Big(v \sum_{i=1}^N (R_i^2 - 2 R_i X_i^\top \theta + (X_i^\top\theta)^2) + \theta^\top \Lambda_{0} \theta - 2 \theta^\top \Lambda_{0} \sum_{\ell=1}^L b_\ell \psi_\ell \\ & \hspace{8cm} +\left(\sum_{\ell=1}^L b_\ell \psi_\ell\right)^\top \Lambda_{0} \left(\sum_{\ell=1}^L b_\ell \psi_\ell\right)\Big)\Big] \,,\\
  &\propto \exp\left[- \frac{1}{2}
  \left( \theta^\top(v\sum_{i=1}^N X_i X_i^\top + \Lambda_{0}) \theta - 2 \theta^\top \left(v \sum_{i=1}^N X_i R_i + \Lambda_{0} \sum_{\ell=1}^L b_\ell \psi_\ell \right)\right)\right]\,,\\
  &\propto \cN\left(\theta; \tilde{\mu}_{N}, \left(\tilde{\Lambda}_{N}\right)^{-1}\right)\,,
\end{align*}
where $\tilde{\Lambda}_{N} = v\sum_{i=1}^N X_i X_i^\top + \Lambda_{0}\,,$ and $ \tilde{\Lambda}_{N} \tilde{\mu}_{N} = v \sum_{i=1}^N X_i R_i + \Lambda_{0} \sum_{\ell=1}^L b_\ell \psi_\ell\,.$
\end{proof}

\section{Regret Proofs}\label{proof:regret_proof}
In this section, we establish a more general version of \cref{thm:regret}. As explained in \cref{subsec:contextual_gaussian_bandits}, we analyze \meTS{} in the linear setting under the assumption of a \emph{fully well-specified} model. That is, the true action parameters and rewards are generated according to the same hierarchical structure assumed by \meTS:
\begin{align}\label{eq:true_contextual_gaussian_model}
    \Psi_* & \sim \cN(\mu_{\Psi}, \Sigma_{\Psi})\,, \\ 
    \theta_{*, a} \mid \Psi_* & \sim \cN\Big( \sum_{\ell =1}^L b_{a, \ell} \psi_{*, \ell} , \,  \Sigma_{0, a}\Big)\,, & \forall a \in \cA\,,\nonumber\\ 
    R_t \mid X_t, A_t, \theta_*, \Psi_* & \sim \cN(X_t^\top \theta_{*, A_t} ,  \sigma^2)\,, & \forall t \in [T]\,, \nonumber
\end{align}
where the subscript $*$ denotes the true action and latent parameters.

To derive the regret bound, we proceed as follows: First, we provide a compact problem formulation in \cref{subsec:regre_proof_1}. Next, we employ total covariance decomposition to derive the posterior covariance of $\theta_{*, a} \mid H_t$ in \cref{subsec:regre_proof_2}. Finally, we present preliminary eigenvalue results in \cref{seubsec:prem_ineq} before completing the proof in \cref{subsec:regret_proof_3}.
\subsection{Problem Reformulation for Regret Analysis}\label{subsec:regre_proof_1}

Here, we aim at rewriting \cref{eq:true_contextual_gaussian_model} in a compact form to simplify regret analysis. We first introduce $K$ independent multivariate Gaussian variables $Z_a \sim \cN(0, \Sigma_{0, a})$ for $a \in [K]$, and the following matrix
\begin{align*}
    \Psi_{*, \text{mat}} &= [\psi_{*, 1}, \ldots, \psi_{*, L}] \in \real^{d \times L}\,.
\end{align*}
First, we have that $\operatorname{Vec}(\Psi_{*, \text{mat}})=\Psi_*$ where $\Psi_*$ is defined in \cref{eq:true_contextual_gaussian_model}. Moreover notice that $\sum_{\ell=1}^L b_{a, \ell} \psi_{*, \ell} = \Psi_{*, \text{mat}} b_a$, where $b_a = (b_{a, \ell})_{\ell \in [L]}$ and thus given matrix $\Psi_{*, \text{mat}}$ we have that
\begin{align}\label{eq:params_per_arm}
\theta_{*, a} &= \Psi_{*, \text{mat}} b_a + Z_a\,, &\forall a \in [K] \,.
\end{align}
We vectorize \cref{eq:params_per_arm} to obtain
\begin{align}
    \theta_{*, a} = \operatorname {Vec}(\theta_{*, a}) = \operatorname {Vec}(\Psi_{*, \text{mat}} b_a + Z_a) =\operatorname {Vec}(\Psi_{*, \text{mat}} b_a) + Z_a\,,
\end{align}
 where we used that if $X \in \real^d$ (a column vector), then $X = \operatorname {Vec}(X)$ and that $\operatorname {Vec}(\cdot)$ is a linear transformation. Also, we know from (c) in \cref{sub:preliminary_results} that $\operatorname {Vec}(\Alpha \Beta)=(\Beta^\top \otimes I_{n})\operatorname {Vec} (\Alpha)$ for any $\Alpha \in \real^{n \times m}, \Beta \in \real^{m \times p}$. Therefore,
\begin{align}\label{eq:vec_params}
    \theta_{*, a} = \Gamma_a \Psi_* + Z_a\,,
\end{align}
where $\Gamma_a = b_a^\top \otimes I_d$ and we used that $\operatorname {Vec}(\Psi_{*, \text{mat}}) = \Psi_*$. It follows that 
\begin{align}\label{eq:reformulation_per_arm}
     \theta_{*, a} \mid \Psi_* \sim \cN(\Gamma_a  \Psi_*, \Sigma_{0, a})\,,
\end{align}

This allows us to rewrite our model as a single-parent hierarchical model
\begin{align}\label{eq:model_rewritten}
    \Psi_{*} & \sim \cN(\mu_{\Psi}, \Sigma_{\Psi})\,, \\ 
         \theta_{*, a} \mid \Psi_* & \sim \cN(\Gamma_a  \Psi_*, \Sigma_{0, a})\,, & \forall a \in [K]\,,\nonumber\\ 
     R_t \mid X_t, A_t, \theta_*, \Psi_* & \sim \cN(X_t^\top \theta_{*, A_t} ,  \sigma^2)\,, & \forall t \in [T]\,. \nonumber
\end{align}

\subsection{Derivation of $\condcov{\theta_{*, a}}{H_t}$}\label{subsec:regre_proof_2}
Let 
\begin{align*}
  G_{t, a}
  &= \sigma^{-2} \sum_{i \in S_{t, a}} X_i X_i^\top \,,
  \quad
  B_{t, a}
  = \sigma^{-2} \sum_{i \in S_{t, a}} R_i X_i \,.
\end{align*}

\begin{lemma}[Expression of $\condcov{\theta_{*, a}}{H_t}$]\label{lemma:gaussian_covariance} Consider the model in \cref{eq:model_rewritten}, then we have
\begin{align*}
  \hat{\Sigma}_{t, a} &= \condcov{\theta_{*, a}}{H_t} = \tilde{\Sigma}_{t, a}  +\tilde{\Sigma}_{t, a} \Sigma_{0, a}^{-1}  \Gamma_a \bar{\Sigma}_{t} \Gamma_a^\top \Sigma_{0, a}^{-1} \tilde{\Sigma}_{t, a}\,, & \forall a \in [K]\,.
\end{align*}
where 
\begin{align*}
    \bar{\Sigma}_{t} &= \Big(\Sigma_\Psi^{-1} + \sum_{a=1}^K b_a b_a^{\top} \otimes \left( \Sigma_{0, a}^{-1} - \Sigma_{0, a}^{-1} (G_{t, a} + \Sigma_{0, a}^{-1})^{-1} \Sigma_{0, a}^{-1} \right)\Big)^{-1}\\
    \tilde{\Sigma}_{t, a} &=\left( G_{t, a} + \Sigma_{0, a}^{-1} \right)^{-1}\,.
\end{align*}
\end{lemma}
\begin{proof}
Before proceeding with the proof, we emphasize that 
$$\condcov{\Psi_*}{H_t} = \condcov{\Psi}{H_t} = \bar{\Sigma}_{t}\,, \qquad \condE{\Psi_*}{H_t} = \condE{\Psi}{H_t} = \bar{\mu}_{t}\,,$$ and 
$$\condcov{\theta_{*, a}}{\Psi_*, H_t} = \condcov{\theta_{a}}{\Psi, H_t}= \tilde{\Sigma}_{t, a}\,, \qquad \condE{\theta_{*, a}}{\Psi_*, H_t} = \condE{\theta_{a}}{\Psi, H_t}= \tilde{\mu}_{t, a}\,,$$
where the explicit expressions of these covariances and expectations are provided in \cref{thm:pt_gaussian} and \cref{thm:pti_gaussian}, respectively. These equalities hold because the true action parameters and rewards are assumed to follow the exact generative process defined by the \meTS{} model.

Now let $\Lambda_{0, a} =  \Sigma_{0, a}^{-1}$. \cref{thm:pti_gaussian} and the fact that $ \sum_{\ell \in [L]} b_{a, \ell} \psi_{*, \ell} = \Gamma_a \Psi_*$ where $\Gamma_a = b_a^\top \otimes I_d$ (\cref{subsec:regre_proof_1}) yield
\begin{align*}
    \condcov{\theta_{*, a}}{\Psi_*, H_t} &= \left( G_{t, a} + \Lambda_{0, a} \right)^{-1} \\
    \condE{\theta_{*, a}}{\Psi_*, H_t} & = \condcov{\theta_{*, a}}{\Psi_*, H_t} \left( B_{t, a}  + \Lambda_{0, a} \Gamma_a \Psi_* \right)
\end{align*}
First, given $H_t$, $\condcov{\theta_{*, a}}{\Psi_*, H_t} =\left( G_{t, a} + \Lambda_{0, a} \right)^{-1}$ is constant (does not depend on $\Psi_*$). Thus 
\begin{align*}   \condE{\condcov{\theta_{*, a}}{\Psi_*, H_t}}{H_t} = \condcov{\theta_{*, a}}{\Psi_*, H_t} = \left( G_{t, a} + \Lambda_{0, a} \right)^{-1}\,.
\end{align*}
In addition, given $H_t$, both $\left( G_{t, a} + \Lambda_{0, a} \right)^{-1}$ and $B_{t, a}$ are constant. Thus
\begin{align*}
    \condcov{\condE{\theta_{*, a}}{\Psi_*, H_t}}{H_t} &= \condcov{\condcov{\theta_{*, a}}{\Psi_*, H_t}\Lambda_{0, a} \Gamma_a \Psi_*}{H_t}\\
    &= \left( G_{t, a} + \Lambda_{0, a} \right)^{-1} \Lambda_{0, a} \Gamma_a \condcov{\Psi_*}{H_t} \Gamma_a^\top \Lambda_{0, a} \left( G_{t, a} + \Lambda_{0, a} \right)^{-1}\\
    &= \left( G_{t, a} + \Lambda_{0, a} \right)^{-1} \Lambda_{0, a} \Gamma_a \bar{\Sigma}_t \Gamma_a^\top \Lambda_{0, a} \left( G_{t, a} + \Lambda_{0, a} \right)^{-1}.
\end{align*}
Finally, total covariance decomposition \citep{weiss05probability} concludes the proof.
\end{proof}

\subsection{Preliminary Eigenvalues Results}\label{seubsec:prem_ineq}
Next we present some preliminary upper bounds on the maximum eigenvalues of our covariance matrices.
\setlist{nolistsep}
\begin{itemize}[noitemsep]
    \item \textbf{Definitions:} Let $\lambda_{1, 0} = \max_{a \in [K]}  \lambda_1(\Sigma_{0, a})\,,$ $\lambda_{d, 0}  = \min_{a \in [K]}  \lambda_d(\Sigma_{0, a})\,,$ $\lambda_{1, \Psi} = \lambda_1(\Sigma_{\Psi})\,,$ and $\kappa_b = \max_{a \in [K]} \normw{b_a}{2}^2 $.
    \item \textbf{upper bound of $\lambda_1(\Gamma_a \Gamma_a^\top)$:}\begin{align}\label{eq:max_eigenvalue_mix_weights}
   \lambda_1(\Gamma_a \Gamma_a^\top) & \leq \kappa_b \,, & \forall a \in [K]\,.
    \end{align} 
    Similarly, we have that
    \begin{align}
   \lambda_1(\Gamma_a^\top \Gamma_a) & \leq \kappa_b \,, & \forall a \in [K]\,.
    \end{align} 
    \item \textbf{upper bound of $\lambda_1(\hat{\Sigma}_{t, a})$:} \begin{align}\label{eq:max_eigenvalue_covariance}
    \lambda_1(\hat{\Sigma}_{t, a}) & \leq \lambda_{1, 0} + \frac{\lambda_{1, 0}^2 \lambda_{1, \Psi} \kappa_b }{\lambda_{d, 0}^2} \,, & \forall a \in [K]\,.
\end{align} 
    \item \textbf{upper bound of $\lambda_1(\Sigma_\Psi^\frac{1}{2} \bar{\Sigma}_{T+1}^{-1} \Sigma_\Psi^\frac{1}{2})$:}\begin{align}\label{eq:max_eigenvalue_effect_covariance}
    \lambda_1(\Sigma_\Psi^\frac{1}{2} \bar{\Sigma}_{T+1}^{-1} \Sigma_\Psi^\frac{1}{2}) & \leq 1 + K \lambda_{1, \Psi} \kappa_b \Big( \frac{1}{\lambda_{d, 0}} - \frac{1}{\lambda_{1, 0}^2\big(\frac{\kappa_x T}{\sigma^2} + \frac{1}{\lambda_{d, 0}} \big)} \Big) \,.
\end{align} 
\end{itemize}

\begin{proof}
We start with \cref{eq:max_eigenvalue_mix_weights}. First, recall that $\Gamma_a = b_a^\top \otimes I_d$ for any $a \in [K]$. Thus $\Gamma_a \Gamma_a^\top = (b_a^\top \otimes I_d) (b_a \otimes I_d) = \normw{b_a}{2}^2 I_d$ for any $a \in [K]$. Then $\lambda_1(\Gamma_a \Gamma_a^\top) = \normw{b_a}{2}^2 \leq \kappa_b\,.$ The second result follows from the fact that $\lambda_1(\Gamma_a \Gamma_a^\top)=\lambda_1(\Gamma_a^\top \Gamma_a)$. 

Now we prove the result in \cref{eq:max_eigenvalue_covariance}. This follows from the expression of $\hat{\Sigma}_{t, a}$ in \cref{lemma:gaussian_covariance}. Precisely, we have that 
\begin{align*}
  \hat{\Sigma}_{t, a} & = \tilde{\Sigma}_{t, a}  +\tilde{\Sigma}_{t, a} \Sigma_{0, a}^{-1}  \Gamma_a \bar{\Sigma}_{t} \Gamma_a^\top \Sigma_{0, a}^{-1} \tilde{\Sigma}_{t, a}\,, & \forall a \in [K]\,.
\end{align*}
where $\tilde{\Sigma}_{t, a} =\left( G_{t, a} + \Sigma_{0, a}^{-1} \right)^{-1}$. Thus Weyl's inequality combined with the properties in \cref{sub:preliminary_results} yields that
\begin{align*}
    \lambda_1(\hat{\Sigma}_{t, a}) & \leq \lambda_1(\tilde{\Sigma}_{t, a}) + \lambda_1(\tilde{\Sigma}_{t, a}) \lambda_1({\Sigma}_{0, a}^{-1})\lambda_1(\Gamma_a\bar{\Sigma}_{t}\Gamma_a^\top) \lambda_1({\Sigma}_{0, a}^{-1}) \lambda_1(\tilde{\Sigma}_{t, a}) \leq  \lambda_{1, 0} + \frac{\lambda_{1, 0}^2 \lambda_{1, \Psi} \kappa_b }{\lambda_{d, 0}^2}
\end{align*} 
In the last inequality, we used that $\lambda_1(\Gamma_a\bar{\Sigma}_{t}\Gamma_a^\top) \leq \lambda_1(\bar{\Sigma}_{t}) \lambda_1(\Gamma_a \Gamma_a^\top),$ ((f) in \cref{sub:preliminary_results}),  $\lambda_1({\Sigma}_{0, a}^{-1}) \leq \frac{1}{\lambda_{d, 0}}\,,$ and $\lambda_1(\tilde{\Sigma}_{t, a}) \leq \lambda_{1,0}$.

Finally, we prove the result in \cref{eq:max_eigenvalue_effect_covariance}. First, we rewrite the precision matrix of the effect posterior $\bar{\Sigma}_t^{-1}$ using the compact notation introduced in \cref{subsec:regre_proof_1}. Precisely, it follows from \cref{eq:effect_posterior_formulas} that
\begin{align*}
    \bar{\Sigma}_t^{-1} &\stackrel{(i)}{=}  \Sigma_\Psi^{-1} + \sum_{a=1}^K\Gamma_a^{\top} \left( \Sigma_{0, a}^{-1} - \Sigma_{0, a}^{-1} (G_{t, a} + \Sigma_{0, a}^{-1})^{-1} \Sigma_{0, a}^{-1} \right)\Gamma_a \,,\nonumber\\
    &\stackrel{(ii)}{=}  \Sigma_\Psi^{-1} + \sum_{a=1}^K\Gamma_a^{\top} \left( \Sigma_{0, a}^{-1} - \Sigma_{0, a}^{-1} \tilde{\Sigma}_{t, a} \Sigma_{0, a}^{-1} \right)\Gamma_a \,.
\end{align*}
Here, $(i)$ and $(ii)$ are the same; $(ii)$ follows from plugging $\tilde{\Sigma}_{t, a} = (G_{t, a} + \Sigma_{0, a}^{-1})^{-1}$ in $(i)$. Then we have that
    \begin{align*}
    &\lambda_1(\Sigma_\Psi^\frac{1}{2} \bar{\Sigma}_{T+1}^{-1} \Sigma_\Psi^\frac{1}{2})\\
    &=\lambda_1\Big(I_{Ld} + \sum_{a=1}^K \Sigma_\Psi^\frac{1}{2}\Gamma_a^{\top} \left( \Sigma_{0, a}^{-1} - \Sigma_{0, a}^{-1} \tilde{\Sigma}_{T+1, a} \Sigma_{0, a}^{-1} \right)\Gamma_a \Sigma_\Psi^\frac{1}{2} \Big) \\
    &\leq 1+ \lambda_{1, \Psi}\sum_{a=1}^K\lambda_1\Big(\Gamma_a^{\top} \left( \Sigma_{0, a}^{-1} - \Sigma_{0, a}^{-1} \tilde{\Sigma}_{T+1, a} \Sigma_{0, a}^{-1} \right)\Gamma_a\Big) \nonumber\,,\\
    &\leq 1+ \lambda_{1, \Psi} \sum_{a=1}^K\lambda_1(\Gamma_a^{\top}\Gamma_a) \lambda_1\left( \Sigma_{0, a}^{-1} - \Sigma_{0, a}^{-1} \tilde{\Sigma}_{T+1, a} \Sigma_{0, a}^{-1} \right) \\
    &\leq 1+ \lambda_{1, \Psi}\sum_{a=1}^K\kappa_b \left( \lambda_1\left( \Sigma_{0, a}^{-1}\right) + \lambda_1\left(-\Sigma_{0, a}^{-1} \tilde{\Sigma}_{T+1, a} \Sigma_{0, a}^{-1} \right) \right)\nonumber\,,\\
    &\leq 1+ \lambda_{1, \Psi} \sum_{a=1}^K\kappa_b \left( \frac{1}{\lambda_{d, 0}} - \lambda_d\left(\Sigma_{0, a}^{-1} \tilde{\Sigma}_{T+1, a} \Sigma_{0, a}^{-1} \right) \right)\nonumber \\
    &\leq  1+ \lambda_{1, \Psi} \sum_{a=1}^K\kappa_b \left( \frac{1}{\lambda_{d, 0}} - \lambda_d\left(\Sigma_{0, a}^{-1}\right)\lambda_d\left( \tilde{\Sigma}_{T+1, a} \right)\lambda_d\left(\Sigma_{0, a}^{-1} \right) \right)\nonumber\,,\\
    &\leq  1+ \lambda_{1, \Psi} \sum_{a=1}^K\kappa_b \left( \frac{1}{\lambda_{d, 0}} - \frac{1}{\lambda_{1, 0}^2}\lambda_d\left( \tilde{\Sigma}_{T+1, a} \right) \right)\nonumber \\
    &= 1+ \lambda_{1, \Psi} \sum_{a=1}^K\kappa_b \left( \frac{1}{\lambda_{d, 0}} - \frac{1}{\lambda_{1, 0}^2\lambda_1\left( G_{T+1, a} + \Sigma_{0, a}^{-1} \right)} \right)\nonumber\,,\\
    &\leq  1+ \lambda_{1, \Psi} \sum_{a=1}^K\kappa_b \left( \frac{1}{\lambda_{d, 0}} - \frac{1}{\lambda_{1, 0}^2\left(\frac{\kappa_x T}{\sigma^2} + \frac{1}{\lambda_{d, 0}} \right)} \right) \\
    &=  1+K \lambda_{1, \Psi} \kappa_b \left( \frac{1}{\lambda_{d, 0}} - \frac{1}{\lambda_{1, 0}^2\left(\frac{\kappa_x T}{\sigma^2} + \frac{1}{\lambda_{d, 0}} \right)} \right)\nonumber\,.
\end{align*}
\end{proof}

\subsection{Regret Proof}\label{subsec:regret_proof_3}
Here we prove a more general version of \cref{thm:regret} where we do not assume that the covariance matrices $\Sigma_{0, a}$ and $\Sigma_{\Psi}$ are diagonal. We still assume that there exists $\kappa_x>0$ such that $\normw{X_t}{2}^2 \leq \kappa_x$ for any $t \in [T]$.

\begin{theorem}[General version of \cref{thm:regret}] For any $\delta \in (0, 1)$, the Bayes regret of \emph{\meTS} in the mixed-effect model in \cref{subsec:contextual_gaussian_bandits} is bounded as
\begin{align}
  \mathcal{B}\mathcal{R}(T)
  \leq \sqrt{2 T 
  \left( \mathcal{R}^{\textsc{a}}(T) + \mathcal{R}^{\textsc{e}}(T) \right) \log(1 / \delta)} +
  c T\delta\,,
\end{align}
with $c = \sqrt{ \frac{2}{\pi} \kappa_x\big( \lambda_{1, 0} + \frac{\lambda_{1, 0}^2 \lambda_{1, \Psi}\kappa_b }{\lambda_{d, 0}^2}\big)}K \,, \, \, \kappa_b = \max_{a \in [K]}\normw{b_a}{2}^2\,, \, \, \lambda_{1, 0}=\max_{a \in [K]}\lambda_1(\Sigma_{0, a})\,, \, \, \lambda_{d, 0} = \min_{a \in [K]}\lambda_d(\Sigma_{0, a})\,,$ $\lambda_{1, \Psi} = \lambda_1(\Sigma_{\Psi})$ and
\begin{align*}
& \mathcal{R}^{\textsc{a}}(T) = dK c_\textsc{a} \log\big(1 + \frac{T\kappa_x \lambda_{1, 0}  }{\sigma^2 d}\big)\,, \, c_\textsc{a} = \frac{\kappa_x \lambda_{1, 0}}{\log(1 + \frac{\kappa_x \lambda_{1, 0}}{\sigma^{2}})}\,,\\ 
&\mathcal{R}^{\textsc{e}}(T) = dL c_\textsc{e} \log\Big(1+K \kappa_b \lambda_{1, \Psi} \Big( \frac{1}{\lambda_{d, 0}} - \frac{1}{\lambda_{1, 0}^2\big(\frac{\kappa_x T}{\sigma^2} + \frac{1}{\lambda_{d, 0}} \big)} \Big)\Big)\,,  \, c_\textsc{e}= \frac{\kappa_x \kappa_b \lambda_{1, 0}^2 \lambda_{1, \Psi} \big(1 + \frac{\kappa_x\lambda_{1, 0}}{\sigma^{2}}  \big)}{\lambda_{d, 0}^2\log\big(1 + \frac{\kappa_x \kappa_b \lambda_{1, 0}^2 \lambda_{1, \Psi}}{\sigma^{2} \lambda_{d, 0}^2}\big)}\,.&
\end{align*}
In particular, the result in \cref{thm:regret} is retrieved when $\lambda_{1, 0}= \lambda_{d, 0}=\sigma_0^2\,, \text{ and }\lambda_{1, \Psi}=\sigma_\Psi^2\,.$
\end{theorem}

\begin{proof}

Consider our model rewritten in \cref{eq:model_rewritten}. Then, the posterior distribution of the action parameter $\theta_{*, a} \mid H_t$ is a multivariate Gaussian distribution $\cN(\hat{\mu}_{t, a} , \hat{\Sigma}_{t, a})$ for some $\hat{\mu}_{t, a} \in \real^{d}$ and $\hat{\Sigma}_{t, a} \in \real^{d \times d}$ (\cref{lemma:gaussian_covariance}). Now we let $\theta_{t,*} = (X_t^\top \theta_{*, a})_{a \in [K]} \in \real^K$ be the concatenation of the expected rewards of actions in round $t$. Notice that the context $X_t$ is known in round $t$, and thus we include it in the history $H_t$. This is important, with slight abuse of notation, $H_t$ now denotes $H_t \gets H_t \cup \{X_t\}$. Then, the joint posterior of the expected rewards, $\theta_{t,*}  \mid H_t$, is also a multivariate Gaussian $\cN(\check{\theta}_t, \check{\Sigma}_t)$ for $\check{\theta}_t = (X_t^\top \hat{\mu}_{t, a})_{a \in [K]} \in \real^K$ and some covariance $\check{\Sigma}_t \in \real^{K \times K}$. This follows from the properties of Gaussian distributions \citep{koller09probabilistic} and the fact that $X_t$ is now included in $H_t$. Let $\bm{A}_{t} \in \{0, 1\}^K$ and $\bm{A}_{t, *} \in \{0, 1\}^K$ be indicator vectors of the taken action $A_t$ and optimal action $A_{t, *}$, respectively (The vector representations are in bold letters while the integer representations are in regular letters). Then the Bayes regret can be rewritten and consequently decomposed following standard analysis \citep{russo14learning} as 
\begin{align}\label{standard_decomposition_app} \mathcal{B}\mathcal{R}(T)
 &= \mathbb{E}\left[ \sum_{t = 1}^T X_t^\top \theta_{*, A_{t, *}} - X_t^\top \theta_{*, A_{t}}\right]\,,\\
 &= \mathbb{E}\Big[ \sum_{t = 1}^T \bm{A}_{t, *}^\top \theta_{t,*} - \bm{A}_{t}^\top \theta_{t,*}\Big]\,,\\
 &= \sum_{t=1}^T \E{}{\condE{\bm{A}_{t, *}^\top (\theta_{t, *} - \check{\theta}_t)}{H_t}} +
  \E{}{\condE{\bm{A}_t^\top (\check{\theta}_t - \theta_{t, *})}{H_t}}\,.\nonumber
\end{align}
This follows from the fact that $\check{\theta}_t = (X_t^\top \hat{\mu}_{t, i})_{i \in [K]}$ is deterministic given $H_t$ (since $H_t$ now includes $X_t$), and that $\bm{A}_{t, *}$ and $\bm{A}_t$ are i.i.d. given $H_t$. Moreover, given $H_t$, $\check{\theta}_t - \theta_{t,*}$ is a zero-mean multivariate random variable independent of $\bm{A}_t$ and thus $\mathbb{E}[\bm{A}_t^\top (\check{\theta}_t - \theta_{t, *})\mid H_t]=0$. Therefore, we only need to bound the first term in \eqref{standard_decomposition_app}. With slight abuse of notation, let $\bm{\cA}$ be the set of all possible indicator vectors of actions $a \in [K]$. Precisely, an action $a \in [K]$ is also represented by an indicator vector $\bm{a} \in \bm{\cA} \subset \{0, 1\}^K$ (in bold letter). Then we define the following events
\begin{align*}
  E_{t, \bm{a}}(\delta) & =
  \set{|\bm{a}^\top (\theta_{t,*} - \check{\theta}_t)|
  \leq \sqrt{2 \log(1 / \delta)} \normw{\bm{a}}{\check{\Sigma}_t}}\,, & \forall \delta \in (0, 1)\,, \, \forall \bm{a} \in \bm{\cA}\,.
\end{align*}
Fix history $H_t$, we split the expectation over the two complementary events $  E_{t, \bm{A}_{t,*}}(\delta)$ and $ \bar{E}_{t, \bm{A}_{t,*}}(\delta)$, and use the Cauchy-Schwarz inequality to obtain
\begin{align}\label{eq:classic_regret_decomposition_2}
  \condE{\bm{A}_{t,*}^\top  (\theta_{t,*} - \check{\theta}_t)}{H_t}
  &\leq \sqrt{2 \log(1 / \delta)} \, \condE{\normw{\bm{A}_{t,*}}{\check{\Sigma}_t}}{H_t} \\  &\hspace{4cm}+
  \condE{\bm{A}_{t,*}^\top (\theta_{t,*} - \check{\theta}_t)
  \I{\bar{E}_{t, \bm{A}_{t,*}}(\delta)}}{H_t}\,.\nonumber
\end{align}
Now the second term in \cref{eq:classic_regret_decomposition_2} can be bounded as follows. For any $\bm{a} \in \bm{\cA}\,,$ let $Z_{\bm{a}} = \bm{a}^\top (\theta_{t,*} - \check{\theta}_t)$. Then we have that  
\begin{align}\label{eq:classic_regret_decomposition_3}
  \condE{\bm{A}_{t,*}^\top (\theta_{t,*} - \check{\theta}_t) \I{ \bar{E}_{t, \bm{A}_{t,*}}(\delta) } }{H_t} \hspace{-5cm} &  \nonumber\\
  &\stackrel{(i)}{=} \condE{Z_{\bm{A}_{t,*}} \I{  |Z_{\bm{A}_{t,*}}| > \sqrt{2 \log(1 / \delta)}\normw{\bm{A}_{t,*}}{\check{\Sigma}_t} } }{H_t}\,, \nonumber \\
& \stackrel{(ii)}{\leq} \condE{|Z_{\bm{A}_{t,*}}| \I{  |Z_{\bm{A}_{t,*}}| > \sqrt{2 \log(1 / \delta)}\normw{\bm{A}_{t,*}}{\check{\Sigma}_t} } }{H_t}\,, \nonumber \\
& \stackrel{(iii)}{\leq} \sum_{\bm{a} \in \bm{\cA}} \condE{|Z_{\bm{a}}| \I{  |Z_{\bm{a}}| > \sqrt{2 \log(1 / \delta)}\normw{\bm{a}}{\check{\Sigma}_t} } }{H_t}\,, \nonumber \\
  & \stackrel{(iv)}{\leq}  \sum_{\bm{a} \in \bm{\cA}}  \frac{2}{\normw{\bm{a}}{\check{\Sigma}_t}\sqrt{2 \pi}}
  \int_{u = \sqrt{2 \log(1 / \delta)} \normw{\bm{a}}{\check{\Sigma}_t}}^\infty
  u \exp\left[- \frac{u^2}{2\normw{\bm{a}}{\check{\Sigma}_t}^2}\right] \dif u\,, \nonumber \\
  & \stackrel{(v)}{\leq}  \sum_{\bm{a} \in \bm{\cA}} \normw{\bm{a}}{\check{\Sigma}_t} \frac{2}{\sqrt{2 \pi}}
  \int_{u = \sqrt{2 \log(1 / \delta)}}^\infty
  u \exp\left[- \frac{u^2}{2}\right] \dif u
   \stackrel{(vi)}{\leq} \sqrt{\frac{2}{\pi}} \lambda_{\max, t}   K \delta\,.
\end{align}
In $(i)$, we simply rewrite the terms using the random variable $Z_{\bm{A}_{t,*}}$. In $(ii)$, we use the fact that $Z_{\bm{A}_{t,*}} \leq |Z_{\bm{A}_{t,*}}|$. In $(iii)$, we upper bound the expectation of the random variable $|Z_{\bm{A}_{t,*}}| \I{  |Z_{\bm{A}_{t,*}}| > \sqrt{2 \log(1 / \delta)}\normw{\bm{A}_{t,*}}{\check{\Sigma}_t}}$ with the sum of the expectations of $|Z_{\bm{a}}| \I{  |Z_{\bm{a}}| > \sqrt{2 \log(1 / \delta)}\normw{\bm{a}}{\check{\Sigma}_t} }$ for $\bm{a} \in \bm{\cA}$ since all these random variables are non-negative. Moreover, $(iv)$ follows from the facts that given $H_t\,,$ $Z_{\bm{a}} \sim \cN(0, \normw{\bm{a}}{\check{\Sigma}_t}^2)$, and that if $Z \sim \cN(0, \sigma^2)$, then for any $\epsilon \geq 0\,,$ $\mathbb{P}(|Z|>\epsilon) \leq 2 \mathbb{P}(Z>\epsilon)$. In $(v)$, we use the change of variables $u \gets u/\normw{\bm{a}}{\check{\Sigma}_t} $. Finally, in $(vi)$, we compute the integral and set $\lambda_{\max, t} = \max_{\bm{a} \in \bm{\cA}} \normw{\bm{a}}{\check{\Sigma}_t}$. We combine \cref{eq:classic_regret_decomposition_2} and \cref{eq:classic_regret_decomposition_3} with the fact that $\bm{A}_t$ and $\bm{A}_{t,*}$ are i.i.d.\ given $H_t$ to obtain that
\begin{align}\label{eq:classic_regret_decomposition_4}
  \condE{\bm{A}_{t,*}^\top (\theta_{t,*} - \check{\theta}_t)}{H_t}
  \leq \sqrt{2 \log(1 / \delta)} \, \condE{\normw{\bm{A}_t}{\check{\Sigma}_t}}{H_t} +
  \sqrt{\frac{2}{\pi}} \lambda_{\max, t}   K \delta\,.
\end{align}
The bound in \cref{eq:classic_regret_decomposition_4} holds for any history $H_t$ and thus we take an additional expectation and get that
\begin{align*}
 \mathcal{B}\mathcal{R}(T) = \E{}{\sum_{t = 1}^T \bm{A}_{t,*}^\top \theta_{t,*} - \bm{A}_t^\top \theta_{t,*}}
  & \leq \sqrt{2 \log(1 / \delta)} \,
  \E{}{\sum_{t = 1}^T \normw{\bm{A}_t}{\check{\Sigma}_t}} +
  \sqrt{\frac{2}{\pi}} K \delta \sum_{t=1}^T \lambda_{\max, t}   \,, \\
  & \stackrel{(i)}{\leq} \sqrt{2 T\log(1 / \delta)} \,
  \E{}{\sqrt{\sum_{t = 1}^T \normw{\bm{A}_t}{\check{\Sigma}_t}^2}} +
  \sqrt{\frac{2}{\pi}} K \delta \sum_{t=1}^T \lambda_{\max, t}   \,, \\
  & \stackrel{(ii)}{\leq} \sqrt{2 T \log(1 / \delta)}
  \sqrt{\E{}{\sum_{t = 1}^T \normw{\bm{A}_t}{\check{\Sigma}_t}^2}} +
  \sqrt{\frac{2}{\pi}} K \delta \sum_{t=1}^T \lambda_{\max, t}  \,,
\end{align*}
where we use the Cauchy-Schwarz inequality in $(i)$, and $(ii)$ follows from the concavity of the square root. Now note that any $\bm{a} \in \bm{\cA}$ is an indicator vector and that $\check{\Sigma}_t$ is the covariance of the joint posterior of the expected rewards $(X_t^\top \theta_{*, a})_{a \in [K] }\mid H_t$. Therefore, for  any $\bm{a} \in \bm{\cA}$, $\normw{\bm{a}}{\check{\Sigma}_t}^2 = \check{\sigma}_a^2$ is the variance of $X_t^\top \theta_{*, a} \mid H_t$. But we know that $\theta_{*, a} \mid H_t$ is a multivariate Gaussian and its covariance is $\hat{\Sigma}_{t,a}$ (\cref{lemma:gaussian_covariance}). Thus the variance of $X^\top \theta_{*, a} \mid H_t$ is $\check{\sigma}_a^2 = X_t^\top \hat{\Sigma}_{t,a}X_t$. It follows that for any 
$\bm{a} \in \bm{\cA}\,,$ $\normw{\bm{a}}{\check{\Sigma}_t}^2= X_t^\top \hat{\Sigma}_{t,a}X_t = \normw{X_t}{\hat{\Sigma}_{t, a}}^2$. In particular,  $\normw{\bm{A}_t}{\check{\Sigma}_t}^2= X_t^\top \hat{\Sigma}_{t,A_t}X_t$. Combining this with \cref{eq:max_eigenvalue_covariance} yields that $\lambda_{\max, t}  = \max_{\bm{a} \in \bm{\cA}} \normw{\bm{a}}{\check{\Sigma}_t} = \max_{a \in \cA} \normw{X_t}{\hat{\Sigma}_{t, a}} \leq  \max_{a \in \cA}\sqrt{\lambda_1(\hat{\Sigma}_{t,a}) \kappa_x} \leq \sqrt{\left( \lambda_{1, 0} + \frac{\lambda_{1, 0}^2 \lambda_{1, \Psi}\kappa_b }{\lambda_{d, 0}^2}\right) \kappa_x}$. Then we let $c = \sqrt{ \frac{2}{\pi}\left( \lambda_{1, 0} + \frac{\lambda_{1, 0}^2 \lambda_{1, \Psi}\kappa_b }{\lambda_{d, 0}^2}\right) \kappa_x} K$ which allows us to write
\begin{align}
   \mkern-22mu \mathcal{B}\mathcal{R}(T) &\leq  \sqrt{2 T \log(1/\delta)}
  \sqrt{\E{}{\sum_{t = 1}^T \normw{X_t}{\hat{\Sigma}_{t,A_t}}^2}} +
  c T\delta\,.
\end{align}
Now we focus on the the term $\sqrt{\E{}{\sum_{t = 1}^T \normw{X_t}{\hat{\Sigma}_{t, A_t}}^2}}$ that we decompose and bound as
\begin{align}\label{eq:sequential proof decomposition}
  & \normw{X_t}{\hat{\Sigma}_{t, A_t}}^2 = \sigma^2 \frac{X_t^\top \hat{\Sigma}_{t, A_t} X_t}{\sigma^2} \stackrel{(i)}{=} \sigma^2 \left(\sigma^{-2} X_t^\top \tilde{\Sigma}_{t, A_t} X_t +
  \sigma^{-2} X_t^\top \tilde{\Sigma}_{t, A_t} \Sigma_{0,A_t}^{-1} \Gamma_{A_t} \bar{\Sigma}_t \Gamma_{A_t}^\top
  \Sigma_{0,A_t}^{-1} \tilde{\Sigma}_{t, A_t} X_t\right)\,,
  \nonumber \\
  & \stackrel{(ii)}{\leq} c_{\textsc{a}} \log(1 + \sigma^{-2} X_t^\top \tilde{\Sigma}_{t, A_t} X_t) +
  c_1 \log(1 + \sigma^{-2} X_t^\top \tilde{\Sigma}_{t, A_t} \Sigma_{0,A_t}^{-1} \Gamma_{A_t} \bar{\Sigma}_t \Gamma_{A_t}^\top
  \Sigma_{0,A_t}^{-1} \tilde{\Sigma}_{t, A_t} X_t)\,,
\end{align}
where $(i)$ follows from $\hat{\Sigma}_{t, A_t} =  \tilde{\Sigma}_{t, A_t}  + \tilde{\Sigma}_{t, A_t} \Sigma_{0,A_t}^{-1} \Gamma_{A_t} \bar{\Sigma}_t \Gamma_{A_t}^\top \Sigma_{0,A_t}^{-1}  \tilde{\Sigma}_{t, A_t}$, and we use the following inequality in $(ii)$
\begin{align*}
  x
  = \frac{x}{\log(1 + x)} \log(1 + x)
  \leq \left(\max_{x \in [0, u]} \frac{x}{\log(1 + x)}\right) \log(1 + x)
  = \frac{u}{\log(1 + u)} \log(1 + x)\,,
\end{align*}
which holds for any $x \in [0, u]$, where constants $c_{\textsc{a}}$ and $c_1$ are derived as
\begin{align*}
  c_{\textsc{a}}
  = \frac{\kappa_x \lambda_{1, 0}}{\log(1 +  \sigma^{-2} \kappa_x \lambda_{1, 0})}\,, \quad
  c_1
  = \frac{c_\Psi}{\log(1 + \sigma^{-2} c_\Psi)}\,, \quad
  c_\Psi
  = \frac{\kappa_x \kappa_b \lambda_{1, 0}^2 \lambda_{1, \Psi}}{\lambda_{d, 0}^2}\,,
\end{align*}
The derivation of $c_{\textsc{a}}$ uses that
\begin{align*}
  X_t^\top \tilde{\Sigma}_{t, A_t} X_t
  \leq \lambda_1(\tilde{\Sigma}_{t, A_t}) \norm{X_t}^2
  \leq  \lambda_d^{-1}(\Sigma_{0,A_t}^{-1} + G_{t, A_t}) \kappa_x
  \leq \lambda_d^{-1}(\Sigma_{0,A_t}^{-1}) \kappa_x
  = \lambda_1(\Sigma_{0,A_t}) \kappa_x \leq  \lambda_{1, 0} \kappa_x \,.
\end{align*}
The derivation of $c_1$ follows from
\begin{align*}
  X_t^\top \tilde{\Sigma}_{t, A_t} \Sigma_{0,A_t}^{-1} \Gamma_{A_t} \bar{\Sigma}_t \Gamma_{A_t}^\top \Sigma_{0,A_t}^{-1} \tilde{\Sigma}_{t, A_t} X_t &\leq \lambda_1^2(\tilde{\Sigma}_{t, A_t}) \lambda_1^2(\Sigma_{0,A_t}^{-1}) \lambda_1(\Gamma_{A_t} \bar{\Sigma}_t \Gamma_{A_t}^\top) \kappa_x\,,\\
  & \leq \frac{\lambda_1^2(\Sigma_{0,A_t}) \lambda_{1, \Psi}\lambda_1(\Gamma_{A_t} \Gamma_{A_t}^\top)  \kappa_x}{\lambda_d^2(\Sigma_{0,A_t})}\,, \\
  &\leq \frac{\lambda_{1, 0}^2 \lambda_{1, \Psi}\kappa_b \kappa_x}{\lambda_{d,0}^2}\,.
\end{align*}
The first inequality follows from Weyl's inequality and the fact that $\lambda_1(\bar{\Sigma}_t) \leq \lambda_1(\Sigma_\Psi) = \lambda_{1, \Psi}  $ and $\lambda_1(\tilde{\Sigma}_{t, A_t}) \leq \lambda_1(\Sigma_{0, A_t})$. Now we focus on bounding the logarithmic terms in \cref{eq:sequential proof decomposition}.

\textbf{First Term in \cref{eq:sequential proof decomposition}} We first rewrite this term as
\begin{align*}
   \log(1 + \sigma^{-2} X_t^\top \tilde{\Sigma}_{t, A_t} X_t) &\stackrel{(i)}{=} \log\det(I_d + \sigma^{-2}\tilde{\Sigma}_{t, A_t}^\frac{1}{2} X_t X_t^\top \tilde{\Sigma}_{t, A_t}^\frac{1}{2})\,,\\
  &= \log\det(\tilde{\Sigma}_{t, A_t}^{-1} + \sigma^{-2} X_t X_t^\top) - \log\det(\tilde{\Sigma}_{t, A_t}^{-1})\,,\\
  &= \log\det(\tilde{\Sigma}_{t+1, A_t}^{-1}) - \log\det(\tilde{\Sigma}_{t, A_t}^{-1})\,,
\end{align*}
where $(i)$ follows from the Weinstein–Aronszajn identity. Now note that for any $a\neq A_t$, the arm-$a$ precision does not update at round $t$, hence
$\tilde{\Sigma}_{t+1,a}^{-1}=\tilde{\Sigma}_{t,a}^{-1}$ and the increment is zero;
therefore we may sum over all $a\in[K]$ without changing the value. Then, we sum over all rounds $ t \in [T]$, and get a telescoping that leads to
\begin{align*}
  \sum_{t = 1}^{T}
   &\log\det(I_d +\sigma^{-2} \tilde{\Sigma}_{t, A_t}^\frac{1}{2} X_t
  X_t^\top \tilde{\Sigma}_{t, A_t}^\frac{1}{2})=\sum_{t = 1}^{T} \log\det(\tilde{\Sigma}_{t+1, A_t}^{-1}) - \log\det(\tilde{\Sigma}_{t, A_t}^{-1})\,,\\
  &=\sum_{t = 1}^{T} \sum_{a=1}^K \log\det(\tilde{\Sigma}_{t+1, a}^{-1}) - \log\det(\tilde{\Sigma}_{t, a}^{-1})=\sum_{a=1}^K \sum_{t = 1}^{T} \log\det(\tilde{\Sigma}_{t+1, a}^{-1}) - \log\det(\tilde{\Sigma}_{t, a}^{-1})\,,\\
  &= \sum_{a=1}^K \log\det(\tilde{\Sigma}_{T+1, a}^{-1}) - \log\det(\tilde{\Sigma}_{1, a}^{-1})\,,\\
  &\stackrel{(i)}{=} \sum_{a=1}^K \log\det(\Sigma_{0, a}^\frac{1}{2} \tilde{\Sigma}_{T+1, a}^{-1} \Sigma_{0, a}^\frac{1}{2}) \stackrel{(ii)}{\leq} \sum_{a=1}^K d \log\left(\frac{1}{d} \operatorname{Tr}(\Sigma_{0, a}^\frac{1}{2} \tilde{\Sigma}_{T+1,a}^{-1}
  \Sigma_{0, a}^\frac{1}{2})\right)\\
  &\leq \sum_{a=1}^K d \log\left(1 + \frac{\kappa_x \lambda_1(\Sigma_{0, a})  T}{\sigma^2 d}\right)\leq K d \log\left(1 + \frac{\kappa_x \lambda_{1, 0} T}{\sigma^2 d}\right)\,.
\end{align*}
where $(i)$ follows from the fact that $\tilde{\Sigma}_{1, a} = \Sigma_{0, a}$ and we use the inequality of arithmetic and geometric means in $(ii)$.

\textbf{Second Term in \cref{eq:sequential proof decomposition}}
First, we rewrite the covariance matrix of the effect posterior $\bar{\Sigma}_t$ using the compact notation introduced in \cref{subsec:regre_proof_1}. Precisely, it follows from \cref{eq:effect_posterior_formulas} that
\begin{align}\label{eq:hyper_posterior_cov_rewritten}
    \bar{\Sigma}_t^{-1} &\stackrel{(i)}{=}  \Sigma_\Psi^{-1} + \sum_{a=1}^K\Gamma_a^{\top} \left( \Sigma_{0, a}^{-1} - \Sigma_{0, a}^{-1} (G_{t, a} + \Sigma_{0, a}^{-1})^{-1} \Sigma_{0, a}^{-1} \right)\Gamma_a \,,\nonumber\\
    &\stackrel{(ii)}{=}  \Sigma_\Psi^{-1} + \sum_{a=1}^K\Gamma_a^{\top} \left( \Sigma_{0, a}^{-1} - \Sigma_{0, a}^{-1} \tilde{\Sigma}_{t, a} \Sigma_{0, a}^{-1} \right)\Gamma_a \,.
\end{align}
Recall that $(i)$ and $(ii)$ are the same; $(ii)$ follows from plugging $\tilde{\Sigma}_{t, a} = (G_{t, a} + \Sigma_{0, a}^{-1})^{-1}$ in $(i)$. Now let $u = \sigma^{-1} \tilde{\Sigma}_{t, A_t}^{\frac{1}{2}} X_t$. Then it follows from $(ii)$ in \cref{eq:hyper_posterior_cov_rewritten} that
\begin{align}
  \bar{\Sigma}_{t + 1}^{-1} - \bar{\Sigma}_t^{-1} & = \Gamma_{A_t}^\top \left(\Sigma_{0,A_t}^{-1} - \Sigma_{0,A_t}^{-1} (\tilde{\Sigma}_{t, A_t}^{-1} + \sigma^{-2} X_t X_t^\top)^{-1} \Sigma_{0,A_t}^{-1} -
  (\Sigma_{0,A_t}^{-1} - \Sigma_{0,A_t}^{-1} \tilde{\Sigma}_{t, A_t} \Sigma_{0,A_t}^{-1})\right)\Gamma_{A_t}\,,
  \nonumber \\
  & = \Gamma_{A_t}^\top \left(\Sigma_{0,A_t}^{-1} (\tilde{\Sigma}_{t, A_t} - (\tilde{\Sigma}_{t, A_t}^{-1} + \sigma^{-2} X_t X_t^\top)^{-1}) \Sigma_{0,A_t}^{-1}\right)\Gamma_{A_t}\,,
  \nonumber \\
  & = \Gamma_{A_t}^\top \left(\Sigma_{0,A_t}^{-1} \tilde{\Sigma}_{t, A_t}^{\frac{1}{2}}
  (I_d - (I_d + \sigma^{-2} \tilde{\Sigma}_{t, A_t}^{\frac{1}{2}} X_t X_t^\top \tilde{\Sigma}_{t, A_t}^{\frac{1}{2}})^{-1})
  \tilde{\Sigma}_{t, A_t}^{\frac{1}{2}} \Sigma_{0,A_t}^{-1}\right)\Gamma_{A_t}\,,
  \nonumber \\
  & = \Gamma_{A_t}^\top \left(\Sigma_{0,A_t}^{-1} \tilde{\Sigma}_{t, A_t}^{\frac{1}{2}}
  (I_d - (I_d + u u^\top)^{-1})
  \tilde{\Sigma}_{t, A_t}^{\frac{1}{2}} \Sigma_{0,A_t}^{-1}\right)\Gamma_{A_t}\,,
  \nonumber \\
  & \stackrel{(i)}{=} \Gamma_{A_t}^\top \left(\Sigma_{0,A_t}^{-1} \tilde{\Sigma}_{t, A_t}^{\frac{1}{2}}
  \frac{u u^\top}{1 + u^\top u}
  \tilde{\Sigma}_{t, A_t}^{\frac{1}{2}} \Sigma_{0,A_t}^{-1}\right)\Gamma_{A_t}\,,\nonumber\\
  &= \sigma^{-2} \Gamma_{A_t}^\top \left( \Sigma_{0,A_t}^{-1} \tilde{\Sigma}_{t, A_t}
  \frac{X_t X_t^\top}{1 + u^\top u}
  \tilde{\Sigma}_{t, A_t} \Sigma_{0,A_t}^{-1}\right)\Gamma_{A_t}\,.
  \label{eq:linear telescoping}
\end{align}
In $(i)$ we use the Sherman-Morrison formula. Moreover, we have that $\norm{X_t}^2 \leq \kappa_x$. Therefore,
\begin{align*}
  1 + u^\top u
  = 1 + \sigma^{-2} X_t^\top \tilde{\Sigma}_{t, A_t} X_t
  \leq 1 + \sigma^{-2} \kappa_x \lambda_1(\Sigma_{0,A_t}) \leq  1 + \sigma^{-2} \kappa_x \lambda_{1, 0} = c_2\,.
\end{align*}
This allows us to bound the second logarithmic term in \cref{eq:sequential proof decomposition} as
\begin{align*}
&\log(1 + \sigma^{-2} X_t^\top \tilde{\Sigma}_{t, A_t} \Sigma_{0,A_t}^{-1} \Gamma_{A_t} \bar{\Sigma}_t \Gamma_{A_t}^\top
  \Sigma_{0,A_t}^{-1} \tilde{\Sigma}_{t, A_t} X_t) \,,\\   & \quad \stackrel{(i)}{\leq} c_2\log(1 + c_2^{-1} \sigma^{-2}  X_t^\top \tilde{\Sigma}_{t, A_t} \Sigma_{0,A_t}^{-1} \Gamma_{A_t} \bar{\Sigma}_t \Gamma_{A_t}^\top
  \Sigma_{0,A_t}^{-1} \tilde{\Sigma}_{t, A_t} X_t)\,, \\
  & \quad \stackrel{(ii)}{=} c_2 \log\det(I_{Ld} +
  c_2^{-1}\sigma^{-2} \bar{\Sigma}^\frac{1}{2}_t \Gamma_{A_t}^\top \Sigma_{0,A_t}^{-1} \tilde{\Sigma}_{t, A_t} X_t X_t^\top
  \tilde{\Sigma}_{t, A_t} \Sigma_{0,A_t}^{-1} \Gamma_{A_t} \bar{\Sigma}^\frac{1}{2}_t ) \,, \\
  & \quad \stackrel{(iii)}{=} c_2 \left[\log\det(\bar{\Sigma}_t^{-1} +
  c_2^{-1}\sigma^{-2} \Gamma_{A_t}^\top \Sigma_{0,A_t}^{-1} \tilde{\Sigma}_{t, A_t} X_t X_t^\top \tilde{\Sigma}_{t, A_t} \Sigma_{0,A_t}^{-1} \Gamma_{A_t}) -
  \log\det(\bar{\Sigma}_t^{-1})\right] \,, \\
  & \quad \stackrel{(iv)}{\leq} c_2 \left[\log\det(\bar{\Sigma}_t^{-1} +
  \sigma^{-2} \Gamma_{A_t}^\top \Sigma_{0,A_t}^{-1} \tilde{\Sigma}_{t, A_t} \frac{X_t X_t^\top}{1+u^\top u}  \tilde{\Sigma}_{t, A_t} \Sigma_{0,A_t}^{-1} \Gamma_{A_t} ) -
  \log\det(\bar{\Sigma}_t^{-1})\right] \,, \\
  & \quad \stackrel{(v)}{=} c_2 \left[\log\det(\bar{\Sigma}_{t + 1}^{-1}) -
  \log\det(\bar{\Sigma}_t^{-1})\right]\,.
\end{align*}
Here $(i)$ follows from the fact that $\log(1 + x) \leq c_2 \log(1 + c_2^{-1}x)$ for any $x \geq 0$ and $c_2 \geq 1$. In $(ii)$, we use the Weinstein–Aronszajn identity. In $(iii)$, we use the $\log$ product formula and the fact that the $\det$ is a multiplicative map. In $(iv)$, we use that $ c_2^{-1} \leq 1 / (1 + u^\top u)$. Finally, $(v)$ follows from \cref{eq:linear telescoping}. Now we sum over all rounds and get telescoping
\begin{align*}
   \sum_{t = 1}^{T} 
  &\log(1 + \sigma^{-2} X_t^\top \tilde{\Sigma}_{t, A_t} \Sigma_{0,A_t}^{-1} \Gamma_{A_t} \bar{\Sigma}_t \Gamma_{A_t}^\top
  \Sigma_{0,A_t}^{-1} \tilde{\Sigma}_{t, A_t} X_t)\,,\\
  &\leq c_2 \left[\log\det(\bar{\Sigma}_{T+1}^{-1}) -
  \log\det(\bar{\Sigma}_1^{-1})\right] = c_2 \log\det(\Sigma_\Psi^\frac{1}{2} \bar{\Sigma}_{T + 1}^{-1} \Sigma_\Psi^\frac{1}{2}) \,,\\ 
  &\stackrel{(i)}{\leq} c_2 L d \log\left(\frac{1}{Ld} \operatorname{Tr}(\Sigma_\Psi^\frac{1}{2} \bar{\Sigma}_{T+1}^{-1}
  \Sigma_\Psi^\frac{1}{2})\right) \,,\\
  &  \stackrel{(ii)}{\leq} c_2 L d
  \log(\lambda_1(\Sigma_\Psi^\frac{1}{2} \bar{\Sigma}_{T+1}^{-1} \Sigma_\Psi^\frac{1}{2}))   \stackrel{(iii)}{\leq} c_2 L d \log\Big(1+K \kappa_b \lambda_{1, \Psi} \big( \frac{1}{\lambda_{d, 0}} - \frac{1}{\lambda_{1, 0}^2\big(\frac{\kappa_x T}{\sigma^2} + \frac{1}{\lambda_{d, 0}} \big)} \big)\Big)\,,\\
\end{align*}
In $(i)$ we use the inequality of arithmetic and geometric means. In $(ii)$ we bound all eigenvalues in the trace by the maximum eigenvalue. In $(iii)$ we use the result in \cref{eq:max_eigenvalue_effect_covariance}. We combine the upper bounds for both logarithmic terms and get
\begin{align*}
  \E{}{\sum_{t = 1}^T \normw{X_t}{\hat{\Sigma}_{t, A_t}}^2}
  \leq K d c_{\textsc{a}} \log\big(1 + \frac{\kappa_x \lambda_{1, 0}  T}{\sigma^2 d}\big) &\\ & \hspace{-3cm}+
  Ld c_1 c_2 \log\Big(1+K \kappa_b \lambda_{1, \Psi} \big( \frac{1}{\lambda_{d, 0}} - \frac{1}{\lambda_{1, 0}^2\big(\frac{\kappa_x T}{\sigma^2} + \frac{1}{\lambda_{d, 0}} \big)} \big)\Big)\,.
\end{align*}
Finally, we set $c_{\textsc{e}} = c_1 c_2$,  which concludes the proof for the general case. To retrieve the result in \cref{thm:regret}, we only need to set $\lambda_{1, 0} = \lambda_{d, 0} = \sigma_0^2$ and $\lambda_{1, \Psi} = \sigma_\Psi^2$ since we assumed that $\Sigma_\Psi = \sigma_\Psi^2 I_{Ld}$ and that $\Sigma_{0, a} = \sigma_0^2 I_d$ for any $a \in [K]$. In that case, the second term simplifies as
\begin{align*}
     \log\Big(1+K \kappa_b \lambda_{1, \Psi} \big( \frac{1}{\lambda_{d, 0}} - \frac{1}{\lambda_{1, 0}^2\big(\frac{\kappa_x T}{\sigma^2} + \frac{1}{\lambda_{d, 0}} \big)} \big)\Big)& =  \log\Big(1+K \kappa_b \sigma_\Psi^2 \big( \frac{1}{\sigma_{0}^2} - \frac{1}{\sigma_{0}^4\big(\frac{\kappa_x T}{\sigma^2} + \frac{1}{\sigma_{0}^2} \big)} \big)\Big)\,,\\
     & =  \log\big(1+K \kappa_b \sigma_\Psi^2 \frac{T\kappa_x}{ T \kappa_x \sigma_{0}^2  + \sigma^2} \big)\,.
\end{align*}
\end{proof}

\section{Additional Experiments}
\label{app:experiments}

We provide additional experiments where we evaluate \meTS using synthetic and real-world problems, and compare it to baselines that either ignore or partially use effect parameters. In each plot, we report the averages and standard errors of the quantities. Both settings are described in \cref{sec:mets-experiments}. 



\subsection{Synthetic Experiments}\label{app:synthetic}

\begin{figure}[H]
  \centering
  \includegraphics[width=0.9\linewidth]{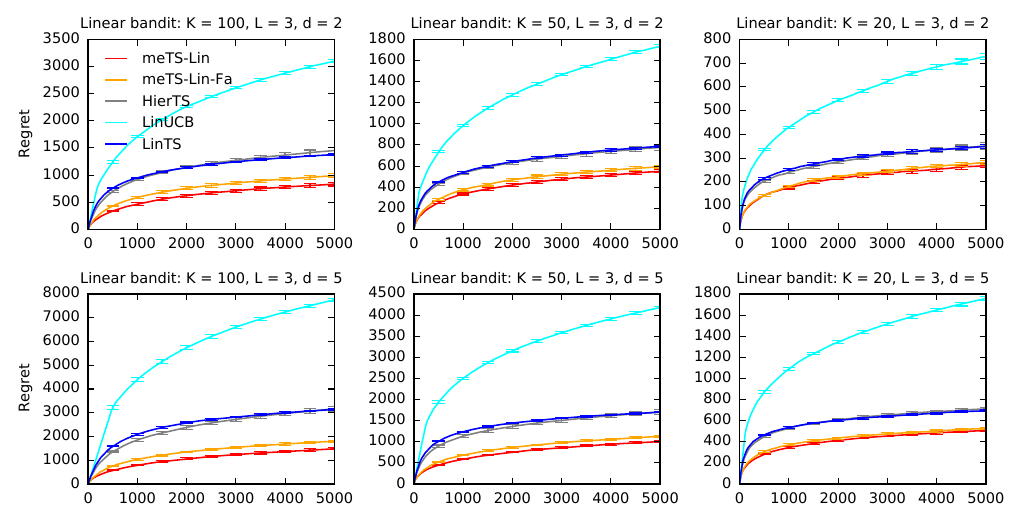}
  \caption{Regret of \alglin on synthetic linear bandit problems with varying feature dimension $d \in \{2, 5\}$ and number of actions $K\in \{20, 50, 100\}$.} 
  \label{fig:app_synthetic_regret_lin}
\end{figure}

\begin{figure}[H]
  \centering
  \includegraphics[width=0.9\linewidth]{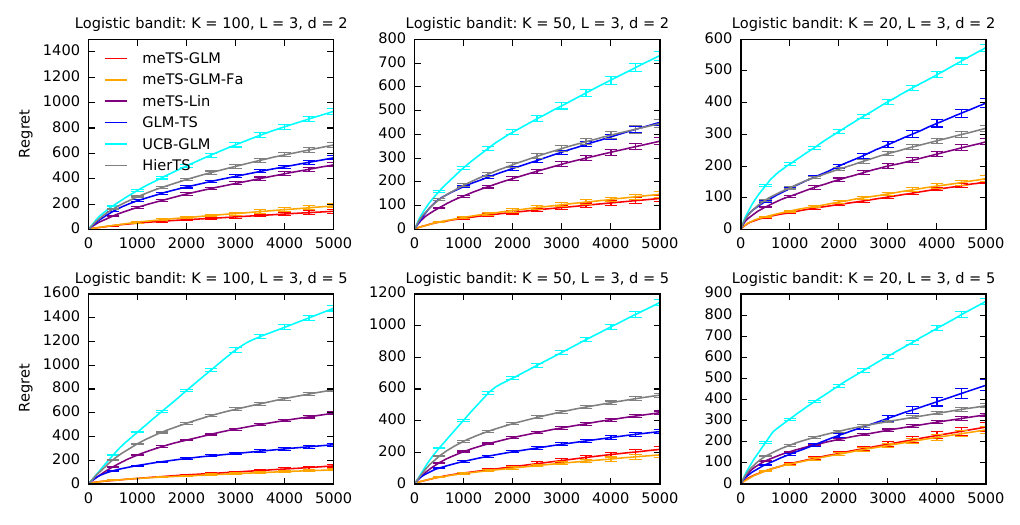}
  \caption{Regret of \algglm on synthetic logistic bandit problems with varying feature dimension $d \in \{2, 5\}$ and number of actions $K\in \{20, 50, 100\}$.} 
  \label{fig:app_synthetic_regret_log}
\end{figure}

In \cref{fig:app_synthetic_regret_lin,fig:app_synthetic_regret_log}, we report regret from 12 experiments with horizon $T=5000$, where we vary $K$ and $d$ and use both linear and logistic rewards. For the linear setting, we compare \alglin (\cref{sec:mets-linear bandit posterior}), \linucb \citep{abbasi2011improved}, \lints \citep{agrawal13thompson} and \hierts \citep{hong22hierarchical}. For the logistic setting, we compare \algglm (\cref{sec:mets-glb bandit posterior}), \alglin (\cref{sec:mets-linear bandit posterior}), \ucbglm \citep{li17provably}, \glmts \citep{chapelle11empirical} and \hierts \citep{hong22hierarchical}. We also include the factored approximation of \meTS (\alglinfa and \algglmfa). In all experiments, we observe that \alglin and \algfa outperform other baselines that ignore the effect parameters or incorporate them partially. We also notice that the gain in performance becomes smaller when $K/L$ decreases.

\subsection{MovieLens Experiments}\label{app:movielens}

\begin{figure}[H]
  \centering
  \includegraphics[width=0.9\linewidth]{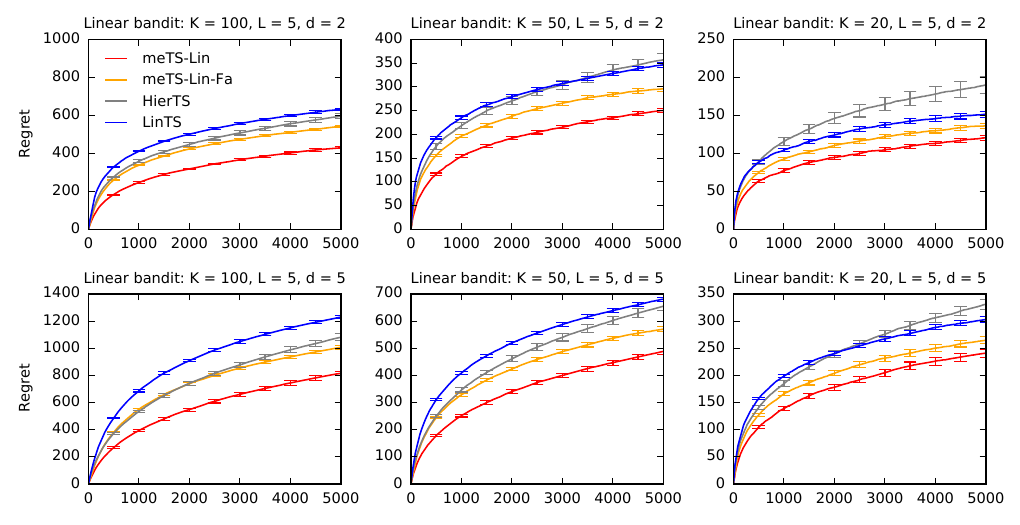}
  \caption{Regret of \alglin on the MovieLens dataset with linear rewards and varying feature dimension $d \in \{2, 5\}$ and number of actions $K\in \{20, 50, 100\}$.} 
  \label{fig:app_movielens_regret_lin}
\end{figure}

\begin{figure}[H]
  \centering
  \includegraphics[width=0.9\linewidth]{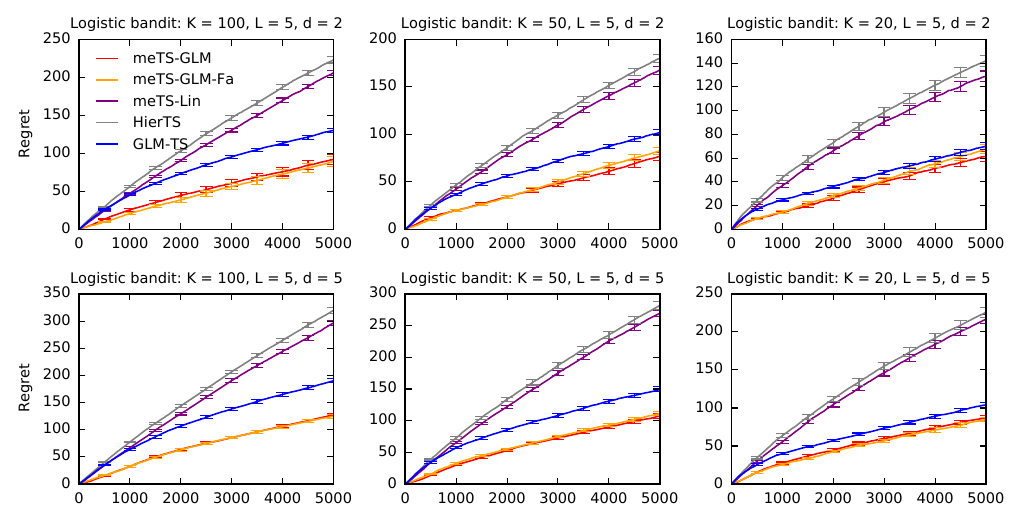}
  \caption{Regret of \algglm on the MovieLens dataset with logistic rewards and varying feature dimension $d \in \{2, 5\}$ and number of actions $K\in \{20, 50, 100\}$.} 
  \label{fig:app_movielens_regret_log}
\end{figure}

We plot the regret of \meTS and the baselines up to $T=5000$ rounds in \cref{fig:app_movielens_regret_lin,fig:app_movielens_regret_log}. We observe that \meTS outperforms the other baselines. This is despite the fact that we did not fine-tune the mixing weights, which attests to the robustness of our approach to model misspecification. Similarly to the synthetic problems, we observe that the gap in performance between \meTS and other baselines is less significant when $K/L$ is small.

\subsection{Robustness to Model Misspecification}\label{app:robustness_model_misspecification}

\begin{figure}[H]
  \centering
  \includegraphics[width=0.9\linewidth]{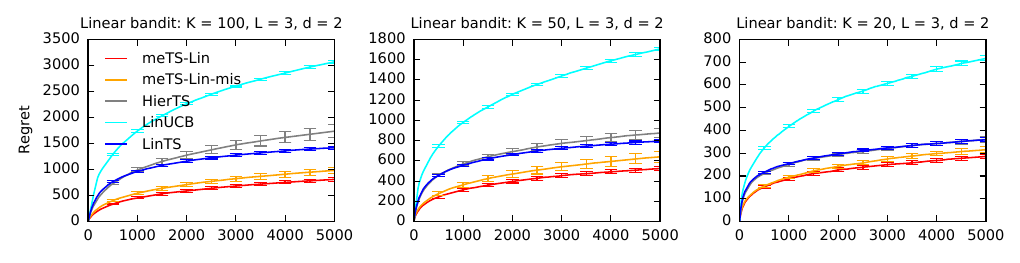}
  \caption{Regret of \emph{misspecified} \alglin on synthetic bandit problems with a varying number of actions $K$. Here, the \emph{misspecified} \meTS, \texttt{meTS-Lin-mis}, is compared to baselines with true hyper-parameters.} 
  \label{fig:misspecified_synthetic_regret}
\end{figure}

We conduct additional synthetic experiments where the hyper-parameters do not match the parameters of the bandit environment to assess the robustness of our approach to misspecification. We provide results for this experiment in \cref{fig:misspecified_synthetic_regret}. Here we consider the setting described in \cref{sec:mets-synthetic experiments} except that the true hyper-parameters are misspecified as follows. At each run, we sample uniformly 4 misspecification constants $c_1, c_2, c_3,$ and $c_4$ from $(0, 2)$ and set the hyper-parameters of \alglin as $c_1 \Sigma_{\Psi}$,  $c_2 \mu_{\Psi}$, $c_3 \Sigma_{0, a},$ and $c_4 b_a$ for any $a \in [K]$; where $ \Sigma_{\Psi}$,  $ \mu_{\Psi}$, $ \Sigma_{0, a},$ and $b_a$ for $a \in [K]$ are the true hyper-parameters. Model misspecification is only applied to \alglin and we refer to it as \texttt{meTS-Lin-mis}. We compare it to \alglin and the other baselines, all with the true hyper-parameters. Although the baselines are not misspecified, \texttt{meTS-Lin-mis} still performs better. \texttt{meTS-Lin-mis} also performs similarly to \alglin (with true hyper-parameters).

\subsection{Effect of Action Uncertainty}\label{app:uncertain_action}
\begin{figure}[H]
  \centering
  \includegraphics[width=0.9\linewidth]{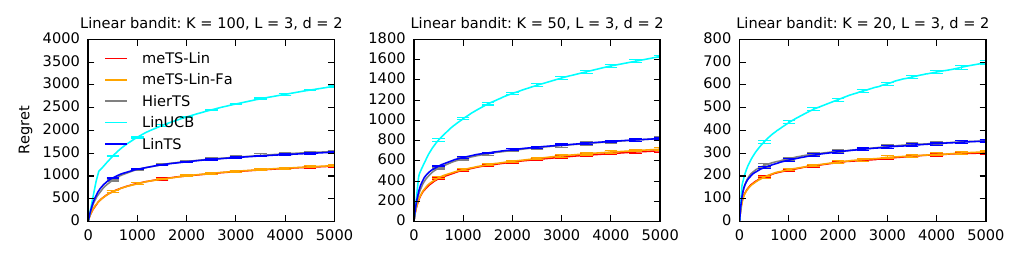}
  \caption{Regret of \alglin on synthetic bandit problems with a varying number of actions $K$, where the action parameters are more uncertain than the effect parameters.} 
  \label{fig:uncertain_synthetic_regret}
\end{figure}

As we mentioned in \cref{sec:mets-synthetic experiments} and predicted by our Bayes regret bound, learning the effect parameters is most beneficial when they are more uncertain than the action parameters. In this section, we support this claim by conducting an experiment where the initial uncertainty of action parameters is greater than the initial uncertainty of the effect parameters.  Precisely, we consider the setting described in \cref{sec:mets-synthetic experiments} except that we set $\Sigma_\Psi=I_{Ld}$ and $\Sigma_{0, a} = 3I_{d}$ for all $a \in [K]$. We report the results in \cref{fig:uncertain_synthetic_regret}. By comparing \cref{fig:uncertain_synthetic_regret} to \cref{fig:synthetic_regret}, we observe that \alglin still outperforms the baselines but the gap in performance shrinks when the action parameters are more uncertain than the effect parameters.

%% file: contents/dts/appendix.tex
\section{Posterior for Linear Diffusion Models}\label{sec:dts_posterior_proofs}
Our posterior approximation builds on the simplified setting where the diffusion model is fully linear, i.e., each link function $f_\ell$ is linear in $\psi_\ell$. This linear case, studied in our earlier workshop paper \citep{aouali2023linear}, serves as the analytical foundation for our posterior approximation used in the general non-linear case. In \cref{app:non_linear_explanation}, we show how the exact posteriors derived in this linear setting inspire our efficient approximation, which extends naturally to practical diffusion models that are typically highly non-linear.

\subsection{Linear Diffusion Models}\label{subsec:dts_dts_gaussian_posterior_derivation}
Here, we assume the link functions $f_\ell$ are linear such as $f_\ell(\psi_{ \ell}) = \W_\ell \psi_{ \ell}$ for $\ell \in [L]$, where $\W_{\ell} \in \real^{d \times d}$ are \emph{known mixing matrices}. Then, \cref{eq:dts_model} becomes a linear Gaussian system (LGS) \citep{Bishop2006} and can be summarized as follows 
\begin{align}\label{eq:dts_contextual_gaussian_model}
  \psi_{ L} & \sim \cN(0, \Sigma_{L+1})\,, \\
  \psi_{ \ell-1} \mid \psi_{ \ell} & \sim \cN(\W_\ell \psi_{ \ell}, \Sigma_\ell)\,, & \forall \ell \in [L]/\{1\} \,, \nonumber\\
  \theta_{ a} \mid \psi_{ 1} & \sim \cN(\W_1 \psi_{ 1}, \Sigma_1)\,, & \forall a \in [K] \,, \nonumber\\
  R_t \mid X_t, A_t, \theta, (\psi_\ell)_{\ell \in [L]} &\sim p(\cdot \mid X_t; \theta_{ A_t})\,, &  \forall t \in [T]\,. \nonumber
\end{align}
This model is important because it yields closed-form posteriors when the reward distribution is linear-Gaussian, i.e., $p(\cdot \mid x; \theta_{ a}) = \cN(\cdot; x^\top \theta_{ a}, \sigma^2)$. This allows bounding the Bayes regret of \alg. For practice, the posterior expressions are used to motivate efficient approximations for the general case in \cref{eq:dts_model} as we show in \cref{subsec:dts_dts_nonlin_posterior_derivation}.

\subsection{Posterior Expressions for Linear Diffusion Models}\label{subsec:dts_dts_posterior_expressions}

Recall that the reward distribution is modeled as a generalized linear model (GLM) \citep{mccullagh89generalized}, allowing for non-linear rewards even when the diffusion links are linear.  
This non-linearity in the reward distribution prevents closed-form posteriors.  
However, since the non-linearity arises only through the reward likelihood, we approximate it by a Gaussian, leading to efficient posterior updates that are exact whenever the reward model itself is Gaussian; a special case of the GLM framework. Precisely, let \(\hat{B}_{t,a}\) and \(\hat{G}_{t,a}\) denote the MLE (see the remark below for practical considerations) and the Hessian of the negative log-likelihood, respectively:
\begin{align}\label{eq:dts_glm_params}
    \hat{B}_{t,a} &= \argmax_{\theta_a \in \real^d} \sum_{i \in S_{t,a}}  \log p(R_i \mid X_i; \theta_a),
    &\qquad
    \hat{G}_{t,a} &= \sum_{i \in S_{t,a}} \dot{g}\!\big(X_i^\top \hat{B}_{t,a}\big) X_i X_i^\top,
\end{align}
where \(S_{t,a} = \{i \in [t-1] : A_i = a\}\) is the set of rounds in which action \(a\) was taken up to round \(t\).  
We approximate the likelihood as
\begin{align}\label{likelihood_app}
    \prod_{i \in S_{t,a}} p(R_i \mid X_i; \theta_a)
\;\propto\;
\exp\!\left(
-\tfrac12(\theta_a-\hat{B}_{t,a})^\top \hat{G}_{t,a}(\theta_a-\hat{B}_{t,a})
\right),
\end{align}
which makes all subsequent posteriors Gaussian. Once this approximation is done, all other derivations of the action posterior and latent posteriors are exact.

\begin{remark}
The MLE may be ill-posed. In practice, we maximize an $\ell_2$-regularized estimator.
\end{remark}

\textbf{Action posterior.}
The conditional action posterior becomes
$$
p(\theta_a \mid \psi_1, H_{t,a})
\;\approx\;
\cN\!\big(\theta_a; \hat{\mu}_{t,a}, \hat{\Sigma}_{t,a}\big),
$$
with parameters
\begin{align}    \label{eq:dts_linear action-posterior}
    \hat{\Sigma}_{t,a}^{-1} &= \Sigma_1^{-1} + \hat{G}_{t,a},
    &\qquad
    \hat{\mu}_{t,a} &= \hat{\Sigma}_{t,a}\!\left(
        \Sigma_1^{-1} \W_1 \psi_1 + \hat{G}_{t,a} \hat{B}_{t,a}
    \right).
\end{align}

\textbf{Latent posteriors.}
For each $\ell \in [L]\setminus\{1\}$, the conditional latent posterior is
$$
p(\psi_{\ell-1} \mid \psi_\ell, H_t)
\;\approx\;
\cN\!\big(\psi_{\ell-1}; \bar{\mu}_{t,\ell-1}, \bar{\Sigma}_{t,\ell-1}\big),
$$
where
\begin{align}\label{eq:dts_hyper_posteriors}
    \bar{\Sigma}_{t,\ell-1}^{-1} &= \Sigma_\ell^{-1} + \bar{G}_{t,\ell-1},
    &\qquad
    \bar{\mu}_{t,\ell-1} &= \bar{\Sigma}_{t,\ell-1}\!\left(
        \Sigma_\ell^{-1} \W_\ell \psi_\ell + \bar{B}_{t,\ell-1}
    \right).
\end{align}
The top-layer posterior is
$$
p(\psi_L \mid H_t)
\;\approx\;
\cN\!\big(\psi_L; \bar{\mu}_{t,L}, \bar{\Sigma}_{t,L}\big),
$$
with
\begin{align}\label{eq:dts_hyper_posterior_L}
    \bar{\Sigma}_{t,L}^{-1} &= \Sigma_{L+1}^{-1} + \bar{G}_{t,L},
    &\qquad
    \bar{\mu}_{t,L} &= \bar{\Sigma}_{t,L}\bar{B}_{t,L}.
\end{align}

\textbf{Recursive updates.}
The matrices $\bar{G}_{t,\ell}$ and $\bar{B}_{t,\ell}$ for $\ell \in [L]$ are defined recursively.  
The base recursion is
\begin{align}\label{eq:dts_basis_gaussian_recursive}
    \bar{G}_{t,1} &= 
    \W_1^\top
    \sum_{a=1}^K
    \big(\Sigma_1^{-1} - \Sigma_1^{-1}\hat{\Sigma}_{t,a}\Sigma_1^{-1}\big)
    \W_1,
    &\qquad
    \bar{B}_{t,1} &=
    \W_1^\top \Sigma_1^{-1}
    \sum_{a=1}^K
    \hat{\Sigma}_{t,a}\hat{G}_{t,a}\hat{B}_{t,a}.
\end{align}
Then, for $\ell \in [L]\setminus\{1\}$, the recursive step is
\begin{align}\label{eq:dts_step_gaussian_recursive}
    \bar{G}_{t,\ell} &=
    \W_\ell^\top\!\left(
        \Sigma_\ell^{-1} - \Sigma_\ell^{-1}\bar{\Sigma}_{t,\ell-1}\Sigma_\ell^{-1}
    \right) \W_\ell,
    &\qquad
    \bar{B}_{t,\ell} &=
    \W_\ell^\top \Sigma_\ell^{-1} \bar{\Sigma}_{t,\ell-1} \bar{B}_{t,\ell-1}.
\end{align}

\textbf{Discussion.}
This completes the derivation of the linear posterior approximation.  
All posteriors are Gaussian and exact whenever the reward distribution follows a linear-Gaussian model, i.e.
$$
p(\cdot \mid x; \theta_a)
= \cN(\cdot; x^\top \theta_a, \sigma^2).
$$
In this case, the above posterior updates coincide with the exact Bayesian updates, while for general GLMs they serve as efficient and accurate approximations.

\section{Posterior for Non-Linear Diffusion Models}\label{app:non_linear_explanation}

The general diffusion model (\cref{eq:dts_model}, which is our case of interest) involves two sources of non-linearity:  
(i)~the reward distribution \( p(\cdot \mid x; \theta) \), which may follow a non-linear generalized linear model (GLM), and  
(ii)~the diffusion links \( f_\ell(\psi_\ell) \), which can be arbitrary non-linear functions.  
Both sources make the posterior intractable, and therefore two approximations are needed.

\textbf{First approximation (likelihood).}  
We first approximate the reward likelihood by a Gaussian density (as we did above in \cref{likelihood_app}). After this substitution, the model becomes conditionally Gaussian given the latent variables. This step is exact when the reward model is linear-Gaussian, and approximate otherwise.

\textbf{Second approximation (diffusion hierarchy).}  
Even after the likelihood is approximated, the diffusion hierarchy remains non-linear because of the non-linear mappings \( f_\ell \).  
To handle this, we reuse the exact Gaussian posteriors derived for the linear diffusion case (\cref{subsec:dts_dts_posterior_expressions}) and generalize them as follows:
\begin{itemize}
    \item Replace each linear mapping \( \W_\ell \psi_\ell \) by its non-linear counterpart \( f_\ell(\psi_\ell) \), which represents the mean of the diffusion prior at layer~\(\ell\).
    \item Remove matrix multiplications involving \( \W_\ell \) in the recursive updates.
\end{itemize}
This step can be viewed as extending the linear-Gaussian posterior updates to a general non-linear setting. This allows fast sampling and updating of the posterior, without heavy standard posterior approximation techniques. Of course, this is a purely empirical and heuristic based approximation that does not come with guarantees but performs very well in practice.

\textbf{Resulting approximation.}  
The two steps above yield a posterior where each conditional factor
\( p(\theta_a \mid \psi_1, H_{t,a}) \) and \( p(\psi_{\ell-1} \mid \psi_\ell, H_t) \)
remains Gaussian with updated means and covariances, while the overall model
retains the hierarchical diffusion structure.  
The approximation satisfies two desirable properties:
it exactly recovers the diffusion prior when no data is available,
and as more data is observed, the likelihood terms dominate and the prior influence fades naturally.

\section{Connection to Two-Level Hierarchies}\label{subsec:dts_dts_two_level_hierarchies}
The linear diffusion \cref{eq:dts_contextual_gaussian_model} can be marginalized into a 2-level hierarchy using two different strategies. To simplify, we let $\Sigma_\ell = \sigma_\ell^2 I_d$. The first one yields,
\begin{align}\label{eq:dts_hier_as_two_lin_model_1}
    \psi_{ L} &\sim \cN(0, \sigma_{L+1}^2 \Beta_L \Beta_L^\top)\,,\\
   \theta_{ a} \mid \psi_{ L}  & \sim \cN( \psi_{ L} , \,  \Omega_1)\,, &\forall a \in [K]\,,\nonumber
\end{align}
with $\Omega_1 = \sigma_1^{2} I_d +  \sum_{\ell =1}^{L-1} \sigma_{\ell+1}^2  \Beta_\ell \Beta_\ell^\top$ and $\Beta_\ell = \prod_{i=1}^\ell\W_i$. The second strategy yields,
\begin{align}\label{eq:dts_hier_as_two_lin_model_2}
    \psi_{ 1} &\sim \cN(0, \Omega_2)\,,\\  \theta_{ a} \mid \psi_{ 1}  & \sim \cN( \psi_{ 1} , \,  \sigma_1^2 I_d)\,, & \forall a \in [K]\nonumber\,,
\end{align}
where $\Omega_2 =  \sum_{\ell=1}^L \sigma_{\ell+1}^2  \Beta_\ell \Beta_\ell^\top$. Recently, \hierts \citep{hong22hierarchical} was developed for such two-level graphical models, and we call \hierts under \cref{eq:dts_hier_as_two_lin_model_1} by \texttt{HierTS-1} and \hierts under \cref{eq:dts_hier_as_two_lin_model_2}  by \texttt{HierTS-2}. Then, we start by highlighting the differences between these two variants of \hierts. First, their regret bounds scale as
\begin{align*}
  &\texttt{HierTS-1}: \tilde{\mathcal{O}}\big(\sqrt{Td(K \sum_{\ell =1}^L\sigma_\ell^2 + L \sigma_{L+1}^2}\big)\,,
  &  \texttt{HierTS-2}: \tilde{\mathcal{O}}\big(\sqrt{T d (K \sigma_1^2 + \sum_{\ell=1}^L \sigma_{\ell+1}^2 )}\big)\,.
\end{align*}
When $K \approx L$, the regret bounds of \texttt{HierTS-1} and \texttt{HierTS-2} are similar. However, when $K > L$, \texttt{HierTS-2} outperforms \texttt{HierTS-1}. This is because \texttt{HierTS-2} puts more uncertainty on a single $d$-dimensional latent parameter $\psi_{ 1}$, rather than $K$ individual $d$-dimensional action parameters $\theta_{ a}$. More importantly, \texttt{HierTS-1} implicitly assumes that action parameters $\theta_{ a}$ are conditionally independent given $\psi_{ L}$, which is not true. Consequently, \texttt{HierTS-2} outperforms \texttt{HierTS-1}. Note that, under the linear diffusion model \cref{eq:dts_contextual_gaussian_model}, \alg and \texttt{HierTS-2} have roughly similar regret bounds. Specifically, their regret bounds dependency on $K$ is identical, where both methods involve multiplying $K$ by $\sigma_1^2$, and both enjoy improved performance compared to \texttt{HierTS-1}. That said, note that \cref{thm:dts_regret,prop:low_rank_regret} provide an understanding of how \alg's regret scales under linear link functions $f_\ell$, and do not say that using \alg is better than using \hierts when the link functions $f_\ell$ are linear since the latter can be obtained by a proper marginalization of latent parameters (i.e., \texttt{HierTS-2} instead of \texttt{HierTS-1}). While such a comparison is not the goal of this work, we still provide it for completeness next.

When the mixing matrices $\W_\ell$ are dense (i.e., assumption \textbf{(A5)} is not applicable), \alg and \texttt{HierTS-2} have comparable regret bounds and computational efficiency. However, under the sparsity assumption \textbf{(A5)} and with mixing matrices that allow for conditional independence of $\psi_{ 1}$ coordinates given $\psi_{ 2}$, \alg enjoys a computational advantage over \texttt{HierTS-2}. This advantage explains why works focusing on multi-level hierarchies typically benchmark their algorithms against two-level structures akin to \texttt{HierTS-1}, rather than the more competitive \texttt{HierTS-2}. This is also consistent with prior works in Bayesian bandits using multi-level hierarchies, such as Tree-based priors \citep{hong22deep}, which compared their method to \texttt{HierTS-1}. In line with this, we also compared \alg with \texttt{HierTS-1} in our experiments. But this is only given for completeness as this is not the aim of \cref{thm:dts_regret,prop:low_rank_regret}. More importantly, \hierts is inapplicable in the general case in \cref{eq:dts_model} with non-linear link functions since the latent parameters cannot be analytically marginalized.

\section{Formal Theory}
\label{sec:dts_analysis}

We analyze \alg assuming that: \textbf{(A0)} The true environment parameters $\theta_*$ and $\psi_{*, \ell}$ are drawn from the same prior distribution used by \alg, as is standard in Bayes regret analysis \citep{russo14learning}; we thus use $\theta_*$ and $\theta$ ($\psi_{*, \ell}$ and $\psi_{\ell}$) interchangeably throughout. \textbf{(A1)} The rewards are linear $p(\cdot \mid x; \theta_{ a}) = \cN(\cdot; x^\top \theta_{ a}, \sigma^2)$. \textbf{(A2)} The link functions $f_\ell$ are linear such as $f_\ell(\psi_{ \ell}) = \W_\ell \psi_{ \ell}$ for $\ell \in [L]$, where $\W_{\ell} \in \real^{d \times d}$ are \emph{known mixing matrices}. This leads to a structure with $L$ layers of linear Gaussian relationships detailed in \cref{subsec:dts_dts_gaussian_posterior_derivation}. In particular, this leads to closed-form posteriors given in \cref{subsec:dts_dts_posterior_expressions} that inspired our approximation and enable theory similar to linear bandits \citep{agrawal13thompson}. However, proofs are not the same, and technical challenges remain (explained in \cref{sec:dts_regret_proof}). 

Although our result holds for milder assumptions, we make additional simplifications for clarity and interpretability. We assume that \textbf{(A3)} Contexts satisfy $\normw{X_t}{2}^2 = 1$ for any $t \in [T]$. Note that \textbf{(A3)} can be relaxed to any contexts $X_t$ with bounded norms $\normw{X_t}{2}$. \textbf{(A4)} Mixing matrices and covariances satisfy $\lambda_1(\W_\ell^\top \W_\ell)= 1$ for any $\ell \in [L]$ and $\Sigma_\ell=\sigma_\ell^2I_d$ for any $\ell \in [L+1]$. In this section, we write $\tilde{\mathcal{O}}$ for the big-O notation up to polylogarithmic factors. We start by stating our bound for \alg.   
\begin{theorem}
\label{thm:dts_regret} Let $\sigma_{\textsc{max}}^2= \max_{\ell\in[L+1]} 1 + \frac{\sigma_\ell^2}{\sigma^2}$. There exists a constant $c>0$ such that for any $\delta \in (0, 1)$, the Bayes regret of \emph{\alg} under \emph{\textbf{(A1)}}, \emph{\textbf{(A2)}}, \emph{\textbf{(A3)}} and \emph{\textbf{(A4)}} is bounded as
\begin{align}\label{eq:dts_regret_terms}
  &\mathcal{B}\mathcal{R}(T)
  \leq 
  \sqrt{2 T \big( \mathcal{R}^{\textsc{act}}(T) + \sum_{\ell =1}^L \mathcal{R}_\ell^{\textsc{lat}} \big) \log(1 / \delta)\Big)} +
  c T\delta\,,\nonumber\\
 & \mathcal{R}^{\textsc{act}}(T) = c_0 dK  \log\big(1 + \frac{T\sigma_1^2}{d \sigma^2} \big), \ c_0 = \frac{ \sigma_1^2}{\log\left(1 + \frac{\sigma_1^2}{\sigma^2} \right)}\,, \nonumber\\ &\mathcal{R}_\ell^{\textsc{lat}} = c_\ell d \log\big(1 +  \frac{\sigma_{\ell+1}^2}{\sigma_{\ell}^2}\big), c_\ell= \frac{\sigma_{\ell+1}^2 \sigma_{\textsc{max}}^{2 \ell}}{\log\left(1 + \frac{\sigma_{\ell+1}^2}{\sigma^2} \right)}, 
\end{align}
\end{theorem}
\cref{eq:dts_regret_terms} holds for any $\delta \in (0, 1)$. In particular, the term $c T\delta$ is constant when $\delta=1/T$. Then, the bound is $\tilde{\mathcal{O}}\Big(\sqrt{T  (dK \sigma_1^2 + d \sum_{\ell=1}^L \sigma_{\ell+1}^2\sigma_{\textsc{max}}^{2 \ell} )}\Big)$, and this dependence on the horizon $T$ aligns with prior Bayes regret bounds. The bound comprises $L+1$ main terms, $\mathcal{R}^{\textsc{act}}(T)$ and $\mathcal{R}_\ell^{\textsc{lat}}$ for $\ell \in [L]$. First, $\mathcal{R}^{\textsc{act}}(T)$ relates to action parameters learning, conforming to a standard form \citep{lu19informationtheoretic}. Similarly, $\mathcal{R}_\ell^{\textsc{lat}}$ is associated with learning the $\ell$-th latent parameter. 

To include more structure, we propose the  \emph{sparsity} assumption \textbf{(A5)} $\W_\ell = (\bar{\W}_\ell, 0_{d, d-d_\ell})$, where $\bar{\W}_\ell \in \real^{d \times d_\ell}$ for any $\ell \in [L]$.  Note that \textbf{(A5)} is not an assumption when $d_\ell=d$ for any $\ell \in [L]$. Notably, \textbf{(A5)} incorporates a plausible structural characteristic that a diffusion model could capture. 
\begin{proposition}[Sparsity]
\label{prop:low_rank_regret}Let $\sigma_{\textsc{max}}^2= \max_{\ell\in[L+1]} 1 + \frac{\sigma_\ell^2}{\sigma^2}$. There exists a constant $c>0$ such that for any $\delta \in (0, 1)$, the Bayes regret of \emph{\alg} under \emph{\textbf{(A1)}}, \emph{\textbf{(A2)}}, \emph{\textbf{(A3)}}, \emph{\textbf{(A4)}} and \emph{\textbf{(A5)}} is bounded as
\begin{align}\label{eq:dts_low_rank_regret_terms}
  &\mathcal{B}\mathcal{R}(T)
  \leq 
  \sqrt{2 T \big( \mathcal{R}^{\textsc{act}}(T) + \sum_{\ell =1}^L \tilde{\mathcal{R}}_\ell^{\textsc{lat}} \big) \log(1 / \delta)\Big)} +
  c T\delta\,, \nonumber\\
 & \tilde{\mathcal{R}}^{\textsc{act}}(T) = c_0 dK  \log\big(1 + \frac{T\sigma_1^2}{d \sigma^2} \big), \ c_0 = \frac{ \sigma_1^2}{\log\left(1 + \frac{\sigma_1^2}{\sigma^2} \right)}\,, \nonumber\\ &\mathcal{R}_\ell^{\textsc{lat}} = c_\ell d_\ell \log\big(1 +  \frac{\sigma_{\ell+1}^2}{\sigma_{\ell}^2}\big), c_\ell= \frac{\sigma_{\ell+1}^2 \sigma_{\textsc{max}}^{2 \ell}}{\log\left(1 + \frac{\sigma_{\ell+1}^2}{\sigma^2} \right)}, \end{align}
\end{proposition}
From \cref{prop:low_rank_regret}, our bounds scales as
\begin{align}
    \mathcal{B}\mathcal{R}(T) = \tilde{\mathcal{O}}\Big(\sqrt{T  (dK \sigma_1^2 + \sum_{\ell=1}^L d_\ell \sigma_{\ell+1}^2\sigma_{\textsc{max}}^{2 \ell} )}\Big)\,.
\end{align}

\section{Regret proof}\label{sec:dts_regret_proof}
\textbf{Important notation clarification.} Throughout this proof, we operate under the standard Bayesian bandit framework where the true environment parameters $\theta_*$ are drawn from the same prior distribution that \alg uses for posterior inference. Specifically, the true action parameters $\theta_{*, a}$ for $a \in [K]$ and the true latent parameters $\psi_{*, \ell}$ for $\ell \in [L]$ are assumed to be sampled according to the generative process in \cref{eq:dts_contextual_gaussian_model}. As a consequence, the true parameters $\theta_*$ and the model parameters $\theta$ used in our derivations follow the same distribution, and we use them interchangeably throughout the proof to simplify notation.

\subsection{Proof Sketch}\label{subsec:dts_dts_sketch1}

We start with the following standard lemma upon which we build our analysis \citep{aouali2022mixed}. 

\begin{lemma}\label{lemma:dts_standard_regret_bound}
Assume that $p(\theta_a \mid H_t) = \cN(\theta_a; \check{\mu}_{t, a}, \check{\Sigma}_{t, a})$ for any $a \in [K]$, then for any $\delta \in (0, 1)$,
\begin{align}\label{standard_regret_bound}
   \mathcal{B}\mathcal{R}(T) &\leq  \sqrt{2 T \log(1/\delta)}
  \sqrt{\E{}{\sum_{t = 1}^T \normw{X_t}{\check{\Sigma}_{t,A_t}}^2}} +
  c T\delta\,, \qquad \text{where $c>0$ is a constant}\,.
\end{align} 
\end{lemma}
Applying \cref{lemma:dts_standard_regret_bound} requires proving that the \emph{marginal} action-posterior densities of $ \theta_a \mid H_t$ in \cref{eq:dts_sampling_equivalence} are Gaussian and computing their covariances, while we only know the \emph{conditional} action-posteriors $p(\theta_a \mid \psi_1, H_t)$ and latent-posteriors $p(\psi_{\ell-1} \mid \psi_\ell, H_t)$. This is achieved by leveraging the preservation properties of the family of Gaussian distributions \citep{koller09probabilistic} and the total covariance decomposition \citep{weiss05probability} which leads to the next lemma. 
\begin{lemma}\label{lemma:dts_gaussian_covariance} Let $t \in [T]$ and $a \in [K]$, then the marginal covariance matrix $\check{\Sigma}_{t, a}$ reads
\begin{align}\label{eq:dts_gaussian_covariance}
  &\check{\Sigma}_{t,a}= \hat{\Sigma}_{t,a} + \sum_{\ell \in [L]} \PP_{a, \ell} \bar{\Sigma}_{t,\ell} \PP_{a, \ell}^\top\,, & \text{where $\, \PP_{a, \ell} = \hat{\Sigma}_{t, a}  \Sigma_1^{-1} \W_1 \prod_{i=1}^{\ell-1} \bar{\Sigma}_{t, i} \Sigma_{i+1}^{-1}  \W_{i+1}$.}
\end{align}
\end{lemma}
The marginal covariance matrix $\check{\Sigma}_{t,a}$ in \cref{eq:dts_gaussian_covariance} decomposes into $L+1$ terms. The first term corresponds to the posterior uncertainty of $\theta_{ a} \mid \psi_{1}$. The remaining $L$ terms capture the posterior uncertainties of $\psi_{ L}$ and $\psi_{ \ell-1} \mid \psi_{ \ell}$ for $\ell \in [L]/\{1\}$. These are then used to quantify the posterior information gain of latent parameters after one round as follows.  
\begin{lemma}[Posterior information gain]\label{lemma:dts_positive_diff} Let $t \in [T]$ and $\ell \in [L]$, then
\begin{align}\label{eq:dts_positive_diff}
  &\bar{\Sigma}_{t+1,\ell}^{-1} -  \bar{\Sigma}_{t,\ell}^{-1} \succeq  \sigma^{-2} \sigma_{\textsc{max}}^{-2 \ell}  \PP_{A_t, \ell}^\top X_t X_t^\top \PP_{A_t, \ell}\,,& \text{where } \, \sigma_{\textsc{max}}^2= \max_{\ell \in [L+1]} 1 + \frac{\sigma_\ell^2}{\sigma^2}\,.
\end{align}
\end{lemma}
Finally, \cref{lemma:dts_gaussian_covariance} is used to decompose $\normw{X_t}{\check{\Sigma}_{t, A_t}}^2$ in \cref{standard_regret_bound} into $L+1$ terms. Each term is bounded thanks to \cref{lemma:dts_positive_diff}. This results in the Bayes regret bound in \cref{thm:dts_regret}.   

\subsection{Proof of lemma~\ref{lemma:dts_gaussian_covariance}}

In this proof, we heavily rely on the total covariance decomposition \citep{weiss05probability}. Also, refer to \citep[Section 5.2]{hong22hierarchical} for a brief introduction to this decomposition. Now, from \cref{eq:dts_linear action-posterior}, we have that
\begin{align*}
    \condcov{\theta_{a}}{H_t, \psi_{ 1}} &=\hat{\Sigma}_{t, a} =  \left( \hat{G}_{t, a} + \Sigma_1^{-1} \right)^{-1} \,,\\
    \condE{\theta_{a}}{H_t, \psi_{ 1}} & =\hat{\mu}_{t, a} =  \hat{\Sigma}_{t, a}  \left(  \hat{G}_{t, a}\hat{B}_{t, a}  + \Sigma_1^{-1} \W_1 \psi_{ 1} \right)\,.
\end{align*}
First, given $H_t$, $\condcov{\theta_{a}}{H_t, \psi_{ 1}} =\left( \hat{G}_{t, a} + \Sigma_1^{-1} \right)^{-1}$ is constant. Thus 
\begin{align*}   \condE{\condcov{\theta_{a}}{H_t, \psi_{ 1}}}{H_t} = \condcov{\theta_{a}}{H_t, \psi_{ 1}} = \left( \hat{G}_{t, a} + \Sigma_1^{-1} \right)^{-1} = \hat{\Sigma}_{t, a} \,.
\end{align*}
In addition, given $H_t$, $\hat{\Sigma}_{t, a}$, $\hat{G}_{t, a}$ and $ \hat{B}_{t, a}$ are constant. Thus
\begin{align*}
   \condcov{\condE{\theta_{a}}{H_t, \psi_{ 1}}}{H_t} & = \condcov{\hat{\Sigma}_{t, a}  \left(  \hat{G}_{t, a}\hat{B}_{t, a}  + \Sigma_1^{-1} \W_1 \psi_{ 1} \right)}{H_t}\,,\\
   &= \condcov{\hat{\Sigma}_{t, a} \Sigma_1^{-1} \W_1 \psi_{ 1}}{H_t}\,,\\
    & = \hat{\Sigma}_{t, a}  \Sigma_1^{-1} \W_1 \condcov{\psi_{ 1}}{H_t} \W_1^\top \Sigma_1^{-1} \hat{\Sigma}_{t, a}\,,\\
    &= \hat{\Sigma}_{t, a} \Sigma_1^{-1} \W_1 \overline{\overline{\Sigma}}_{t, 1} \W_1^\top \Sigma_1^{-1} \hat{\Sigma}_{t, a} \,,
\end{align*}
where $\overline{\overline{\Sigma}}_{t, 1} = \condcov{\psi_{ 1}}{H_t}$ is the marginal posterior covariance of $\psi_{ 1}$. Finally, the total covariance decomposition \citep{weiss05probability,hong22hierarchical} yields that
\begin{align}\label{eq:dts_helper_cov_decomp}
      \check{\Sigma}_{t,a} =  \condcov{\theta_{a}}{H_t}&=\condE{\condcov{\theta_{a}}{H_t, \psi_{ 1}}}{H_t} + \condcov{\condE{\theta_{a}}{H_t, \psi_{ 1}}}{H_t}\,,\nonumber\\
        &=\hat{\Sigma}_{t, a} + \hat{\Sigma}_{t, a}  \Sigma_1^{-1} \W_1 \overline{\overline{\Sigma}}_{t, 1} \W_1^\top \Sigma_1^{-1} \hat{\Sigma}_{t, a}\,,
\end{align}
However, $\overline{\overline{\Sigma}}_{t, 1} = \condcov{\psi_{ 1}}{H_t}$ is different from $\bar{\Sigma}_{t, 1} = \condcov{\psi_{ 1}}{H_t, \psi_{ 2}}$ that we already derived in \cref{eq:dts_hyper_posteriors}. Thus we do not know the expression of $\overline{\overline{\Sigma}}_{t, 1}$. But we can use the same total covariance decomposition trick to find it. Precisely, let  $\overline{\overline{\Sigma}}_{t, \ell} = \condcov{\psi_{ \ell}}{H_t}$ for any $\ell \in [L]$. Then we have that 
\begin{align*}
   \bar{\Sigma}_{t, 1} =  \condcov{\psi_{ 1}}{H_t, \psi_{ 2}} &=  \big( \Sigma_2^{-1} +  \bar{G}_{t, 1} \big)^{-1} \,,\\
  \bar{\mu}_{t, 1} =   \condE{\psi_{ 1}}{H_t, \psi_{ 2}} & = \bar{\Sigma}_{t, 1} \Big( \Sigma_2^{-1}  \W_2 \psi_{ 2} + \bar{B}_{t, 1} \Big)\,.
\end{align*}
First, given $H_t$, $ \condcov{\psi_{ 1}}{H_t, \psi_{ 2}} =  \big( \Sigma_2^{-1} +  \bar{G}_{t, 1} \big)^{-1}$ is constant. Thus 
\begin{align*}   
\condE{\condcov{\psi_{ 1}}{H_t, \psi_{ 2}}}{H_t} = \condcov{\psi_{ 1}}{H_t, \psi_{ 2}} = \bar{\Sigma}_{t, 1} \,.
\end{align*}
In addition, given $H_t$, $\bar{\Sigma}_{t, 1}$, $\tilde{\Sigma}_{t, 1}$ and $\bar{B}_{t, 1}$ are constant. Thus
\begin{align*}
    \condcov{ \condE{\psi_{ 1}}{H_t, \psi_{ 2}}}{H_t} &= \condcov{\bar{\Sigma}_{t, 1} \Big( \Sigma_2^{-1}  \W_2 \psi_{ 2} + \bar{B}_{t, 1} \Big)}{H_t}\,,\\
    &= \condcov{\bar{\Sigma}_{t, 1} \Sigma_2^{-1}  \W_2 \psi_{ 2}}{H_t}\,,\\
    &= \bar{\Sigma}_{t, 1} \Sigma_2^{-1}  \W_2 \condcov{\psi_{ 2}}{H_t} \W_2^\top \Sigma_2^{-1}   \bar{\Sigma}_{t, 1}\,,\\
    &= \bar{\Sigma}_{t, 1} \Sigma_2^{-1}  \W_2 \overline{\overline{\Sigma}}_{t, 2} \W_2^\top \Sigma_2^{-1}   \bar{\Sigma}_{t, 1}\,.
\end{align*}
Finally, total covariance decomposition \citep{weiss05probability,hong22hierarchical} leads to
\begin{align*}
      \overline{\overline{\Sigma}}_{t, 1} =  \condcov{\psi_{ 1}}{H_t}&= \condE{\condcov{\psi_{ 1}}{H_t, \psi_{ 2}}}{H_t} + \condcov{ \condE{\psi_{ 1}}{H_t, \psi_{ 2}}}{H_t}\,,\\
        &=\bar{\Sigma}_{t, 1} + \bar{\Sigma}_{t, 1} \Sigma_2^{-1}  \W_2 \overline{\overline{\Sigma}}_{t, 2} \W_2^\top \Sigma_2^{-1}   \bar{\Sigma}_{t, 1}\,.
\end{align*}
Now using the techniques, this can be generalized using the same technique as above to 
\begin{align*}
      &\overline{\overline{\Sigma}}_{t,\ell} =\bar{\Sigma}_{t, \ell} + \bar{\Sigma}_{t, \ell} \Sigma_{\ell+1}^{-1}  \W_{\ell+1} \overline{\overline{\Sigma}}_{t, \ell+1} \W_{\ell+1}^\top \Sigma_{\ell+1}^{-1}   \bar{\Sigma}_{t, \ell}\,, &\forall \ell \in [L-1]\,.
\end{align*}
Then, by induction, we get that 
\begin{align*}
      &\overline{\overline{\Sigma}}_{t,1} = \sum_{\ell \in [L]} \bar{\PP}_\ell \bar{\Sigma}_{t, \ell}\bar{\PP}_\ell^\top\,, &\forall \ell \in [L-1]\,, 
\end{align*}
where we use that by definition $\overline{\overline{\Sigma}}_{t, L}= \condcov{\psi_{ L}}{H_t} = \bar{\Sigma}_{t, L}$ and set $\bar{\PP}_1 = I_d$ and $\bar{\PP}_\ell = \prod_{i=1}^{\ell-1} \bar{\Sigma}_{t, i} \Sigma_{i+1}^{-1}  \W_{i+1}$ for any $\ell \in [L]/\{1\}$. Plugging this in \cref{eq:dts_helper_cov_decomp} leads to 
\begin{align*}
      \check{\Sigma}_{t,a} &= \hat{\Sigma}_{t, a} +  \sum_{\ell \in [L]} \hat{\Sigma}_{t, a}  \Sigma_1^{-1} \W_1 \bar{\PP}_\ell \bar{\Sigma}_{t, \ell}\bar{\PP}_\ell^\top \W_1^\top \Sigma_1^{-1} \hat{\Sigma}_{t, a}\,,\\
      &= \hat{\Sigma}_{t, a} +  \sum_{\ell \in [L]} \hat{\Sigma}_{t, a}  \Sigma_1^{-1} \W_1 \bar{\PP}_\ell \bar{\Sigma}_{t, \ell} (\hat{\Sigma}_{t, a}  \Sigma_1^{-1} \W_1)^\top\,,\\
        &= \hat{\Sigma}_{t, a} +  \sum_{\ell \in [L]} \PP_{a, \ell} \bar{\Sigma}_{t, \ell} \PP_{a, \ell}^\top\,,
\end{align*}
where $\PP_{a, \ell} = \hat{\Sigma}_{t, a}  \Sigma_1^{-1} \W_1 \bar{\PP}_\ell = \hat{\Sigma}_{t, a}  \Sigma_1^{-1} \W_1 \prod_{i=1}^{\ell-1} \bar{\Sigma}_{t, i} \Sigma_{i+1}^{-1}  \W_{i+1}$.

\subsection{Proof of lemma~\ref{lemma:dts_positive_diff}}
We prove this result by induction. We start with the base case when $\ell=1$.

\textbf{(I) Base case.} Let $u =  \sigma^{-1} \hat{\Sigma}_{t, A_t}^{\frac{1}{2}} X_t$ From the expression of $\bar{\Sigma}_{t, 1}$ in \cref{eq:dts_hyper_posteriors}, we have that
\begin{align}
  \bar{\Sigma}_{t + 1, 1}^{-1} - \bar{\Sigma}_{t, 1}^{-1} & = \W_1^\top \left(\Sigma_1^{-1} - \Sigma_1^{-1} (\hat{\Sigma}_{t, A_t}^{-1} + \sigma^{-2} X_t X_t^\top)^{-1} \Sigma_1^{-1} -
  (\Sigma_1^{-1} - \Sigma_1^{-1} \hat{\Sigma}_{t, A_t} \Sigma_1^{-1})\right)\W_1\,,
  \nonumber \\
  & = \W_1^\top \left(\Sigma_1^{-1} (\hat{\Sigma}_{t, A_t} - (\hat{\Sigma}_{t, A_t}^{-1} + \sigma^{-2} X_t X_t^\top)^{-1}) \Sigma_1^{-1}\right)\W_1\,,
  \nonumber \\
  & = \W_1^\top \left(\Sigma_1^{-1} \hat{\Sigma}_{t, A_t}^{\frac{1}{2}}
  (I_d - (I_d + \sigma^{-2} \hat{\Sigma}_{t, A_t}^{\frac{1}{2}} X_t X_t^\top \hat{\Sigma}_{t, A_t}^{\frac{1}{2}})^{-1})
  \hat{\Sigma}_{t, A_t}^{\frac{1}{2}} \Sigma_1^{-1}\right)\W_1\,,
  \nonumber \\
  & = \W_1^\top \left(\Sigma_1^{-1} \hat{\Sigma}_{t, A_t}^{\frac{1}{2}}
  (I_d - (I_d + u u^\top)^{-1})
  \hat{\Sigma}_{t, A_t}^{\frac{1}{2}} \Sigma_1^{-1}\right)\W_1\,,
  \nonumber \\
  & \stackrel{(i)}{=} \W_1^\top \left(\Sigma_1^{-1} \hat{\Sigma}_{t, A_t}^{\frac{1}{2}}
  \frac{u u^\top}{1 + u^\top u}
  \hat{\Sigma}_{t, A_t}^{\frac{1}{2}} \Sigma_1^{-1}\right)\W_1\,,\nonumber\\
  & \stackrel{(ii)}{=} \sigma^{-2} \W_1^\top \Sigma_1^{-1} \hat{\Sigma}_{t, A_t}
  \frac{X_t X_t^\top}{1 + u^\top u}
  \hat{\Sigma}_{t, A_t} \Sigma_1^{-1}\W_1\,.
  \label{eq:dts_linear telescoping}
\end{align}
In $(i)$ we use the Sherman-Morrison formula. Note that $(ii)$ says that $\bar{\Sigma}_{t + 1, 1}^{-1} - \bar{\Sigma}_{t, 1}^{-1}$ is one-rank which we will also need in induction step. Now, we have that $\norm{X_t}^2 = 1$. Therefore,
\begin{align*}
  1 + u^\top u
  = 1 + \sigma^{-2} X_t^\top \hat{\Sigma}_{t, A_t} X_t
  \leq 1 + \sigma^{-2} \lambda_1(\Sigma_1) \norm{X_t}^2 =  1 + \sigma^{-2}  \sigma_1^2 \leq \sigma_{\textsc{max}}^2\,,
\end{align*}
where we use that by definition of $\sigma_{\textsc{max}}^2 $ in \cref{lemma:dts_positive_diff}, we have that $\sigma_{\textsc{max}}^2 \geq 1 + \sigma^{-2}  \sigma_1^2$. Therefore, by taking the inverse, we get that $\frac{1}{ 1 + u^\top u} \geq \sigma_{\textsc{max}}^{-2}$. Combining this with  \cref{eq:dts_linear telescoping} leads to
\begin{align*}
    \bar{\Sigma}_{t + 1, 1}^{-1} - \bar{\Sigma}_{t, 1}^{-1} \succeq   \sigma^{-2} \sigma_{\textsc{max}}^{-2} \W_1^\top \Sigma_1^{-1} \hat{\Sigma}_{t, A_t}
  X_t X_t^\top
  \hat{\Sigma}_{t, A_t} \Sigma_1^{-1}\W_1
\end{align*}
Noticing that $\PP_{A_t, 1} = \hat{\Sigma}_{t, A_t} \Sigma_1^{-1}\W_1$ concludes the proof of the base case when $\ell=1$.

\textbf{(II) Induction step.} Let $\ell \in [L]/\{1\}$ and suppose that $\bar{\Sigma}_{t+1,\ell-1}^{-1} -  \bar{\Sigma}_{t,\ell-1}^{-1}$ is one-rank and that it holds for $\ell-1$ that
\begin{align*}
  &\bar{\Sigma}_{t+1,\ell-1}^{-1} -  \bar{\Sigma}_{t,\ell-1}^{-1} \succeq  \sigma^{-2} \sigma_{\textsc{max}}^{-2(\ell-1)}  \PP_{A_t, \ell-1}^\top X_t X_t^\top \PP_{A_t, \ell-1}\,,& \text{where } \, \sigma_{\textsc{max}}^{2} = \max_{\ell \in [L+1]} 1 +  \sigma^{-2} \sigma_\ell^2.
\end{align*}
Then, we want to show that $\bar{\Sigma}_{t+1,\ell}^{-1} -  \bar{\Sigma}_{t,\ell}^{-1}$ is also one-rank and that it holds that
\begin{align*}
  &\bar{\Sigma}_{t+1,\ell}^{-1} -  \bar{\Sigma}_{t,\ell}^{-1} \succeq  \sigma^{-2} \sigma_{\textsc{max}}^{-2\ell}  \PP_{A_t, \ell}^\top X_t X_t^\top \PP_{A_t, \ell}\,,& \text{where } \, \sigma_{\textsc{max}}^{2} = \max_{\ell \in [L+1]} 1 +  \sigma^{-2} \sigma_\ell^2.
\end{align*}
This is achieved as follows. Define the precision increment at level $\ell-1$ by
$$
\Delta_{t,\ell-1} := \bar{\Sigma}_{t+1,\ell-1}^{-1}-\bar{\Sigma}_{t,\ell-1}^{-1}.
$$
By the induction hypothesis, $\Delta_{t,\ell-1}$ is rank-one PSD, hence there exists $u\in\R^d$ such that
$$
\Delta_{t,\ell-1} = u u^\top.
$$
Using \cref{eq:dts_step_gaussian_recursive}, we have for any $s\in\{t,t+1\}$:
$$
\bar{\Sigma}_{s,\ell}^{-1}
= \Sigma_{\ell+1}^{-1} + \bar{G}_{s,\ell}
= \Sigma_{\ell+1}^{-1} + \W_\ell^\top\!\left(\Sigma_\ell^{-1}-\Sigma_\ell^{-1}\bar{\Sigma}_{s,\ell-1}\Sigma_\ell^{-1}\right)\!\W_\ell .
$$
Therefore,
\begin{align*}
\bar{\Sigma}_{t+1,\ell}^{-1}-\bar{\Sigma}_{t,\ell}^{-1}
&= \bar{G}_{t+1,\ell}-\bar{G}_{t,\ell}\\
&= \W_\ell^\top \Sigma_\ell^{-1}\big(\bar{\Sigma}_{t,\ell-1}-\bar{\Sigma}_{t+1,\ell-1}\big)\Sigma_\ell^{-1}\W_\ell .
\end{align*}
Since $\bar{\Sigma}_{t+1,\ell-1}^{-1}=\bar{\Sigma}_{t,\ell-1}^{-1}+u u^\top$, Sherman--Morrison yields
$$
\bar{\Sigma}_{t+1,\ell-1}
=\left(\bar{\Sigma}_{t,\ell-1}^{-1}+u u^\top\right)^{-1}
=\bar{\Sigma}_{t,\ell-1}-\frac{\bar{\Sigma}_{t,\ell-1}u u^\top \bar{\Sigma}_{t,\ell-1}}{1+u^\top\bar{\Sigma}_{t,\ell-1}u}.
$$
Hence,
$$
\bar{\Sigma}_{t,\ell-1}-\bar{\Sigma}_{t+1,\ell-1}
=\frac{\bar{\Sigma}_{t,\ell-1}u u^\top \bar{\Sigma}_{t,\ell-1}}{1+u^\top\bar{\Sigma}_{t,\ell-1}u},
$$
and plugging this back gives
\begin{align*}
\bar{\Sigma}_{t+1,\ell}^{-1}-\bar{\Sigma}_{t,\ell}^{-1}
&=\W_{\ell}^\top \Sigma_{\ell}^{-1} \bar{\Sigma}_{t, \ell-1}
\frac{u u^{\top}}{1+u^{\top} \bar{\Sigma}_{t, \ell-1} u}
\bar{\Sigma}_{t, \ell-1} \Sigma_{\ell}^{-1} \W_{\ell}.
\end{align*}
In particular, this increment is rank-one PSD.

However, it follows from the induction hypothesis that
$$
u u^\top
= \bar{\Sigma}_{t+1,\ell-1}^{-1}-\bar{\Sigma}_{t,\ell-1}^{-1}
\succeq \sigma^{-2}\sigma_{\textsc{max}}^{-2(\ell-1)}\,
\PP_{A_t,\ell-1}^\top X_t X_t^\top \PP_{A_t,\ell-1}.
$$
Therefore, 
\begin{align*}
    \bar{\Sigma}_{t+1,\ell}^{-1} -  \bar{\Sigma}_{t,\ell}^{-1} & =\W_{\ell}^\top \Sigma_{\ell}^{-1} \bar{\Sigma}_{t, \ell-1} \frac{u u^{\top}}{1+u^{\top} \bar{\Sigma}_{t, \ell-1} u} \bar{\Sigma}_{t, \ell-1} \Sigma_{\ell}^{-1} \W_{\ell}\,,\\
   &\succeq  \W_{\ell}^\top \Sigma_{\ell}^{-1} \bar{\Sigma}_{t, \ell-1} \frac{ \sigma^{-2} \sigma_{\textsc{max}}^{-2(\ell-1)}  \PP_{A_t, \ell-1}^\top X_t X_t^\top \PP_{A_t, \ell-1}}{1+u^{\top} \bar{\Sigma}_{t, \ell-1} u} \bar{\Sigma}_{t, \ell-1} \Sigma_{\ell}^{-1} \W_{\ell}\,,\\
      &= \frac{\sigma^{-2} \sigma_{\textsc{max}}^{-2(\ell-1)}}{1+u^{\top} \bar{\Sigma}_{t, \ell-1} u}  \W_{\ell}^\top \Sigma_{\ell}^{-1} \bar{\Sigma}_{t, \ell-1}   \PP_{A_t, \ell-1}^\top X_t X_t^\top \PP_{A_t, \ell-1} \bar{\Sigma}_{t, \ell-1} \Sigma_{\ell}^{-1} \W_{\ell}\,,\\
      &=\frac{\sigma^{-2} \sigma_{\textsc{max}}^{-2(\ell-1)}}{1+u^{\top} \bar{\Sigma}_{t, \ell-1} u}\PP_{A_t, \ell}^\top X_t X_t^\top \PP_{A_t, \ell}\,.
\end{align*}
Finally, we use that $1+u^{\top} \bar{\Sigma}_{t, \ell-1} u \leq 1 + \normw{u}{2}^2 \lambda_1(\bar{\Sigma}_{t, \ell-1})   \leq 1 + \sigma^{-2} \sigma_\ell^2$. Here we use that $\normw{u}{2}^2 \leq \sigma^{-2}$, which can also be proven by induction, and that $\lambda_1(\bar{\Sigma}_{t, \ell-1}) \leq \sigma_\ell^2$, which follows from the expression of $\bar{\Sigma}_{t, \ell-1}$ in \cref{subsec:dts_dts_posterior_expressions}. Therefore, we have that 
\begin{align*}
    \bar{\Sigma}_{t+1,\ell}^{-1} -  \bar{\Sigma}_{t,\ell}^{-1}       & \succeq \frac{\sigma^{-2} \sigma_{\textsc{max}}^{-2(\ell-1)}}{1+u^{\top} \bar{\Sigma}_{t, \ell-1} u}\PP_{A_t, \ell}^\top X_t X_t^\top \PP_{A_t, \ell}\,,\\
    & \succeq \frac{\sigma^{-2} \sigma_{\textsc{max}}^{-2(\ell-1)}}{1 + \sigma^{-2} \sigma_\ell^2}\PP_{A_t, \ell}^\top X_t X_t^\top \PP_{A_t, \ell}\,,\\
     & \succeq \sigma^{-2} \sigma_{\textsc{max}}^{-2\ell}\PP_{A_t, \ell}^\top X_t X_t^\top \PP_{A_t, \ell}\,,
\end{align*}
where the last inequality follows from the definition of $\sigma_{\textsc{max}}^2 = \max_{\ell \in [L+1]} 1 + \sigma^{-2}\sigma_\ell^2$. This concludes the proof.

\subsection{Proof of theorem~\ref{thm:dts_regret}}\label{subsec:dts_dts_proof_thm}
We start with the following standard result which we borrow from \citep{hong22deep, aouali2022mixed},
\begin{align}\label{eq:dts_helper_standard_regret_bound}
   \mathcal{B}\mathcal{R}(T) &\leq  \sqrt{2 T \log(1/\delta)}
  \sqrt{\E{}{\sum_{t = 1}^T \normw{X_t}{\check{\Sigma}_{t,A_t}}^2}} +
  c T\delta\,, \qquad \text{where $c>0$ is a constant}\,.
\end{align} 
Then we use \cref{lemma:dts_gaussian_covariance} and express the marginal covariance $\check{\Sigma}_{t,A_t}$ as
\begin{align}\label{eq:dts_helper_gaussian_covariance}
  &\check{\Sigma}_{t,a}= \hat{\Sigma}_{t,a} + \sum_{\ell \in [L]} \PP_{a, \ell} \bar{\Sigma}_{t,\ell} \PP_{a, \ell}^\top\,, & \text{where $\, \PP_{a, \ell} = \hat{\Sigma}_{t, a}  \Sigma_1^{-1} \W_1 \prod_{i=1}^{\ell-1} \bar{\Sigma}_{t, i} \Sigma_{i+1}^{-1}  \W_{i+1}$.}
\end{align}
Therefore, we can decompose $\normw{X_t}{\check{\Sigma}_{t,A_t}}^2$ as \begin{align}\label{eq:dts_sequential proof decomposition}
  & \normw{X_t}{\check{\Sigma}_{t, A_t}}^2 = \sigma^2 \frac{X_t^\top \check{\Sigma}_{t, A_t} X_t}{\sigma^2} \stackrel{(i)}{=} \sigma^2 \big(\sigma^{-2} X_t^\top \hat{\Sigma}_{t, A_t} X_t +
  \sigma^{-2} \sum_{\ell \in [L]} X_t^\top \PP_{A_t, \ell} \bar{\Sigma}_{t,\ell} \PP_{A_t, \ell}^\top X_t\big)\,,
  \nonumber \\
  & \stackrel{(ii)}{\leq} c_0 \log(1 + \sigma^{-2} X_t^\top \hat{\Sigma}_{t, A_t} X_t) + 
  \sum_{\ell \in [L]}  c_\ell \log(1 + \sigma^{-2} X_t^\top \PP_{A_t, \ell} \bar{\Sigma}_{t,\ell} \PP_{A_t, \ell}^\top X_t)\,,
\end{align}
where $(i)$ follows from \cref{eq:dts_helper_gaussian_covariance}, and we use the following inequality in $(ii)$
\begin{align*}
  x
  = \frac{x}{\log(1 + x)} \log(1 + x)
  \leq \left(\max_{x \in [0, u]} \frac{x}{\log(1 + x)}\right) \log(1 + x)
  = \frac{u}{\log(1 + u)} \log(1 + x)\,,
\end{align*}
which holds for any $x \in [0, u]$, where constants $c_0$ and $c_\ell$ are derived as
\begin{align*}
  c_0
  = \frac{\sigma_1^2}{\log(1 +  \frac{\sigma_1^2}{\sigma^{2}})}\,, \quad
  c_\ell
  = \frac{\sigma_{\ell+1}^2}{\log(1 + \frac{\sigma_{\ell+1}^2}{\sigma^{2}})}\,.
\end{align*}
The derivation of $c_0$ uses that
\begin{align*}
  X_t^\top \hat{\Sigma}_{t, A_t} X_t
  \leq \lambda_1(\hat{\Sigma}_{t, A_t}) \norm{X_t}^2
  \leq  \lambda_d^{-1}(\Sigma_{1}^{-1} + G_{t, A_t})
  \leq \lambda_d^{-1}(\Sigma_{1}^{-1})
  = \lambda_1(\Sigma_{1}) =  \sigma_1^2 \,.
\end{align*}
The derivation of $c_\ell$ follows from
\begin{align*}
 X_t^\top \PP_{A_t, \ell} \bar{\Sigma}_{t,\ell} \PP_{A_t, \ell}^\top X_t
  \leq \lambda_1(\PP_{A_t, \ell} \PP_{A_t, \ell}^\top) \lambda_1(\bar{\Sigma}_{t,\ell}) \norm{X_t}^2
  & \leq \sigma_{\ell+1}^2\,.
\end{align*}
Therefore, from \cref{eq:dts_sequential proof decomposition} and \cref{eq:dts_helper_standard_regret_bound}, we get that 
\begin{align}\label{eq:dts_main_terms}
     \mathcal{B}\mathcal{R}(T) &\leq \sqrt{2 T \log(1/\delta)}
  \Big(\mathbb{E}\Big[ c_0 \sum_{t = 1}^T \log(1 + \sigma^{-2} X_t^\top \hat{\Sigma}_{t, A_t} X_t) \nonumber\\ & + 
  \sum_{\ell \in [L]}  c_\ell \sum_{t = 1}^T\log(1 + \sigma^{-2} X_t^\top \PP_{A_t, \ell} \bar{\Sigma}_{t,\ell} \PP_{A_t, \ell}^\top X_t) \Big]\Big)^{\frac{1}{2}} +
  c T\delta 
\end{align}

Now we focus on bounding the logarithmic terms in \cref{eq:dts_main_terms}.

\textbf{(I) First term in \cref{eq:dts_main_terms}} We first rewrite this term as
\begin{align*}
   \log(1 & + \sigma^{-2}  X_t^\top \hat{\Sigma}_{t, A_t} X_t) \stackrel{(i)}{=} \log\det(I_d + \sigma^{-2}\hat{\Sigma}_{t, A_t}^\frac{1}{2} X_t X_t^\top \hat{\Sigma}_{t, A_t}^\frac{1}{2})\,,\\
  &= \log\det(\hat{\Sigma}_{t, A_t}^{-1} + \sigma^{-2} X_t X_t^\top) - \log\det(\hat{\Sigma}_{t, A_t}^{-1})
  = \log\det(\hat{\Sigma}_{t+1, A_t}^{-1}) - \log\det(\hat{\Sigma}_{t, A_t}^{-1})\,,
\end{align*}
where $(i)$ follows from the Weinstein-Aronszajn identity. Then we sum over all rounds $ t \in [T]$, and get a telescoping
\begin{align*}
  \sum_{t = 1}^{T}
   &\log\det(I_d +\sigma^{-2} \hat{\Sigma}_{t, A_t}^\frac{1}{2} X_t
  X_t^\top \hat{\Sigma}_{t, A_t}^\frac{1}{2})=\sum_{t = 1}^{T} \log\det(\hat{\Sigma}_{t+1, A_t}^{-1}) - \log\det(\hat{\Sigma}_{t, A_t}^{-1})\,,\\
  &=\sum_{t = 1}^{T} \sum_{a=1}^K \log\det(\hat{\Sigma}_{t+1, a}^{-1}) - \log\det(\hat{\Sigma}_{t, a}^{-1})=\sum_{a=1}^K \sum_{t = 1}^{T} \log\det(\hat{\Sigma}_{t+1, a}^{-1}) - \log\det(\hat{\Sigma}_{t, a}^{-1})\,,\\
  &= \sum_{a=1}^K \log\det(\hat{\Sigma}_{T+1, a}^{-1}) - \log\det(\hat{\Sigma}_{1, a}^{-1})
  \stackrel{(i)}{=} \sum_{a=1}^K \log\det(\Sigma_1^\frac{1}{2} \hat{\Sigma}_{T+1, a}^{-1} \Sigma_1^\frac{1}{2})\,,
\end{align*}
where $(i)$ follows from the fact that $\hat{\Sigma}_{1, a} = \Sigma_1$. Now we use the inequality of arithmetic and geometric means and get 
\begin{align}\label{eq:dts_herlper_}
    \sum_{t = 1}^{T}
   \log\det(I_d +\sigma^{-2} \hat{\Sigma}_{t, A_t}^\frac{1}{2} X_t
  X_t^\top \hat{\Sigma}_{t, A_t}^\frac{1}{2}) & =\sum_{a=1}^K \log\det(\Sigma_1^\frac{1}{2} \hat{\Sigma}_{T+1, a}^{-1} \Sigma_1^\frac{1}{2})  \,,\nonumber\\
  & \leq \sum_{a=1}^K d \log\left(\frac{1}{d} \operatorname{Tr}(\Sigma_1^\frac{1}{2} \hat{\Sigma}_{T+1,a}^{-1}
  \Sigma_1^\frac{1}{2})\right)\,,\\
  &\leq \sum_{a=1}^K d \log\left(1 + \frac{   T}{ d} \frac{\sigma_1^2}{\sigma^2}\right) = Kd\log\left(1 + \frac{   T}{ d} \frac{\sigma_1^2}{\sigma^2}\right)\nonumber\,.
\end{align}

\textbf{(II) Remaining terms in \cref{eq:dts_main_terms}} Let $\ell \in [L]$. Then we have that
\begin{align*}
   \log(1 &+ \sigma^{-2} X_t^\top \PP_{A_t, \ell} \bar{\Sigma}_{t,\ell} \PP_{A_t, \ell}^\top X_t) = \sigma_{\textsc{max}}^{2\ell} \sigma_{\textsc{max}}^{-2\ell}\log(1 + \sigma^{-2} X_t^\top \PP_{A_t, \ell} \bar{\Sigma}_{t,\ell} \PP_{A_t, \ell}^\top X_t)\,,\\
  &\leq \sigma_{\textsc{max}}^{2\ell}\log(1 + \sigma^{-2} \sigma_{\textsc{max}}^{-2\ell} X_t^\top \PP_{A_t, \ell} \bar{\Sigma}_{t,\ell} \PP_{A_t, \ell}^\top X_t)\,,\\
  &\stackrel{(i)}{=}\sigma_{\textsc{max}}^{2\ell}\log\det(I_d + \sigma^{-2} \sigma_{\textsc{max}}^{-2\ell} \bar{\Sigma}_{t,\ell}^{\frac{1}{2}} \PP_{A_t, \ell}^\top X_t X_t^\top \PP_{A_t, \ell}  \bar{\Sigma}_{t,\ell}^{\frac{1}{2}})\,,\\
  &=\sigma_{\textsc{max}}^{2\ell}\Big(\log\det(\bar{\Sigma}_{t,\ell}^{-1} + \sigma^{-2} \sigma_{\textsc{max}}^{-2\ell}  \PP_{A_t, \ell}^\top X_t X_t^\top \PP_{A_t, \ell})- \log\det(\bar{\Sigma}_{t,\ell}^{-1})\Big)\,,
\end{align*}
where we use the Weinstein-Aronszajn identity in $(i)$. Now we know from \cref{lemma:dts_positive_diff} that the following inequality holds $\sigma^{-2} \sigma_{\textsc{max}}^{-2\ell}  \PP_{A_t, \ell}^\top X_t X_t^\top \PP_{A_t, \ell} \preceq \bar{\Sigma}_{t+1,\ell}^{-1} - \bar{\Sigma}_{t,\ell}^{-1}$. As a result, we get that $\bar{\Sigma}_{t,\ell}^{-1} + \sigma^{-2} \sigma_{\textsc{max}}^{-2\ell}  \PP_{A_t, \ell}^\top X_t X_t^\top \PP_{A_t, \ell} \preceq \bar{\Sigma}_{t+1,\ell}^{-1}$. Thus,  
\begin{align*}
   \log(1 + \sigma^{-2} X_t^\top \PP_{A_t, \ell} \bar{\Sigma}_{t,\ell} \PP_{A_t, \ell}^\top X_t) &\leq \sigma_{\textsc{max}}^{2\ell}\Big(\log\det(\bar{\Sigma}_{t+1,\ell}^{-1})- \log\det(\bar{\Sigma}_{t,\ell}^{-1})\Big)\,,
\end{align*}

Then we sum over all rounds $ t \in [T]$, and get a telescoping
\begin{align*}
   \sum_{t=1}^T \log(1 + \sigma^{-2} X_t^\top \PP_{A_t, \ell} \bar{\Sigma}_{t,\ell} \PP_{A_t, \ell}^\top X_t) &\leq \sigma_{\textsc{max}}^{2\ell} \sum_{t=1}^T \log\det(\bar{\Sigma}_{t+1,\ell}^{-1})- \log\det(\bar{\Sigma}_{t,\ell}^{-1}) \,,\\
   &=  \sigma_{\textsc{max}}^{2\ell} \Big( \log\det(\bar{\Sigma}_{T+1,\ell}^{-1})- \log\det(\bar{\Sigma}_{1,\ell}^{-1})\Big)\,,\\
    &\stackrel{(i)}{=}  \sigma_{\textsc{max}}^{2\ell} \Big( \log\det(\bar{\Sigma}_{T+1,\ell}^{-1})- \log\det(\Sigma_{\ell+1}^{-1})\Big)\,,\\
    &=  \sigma_{\textsc{max}}^{2\ell} \Big( \log\det(\Sigma_{\ell+1}^{\frac{1}{2}}\bar{\Sigma}_{T+1,\ell}^{-1}\Sigma_{\ell+1}^{\frac{1}{2}})\Big)\,,\\
\end{align*}
where we use that $\bar{\Sigma}_{1,\ell} = \Sigma_{\ell+1}$ in $(i)$. Finally, we use the inequality of arithmetic and geometric means and get that
\begin{align}\label{eq:dts_part_to_change_for_sparsity}
   \sum_{t=1}^T \log(1 + \sigma^{-2} X_t^\top \PP_{A_t, \ell} \bar{\Sigma}_{t,\ell} \PP_{A_t, \ell}^\top X_t) &\leq \sigma_{\textsc{max}}^{2\ell} \Big( \log\det(\Sigma_{\ell+1}^{\frac{1}{2}}\bar{\Sigma}_{T+1,\ell}^{-1}\Sigma_{\ell+1}^{\frac{1}{2}})\Big)\,,\nonumber\\
    & \leq d \sigma_{\textsc{max}}^{2\ell}  \log\left(\frac{1}{d} \operatorname{Tr}(\Sigma_{\ell+1}^\frac{1}{2} \bar{\Sigma}_{T+1,\ell}^{-1}
  \Sigma_{\ell+1}^\frac{1}{2})\right)\,,\\
    & \leq d \sigma_{\textsc{max}}^{2\ell}  \log\left(1 +  \frac{\sigma_{\ell+1}^2}{\sigma_{\ell}^2}\right)\,,\nonumber
\end{align}
The last inequality follows from the expression of $\bar{\Sigma}_{T+1,\ell}^{-1}$ in \cref{eq:dts_hyper_posteriors} that leads to 
\begin{align}\label{eq:dts_helper_expression}
    \Sigma_{\ell+1}^\frac{1}{2} \bar{\Sigma}_{T+1,\ell}^{-1}
  \Sigma_{\ell+1}^\frac{1}{2} &=  I_d + \Sigma_{\ell+1}^\frac{1}{2}  \bar{G}_{T+1, \ell} \Sigma_{\ell+1}^\frac{1}{2}\,,\nonumber\\
  &= I_d + \Sigma_{\ell+1}^\frac{1}{2} \W_{\ell}^\top \big(\Sigma_{\ell}^{-1}-  \Sigma_{\ell}^{-1}\bar{\Sigma}_{T+1, \ell-1}\Sigma_{\ell}^{-1}\big)\W_{\ell}\Sigma_{\ell+1}^\frac{1}{2}\,,
\end{align}
since $ \bar{G}_{T+1, \ell}= \W_{\ell}^\top \big(\Sigma_{\ell}^{-1}-  \Sigma_{\ell}^{-1}\bar{\Sigma}_{T+1, \ell-1}\Sigma_{\ell}^{-1}\big)\W_{\ell}$. This allows us to bound $\frac{1}{d} \operatorname{Tr}(\Sigma_{\ell+1}^\frac{1}{2} \bar{\Sigma}_{T+1,\ell}^{-1}
  \Sigma_{\ell+1}^\frac{1}{2})$ as
  \begin{align}\label{eq:dts_useful_derivations}
      \frac{1}{d} \operatorname{Tr}(\Sigma_{\ell+1}^\frac{1}{2} \bar{\Sigma}_{T+1,\ell}^{-1}
  \Sigma_{\ell+1}^\frac{1}{2})& = \frac{1}{d} \operatorname{Tr}(I_d + \Sigma_{\ell+1}^\frac{1}{2} \W_{\ell}^\top \big(\Sigma_{\ell}^{-1}-  \Sigma_{\ell}^{-1}\bar{\Sigma}_{T+1, \ell-1}\Sigma_{\ell}^{-1}\big)\W_{\ell}\Sigma_{\ell+1}^\frac{1}{2})  \,, \nonumber\\
  & = \frac{1}{d} ( d + \operatorname{Tr}(\Sigma_{\ell+1}^\frac{1}{2} \W_{\ell}^\top \big(\Sigma_{\ell}^{-1}-  \Sigma_{\ell}^{-1}\bar{\Sigma}_{T+1, \ell-1}\Sigma_{\ell}^{-1}\big)\W_{\ell}\Sigma_{\ell+1}^\frac{1}{2})\,,\nonumber\\
  & \leq  1 + \frac{1}{d}\sum_{i=1}^d\lambda_1(\Sigma_{\ell+1}^\frac{1}{2} \W_{\ell}^\top \big(\Sigma_{\ell}^{-1}-  \Sigma_{\ell}^{-1}\bar{\Sigma}_{T+1, \ell-1}\Sigma_{\ell}^{-1}\big)\W_{\ell}\Sigma_{\ell+1}^\frac{1}{2}\,,\nonumber\\
    & \leq  1 + \frac{1}{d}\sum_{i=1}^d\lambda_1(\Sigma_{\ell+1}) \lambda_1(\W_{\ell}^\top \W_{\ell}) \lambda_1\big(\Sigma_{\ell}^{-1}-  \Sigma_{\ell}^{-1}\bar{\Sigma}_{T+1, \ell-1}\Sigma_{\ell}^{-1}\big)\,,\nonumber\\
    & \leq  1 + \frac{1}{d}\sum_{i=1}^d\lambda_1(\Sigma_{\ell+1}) \lambda_1(\W_{\ell}^\top \W_{\ell}) \lambda_1\big(\Sigma_{\ell}^{-1}\big)\,,\nonumber\\
     & \leq  1 + \frac{1}{d}\sum_{i=1}^d \frac{\sigma_{\ell+1}^2}{\sigma_{\ell}^2} = 1 +  \frac{\sigma_{\ell+1}^2}{\sigma_{\ell}^2} \,,
  \end{align}
where we use the assumption that $\lambda_1(\W_\ell^\top \W_\ell)=1$ \textbf{(A4)} and that $\lambda_1(\Sigma_{\ell+1}) = \sigma_{\ell+1}^2$ and $\lambda_1(\Sigma_{\ell}^{-1}) = 1/\sigma_{\ell}^2$. This is because $\Sigma_\ell = \sigma_\ell^2I_d$ for any $\ell \in [L+1]$. Finally, plugging \cref{eq:dts_herlper_,eq:dts_part_to_change_for_sparsity} in \cref{eq:dts_main_terms} concludes the proof.

\subsection{Proof of proposition~\ref{prop:low_rank_regret}}
We use exactly the same proof in \cref{subsec:dts_dts_proof_thm}, with one change to account for the sparsity assumption \textbf{(A5)}. The change corresponds to \cref{eq:dts_part_to_change_for_sparsity}. First, recall  that \cref{eq:dts_part_to_change_for_sparsity} writes
\begin{align}
   \sum_{t=1}^T \log(1 + \sigma^{-2} X_t^\top \PP_{A_t, \ell} \bar{\Sigma}_{t,\ell} \PP_{A_t, \ell}^\top X_t) &\leq \sigma_{\textsc{max}}^{2\ell} \Big( \log\det(\Sigma_{\ell+1}^{\frac{1}{2}}\bar{\Sigma}_{T+1,\ell}^{-1}\Sigma_{\ell+1}^{\frac{1}{2}})\Big)\,,\nonumber
\end{align}
where 
\begin{align}\label{eq:dts_matrix_identity}
    \Sigma_{\ell+1}^\frac{1}{2} \bar{\Sigma}_{T+1,\ell}^{-1}
  \Sigma_{\ell+1}^\frac{1}{2} &= I_d + \Sigma_{\ell+1}^\frac{1}{2} \W_{\ell}^\top \big(\Sigma_{\ell}^{-1}-  \Sigma_{\ell}^{-1}\bar{\Sigma}_{t, \ell-1}\Sigma_{\ell}^{-1}\big)\W_{\ell}\Sigma_{\ell+1}^\frac{1}{2}\,,\nonumber\\
  &= I_d + \sigma_{\ell+1}^2 \W_{\ell}^\top \big(\Sigma_{\ell}^{-1}-  \Sigma_{\ell}^{-1}\bar{\Sigma}_{t, \ell-1}\Sigma_{\ell}^{-1}\big)\W_{\ell}\,,
\end{align}
where the second equality follows from the assumption that $\Sigma_{\ell+1}=\sigma_{\ell+1}^2 I_d$. But notice that in our assumption, \textbf{(A5)}, we assume that $\W_\ell = (\bar{\W}_\ell, 0_{d, d-d_\ell})$, where $\bar{\W}_\ell \in \real^{d \times d_\ell}$ for any $\ell \in [L]$. Therefore, we have that for any $d \times d$ matrix $\Beta \in \real^{d \times d}$, the following holds, $\W_\ell^\top \Beta \W_\ell = {\begin{pmatrix}\bar{\W}_\ell^\top \Beta \bar{\W}_\ell&0_{d_\ell, d-d_\ell}\\0_{d-d_\ell, d_\ell}&0_{d-d_\ell, d-d_\ell}\end{pmatrix}}$. In particular, we have that 
\begin{align}
   \W_{\ell}^\top \big(\Sigma_{\ell}^{-1}-  \Sigma_{\ell}^{-1}\bar{\Sigma}_{t, \ell-1}\Sigma_{\ell}^{-1}\big)\W_{\ell} = {\begin{pmatrix}\bar{\W}_\ell^\top \big(\Sigma_{\ell}^{-1}-  \Sigma_{\ell}^{-1}\bar{\Sigma}_{t, \ell-1}\Sigma_{\ell}^{-1}\big) \bar{\W}_\ell&0_{d_\ell, d-d_\ell}\\0_{d-d_\ell, d_\ell}&0_{d-d_\ell, d-d_\ell}\end{pmatrix}}\,.
\end{align}
Therefore, plugging this in \cref{eq:dts_matrix_identity} yields that
\begin{align}
  \Sigma_{\ell+1}^\frac{1}{2} \bar{\Sigma}_{T+1,\ell}^{-1}
  \Sigma_{\ell+1}^\frac{1}{2} = {\begin{pmatrix}I_{d_\ell} + \sigma_{\ell+1}^2\bar{\W}_\ell^\top \big(\Sigma_{\ell}^{-1}-  \Sigma_{\ell}^{-1}\bar{\Sigma}_{t, \ell-1}\Sigma_{\ell}^{-1}\big) \bar{\W}_\ell&0_{d_\ell, d-d_\ell}\\0_{d-d_\ell, d_\ell}&I_{d-d_\ell}\end{pmatrix}}\,.
\end{align}
As a result, $\det(\Sigma_{\ell+1}^\frac{1}{2} \bar{\Sigma}_{T+1,\ell}^{-1}
  \Sigma_{\ell+1}^\frac{1}{2})=\det(I_{d_\ell} + \sigma_{\ell+1}^2\bar{\W}_\ell^\top \big(\Sigma_{\ell}^{-1}-  \Sigma_{\ell}^{-1}\bar{\Sigma}_{t, \ell-1}\Sigma_{\ell}^{-1}\big) \bar{\W}_\ell)$. This allows us to move the problem from a $d$-dimensional one to a $ d_\ell$-dimensional one. Then we use the inequality of arithmetic and geometric means and get that
  \begin{align}\label{eq:dts_new_term}
   \sum_{t=1}^T \log(1 &+ \sigma^{-2} X_t^\top \PP_{A_t, \ell} \bar{\Sigma}_{t,\ell} \PP_{A_t, \ell}^\top X_t) \leq \sigma_{\textsc{max}}^{2\ell} \Big( \log\det(\Sigma_{\ell+1}^{\frac{1}{2}}\bar{\Sigma}_{T+1,\ell}^{-1}\Sigma_{\ell+1}^{\frac{1}{2}})\Big)\,,\nonumber\\
   &= \sigma_{\textsc{max}}^{2\ell} \log\det(I_{d_\ell} + \sigma_{\ell+1}^2\bar{\W}_\ell^\top \big(\Sigma_{\ell}^{-1}-  \Sigma_{\ell}^{-1}\bar{\Sigma}_{t, \ell-1}\Sigma_{\ell}^{-1}\big) \bar{\W}_\ell)\,,\nonumber\\
 & \leq d_\ell \sigma_{\textsc{max}}^{2\ell}  \log\left(\frac{1}{d_\ell} \operatorname{Tr}(I_{d_\ell} + \sigma_{\ell+1}^2\bar{\W}_\ell^\top \big(\Sigma_{\ell}^{-1}-  \Sigma_{\ell}^{-1}\bar{\Sigma}_{t, \ell-1}\Sigma_{\ell}^{-1}\big) \bar{\W}_\ell)\right)\,,\nonumber\\
  & \leq d_\ell \sigma_{\textsc{max}}^{2\ell}  \log\left(1 + \frac{\sigma_{\ell+1}^2}{\sigma_{\ell}^2} \right)\,.
\end{align}
To get the last inequality, we use derivations similar to the ones we used in \cref{eq:dts_useful_derivations}. Finally, the desired result in obtained by replacing \cref{eq:dts_part_to_change_for_sparsity} by \cref{eq:dts_new_term} in the previous proof in \cref{subsec:dts_dts_proof_thm}.

\section{Additional Experiments}\label{app:exp}

\subsection{Swiss roll data}\label{app:swiss_roll_data}

\begin{figure}[H]
     \centering
     \begin{subfigure}[b]{0.32\textwidth}
         \centering
         \includegraphics[width=\textwidth]{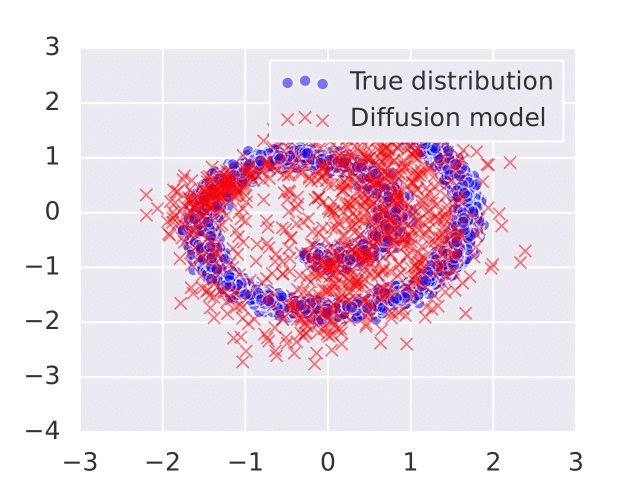}
         \caption{Diffusion pre-trained on $50$ samples from the Swiss roll dataset.}
     \end{subfigure}
     \hfill
     \begin{subfigure}[b]{0.32\textwidth}
         \centering
         \includegraphics[width=\textwidth]{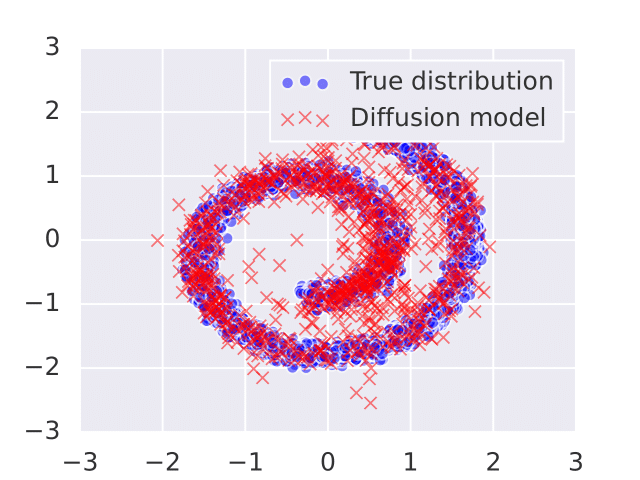}
         \caption{Diffusion pre-trained on $10^3$ samples from the Swiss roll dataset.}
     \end{subfigure}
     \hfill
     \begin{subfigure}[b]{0.32\textwidth}
         \centering
         \includegraphics[width=\textwidth]{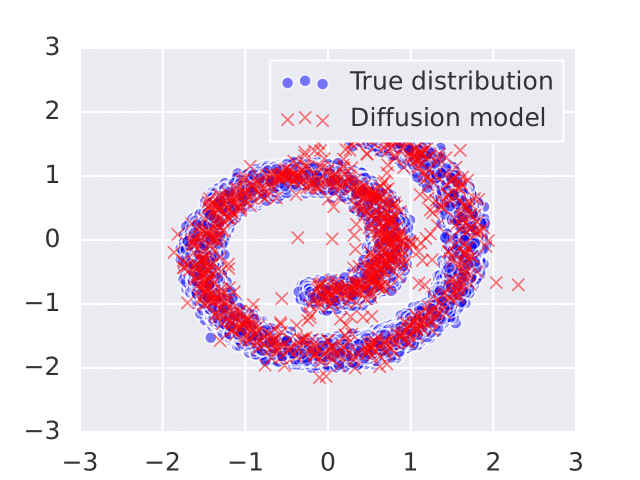}
         \caption{Diffusion pre-trained on $10^4$ samples from the Swiss roll dataset.}
     \end{subfigure}
    \caption{True distribution of action parameters (blue) vs. distribution of pre-trained diffusion model (red).}\label{fig:dts_swiss_roll_data}
\end{figure}

\cref{fig:dts_swiss_roll_data} shows samples from the Swiss roll data and samples from generated by the pre-trained diffusion model for different pre-training sample sizes.

\subsection{Diffusion models pre-training}\label{app:pretraining}

We used JAX for diffusion model pre-training, summarized as follows:
\begin{itemize}
    \item  \textbf{Parameterization:} Functions $f_{\ell}$ are parameterized with a fully connected 2-layer neural network (NN) with ReLU activation. The step $\ell$ is provided as input to capture the current sampling stage. Covariances are fixed (not learned) as $\Sigma_{\ell}=\sigma_{\ell}^2 I_d$ with $\sigma_{\ell}$ increasing with $\ell$.
    \item \textbf{Loss:} Offline data samples are progressively noised over steps $\ell \in[L]$, creating increasingly noisy versions of the data following a predefined noise schedule \citep{ho2020denoising}. The NN is trained to reverse this noise (i.e., denoise) by predicting the noise added at each step. The loss function measures the $L_2$ norm difference between the predicted and actual noise at each step, as explained in \citet{ho2020denoising}.
    \item \textbf{Optimization:} Adam optimizer with a $10^{-3}$ learning rate was used. The NN was trained for 20,000 epochs with a batch size of min(2048, pre-training sample size). We used CPUs for pre-training, which was efficient enough to conduct multiple ablation studies.
    \item \textbf{After pre-training:} The pre-trained diffusion model is used as a prior for \alg and compared to \texttt{LinTS} as the reference baseline. In our ablation study, we plot the cumulative regret of \texttt{LinTS} in the last round divided by that of \alg. A ratio greater than 1 indicates that \alg outperforms \texttt{LinTS}, with higher values representing a larger performance gap.
\end{itemize}

\subsection{Quality of our posterior approximation}\label{app:quality_posterior}

\begin{figure}[H]
     \centering
     \begin{subfigure}[b]{0.45\textwidth}
         \centering
         \includegraphics[width=\textwidth]{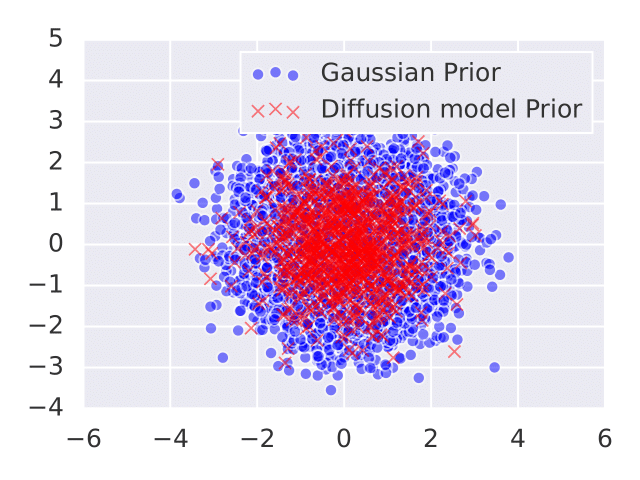}
         \caption{Gaussian distribution vs. diffusion model pre-trained on $10^3$ samples drawn from it.}\label{fig:dts_prior}
     \end{subfigure}
     \hfill
     \begin{subfigure}[b]{0.45\textwidth}
         \centering
         \includegraphics[width=\textwidth]{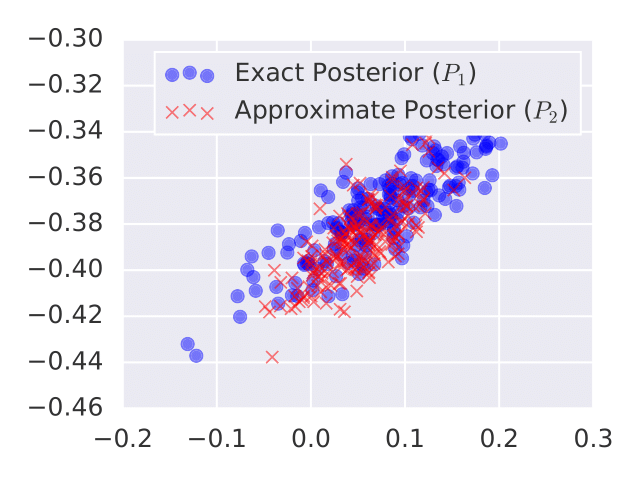}
         \caption{Exact posterior $P_1$ vs. approximate posterior $P_2$ after $T=100$ rounds of interactions.}\label{fig:dts_pos}
     \end{subfigure}     
    \caption{Assessing the quality of our posterior approximation.}
\end{figure}

To assess the quality of our posterior approximation, we consider the scenario where the true distribution of action parameters is $\mathcal{N}\left(0_d, I_d\right)$ with $d=2$ and rewards are linear. We pre-train a diffusion model using samples drawn from $\mathcal{N}\left(0_d, I_d\right)$. We then consider two priors: the true prior $\mathcal{N}\left(0_d, I_d\right)$ and the pre-trained diffusion model prior. This yields two posteriors:
\begin{itemize}
    \item $P_1$ : Uses $\mathcal{N}\left(0_d, I_d\right)$ as the prior. $P_1$ is an exact posterior since the prior is Gaussian and rewards are linear-Gaussian.
    \item $P_2$ : Uses the pre-trained diffusion model as the prior. $P_2$ is our approximate posterior.
\end{itemize}
The learned diffusion model prior matches the true Gaussian prior (as seen in \cref{fig:dts_prior}). Thus, if our approximation is accurate, their posteriors $P_1$ and $P_2$ should also be similar. This is observed in \cref{fig:dts_pos} where the approximate posterior $P_2$ nearly matches the exact posterior $P_1$.

\subsection{CIFAR Ablation}\label{app:cifar}
\textbf{CIFAR.} In \cref{fig:dts_effect_n} in \cref{subsec:dts_dts_effect_pretraining}, we showed that with only 10 pre-training samples, \alg outperforms \texttt{LinTS} on the Swiss-roll benchmark. 
We now extend this analysis to the vision dataset CIFAR \citep{krizhevsky2009learning} (similar results were obtained on MNIST \citep{mnist}). 
Our setting is similar to that in \citet{hong22deep} and we use \alg's variant that uses a single shared parameter $\theta \in \real^d$ (\cref{remark,remark_section}) because it is more suited for this setting. These additional ablations on CIFAR confirm that \alg consistently benefits from offline pre-training, even when the true prior is not a diffusion model. 
Specifically, we vary the percentage of offline data used to train the prior and compare against both \texttt{HierTS} and \texttt{LinTS}.

\begin{figure}[h]
\centering
\captionof{table}{Regret improvement  (\%) of \alg on CIFAR.}
\vspace{0.3em}
\begin{tabular}{lcc}
\toprule
\textbf{Offline Data (\%)} & \textbf{vs. \texttt{HierTS}} & \textbf{vs. \texttt{LinTS}} \\
\midrule
1\%  & 69.11\% & 87.74\% \\
5\%  & 79.56\% & 92.18\% \\
25\% & 80.65\% & 92.48\% \\
50\% & 81.67\% & 92.88\% \\
\bottomrule
\end{tabular}
\label{tab:offline_data_comparison}
\end{figure}

\subsection{Bound comparison}\label{app:bound_comparison}
Here, we compare our bound in \cref{thm:dts_regret} to bounds of \texttt{LinTS} from the literature. 
\begin{figure}[H]
     \centering
     \begin{subfigure}[b]{0.45\textwidth}
         \centering
         \includegraphics[width=\textwidth]{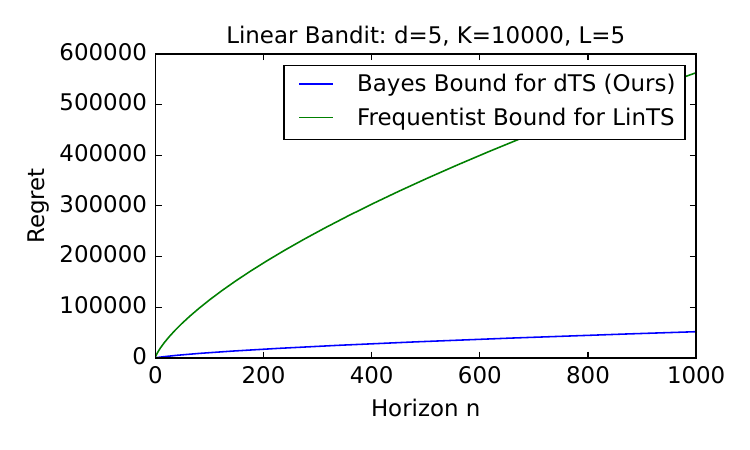}
         \caption{Our bound vs. the frequentist bound of \texttt{LinTS} in \citet{abeille17linear}.}
     \end{subfigure}
     \hfill
     \begin{subfigure}[b]{0.45\textwidth}
         \centering
         \includegraphics[width=\textwidth]{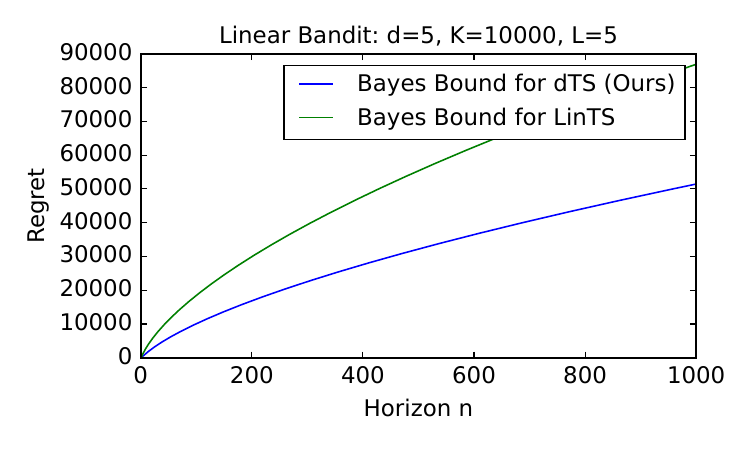}
         \caption{Our bound vs. the standard Bayesian bound of \texttt{LinTS}.}
     \end{subfigure}     
    \caption{Comparing our Bayesian regret bound of \dTS to the frequentist and Bayesian bounds of \texttt{LinTS}.}
\end{figure}

%% file: contents/sdm/appendix.tex
\section{Posterior Derivations Under Standard Priors}\label{app:sdm-posterior_derivations_standard}
Here we derive the posterior under the standard prior in \cref{eq:sdm-basic_model}. These are standard derivations and we present them here for the sake of completeness. But first, we state the following standard assumption that allows posterior derivations. 
\begin{assumption}[Independence]\label{ass:independence}
 $(X, A)$ is independent of $\theta$, and the $\theta_a$, for $a \in \cA$ are independent. 
\end{assumption}

\begin{proof}[Derivation of 
 $   p(\theta_a \mid \cD_n)$ for the standard prior in \cref{eq:sdm-basic_model}] We start by recalling the standard prior in \cref{eq:sdm-basic_model}
\begin{align}\label{eq:sdm-app_basic_model}
  \theta_a &\sim \cN(\mu_{a}, \Sigma_{a})  , &  \forall a \in \cA  , \\
    R \mid \theta, X, A &\sim \cN(\phi(X)^\top \theta_A , \sigma^2)  ,& \nonumber
\end{align}
where $\cN(\mu_{a}, \Sigma_{a})$ is the prior on the action parameter $\theta_a$. Let $\theta = (\theta_a)_{a \in \cA} \in \real^{dK}$, $\Sigma_{\cA} = {\rm diag}(\Sigma_a)_{a \in \cA} \in \real^{dK \times dK}$ and $\mu_{\cA} = (\mu_a)_{a\in \cA} \in \real^{dK}$. Also, let $u_a \in \set{0, 1}^K$ be the binary vector representing the action $a$. That is, $u_{a, a} = 1$ and $u_{a, a^\prime} = 0$ for all $a^\prime \neq a$. Then we can rewrite the model in \cref{eq:sdm-app_basic_model} as 
\begin{align}\label{eq:sdm-app_joint_basic_model}
  \theta &\sim \cN(\mu_{\cA}, \Sigma_{\cA})  , \\
    R \mid \theta, X, A &\sim \cN((u_A \otimes \phi(X))^\top \theta , \sigma^2)  . \nonumber
\end{align}
Then the \emph{joint} action posterior $p(\theta \mid \cD_n)$ decomposes as
\begin{align*}
    p(&\theta \mid \cD_n) = p(\theta \mid (X_i, A_i, R_i)_{i \in [n]}) \stackrel{(i)}{\propto} p((R_i)_{i \in [n]} \mid \theta, (X_i, A_i)_{i \in [n]}) p(\theta \mid  (X_i, A_i)_{i \in [n]})   ,  \\
    &\stackrel{(ii)}{=} p((R_i)_{i \in [n]} \mid \theta, (X_i, A_i)_{i \in [n]}) p(\theta) \stackrel{(iii)}{=} \prod_{i \in [n]}p(R_i \mid \theta, X_i, A_i) p(\theta)   ,  \\
    & \stackrel{(iv)}{=}  \prod_{i \in [n]} \cN(R_i; (u_{A_i} \otimes  \phi(X_i))^\top \theta, \sigma^2) \cN(\theta; \mu_{\cA} , \Sigma_{\cA})  ,\\
  &\stackrel{(v)}{\propto} \cN\left(\theta; \hat{\mu}_{\cA}, \left(\hat{\Lambda}_{\cA}\right)^{-1}\right)  .
\end{align*}
In $(i)$, we apply Bayes rule. In $(ii)$, we use that $\theta$ is independent of $(X, A)$, and $(iii)$ follows from the assumption that $R_i \mid \theta, X_i, A_i$ are independent. Finally, in $(iv)$, we replace the distribution by their Gaussian form, and in $(v)$, we set $\hat{\Lambda}_{\cA} = v\sum_{i =1}^n (u_{A_i}u_{A_i}^\top \otimes \phi(X_i) \phi(X_i)^\top) + \Lambda_{\cA}  ,$ and $  \hat{\mu}_{\cA}  = \hat{\Lambda}_{\cA}^{-1} \big( v \sum_{i =1}^n (u_{A_i} \otimes \phi(X_i)) R_i + \Lambda_{\cA} \mu_{\cA}\Big)$ where $v = \sigma^{-2}$ and $\Lambda_{\cA} = \Sigma_{\cA}^{-1} = \mathrm{diag}(\Sigma_a^{-1})_{a \in \cA}$. Now notice that $\hat{\Lambda}_{\cA} = {\rm diag}(\Sigma_a^{-1} + G_a)_{a \in \cA}$. Thus, $p(\theta \mid \cD_n) = \cN(\theta; \hat{\mu}_{\cA}, \hat{\Lambda}_{\cA}^{-1})$ where $\hat{\mu}_{\cA} = (\hat{\mu}_{a})_{a \in \cA}$ and $ \hat{\Lambda}_{\cA} = {\rm diag}(\hat{\Lambda}_{a})_{a \in \cA}  ,$ with
\begin{align*}
    &\hat{\Lambda}_{a} = \Sigma_a^{-1} + G_a  , &\hat{\Lambda}_{a}\hat{\mu}_{a} &=  \Sigma_a^{-1} \mu_a + B_a   .
\end{align*}
Since the covariance matrix of $p(\theta \mid \cD_n)$ is diagonal by block, we know that the marginals $\theta_a \mid \cD_n$ also have a Gaussian density  $p(\theta_a \mid \cD_n) = \cN(\theta_a; \hat{\mu}_{a}, \hat{\Sigma}_{a})$ where $\hat{\Sigma}_{a} = \hat{\Lambda}_{a}^{-1}$.

\end{proof}

\section{Posterior Derivations Under Structured Priors}\label{app:sdm-posterior_derivations}
Here we derive the posteriors under the structured prior in \cref{eq:sdm-contextual_gaussian_model}. Precisely, we derive the latent posterior density of $\psi \mid \cD_n$, the conditional posterior density of $\theta \mid \cD_n, \psi$. Then, we derive the marginal posterior $\theta \mid \cD_n$. Posterior derivations rely on the following assumption. 

\begin{assumption}[Structured Independence]\label{ass:indep_assumptions} \textit{(i)} $(X, A)$ is independent of $\psi$ and given $\psi$, $(X, A)$ is independent of $\theta$. \textit{(ii)} Given $\psi$, the $\theta_a,$ for all $a \in \cA$ are independent.  \end{assumption}

\subsection{Latent Posterior}\label{app:sdm-latent posterior-derivation}

\begin{proof}[Derivation of 
 $   p(\psi \mid \cD_n)$]
First, recall that our model in \cref{eq:sdm-contextual_gaussian_model} reads
\begin{align}
    \psi & \sim \cN(\mu, \Sigma)  , \nonumber\\ 
    \theta_a \mid  \psi& \sim \cN\Big( \W_a \psi ,     \Sigma_{a}\Big)  , & \forall a \in \cA  ,\nonumber\\ 
    R  \mid \psi, \theta, X, A & \sim \cN(\phi(X)^\top \theta_A ,  \sigma^2)  .
\end{align}
Then we first rewrite it as
\begin{align}\label{eq:sdm-model_rewritten_joint}
    \psi & \sim \cN(\mu, \Sigma)  , \nonumber \\ 
    \theta \mid  \psi& \sim \cN\Big( \W_{\cA} \psi ,     \Sigma_{\cA}\Big)  ,\nonumber\\ 
   R  \mid \psi, \theta, X, A  & \sim \cN((u_A \otimes \phi(X))^\top \theta ,  \sigma^2)  .
\end{align}
Then the latent posterior is
\begin{align*}
   p(\psi \mid (X_i, A_i, R_i)_{i \in [n]})  & \propto p((R_i)_{i \in [n]} \mid \psi, (X_i, A_i)_{i \in [n]})p(\psi \mid (X_i, A_i)_{i \in [n]})   ,\\
    &\stackrel{(i)}{=} p((R_i)_{i \in [n]} \mid \psi, (X_i, A_i)_{i \in [n]}) q(\psi)  ,\\
    &= \int_{\theta} p((R_i)_{i \in [n]}, \theta \mid \psi, (X_i, A_i)_{i \in [n]}) \dif \theta    q(\psi)  ,\\
    &= \int_{\theta} p((R_i)_{i \in [n]} \mid \psi, \theta, (X_i, A_i)_{i \in [n]}) p(\theta \mid \psi, (X_i, A_i)_{i \in [n]}) \dif \theta    q(\psi)  ,\\
    &\stackrel{(ii)}{=} \int_{\theta} p((R_i)_{i \in [n]} \mid \psi, \theta, (X_i, A_i)_{i \in [n]}) p(\theta \mid \psi) \dif \theta    q(\psi)  ,\\
\end{align*}
In $(i)$, we use that $(X, A)$ is independent of $\psi$, which follows from \cref{ass:indep_assumptions}. Similarly, in $(ii)$, we use that $\theta$ is conditionally independent of $(X, A)$ given $\psi$. Now we know that given $\theta$, $R_i \mid X_i, A_i$ are i.i.d. and hence $p((R_i)_{i \in [n]} \mid \psi, \theta, (X_i, A_i)_{i \in [n]}) = \prod_{a \in \cA} \LL_a(\theta_a)$. Moreover, $\theta_a$ for $a \in \cA$ are conditionally independent given $\psi$. Thus $p(\theta  \mid \psi) = \prod_{a \in \cA} p_a\left(\theta_a; f_a(\psi) \right)$, where we also used that $\theta_a \mid \psi \sim  p_a\left(\cdot; f_a(\psi) \right)$. This leads to 
\begin{align*}
     p(\psi \mid (X_i, A_i, R_i)_{i \in [n]})  & \propto \int_{\theta}  \prod_{a \in \cA} \LL_a(\theta_a) p_a\left(\theta_a ; f_a(\psi) \right) \dif \theta    q(\psi)  ,\\
    &\stackrel{(i)}{=} \prod_{a \in \cA} \int_{\theta_a} \LL_a(\theta_a) \cN \left(\theta_a ; \W_a \psi, \Sigma_a\right) \dif \theta_a \cN(\psi;\mu, \Sigma)  , \nonumber\\
   & \stackrel{(ii)}{=} \prod_{a \in \cA} \int_{\theta_a}  \Big(\prod_{i \in I_a} \cN(R_i; \phi(X_i)^\top \theta_a, \sigma^2)\Big) \cN \left(\theta_a ; \W_a \psi, \Sigma_a\right) \dif \theta_a \cN(\psi;\mu, \Sigma)  .
\end{align*}
In $(i)$, we notice that $\theta=(\theta_a)_{a \in \cA}$ and apply Fubini's Theorem. In $(ii)$, we let $I_a = \{i \in [n]; A_i=a\}$ as the rounds where action $a$ appears in the sample set $\cD_n$.   Now let $h_a(\psi) = \int_{\theta_a} \left(\prod_{i \in I_a} \cN(R_i; \phi(X_i)^\top \theta_a, \sigma^2)\right) \cN \left(\theta_a; \W_a \psi, \Sigma_a\right) \dif \theta_a$. Then we have that 
\begin{align}\label{eq:sdm-helper}
    p(\psi \mid \cD_n)  & \propto \prod_{a \in \cA} h_a(\psi) \cN(\psi;\mu, \Sigma)  .
\end{align}
We start by computing $h_a$. To reduce clutter, let $v = \sigma^{-2}$ and $\Lambda_a = \Sigma_a^{-1}$. Then we compute $h_a$ as 
\begin{align*}
& h_a(\psi)  =  \int_{\theta_a} \left(\prod_{i \in I_a} \cN(R_i; \phi(X_i)^\top \theta_a, \sigma^2)\right)
  \cN(\theta_a; \W_a \psi, \Sigma_a) \dif\theta_a  , \\
  & \propto \int_{\theta_a} \exp\left[
  - \frac{1}{2} v \sum_{i \in I_a} (R_i - \phi(X_i)^\top \theta_a)^2 -
  \frac{1}{2} (\theta_a- \W_a \psi)^\top \Lambda_a (\theta_a- \W_a \psi)\right] \dif \theta_a  ,\\
  & = \int_{\theta_a} \exp\Big[- \frac{1}{2}
  \Big(v \sum_{i \in I_a} (R_i^2 - 2 R_i \theta_a^\top \phi(X_i) + (\theta_a^\top \phi(X_i))^2) +
  \theta_a^\top \Lambda_a \theta_a- 2 \theta_a^\top \Lambda_a \W_a \psi \\& \hspace{7cm} + (\W_a \psi)^\top \Lambda_a (\W_a \psi) \Big)\Big] \dif \theta_a  ,\\
  & \propto \int_{\theta_a} \exp\Big[- \frac{1}{2}
  \Big(\theta_a^\top \left( v \sum_{i \in I_a} \phi(X_i) \phi(X_i)^\top + \Lambda_a \right) \theta_a- 2 \theta_a^\top \left(v \sum_{i \in I_a} R_i \phi(X_i) + \Lambda_a \W_a \psi\right) \\ & \hspace{7cm} + (\W_a \psi)^\top \Lambda_a (\W_a \psi)\Big) \Big] \dif \theta_a  .
 \end{align*}
Now recall that $G_a = v \sum_{i \in I_a} \phi(X_i) \phi(X_i)^\top$ and $B_a = v \sum_{i \in I_a} R_i \phi(X_i)$ and let $V_a = \left(G_a + \Lambda_a\right)^{-1}$, $U_a = V_a^{-1}$, and $\beta_a = V_a (B_a  + \Lambda_a \W_a \psi)$. Then have that $U_a V_a = V_a U_a = I_{d}  ,$ and thus
 \begin{align*}
  h_a(\psi) & \propto \int_{\theta_a} \exp\left[- \frac{1}{2}
  \left(\theta_a^\top U_a \theta_a- 2 \theta_a^\top U_a V_a\left(B_a + \Lambda_a \W_a \psi\right) + (\W_a \psi)^\top \Lambda_a (\W_a \psi)\right) \right] \dif \theta_a  ,\\
  & = \int_{\theta_a} \exp\left[- \frac{1}{2}
  \left(\theta_a^\top U_a \theta_a- 2 \theta_a^\top U_a \beta_a +  (\W_a \psi)^\top \Lambda_a (\W_a \psi)\right) \right] \dif \theta_a  ,\\
  & = \int_{\theta_a} \exp\left[- \frac{1}{2}
  \left( (\theta_a- \beta_a)^\top U_a (\theta_a- \beta_a) - \beta_a^\top U_a \beta_a + (\W_a \psi)^\top \Lambda_a (\W_a \psi)\right) \right] \dif \theta_a  , \\
  & \propto  \exp\left[- \frac{1}{2}
  \left( - \beta_a^\top U_a \beta_a + (\W_a \psi)^\top \Lambda_a (\W_a \psi)\right) \right]  ,\\
  & =  \exp\left[- \frac{1}{2}
  \left( -  \left(B_a + \Lambda_a \W_a \psi\right)^\top  V_a \left(B_a + \Lambda_a \W_a \psi\right) + (\W_a \psi)^\top \Lambda_a (\W_a \psi)\right) \right]   ,\\
  & \propto  \exp\left[- \frac{1}{2}
  \left( \psi^{\top} \W_a^{\top} \left( \Lambda_a - \Lambda_a V_a \Lambda_a \right) \W_a \psi - 2 \psi^\top \left(\W_a^\top \Lambda_a V_a B_a\right) \right) \right]   ,\\
    & \propto  \exp\left[- \frac{1}{2}
  \left( \psi^{\top} \bar{\Lambda}_a \psi - 2 \psi^\top \bar{\Lambda}_a \bar{\mu}_{a} \right) \right]   ,
\end{align*}
where 
\begin{align}\label{eq:sdm-latent_posterior_formulas}
      \bar{\Lambda}_a & = \W_a^{\top} \left( \Lambda_a - \Lambda_a V_a \Lambda_a \right)\W_a = \W_a^{\top} \left( \Sigma_a^{-1} - \Sigma_a^{-1} (G_a + \Sigma_a^{-1})^{-1} \Sigma_a^{-1} \right)\W_a  , \nonumber \\
    \bar{\Lambda}_a \bar{\mu}_{a} & =  \W_a^\top \Lambda_a V_a B_a =   \W_a^\top \Sigma_a^{-1} (G_a + \Sigma_a^{-1})^{-1}  B_a  .
\end{align} 
However, we know from \cref{eq:sdm-helper} that $p(\psi \mid \cD_n)  \propto \prod_{a \in \cA} h_a(\psi) \cN(\psi;\mu, \Sigma)$. But $h_a(\psi)$ is proportional to $ \exp\left[- \frac{1}{2}
  \left( \psi^{\top} \bar{\Lambda}_a \psi - 2 \psi^\top \bar{\Lambda}_a \bar{\mu}_{a} \right) \right]$ for any $a$. Thus $ p(\psi \mid \cD_n)$ can be seen as the product of $K+1$ Gaussian kernels. Thus, $ p(\psi \mid \cD_n)$ is a multivariate Gaussian distribution $\cN(\bar{\mu}, \bar{\Sigma})$, with
\begin{align}\label{eq:sdm-without_woodbury}
    \bar{\Sigma}^{-1} &=  \Sigma^{-1} + \sum_{a \in \cA}\bar{\Lambda}_a = \Sigma^{-1} + \sum_{a \in \cA} \W_a^{\top} \left( \Sigma_a^{-1} - \Sigma_a^{-1} (G_a + \Sigma_a^{-1})^{-1} \Sigma_a^{-1} \right)\W_a  , \\
   \bar{\Sigma}^{-1}\bar{\mu} &=  \Sigma^{-1} \mu +\sum_{a \in \cA}\bar{\Lambda}_a\bar{\mu}_{a} = \Sigma^{-1} \mu +\sum_{a \in \cA}   \W_a^\top \Sigma_a^{-1} (G_a + \Sigma_a^{-1})^{-1}  B_a  .
\end{align}
\end{proof}

\subsection{Conditional Posterior}\label{app:sdm-conditional-posterior-derivation}

\begin{proof}[Derivation of 
 $   p(\theta_a \mid \psi, \cD_n)$]
Let $v
  = \sigma^{-2}  , \quad
  \Lambda_a
  = \Sigma_a^{-1}  .$ We consider the model rewritten in \cref{eq:sdm-model_rewritten_joint},  then the \emph{joint} conditional action posterior $p(\theta \mid \psi, \cD_n)$ decomposes as
\begin{align*}
   & p(\theta \mid \psi,  \cD_n) = p(\theta \mid \psi, (X_i, A_i, R_i)_{i \in [n]}) \stackrel{(i)}{\propto} p((R_i)_{i \in [n]} \mid \theta,  \psi, (X_i, A_i)_{i \in [n]}) p(\theta \mid \psi,  (X_i, A_i)_{i \in [n]})   ,  \\
    &\stackrel{(ii)}{=} p((R_i)_{i \in [n]} \mid \theta, (X_i, A_i)_{i \in [n]}) p(\theta \mid \psi) \stackrel{(iii)}{=} \prod_{i \in [n]}p(R_i \mid \theta, X_i, A_i) p(\theta \mid \psi)   ,  \\
    & \stackrel{(iv)}{=}  \prod_{i \in [n]} \cN(R_i; (u_{A_i} \otimes \phi(X_i))^\top \theta, \sigma^2) \cN(\theta; \W_{\cA} \psi , \Sigma_{\cA})  ,\\
    & = \exp\Big[- \frac{1}{2}
  \Big(v \sum_{i =1}^n (R_i^2 - 2 R_i (u_{A_i} \otimes \phi(X_i))^\top \theta + ((u_{A_i} \otimes \phi(X_i))^\top \theta)^2) + \theta^\top \Lambda_{\cA} \theta - 2 \theta^\top \Lambda_{\cA} \W_{\cA} \psi\\ 
  & \hspace{9cm} + \psi^\top  \W_{\cA} ^\top \Lambda_{\cA} \W_{\cA} \psi\Big)\Big]   ,\\
  &\propto \exp\Big[- \frac{1}{2}
  \Big( \theta^\top(v\sum_{i =1}^n (u_{A_i} \otimes \phi(X_i)) (u_{A_i} \otimes \phi(X_i))^\top + \Lambda_{\cA}) \theta \\ 
  & \hspace{7cm} -2 \theta^\top \left(v \sum_{i =1}^n (u_{A_i} \otimes \phi(X_i)) R_i + \Lambda_{\cA} \W_{\cA} \psi \right)\Big)\Big]  ,\\
    &= \exp\Big[- \frac{1}{2}
  \Big( \theta^\top(v\sum_{i =1}^n (u_{A_i}u_{A_i}^\top \otimes \phi(X_i) \phi(X_i)^\top)+ \Lambda_{\cA}) \theta - \\ 
  & \hspace{7cm}2 \theta^\top \left(v \sum_{i =1}^n (u_{A_i} \otimes \phi(X_i)) R_i + \Lambda_{\cA} \W_{\cA} \psi \right)\Big)\Big]  ,\\
  &\stackrel{(v)}{\propto} \cN\left(\theta; \tilde{\mu}_{\cA}, \left(\tilde{\Lambda}_{\cA}\right)^{-1}\right)  ,
\end{align*}
where we use Bayes rule in $(i)$, $(ii)$ uses two assumptions. First, Given $\theta, X, A$, $R$ is independent of $\psi$. Second, given $\psi$, $\theta$ is independent of $(X, A)$. Moreover, $(iii)$ follows from the assumption that $R_i \mid \theta, X_i, A_i$ are independent. Finally, in $iv$, we replace the distribution by their Gaussian form, and in $(v)$, we set $\tilde{\Lambda}_{\cA} = v\sum_{i =1}^n (u_{A_i}u_{A_i}^\top \otimes \phi(X_i) \phi(X_i)^\top) + \Lambda_{\cA}  ,$ and $  \tilde{\mu}_{\cA} = \tilde{\Lambda}_{\cA}^{-1} \big( v \sum_{i =1}^n (u_{A_i} \otimes \phi(X_i)) R_i + \Lambda_{\cA} \W_{\cA} \psi\Big)$, where $\Lambda_{\cA} = \Sigma_{\cA}^{-1} = \mathrm{diag}(\Sigma_a^{-1})_{a \in \cA}$. Now notice that $\tilde{\Lambda}_{\cA} = {\rm diag}(\Sigma_a^{-1} + G_a)_{a \in \cA}$. Thus, $p(\theta \mid \psi, \cD_n) = \cN(\theta; \tilde{\mu}_{\cA}, \tilde{\Lambda}_{\cA}^{-1})$ where $\tilde{\mu}_{\cA} = (\tilde{\mu}_{a})_{a \in \cA}$ and $ \tilde{\Lambda}_{\cA} = {\rm diag}(\tilde{\Lambda}_{a})_{a \in \cA}  ,$ with
\begin{align*}
    \tilde{\Lambda}_{a} &= \Sigma_a^{-1} + G_a  , \nonumber\\
      \tilde{\Lambda}_{a}\tilde{\mu}_{a} &=  \Sigma_a^{-1} \W_{a} \psi + B_a   .
\end{align*}
The covariance matrix of $p(\theta \mid \psi, \cD_n)$ is diagonal by block. Thus $\theta_a \mid \psi, \cD_n$ for $a \in \cA$ are independent and have a Gaussian density  $p(\theta_a \mid \psi, \cD_n) = \cN(\theta_a; \tilde{\mu}_{a}, \tilde{\Sigma}_{a})$ where $\tilde{\Sigma}_{a} = \tilde{\Lambda}_{a}^{-1}$.
\end{proof}

\subsection{Action Posterior}\label{app:sdm-marginal-action-posterior}
\begin{proof}[Derivation of 
 $   p(\theta_a \mid \cD_n)$]
We know that $\theta_a \mid \cD_n, \psi \sim \cN(\tilde{\mu}_a, \tilde{\Sigma}_a)$ and $\psi \mid \cD_n \sim \cN(\bar{\mu}, \bar{\Sigma})$. Thus the posterior density of $\theta_a \mid \cD_n$ is also Gaussian since Gaussianity is preserved after marginalization \citep{koller09probabilistic}. We let $\theta_a \mid \cD_n \sim \cN(\hat{\mu}_a, \hat{\Sigma}_a)$. Then, we can compute $\hat{\mu}_a$ and $ \hat{\Sigma}_a$ using the total expectation and total covariance decompositions. Let $\Lambda_a =  \Sigma_a^{-1}$. Then we have that
\begin{align*}
    \tilde{\Sigma}_a &= \left( G_a + \Lambda_a \right)^{-1} \\
    \condE{\theta_a}{\psi, \cD_n} & = \tilde{\Sigma}_a \left( B_a  + \Lambda_a \W_a \psi \right)
\end{align*}
First, given $\cD_n$, $\tilde{\Sigma}_a =\left( G_a + \Lambda_a \right)^{-1}$ and $B_a$ are constant (do not depend on $\psi$). Thus 
\begin{align*}
    \hat{\mu}_a = \condE{\theta_a}{ \cD_n}= \E{}{\condE{\theta_a}{\psi, \cD_n} \mid \cD_n} &= \E{\psi \sim \cN(\bar \mu, \bar \Sigma)}{\tilde{\Sigma}_a \left( B_a  + \Lambda_a \W_a \psi \right)}\\
    &= \tilde{\Sigma}_a \left( B_a  + \Lambda_a \W_a \E{\psi \sim \cN(\bar \mu, \bar \Sigma)}{\psi} \right)  ,\\
    &= \tilde{\Sigma}_a \left( B_a  + \Lambda_a \W_a \bar \mu \right)  .
\end{align*}
This concludes the computation of $\hat{\mu}_a$. Similarly, given $\cD_n$, $\tilde{\Sigma}_a =\left( G_a + \Lambda_a \right)^{-1}$ and $B_a$ are constant (do not depend on $\psi$), yields two things. First,
\begin{align*}   \condE{\condcov{\theta_a}{\psi, \cD_n}}{\cD_n} = \condE{\tilde{\Sigma}_a}{\cD_n} = \tilde{\Sigma}_a  .
\end{align*}
Second,
\begin{align*}
    \condcov{\condE{\theta_a}{\psi, \cD_n}}{\cD_n} &= \condcov{\tilde{\Sigma}_a\Lambda_a \W_a \psi}{\cD_n}\\
    &= \tilde{\Sigma}_a \Lambda_a \W_a \condcov{\psi}{\cD_n} \W_a^\top \Lambda_a \tilde{\Sigma}_a\\
    &= \tilde{\Sigma}_a \Lambda_a \W_a \bar{\Sigma} \W_a^\top \Lambda_a \tilde{\Sigma}_a  .
\end{align*}
Finally, the total covariance decomposition \citep{weiss05probability} yields that 
\begin{align*}
 \hat{\Sigma}_a =   \condcov{\theta_a}{\cD_n} &= \condE{\condcov{\theta_a}{\psi, \cD_n}}{\cD_n}  +  \condcov{\condE{\theta_a}{\psi, \cD_n}}{\cD_n}\\
 &= \tilde{\Sigma}_a+ \tilde{\Sigma}_a \Lambda_a \W_a \bar{\Sigma} \W_a^\top \Lambda_a \tilde{\Sigma}_a  .
\end{align*}
This concludes the proof.

\end{proof}

\section{Proofs}\label{sec:sdm-ope_proofs}

\subsection{Main Result}\label{subsec:sdm-main_result_app}
In this section, we prove \cref{thm:main_thm_1}. Recall that we make the following well-specified prior assumption.

\begin{assumption}[Well-specified priors]\label{assum:well-specified_app}
Action parameters $\theta_{*, a}$ and rewards are drawn from \cref{eq:sdm-contextual_gaussian_model}. 
\end{assumption}

\begin{assumption}[Diagonal covariances for simplicity]\label{assum:isotropic_app}
We assume $\Sigma_a = \sigma_0^2 I_d$, $\Sigma = \tau^2 I_{d'}$, $\normw{\phi(x)}{2} \leq 1$, and the matrices $\W_a$ are normalized such that $\lambda_1(\W_a \W_a^\top) = \lambda_d(\W_a \W_a^\top) = 1$.
\end{assumption}

\begin{proof}
First, given $x \in \cX$, by definition of the optimal policy, we know that it is deterministic. That is, there exists $a_{x, \theta_*} \in [K]$ such that $\pi_*(a_{x, \theta_*} \mid x) = 1$. To simplify the notation and since $\pi_*$ is deterministic, we let $\pi_*(x) =a_{x, \theta_*} $. Also, we know that the greedy policy is deterministic in $\hat{a}_{x} = \argmax_{b \in \cA} \hat{r}(x, b)$. That is $\hat{\pi}_{\textsc{g}}(\hat{a}_{x} \mid x)=1$. Similarly, we let $\hat{\pi}_{\textsc{g}}(x) = \hat{a}_{x}$. Moreover, we let $\Phi(x, a) = e_a \otimes \phi(x) \in \real^{dK}$ where $e_a \in \real^K$ is the indicator vector of action $a$, such that $e_{a, b}=0$ for any $b \in \cA/\{a\}$ and $e_{a, a}=1$. Also, recall that $\hat{\mu}=(\hat{\mu}_a)_{a\in \cA}$ is the concatenation of the posterior means. 
\begin{align*}
      \textsc{Bso}(\hat{\pi}_{\textsc{g}})&=   \E{}{V(\pi_*;\theta_*) - V(\hat{\pi}_{\textsc{g}};\theta_*)}  ,\\
      &= \E{}{r(X,  \pi_*(X); \theta_*) - r(X,  \hat{\pi}_{\textsc{g}}(X); \theta_*)  }  ,\\
      &= \E{}{r(X,  \pi_*(X); \theta_*) - r(X,  \hat{\pi}_{\textsc{g}}(X); \hat{\mu}) + r(X,  \hat{\pi}_{\textsc{g}}(X); \hat{\mu}) - r(X,  \hat{\pi}_{\textsc{g}}(X); \theta_*)  }  ,\\
      &\leq \E{}{r(X,  \pi_*(X); \theta_*) - r(X,   \pi_*(X); \hat{\mu}) + r(X,  \hat{\pi}_{\textsc{g}}(X); \hat{\mu}) - r(X,  \hat{\pi}_{\textsc{g}}(X); \theta_*)  }  ,\\
        &\leq \E{}{r(X,  \pi_*(X); \theta_*) - r(X,   \pi_*(X); \hat{\mu})} + \E{}{ r(X,  \hat{\pi}_{\textsc{g}}(X); \hat{\mu}) - r(X,  \hat{\pi}_{\textsc{g}}(X); \theta_*)}  .
\end{align*}
Now we start by proving that $\E{}{ r(X,  \hat{\pi}_{\textsc{g}}(X); \hat{\mu}) - r(X,  \hat{\pi}_{\textsc{g}}(X); \theta_*)}=0$. This is achieved as follows
\begin{align*}
   \E{}{ r(X,  \hat{\pi}_{\textsc{g}}(X); \hat{\mu}) - r(X,  \hat{\pi}_{\textsc{g}}(X); \theta_*)} &= \E{}{\E{}{ r(X,  \hat{\pi}_{\textsc{g}}(X); \hat{\mu}) - r(X,  \hat{\pi}_{\textsc{g}}(X); \theta_*) \mid X, \cD_n}}   ,\\
   &= \E{}{\E{}{ \phi(X)^\top \hat{\mu}_{\hat{\pi}_{\textsc{g}}(X)} - \phi(X)^\top \theta_{*, \hat{\pi}_{\textsc{g}}(X)} \mid X, \cD_n}}   ,\\
    &\stackrel{(i)}{=}  \E{}{\E{}{ \Phi(X, \hat{\pi}_{\textsc{g}}(X))^\top \hat{\mu} - \Phi(X, \hat{\pi}_{\textsc{g}}(X))^\top \theta_{*} \mid X, \cD_n}}   ,\\
    &\stackrel{(ii)}{=}  \E{}{\Phi(X, \hat{\pi}_{\textsc{g}}(X))^\top \E{}{ \hat{\mu} - \theta_{*} \mid X, \cD_n}}   ,\\
    &\stackrel{(iii)}{=}  \E{}{\Phi(X, \hat{\pi}_{\textsc{g}}(X))^\top ( \hat{\mu} - \E{}{\theta_{*} \mid X, \cD_n})}   ,\\
    &\stackrel{(iv)}{=}  0   .
\end{align*}
In $(i)$, we used that by definition of $\Phi(x, a) = e_a \otimes \phi(X) \in \real^{dK}$, we have $\phi(x)^\top \hat{\mu}_a = \Phi(x, a)^\top \hat{\mu}$ for any $(x, a)$, and the same holds for $\theta_*$. In $(ii)$, we used that $\Phi(X, \hat{\pi}_{\textsc{g}}(X))$ is deterministic given $X$ and $\cD_n$. In $(iii)$, we used that $\hat{\mu}$ is deterministic given $X$ and $\cD_n$. Finally, in $(iv)$, we used that $\E{}{\theta_{*} \mid X, \cD_n} = \E{}{\theta_{*} \mid \cD_n} = \hat{\mu}$, which follows from the assumption that $\theta_*$ does not depend on $X$ and the assumption that $\theta_*$ is drawn from the prior, and hence when conditioned on $\cD_n$, it is drawn from the posterior whose mean is $\hat{\mu}$. Therefore, $\E{}{ r(X,  \hat{\pi}_{\textsc{g}}(X); \hat{\mu}) - r(X,  \hat{\pi}_{\textsc{g}}(X); \theta_*)}=0$ which leads to 
\begin{align*}
      \textsc{Bso}(\hat{\pi}_{\textsc{g}})& \leq \E{}{r(X,  \pi_*(X); \theta_*) - r(X,   \pi_*(X); \hat{\mu})}  .
\end{align*}

Now let $\delta\in(0,1)$ and define the \emph{joint parameter event}
\begin{align*}
\mathcal E_\alpha
=
\Big\{
\forall a\in\cA:\ \|\theta_{*,a}-\hat\mu_a\|_{\hat\Sigma_a^{-1}}\le \alpha
\Big\}.
\end{align*}
Recall $Z_a = r(X,a;\theta_*)-r(X,a;\hat\mu)=\phi(X)^\top(\theta_{*,a}-\hat\mu_a)$. By Cauchy-Schwarz, on $\mathcal E_\alpha$ we have for all $a$,
$$
|Z_a|
=
|\phi(X)^\top(\theta_{*,a}-\hat\mu_a)|
\le
\|\phi(X)\|_{\hat\Sigma_a}  \|\theta_{*,a}-\hat\mu_a\|_{\hat\Sigma_a^{-1}}
\le
\alpha\|\phi(X)\|_{\hat\Sigma_a},
$$
hence in particular $|Z_{\pi_*(X)}|\le \alpha\|\phi(X)\|_{\hat\Sigma_{\pi_*(X)}}$ on $\mathcal E_\alpha$.
Therefore,
\begin{align*}
\textsc{Bso}(\hat\pi_{\textsc{g}})
&\le \E{}{|Z_{\pi_*(X)}|}\\
&= \E{}{|Z_{\pi_*(X)}|\I{\mathcal E_\alpha}}
 + \E{}{|Z_{\pi_*(X)}|\I{\bar{\mathcal E}_\alpha}}\\
&\le \alpha  \E{}{\|\phi(X)\|_{\hat\Sigma_{\pi_*(X)}}}
 + \E{}{|Z_{\pi_*(X)}|\I{\bar{\mathcal E}_\alpha}}.
\end{align*}
We now bound $\mathbb{P}(\bar{\mathcal E}_\alpha\mid \cD_n)$.
Under the well-specified Bayes assumption, conditional on $\cD_n$ each marginal posterior satisfies
$\theta_{*,a}\mid\cD_n\sim\cN(\hat\mu_a,\hat\Sigma_a)$. This means that $\theta_{*, a} - \hat{\mu}_a \mid \cD_n \sim \cN(0, \hat{\Sigma}_a)$. Thus, $\hat{\Sigma}_a^{-\frac{1}{2}}(\theta_{*, a} - \hat{\mu}_a) \sim \cN(0, I_d)$. But notice that $\left\|\theta_{*, a}-\hat{\mu}_a\right\|_{\hat{\Sigma}_a^{-1}} = \|\hat{\Sigma}_a^{-\frac{1}{2}}(\theta_{*, a} - \hat{\mu}_a)\|$. Thus we apply \citet[Lemma~1]{laurent2000adaptive} and get that 
\begin{align*}
        \condprob{\left\|\theta_{*, a}-\hat{\mu}_a\right\|_{\hat{\Sigma}_a^{-1}} \leq \alpha}{\cD_n} \geq 1- \frac{\delta}{K}  ,
\end{align*}
where $\alpha = \sqrt{d + 2 \sqrt{d \log \frac{K}{\delta}} + 2 \log \frac{K}{\delta}}$. This means that for every $a$, $\mathbb{P}(\|\theta_{*,a}-\hat\mu_a\|_{\hat\Sigma_a^{-1}}>\alpha\mid\cD_n)\le \delta/K$,
and by a union bound,
\begin{align}
\mathbb{P}(\bar{\mathcal E}_\alpha\mid \cD_n)\le \delta.
\label{eq:joint_event_prob}
\end{align}
and therefore $\mathbb{P}(\bar{\mathcal E}_\alpha) = \mathbb{E}[\mathbb{P}(\bar{\mathcal E}_\alpha\mid \cD_n)]\le \delta$. Finally, control the bad-event contribution by Cauchy-Schwarz:
\begin{align}
\E{}{|Z_{\pi_*(X)}|\I{\bar{\mathcal E}_\alpha}}
\le
\sqrt{\E{}{Z_{\pi_*(X)}^2}}  \sqrt{\mathbb{P}(\bar{\mathcal E}_\alpha)}
\le
\sqrt{\E{}{Z_{\pi_*(X)}^2}}  \sqrt{\delta}.
\label{eq:bad_event_cs}
\end{align}
Putting the pieces together gives
\begin{align}
\textsc{Bso}(\hat\pi_{\textsc{g}})
\le
\alpha  \E{}{\|\phi(X)\|_{\hat\Sigma_{\pi_*(X)}}}
+
\sqrt{\delta}  \sqrt{\E{}{Z_{\pi_*(X)}^2}},
\label{eq:bso_joint_event_bound}
\end{align}
with $\alpha^2 = d + 2\sqrt{d\log(K/\delta)} + 2\log(K/\delta)$.

We now upper bound $\E{}{Z_{\pi_*(X)}^2}$ in \eqref{eq:bso_joint_event_bound}. We have that
\begin{align}
Z_{\pi_*(X)}^2 \le \max_{a\in\cA} Z_a^2,
\qquad\text{hence}\qquad
\E{}{Z_{\pi_*(X)}^2} \le \E{}{\max_{a\in\cA} Z_a^2}.
\label{eq:zpistar_to_max}
\end{align}
Fix $(X,\cD_n)$. Under the well-specified Bayes assumption, the conditional joint posterior
$(\theta_{*,a})_{a\in\cA}\mid \cD_n$ is Gaussian, and each marginal satisfies
$\theta_{*,a}\mid \cD_n\sim \cN(\hat\mu_a,\hat\Sigma_a)$ (with $\hat\Sigma_a$ as derived in
\cref{app:sdm-marginal-action-posterior}). Therefore for each fixed $a$,
\begin{align*}
Z_a \mid X,\cD_n \sim \cN \big(0,   s_a^2\big),
\qquad
s_a^2=\|\phi(X)\|_{\hat\Sigma_a}^2.
\end{align*}
Let $s_{\max}^2=\max_{a\in\cA} s_a^2$. Then for any $t\ge 0$, by a union bound and the
Gaussian tail bound,
\begin{align*}
\mathbb{P} \left(\max_{a\in\cA} |Z_a| \ge t   \middle|   X,\cD_n\right)
&\le \sum_{a\in\cA} \mathbb{P} \left(|Z_a|\ge t   \middle|   X,\cD_n\right)
\le 2K \exp \left(-\frac{t^2}{2s_{\max}^2}\right).
\end{align*}
Using the identity $\E{}{W^2}=\int_0^\infty \mathbb{P}(W^2\ge u)  \mathrm{d} u$ for $W\ge 0$ and setting
$W=\max_{a}|Z_a|$, we obtain
\begin{align*}
\E{}{\max_{a\in\cA} Z_a^2   \middle|   X,\cD_n}
&= \int_0^\infty \mathbb{P} \left(\max_{a\in\cA} |Z_a| \ge \sqrt{u}  \middle|  X,\cD_n\right)\mathrm{d} u \\
&\le \int_0^\infty \min \left\{1,\ 2K \exp \left(-\frac{u}{2s_{\max}^2}\right)\right\}\mathrm{d} u \\
&= \int_0^{2s_{\max}^2\log(2K)} 1  \mathrm{d} u
   + \int_{2s_{\max}^2\log(2K)}^\infty 2K \exp \left(-\frac{u}{2s_{\max}^2}\right)\mathrm{d} u \\
&= 2s_{\max}^2\log(2K) + 2s_{\max}^2
= \big(2\log(2K)+2\big)  s_{\max}^2.
\end{align*}
Taking expectation over $(X,\cD_n)$ yields
\begin{align}
\E{}{\max_{a\in\cA} Z_a^2}
\le \big(2\log(2K)+2\big)  \E{}{\max_{a\in\cA}\|\phi(X)\|_{\hat\Sigma_a}^2}.
\label{eq:maxZ2_bound}
\end{align}
Combining \eqref{eq:zpistar_to_max} and \eqref{eq:maxZ2_bound} gives
\begin{align}
\E{}{Z_{\pi_*(X)}^2}
\le \big(2\log(2K)+2\big)  \E{}{\max_{a\in\cA}\|\phi(X)\|_{\hat\Sigma_a}^2}.
\label{eq:zpistar2_bound}
\end{align}
Using the assumption that $\|\phi(X)\|_2\le 1$ yields that 
$\|\phi(X)\|_{\hat\Sigma_a}^2 \le   \lambda_1(\hat\Sigma_a)$, hence
\begin{align}
\E{}{Z_{\pi_*(X)}^2}
\le \big(2\log(2K)+2\big)    \max_{a\in\cA}\lambda_1(\hat\Sigma_a).
\label{eq:zpistar2_bound_L}
\end{align}
Now recall that to simplify, we also assumed that $\Sigma_a = \sigma_0^2 I_d$ for any $a \in \cA$ and that $\Sigma=\tau^2 I_{d'}$. As a result, we have that:
\begin{align*}
    \hat{\Sigma}_a = \tilde{\Sigma}_{a}  + \sigma_0^{-4}\tilde{\Sigma}_{a}  \W_a \bar{\Sigma} \W_a^\top \tilde{\Sigma}_{a}  ,
\end{align*}
and we also have that $\lambda_1(\tilde{\Sigma}_{a}) \leq \sigma_0^2$ and that  
\begin{align}\label{eq:sdm-max_eigenvalue_covariance}
    \lambda_1(\hat{\Sigma}_a) & \leq 
\sigma_0^2 + \tau^2   , & \forall a \in \cA  ,
\end{align} 
Plugging \eqref{eq:zpistar2_bound} (or \eqref{eq:zpistar2_bound_L}) into
\eqref{eq:bso_joint_event_bound} yields
\begin{align}
\textsc{Bso}(\hat\pi_{\textsc{g}})
& \le
\alpha  \E{}{\|\phi(X)\|_{\hat\Sigma_{\pi_*(X)}}}
+
\sqrt{(2\log(2K)+2) (\sigma_0^2 + \tau^2) \delta},
\label{eq:bso_joint_event_final_L}
\end{align}
with $\alpha^2 = d + 2\sqrt{d\log(K/\delta)} + 2\log(K/\delta)$ as defined above. Choosing $\delta = 1/n$ in \eqref{eq:bso_joint_event_final_L} yields
\begin{align}
\textsc{Bso}(\hat\pi_{\textsc{g}})
& \le
\alpha_n  \E{}{\|\phi(X)\|_{\hat\Sigma_{\pi_*(X)}}}
+
\sqrt{\frac{(2\log(2K)+2)(\sigma_0^2 + \tau^2)}{n}},
\label{eq:bso_joint_event_delta_1_over_n}
\end{align}
where
\begin{align}
\alpha_n^2
=
d + 2\sqrt{d\log(Kn)} + 2\log(Kn).
\label{eq:alpha_delta_1_over_n}
\end{align}

This concludes the proof.
\end{proof}

\subsection{Explicit Bound}\label{subsec:sdm-explicit_bound}

\textbf{Additional simplification.} To further simplify the exposition, we just set $\phi(x) = x$ for any $x \in \cX$.

\begin{assumption}\label{assum:covariate-covariance}
Let $G = \mathbb{E}_{}[X X^\top]$ with $g = \lambda_d(G)$. We assume that $g > 0$.
\end{assumption}

\begin{assumption}[Context-independent logging policy]\label{assum:independent-logging}
$A$ is independent of $X$, i.e., $\pi_0(a \mid x) = p_a$ for all $x$ and $a$. Equivalently, $(X_i)$ are i.i.d.\ $\sim \nu$ and independent of $(A_i)$, with $\mathbb{P}(A = a) = p_a$.
\end{assumption}

\begin{theorem}[Explicit Bound]\label{thm:main_thm_app}
Let $\pi_*(x)$ be the optimal action for context $x$. Then,
the BSO of \emph{\alg} under the structured prior \cref{eq:sdm-contextual_gaussian_model} satisfies
\begin{align*}
\textsc{Bso}(\hat\pi_{\textsc{g}})
&\le
\alpha_n  \sqrt{\E{}{
\frac{1}{\lambda_X} + \frac{\tau^2}{\sigma_0^{4}\lambda_X^2}
+ (\sigma_0^2+\tau^2)\left((d+1)e^{-n\rho_X^2/2}
+ d\left(\tfrac{2}{e}\right)^{g m_X/2}\right)
}}\\
&+
\sqrt{\frac{(2\log(2K)+2)(\sigma_0^2 + \tau^2)}{n}},
\end{align*}
where $$\alpha_n = \sqrt{d + 2\sqrt{d\log(Kn)} + 2\log(Kn)}  ,$$
and
$$
\rho_X = p_{\pi_*(X)},\qquad
m_X = \left\lfloor \frac{n\rho_X}{2}\right\rfloor,\qquad
\lambda_X = \sigma^{-2}g\frac{m_X}{2}+\sigma_0^{-2}.
$$
\end{theorem}

\textbf{Scaling with $n$.}
Recall $m_X=\lfloor n\rho_X/2\rfloor$ and
$\lambda_X=\sigma^{-2}g  \frac{m_X}{2}+\sigma_0^{-2}$, where
$\rho_X=\pi_0(\pi_*(X))$.
For $n$ large enough (so that $m_X$ is roughly 
$n\rho_X$), we have
$\lambda_X=\Theta(n\rho_X+1)$ and the exponentially small tail terms can be ignored at the level of leading-order scaling.
Consequently, up to absolute constants and polylogarithmic factors,
$$
\textsc{Bso}(\hat{\pi}_{\textsc{g}})
=
\tilde{\mathcal O} \left(
\alpha_n  \sqrt{\mathbb E \left[\frac{1}{n\rho_X+1}\right]}
 + 
\sqrt{\frac{\log K}{n}}
\right),
$$
and using $\alpha_n=\tilde{\Theta}(\sqrt d)$ this can be summarized as
$$
\textsc{Bso}(\hat{\pi}_{\textsc{g}})
=
\tilde{\mathcal O} \left(
 \sqrt{\mathbb E \left[\frac{d}{n\rho_X+1}\right]}
 + 
\sqrt{\frac{\log K}{n}}
\right),
$$

   \begin{proof}
Let's focus on the main term of \cref{thm:main_thm_1}, and we start by bounding 
\begin{align}
\mathbb{E} \left[\|X\|_{\hat{\Sigma}_{\pi_*(X)}}^2\right],
\label{eq:bso-cs}
\end{align}
Now recall that to simplify, we also assumed that $\Sigma_a = \sigma_0^2 I_d$ for any $a \in \cA$ and that $\Sigma=\tau^2 I_{d'}$. As a result, we have that:
\begin{align*}
    \hat{\Sigma}_a = \tilde{\Sigma}_{a}  + \sigma_0^{-4}\tilde{\Sigma}_{a}  \W_a \bar{\Sigma} \W_a^\top \tilde{\Sigma}_{a}  ,
\end{align*}
and we also have that $\lambda_1(\tilde{\Sigma}_{a}) \leq \sigma_0^2$ and that  
\begin{align}
    \lambda_1(\hat{\Sigma}_a) & \leq 
\sigma_0^2 + \tau^2   , & \forall a \in \cA  ,
\end{align} 
since $\lambda_1(\W_a \W_a^\top) = 1$. These are obtained using Weyl's inequalities.

Now let 
\begin{align*}
    N_a = \sum_{i=1}^n \mathds{1}\{A_i=a\},
\qquad
p_a = \mathbb{P}(A=a)=\pi_0(a),
\qquad
\rho_x = p_{\pi_*(x)},
\qquad
\rho_X = p_{\pi_*(X)}.
\end{align*}
Define
$$
m_x = \left\lfloor \frac{n\rho_x}{2}\right\rfloor,
\qquad
m_X = \left\lfloor \frac{n\rho_X}{2}\right\rfloor.
$$

Then, under \cref{assum:independent-logging}, $N_a\sim \mathrm{Bin}(n,p_a)$ and it is independent of context $X$.
Therefore, Hoeffding gives, for $t=n\rho_X/2$,
\begin{align}
\mathbb{P} \left(N_{\pi_*(X)} < \frac{n\rho_X}{2}  \middle|   X, \theta_*\right)
\le \exp \left(-\frac{n\rho_X^2}{2}\right).
\label{eq:count-failure}
\end{align}
Let
$$
\Omega_{X,1} = \left\{ N_{\pi_*(X)} \ge m_X \right\}.
$$
Then we have that 
\begin{align}\label{part11}
    \mathbb{P} \left(\bar{\Omega}_{X,1}  \middle|   X, \theta_*\right)
\le \exp \left(-\frac{n\rho_X^2}{2}\right).
\end{align}

\begin{lemma}[Matrix Chernoff, {\citet[Theorem~1.1]{tropp2012user}}]\label{lem:matrix-chernoff}
Let $(Y_k)_{k=1}^m$ be independent PSD matrices with $\lambda_1(Y_k)\le R$ a.s.
Let $\mu_{\min}=\lambda_d(\sum_{k=1}^m \mathbb E[Y_k])$.
Then for $\delta\in[0,1]$,
$$
\mathbb P \left(\lambda_d\Big(\sum_{k=1}^m Y_k\Big)\le (1-\delta)\mu_{\min}\right)
\le d\left[\frac{e^{-\delta}}{(1-\delta)^{1-\delta}}\right]^{\mu_{\min}/R}.
$$
\end{lemma}
Define 
$$
\Omega_{X,2}
=
\left\{
\lambda_d \Big(\sum_{i=1}^n \mathds{1}\{A_i=\pi_*(X)\}X_iX_i^\top\Big)
\ge \frac{1}{2}  N_{\pi_*(X)}  g
\right\},
\qquad
\Omega_X = \Omega_{X,1}\cap \Omega_{X,2}.
$$

We now bound $\mathbb{P}(\bar{\Omega}_{X,2}\mid X, \theta_*)$.
Recall that
$$
\Omega_{X,1} = \left\{N_{\pi_*(X)} \ge m_X\right\},\qquad
m_X = \left\lfloor \frac{n\rho_X}{2}\right\rfloor.
$$

Under \cref{assum:independent-logging}, $(A_i)$ is independent of $(X_i)$, hence conditional on the index set
$S_{\pi_*(X)}=\{i\in[n]:A_i=\pi_*(X)\}$, the matrices
$\{X_iX_i^\top:i\in S_{\pi_*(X)}\}$ are i.i.d. with the same law as $XX^\top$.
Moreover, since the Chernoff bound below depends on $S_{\pi_*(X)}$ only through
$|S_{\pi_*(X)}|=N_{\pi_*(X)}$, the same bound holds when conditioning on $N_{\pi_*(X)}$. Moreover, we have that $\phi(x) = x$ and that $\|\phi(x)\| \le 1$. Thus, $\|x\|\leq 1$ and hence  $\lambda_1(X_iX_i^\top) \le 1$. Applying Lemma~\ref{lem:matrix-chernoff} with $R=1$, $\mathbb{E}[XX^\top]=G$, and
$$
\mu_{\min}=\lambda_d \left(N_{\pi_*(X)}G\right)=N_{\pi_*(X)}g,
$$
and choosing $\delta=\tfrac12$, we obtain the conditional bound
$$
\mathbb{P}(\bar{\Omega}_{X,2}\mid X, N_{\pi_*(X)}, \theta_*)
\le d\left(\sqrt{2/e}\right)^{gN_{\pi_*(X)}}.
$$
Taking conditional expectation $N_{\pi_*(X)}$ given $X$ and $\theta_*$,
$$
\mathbb{P}(\bar{\Omega}_{X,2}\mid X, \theta_*)
= \mathbb{E} \left[\mathbb{P}(\bar{\Omega}_{X,2}\mid X, N_{\pi_*(X)}, \theta_*)\mid X, \theta_*\right]
\le d  \mathbb{E} \left[\left(\sqrt{2/e}\right)^{gN_{\pi_*(X)}}\mid X, \theta_*\right].
$$
Splitting on $\Omega_{X,1}$ yields
\begin{align*}
\mathbb{E} \left[\left(\sqrt{2/e}\right)^{gN_{\pi_*(X)}}\mid X, \theta_*\right]
&= \mathbb{E} \left[\left(\sqrt{2/e}\right)^{gN_{\pi_*(X)}}\mathds{1}\{\Omega_{X,1}\}\mid X, \theta_*\right]
  + \mathbb{E} \left[\left(\sqrt{2/e}\right)^{gN_{\pi_*(X)}}\mathds{1}\{\bar{\Omega}_{X,1}\}\mid X, \theta_*\right] \\
&\le \left(\sqrt{2/e}\right)^{gm_X} + \mathbb{P}(\bar{\Omega}_{X,1}\mid X, \theta_*),
\end{align*}
since on $\Omega_{X,1}$ we have $N_{\pi_*(X)}\ge m_X$ and thus $\left(\sqrt{2/e}\right)^{gN_{\pi_*(X)}}\le \left(\sqrt{2/e}\right)^{gm_X}$, while on $\bar{\Omega}_{X,1}$
we have $\left(\sqrt{2/e}\right)^{gN_{\pi_*(X)}}\le 1$. Therefore,
\begin{align}
\mathbb{P}(\bar{\Omega}_{X,2}\mid X, \theta_*)
\le d\left(\sqrt{2/e}\right)^{g m_X} + d \mathbb{P}(\bar{\Omega}_{X,1}\mid X, \theta_*).
\label{eq:design-failure}
\end{align}
Combining \eqref{eq:design-failure} with \eqref{eq:count-failure} yields
\begin{align}
\mathbb{P}(\bar{\Omega}_{X}\mid X, \theta_*)
\le \mathbb{P}(\bar{\Omega}_{X,1}\mid X, \theta_*) + \mathbb{P}(\bar{\Omega}_{X,2}\mid X, \theta_*)
\le (d+1) \exp \left(-\frac{n\rho_X^2}{2}\right) + d\left(\sqrt{2/e}\right)^{g m_X}.
\end{align}
Finally, let
$$
I_1 = \mathbb{E} \left[\|X\|_{\hat{\Sigma}_{\pi_*(X)}}^2  \mathds{1}\{\Omega_X\}\mid X, \theta_*\right],
\qquad
I_2 = \mathbb{E} \left[\|X\|_{\hat{\Sigma}_{\pi_*(X)}}^2  \mathds{1}\{\bar{\Omega}_X\}\mid X, \theta_*\right].
$$
Then
$$
\mathbb{E} \left[\|X\|_{\hat{\Sigma}_{\pi_*(X)}}^2\right]
= \mathbb{E}[I_1]+\mathbb{E}[I_2].
$$

Using $\|X\|_2\le 1$ and \eqref{eq:sdm-max_eigenvalue_covariance},
$$
\|X\|_{\hat{\Sigma}_{\pi_*(X)}}^2
= X^\top \hat{\Sigma}_{\pi_*(X)} X
\le \lambda_1(\hat{\Sigma}_{\pi_*(X)})\|X\|_2^2
\le \sigma_0^2+\tau^2.
$$
Let $c_1=\sigma_0^2+\tau^2$. Then
\begin{align}
I_2 \le c_1  \mathbb{P}(\bar{\Omega}_X\mid X, \theta_*)
\le c_1 \left((d+1) \exp \left(-\frac{n\rho_X^2}{2}\right) + d\left(\sqrt{2/e}\right)^{g m_X}\right).
\label{eq:I2}
\end{align}

Moreover, fix $X$ and $\theta_*$, on $\Omega_X$,
$$
\lambda_d\Big(\sum_{i=1}^n \mathds{1}\{A_i=\pi_*(X)\}X_iX_i^\top\Big)
\ge \frac12 N_{\pi_*(X)} g
\ge \frac12 m_X g,
$$
so
$$
\lambda_d(\hat{G}_{\pi_*(X)})
= \sigma^{-2}\lambda_d\Big(\sum_{i=1}^n \mathds{1}\{A_i=\pi_*(X)\}X_iX_i^\top\Big)
\ge \sigma^{-2}\frac12 m_X g.
$$
Hence
$$
\lambda_d(\hat{G}_{\pi_*(X)}+\sigma_0^{-2}I_d)
\ge \sigma^{-2}\frac12 m_X g + \sigma_0^{-2}.
$$
Let
$$
\lambda_X = \sigma^{-2} g \frac{m_X}{2} + \sigma_0^{-2}.
$$
Since $\tilde{\Sigma}_{\pi_*(X)}=(\hat{G}_{\pi_*(X)}+\sigma_0^{-2}I_d)^{-1}$, we obtain on $\Omega_X$:
$$
\lambda_1(\tilde{\Sigma}_{\pi_*(X)}) \leq  \frac{1}{\lambda_X}.
$$
Moreover, on $\Omega_X$ we have that
$$
\lambda_1(\hat{\Sigma}_{\pi_*(X)})
\leq  \lambda_1(\tilde{\Sigma}_{\pi_*(X)}) + \sigma_0^{-4}\tau^2 \lambda_1(\tilde{\Sigma}_{\pi_*(X)})^2
\leq  \frac{1}{\lambda_X} + \frac{\sigma_0^{-4}\tau^2}{\lambda_X^2}.
$$
Therefore, since $\|X\|_2\leq  1$,
\begin{align}
I_1
\leq  \left(\frac{1}{\lambda_X} + \frac{\sigma_0^{-4}\tau^2}{\lambda_X^2}\right).
\label{eq:I1}
\end{align}

Combining \eqref{eq:I2} and \eqref{eq:I1},
$$
\mathbb{E} \left[\|X\|_{\hat{\Sigma}_{\pi_*(X)}}^2\right]
\leq 
\mathbb{E} \left[
\frac{1}{\lambda_X} + \frac{\sigma_0^{-4}\tau^2}{\lambda_X^2}
+ (\sigma_0^2+\tau^2)\left((d+1)\exp \left(-\frac{n\rho_X^2}{2}\right)
+ d\left(\sqrt{2/e}\right)^{g m_X}\right)
\right].
$$
But from \cref{thm:main_thm_1}, we know that
\begin{align}\label{sdm:helper}
\textsc{Bso}(\hat\pi_{\textsc{g}})
& \le
\alpha_n  \E{}{\|X\|_{\hat\Sigma_{\pi_*(X)}}}
+
\sqrt{\frac{(2\log(2K)+2)(\sigma_0^2 + \tau^2)}{n}},
\end{align}
where $\alpha_n
=
\sqrt{d + 2\sqrt{d\log(Kn)} + 2\log(Kn)}.$

Finally, by Jensen's inequality, $\E{}{\|X\|_{\hat\Sigma_{\pi_*(X)}}} \leq \sqrt{\E{}{\|X\|_{\hat{\Sigma}_{\pi_*(X)}}^2}}$. Substituting into \eqref{sdm:helper} and combining with \eqref{eq:I2} and \eqref{eq:I1}, we obtain
\begin{align*}
\textsc{Bso}(\hat\pi_{\textsc{g}})
&\le
\alpha_n  \sqrt{\E{}{
\frac{1}{\lambda_X} + \frac{\tau^2}{\sigma_0^{4}\lambda_X^2}
+ (\sigma_0^2+\tau^2)\left((d+1)e^{-n\rho_X^2/2}
+ d\left(\tfrac{2}{e}\right)^{g m_X/2}\right)
}}\\
&+
\sqrt{\frac{(2\log(2K)+2)(\sigma_0^2 + \tau^2)}{n}},
\end{align*}
where $\alpha_n = \sqrt{d + 2\sqrt{d\log(Kn)} + 2\log(Kn)}$.

\end{proof}

\subsection{Optimality of Greedy 
Policies}\label{subsec:sdm-greedy_pessimism}
Here, we show that Greedy policy $\hat{\pi}_{\textsc{g}}$ should be preferred to any other choice of policies when considering the BSO as our performance metric. This is because $\hat{\pi}_{\textsc{g}}$ minimizes the BSO. To see this, note that by definition the Greedy policy $\hat{\pi}_{\textsc{g}}$ is deterministic, that is for any context $x \in \cX$, there exists $\hat{a}_{\textsc{g}}$, such that $\hat{\pi}_{\textsc{g}}(\hat{a}_{\textsc{g}}\mid x) =1$. Thus, for any context $x \in \cX$, we simplify the notation by letting $\hat{\pi}_{\textsc{g}}(x)$ denote the action that has a mass equal to $1$. Then, we have that 
\begin{align}\label{eq:sdm-helper_11}
 \E{A \sim \hat{\pi}_{\textsc{g}}(\cdot | x)}{\E{\theta_*}{r(x, A; \theta_*)\mid \cD_n}} &= \E{\theta_*}{r(x, \hat{\pi}_{\textsc{g}}(x); \theta_*) \mid \cD_n}  ,\\
 &\geq \E{}{ r(x, a; \theta_*)\mid \cD_n} & \forall x, a \in \cX \times \cA  .
\end{align}
where this follows from the definition of $\hat{r}(x, a) = \E{}{ r(x, a; \theta)\mid \cD_n}$, the definition of $\hat{\pi}_{\textsc{g}}$ and the fact that $\theta_*$ is sampled from the prior, which leads to $\E{}{ r(x, a; \theta)\mid \cD_n} = \E{}{ r(x, a; \theta_*)\mid \cD_n}$. Now \cref{eq:sdm-helper_11} holds for any $x \in \cX$ and $a \in \cA$, and hence it holds in expectation under $X\sim \nu$ and $A \sim \pi(\cdot \mid X)$ for any policy $\pi$. That is, 
\begin{align}\label{eq:sdm-helper_12}
 \E{X \sim \nu, A \sim \hat{\pi}_{\textsc{g}}(\cdot | X)}{\E{\theta_*}{r(x, A; \theta_*)\mid \cD_n}} \geq  \E{X \sim \nu, A \sim \pi(\cdot | X)}{\E{\theta_*}{r(x, A; \theta_*)\mid \cD_n}}  .
\end{align}
Taking another expectation w.r.t. the sample set $\cD_n$ and using Fubini's theorem and the tower rule leads to $\E{}{V(\hat{\pi}_{\textsc{g}}; \theta_*)} \geq \E{}{V(\pi; \theta_*)}$ for any stationary policy $\pi$. Then, subtracting $\E{}{V(\pi_*; \theta_*)}$ from both sides of the previous inequality yields that the BSO is minimized by $\hat{\pi}_{\textsc{g}}$ compared to any stationary policy $\pi$, in particular, compared to the policy $\pi_{\textsc{p}}$ induced by pessimism.

\section{Additional Experiments}\label{app:sdm-experiments}
As mentioned in \cref{app:sdm-experiments}, our experiments were conducted on internal machines with 30 CPUs and thus they required a moderate amount of computation. These experiments are also reproducible with minimal computational resources.

\subsection{Implementation Details of Baselines}\label{app:sdm-details_imp}

We implement the baselines as follows.

\begin{itemize}
    \item \textbf{IPS.} 
    \begin{align}
    &\argmax_\pi \frac{1}{n} \sum_{i=1}^n \frac{\pi(A_i|X_i)}{\max\{\pi_0(A_i|X_i), \tau\}} R_i   ,
\end{align}
where $\tau \in [0, 1]$ is a hyper-parameter.
    \item \textbf{snIPS.} 
    \begin{align}
    \argmax_\pi \frac{1}{\sum_{i=1}^n \frac{\pi(A_i|X_i)}{\pi_0(A_i|X_i)}} \sum_{i=1}^n \frac{\pi(A_i|X_i)}{\pi_0(A_i|X_i)} R_i  ,
\end{align}

    \item \textbf{MIPS.} We cluster actions into $L$ groups and let $h(a)$ be the cluster of action $a$. Let $C_i$ be the cluster of action $A_i$, then we use
    \begin{align}
    \argmax_\pi \frac{1}{n} \sum_{i=1}^n \frac{\pi(C_i|X_i)}{\pi_0(C_i|X_i)}R_i   ,
\end{align}
where $C_i = h(A_i)$ for any $ i\in [n]$, and $\pi(c|x) = \sum_{a \in \mathcal{A}}\mathds{1}\left[h(a) = c \right]\pi(a|x)$.
    \item \textbf{PC.} We use the Knn implementation of \texttt{PC}. Let $N(a, k)$ be the set of k-nearest neighbors of $a$, then 
    \begin{align}
    \argmax_\pi \frac{1}{n} \sum_{i=1}^n \frac{\sum_{a \in \mathcal{A}}\mathds{1}\left[a \in N(A_i, k) \right]\pi(a|x)}{\sum_{a \in \mathcal{A}}\mathds{1}\left[a \in N(A_i, k)\right]\pi_0(a|x)}R_i   .
\end{align}

    \item \textbf{DM (Freq).} This DM uses the linear-Gaussian likelihood model $R \mid \theta, X, A \sim \cN(\phi(X)^\top \theta_A , \sigma^2)$ and learn the parameters $\theta_a$ using the maximum likelihood principle leading to 
    \begin{align}
        \hat{r}(x, a) = \phi(x)^\top \hat{\mu}_a  ,
    \end{align}
    where the MLE is \(\hat{\mu}_{a} = (G_a + \lambda I_d)^{-1} B_a \), with \( G_{a} = \sum_{i \in [n]} \mathbb{I}_{\{A_i=a\}} \phi(X_i) \phi(X_i)^\top \) and \( B_{a} = \sum_{i \in [n]} \mathbb{I}_{\{A_i=a\}} R_i \phi(X_i) \), and $\lambda$ is a regularization hyper-parameter.
    
    \item \textbf{DM (Bayes).} This DM uses the linear-Gaussian likelihood model combined with Gaussian priors as
\begin{align}
  \theta_a &\sim \cN(\mu_{a}, \Sigma_{a})  , & \forall a \in \cA  ,\\
    R \mid \theta, X, A &\sim \cN(\phi(X)^\top \theta_A , \sigma^2)  ,\nonumber
\end{align}
Under this prior, each action \( a \) has an associated parameter \( \theta_a \). Given the prior in \cref{eq:sdm-basic_model}, the posterior distribution of an action parameter follows a multivariate Gaussian: \( \theta_a \mid \cD_n \sim \mathcal{N}(\hat{\mu}_a, \hat{\Sigma}_a) \), where \( \hat{\Sigma}_{a}^{-1} = \Sigma_{a}^{-1} + G_{a} \) and \( \hat{\Sigma}_{a}^{-1} \hat{\mu}_{a} = \Sigma_{a}^{-1} \mu_a + B_a \). Here, \( G_{a} = \sigma^{-2} \sum_{i \in [n]} \mathbb{I}_{\{A_i=a\}} \phi(X_i) \phi(X_i)^\top \) and \( B_{a} = \sigma^{-2} \sum_{i \in [n]} \mathbb{I}_{\{A_i=a\}} R_i \phi(X_i) \). Then, the reward estimate is 
    \begin{align}
        \hat{r}(x, a) = \phi(x)^\top \hat{\mu}_a  ,
    \end{align}
    \item \textbf{DR.} 
    \begin{align}
    \argmax_\pi \frac{1}{n} \sum_{i=1}^n &\frac{\pi(A_i|X_i)}{\max\{\pi_0(A_i|X_i), \tau\}} \left(R_i - \hat{r}(X_i, A_i) \right)+
 \E{A \sim \pi(\cdot | X_i)}{\hat{r}(X_i, A)}  ,
\end{align}
with $\tau \in [0, 1]$ and $\hat{r}$ is the reward model obtained using \texttt{DM (Freq)}.
\end{itemize}

\subsection{Robustness to Likelihood Misspecification}\label{app:sdm-misspecification_experiments}

We strengthened our evaluation by assessing \alg's robustness to likelihood misspecification below (robustness to prior misspecification is provided in \cref{app:sdm-mips_experiments}). In these experiments, the true data-generating process (same as the synthetic experiments in \cref{sec:sdm-experiments}) differed from \alg's assumptions in two different ways: either the likelihood is misspecified

\begin{figure}[H]
\begin{center}
\centerline{\includegraphics[width=0.9\linewidth]{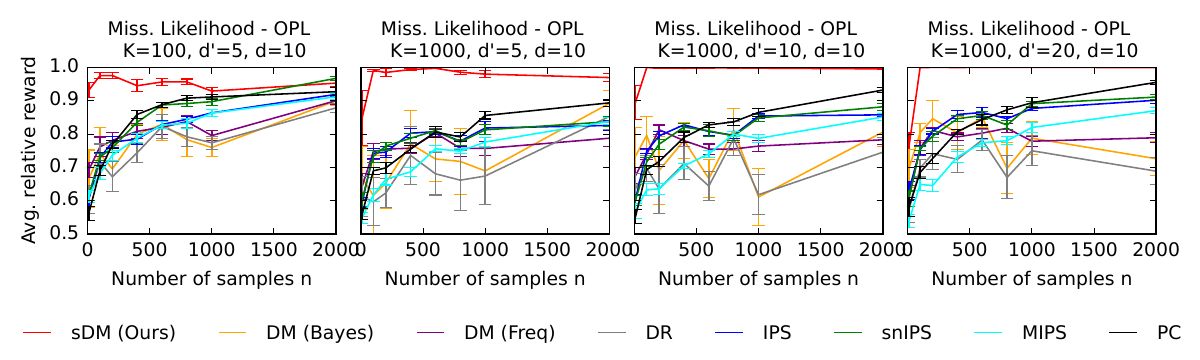}}
\vskip -0.1in
\caption{Effect of likelihood misspecification: The relative reward of the learned policy on synthetic problems using misspecified likelihood with varying $n$ and $K$.}
\label{fig:sdm-likelihood_misspecfication_opl}
\end{center}
\end{figure}

\textbf{Misspecified likelihood (\cref{fig:sdm-likelihood_misspecfication_opl}).} We also simulate when the true reward distribution differed from the likelihood assumed by \alg. For example, we simulated binary rewards using a Bernoulli-logistic model while \alg used a linear-Gaussian likelihood. Other DMs: \texttt{DM (Bayes)} and \texttt{DM (Freq)} also use a misspecified likelihood model and to emphasize this we add the suffix \texttt{Lin} to all DMs names. Overall, \alg still outperforms all methods by a large margin despite misspecification.

\subsection{Robustness to Prior Misspecification}\label{app:sdm-mips_experiments}

\begin{figure}[H]
  \centering  \includegraphics[width=0.9\linewidth]{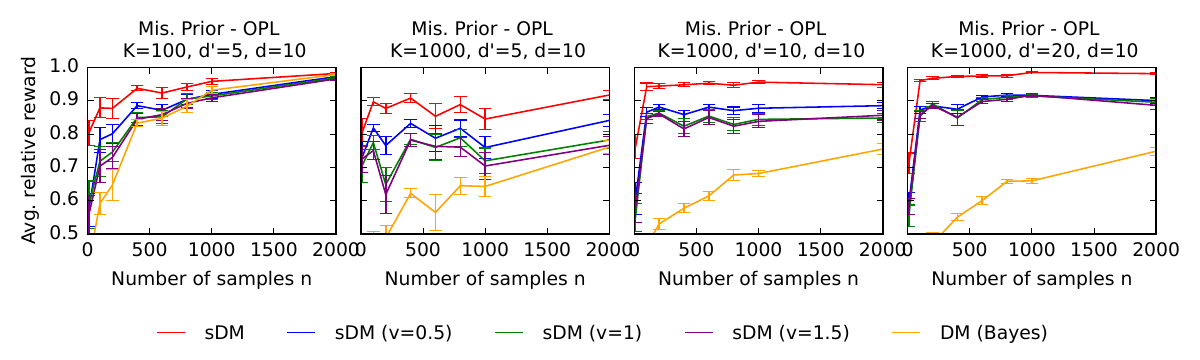}
  \caption{Effect of prior mean and covariance misspecification: The average relative reward of the learned policy on synthetic problems using both misspecified prior means and covariances with varying $n$ and $K$ and $d'$.} 
  \label{fig:sdm-mis_prior_mean_covariance}
\end{figure}

\textbf{Misspecified prior means and covariances(\cref{fig:sdm-mis_prior_mean_covariance}).}  This is achieved by adding uniformly sampled noise from \([v, v+0.5]\) to both the true prior mean and covariance parameters \(\mu, \Sigma, \W_a, \Sigma_a\), with \(v\) controlling the level of misspecification. We varied \(v \in \{0.5, 1, 1.5\}\) and analyzed its impact on \alg's performance. For comparison, we included the well-specified \alg and the most competitive baseline, \texttt{DM (Bayes)}, while omitting other baselines to reduce clutter. \alg's performance decreases with increasing misspecification, yet \alg with misspecification still outperforms the most competitive baseline, especially when \(K\) is large. We also observe that the impact of prior covariance misspecification is less significant compared to prior mean misspecification.

\subsection{Comparison of Greedy and Pessimistic Policies}\label{app:sdm-greedy_pessimism_experiments}

To validate our theory that a greedy policy should be preferred over the commonly adopted pessimistic policy in our Bayesian setting, we used a performance metric averaged over multiple bandit problems sampled from the prior. To verify this, we considered the same OPL synthetic setting as in \cref{sec:sdm-experiments} and compared \alg with a greedy policy to \alg with a pessimistic policy. Recall that a greedy policy with respect to our reward estimate writes
\begin{align}
     \hat{\pi}_{\textsc{g}}(a \mid x) = \mathds{1}{\{a = \argmax_{b \in \cA} \hat{r}(x, b)\}}  ,
\end{align}
while a pessimistic one writes
\begin{align}
     \hat{\pi}_{\textsc{p}}(a \mid x) = \mathds{1}{\{a = \argmax_{b \in \cA} \hat{r}(x, b) - u(x, a)\}}  ,
\end{align}
where $u(x, a) = \alpha(d, \delta) \|\phi(X)\|_{\hat{\Sigma}_a}$ with $\alpha(d, \delta) = \sqrt{d + 2 \sqrt{d \log \frac{1}{\delta}} + 2 \log \frac{1}{\delta}}$. As predicted by our theory, the results show that the greedy policy has better average performance over multiple bandit instances sampled from the prior.

\begin{figure}[H]
  \centering  \includegraphics[width=\linewidth]{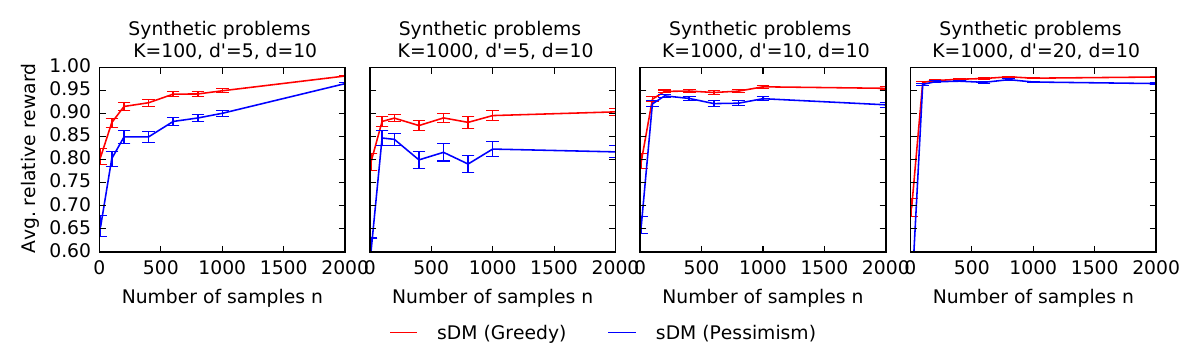}
  \caption{Comparison of \alg with greedy policy and \alg with pessimistic policy in OPL: The average MSE of an $\epsilon$-greedy policy on synthetic problems with varying $n$, $K$, and $d^\prime$.} 
  \label{fig:sdm-greedy_pessimism_results}
\end{figure}

%% file: contents/las/appendix.tex
\section{Proofs for Oracle Policies}\label{app:las-asymptotic_solutions_proofs}

\subsection{Oracle Policies for IPS-Based Objectives}

\textbf{(IPS), cIPS and ES.} Recall the definition of the (logging propensity) clipped IPS estimator with $\tau \in [0, 1]$: 

\begin{align*}
    \hat{V}_{\textsc{cips}}(\pi) = \frac{1}{n} \sum_{i=1}^n \frac{\pi(A_i|X_i)}{\max\{\pi_0(A_i|X_i), \tau\}} R_i \,.
\end{align*}
Taking $n \rightarrow \infty$, one obtains:
\begin{align*}
    V_{\textsc{cips}}(\pi) &= \E{X \sim \nu, A \sim \pi_0(\cdot|X)}{\frac{\pi(A|X)}{\max\{\pi_0(A|X), \tau\}}r(X, A) } \\
    &= \E{X \sim \nu, A \sim \pi(\cdot|X)}{\frac{\pi_0(A|X)}{\max\{\pi_0(A|X), \tau\}}r(X, A) }.
\end{align*}
As the objective is linear in the policy $\pi$, the optimal policy should put for any $x \in \mathcal{X}$, all the mass on the action $a$ that maximizes the weighted reward, giving:
\begin{align*}
    \pi^{\texttt{cIPS}}_*(a|x) = \mathbbm{1}\Big[ a = \argmax_{a' \in \mathcal{A}} \frac{\pi_0(a'|x) r(x, a') }{\max\{\pi_0(a'|x), \tau\}}  \Big]\,.
\end{align*}
We recover the solution for \texttt{IPS} when we let $\tau \rightarrow 0$:
\begin{align*}
    \pi^{\texttt{IPS}}_*(a|x) = \mathbbm{1}\Big[ a = \argmax_{a' \in \mathcal{A}} r(x, a') \mathbbm{1}\left[ \pi_0(a'|x) > 0\right]  \Big]\,.
\end{align*}
We also recover the solution of \texttt{ES} just by replacing the clipping function by an exponential function of factor $\alpha$, obtaining:
\begin{align*}
    \pi^{\texttt{ES}}_*(a|x) = \mathbbm{1}\Big[ a = \argmax_{a' \in \mathcal{A}} r(x, a') \pi_0(a'|x)^{1 - \alpha}  \Big]\,.
\end{align*}

\textbf{Doubly Robust (DR).} The doubly robust estimator converges to the following quantity:
\begin{align*}
    V_{\textsc{dr}}(\pi)
    &= \E{X \sim \nu, A \sim \pi(\cdot|X)}{ (r(X, A) - \hat{r}(X, A)) \frac{\pi_0(A|X)}{\max\{\pi_0(A|X), \tau\}} + \hat{r}(X, A)}.
\end{align*}
The objective is linear in $\pi$ and is thus maximized by the following deterministic decision rule:
\begin{align*}
    \pi^{\texttt{DR}}_*(a|x) = \mathbbm{1}\left[  a =  \argmax_{a' \in \mathcal{A}} \hat{r}(x, a') + (r(x, a') - \hat{r}(x, a'))\frac{\pi_0(a'|x)}{\max\{\pi_0(a'|x), \tau\}}\right]
\end{align*}

\paragraph{\textbf{Marginalized IPS (MIPS) with clusters.}} We adopt the same approach to look for the maximizer of \texttt{MIPS}. We generalize the clustering function $h$ to also account for context. We write down the estimator:
\begin{align*}
    \hat{V}_{\textsc{mips}}(\pi) = \frac{1}{n} \sum_{i=1}^n \frac{\sum_{a'} \mathbbm{1}\left[h(a', X_i) = h(A_i, X_i)  \right]\pi(a'|X_i)}{\sum_{a''} \mathbbm{1}\left[h(a'', X_i) = h(A_i, X_i)  \right]\pi_0(a''|X_i)}R_i = \frac{1}{n} \sum_{i=1}^n \frac{\pi(C_i|X_i)}{\pi_0(C_i|X_i)}R_i \,,
\end{align*}
with which, we recover when $n \rightarrow \infty$:
\begin{align*}
    V_{\textsc{mips}}(\pi) &= \E{X \sim \nu, A \sim \pi_0(\cdot|X)}{\frac{\sum_{a'} \mathbbm{1}\left[h(a', X) = h(A, X)  \right]\pi(a'|X)}{\sum_{a''} \mathbbm{1}\left[h(a'', X) = h(A, X)  \right]\pi_0(a''|X)}r(X,A)} \\
    &= \E{X \sim \nu}{\sum_{a} \pi_0(a|X)\frac{\sum_{a'} \mathbbm{1}\left[h(a', X) = h(a, X)  \right]\pi(a'|X)}{\sum_{a''} \mathbbm{1}\left[h(a'', X) = h(a, X)  \right]\pi_0(a''|X)}r(X,a)} \\
    &= \E{X \sim \nu}{\sum_{a'} \pi(a'|X)\sum_{a}\pi_0(a|X)\frac{\mathbbm{1}\left[h(a', X) = h(a, X)  \right]}{\sum_{a''} \mathbbm{1}\left[h(a'', X) = h(a, X)  \right]\pi_0(a''|X)}r(X,a)} \\
    &= \E{X \sim \nu}{\sum_{a'} \pi(a'|X)\E{A \sim \pi_0(\cdot|X)}{\frac{\mathbbm{1}\left[h(a', X) = h(A, X)  \right] r(X,A)}{\E{A'' \sim \pi_0(\cdot|X)}{\mathbbm{1}\left[h(A'', X) = h(A, X)  \right]}}}}.
\end{align*}
The objective is linear in $\pi$, and depends on the action $a'$ through its cluster $h(a', \cdot)$ alone. This means that multiple solutions are maximizers as long as the policy chooses the best cluster $c$. We thus write down the oracle policy for \texttt{MIPS} in the cluster level, giving:
\begin{align*}
    \pi^{\texttt{\texttt{MIPS}}}_*(c|x) &= \mathbbm{1}\Big[ c = \argmax_{c' \in \mathcal{C}} \Big\{ \E{A \sim \pi_0(\cdot|x)}{\frac{r(x, A)\mathbbm{1}[h(A,x) = c']}{\E{A'' \sim \pi_0(\cdot|x)}{\mathbbm{1}\left[h(A'', x) = h(A, x)  \right]}}} \Big\} \Big]\\
    &= \mathbbm{1}\Big[ c = \argmax_{c' \in \mathcal{C}} \Big\{ \E{A \sim \pi_0(\cdot|x)}{\frac{r(x, A)\mathbbm{1}[h(A,x) = c']}{\E{A'' \sim \pi_0(\cdot|x)}{\mathbbm{1}\left[h(A'', x) = c'  \right]}}} \Big\} \Big]\\
    &= \mathbbm{1}\Big[ c = \argmax_{c' \in \mathcal{C}} \Big\{ \frac{\E{A \sim \pi_0(\cdot|x)}{r(x, A)\mathbbm{1}[h(A,x) = c']}}{\E{A \sim \pi_0(\cdot|x)}{\mathbbm{1}\left[h(A, x) = c'  \right]}} \Big\} \Big]\,,
\end{align*}
which ends the proof.

\paragraph{\textbf{Conjunct Effect Modeling (OffCEM).}} This estimator can be seen as the natural, doubly robust extension of the \texttt{MIPS} estimator. Combining similar techniques to the ones employed for \texttt{MIPS} and \texttt{DR} yields
\begin{align*}
   & \pi^{\texttt{\texttt{OffCEM}}}_*(a|x) = \mathbbm{1}\Big[ a = \argmax_{a' \in \mathcal{A}} \Big\{ \hat{r}(x, a') +  \frac{\E{\bar{A} \sim \pi_0(\cdot|x)}{(r(x, \bar{A}) - \hat{r}(x, \bar{A})) \mathbbm{1}[h(x, \bar{A}) = h(x, a')]}}{\pi_0(h(x, a')|x)}\Big\}\Big]\,.
\end{align*}

\paragraph{\textbf{Two Stage Decomposition (POTEC).}} This is an \emph{optimization strategy} for \texttt{OffCEM}. It restricts the policy to a cluster-informed form,
\begin{align*}
    \pi(a \mid x) = \sum_{c \in \mathcal{C}} \pi^{\textsc{rm}}(a \mid x,c)\pi^{\textsc{cl}}(c\mid x) ,
\end{align*}
where $\pi^{\textsc{rm}}(a \mid x, c) = \mathbbm{1}[a = \argmax_{a' \in c} \hat{r}(x, a')]$ is fixed, model-based policy that deterministically selects the best action within each cluster. Learning is then simplified to finding the optimal cluster-level policy $\pi^{\textsc{cl}}$ that maximizes the \texttt{OffCEM} objective:
\begin{align*}
\hat{V}_{\textsc{potec}}(\pi^{\textsc{cl}}) 
&= \frac{1}{n} \sum_{i=1}^n \left(
\frac{\pi^{\textsc{cl}}(C_i\mid X_i)}{\pi_0(C_i\mid X_i)}\left(R_i - \hat{r}(X_i, A_i)\right) 
+ \sum_{c\in\mathcal C} \pi^{\textsc{cl}}(c\mid X_i) \hat r^*_c(X_i)
\right),
\end{align*}
where $\hat r^*_c(x) = \max_{a\in c}\hat r(x,a)$ is the estimated reward of the best action in cluster $c$. This is exactly the Doubly Robust version of \texttt{MIPS} on the cluster level, the oracle policy on the cluster level can be followed in the same fashion:
\begin{align*}
    \pi^{\textsc{cl}}_*(c \mid x) =  \mathbbm{1}\Big[ c = \argmax_{c' \in \mathcal{C}} \Big\{ \frac{\E{A \sim \pi_0(\cdot|x)}{(r(x, A) - \hat{r}(x, A))\mathbbm{1}[h(A,x) = c']}}{\E{A \sim \pi_0(\cdot|x)}{\mathbbm{1}\left[h(A, x) = c'  \right]}} + \hat{r}_{c'}^*(x)\Big\} \Big]\,.
\end{align*}

The optimal policy for the \texttt{POTEC} optimization strategy unfolds as:
\begin{align*}
    \pi^{\texttt{POTEC}}_*(a|x) &= \sum_{c \in \mathcal{C}} \pi^{\textsc{rm}}(a \mid x,c)\pi_*^{\textsc{cl}}(c\mid x)\,.
\end{align*}
At first glance, it might be hard to see the connection between \texttt{POTEC} and \texttt{OffCEM} solutions, but they are equivalent. For ease of notation, let us denote by $D_{\hat{r}, x}(c)$:
\begin{align*}
    D_{\hat{r}, x}(c) = \frac{\E{A \sim \pi_0(\cdot|x)}{(r(x, A) - \hat{r}(x, A))\mathbbm{1}[h(A,x) = c]}}{\E{A \sim \pi_0(\cdot|x)}{\mathbbm{1}\left[h(A, x) = c  \right]}}\,.
\end{align*}
and recall that the optimal policy of \texttt{OffCEM} finds the action $a$ that maximizes:
\begin{align*}
    \tilde{V}(x,a) = \hat{r}(x, a) + D_{\hat{r}, x}(h(a,x))\,.
\end{align*}
For any context $x$, the optimal action $a^*$ of \texttt{POTEC} verifies:
\begin{itemize}
    \item $a^*$ is in the optimal cluster: $h(a^*, x) = c_*(x)$ with  $c_*(x) = \argmax_{c \in \mathcal{C}} D_{\hat{r}, x}(c) + \hat{r}_{c}^*(x)$.
    \item $a^*$ is optimal within that cluster: $a = \argmax_{a \in c_*(x)} \hat{r}(x, a)$.
\end{itemize}
This means that for all actions $a$ with $h(a, x) \neq c_*(x)$, we have:
\begin{align*}
    \tilde{V}(x,a) &= D_{\hat{r}, x}(h(a,x)) + \hat{r}(x, a) \\
    &\le  D_{\hat{r}, x}(h(a,x)) + \hat{r}_{h(a,x)}^*(x) \\
    &\le D_{\hat{r}, x}(c_*(x)) + \hat{r}_{c_*(x)}^*(x)  \\
    &= D_{\hat{r}, x}(h(x, a^*)) + \hat{r}(x, a^*) = \tilde{V}(x,a^*)\,.
\end{align*}
In addition, for all actions $a$ with $h(a, x) = c_*(x)$, we have:
\begin{align*}
    \tilde{V}(x,a) &= D_{\hat{r}, x}(h(a,x)) + \hat{r}(x, a) \\&=  D_{\hat{r}, x}(c_*(x)) + \hat{r}(x, a) \\
    &\le D_{\hat{r}, x}(c_*(x)) + \hat{r}_{c_*(x)}^*(x) = \tilde{V}(x,a^*)\,.
\end{align*}
This means that the optimal action $a^*$ for \texttt{POTEC} is the maximizer of $\tilde{V}(x, a)$, which is exactly the solution of \texttt{OffCEM}.

\textbf{Policy Convolution (PC).} This estimator uses a nearest neighbors function to aggregate the propensities of similar actions, making the hypothesis that similar actions will result in similar reward signal. The estimator writes:
\begin{align*}
    \hat{V}_{\textsc{pc}}(\pi) 
    = \frac{1}{n} \sum_{i=1}^n \frac{\pi(N_\epsilon(A_i)\mid X_i)}{\pi_0(N_\epsilon(A_i)\mid X_i)} R_i, 
    \quad \text{with } \pi(N_\epsilon(a)\mid x) = \sum_{a' \in N_\epsilon(a)} \pi(a' \mid x).
\end{align*}
This estimator is equivalent to the following when $n \rightarrow \infty$:
\begin{align*}
    V^{\texttt{\texttt{PC}}}(\pi) &= \E{X \sim \nu, A \sim \pi_0(\cdot|X)}{\frac{\sum_{a'}\pi(a'|X) \mathbbm{1}\left[a'\in N_\epsilon(A) \right]}{\pi_0(N_\epsilon(A)|X)}r(X, A)}  \\
    &= \E{X \sim \nu, A \sim \pi(\cdot|X)}{ \E{\bar{A} \sim \pi_0(\cdot|X)}{\frac{r(x, \bar{A})\mathbbm{1}\left[A \in N_\epsilon(\bar{A}) \right]}{\pi_0(N_\epsilon(\bar{A})|X)}}}\,.
\end{align*}

The same argument of linearity applies here, giving us the corresponding oracle policy:
\begin{align*}
    \pi^{\texttt{\texttt{PC}}}_*(a|x) = \mathbbm{1}\Big[ a = \argmax_{a' \in \mathcal{A}} \Big\{ \E{\bar{A} \sim \pi_0(\cdot|x)}{\frac{r(x, \bar{A})\mathbbm{1}[a'\in N_\epsilon(\bar{A})]}{\pi_0(N_\epsilon(\bar{A})|x)}} \Big\} \Big]\,.
\end{align*}

\subsection{Oracle Policies for PWLL-Based Objectives}

Our objectives can be written in the same form, only choosing for each a different function $g$:
\begin{align*}
    \hat{U}_g(\pi) = \frac{1}{n}\sum_{i=1}^n 
    g(X_i, A_i, R_i)\log \pi(A_i \mid X_i)\,.
\end{align*} 
Since we are looking at oracle policies, we consider the expectation
\begin{align*}
    U_g(\pi)
    =
    \E{X \sim \nu,\, A\sim \pi_0(\cdot\mid X),\, R\sim p(\cdot\mid X,A)}{
        g(X,A,R)\log \pi(A\mid X)
    }.
\end{align*}
The maximization decomposes over contexts. Fix $x$ and define the nonnegative weights
\begin{align*}
    w_x(a) = \E{R\sim p(\cdot\mid x,a)}{g(x,a,R)} \ge 0.
\end{align*}
For each $x$, we thus consider
\begin{align*}
    &\max_{\pi(\cdot\mid x)}\;\; \sum_{a\in\mathcal A}\pi_0(a\mid x)\,w_x(a)\,\log \pi(a\mid x) \\
    &\text{s.t.}\quad \sum_{a\in\mathcal A}\pi(a\mid x)=1,\qquad \forall a\in\mathcal A,\;\pi(a\mid x)\ge 0.
\end{align*}
Let $v_x(a)= \pi_0(a\mid x)\,w_x(a)\ge 0$. The Lagrangian (with equality multiplier $\lambda\in\mathbb R$
and inequality multipliers $\{\mu(a)\}_{a\in\mathcal A}$, $\mu(a)\ge 0$) is
\begin{align*}
    \mathcal{L}(\pi,\lambda,\mu)
    =
    \sum_{a\in\mathcal A} v_x(a)\log \pi(a\mid x)
    +\lambda\Big(\sum_{a\in\mathcal A}\pi(a\mid x)-1\Big)
    +\sum_{a\in\mathcal A}\mu(a)\,\pi(a\mid x).
\end{align*}
By KKT conditions, at an optimum $\pi_*^g(\cdot\mid x)$ we have for all $a\in\mathcal A$:
\begin{align*}
    \frac{\partial \mathcal{L}}{\partial \pi(a\mid x)}
    =
    \frac{v_x(a)}{\pi(a\mid x)}+\lambda+\mu(a)=0,
    \qquad
    \text{and}\quad \mu(a)\,\pi(a\mid x)=0.
\end{align*}
For any action with $\pi_*^g(a\mid x)>0$, we get that $\mu(a)=0$, and hence
\begin{align*}
    \pi_*^g(a\mid x) = -\frac{v_x(a)}{\lambda}.
\end{align*}
Normalizing with $\sum_a \pi_*^g(a\mid x)=1$ gives $\lambda=-\sum_{a'} v_x(a')$ and therefore
\begin{align*}
    \pi_*^g(a\mid x)
    =
    \frac{v_x(a)}{\sum_{a'\in\mathcal A} v_x(a')}
    =
    \frac{\pi_0(a\mid x)\,\E{R\sim p(\cdot\mid x,a)}{g(x,a,R)}}
    {\sum_{a'\in\mathcal A}\pi_0(a'\mid x)\,\E{R\sim p(\cdot\mid x,a')}{g(x,a',R)}}.
\end{align*}
This concludes the proof.

\section{Proofs for Optimization Properties}\label{app:las-additional_proofs}

In this section, we prove the propositions about the optimization landscape of IPS-based and PWLL learning approaches. We start by stating the following lemmas, that will be helpful to prove our propositions.

\begin{lemma}{\citep[Lemma 2]{lemma_softmax}}\label{lem:smoothness}
Consider the single context case. With a slight abuse of notation, we drop the dependence on $x$ and write $r(a)$ instead of $r(x, a)$, $\pi_\theta(a)$ instead of $\pi_\theta(a|x)$, and $\hat{r}(a)$ instead of $\hat{r}(x, a)$. Let $\pi_\theta$ be a softmax policy parameterized by $ \theta$. Then, for any $\hat{\bm{r}} \in [0, 1]^K$, and any estimator $\hat{V}$ linear in $\pi_\theta$, the mapping $\theta \mapsto \hat{V}(\pi_{\theta}) = \langle \hat{\bm{r}}, \pi_\theta \rangle$ is $5/2$-smooth.
\end{lemma}

\begin{lemma}\label{app:las-linearisation} All the action level estimators \texttt{EST} in (\texttt{IPS}, \texttt{cIPS}, \texttt{DR}, \texttt{PC}) can be written, for any policy $\pi$, in the form: \begin{align} \hat{V}_{\textsc{est}}(\pi) = \frac{1}{n}\sum_{i = 1}^n \mathbb{E}_{A \sim \pi(\cdot|X_i)}\left[\hat{r}_{\textsc{est}, i}(A, X_i) \right]\,, \end{align} For the cluster level estimators/approaches \texttt{EST-C} in (\texttt{MIPS}, \texttt{OffCEM}, \texttt{POTEC}), we also have \begin{align} \hat{V}_{\textsc{est-c}}(\pi) = \frac{1}{n}\sum_{i = 1}^n \mathbb{E}_{C \sim \pi(\cdot|X_i)}\left[\hat{r}_{\textsc{est-c}, i}(C, X_i) \right]\,, \end{align} meaning that all these estimators are linear in $\pi$. \end{lemma} \begin{proof} This is straightforward to prove. We begin by the action level estimators and take \texttt{DR} as a representative. For \texttt{DR}, we have the following: \begin{align*} \hat{r}_{\texttt{DR}, i}(a, X_i) = \hat{r}(a, X_i) + \mathbb{I}[a = A_i] \frac{R_i - \hat{r}(A_i,X_i)}{\max(\tau, \pi_0(A_i|X_i))} \end{align*} verifies the equation. Solutions for \texttt{cIPS} and \texttt{IPS} can be recovered directly, and \texttt{PC} follows the same construction. For the cluster level approaches, we take \texttt{POTEC} as a representative, and we have: \begin{align*} \hat{r}_{\texttt{POTEC}, i}(c, X_i) = \hat{r}^\star_c(X_i) + \mathbb{I}[c = C_i] \frac{R_i - \hat{r}(A_i,X_i)}{\pi_0(C_i|X_i)}\,, \end{align*} The $\hat{r}_{\texttt{MIPS}, i}$ follows as a special case when $\hat{r} = 0$. \end{proof}

\begin{lemma}\label{app:las-existence}
Consider the single-context case and assume a finite action set $\mathcal A$.
For any estimator \texttt{EST} in (\texttt{IPS}, \texttt{cIPS}, \texttt{DR}, \texttt{OffCEM}, \texttt{MIPS}, \texttt{PC}),
there exists a problem instance (i.e., a choice of $r$ and $\pi_0$; and when relevant, a choice of auxiliary objects such as $\hat r$, $h$, $N_\epsilon$)
such that, in the large-$n$ limit,
$$
\hat r_{\textsc{est}}(a) = \mathbbm{1}[a=a_K]
$$
for some optimal action $a_K$.
Similarly, for cluster-based approaches (e.g., \texttt{POTEC} and \texttt{MIPS}), there exists an instance such that
$$
\hat r_{\textsc{est-c}}(c)=\mathbbm{1}[c=c_{|\mathcal C|}]
$$
for some optimal cluster $c_{|\mathcal C|}$.
\end{lemma}

\begin{proof}
We give explicit constructions for \texttt{cIPS} (action-level) and \texttt{POTEC} (cluster-level). The other estimators follow by the same idea:
choose a setting where the estimator becomes linear in $\pi$ with some deterministic coefficient, and pick $r$ (and possibly $\hat r$, $h$, $N_\epsilon$)
so that the resulting linearized reward is one-hot.

\paragraph{Action-level: \texttt{cIPS}.}
Fix $\tau\in[0,1)$.\footnote{If $\tau=1$ and $|\mathcal A|>1$, the simplifying assumption $\pi_0(a)>0$ for all $a$ is incompatible with having some $\pi_0(a)\ge\tau$.
In practice $\tau\ll 1$.}
Choose a logging policy $\pi_0$ with full support and such that
\begin{equation}\label{eq:pi0_tau}
\max_{a\in\mathcal A}\pi_0(a)\ \ge\ \tau,
\end{equation}
Let
$$
a_K \in \arg\max_{a\in\mathcal A}\frac{\pi_0(a)}{\max\{\pi_0(a),\tau\}}.
$$
Under \cref{eq:pi0_tau}, there exists at least one action with $\pi_0(a)\ge\tau$, for which the ratio equals $1$, hence the maximizer satisfies $\pi_0(a_K)\ge\tau$ and therefore
$$
\frac{\pi_0(a_K)}{\max\{\pi_0(a_K),\tau\}}=1.
$$
Now define the reward function
$$
r(a)\ =\ \mathbbm{1}[a=a_K]\frac{\max\{\pi_0(a),\tau\}}{\pi_0(a)}.
$$
This satisfies $r(a)\in[0,1]$ for all $a$ because $r(a)=0$ for $a\neq a_K$, and
$$
r(a_K)=\frac{\max\{\pi_0(a_K),\tau\}}{\pi_0(a_K)}=1
\quad(\text{since }\pi_0(a_K)\ge\tau).
$$
For \texttt{cIPS}, the large-$n$ linearized reward is
$$
\hat r_{\textsc{cips}}(a)=\frac{\pi_0(a)}{\max\{\pi_0(a),\tau\}}\,r(a),
$$
hence
$$
\hat r_{\textsc{cips}}(a)=\frac{\pi_0(a)}{\max\{\pi_0(a),\tau\}}\,
\mathbbm{1}[a=a_K]\frac{\max\{\pi_0(a),\tau\}}{\pi_0(a)}
=\mathbbm{1}[a=a_K],
$$
as desired.

\paragraph{Cluster-level: \texttt{POTEC}.}
We work in the single-context case and consider a clustering map $h:\mathcal A\to\mathcal C$.
Choose $h$ so that $a_K$ forms a singleton cluster:
$$
c_{|\mathcal C|}=h(a_K)=\{a_K\},
\qquad
h(a)\neq c_{|\mathcal C|}\ \ \forall a\neq a_K.
$$
Let rewards be $r(a_K)=1$ and $r(a)=0$ for $a\neq a_K$.
Pick any $\varepsilon\in(0,1/2]$ and define a reward model
$$
\hat r(a_K)=1-\varepsilon,
\qquad
\hat r(a)=\varepsilon \ \ \forall a\neq a_K.
$$
For \texttt{POTEC}, the induced (cluster-level) linearized reward takes the form
$$
\hat r_{\texttt{POTEC}}(c)
=
\max_{a\in c}\hat r(a)
+
\frac{\sum_{a\in c}\pi_0(a)\big(r(a)-\hat r(a)\big)}{\pi_0(c)}.
$$
For the singleton cluster $c_{|\mathcal C|}=\{a_K\}$, we get
$$
\hat r_{\texttt{POTEC}}(c_{|\mathcal C|})
=
(1-\varepsilon)+\frac{\pi_0(a_K)\big(1-(1-\varepsilon)\big)}{\pi_0(a_K)}
=(1-\varepsilon)+\varepsilon=1.
$$
For any other cluster $c\neq c_{|\mathcal C|}$, all its actions satisfy $r(a)=0$ and $\hat r(a)=\varepsilon$, hence
$$
\hat r_{\texttt{POTEC}}(c)
=
\varepsilon
+
\frac{\sum_{a\in c}\pi_0(a)(-\varepsilon)}{\pi_0(c)}
=
\varepsilon-\varepsilon=0.
$$
Therefore $\hat r_{\texttt{POTEC}}(c)=\mathbbm{1}[c=c_{|\mathcal C|}]$.

This concludes the constructions for \texttt{cIPS} and \texttt{POTEC}. The remaining estimators can be handled analogously by choosing
$\pi_0$ (and when relevant, $h$ or $N_\epsilon$) so that the estimator’s linear coefficient on $r(a)$ equals $1$ at a chosen $a_K$
(and equals something finite elsewhere), and then defining $r$ (and possibly $\hat r$) to make the resulting $\hat r_{\textsc{est}}$ one-hot.
\end{proof}

Now we restate \cref{prop:plateaus} and proceed to its proof.

\begin{proposition}[Plateau for linear-in-$\pi$ objectives under softmax]
Consider the single-context case. Let $\hat V(\pi)$ be any objective linear in $\pi$ and let $\pi^\star \in \arg\max_{\pi } \hat V(\pi)
$ denote a maximizer over the probability simplex on the effective action space $\mathcal A_{\mathrm{eff}}$
(of size $K_{\mathrm{eff}}=|\mathcal A_{\mathrm{eff}}|$). Let $\{\pi_{\theta_t}\}_{t\ge 1}$ be the iterates of gradient ascent
on $\theta \mapsto \hat V(\pi_\theta)$ with a linear softmax policy
$\pi_\theta(a)=\exp(\theta_a)/\sum_{a'\in\mathcal A_{\mathrm{eff}}}\exp(\theta_{a'})$ and step sizes $\eta_t\in(0,1]$.
Then there exists a problem instance such that gradient ascent cannot escape a suboptimal region before
$t_0=C\,K_{\mathrm{eff}}=\mathcal O(K_{\mathrm{eff}})$ iterations, in the sense that
$$
\forall t\le t_0:\qquad \hat V(\pi^\star)-\hat V(\pi_{\theta_t}) \ge 0.9.
$$
\end{proposition}

\begin{proof}
The proof follows the same technique as \citep[Theorem 1]{gravitational_pull}.
By Lemma~\ref{app:las-existence}, there exists an instance (single context) for which the linearized reward is one-hot:
$$
\hat r_{\textsc{est}}(a)=\mathbbm{1}[a=a_K]
\qquad\text{for some }a_K\in\mathcal A_{\mathrm{eff}}.
$$
Hence, for any policy $\pi$ supported on $\mathcal A_{\mathrm{eff}}$,
$$
\hat V(\pi)=\sum_{a\in\mathcal A_{\mathrm{eff}}}\pi(a)\hat r_{\textsc{est}}(a)=\pi(a_K).
$$
The maximizer over the simplex is therefore $\pi^\star=\delta_{a_K}$, and
$$
\hat V(\pi^\star)=1,
\qquad
\hat V(\pi^\star)-\hat V(\pi_{\theta})=1-\pi_\theta(a_K).
$$
(Notice that $\sup_{\theta}\hat V(\pi_\theta)=1$ as well, although the supremum is not attained by any finite $\theta$
when $K_{\mathrm{eff}}\ge 2$). We now upper bound the gradient norm. For the softmax parametrization,
$$
\Big\|\nabla_\theta \hat V(\pi_\theta)\Big\|_2
\le \sqrt{2}\,\pi_\theta(a_K)\big(1-\pi_\theta(a_K)\big),
$$
where the bound follows by a direct computation (as in \citep{gravitational_pull}).

Define the update $\theta_{t+1}=\theta_t+\eta_t\nabla_\theta \hat V(\pi_{\theta_t})$ and
split iterations into
$$
t_{\mathrm{good}}=\{t\ge 1:\ \pi_{\theta_{t+1}}(a_K)>\pi_{\theta_t}(a_K)\},
\qquad
t_{\mathrm{bad}}=\{t\ge 1:\ \pi_{\theta_{t+1}}(a_K)\le \pi_{\theta_t}(a_K)\}.
$$
For $t\in t_{\mathrm{bad}}$,
$$
\frac{1}{\pi_{\theta_t}(a_K)}-\frac{1}{\pi_{\theta_{t+1}}(a_K)}\le 0.
$$
For $t\in t_{\mathrm{good}}$, using Lemma~\ref{lem:smoothness} (the $5/2$-smoothness of
$\theta\mapsto \hat V(\pi_\theta)$) and $\eta_t\in(0,1]$, we obtain
$$
\pi_{\theta_{t+1}}(a_K)-\pi_{\theta_t}(a_K)
\le \frac{9}{2}\,\pi_{\theta_t}(a_K)^2,
$$
and therefore (since $\pi_{\theta_{t+1}}(a_K)\ge \pi_{\theta_t}(a_K)>0$),
$$
\frac{1}{\pi_{\theta_t}(a_K)}-\frac{1}{\pi_{\theta_{t+1}}(a_K)}
=
\frac{\pi_{\theta_{t+1}}(a_K)-\pi_{\theta_t}(a_K)}{\pi_{\theta_{t+1}}(a_K)\pi_{\theta_t}(a_K)}
\le \frac{9}{2}.
$$
Summing over $s=1,\dots,t-1$ yields
$$
\frac{1}{\pi_{\theta_1}(a_K)}-\frac{1}{\pi_{\theta_t}(a_K)}
=\sum_{s=1}^{t-1}\left(\frac{1}{\pi_{\theta_s}(a_K)}-\frac{1}{\pi_{\theta_{s+1}}(a_K)}\right)
\le \frac{9}{2}\,t.
$$

Assume a standard symmetric initialization so that $\pi_{\theta_1}(a_K)=1/K_{\mathrm{eff}}$.
Pick any constant $c\ge 11$ and take $K_{\mathrm{eff}}$ large enough so that $\pi_{\theta_1}(a_K)\le 1/c$.
If $t\le \frac{2}{9c}\,K_{\mathrm{eff}}$, then
$$
\frac{1}{\pi_{\theta_t}(a_K)}
\ge \frac{1}{\pi_{\theta_1}(a_K)}-\frac{9}{2}\,t
\ge \frac{1}{\pi_{\theta_1}(a_K)}\left(1-\frac{1}{c}\right)
\ge c-1\ge 10,
$$
hence $\pi_{\theta_t}(a_K)\le 1/10$, and thus
$$
\hat V(\pi^\star)-\hat V(\pi_{\theta_t})
=1-\pi_{\theta_t}(a_K)\ge 0.9.
$$
This proves the claim with $t_0=\frac{2}{9c}K_{\mathrm{eff}}$.
\end{proof}

\begin{proposition}
Even for a single context $x$, deterministic rewards, there is problem where IPS-based learning with a linear softmax policy $\pi_\theta(a) \propto \exp(\langle \theta, \phi(x, a) \rangle) \mathbb{I}[a \in \mathcal{A}_\text{eff}]$ can have a number of local maxima exponential in the number of effective actions $K_\text{eff}$.
\end{proposition}

\begin{proof}
Let \texttt{EST} an off-policy estimators considered in the paper with an action-level policy. By Lemma~\ref{app:las-linearisation}, we have:
\begin{align}
    \hat{V}_{\textsc{est}}(\pi) = \frac{1}{n}\sum_{i = 1}^n \mathbb{E}_{a \sim \pi(\cdot|x_i)}\left[\hat{r}_{\textsc{est}, i}(a, x_i) \right]\,,
\end{align}
In a single context setting, it becomes:
\begin{align}
    \hat{V}_{\textsc{est}}(\pi_\theta) &=  \mathbb{E}_{a \sim \pi_\theta(\cdot)}\left[\frac{1}{n}\sum_{i = 1}^n\hat{r}_{\textsc{est}, i}(a) \right]\,, \\
    &=  \langle \frac{1}{n}\sum_{i = 1}^n\hat{r}_{\textsc{est}, i}, \pi_\theta \rangle\,.
\end{align}
This also holds for estimators with policies in the cluster level, as we still have:
\begin{align}
    \hat{V}_{\textsc{est-c}}(\pi_\theta) &=  \mathbb{E}_{c \sim \pi_\theta(\cdot)}\left[\frac{1}{n}\sum_{i = 1}^n\hat{r}_{\textsc{est-c}, i}(c) \right]\,, \\
    &=  \langle\frac{1}{n}\sum_{i = 1}^n\hat{r}_{\textsc{est-c}, i}, \pi_\theta \rangle\,.
\end{align}
These softmax policies are all defined on the effective action space $\mathcal{A}_\text{eff}$, be it a subset of the action space $\mathcal{A}$ or the discrete cluster space $\mathcal{C}$. Using the linearity of the objective, we can directly apply Theorem 1 from \cite{surrogate} and obtain our result.
\end{proof}

Finally, we also restate \cref{prop:concave}, and provide its proof.

\begin{proposition}
For an $\ell_2$ regularized (substituting $\frac{\lambda}{2} ||\theta||^2$, with $\lambda >0$), linear softmax policy $\pi_\theta$, the PWLL objective $\hat{U}^{\texttt{g}}(\pi_\theta)$ defined as:
\begin{align*}
\hat{U}^{\texttt{g}}(\pi) = \frac{1}{n} \sum_{i=1}^n g(R_i, \pi_0(A_i\mid X_i)) \log \pi(A_i\mid X_i) \,,
\end{align*}
is $\lambda$-strongly concave. Without regularization, the objective is concave.
\end{proposition}

\begin{proof}

For any $x$ and $a \in \mathcal{A}_\text{eff}(x)$, we have: 
$$\pi_\theta(a|x) = \frac{\exp(\langle \theta, \phi(x,a)\rangle)}{\sum_{a' \in \mathcal{A}_\text{eff}(x)} \exp(\langle \theta, \phi(x,a')\rangle)}\,,$$
optimizing an $\ell_2$ regularized linear softmax, giving:
\begin{align*}
\hat{L}^{g,\lambda}(\pi) = \hat{U}^{\texttt{g}}(\pi) -  \frac{\lambda}{2} ||\theta||^2 \,,
\end{align*}
with $\lambda > 0$ and recall that $g \ge 0$. For strong concavity, we need to show that the Hessian $\nabla^2_\theta \hat{U}^{\texttt{g}}(\pi_\theta)$ is negative definite with eigenvalues bounded away from zero.

The gradient with respect to $\theta$ is:
$\nabla_\theta \hat{U}^{\texttt{g}}(\pi_\theta) = \frac{1}{n} \sum_{i=1}^n g(R_i, \pi_0(A_i\mid X_i)) \nabla_\theta \log \pi_\theta(A_i|X_i) - \lambda \theta$

For the softmax policy:
$$\nabla_\theta \log \pi_\theta(a|x) = \phi(x,a) - \sum_{a'} \pi_\theta(a'|x) \phi(x,a') = \phi(x,a) - \mathbb{E}_{A\sim \pi_\theta(\cdot|x)}[\phi(x,A)]$$

Therefore:
$\nabla_\theta \hat{U}^{\texttt{g}}(\pi_\theta) = \frac{1}{n} \sum_{i=1}^n g(R_i, \pi_0(A_i\mid X_i)) \left(\phi(X_i,A_i) - \mathbb{E}_{A \sim \pi_\theta(\cdot|X_i)}[\phi(X_i,A)]\right) - \lambda \theta$

Taking the second derivative:
$\nabla^2_\theta \hat{U}^{\texttt{g}}(\pi_\theta) = -\frac{1}{n} \sum_{i=1}^n g(R_i, \pi_0(A_i\mid X_i)) \nabla_\theta \mathbb{E}_{A \sim \pi_\theta(\cdot|X_i)}[\phi(X_i,A)] - \lambda I_d$, where $I_d$ is the $d \times d$ identity matrix. The gradient of the expectation is:
$$\nabla_\theta \mathbb{E}_{A \sim \pi_\theta(\cdot|x)}[\phi(x,A)] = \sum_{a} \nabla_\theta \pi_\theta(a|x) \phi(x,a)$$

Using $\nabla_\theta \pi_\theta(a|x) = \pi_\theta(a|x)(\phi(x,a) - \mathbb{E}_{A \sim \pi_\theta(\cdot|x)}[\phi(x,A)])$:

$$\nabla_\theta \mathbb{E}_{A \sim \pi_\theta(\cdot|x)}[\phi(x,A)] = \sum_{a} \pi_\theta(a|x)(\phi(x,a) - \mathbb{E}_{A \sim \pi_\theta(\cdot|x)}[\phi(x,A)]) \phi(x,a)^\top$$

This simplifies to:
$$\nabla_\theta \mathbb{E}_{A \sim \pi_\theta(\cdot|x)}[\phi(x,A)] = \text{Cov}_{A \sim \pi_\theta(\cdot|x)}[\phi(x, A)]$$

where $\text{Cov}_{A \sim \pi_\theta(\cdot|x)}[\phi(x, A)] = \mathbb{E}_{A \sim \pi_\theta(\cdot|x)}[\phi(x, A)\phi(x, A)^\top] - \mathbb{E}_{A \sim \pi_\theta(\cdot|x)}[\phi(x,A)]\mathbb{E}_{A \sim \pi_\theta(\cdot|x)}[\phi(x,A)]^\top$

Therefore:
$$\nabla^2_\theta \hat{U}^{\texttt{g}}(\pi_\theta) = -\frac{1}{n} \sum_{i=1}^n g(R_i, \pi_0(A_i\mid X_i)) \text{Cov}_{A \sim \pi_\theta(\cdot|X_i)}[\phi(X_i, A)] - \lambda I_d$$

We can write this as:
$\nabla^2_\theta \hat{U}^{\texttt{g}}(\pi_\theta) = -H - \lambda I_d$

where $H = \frac{1}{n} \sum_{i=1}^n g(R_i, \pi_0(A_i\mid X_i)) \text{Cov}_{A \sim \pi_\theta(\cdot|X_i)}[\phi(X_i, A)]$ is positive semi-definite. To see this explicitly, for any vector $v \in \mathbb{R}^d$:
$$v^\top \text{Cov}_{A \sim \pi_\theta(\cdot|X_i)}[\phi(X_i, A)] v = \text{Var}_{A \sim \pi_\theta(\cdot|X_i)}[v^\top \phi(X_i, A)] \geq 0\,,$$
with the positivity of $g$, this ensures $H$ is positive semi-definite. Then we have:
$$v^\top \nabla^2_\theta \hat{U}^{\texttt{g}}(\pi_\theta) v = -v^\top H v - \lambda v^\top v = -v^\top H v - \lambda \|v\|^2\,,$$
meaning that when $v \neq 0$, we get
$v^\top \nabla^2_\theta \hat{U}^{\texttt{g}}(\pi_\theta) v \leq -\lambda \|v\|^2 < 0\,.$

This shows the Hessian is negative definite with all eigenvalues bounded above by $-\lambda < 0$. Therefore, $\ell_2$ regularized $\hat{U}^{\texttt{g}}(\pi_\theta)$ is $\lambda$-strongly concave. In addition, when $\lambda = 0$, the hessian is negative semi-definite, giving simple concavity.
\end{proof}

\section{Stochastic Optimization Convergence Guarantees for PWLL}\label{sec:las-optimization-convergence}

We analyze the convergence rates of stochastic gradient methods on the PWLL objective. We formulate this as the minimization of the finite-sum loss $f(\theta) = -\hat{U}_g(\pi_\theta)$:
\begin{align}
    f(\theta) = \frac{1}{n} \sum_{i=1}^n f_i(\theta), \quad \text{where } f_i(\theta) = -g_i \log \pi_\theta(A_i \mid X_i),
\end{align}
where $g_i = g(R_i, \pi_0(A_i |X_i))$. We adopt the linear softmax policy parametrization in \cref{eq:las-softmax} with $s_\theta(x, a) = \phi(x, a)^\top \theta$ (lightweight parametrization in \cref{eq:las-softmax_scores}). We note that our analysis extends naturally to the heavyweight parametrization in \cref{eq:las-softmax_scores}.

\subsection{Assumptions and Regularity}
\label{sec:las-assumptions}

To establish problem-dependent convergence bounds, we rely on the following structural assumptions regarding the feature space and the importance weights.

\begin{assumption}[Bounded features]
\label{ass:bounded-features}
For all context-action pairs $(x, a) \in \mathcal{X} \times \mathcal{A}$, the feature representations are bounded in Euclidean norm:
$$
\|\phi(x, a)\|_2 \leq H.
$$
\end{assumption}

\begin{assumption}[Bounded weighting function]
\label{ass:bounded-weights}
The weights $g_i = g(R_i, \pi_0(A_i|X_i))$ computed on the static dataset are strictly positive and bounded. That is, for all $i \in \{1, \dots, n\}$:
$$
0 < g_i \leq G_{\max}.
$$
\end{assumption}

Assumptions \ref{ass:bounded-features} and \ref{ass:bounded-weights} are sufficient to establish the smoothness and bounded variance of the objective $f(\theta)$. We formally derive these properties in the following proposition.

\begin{proposition}[Regularity and Variance Bounds]
\label{prop:regularity}
Under Assumptions \ref{ass:bounded-features} and \ref{ass:bounded-weights}, the objective $f(\theta)$ satisfies the following properties:
\begin{enumerate}
    \item \textbf{Global Smoothness:} The objective is $\bar{L}$-smooth with $\bar{L} = G_{\max} H^2$.
    \item \textbf{Bounded Single-Sample Variance:} The variance of the stochastic gradient for a single sample is bounded by $\bar{\sigma}^2 = 4 G_{\max}^2 H^2$.
    \item \textbf{Bounded Mini-Batch Variance:} For a mini-batch of size $b$, the variance is bounded by $\bar{\sigma}_b^2 = \frac{4 G_{\max}^2 H^2}{b}$.
\end{enumerate}
\end{proposition}

\begin{proof}
\textit{1. Smoothness:} The Hessian of the objective is the weighted sum of the feature covariance matrices under the policy $\pi_\theta$:
$$
\nabla^2 f(\theta) = \frac{1}{n} \sum_{i=1}^n g_i \text{Cov}_{A \sim \pi_\theta(\cdot|X_i)} [\phi(X_i, A)].
$$
The spectral norm of a covariance matrix is bounded by the maximum squared norm of its random vectors. Thus, using \cref{ass:bounded-features} we get that $\|\nabla^2 f(\theta)\|_{\text{op}} \leq \frac{1}{n} \sum_{i=1}^n g_i H^2 \leq G_{\max} H^2$.

\textit{2. Single-Sample Variance:} We first bound the norm of the gradient for an arbitrary sample $i$. The gradient is $\nabla f_i(\theta) = -g_i (\phi(X_i, A_i) - \mathbb{E}_{A \sim \pi_\theta(\cdot \mid X_i)}[\phi(X_i, A)])$. Using the triangle inequality and Assumption \ref{ass:bounded-features}:
$$
\|\nabla f_i(\theta)\|_2 \leq g_i \left( \|\phi(X_i, A_i)\|_2 + \|\mathbb{E}_{A \sim \pi_\theta(\cdot \mid X_i)}[\phi(X_i, A)]\|_2 \right) \leq G_{\max}(H + H) = 2 G_{\max} H.
$$
Let $\xi = \nabla f_I(\theta)$ be the stochastic gradient sampled uniformly from the dataset. The variance is bounded by the second moment:
$$
\mathrm{Var}(\xi) \leq \mathbb{E}[\|\xi\|^2] = \frac{1}{n}\sum_{i=1}^n \|\nabla f_i(\theta)\|^2 \leq (2 G_{\max} H)^2 = 4 G_{\max}^2 H^2.
$$

\textit{3. Mini-Batch Variance:} Let the mini-batch gradient be $\bar{g}_t = \frac{1}{b} \sum_{j=1}^b \nabla f_{i_j}(\theta)$, where indices are sampled independently with replacement. Using the standard variance reduction property for independent variables:
$$
\mathbb{E}[\|\bar{g}_t - \nabla f(\theta)\|^2] = \frac{1}{b} \mathbb{E}[\|\nabla f_I(\theta) - \nabla f(\theta)\|^2] \leq \frac{4 G_{\max}^2 H^2}{b}.
$$
\end{proof}

Based on Proposition \ref{prop:regularity}, we define the following global problem-dependent constants on which our convergence rates depend:
\begin{itemize}
    \item $\bar{L} = G_{\max}H^2$: Smoothness constant.
    \item $\bar{\sigma}^2 = 4G_{\max}^2H^2$: Upper bound on the gradient variance for a single sample.
    \item $\bar{\sigma}_b^2 = \frac{4G_{\max}^2H^2}{b}$: Upper bound on the gradient variance for a mini-batch of size $b$.
\end{itemize}

\subsection{PWLL without $\ell_2$ regularization}
We begin by analyzing the standard unregularized PWLL objective. Here, the objective $f(\theta)$ is convex but not necessarily strongly convex. This implies the loss landscape may contain multiple minimizers rather than a unique global minimum. Consequently, we characterize convergence in terms of $\hat{U}_g(\pi_{\theta^{\text{opt}}}) - \hat{U}_g(\pi_{\bar{\theta}_T})$ (instead of $\|\theta_t - \theta_{n}^{\text{opt}}\|$). Here, $\theta^{\text{opt}} \in \arg\max_\theta \hat{U}_g(\pi_\theta)$ is an optimal parameter and $\bar{\theta}_T$ is the average of the SGA iterates.  

\begin{proposition}
\label{thm:convex-constant}
Let $\theta^{\text{opt}} \in \arg\max_\theta \hat{U}_g(\pi_\theta)$ be an optimal parameter. If the learning rate satisfies $0 < \eta \leq \frac{1}{4\bar{L}}$, then by \cite[Theorem 6.9]{garrigos2023}, the iterates of mini-batch SGA satisfy:
$$
\mathbb{E}\left[\hat{U}_g(\pi_{\theta^{\text{opt}}}) - \hat{U}_g(\pi_{\bar{\theta}_T})\right] \leq \frac{\|\theta_0 - \theta^{\text{opt}}\|^2}{\eta T} + \frac{8\eta G_{\max}^2 H^2}{b}
$$
where $\bar{\theta}_T$ is the average of the iterates.
\end{proposition}

Proposition \ref{thm:convex-constant} highlights the trade-off inherent to constant step-size SGA: a larger $\eta$ accelerates the decay of the initial error (first term) but increases the asymptotic noise floor (second term). For a fixed horizon $T$, one can recover a convergence rate of $\mathcal{O}(1/\sqrt{T})$ by setting $\eta \propto 1/\sqrt{T}$, which balances both terms.

\subsection{PWLL with $\ell_2$ regularization}
We now move to the $\ell_2$-regularized case where the PWLL objective is strongly concave (\cref{prop:concave}). Precisely, we consider the regularized objective $\tilde{U}^\lambda(\theta) = \hat{U}_g(\pi_\theta) - \frac{\lambda}{2}\|\theta\|^2$.

Strong convexity implies the existence of a unique global minimizer. This allows us to guarantee convergence of the parameters $\theta_t$ themselves, which is a stronger condition than value convergence.

\begin{proposition}
\label{thm:strongly-convex-constant}
Let $\theta_{n,\lambda}^{\text{opt}} = \arg\max_\theta \tilde{U}^\lambda(\theta)$ be the unique optimal parameter. If the learning rate satisfies $0 < \eta \leq \frac{1}{2 (G_{\max}H^2 + \lambda)}$, then by \cite[Theorem 6.12]{garrigos2023}:
$$
\mathbb{E}\left[\|\theta_t - \theta_{n,\lambda}^{\text{opt}}\|^2\right] \leq (1-\eta\lambda)^t \|\theta_0 - \theta_{n,\lambda}^{\text{opt}}\|^2 + \frac{8\eta G_{\max}^2 H^2}{\lambda b}
$$
\end{proposition}

The regularized case demonstrates a convergence rate that is significantly faster than the rate of the unregularized case.

\section{Additional 
Experiments}\label{app:las-additional_experiments}

\subsection{Detailed Experimental Setting}

\textbf{Experimental Setting.} 
\renewcommand{\arraystretch}{1.5}
\begin{table}[H]
\caption{Statistics of Post Processed Datasets}
\label{tab:las-stats}
\begin{center}
{\small
\begin{tabular}{lcc}
\toprule
\textbf{Dataset}    & \textbf{Num. of actions} & \textbf{Num. of samples}\\
\midrule
\texttt{MovieLens}  & $60,000$ &   $132,744$ \\
 \texttt{Twitch}  & $200,000$ &  $400,000$ \\
\texttt{GoodReads}  & $1,000,000$  & $400,000$   \\
\bottomrule
\end{tabular}
}
\end{center}
\vskip -0.1in
\end{table}

Our experimental setup is designed to study the behavior of the different policy learning paradigms in large action spaces. To this end, we use three large action spaces collaborative filtering datasets: Movielens \citep{movielens}, Twitch \citep{twitch} and GoodReads \citep{gr2} that are preprocessed to obtain a user-item interaction matrix. We follow the exact procedure of \citet{sakhi2023fast} to pre-process the datasets. The statistics of the obtained datasets are described in Table~\ref{tab:las-stats}. For each user, we keep half of its history as the context $x$, and use the other half of the history as the products with positive reward, which align the learned policies to recommend new and relevant items. We direct the interested readers to \citet{sakhi2023fast} for a detailed description of the experimental setup. 

The large action space scenario restricts the policies used to the inner product parametrization \citep{aouali2022reward}. This parametrization is essential to leverage Maximum Inner Product Search algorithms \citep{mips} for fast query response. In particular, we adopt policies of the following form:
\begin{align*}
    \pi_\theta(a|x) \propto \exp(\langle \phi_\Gamma(x), \beta_a\rangle)\,,
\end{align*}
with the learnable parameter $\theta = [\Gamma, \beta]$, $\phi_\Gamma: \mathcal{X} \rightarrow \mathbb{R}^\ell$ defines the context embedding function in $\mathbb{R}^\ell$ and $\beta$ the actions embeddings of size $K \times \ell$. To define our policies, we start by extracting action embeddings $\beta_0$ using an SVD decomposition of the user-item matrix. These embeddings help us define the context embedding function $\phi_\Gamma$ and our logging policy $\pi_0$. $\phi_0$ is set to the average embeddings of the observed actions in the contexts and is fixed for the logging policy $\pi_0$. Using the SVD action embeddings $\beta_0$, we define our logging policy $\pi_0$ as:
\begin{align*}
    \pi_0(a|x) \propto \exp\left(\frac{1}{t}\langle \phi_0(x), \beta_{0, a}\rangle \right) \mathbb{I}\left[a \in \textsc{top}^{k_0}(x) \right]\,,
\end{align*}
with $t$ the temperature of the logging policy, and $k_0$ define the support of the logging policy, concentrating on the top $k_0$ actions with:
$\textsc{top}^{k_0}(x) = \argsort_{a_1, \cdots, a_{k_0}} \langle \phi_0(x), \beta_{0, a}\rangle $.

If not explicitly stated, $k_0$ is set to $100$  and the temperature at $t = 1$ in all experiments. This policy is used to collect the offline dataset $\mathcal{D}_n = \{X_i, A_i, R_i \}_{i \in [n]}$ on which all trainings are conducted. For each $i\in [n]$ in the processed dataset, $X_i$ is the user history, $A_i$ is the action played by the logging policy $\pi_0(\cdot|X_i)$ and $R_i = \mathbbm{1}[A_i \in H_i]$ the observed reward, which is if the action played is in the hidden items of user $i$.

\textbf{Trained Policies Parameterizations.} We adopt two parameterizations of the trained policies. The first one is a  \textbf{heavyweight} parametrization, and focuses on learning the embeddings of the actions $\beta$ (be it $\mathcal{A}$ of size $K$ or $\mathcal{C}$ of size $|\mathcal{C}|$), meaning that $\theta$ in this case is $\beta$. For action-level policies, this gives $\beta \in \mathbb{R}^{K \times \ell}$ and for any $x$:
\begin{align*}
    \pi_\beta(a|x) = \frac{\exp(\langle \phi_0(x), \beta_a\rangle)}{\sum_{a' \in \mathcal{A}_{\text{eff}}(x)}\exp(\langle \phi_0(x), \beta_{a'}\rangle)}\,,
\end{align*}
with $\mathcal{A}_{\text{eff}}(x) \subset \mathcal{A}$, which depends on the choice of the practitioner, for example $\mathcal{A}_{\text{eff}}(x) = S_0(x)$, the support of $\pi_0$ for context $x$ when we optimize IPS objectives. For cluster-level policies, this gives a $\beta \in \mathbb{R}^{|\mathcal{C}| \times \ell}$ and for any $x$:
\begin{align*}
    \pi_\beta(c|x) = \frac{\exp(\langle \phi_0(x), \beta_c\rangle)}{\sum_{c' \in \mathcal{C}}\exp(\langle \phi_0(x), \beta_{c'}\rangle)}\,.
\end{align*}
This is used by default if nothing is explicitly stated. 

We have also define a \textbf{lightweight} parametrization, where only a small projection $W \in \mathbb{R}^{\ell \times \ell}$ is learned, giving in action level policies:
\begin{align*}
    \pi_W(a|x) = \frac{\exp(\langle \phi_0(x)W, \beta_{a, 0}\rangle)}{\sum_{a' \in \mathcal{A}_{\text{eff}}(x)}\exp(\langle \phi_0(x)W, \beta_{a', 0}\rangle)}\,,
\end{align*}
using $\beta_0$, the embeddings of $\pi_0$. For cluster level policies, we first define $\bar{\beta}_0 \in \mathbb{R}^{|\mathcal{C}| \times \ell}$ with $\bar{\beta}_{0, c} = \frac{1}{|c|}\sum_{a \in c} \beta_{0, a}$, and use it to define the cluster level policy:
\begin{align*}
    \pi_W(c|x) = \frac{\exp(\langle \phi_0(x)W, \bar{\beta}_{c, 0}\rangle)}{\sum_{c' \in \mathcal{C}}\exp(\langle \phi_0(x)W, \bar{\beta}_{c', 0}\rangle)}\,.
\end{align*}

\textbf{Reward Model.} The reward model used $\hat{r}$ is learned using regularized linear regression the collected interaction data, with $\hat{r}(x, a) = \langle \phi(x), \theta_a \rangle$.

\textbf{Clustering and $\epsilon$ used.} We use the embeddings $\beta_0$, combined with K-means clustering to find our clusters. The number of clusters is set to $2000$ for all datasets and experiments. For \texttt{PC}, the $\ell_2$ threshold $\epsilon$ is set to $0.1$.

\subsection{Additional results}\label{app:las-additional_results}

\begin{figure}[H]
  \centering
  \includegraphics[width=0.8\linewidth]{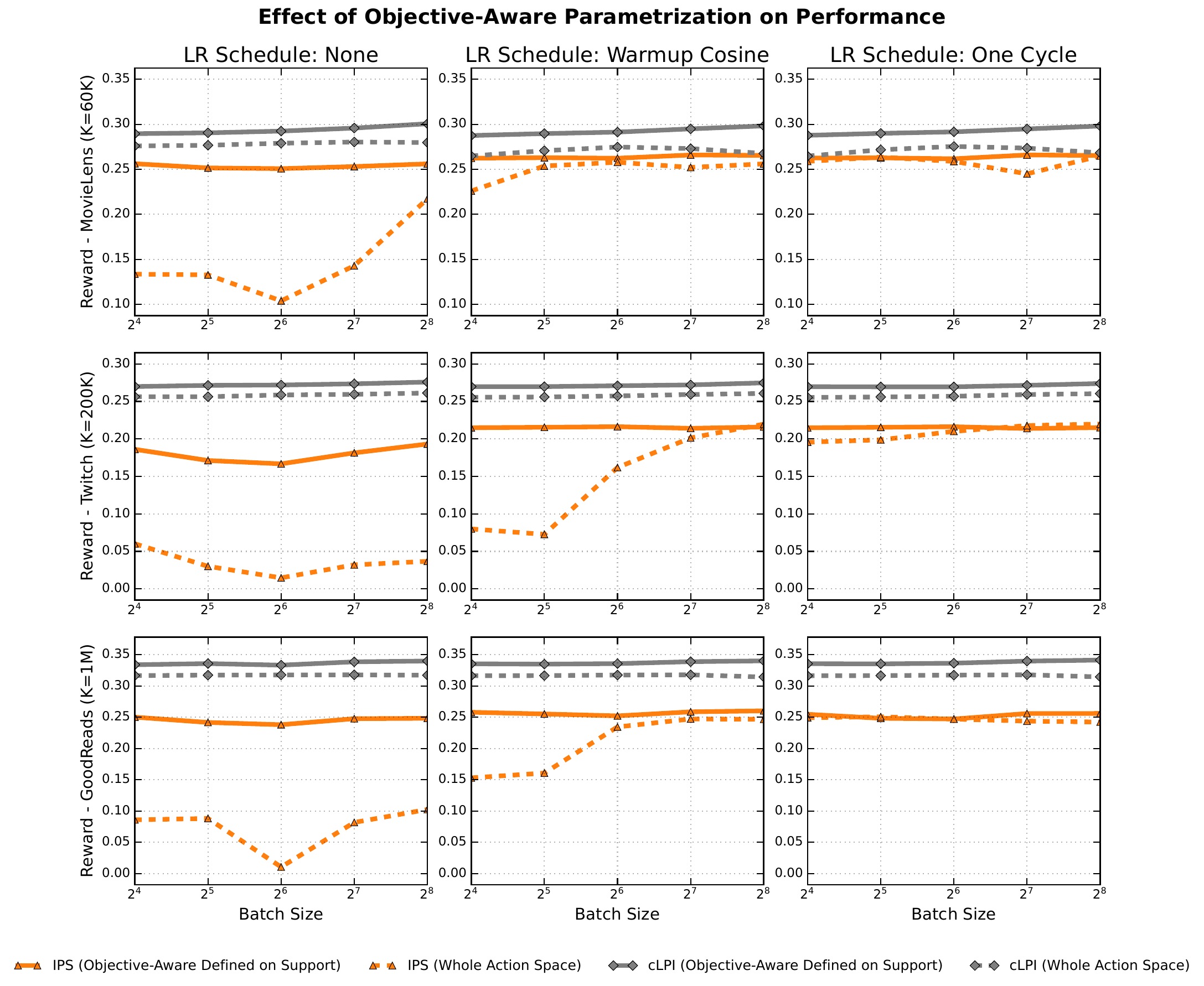}
  \caption{The effect of objective-aware parametrization for \texttt{IPS} and \texttt{cLPI} on three large-scale datasets}
  \label{fig:las-ips_support_comparison-all}
\end{figure}

\textbf{Benefits of objective-aware parametrization.} \cref{fig:las-ips_support_comparison-all} shows the effect of objective-aware policy parameterizations for two different objectives and three large action space datasets.

\begin{figure}[H]
  \centering
  \includegraphics[width=0.7\linewidth]{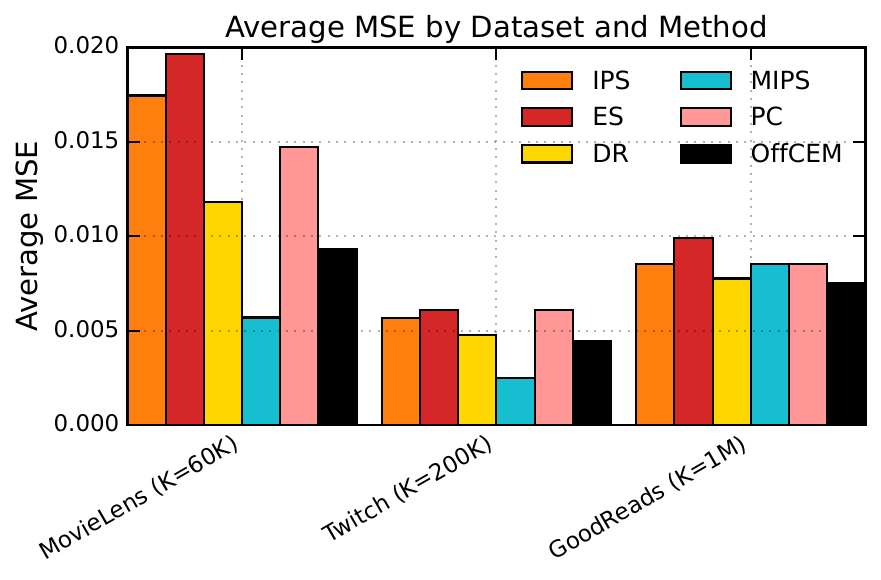}
  \caption{Average MSE by Dataset and Method. Several methods are excluded from the figure, as their high MSE values would distort the scale and obscure the comparison.}
  \label{fig:las-mse-average}
\end{figure}

\textbf{Average MSE.} \cref{fig:las-mse-average} shows the average MSE by dataset and method. Several methods are excluded from the figure, as their high MSE values would distort the scale and obscure the comparison.

\begin{figure}[H]
    \centering

    \begin{subfigure}{0.32\linewidth}
        \centering
        \includegraphics[width=\linewidth]{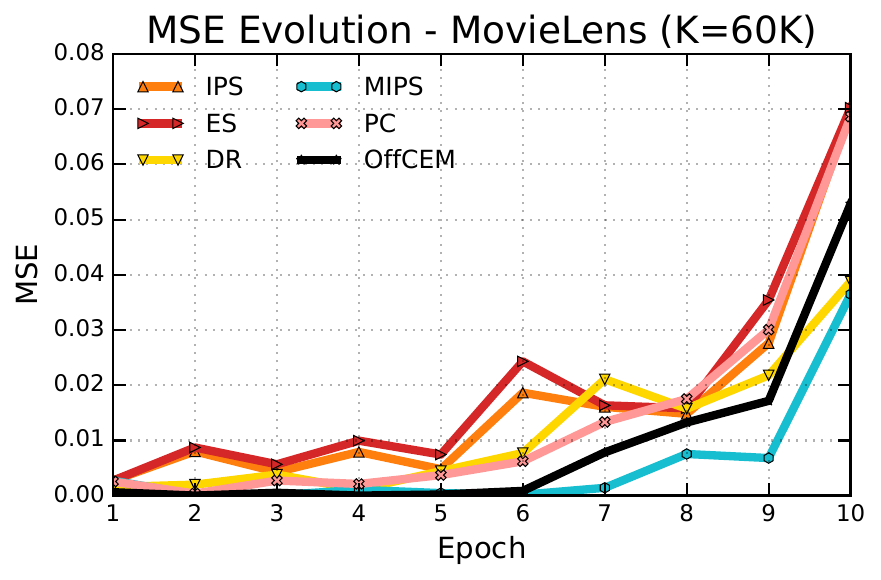}
        \caption{\texttt{MovieLens}}
        \label{fig:las-mse-progress-movielens}
    \end{subfigure}
    \hfill
    \begin{subfigure}{0.32\linewidth}
        \centering
        \includegraphics[width=\linewidth]{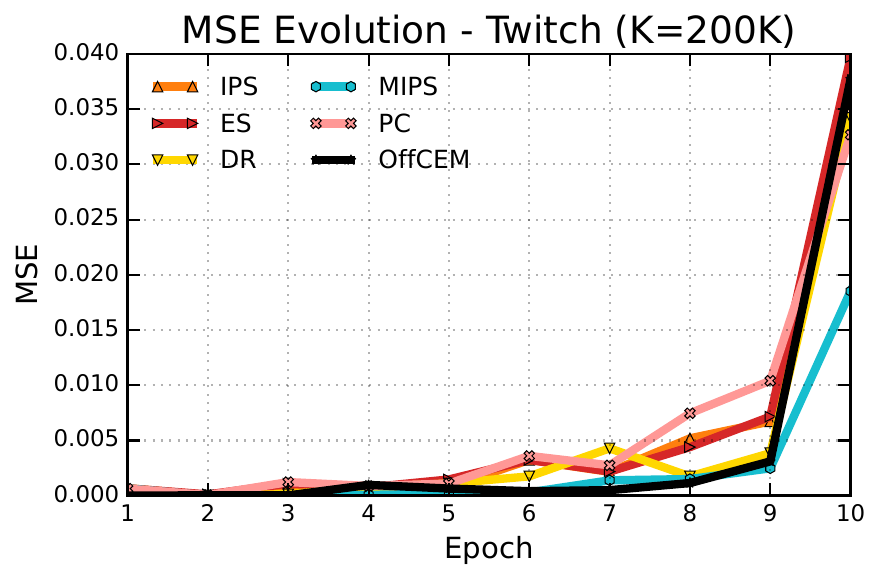}
        \caption{\texttt{Twitch}}
        \label{fig:las-mse-progress-twitch}
    \end{subfigure}
    \hfill
    \begin{subfigure}{0.32\linewidth}
        \centering
        \includegraphics[width=\linewidth]{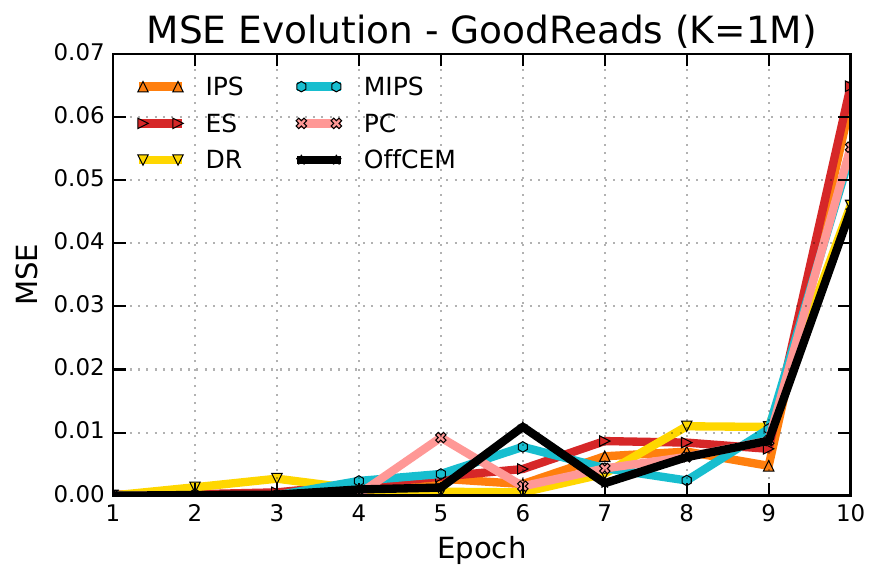}
        \caption{\texttt{GoodReads}}
        \label{fig:las-mse-progress-goodreads}
    \end{subfigure}

    \caption{MSE progression over 10 epochs across datasets.}
    \label{fig:las-mse-progress-all}
\end{figure}

\textbf{MSE progress during training.} \cref{fig:las-mse-progress-movielens,fig:las-mse-progress-twitch,fig:las-mse-progress-goodreads} show the progress of the MSE over 10 epochs on all three datasets. Several methods are excluded from the figure, as their high MSE values would distort the scale and obscure the comparison.

\subsection{Results Averaging Different Seeds}

In this experiment, we analyze the reward evolution of representative PWLL and IPS-based methods on the three considered datasets. We compare two distinct optimization configurations: (i) a standard off-the-shelf Adam optimizer, and (ii) a carefully tuned setup using Adam with an optimized batch size and a one-cycle learning-rate scheduler. This comparison enables us to isolate the effect of optimization on stability and convergence. Each method is evaluated over 5 random seeds, and we report the mean reward along with a shaded standard deviation region to visualize sensitivity to optimization randomness.

\begin{figure}[H]
  \centering
  \includegraphics[width=0.8\linewidth]{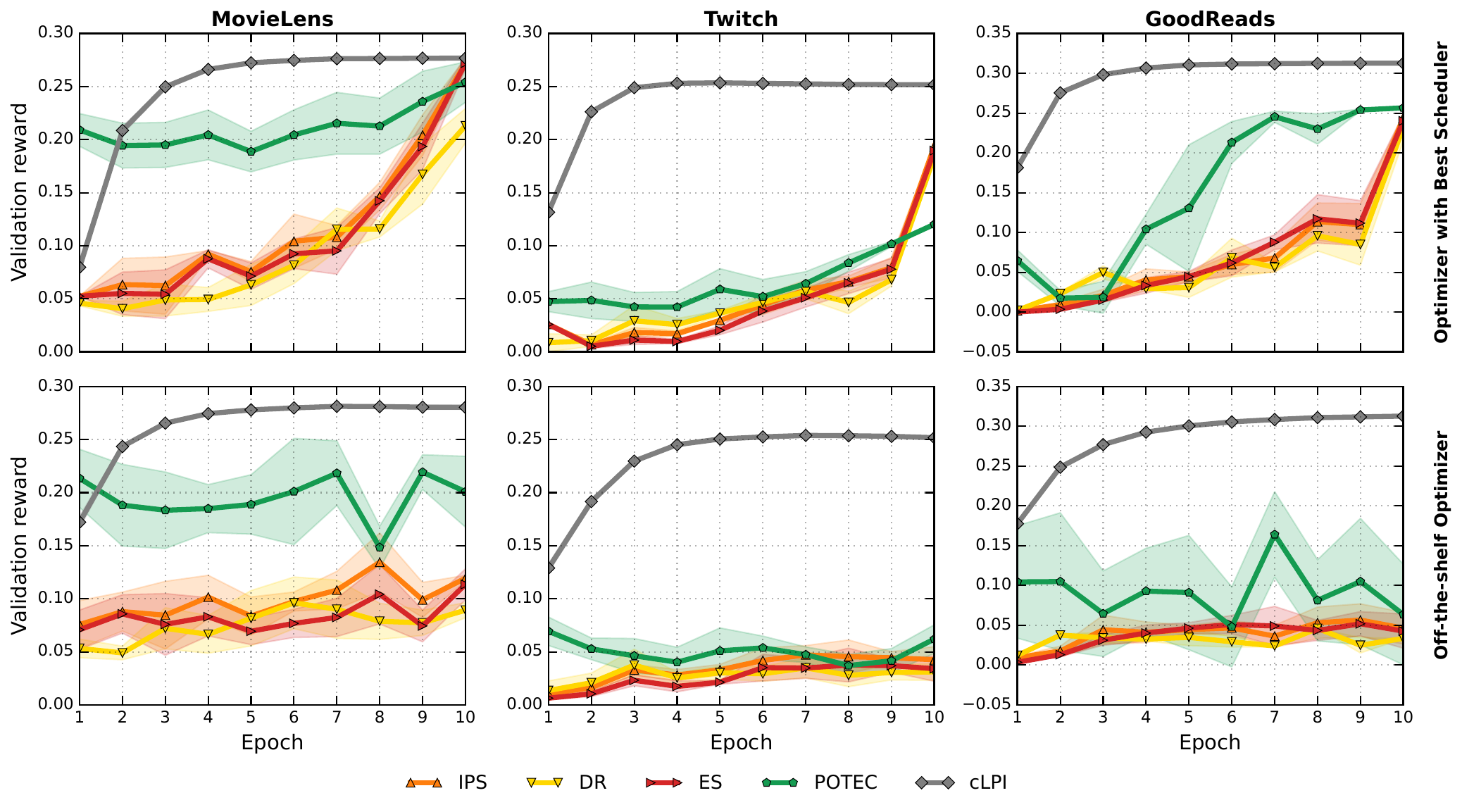}
  \caption{\texttt{cLPI} vs IPS-Based methods: Evolution of rewards averaged over 5 different seeds. \texttt{cLPI} is more stable to optimize and reaches better policies.}
  \label{fig:las-ablation_seeds_cLPI}
\end{figure}

In Figure~\ref{fig:las-ablation_seeds_cLPI}, across all datasets and optimization settings, we observe that IPS-based methods (\texttt{cIPS}, \texttt{IX}, and even \texttt{POTEC}) not only reach inferior performance but also suffer from considerably higher variance. Their uncertainty bands are significantly wider, indicating unstable optimization. In contrast, PWLL-based methods exhibit near-invisible variance bands, with standard deviations roughly an order of magnitude smaller on average.

\begin{figure}[H]
  \centering
  \includegraphics[width=0.9\linewidth]{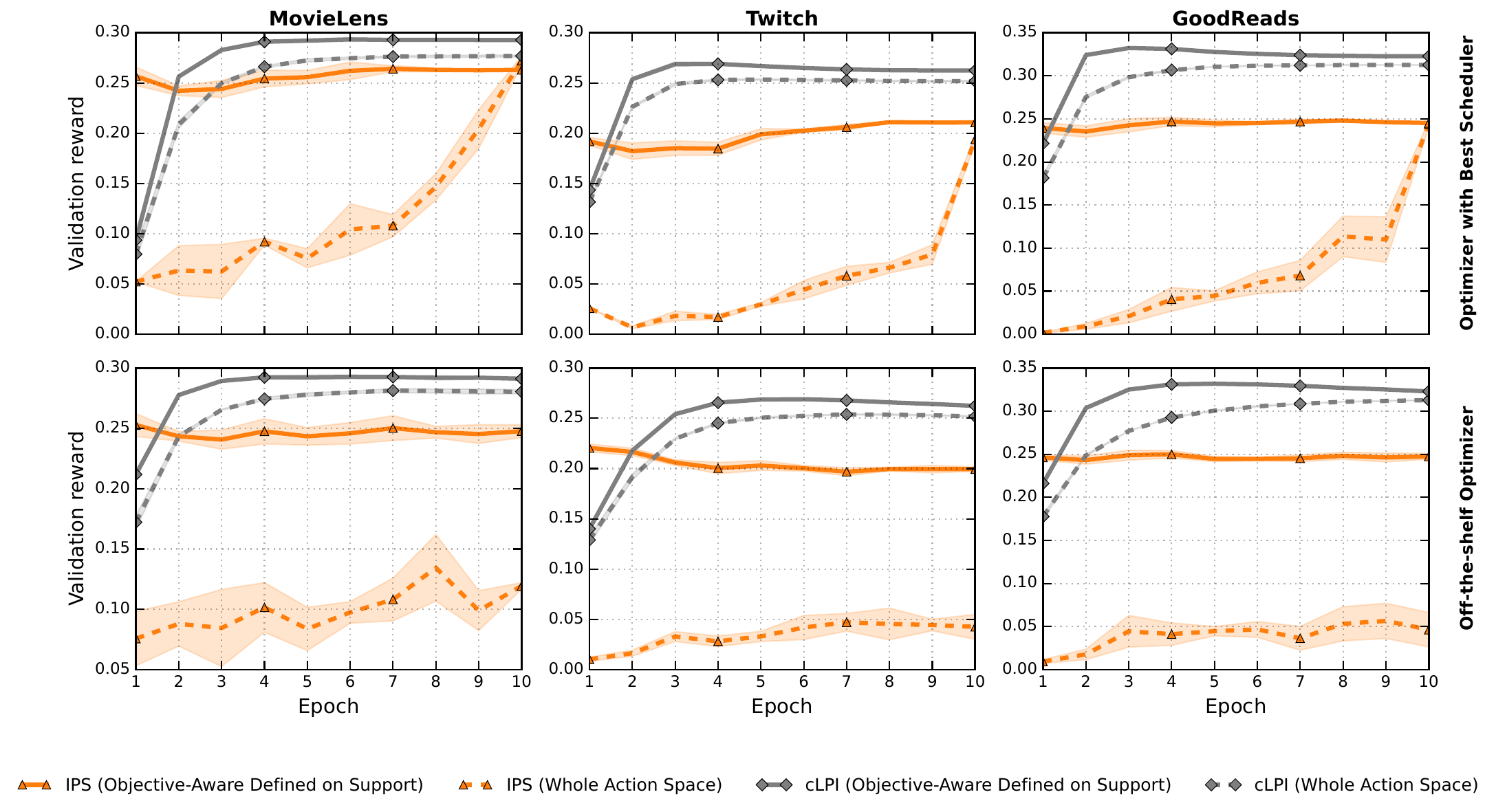}
  \caption{Objective Aware Parametrisation: Evolution of rewards averaged over 5 different seeds. Objective Aware Parametrization stabilizes and improves performance for PWLL and IPS methods.}
  \label{fig:las-ablation_seeds_OW}
\end{figure}

Finally, in Figure~\ref{fig:las-ablation_seeds_OW}, we observe that adopting an Objective-Aware parametrization yields further performance and stability improvements. For example, \texttt{cIPS} with Objective-Aware parametrization surpasses \texttt{cIPS} while maintaining lower variability, and \texttt{cLPI} in its Objective-Aware form consistently achieves the best overall performance. These results demonstrate that the combination of PWLL objectives and clever parametrization leads to more robust and more effective learned policies, while being very simple to implement.

\subsection{Ablation - Sensitivity to Reward Noise}



In the original evaluation setup (see \cref{app:las-additional_experiments}), the observed reward is deterministic; for user $i$, we have 
$R_i = \mathbbm{1}[A_i \in H_i]$, meaning that a positive reward is returned only when the selected action belongs to the user’s hidden set $H_i$.  
In this section, we investigate robustness to reward noise by introducing stochasticity in the form:
$$
R_i \sim \mathbbm{1}[A_i \in H_i]\,(1 - B(\epsilon)) + B(\epsilon)\,s,
$$
where $B(\epsilon)$ is a Bernoulli random variable with parameter $\epsilon$, and $s \in [0,1]$ is a shift.  
This results in noisy rewards supported on $[0,1]$. Note that any reward scaling can be normalized to this range via $R/R_{\max}$ when $R_{\max}>1$.

\begin{figure}[H]
  \centering
  \includegraphics[width=1.\linewidth]{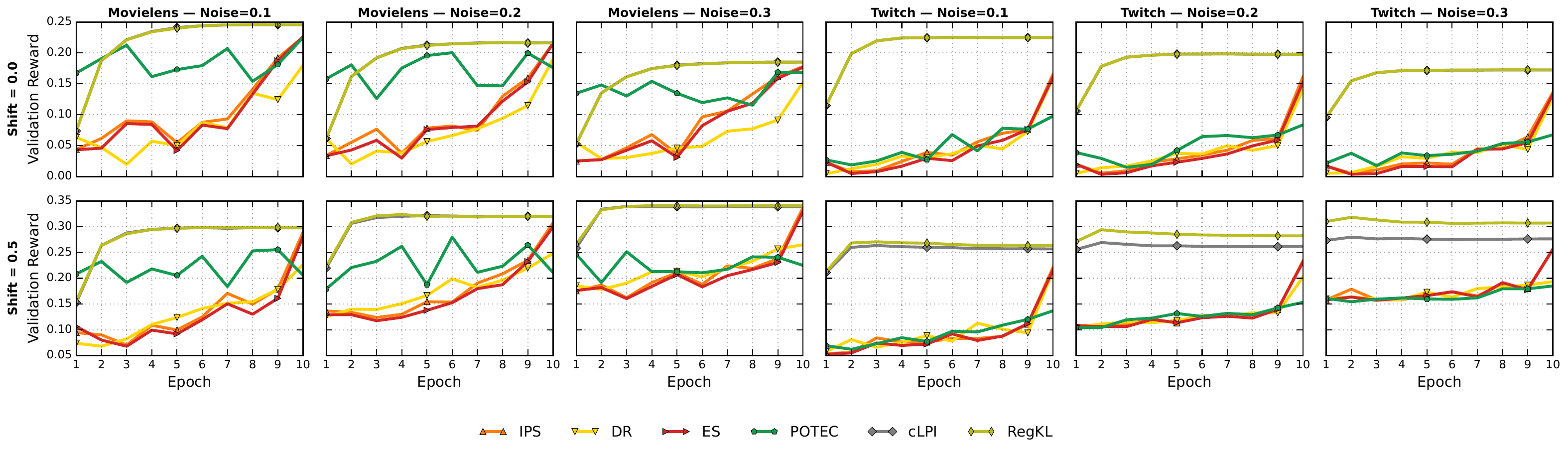}
  \caption{Ablation - Sensitivity to reward noise}
  \label{fig:las-ablation_noise}
\end{figure}

We evaluate six configurations defined by noise parameters $\epsilon \in \{0.1, 0.2, 0.3\}$ and reward shifts $s \in \{0, 0.5\}$.  
All methods are trained using the best-performing optimization schedule (one-cycle) to isolate the effect of noise.  
Results are reported in \cref{fig:las-ablation_noise}.

We observe that increasing the noise level when $s = 0$ consistently harms all methods, as expected from a more stochastic reward signal. In contrast, when $s = 0.5$, higher noise tends to increase the overall reward level, since the shift raises the baseline reward. Across all noise–shift conditions, PWLL-based objectives maintain a clear advantage over IPS-based methods. When $s = 0$, \texttt{RegKL} and \texttt{cLPI} perform similarly, confirming that both benefit from the logarithmic reparameterization. However, as both noise and shift increase, \texttt{RegKL} begins to outperform \texttt{cLPI}, suggesting that, with an appropriately chosen regularization weight $\beta$, \texttt{RegKL} remains highly competitive even under reward high stochasticity.

\textbf{Conclusion.} PWLL methods demonstrate robustness to reward noise, leading to improved stability and performance compared to traditional IPS-based objectives, even in challenging noise regimes.

\subsection{Ablation Study on Hyperparameters and Log Transform}

\begin{figure}[H]
    \centering

    \begin{subfigure}{0.49\linewidth}
        \centering
        \includegraphics[width=\linewidth]{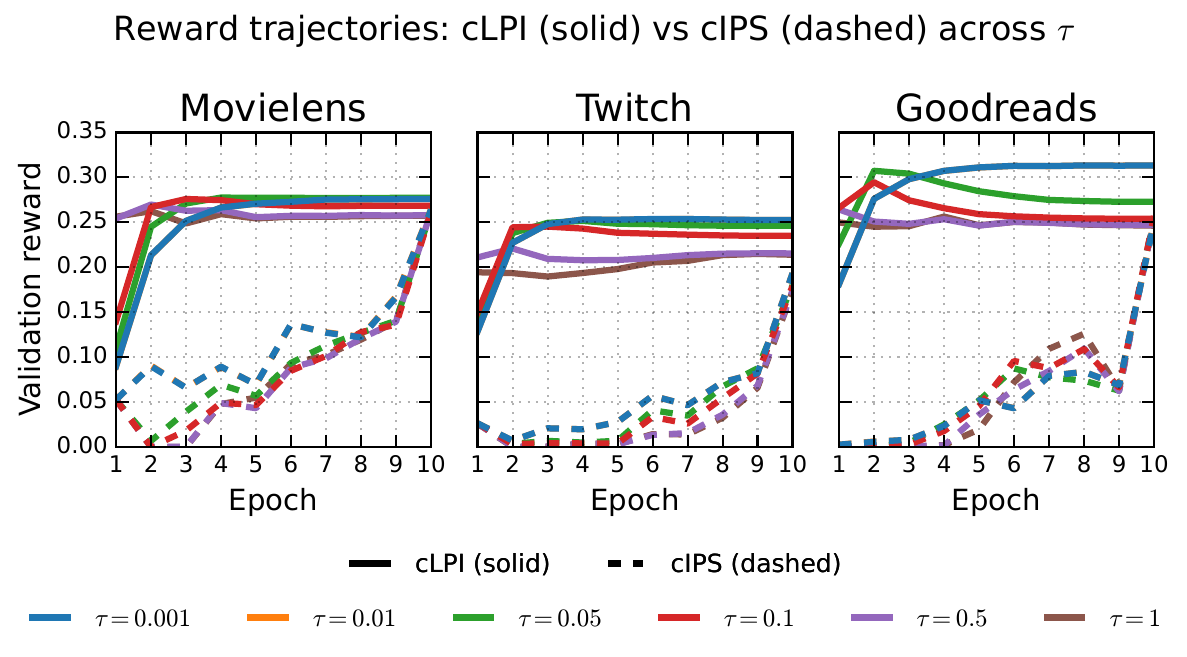}
        \caption{\texttt{cIPS} (linear) vs \texttt{cLPI} (PWLL) w.r.t $\tau$}
        \label{fig:las-left}
    \end{subfigure}
    \hfill
    \begin{subfigure}{0.49\linewidth}
        \centering
        \includegraphics[width=\linewidth]{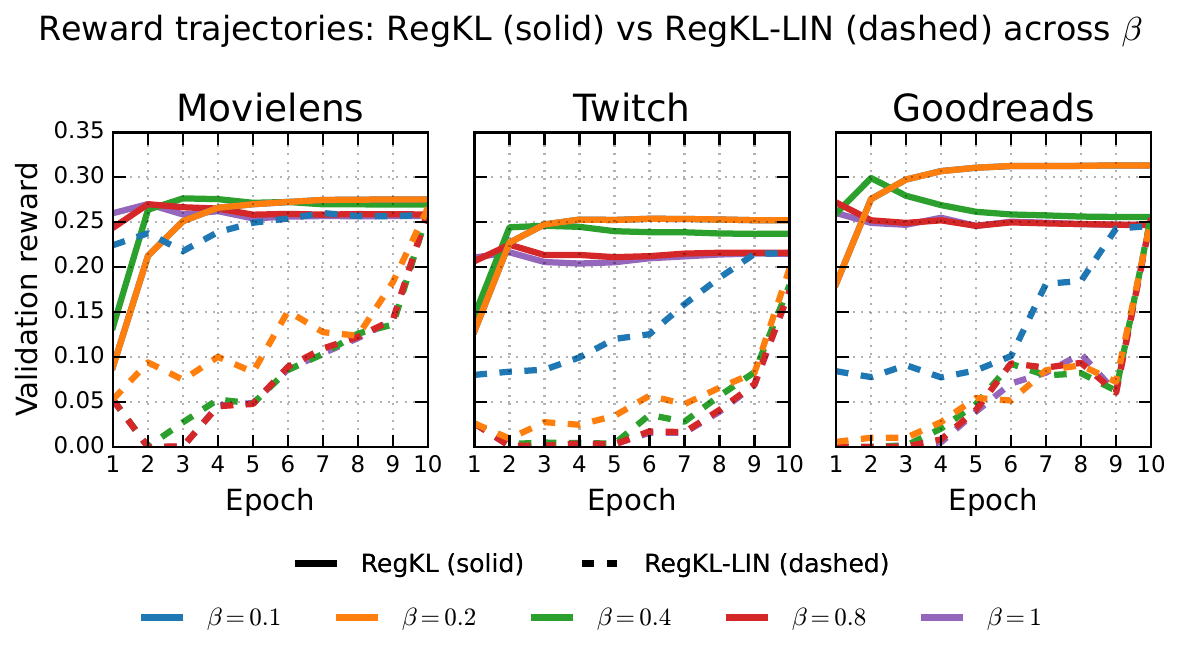}
        \caption{\texttt{RegKL-LIN} (linear) vs \texttt{RegKL} (PWLL) w.r.t $\beta$}
        \label{fig:las-right}
    \end{subfigure}

    \caption{Ablation Study on hyper-parameters and Log Transform}
    \label{fig:las-hp_log_ablation}
\end{figure}

In this section, we evaluate the impact of hyperparameter choices on \texttt{cIPS}, \texttt{cLPI}, \texttt{RegKL}, and \texttt{RegKL-LIN} (the non-logarithmic variant of \texttt{RegKL}). All methods are run using the best-performing optimization configuration (optimizer + learning rate scheduler), ensuring that differences are driven solely by hyperparameter values and by whether the policy transformation is linear or logarithmic. The results are shown in Figure~\ref{fig:las-hp_log_ablation}.

\begin{itemize}
    \item \textbf{cIPS} consistently fails to reach competitive performance across all values of $\tau$, especially in large action spaces. Its PWLL counterpart, \texttt{cLPI}, dominates for every $\tau$, converging faster and achieving superior results.
    
    \item For the KL-based objectives, we restrict to $\beta \ge 0.1$ in order to avoid numerical instability from the exponential term ($\exp(1/\beta) > 2\cdot10^5$ for $\beta < 0.1$). The same trend is observed: the PWLL variant (\texttt{RegKL}) reliably outperforms its linear analogue (\texttt{RegKL-LIN}) across all $\beta$, exhibiting more stable training dynamics, faster convergence and better performance.
\end{itemize}

\textbf{PWLL dominates.} Across both objective families, replacing linear weights with \textit{log-transformed} policy weights (PWLL) consistently provides \textbf{greater robustness to hyperparameters}, \textbf{faster optimization}, and \textbf{higher final performance}, even more in challenging large-action-space settings.




\subsection{Ablation: PWLL in Smaller Action Spaces}

We have shown that PWLL provides a more benign optimization landscape and yields stronger policies than IPS-based objectives in large action spaces. Here, we examine whether these benefits also extend to smaller action spaces. We construct a reduced version of \texttt{Movielens} by subsampling the action space to $K \in \{100, 500, 1000, 5000\}$ items.

\begin{figure}[H]
  \centering
  \includegraphics[width=1.\linewidth]{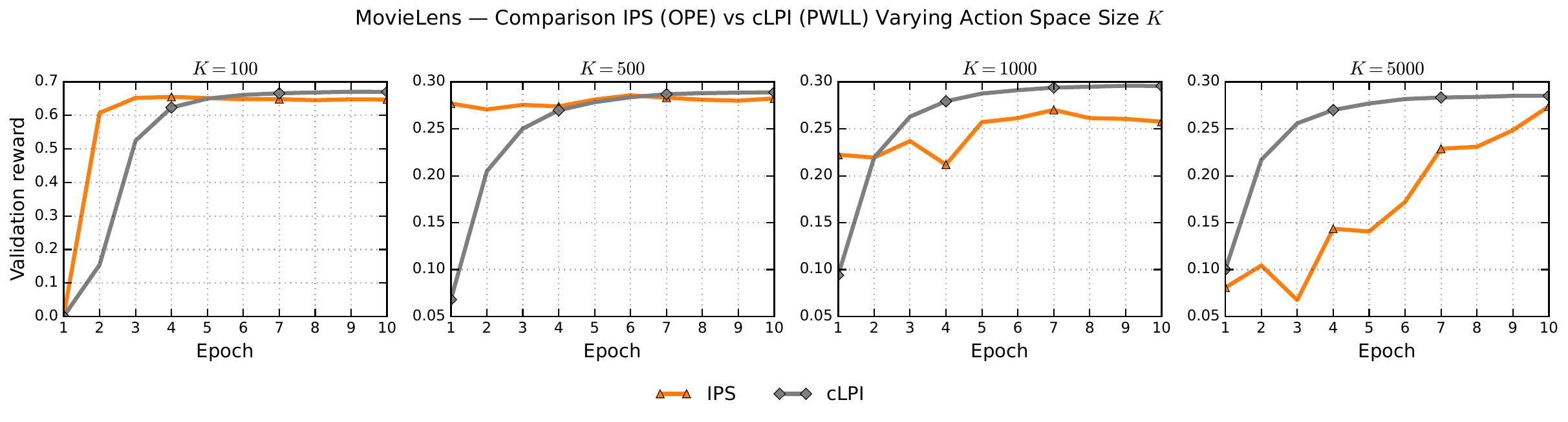}
  \caption{PWLL (\texttt{cLPI}) vs IPS-based (\texttt{IPS}) in smaller action spaces.}
  \label{fig:las-ablation_K}
\end{figure}

Figure~\ref{fig:las-ablation_K} reports performance across varying $K$ for \texttt{cIPS} (linear) and its PWLL-enhanced counterpart \texttt{cLPI} (log). In small action space settings ($K \le 500$), \texttt{cIPS} convergences faster than \texttt{cLPI}, but \texttt{cLPI} identifies a better maxima by the end of the 10 epochs. For medium action spaces ($K \ge 500$),  \texttt{cLPI} consistently outperforms \texttt{cIPS}, converging faster and identifying a better maximum. These results indicate that the optimization advantages of PWLL can still be beneficial in medium sized action space settings.

\subsection{Ablation - Sensitivity to the number of clusters}

\begin{figure}[H]
  \centering
  \includegraphics[width=0.75\linewidth]{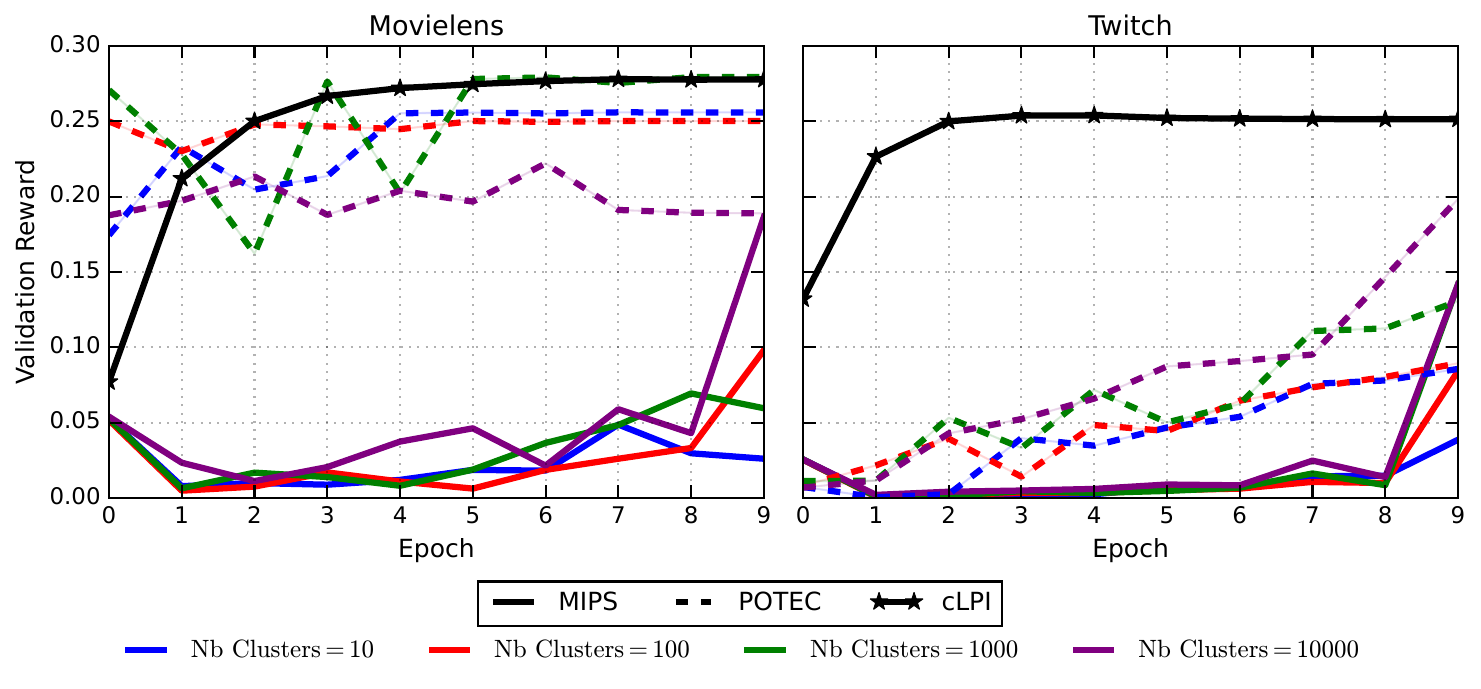}
  \caption{PWLL (\texttt{cLPI}) vs \texttt{POTEC} and \texttt{MIPS}, changing the number of  clusters.}
  \label{fig:las-ablation_CS}
\end{figure}

In this study, we compare our simple PWLL objective \texttt{cLPI} against \texttt{MIPS} and \texttt{POTEC}, two more complex IPS-based methods specifically designed for large action spaces. These baselines rely on a clustering function to reduce variance, and \texttt{POTEC} additionally leverages a reward model $\hat{r}$. We examine how the number of clusters affects their optimization performance. Figure~\ref{fig:las-ablation_CS} reports the results.

\texttt{POTEC} generally outperforms \texttt{MIPS} for all numbers of clusters. However, both methods exhibit optimization instability across settings. While \texttt{POTEC} can occasionally match the final performance of \texttt{cLPI} on Movielens for a carefully selected number of clusters ($1000$), it consistently falls short on Twitch regardless of the cluster configuration.

\textbf{Conclusion.} These findings demonstrate that focusing on optimization properties pays off: despite its simplicity, the PWLL objective \texttt{cLPI} can consistently outperform intricate IPS-based approaches tailored to large action spaces, even with the best finetuning.

\subsection{Ablation - Different Logging Supports}

\begin{figure}[H]
  \centering
  \includegraphics[width=0.6\linewidth]{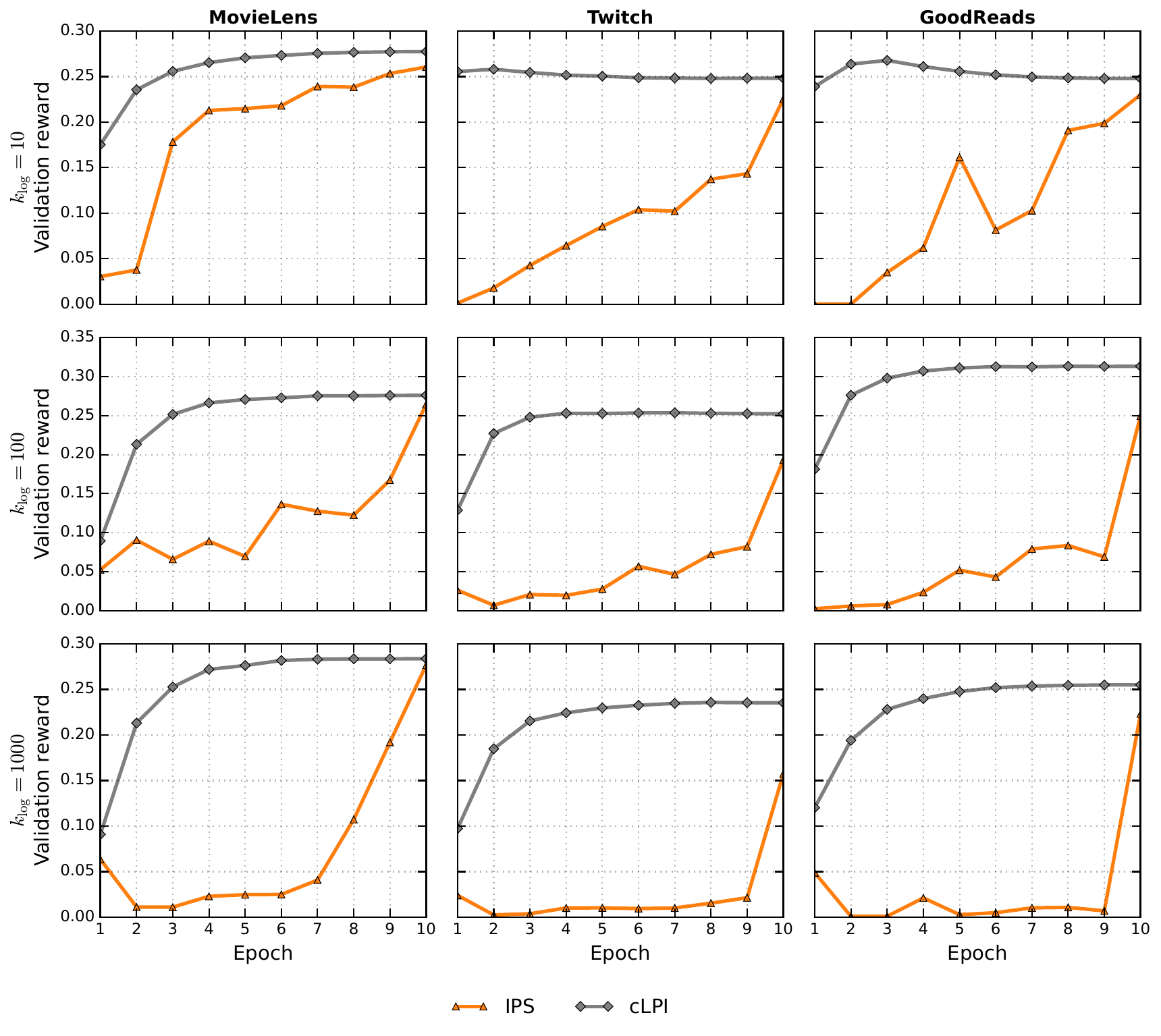}
  \caption{PWLL vs IPS: Different Logging Support sizes $k_{\log}$}
  \label{fig:las-ablation_k_log}
\end{figure}

We conduct experiments to quantify how increasing or restricting the support of the logging policy affects policy learning, comparing PWLL and IPS-based methods. Figure~\ref{fig:las-ablation_k_log} compiles the results and show that PWLL is still better than IPS-based approaches for different logging support sizes.

\subsection{Pessimism Does not Solve Optimization Problems}

\begin{figure}[H]
  \centering
  \includegraphics[width=0.8\linewidth]{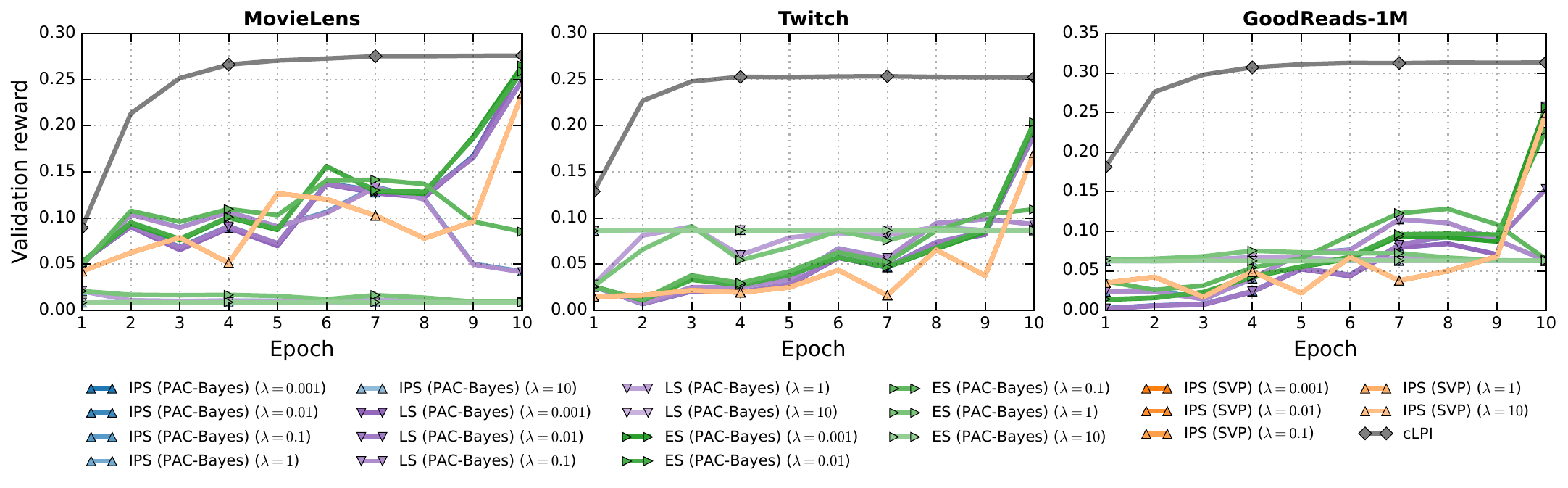}
  \caption{\texttt{cLPI} outperforms the pessimistic approaches. Some methods do not appear in the plot because their curves overlap.}
  \label{fig:las-pessimism_ablation}
\end{figure}

Pessimism in face of uncertainty \citet{jin2021pessimism} is motivated through a pure statistical learning rationale and is used to provide better statistical guarantees and more controlled excess risk. In the context of OPL, pessimistic strategies are derived combining concentration bounds with class complexity measures, be it VC dimension \citep{swaminathan2015batch} or PAC-Bayesian tools \citep{london2019bayesian, aouali23a, Sakhi2024LS}. For example, in its PAC-Bayesian formulation, the pessimistic objectives are all written in the following form:
$$\arg\max_{\pi_\theta} \quad \hat{V}(\pi_\theta) - \frac{\lambda}{n} \lvert\lvert \theta - \theta_0 \rvert\rvert^2_2\,,$$
Adding an $\ell_2$ regularization term that pulls the parameters $\theta$ towards the behavior policy parameters $\theta_0$ (defining $\pi_0$) induces pessimism by encouraging the learned policy to stay close to $\pi_0$ in parameter space. However, the optimization landscape of this objective becomes concave only when the regularization weight $\lambda$ is sufficiently large for the $\ell_2$ term to dominate. In that regime, the objective is indeed easier to optimize, but becomes overly conservative, yielding policies that remain too close to $\pi_0$ and under-exploit potential improvements. Figure~\ref{fig:las-pessimism_ablation} confirms this empirically: pessimistic approaches, whether based on Sample Variance Penalisation (SVP)~\citep{swaminathan2015batch}, PAC-Bayesian learning with clipped IPS~\citep{london2019bayesian}, Exponential Smoothing~\citep{aouali23a}, or Logarithmic Smoothing~\citep{Sakhi2024LS}, fail to outperform \texttt{cLPI} for any value of $\lambda$.

%% file: contents/exp_smoothing/appendix.tex
\section*{Notation Clarification: Value vs. Risk Formulation}

\textbf{Important:} Throughout the main chapter, we present our results using the \emph{value} formulation, where the goal is to maximize the expected reward $V(\pi) = \mathbb{E}_{X \sim \nu, A \sim \pi(\cdot | X)}[r(X, A)]$ with rewards $R \in [0, 1]$. In contrast, the appendix uses the equivalent \emph{risk} (or cost) formulation, where the goal is to minimize the expected cost $\mathcal{L}(\pi) = \mathbb{E}_{X \sim \nu, A \sim \pi(\cdot | X)}[c(X, A)]$ with costs $C \in [-1, 0]$.

These two formulations are related by a simple sign change:
\begin{align*}
    \mathcal{L}(\pi) = -V(\pi) \quad \text{and} \quad C = -R\,,
\end{align*}
where $r(x,a) = -c(x,a)$ for any $(x,a) \in \cX \times \cA$. Consequently:
\begin{itemize}
    \item Maximizing the value $V(\pi)$ is equivalent to minimizing the risk $\mathcal{L}(\pi)$.
    \item Upper bounds on $|V(\pi) - \hat{V}(\pi)|$ translate directly to upper bounds on $|\mathcal{L}(\pi) - \hat{\mathcal{L}}(\pi)|$.
    \item All theoretical guarantees derived in the appendix using the risk formulation apply equivalently to the value formulation presented in the corresponding chapter.
\end{itemize}

We adopt the risk formulation in the appendix as it aligns with the standard convention in statistical learning theory, where one typically minimizes a loss or risk function. The reader should keep this equivalence in mind when relating the appendix results to the main paper.

\section{Bias and Variance Trade-Off}\label{proofs:es-ope}
\textbf{Notation reminder:} This section uses the risk formulation with costs $C \in [-1, 0]$, which corresponds to the value formulation with rewards $R = -C \in [0, 1]$ in the main paper. The estimator $\hat{\mathcal{L}}_n^\alpha(\pi)$ here corresponds to $-\hat{V}^\alpha(\pi)$ in the main paper.

In this section, we provide additional results on how $\alpha$ controls the bias and variance of $\hat{\mathcal{L}}_n^\alpha(\cdot)$.

\subsection{Bias and Variance of $\texttt{IPS-}\alpha$}\label{app:alpha_bias_variance}

The following proposition states the bias-variance trade-off for $\hat{\mathcal{L}}_n^\alpha(\cdot)$. 

\begin{proposition}[Bias and variance of $\texttt{IPS-}\alpha$]\label{prop:alpha_bias_variance} Let $\alpha \in [0, 1]$, the following holds for any evaluation policy $\pi \in \Pi$ that is absolutely continuous with respect to $\pi_0$
\begin{align*}
    |\mathbb{B}(\hat{\mathcal{L}}_n^\alpha(\pi))|  & \leq \E{X \sim \nu, A \sim  \pi(\cdot | X)}{1 - \pi_0(A | X)^{1-\alpha}} \,,\nonumber\\
\mathbb{V}\left[\hat{\mathcal{L}}_n^\alpha(\pi)\right] & \leq \frac{1}{n} \mathbb{E}_{X \sim \nu, A \sim  \pi(\cdot | X)}\big[ \frac{\pi(A | X)}{\pi_0(A | X)^{2\alpha-1}} \big] \,.
\end{align*}
\end{proposition}

\begin{proof}
We first bound the bias as 
\begin{align*}
 \mathbb{B}(\hat{\mathcal{L}}_n^\alpha(\pi))  & = \E{}{\hat{\mathcal{L}}_n^\alpha(\pi)} - \mathcal{L}(\pi)\,,\\
 &= \frac{1}{n} \sum_{i=1}^n \E{X_i \sim \nu, A_i \sim \pi_0(\cdot | X_i), C_i \sim p(\cdot | X_i, A_i)}{C_i \frac{\pi(A_i | X_i)}{\pi_0(A_i | X_i)^\alpha}} - \mathcal{L}(\pi)\,,\\
    & \stackrel{(i)}{=} \E{(X, A, C) \sim \mu_{\pi_0}}{C \frac{\pi(A | X)}{\pi_0(A | X)^\alpha}} - \mathcal{L}(\pi)\,,\\
    & = \E{X \sim \nu}{\sum_{a \in \cA}c(X, a) \frac{\pi(a | X)}{\pi_0(a | X)^{\alpha-1}}} - \E{X \sim \nu}{\sum_{a \in \cA}c(X, a)\pi(a | X)}\,,\\
    & = \E{X \sim \nu}{\sum_{a \in \cA}c(X, a) \pi(a | X) ( \pi_0(a | X)^{1-\alpha}-1)}\,,\\
    & = \E{X \sim \nu, A \sim  \pi(\cdot | X)}{c(X, A) ( \pi_0(A | X)^{1-\alpha}-1)}\,,\\
\end{align*}
where $(i)$ follows from the i.i.d. assumption. Since $\pi_0(A | X)^{1-\alpha} \leq 1$ for any $x \in \cX$ and $a \in \cA$, we have that
\begin{align*}
 |\mathbb{B}(\hat{\mathcal{L}}_n^\alpha(\pi))|     & \leq \E{X \sim \nu, A \sim  \pi(\cdot | X)}{ |c(X, A)||\pi_0(A | X)^{1-\alpha}-1|}\,,\\
&\leq \E{X \sim \nu, A \sim  \pi(\cdot | X)}{ 1-\pi_0(A | X)^{1-\alpha}}\,.
\end{align*}

The variance is bounded as 
\begin{align*}
\mathbb{V}\left[\hat{\mathcal{L}}_n^\alpha(\pi)\right] &= \frac{1}{n^2} \sum_{i=1}^n \mathbb{V}_{X_i \sim \nu, A_i \sim \pi_0(\cdot | X_i), C_i \sim p(\cdot | X_i, A_i)}\Big[C_i \frac{\pi(A_i | X_i)}{\pi_0(A_i | X_i)^\alpha}\Big] \,,\\
   &= \frac{1}{n} \mathbb{V}_{(X, A, C) \sim \mu_{\pi_0}}\Big[C \frac{\pi(A | X)}{\pi_0(A | X)^\alpha}\Big] \,,\\
   &\leq\frac{1}{n} \mathbb{E}_{(X, A, C) \sim \mu_{\pi_0}}\Big[C^2 \frac{\pi(A | X)^2}{\pi_0(A | X)^{2\alpha}}\Big] \,,\\
   &\leq\frac{1}{n} \mathbb{E}_{X \sim \nu, A \sim  \pi_0(\cdot | X)}\Big[\frac{\pi(A | X)^2}{\pi_0(A | X)^{2\alpha}}\Big] \,,\\
    &=\frac{1}{n} \mathbb{E}_{X \sim \nu}\Big[ \sum_{a \in \cA} \frac{\pi(a | X)^{2}}{\pi_0(a | X)^{2\alpha-1}}\Big] \,,\\
    &=\frac{1}{n} \mathbb{E}_{X \sim \nu, A \sim  \pi(\cdot | X)}\Big[ \frac{\pi(A | X)}{\pi_0(A | X)^{2\alpha-1}}\Big] \,.
\end{align*}
\end{proof}

\section{Proofs for Off-Policy Learning}\label{proofs:es-opl}
In this section, we provide the complete proofs for our OPL results in \cref{sec:es-opl}. We start with proving \cref{thm:es-main_result} in \cref{proof:main_thm_proof}. We then state the extension of \cref{thm:es-main_result} along with its proof in \cref{app:thm_extension}. After that, in \cref{proof:direct_application}, we provide the proof for \cref{prop:direct_application}.  Finally, in \cref{proof:practice_theory}, we discuss in detail and prove our claims regarding the number of samples needed so that the performance of the learned policy is close to that of the optimal policy.

\textbf{Notation:} This section uses the risk formulation with $\mathcal{L}(\pi) = -V(\pi)$ and $\hat{\mathcal{L}}_n^\alpha(\pi) = -\hat{V}^\alpha(\pi)$. All bounds translate directly to the value formulation in the main paper. Recall that we assume the costs to be deterministic for simplicity: $C_i = c(X_i, A_i)$.

\subsection{Proof of \cref{thm:es-main_result}}\label{proof:main_thm_proof}
In this section, we prove \cref{thm:es-main_result}.
\begin{proof}
    
First, we decompose the difference $\mathcal{L}(\pi_{\mathbb{Q}})-\hat{\mathcal{L}}_n^\alpha(\pi_{\mathbb{Q}})$ as
\begin{align*}
    \mathcal{L}(\pi_{\mathbb{Q}})-\hat{\mathcal{L}}_n^\alpha(\pi_{\mathbb{Q}}) =  \underbrace{\mathcal{L}(\pi_{\mathbb{Q}}) - \frac{1}{n}\sum_{i=1}^n \mathcal{L}(\pi_{\mathbb{Q}} | X_i)}_{I_1} &+ \underbrace{\frac{1}{n} \sum_{i=1}^n \mathcal{L}(\pi_{\mathbb{Q}} | X_i) - \frac{1}{n}\sum_{i=1}^n\mathcal{L}^\alpha(\pi_{\mathbb{Q}} | X_i)}_{I_2} \\
    &+ \underbrace{\frac{1}{n}\sum_{i=1}^n\mathcal{L}^\alpha(\pi_{\mathbb{Q}} | X_i) - \hat{\mathcal{L}}_n^\alpha(\pi_{\mathbb{Q}})}_{I_3}\,,
\end{align*}
where 
\begin{align*}
    & \mathcal{L}(\pi_{\mathbb{Q}}) = \E{X \sim \nu\,, A \sim  \pi_{\mathbb{Q}}(\cdot | X)}{c(X, A)}\,,\\
    & \mathcal{L}(\pi_{\mathbb{Q}} | X_i)    = \E{A \sim  \pi_{\mathbb{Q}}(\cdot | X_i)}{c(X_i, A)}\,,\\ 
   &\mathcal{L}^\alpha(\pi_{\mathbb{Q}} | X_i) = \E{A \sim  \pi_0(\cdot | X_i)}{\frac{\pi_{\mathbb{Q}}(A | X_i)}{\pi_0(A | X_i)^\alpha}c(X_i, A)}\,, \\
   & \hat{\mathcal{L}}_n^\alpha(\pi) = \frac{1}{n} \sum_{i=1}^n \frac{\pi(A_i | X_i)}{\pi_0(A_i | X_i)^\alpha}C_i \,.
\end{align*}
Our goal is to bound $|\mathcal{L}(\pi_{\mathbb{Q}})-\hat{\mathcal{L}}_n^\alpha(\pi_{\mathbb{Q}}) |$ and thus we need to bound $|I_1| + |I_2| + |I_3| $. We start with $|I_1|$, \citet[Theorem 3.3]{alquier2021user} yields that following inequality holds with probability at least $1 -\delta/2$ for any distribution $\mathbb{Q}$ on $\mathcal{H}$ 
\begin{align}\label{eq:es-app_i1}
    |I_1| \leq \sqrt{ \frac{D_{\mathrm{KL}}(\mathbb{Q} \| \mathbb{P})+\log \frac{4\sqrt{n}}{\delta}}{2n}} \,.
\end{align}
Moreover, $|I_2|$ can be bounded by decomposing it as 
\begin{align*}
    |I_2|&=\left|\frac{1}{n} \sum_{i=1}^{n}  \E{A \sim  \pi_{\mathbb{Q}}(\cdot | X_i)}{c(X_i, A)}-\frac{1}{n} \sum_{i=1}^{n}\mathbb{E}_{A \sim  \pi_0\left(\cdot | X_i\right)}\left[ \frac{\pi_{\mathbb{Q}}(A | X_i)}{\pi_0^\alpha(A | X_i)}c(X_i, A) \right]\right| \\
    &=\left|\frac{1}{n} \sum_{i=1}^{n} \sum_{a \in \cA} \pi_{\mathbb{Q}}(a | X_i) c(X_i, a) -\pi_0(a | X_i)\frac{\pi_{\mathbb{Q}}(a | X_i)}{\pi_0^\alpha(a | X_i)}c(X_i, a)\right| \\
    &=\left|\frac{1}{n} \sum_{i=1}^{n} \sum_{a \in \cA} \Big(\pi_{\mathbb{Q}}(a | X_i) - \frac{\pi_{\mathbb{Q}}(a | X_i)}{\pi_0^{\alpha-1}(a | X_i)}\Big)c(X_i, a)\right| \\
    &=\left|\frac{1}{n} \sum_{i=1}^{n} \sum_{a \in \cA} \Big(1 - \pi_0^{1-\alpha}(a | X_i)\Big)\pi_{\mathbb{Q}}(a | X_i)c(X_i, a)\right|\,,\\
    &\leq\frac{1}{n} \sum_{i=1}^{n} \sum_{a \in \cA} \left|1 - \pi_0^{1-\alpha}(a | X_i)\right|\pi_{\mathbb{Q}}(a | X_i)\left|c(X_i, a)\right|\,.
\end{align*}
But $1 - \pi_0^{1-\alpha}(a | x) \geq 0$ and $|c(x, a)| \leq 1$ for any $a \in \cA$ and $x \in \cX$. Thus 
\begin{align}\label{eq:es-app_i2}
   |I_2| \leq \frac{1}{n} \sum_{i=1}^{n} \E{A \sim  \pi_{\mathbb{Q}}(\cdot | X_i)}{1 - \pi_0^{1-\alpha}(A | X_i)}\,.
\end{align}
Finally, we need to bound the main term $|I_3|$. To achieve this, we borrow the following technical lemma from \citet{haddouche2022pac}. It is slightly different from the one in \citet{haddouche2022pac}; their result holds for any $n \geq 1$ while we state a simpler version where $n$ is fixed in advance.

\begin{lemma}\label{lemma:es-app_maxime}
Let $\mathcal{Z}$ be an instance space and let $S_n=\left(z_i\right)_{i \in [n]}$ be an $n$-sized dataset for some $n \geq 1$. Let $\left(\mathcal{F}_i\right)_{i \in\{0\} \cup [n] }$ be a filtration adapted to $S_n$. Also, let $\mathcal{H}$ be a hypothesis space and $(f_i\left(S_i, h\right))_{i \in [n]}$ be a martingale difference sequence for any $h \in \mathcal{H}$, that is for any $i \in [n]$, and $h \in \mathcal{H}\,,$ we have that $ \mathbb{E}\left[f_i\left(S_i, h\right) | \mathcal{F}_{i-1}\right]=0$. Moreover, for any $h \in \mathcal{H}$, let $M_n(h)=\sum_{i=1}^n f_i\left(S_i, h\right)$. Then for any fixed prior, $\mathbb{P}$, on $\mathcal{H}$, any $\lambda>0$, the following holds with probability $1-\delta$ over the sample $S_n$, simultaneously for any $\mathbb{Q}$, on $\mathcal{H}$
\begin{align*}
    \left|\E{h \sim \mathbb{Q}}{M_n(h)}\right| \leq \frac{D_{\mathrm{KL}}(\mathbb{Q} \| \mathbb{P})+\log (2 / \delta)}{\lambda}+\frac{\lambda}{2}\left(\E{h \sim \mathbb{Q}}{\langle M\rangle_n(h) + [M]_n(h)}\right)\,,
\end{align*}
where $ \langle M\rangle_n(h)=\sum_{i=1}^n \mathbb{E}\left[f_i\left(S_i, h\right)^2 | \mathcal{F}_{i-1}\right]$ and $[M]_n(h)=\sum_{i=1}^n f_i\left(S_i, h\right)^2$.
\end{lemma}

To apply \cref{lemma:es-app_maxime}, we need to construct an adequate martingale difference sequence $(f_i(S_i, h))_{i \in [n]}$ for $h \in \mathcal{H}$ that allows us to retrieve $|I_3|$. To achieve this, we define $S_n = (A_i)_{i \in [n]}$ as the set of $n$ taken actions. Also, we let $(\mathcal{F}_i)_{i \in \{0\}\cup[n]}$ be a filtration adapted to $S_n$. For $h \in \mathcal{H}$, we define $f_i\left(S_i, h\right)$ as 
\begin{align*}
    f_i\left(S_i, h\right) = f_i\left(A_i, h\right) = \E{A \sim  \pi_0(\cdot | X_i)}{\frac{\mathbb{I}_{\{h(X_i)=A\}}}{\pi_0(A | X_i)^\alpha}c(X_i, A)} - \frac{\mathbb{I}_{\{h(X_i)=A_i\}}}{\pi_0(A_i | X_i)^\alpha}c(X_i, A_i)\,.
\end{align*}
We stress that $f_i(S_i, h)$ only depends on the last action in $S_i$, $A_i$, and the predictor $h$. For this reason, we denote it by $f_i(A_i, h)$. The function $f_i$ is indexed by $i$ since it depends on the fixed $i$-th context, $X_i$. The context $X_i$ is fixed and thus randomness only comes from $A_i \sim \pi_0(\cdot | X_i)$. It follows that the expectations are under $A_i \sim \pi_0(\cdot | X_i)$. First, we have that $\mathbb{E}\left[f_i\left(A_i, h\right) | \mathcal{F}_{i-1}\right]= 0$ for any $i \in [n]\,, h \in \mathcal{H}$. This follows from 
\begin{align*}
    &\mathbb{E}\left[f_i\left(A_i, h\right) | \mathcal{F}_{i-1}\right]  = \mathbb{E}_{A_i \sim \pi_0(\cdot | X_i)}\left[f_i\left(A_i, h\right) \Big| A_1, \ldots, A_{i-1}\right] \,,\\
     &=  \mathbb{E}_{A_i \sim \pi_0(\cdot | X_i)}\left[\E{A \sim  \pi_0(\cdot | X_i)}{\frac{\mathbb{I}_{\{h(X_i)=A\}}}{\pi_0(A | X_i)^\alpha}c(X_i, A)} - \frac{\mathbb{I}_{\{h(X_i)=A_i\}}}{\pi_0(A_i | X_i)^\alpha}c(X_i, A_i) \Big| A_1, \ldots, A_{i-1}\right]\,,\\
     &\stackrel{(i)}{=}  \E{A \sim  \pi_0(\cdot | X_i)}{\frac{\mathbb{I}_{\{h(X_i)=A\}}}{\pi_0(A | X_i)^\alpha}c(X_i, A)} - \mathbb{E}_{A_i \sim \pi_0(\cdot | X_i)}\left[\frac{\mathbb{I}_{\{h(X_i)=A_i\}}}{\pi_0(A_i | X_i)^\alpha}c(X_i, A_i) \Big| A_1, \ldots, A_{i-1}\right]\,.
\end{align*}
In $(i)$ we use the fact that given $X_i$, $ \E{A \sim  \pi_0(\cdot | X_i)}{\frac{\mathbb{I}_{\{h(X_i)=A\}}}{\pi_0(A | X_i)^\alpha}c(X_i, A)}$ is deterministic. Now $A_i$ does not depend on $A_1, \ldots, A_{i-1}$ since logged data is i.i.d. Hence 
\begin{align*}
    \mathbb{E}_{A_i \sim \pi_0(\cdot | X_i)}\left[\frac{\mathbb{I}_{\{h(X_i)=A_i\}}}{\pi_0(A_i | X_i)^\alpha}c(X_i, A_i) \Big| A_1, \ldots, A_{i-1}\right] &= \mathbb{E}_{A_i \sim \pi_0(\cdot | X_i)}\left[\frac{\mathbb{I}_{\{h(X_i)=A_i\}}}{\pi_0(A_i | X_i)^\alpha}c(X_i, A_i)\right]\,,\\ &= \mathbb{E}_{A \sim  \pi_0(\cdot | X_i)}\left[\frac{\mathbb{I}_{\{h(X_i)=A\}}}{\pi_0(A | X_i)^\alpha}c(X_i, A)\right]\,.
\end{align*}
It follows that 
\begin{align*}
    & \mathbb{E}\left[f_i\left(A_i, h\right) | \mathcal{F}_{i-1}\right] \\
     &=  \E{A \sim  \pi_0(\cdot | X_i)}{\frac{\mathbb{I}_{\{h(X_i)=A\}}}{\pi_0(A | X_i)^\alpha}c(X_i, A)} - \mathbb{E}_{A_i \sim \pi_0(\cdot | X_i)}\left[\frac{\mathbb{I}_{\{h(X_i)=A_i\}}}{\pi_0(A_i | X_i)^\alpha}c(X_i, A_i) \Big| A_1, \ldots, A_{i-1}\right]\,,\\
    &= \E{A \sim  \pi_0(\cdot | X_i)}{\frac{\mathbb{I}_{\{h(X_i)=A\}}}{\pi_0(A | X_i)^\alpha}c(X_i, A)} - \mathbb{E}_{A \sim  \pi_0(\cdot | X_i)}\left[\frac{\mathbb{I}_{\{h(X_i)=A\}}}{\pi_0(A | X_i)^\alpha}c(X_i, A)\right]\,,\\
    &= 0\,.
\end{align*}
Therefore, for any $h \in \mathcal{H}$, $(f_i(A_i, h))_{i \in [n]}$ is a martingale difference sequence. Hence we apply \cref{lemma:es-app_maxime} and obtain that the following inequality holds with probability at least $1-\delta/2$ for any $\mathbb{Q}$ on $\mathcal{H}$
\begin{align}\label{eq:es-app_proof_0}
    \left|\E{h \sim \mathbb{Q}}{M_n(h)}\right| \leq \frac{D_{\mathrm{KL}}(\mathbb{Q} \| \mathbb{P})+\log (4 / \delta)}{\lambda}+\frac{\lambda}{2}\left(\E{h \sim \mathbb{Q}}{\langle M\rangle_n(h) + [M]_n(h)}\right)\,,
\end{align}
where 
\begin{align*}
    M_n(h)&=\sum_{i=1}^n f_i\left(A_i, h\right)\,,\\
     \langle M\rangle_n(h)&=\sum_{i=1}^n \mathbb{E}\left[f_i\left(A_i, h\right)^2 | \mathcal{F}_{i-1}\right]\,,\\
     [M]_n(h)&=\sum_{i=1}^n f_i\left(A_i, h\right)^2
\end{align*}
Now these terms can be decomposed as 
\begin{align*}
    \E{h \sim \mathbb{Q}}{M_n(h)} &= \sum_{i=1}^n\E{h \sim \mathbb{Q}}{f_i\left(A_i, h\right)}\,,\\
    &=  \sum_{i=1}^n \E{h \sim \mathbb{Q}}{\E{A \sim  \pi_0(\cdot | X_i)}{\frac{\mathbb{I}_{\{h(X_i)=A\}}}{\pi_0(A | X_i)^\alpha}c(X_i, A)} - \frac{\mathbb{I}_{\{h(X_i)=A_i\}}}{\pi_0(A_i | X_i)^\alpha}c(X_i, A_i)}\,,\\
    &\stackrel{(i)}{=} \sum_{i=1}^n \E{h \sim \mathbb{Q}}{\E{A \sim  \pi_0(\cdot | X_i)}{\frac{\mathbb{I}_{\{h(X_i)=A\}}}{\pi_0(A | X_i)^\alpha}c(X_i, A)}} - \E{h \sim \mathbb{Q}}{\frac{\mathbb{I}_{\{h(X_i)=A_i\}}}{\pi_0(A_i | X_i)^\alpha}c(X_i, A_i)}\,,\\
    &\stackrel{(ii)}{=} \sum_{i=1}^n\E{A \sim  \pi_0(\cdot | X_i)}{\frac{ \E{h \sim \mathbb{Q}}{\mathbb{I}_{\{h(X_i)=A\}}}}{\pi_0(A | X_i)^\alpha}c(X_i, A)} - \frac{\E{h \sim \mathbb{Q}}{\mathbb{I}_{\{h(X_i)=A_i\}}}}{\pi_0(A_i | X_i)^\alpha}c(X_i, A_i)\,,\\
    &\stackrel{(iii)}{=} \sum_{i=1}^n\E{A \sim  \pi_0(\cdot | X_i)}{\frac{ \pi_{\mathbb{Q}}(A | X_i) }{\pi_0(A | X_i)^\alpha}c(X_i, A)} - \sum_{i=1}^n \frac{ \pi_{\mathbb{Q}}(A_i | X_i)}{\pi_0(A_i | X_i)^\alpha}c(X_i, A_i)\,,
\end{align*}
where we use the linearity of the expectation in both $(i) $  and $(ii)$. In $(iii)$, we use our definition of policies in \eqref{eq:es-pac_policies}. Therefore, we have that 
\begin{align}\label{eq:es-app_proof_1}
    \E{h \sim \mathbb{Q}}{M_n(h)}  &=\sum_{i=1}^n\E{A \sim  \pi_0(\cdot | X_i)}{\frac{ \pi_{\mathbb{Q}}(A | X_i) }{\pi_0(A | X_i)^\alpha}c(X_i, A)} - \sum_{i=1}^n \frac{ \pi_{\mathbb{Q}}(A_i | X_i)}{\pi_0(A_i | X_i)^\alpha}c(X_i, A_i)\,,\nonumber\\
    & \stackrel{(i)}{=} \sum_{i=1}^n\mathcal{L}^\alpha(\pi_{\mathbb{Q}} | X_i) - n\hat{\mathcal{L}}_n^\alpha(\pi_{\mathbb{Q}})\,,\nonumber\\
    &= nI_3\,,
\end{align}
where we used the fact that $C_i = c(X_i, A_i)$ for any $i \in [n]$ in $(i)$. 

Now we focus on the terms $\langle M\rangle_n(h)$ and $ [M]_n(h)$. First, we have that 
\begin{align}\label{eq:es-app_proof_2}
    f_i\left(A_i, h\right)^2 &= \Big(\E{A \sim  \pi_0(\cdot | X_i)}{\frac{\mathbb{I}_{\{h(X_i)=A\}}}{\pi_0(A | X_i)^\alpha}c(X_i, A)} - \frac{\mathbb{I}_{\{h(X_i)=A_i\}}}{\pi_0(A_i | X_i)^\alpha}c(X_i, A_i)\Big)^2\,,\\
    &= \E{A \sim  \pi_0(\cdot | X_i)}{\frac{\mathbb{I}_{\{h(X_i)=A\}}}{\pi_0(A | X_i)^\alpha}c(X_i, A)}^2 + \Big(\frac{\mathbb{I}_{\{h(X_i)=A_i\}}}{\pi_0(A_i | X_i)^\alpha}c(X_i, A_i)\Big)^2 \nonumber \\ & \hspace{2cm} - 2   \E{A \sim  \pi_0(\cdot | X_i)}{\frac{\mathbb{I}_{\{h(X_i)=A\}}}{\pi_0(A | X_i)^\alpha}c(X_i, A)} \frac{\mathbb{I}_{\{h(X_i)=A_i\}}}{\pi_0(A_i | X_i)^\alpha}c(X_i, A_i)\,,\nonumber\\
    &= \E{A \sim  \pi_0(\cdot | X_i)}{\frac{\mathbb{I}_{\{h(X_i)=A\}}}{\pi_0(A | X_i)^\alpha}c(X_i, A)}^2 + \frac{\mathbb{I}_{\{h(X_i)=A_i\}}}{\pi_0(A_i | X_i)^{2\alpha}}c(X_i, A_i)^2 \nonumber \\ & \hspace{2cm} - 2   \E{A \sim  \pi_0(\cdot | X_i)}{\frac{\mathbb{I}_{\{h(X_i)=A\}}}{\pi_0(A | X_i)^\alpha}c(X_i, A)} \frac{\mathbb{I}_{\{h(X_i)=A_i\}}}{\pi_0(A_i | X_i)^\alpha}c(X_i, A_i)\,.\nonumber
\end{align}
Moreover, $f_i\left(A_i, h\right)^2$ does not depend on $A_1, \ldots, A_{i-1}$. Thus, 
\begin{align*}
    \mathbb{E}\left[f_i\left(A_i, h\right)^2 | \mathcal{F}_{i-1}\right] &= \mathbb{E}_{A_i \sim \pi_0(\cdot | X_i)}\left[f_i\left(A_i, h\right)^2 | \mathcal{F}_{i-1}\right]\,,\\
    & = \mathbb{E}_{A_i \sim \pi_0(\cdot | X_i)}\left[f_i\left(A_i, h\right)^2\right]= \mathbb{E}_{A \sim  \pi_0(\cdot | X_i)}\left[f_i\left(A, h\right)^2\right]\,.
\end{align*}
Computing $\mathbb{E}_{A \sim  \pi_0(\cdot | X_i)}\left[f_i\left(A, h\right)^2\right]$ using the decomposition in \eqref{eq:es-app_proof_2} yields
\begin{align}\label{eq:es-app_proof_3}
    &\mathbb{E}\left[f_i\left(A_i, h\right)^2 | \mathcal{F}_{i-1}\right] = \mathbb{E}_{A \sim  \pi_0(\cdot | X_i)}\left[f_i\left(A, h\right)^2\right] \,,\nonumber\\
    &= -  \E{A \sim  \pi_0(\cdot | X_i)}{\frac{\mathbb{I}_{\{h(X_i)=A\}}}{\pi_0(A | X_i)^\alpha}c(X_i, A)}^2
+ \E{A \sim  \pi_0(\cdot | X_i)}{\frac{\mathbb{I}_{\{h(X_i)=A\}}}{\pi_0(A | X_i)^{2\alpha}}c(X_i, A)^2}
\end{align}
Combining \eqref{eq:es-app_proof_2} and \eqref{eq:es-app_proof_3} leads to
\begin{align}\label{eq:es-app_proof_4}
    \mathbb{E}\left[f_i\left(A_i, h\right)^2 | \mathcal{F}_{i-1}\right] + f_i\left(A_i, h\right)^2 &= \E{A \sim  \pi_0(\cdot | X_i)}{\frac{\mathbb{I}_{\{h(X_i)=A\}}}{\pi_0(A | X_i)^{2\alpha}}c(X_i, A)^2}+ \frac{\mathbb{I}_{\{h(X_i)=A_i\}}}{\pi_0(A_i | X_i)^{2\alpha}}c(X_i, A_i)^2 \nonumber\\ &- 2   \E{A \sim  \pi_0(\cdot | X_i)}{\frac{\mathbb{I}_{\{h(X_i)=A\}}}{\pi_0(A | X_i)^\alpha}c(X_i, A)} \frac{\mathbb{I}_{\{h(X_i)=A_i\}}}{\pi_0(A_i | X_i)^\alpha}c(X_i, A_i)\,,\nonumber\\
    & \stackrel{(i)}{\leq} \E{A \sim  \pi_0(\cdot | X_i)}{\frac{\mathbb{I}_{\{h(X_i)=A\}}}{\pi_0(A | X_i)^{2\alpha}}c(X_i, A)^2}+ \frac{\mathbb{I}_{\{h(X_i)=A_i\}}}{\pi_0(A_i | X_i)^{2\alpha}}c(X_i, A_i)^2\,.
\end{align}
The inequality in $(i)$ holds because $- 2   \E{A \sim  \pi_0(\cdot | X_i)}{\frac{\mathbb{I}_{\{h(X_i)=A\}}}{\pi_0(A | X_i)^\alpha}c(X_i, A)} \frac{\mathbb{I}_{\{h(X_i)=A_i\}}}{\pi_0(A_i | X_i)^\alpha}c(X_i, A_i) \leq 0$. Therefore, we have that
\begin{align*}
    \langle M\rangle_n(h) + [M]_n(h) \leq \sum_{i=1}^n  \E{A \sim  \pi_0(\cdot | X_i)}{\frac{\mathbb{I}_{\{h(X_i)=A\}}}{\pi_0(A | X_i)^{2\alpha}}c(X_i, A)^2}+ \frac{\mathbb{I}_{\{h(X_i)=A_i\}}}{\pi_0(A_i | X_i)^{2\alpha}}c(X_i, A_i)^2\,.
\end{align*}
Finally, by using the linearity of the expectation and the definition of policies in \eqref{eq:es-pac_policies}, we get that 
\begin{align}\label{eq:es-app_proof_5}
    &\E{h \sim \mathbb{Q}}{\langle M\rangle_n(h) + [M]_n(h)} \\
    & \leq \sum_{i=1}^n  \E{A \sim  \pi_0(\cdot | X_i)}{\frac{\E{h \sim \mathbb{Q}}{\mathbb{I}_{\{h(X_i)=A\}}}}{\pi_0(A | X_i)^{2\alpha}}c(X_i, A)^2}+ \frac{\E{h \sim \mathbb{Q}}{\mathbb{I}_{\{h(X_i)=A_i\}}}}{\pi_0(A_i | X_i)^{2\alpha}}c(X_i, A_i)^2\,,\nonumber\\
    & = \sum_{i=1}^n  \E{A \sim  \pi_0(\cdot | X_i)}{\frac{\pi_{\mathbb{Q}}(A | X_i)}{\pi_0(A | X_i)^{2\alpha}}c(X_i, A)^2}+ \frac{\pi_{\mathbb{Q}}(A_i | X_i)}{\pi_0(A_i | X_i)^{2\alpha}}c(X_i, A_i)^2\,.
\end{align}
Combining \eqref{eq:es-app_proof_0} and \eqref{eq:es-app_proof_5} yields
\begin{align}\label{eq:es-app_proof_6}
  & n |I_3| = | \sum_{i=1}^n\mathcal{L}^\alpha(\pi_{\mathbb{Q}} | X_i) - n\hat{\mathcal{L}}_n^\alpha(\pi_{\mathbb{Q}})  |\, \nonumber\\
   &\leq \frac{D_{\mathrm{KL}}(\mathbb{Q} \| \mathbb{P})+\log (4 / \delta)}{\lambda} + \frac{\lambda}{2}\sum_{i=1}^n  \E{A \sim  \pi_0(\cdot | X_i)}{\frac{\pi_{\mathbb{Q}}(A | X_i)}{\pi_0(A | X_i)^{2\alpha}}c(X_i, A)^2}+ \frac{\pi_{\mathbb{Q}}(A_i | X_i)}{\pi_0(A_i | X_i)^{2\alpha}}c(X_i, A_i)^2\,.
\end{align}
This means that the following inequality holds with probability at least $1-\delta/2$ for any distribution $\mathbb{Q}$ on $\mathcal{H}$
\begin{align}
   \left|I_3  \right| \leq \frac{D_{\mathrm{KL}}(\mathbb{Q} \| \mathbb{P})+\log (4 / \delta)}{n\lambda} &+ \frac{\lambda}{2n}\sum_{i=1}^n  \E{A \sim  \pi_0(\cdot | X_i)}{\frac{\pi_{\mathbb{Q}}(A | X_i)}{\pi_0(A | X_i)^{2\alpha}}c(X_i, A)^2} \nonumber \\&+ \frac{\lambda}{2n} \sum_{i=1}^n\frac{\pi_{\mathbb{Q}}(A_i | X_i)}{\pi_0(A_i | X_i)^{2\alpha}}c(X_i, A_i)^2\,.
\end{align}
However we know that $c(x, a)^2 \leq 1$ for any $x\in \cX$ and $a \in \cA$ and that $c(X_i, A_i) = C_i$ for any $i \in [n]$. Thus the following inequality holds with probability at least $1-\delta/2$ for any distribution $\mathbb{Q}$ on $\mathcal{H}$
\begin{align}\label{eq:es-app_i3}
   \left|I_3  \right| \leq \frac{D_{\mathrm{KL}}(\mathbb{Q} \| \mathbb{P})+\log (4 / \delta)}{n\lambda} + \frac{\lambda}{2n}\sum_{i=1}^n  \E{A \sim  \pi_0(\cdot | X_i)}{\frac{\pi_{\mathbb{Q}}(A | X_i)}{\pi_0(A | X_i)^{2\alpha}}}+ \frac{\lambda}{2n} \sum_{i=1}^n\frac{\pi_{\mathbb{Q}}(A_i | X_i)}{\pi_0(A_i | X_i)^{2\alpha}}C_i^2\,.
\end{align}
The union bound of \eqref{eq:es-app_i1} and \eqref{eq:es-app_i3} combined with the deterministic result in \eqref{eq:es-app_i2} yields that the following inequality holds with probability at least $1-\delta$ for any distribution $\mathbb{Q}$ on $\mathcal{H}$
\begin{align}\label{eq:es-app_main_inequality}
    |\mathcal{L}(\pi_{\mathbb{Q}}) -\hat{\mathcal{L}}_n^\alpha(\pi_{\mathbb{Q}})|  \leq  \sqrt{ \frac{D_{\mathrm{KL}}(\mathbb{Q} \| \mathbb{P})+\log \frac{4\sqrt{n}}{\delta}}{2n}} + \frac{1}{n} \sum_{i=1}^{n} \E{A \sim  \pi_{\mathbb{Q}}(\cdot | X_i)}{1 - \pi_0^{1-\alpha}(A | X_i)}\nonumber\\  + \frac{D_{\mathrm{KL}}(\mathbb{Q} \| \mathbb{P})+\log (4 / \delta)}{n\lambda}
    + \frac{\lambda}{2n}\sum_{i=1}^n  \E{A \sim  \pi_0(\cdot | X_i)}{\frac{\pi_{\mathbb{Q}}(A | X_i)}{\pi_0(A | X_i)^{2\alpha}}}+ \frac{\lambda}{2n} \sum_{i=1}^n\frac{\pi_{\mathbb{Q}}(A_i | X_i)}{\pi_0(A_i | X_i)^{2\alpha}}C_i^2\,.
\end{align}
\end{proof}

\subsection{Extensions of \cref{thm:es-main_result}}\label{app:thm_extension}

\begin{proposition}[Extension of \cref{thm:es-main_result} to hold simultaneously for any $\lambda \in (0, 1)$]\label{prop:thm_any_lambda} Let $n \ge 1$, $\delta \in [0, 1]$, $\alpha \in [0, 1]$, and let $\mathbb{P}$ be a fixed prior on $\mathcal{H}$, then with probability at least $1-\delta$ over draws $\cD_n \sim \mu_{\pi_0}^n$, the following holds simultaneously for any posterior $\mathbb{Q}$ on $\mathcal{H}$, and for any $\lambda \in (0, 1)$ that 
\begin{align*}
    |\mathcal{L}(\pi_{\mathbb{Q}}) -\hat{\mathcal{L}}_n^\alpha(\pi_{\mathbb{Q}})| \leq \sqrt{ \frac{{\textsc{kl}^\prime}_{1}(\pi_{\mathbb{Q}}, \lambda)}{2n} } + B_n^\alpha(\pi_{\mathbb{Q}})  +
\frac{{\textsc{kl}^\prime}_{2}(\pi_{\mathbb{Q}}, \lambda)}{n \lambda } + \frac{\lambda}{2}\operatorname{Var}_n^\alpha(\pi_{\mathbb{Q}})\,.
\end{align*}
where 
\begin{align*}
    & {\textsc{kl}^\prime}_{1}(\pi_{\mathbb{Q}}, \lambda)  =D_{\mathrm{KL}}(\mathbb{Q} \| \mathbb{P})+\log \frac{8\sqrt{n}}{\delta \lambda} \,,\\
    &{\textsc{kl}^\prime}_{2}(\pi_{\mathbb{Q}}, \lambda)  =  2\big( D_{\mathrm{KL}}(\mathbb{Q} \| \mathbb{P})+\log \frac{8 }{\delta \lambda}\big)\,,\\
    & B_n^\alpha(\pi_{\mathbb{Q}}) = 1 - \frac{1}{n}\sum_{i=1}^{n} \E{A \sim  \pi_{\mathbb{Q}}(\cdot | X_i)}{\pi_0^{1-\alpha}(A | X_i)}\,,\\
    &\operatorname{Var}_n^\alpha(\pi_{\mathbb{Q}}) = \frac{1}{n}\sum_{i=1}^n  \E{A \sim  \pi_0(\cdot | X_i)}{\frac{\pi_{\mathbb{Q}}(A | X_i)}{\pi_0(A | X_i)^{2\alpha}}} + \frac{\pi_{\mathbb{Q}}(A_i | X_i)}{\pi_0(A_i | X_i)^{2\alpha}}C_i^2.
\end{align*}
\end{proposition}

\begin{proof}
Let $\delta \in (0, 1)$. For any $i \geq 1$, we define $\lambda_i=2^{-i}$ and let $\delta_i = \delta \lambda_i$. Then \cref{thm:es-main_result} yields that for any $i \geq 1$, the following inequality holds with probability at least $1 - \delta_i$ for any $\mathbb{Q}$ on $\mathcal{H}$
\begin{align*}
    |\mathcal{L}(\pi_{\mathbb{Q}}) -\hat{\mathcal{L}}_n^\alpha(\pi_{\mathbb{Q}})| \leq \sqrt{ \frac{D_{\mathrm{KL}}(\mathbb{Q} \| \mathbb{P})+\log \frac{4\sqrt{n}}{\delta_i}}{2n} } + B_n^\alpha(\pi_{\mathbb{Q}})  +
\frac{D_{\mathrm{KL}}(\mathbb{Q} \| \mathbb{P})+\log \frac{4}{\delta_i}}{n \lambda_i } + \frac{\lambda_i}{2}\operatorname{Var}_n^\alpha(\pi_{\mathbb{Q}})\,.
\end{align*}
Now notice that $\sum_{i =1}^\infty \lambda_i = 1$, and hence $\sum_{i =1}^\infty \delta_i = \delta$. Therefore, the union bound of the above inequalities over $i \geq 1$ yields that with probability at least $1-\delta$, the following inequality holds with probability at least $1 - \delta$ for any $\mathbb{Q}$ on $\mathcal{H}$ and for any $i \geq 1$
\begin{align}\label{eq:es-for_any_i}
    |\mathcal{L}(\pi_{\mathbb{Q}}) -\hat{\mathcal{L}}_n^\alpha(\pi_{\mathbb{Q}})| \leq \sqrt{ \frac{D_{\mathrm{KL}}(\mathbb{Q} \| \mathbb{P})+\log \frac{4\sqrt{n}}{\delta_i}}{2n} } + B_n^\alpha(\pi_{\mathbb{Q}})  +
\frac{D_{\mathrm{KL}}(\mathbb{Q} \| \mathbb{P})+\log \frac{4}{\delta_i}}{n \lambda_i } + \frac{\lambda_i}{2}\operatorname{Var}_n^\alpha(\pi_{\mathbb{Q}})\,.
\end{align}
Let $\lceil \cdot \rceil$ denote the ceiling function, then we have that for any $\lambda \in (0, 1)$, there exists $j = \lceil \frac{-\log \lambda}{\log 2} \rceil \geq 1$ such that $\lambda/2 \leq \lambda_j \leq \lambda$. Since \eqref{eq:es-for_any_i} holds for any $i \geq 1$, it holds in particular for $j$. In addition to this, we have that $\frac{1}{\lambda_j} \leq \frac{2}{\lambda}$, that $\lambda_j \leq \lambda$ and that $\frac{1}{\delta_j} = \frac{1}{\lambda_j \delta} \leq \frac{2}{\delta \lambda}$. This yields that the following inequality holds with probability at least $1 - \delta$ for any $\mathbb{Q}$ on $\mathcal{H}$ and for any $\lambda \in (0, 1)$
\begin{align}\label{eq:es-any_lambda}
    |\mathcal{L}(\pi_{\mathbb{Q}}) -\hat{\mathcal{L}}_n^\alpha(\pi_{\mathbb{Q}})| \leq \sqrt{ \frac{D_{\mathrm{KL}}(\mathbb{Q} \| \mathbb{P})+\log \frac{8\sqrt{n}}{\delta \lambda}}{2n} } + B_n^\alpha(\pi_{\mathbb{Q}})  +
2 \frac{D_{\mathrm{KL}}(\mathbb{Q} \| \mathbb{P})+\log \frac{8}{\delta \lambda}}{n \lambda } + \frac{\lambda}{2}\operatorname{Var}_n^\alpha(\pi_{\mathbb{Q}})\,.
\end{align}
The additional $2$ in $2 \frac{D_{\mathrm{KL}}(\mathbb{Q} \| \mathbb{P})+\log \frac{8}{\delta \lambda}}{n \lambda }$ appears since we used that $\frac{1}{\lambda_j} \leq \frac{2}{\lambda}$. Similarly, the additional $\frac{2}{\lambda}$ in the logarithmic terms is due to the fact that $\frac{1}{\delta_j} \leq \frac{2}{\delta \lambda}$. Finally, setting
\begin{align*}
    & {\textsc{kl}^\prime}_{1}(\pi_{\mathbb{Q}}, \lambda)  =D_{\mathrm{KL}}(\mathbb{Q} \| \mathbb{P})+\log \frac{8\sqrt{n}}{\delta \lambda} \,,\\
    &{\textsc{kl}^\prime}_{2}(\pi_{\mathbb{Q}}, \lambda)  =  2\big( D_{\mathrm{KL}}(\mathbb{Q} \| \mathbb{P})+\log \frac{8 }{\delta \lambda}\big)\,,
\end{align*}
concludes the proof.
\end{proof}

Next, we provide a similar proof to extend \cref{thm:es-main_result} to any $\alpha \in (0, 1]$. While we only provide a one-sided inequality, the same covering technique can be used to obtain the other side of the inequality. 

\begin{proposition}[One-sided extension of \cref{thm:es-main_result} to hold simultaneously for any $\alpha \in (0,1)\cup \{1\}\,$]\label{prop:thm_alpha} Let $n \ge 1$, $\delta \in [0, 1]$, $\lambda >0 $, and let $\mathbb{P}$ be a fixed prior on $\mathcal{H}$, then with probability at least $1-\delta$ over draws $\cD_n \sim \mu_{\pi_0}^n$, the following holds simultaneously for any posterior $\mathbb{Q}$ on $\mathcal{H}$, and for any $\alpha \in (0, 1]$ that 
\begin{align*}
    \mathcal{L}(\pi_{\mathbb{Q}}) \leq \hat{\mathcal{L}}_n^\alpha(\pi_{\mathbb{Q}}) + \sqrt{ \frac{{\textsc{kl}^{\prime \prime}}_{1}(\pi_{\mathbb{Q}}, \alpha)}{2n} } + B_n^\alpha(\pi_{\mathbb{Q}})  +
\frac{{\textsc{kl}^{\prime \prime}}_{2}(\pi_{\mathbb{Q}}, \alpha)}{n \lambda } + \frac{\lambda}{2}\operatorname{Var}_n^{2\alpha}(\pi_{\mathbb{Q}})\,.
\end{align*}
where 
\begin{align*}
    & {\textsc{kl}^{\prime \prime}}_{1}(\pi_{\mathbb{Q}}, \alpha)  =D_{\mathrm{KL}}(\mathbb{Q} \| \mathbb{P})+\log \frac{8\sqrt{n}}{\delta \alpha} \,,\\
    &{\textsc{kl}^{\prime \prime}}_{2}(\pi_{\mathbb{Q}}, \alpha)  =  D_{\mathrm{KL}}(\mathbb{Q} \| \mathbb{P})+\log \frac{8 }{\delta \alpha}\,,\\
   & B_n^\alpha(\pi_{\mathbb{Q}}) = 1 - \frac{1}{n}\sum_{i=1}^{n} \E{A \sim  \pi_{\mathbb{Q}}(\cdot | X_i)}{\pi_0^{1-\alpha}(A | X_i)}\,,\\
    &\operatorname{Var}_n^\alpha(\pi_{\mathbb{Q}}) = \frac{1}{n}\sum_{i=1}^n  \E{A \sim  \pi_0(\cdot | X_i)}{\frac{\pi_{\mathbb{Q}}(A | X_i)}{\pi_0(A | X_i)^{2\alpha}}} + \frac{\pi_{\mathbb{Q}}(A_i | X_i)}{\pi_0(A_i | X_i)^{2\alpha}}C_i^2.
\end{align*}
\end{proposition}
\begin{proof}
    
Let $\delta \in (0, 1)$. For any $i \geq 0$, we define $\alpha_i=2^{-i}$ and let $\delta_i = \delta \alpha_i / 2$. Then \cref{thm:es-main_result} yields that for any $i \geq 0$, the following inequality holds with probability at least $1 - \delta_i$ for any $\mathbb{Q}$ on $\mathcal{H}$
\begin{align*}
    |\mathcal{L}(\pi_{\mathbb{Q}}) -\hat{\mathcal{L}}_n^{\alpha_i}(\pi_{\mathbb{Q}})| \leq \sqrt{ \frac{D_{\mathrm{KL}}(\mathbb{Q} \| \mathbb{P})+\log \frac{4\sqrt{n}}{\delta_i}}{2n} } + B_n^{\alpha_i}(\pi_{\mathbb{Q}})  +
\frac{D_{\mathrm{KL}}(\mathbb{Q} \| \mathbb{P})+\log \frac{4}{\delta_i}}{n \lambda } + \frac{\lambda}{2}\operatorname{Var}_n^{\alpha_i}(\pi_{\mathbb{Q}})\,.
\end{align*}
Now notice that $\sum_{i =0}^\infty \alpha_i = 2$, and hence by definition of $\delta_i$, we have $\sum_{i =0}^\infty \delta_i = \delta$. Therefore, the union bound of the above inequalities over $i \geq 0$ yields that with probability at least $1-\delta$, the following inequality holds with probability at least $1 - \delta$ for any $\mathbb{Q}$ on $\mathcal{H}$ and for any $i \geq 0$
\begin{align}\label{eq:es-for_any_alpha_i}
    |\mathcal{L}(\pi_{\mathbb{Q}}) -\hat{\mathcal{L}}_n^{\alpha_i}(\pi_{\mathbb{Q}})| \leq \sqrt{ \frac{D_{\mathrm{KL}}(\mathbb{Q} \| \mathbb{P})+\log \frac{4\sqrt{n}}{\delta_i}}{2n} } + B_n^{\alpha_i}(\pi_{\mathbb{Q}})  +
\frac{D_{\mathrm{KL}}(\mathbb{Q} \| \mathbb{P})+\log \frac{4}{\delta_i}}{n \lambda } + \frac{\lambda}{2}\operatorname{Var}_n^{\alpha_i}(\pi_{\mathbb{Q}})\,.
\end{align}
Let $\lfloor \cdot \rfloor$ denote the floor function, then we have that for any $\alpha \in (0, 1]$, there exists $j = \lfloor \frac{-\log \alpha}{\log 2} \rfloor \geq 0$ such that $\alpha \leq \alpha_j \leq 2 \alpha$. Since \eqref{eq:es-for_any_alpha_i} holds for any $i \geq 0$, it holds in particular for $j$. In addition, we have that $B_n^\alpha(\pi_{\mathbb{Q}})$ and $\hat{\mathcal{L}}_n^\alpha(\pi_{\mathbb{Q}})$ are decreasing in $\alpha$ while $\operatorname{Var}_n^\alpha(\pi_{\mathbb{Q}})$ is increasing in $\alpha$. Therefore, we have that $\hat{\mathcal{L}}_n^{\alpha_j}(\pi_{\mathbb{Q}}) \leq \hat{\mathcal{L}}_n^\alpha(\pi_{\mathbb{Q}})\,,$ $B_n^{\alpha_j}(\pi_{\mathbb{Q}}) \leq B_n^\alpha(\pi_{\mathbb{Q}})\,,$ and $\operatorname{Var}_n^{\alpha_j}(\pi_{\mathbb{Q}}) \leq \operatorname{Var}_n^{2\alpha}(\pi_{\mathbb{Q}})$. Moreover, we have that $\frac{1}{\delta_j}  \leq \frac{2}{\delta \alpha}$. This yields that the following inequality holds with probability at least $1 - \delta$ for any $\mathbb{Q}$ on $\mathcal{H}$ and for any $\alpha \in (0, 1]$
\begin{align}\label{eq:es-any_alpha}
    \mathcal{L}(\pi_{\mathbb{Q}})  \leq \hat{\mathcal{L}}_n^\alpha(\pi_{\mathbb{Q}}) + \sqrt{ \frac{D_{\mathrm{KL}}(\mathbb{Q} \| \mathbb{P})+\log \frac{8\sqrt{n}}{\delta \alpha}}{2n} } + B_n^\alpha(\pi_{\mathbb{Q}})  +
\frac{D_{\mathrm{KL}}(\mathbb{Q} \| \mathbb{P})+\log \frac{8}{\delta \alpha}}{n \lambda } + \frac{\lambda}{2}\operatorname{Var}_n^{2\alpha}(\pi_{\mathbb{Q}})\,.
\end{align}
Finally, setting
\begin{align*}
    & {\textsc{kl}^{\prime \prime}}_{1}(\pi_{\mathbb{Q}}, \alpha)  =D_{\mathrm{KL}}(\mathbb{Q} \| \mathbb{P})+\log \frac{8\sqrt{n}}{\delta \alpha} \,,\\
    &{\textsc{kl}^{\prime \prime}}_{2}(\pi_{\mathbb{Q}}, \alpha)  =  D_{\mathrm{KL}}(\mathbb{Q} \| \mathbb{P})+\log \frac{8 }{\delta \alpha}\,,
\end{align*}
concludes the proof.
\end{proof}

\subsection{Proof of \cref{prop:direct_application} }\label{proof:direct_application}

\citet[Theorem 7]{haddouche2022pac} provides an application of \cref{lemma:es-app_maxime} to the general PAC-Bayes learning problems in \cref{subsec:es-pac_bayes_framekwork}. We cannot apply their theorem directly to get \cref{prop:direct_application} for two reasons. They assume that the loss function is non-negative and they derive a one-sided generalization bound. In our case, the loss function is negative and we want to derive a two-sided generalization bound. Fortunately, we show with a slight modification of their proof that the result can be extended to two-sided inequalities with negative losses. In fact, the only requirement is that the sign of loss is fixed. We show next how this is achieved.
\begin{proof}
First, note that \cref{lemma:es-app_maxime} does not make any assumption on the sign of the martingale difference sequence $(f_i(S_i, h))_{i \in [n]}$ nor on the sign of the terms that decompose it. Now similarly to the proof in \cref{proof:main_thm_proof}, we define $S_n = (X_i, A_i)_{i \in [n]}$ as the set of $n$ observed contexts and taken actions. Also, we let $(\mathcal{F}_i)_{i \in \{0\}\cup[n]}$ be a filtration adapted to $S_n$. For $h \in \mathcal{H}$, we define $f_i\left(S_i, h\right)$ as 
\begin{align*}
    f_i\left(S_i, h\right) &= f_i\left(X_i, A_i, h\right) = f\left(X_i, A_i, h\right)\,,\\& = \E{X \sim \nu, A \sim  \pi_0(\cdot | X)}{\frac{\mathbb{I}_{\{h(X)=A\}}}{\pi_0(A | X)^\alpha}c(X, A)} - \frac{\mathbb{I}_{\{h(X_i)=A_i\}}}{\pi_0(A_i | X_i)^\alpha}c(X_i, A_i)\,.
\end{align*}
Here $f_i(S_i, h)$ only depends on the last samples $X_i, A_i$ and the predictor $h$. For this reason, we denote it by $f_i\left(X_i, A_i, h\right)$. Also, the function $f_i$ does not depend on $i$ and this is why we simplify the notation as $f_i\left(X_i, A_i, h\right) = f\left(X_i, A_i, h\right)$. Moreover, the randomness in $f\left(X_i, A_i, h\right)$ is only due $X_i \sim \nu$ and $A_i \sim \pi_0(\cdot | X_i)$; all other terms are deterministic. Thus the expectations are under $X_i \sim \nu, A_i \sim \pi_0(\cdot | X_i)$. Now similarly to the proof in \cref{proof:main_thm_proof}, we have that $\mathbb{E}\left[f\left(X_i, A_i, h\right) | \mathcal{F}_{i-1}\right]= 0$ for any $i \in [n]\,, h \in \mathcal{H}$. Therefore, $(f(X_i, A_i, h))_{i \in [n]}$ is a martingale difference sequence for any $h \in \mathcal{H}$. Thus we apply \cref{lemma:es-app_maxime} and get that that with probability at least $1-\delta$, the following holds simultaneously for any distribution $\mathbb{Q}$ on $\mathcal{H}$
\begin{align}\label{eq:es-app_direct_proof_0}
    \left|\E{h \sim \mathbb{Q}}{M_n(h)}\right| \leq \frac{D_{\mathrm{KL}}(\mathbb{Q} \| \mathbb{P})+\log (2 / \delta)}{\lambda}+\frac{\lambda}{2}\left(\E{h \sim \mathbb{Q}}{\langle M\rangle_n(h) + [M]_n(h)}\right)\,,
\end{align} 
where 
\begin{align*}
    &M_n(h)=\sum_{i=1}^n f\left(X_i, A_i, h\right)\,,\\
    &\langle M\rangle_n(h)=\sum_{i=1}^n \mathbb{E}\left[f\left(X_i, A_i, h\right)^2 | \mathcal{F}_{i-1}\right]\,,\\
    &[M]_n(h)=\sum_{i=1}^n f\left(X_i, A_i, h\right)^2\,.
\end{align*}
Now we compute $\E{h \sim \mathbb{Q}}{M_n(h)} $ as
\begin{align}\label{eq:es-app_direct_proof_1}
    \E{h \sim \mathbb{Q}}{M_n(h)}  &=\sum_{i=1}^n\E{X \sim \nu, A \sim  \pi_0(\cdot | X)}{\frac{ \pi_{\mathbb{Q}}(A | X) }{\pi_0(A | X)^\alpha}c(X, A)} - \frac{ \pi_{\mathbb{Q}}(A_i | X_i)}{\pi_0(A_i | X_i)^\alpha}c(X_i, A_i)\,,\nonumber\\
    &=n \E{X \sim \nu, A \sim  \pi_0(\cdot | X)}{\frac{ \pi_{\mathbb{Q}}(A | X) }{\pi_0(A | X)^\alpha}c(X, A)} - \sum_{i=1}^n \frac{ \pi_{\mathbb{Q}}(A_i | X_i)}{\pi_0(A_i | X_i)^\alpha}c(X_i, A_i)\,,
\end{align}
where we used the linearity of the expectation $\E{h \sim \mathbb{Q}}{\cdot}$ and the definition of policies in \eqref{eq:es-pac_policies}. Moreover, similarly to the proof in \cref{proof:main_thm_proof}, we have that
\begin{align}\label{eq:es-app_direct_proof_4}
  \langle M\rangle_n(h) + [M]_n(h)  & = \sum_{i=1}^n \mathbb{E}\left[f\left(X_i, A_i, h\right)^2 | \mathcal{F}_{i-1}\right] + f\left(X_i, A_i, h\right)^2\nonumber\\
    &=  \sum_{i=1}^n\E{X \sim \nu, A \sim  \pi_0(\cdot | X)}{\frac{\mathbb{I}_{\{h(X)=A\}}}{\pi_0(A | X)^{2\alpha}}c(X, A)^2}+ \frac{\mathbb{I}_{\{h(X_i)=A_i\}}}{\pi_0(A_i | X_i)^{2\alpha}}c(X_i, A_i)^2 \nonumber\\
    & \hspace{1.5cm} - 2   \E{X \sim \nu, A \sim  \pi_0(\cdot | X)}{\frac{\mathbb{I}_{\{h(X)=A\}}}{\pi_0(A | X)^\alpha}c(X, A)} \frac{\mathbb{I}_{\{h(X_i)=A_i\}}}{\pi_0(A_i | X_i)^\alpha}c(X_i, A_i)\,,\nonumber\\
    & \stackrel{(i)}{\leq} n\E{X \sim \nu, A \sim  \pi_0(\cdot | X)}{\frac{\mathbb{I}_{\{h(X)=A\}}}{\pi_0(A | X)^{2\alpha}}c(X, A)^2} +  \sum_{i=1}^n\frac{\mathbb{I}_{\{h(X_i)=A_i\}}}{\pi_0(A_i | X_i)^{2\alpha}}c(X_i, A_i)^2\,,
\end{align}
where $(i)$ holds since $- 2   \E{X \sim \nu, A \sim  \pi_0(\cdot | X)}{\frac{\mathbb{I}_{\{h(X)=A\}}}{\pi_0(A | X)^\alpha}c(X, A)} \frac{\mathbb{I}_{\{h(X_i)=A_i\}}}{\pi_0(A_i | X_i)^\alpha}c(X_i, A_i) \leq 0$ for any $i \in [n]$. This is where the non-negative loss assumption is not needed. Our loss $L_\alpha(h, x, a, c) = \frac{\mathbb{I}_{\{h(X)=A\}}}{\pi_0(a | x)^\alpha}c\,$ is negative since $c \in [-1, 0]$. However, we only need the product between the loss and its expectation to be non-negative. This holds in particular when the loss has a fixed sign. In that case, the expectation of the loss and the loss itself will have the same sign and thus their product will be non-negative. In our case, the loss has a fixed negative sign and this is all we needed. Now notice that 
\begin{align*}
    &n\E{X \sim \nu, A \sim  \pi_0(\cdot | X)}{\frac{ \pi_{\mathbb{Q}}(A | X) }{\pi_0(A | X)^\alpha}c(X, A)} = nR^\alpha(\pi_{\mathbb{Q}})\,,\\
   & \sum_{i=1}^n \frac{ \pi_{\mathbb{Q}}(A_i | X_i)}{\pi_0(A_i | X_i)^\alpha}c(X_i, A_i) = n \hat{\mathcal{L}}_n^\alpha(\pi_{\mathbb{Q}})\,,
\end{align*}
where we used that $c(X_i, A_i)=C_i$ for any $i \in [n]$ in the second equality. Using these two equalities and plugging \eqref{eq:es-app_direct_proof_1} and \eqref{eq:es-app_direct_proof_4} in \eqref{eq:es-app_direct_proof_0} yields that with probability at least $1 -\delta$, the following holds simultaneously for any distribution $\mathbb{Q}$ on $\mathcal{H}$
\begin{align}
    n\left|\mathcal{L}^\alpha(\pi_{\mathbb{Q}}) - \hat{\mathcal{L}}_n^\alpha(\pi_{\mathbb{Q}})\right|  \leq \frac{D_{\mathrm{KL}}(\mathbb{Q} \| \mathbb{P})+\log (2 / \delta)}{\lambda}+\frac{\lambda}{2} \Big(n\E{X \sim \nu, A \sim  \pi_0(\cdot | X)}{\frac{ \pi_{\mathbb{Q}}(A | X)}{\pi_0(A | X)^{2\alpha}}c(X, A)^2}\nonumber\\ +  \sum_{i=1}^n\frac{ \pi_{\mathbb{Q}}(A_i | X_i)}{\pi_0(A_i | X_i)^{2\alpha}}c(X_i, A_i)^2\Big)\,.
\end{align} 
Again we used the linearity of the expectation $\E{h \sim \mathbb{Q}}{\cdot}$ and the definition of policies in \eqref{eq:es-pac_policies}. Finally, we have that $c(X_i, A_i)=C_i$ for any $i \in [n]$. Thus with probability at least $1 -\delta$ the following inequality holds for any distribution $\mathbb{Q}$ on $\mathcal{H}$
\begin{align}
    \left|\mathcal{L}^\alpha(\pi_{\mathbb{Q}}) - \hat{\mathcal{L}}_n^\alpha(\pi_{\mathbb{Q}})\right| \leq \frac{D_{\mathrm{KL}}(\mathbb{Q} \| \mathbb{P})+\log (2 / \delta)}{n \lambda} +\frac{\lambda}{2} \E{X \sim \nu, A \sim  \pi_0(\cdot | X)}{\frac{\pi_{\mathbb{Q}}(A | X)}{\pi_0(A | X)^{2\alpha}}c(X, A)^2}   \nonumber\\ + \frac{\lambda}{2n}\sum_{i=1}^n\frac{\pi_{\mathbb{Q}}(A_i | X_i)}{\pi_0(A_i | X_i)^{2\alpha}}C_i^2\,.
\end{align} 
This concludes the proof.
\end{proof}

\subsection{Sample Complexity}\label{proof:practice_theory}
\textbf{Notation reminder:} Minimizing risk $\mathcal{L}(\pi)$ is equivalent to maximizing value $V(\pi) = -\mathcal{L}(\pi)$. Achieving $\mathcal{L}(\hat{\pi}) \leq \mathcal{L}(\pi_*) + \epsilon$ is equivalent to $V(\hat{\pi}) \geq V(\pi_*) - \epsilon$.

\begin{proposition}\label{prop:samples_oracle}
Let $\mathcal{M}_1(\mathcal{H})$ be the set of probability distributions on the hypothesis space $\mathcal{H}$, and let $\lambda>0$,  $n \ge 1$, $\delta \in [0, 1]$, $\alpha \in [0, 1]$, and let $\mathbb{P}$ be a fixed prior on $\mathcal{H}$, then with probability at least $1-\delta$ over draws $\cD_n \sim \mu_{\pi_0}^n$, we have
\begin{align*}
    \mathcal{L}( \pi_{\hat{\mathbb{Q}}_n}) \leq \mathcal{L}(\pi_{\mathbb{Q}_*})  + 2 \sqrt{ \frac{{\textsc{kl}}_{1}(\pi_{\mathbb{Q}_*})}{2n} } + 2B_n^\alpha(\pi_{\mathbb{Q}_*})  +
2\frac{{\textsc{kl}}_{2}(\pi_{\mathbb{Q}_*})}{n \lambda } + \lambda \operatorname{Var}_n^\alpha(\pi_{\mathbb{Q}_*})\,.
\end{align*}
where $\pi_{\hat{\mathbb{Q}}_n}\,$ is the learned policy with $\, \hat{\mathbb{Q}}_n =  \argmin_{\mathbb{Q} \in \mathcal{M}_1(\mathcal{H})} \hat{\mathcal{L}}_n^\alpha(\pi_{\mathbb{Q}}) + \sqrt{ \frac{{\textsc{kl}}_{1}(\pi_{\mathbb{Q}})}{2n} } + B_n^\alpha(\pi_{\mathbb{Q}})  +
\frac{{\textsc{kl}}_{2}(\pi_{\mathbb{Q}})}{n \lambda } + \frac{\lambda}{2}\operatorname{Var}_n^\alpha(\pi_{\mathbb{Q}})\,,$ $\mathbb{Q}_* =  \argmin_{\mathbb{Q} \in \mathcal{M}_1(\mathcal{H})} \mathcal{L}(\pi_{\mathbb{Q}})$, and
\begin{align*}
    &{\textsc{kl}}_{1}(\pi_{\mathbb{Q}})  =D_{\mathrm{KL}}(\mathbb{Q} \| \mathbb{P})+\log \frac{4\sqrt{n}}{\delta}\,, \qquad {\textsc{kl}}_{2}(\pi_{\mathbb{Q}})  =  D_{\mathrm{KL}}(\mathbb{Q} \| \mathbb{P})+\log \frac{4}{\delta}\,,\\
   & B_n^\alpha(\pi_{\mathbb{Q}}) = 1 - \frac{1}{n}\sum_{i=1}^{n} \E{A \sim  \pi_{\mathbb{Q}}(\cdot | X_i)}{\pi_0^{1-\alpha}(A | X_i)}\,, \\ &\operatorname{Var}_n^\alpha(\pi_{\mathbb{Q}}) = \frac{1}{n}\sum_{i=1}^n  \E{A \sim  \pi_0(\cdot | X_i)}{\frac{\pi_{\mathbb{Q}}(A | X_i)}{\pi_0(A | X_i)^{2\alpha}}} + \frac{\pi_{\mathbb{Q}}(A_i | X_i)C_i^2}{\pi_0(A_i | X_i)^{2\alpha}}\,.
\end{align*}
\end{proposition}

\begin{proof}
First, \cref{thm:es-main_result} holds for any potentially data dependent distribution $\mathbb{Q}$ on $\mathcal{H}$. In particular, we have that with probability at least $1-\delta$ the following inequalities hold simultaneously for $\hat{\mathbb{Q}}_n$ and $\mathbb{Q}_*$
\begin{align*}
    & |\mathcal{L}(\pi_{\hat{\mathbb{Q}}_n}) -\hat{\mathcal{L}}_n^\alpha(\pi_{\hat{\mathbb{Q}}_n})| \leq \sqrt{ \frac{{\textsc{kl}}_{1}(\pi_{\hat{\mathbb{Q}}_n})}{2n} } + B_n^\alpha(\pi_{\hat{\mathbb{Q}}_n})  +
\frac{{\textsc{kl}}_{2}(\pi_{\hat{\mathbb{Q}}_n})}{n \lambda } + \frac{\lambda}{2}\operatorname{Var}_n^\alpha(\pi_{\hat{\mathbb{Q}}_n})\,,\\
&|\mathcal{L}(\pi_{\mathbb{Q}_*}) -\hat{\mathcal{L}}_n^\alpha(\pi_{\mathbb{Q}_*})| \leq \sqrt{ \frac{{\textsc{kl}}_{1}(\pi_{\mathbb{Q}_*})}{2n} } + B_n^\alpha(\pi_{\mathbb{Q}_*})  +
\frac{{\textsc{kl}}_{2}(\pi_{\mathbb{Q}_*})}{n \lambda } + \frac{\lambda}{2}\operatorname{Var}_n^\alpha(\pi_{\mathbb{Q}_*})\,.
\end{align*}
Taking only one side of these inequalities yields that with probability at least $1-\delta$ the following inequalities hold simultaneously for $\hat{\mathbb{Q}}_n$ and $\mathbb{Q}_*$
\begin{align*}
    &\mathcal{L}(\pi_{\hat{\mathbb{Q}}_n}) \leq \underbrace{\hat{\mathcal{L}}_n^\alpha(\pi_{\hat{\mathbb{Q}}_n}) + \sqrt{ \frac{{\textsc{kl}}_{1}(\pi_{\hat{\mathbb{Q}}_n})}{2n} } + B_n^\alpha(\pi_{\hat{\mathbb{Q}}_n})  +
\frac{{\textsc{kl}}_{2}(\pi_{\hat{\mathbb{Q}}_n})}{n \lambda } + \frac{\lambda}{2}\operatorname{Var}_n^\alpha(\pi_{\hat{\mathbb{Q}}_n})}_{(I)}\,,\\
& \hat{\mathcal{L}}_n^\alpha(\pi_{\mathbb{Q}_*}) \leq \mathcal{L}(\pi_{\mathbb{Q}_*}) +  \sqrt{ \frac{{\textsc{kl}}_{1}(\pi_{\mathbb{Q}_*})}{2n} } + B_n^\alpha(\pi_{\mathbb{Q}_*})  +
\frac{{\textsc{kl}}_{2}(\pi_{\mathbb{Q}_*})}{n \lambda } + \frac{\lambda}{2}\operatorname{Var}_n^\alpha(\pi_{\mathbb{Q}_*})\,.
\end{align*}
Now using the definition of $\pi_{\hat{\mathbb{Q}}_n}$, we know that 
\begin{align*}
   I \leq \hat{\mathcal{L}}_n^\alpha(\pi_{\mathbb{Q}_*})  +  \sqrt{ \frac{{\textsc{kl}}_{1}(\pi_{\mathbb{Q}_*})}{2n} } + B_n^\alpha(\pi_{\mathbb{Q}_*})  +
\frac{{\textsc{kl}}_{2}(\pi_{\mathbb{Q}_*})}{n \lambda } + \frac{\lambda}{2}\operatorname{Var}_n^\alpha(\pi_{\mathbb{Q}_*})\,.
\end{align*}
This yields that with probability at least $1-\delta$ the following inequalities hold simultaneously for $\hat{\mathbb{Q}}_n$ and $\mathbb{Q}_*$
\begin{align*}
    &\mathcal{L}(\pi_{\hat{\mathbb{Q}}_n}) \leq \hat{\mathcal{L}}_n^\alpha(\pi_{\mathbb{Q}_*})  +  \sqrt{ \frac{{\textsc{kl}}_{1}(\pi_{\mathbb{Q}_*})}{2n} } + B_n^\alpha(\pi_{\mathbb{Q}_*})  +
\frac{{\textsc{kl}}_{2}(\pi_{\mathbb{Q}_*})}{n \lambda } + \frac{\lambda}{2}\operatorname{Var}_n^\alpha(\pi_{\mathbb{Q}_*})\,,\\
& \hat{\mathcal{L}}_n^\alpha(\pi_{\mathbb{Q}_*}) \leq \mathcal{L}(\pi_{\mathbb{Q}_*}) +  \sqrt{ \frac{{\textsc{kl}}_{1}(\pi_{\mathbb{Q}_*})}{2n} } + B_n^\alpha(\pi_{\mathbb{Q}_*})  +
\frac{{\textsc{kl}}_{2}(\pi_{\mathbb{Q}_*})}{n \lambda } + \frac{\lambda}{2}\operatorname{Var}_n^\alpha(\pi_{\mathbb{Q}_*})\,.
\end{align*}
Computing the sum of these two inequalities concludes the proof.
\end{proof}

\begin{corollary}[Special case of \cref{prop:samples_oracle}] Let $\mathcal{H} = \set{h_{\theta} \, ; \theta \in \real^{dK}}$ of mappings $h_{\theta}(x) = \argmax_{a \in \cA} \phi(x)^\top \theta_a$ for any $x \in \cX$. Let $n \ge 1$, $\delta \in [0, 1]$, $\alpha \in [0, 1]$, and let $\mathbb{P} = \cN(\mu_0, I_{dK})$ be a fixed prior on $\mathcal{H}$, then with probability at least $1-\delta$ over draws $\cD_n \sim \mu_{\pi_0}^n$, we have that
\begin{align*}
    \mathcal{L}( \pi_{\hat{\mathbb{Q}}_n}) \leq \mathcal{L}(\pi_{\mathbb{Q}_*})  &+  \frac{ \sqrt{\norm{\mu_* - \mu_0}^2 + 2\log\frac{4 \sqrt{n}}{\delta}} }{\sqrt{n}} \\ &+ 2(1 - K^{\alpha-1})  +
\frac{ \norm{\mu_* - \mu_0}^2 + 2\log\frac{4}{\delta}}{\sqrt{n}} + \frac{K^{2\alpha-1} + K^{2\alpha}}{\sqrt{n}}\,.
\end{align*}
where $\pi_{\hat{\mathbb{Q}}_n}\,$ is the learned policy with $\, \hat{\mathbb{Q}}_n =  \argmin_{\mathbb{Q} = \cN(\mu, I_{dK}) } \hat{\mathcal{L}}_n^\alpha(\pi_{\mathbb{Q}}) + \sqrt{ \frac{{\textsc{kl}}_{1}(\pi_{\mathbb{Q}})}{2n} } + B_n^\alpha(\pi_{\mathbb{Q}})  +
\frac{{\textsc{kl}}_{2}(\pi_{\mathbb{Q}})}{n \lambda } + \frac{\lambda}{2}\operatorname{Var}_n^\alpha(\pi_{\mathbb{Q}})\,,$ $\mathbb{Q}_* =  \argmin_{\mathbb{Q} = \cN(\mu, I_{dK})} \mathcal{L}(\pi_{\mathbb{Q}})$.
\end{corollary}

\begin{proof}
    This result follows from the general \cref{prop:samples_oracle} by simply setting $\mathbb{P} = \cN(\mu_0, I_{dK})$ and $\mathbb{Q}_* = \cN(\mu_*, I_{dK})$. First, since the covariance matrices of both distributions are $I_{dK}$, their KL divergence is $D_{\mathrm{KL}}(\mathbb{Q} \| \mathbb{P}) = \norm{\mu_* - \mu_0}^2/2$. Moreover, since the logging policy is uniform then $B_n^\alpha(\pi_{\mathbb{Q}}) = (1-K^{\alpha-1})$ and $ \operatorname{Var}_n^\alpha(\pi_{\mathbb{Q}})
 \leq K^{2\alpha-1} + K^{2\alpha}$.  Using these quantities, setting $\lambda = 1/\sqrt{n}$ and applying \cref{prop:samples_oracle} yields that with probability at least $1-\delta$ over draws $\cD_n \sim \mu_{\pi_0}^n$, we have that
\begin{align*}
    \mathcal{L}( \pi_{\hat{\mathbb{Q}}_n}) \leq \mathcal{L}(\pi_{\mathbb{Q}_*}) & +  \frac{ \sqrt{\norm{\mu_* - \mu_0}^2 + 2\log\frac{4 \sqrt{n}}{\delta}} }{\sqrt{n}}+ 2(1 - K^{\alpha-1}) \\
    &+ \frac{ \norm{\mu_* - \mu_0}^2 + 2\log\frac{4}{\delta}}{\sqrt{n}} + \frac{K^{2\alpha-1} + K^{2\alpha}}{\sqrt{n}}\,.
\end{align*}
This concludes the proof.
\end{proof}

The above corollary allows us to give insights into the sample complexity of our procedure. That is, the number of samples needed so that the performance of the learned policy $\pi_{\hat{\mathbb{Q}}_n}$ is close to that of the optimal one. Let $\epsilon> 2(1 - K^{\alpha-1})$ for $\alpha \in [1- \log 2 / \log K, 1]$. This condition on $\alpha$ ensures that $\epsilon \in [0, 1]$ and it is mild as $\alpha$ is often close to 1. Let $\delta$, then the following implication holds
\begin{align}\label{eq:es-inequality1}
    \epsilon \geq \frac{ \sqrt{\norm{\mu_* - \mu_0}^2 + 2\log\frac{4 \sqrt{n}}{\delta}} }{\sqrt{n}}+ 2(1 - K^{\alpha-1}) + \frac{ \norm{\mu_* - \mu_0}^2 + 2\log\frac{4}{\delta}}{\sqrt{n}} + \frac{K^{2\alpha-1} + K^{2\alpha}}{\sqrt{n}}  \nonumber \\ \implies \mathbb{P}(\mathcal{L}(\pi_{\hat{\mathbb{Q}}_n}) \leq \mathcal{L}(\pi_{\mathbb{Q}_*}) + \epsilon) \geq 1-\delta\,.
\end{align}
First, we use that $\sqrt{\norm{\mu_* - \mu_0}^2 + 2\log\frac{4 \sqrt{n}}{\delta}} \leq \norm{\mu_* - \mu_0} + \sqrt{2\log\frac{4 \sqrt{n}}{\delta}}$. Moreover we bound $K^{2\alpha-1} + K^{2\alpha} \leq 2K^{2 \alpha}$. Then the implication in \eqref{eq:es-inequality1} becomes
\begin{align}\label{eq:es-inequality2}
   \sqrt{n} \geq \frac{\norm{\mu_* - \mu_0} + \norm{\mu_* - \mu_0}^2 + 2\log\frac{4}{\delta} + \sqrt{2\log\frac{4 \sqrt{n}}{\delta}} + 2K^{2\alpha} }{\epsilon - 2(1 - K^{\alpha-1})}  \implies \mathbb{P}(\mathcal{L}(\pi_{\hat{\mathbb{Q}}_n}) \leq \mathcal{L}(\pi_{\mathbb{Q}_*}) + \epsilon) \geq 1-\delta\,.
\end{align}
We only provide intuition on the sample complexity and aim at having easy-to-interpret terms. Thus we omit the logarithmic terms in \eqref{eq:es-inequality2} and assume that $\norm{\mu_* - \mu_0}^2 \geq \norm{\mu_* - \mu_0}$. This leads to the claim made in \cref{subsec:es-interpretation}. Of course, a more precise sample complexity analysis can be made by studying the function $h(x) = \sqrt{x} -  \sqrt{2\log\frac{4 \sqrt{x}}{\delta}}/(\epsilon - 2(1 - K^{\alpha-1}))$ and finding $x$ such that $f(x) \geq \frac{\norm{\mu_* - \mu_0} + \norm{\mu_* - \mu_0}^2 + 2\log\frac{4}{\delta} + 2K^{2\alpha} }{\epsilon - 2(1 - K^{\alpha-1})}$.
\section{Experiments}\label{app:all_experiments}

\subsection{Setup}\label{app:setup}
We consider the standard supervised-to-bandit conversion \citep{agarwal2014taming}. Precisely, let $\mathcal{S}^{\textsc{tr}}_{n}$ and $\mathcal{S}^{\textsc{ts}}_{n_{\textsc{ts}}}$ be the training and testing set of a classification dataset, respectively. First, we transform the training set $\mathcal{S}^{\textsc{tr}}_{n}$ to a logged bandit data $\cD_n$ as described in \cref{alg:es-supervised_to_bandit}. The resulting logged data $\cD_n$ is then used to train our policies. After that, the learned policies are tested on $\mathcal{S}^{\textsc{ts}}_{n_{\textsc{ts}}}$ as described in \cref{alg:es-supervised_to_bandit_test}. We consider that the resulting reward in \cref{alg:es-supervised_to_bandit_test} is a good proxy for the unknown true reward of the learned policies. This will be our performance metric, the higher the better.

In our experiments, we use the following image classification datasets \texttt{MNIST} \citep{lecun1998gradient}, \texttt{FashionMNIST} \citep{xiao2017fashion},  \texttt{EMNIST} \citep{cohen2017emnist} and \texttt{CIFAR100} \citep{krizhevsky2009learning}. We provide a summary of the statistics of these datasets in \cref{tab:stats}. \cref{alg:es-supervised_to_bandit} takes as input a logging policy $\pi_0$ which we define as
\begin{align}\label{eq:es-logging}
    &\pi_0(a| x) = \frac{ \exp(\eta_0 \cdot \phi(x)^\top \mu_{0, a})}{\sum_{a^\prime \in \cA}  \exp(\eta_0 \cdot \phi(x)^\top \mu_{0, a^\prime})}\,, & \forall (x,a) \in \cX \times \cA\,.
\end{align}
Here $\phi(x) \in \real^d$ is the feature transformation function that outputs a $d$-dimensional vector, $\mu_0 = (\mu_{0,a})_{a \in \cA} \in \real^{dK}$ are learnable parameters and $\eta_0$ is an inverse-temperature parameter for the \texttt{softmax} in \eqref{eq:es-logging}. We explain next how these quantities are derived in detail.

\textbf{The feature transformation function $\phi(x) \in \real^d$:} for all the datasets, except \texttt{CIFAR100}, the feature transformation function $\phi(\cdot)$ is defined as $\phi(x) = \frac{x}{\norm{x}}$ for any $x \in \cX$. That is, we simply normalize the features $x \in \cX$ by their $L_2$ norm $\norm{x}$. In contrast, \texttt{CIFAR100} is a more challenging problem. Thus we use transfer learning to extract features $\phi(x)$ expressive enough so that a linear \texttt{softmax} model would enjoy a reasonable performance. Precisely, we retrieve the last hidden layer of a \texttt{ResNet-50} network, pre-trained on the ImageNet dataset, to output 2048-dimensional features. Finally, the obtained features are normalized as $\frac{x}{\norm{x}}$ and this whole process (\texttt{ResNet-50}  + normalization) corresponds to $\phi(\cdot)$ for \texttt{CIFAR100}.

\textbf{The parameters $\mu_0 = (\mu_{0,a})_{a \in \cA} \in \real^{dK}$:} we learn the parameters $\mu_0$ using 5\% of the training set $\mathcal{S}^{\textsc{tr}}_{n}$. Precisely, we use the cross-entropy loss with an $L_2$ regularization of $10^{-6}$ to prevent the logging policy $\pi_0$ from being degenerate.  This ensures that the learning policies are absolutely continuous with respect to the logging policy $\pi_0$, a condition under which standard IPS is unbiased. In optimization, we use Adam \citep{kingma2014adam} with a learning rate of $0.1$ for $10$ epochs. In all the experiments, we set the prior $\mathbb{P} = \cN(\eta_0 \mu_0, I_{dK})$ for the Gaussian policies in \eqref{eq:es-gaussian_pac_bayes} and we set it as $\mathbb{P} = \cN(\eta_0 \mu_0, I_{dK}) \times {\rm G}(0, 1)^K$ for the mixed-logit policies in \eqref{eq:es-logit_pac_bayes}. Our theory requires that the prior does not depend on data. Given that $\mu_0$ is learned on the $5\%$ portion of data, we only train our learning policies on the remaining  $95\%$ portion of the data to match our theoretical requirements.

\textbf{The inverse-temperature parameter $\eta_0 \in \real$:} this controls the performance of the logging policy. A high positive value of $\eta_0$ leads to a well-performing logging policy, while a negative one leads to a low-performing logging policy. When $\eta_0=0$, $\pi_0$ is identical to the uniform policy. In our experiments $\eta_0$ varies between $0$ and $1$.

\begin{algorithm}
\caption{Supervised-to-bandit: creating logged data}
\label{alg:es-supervised_to_bandit}
\textbf{Input:} training classification set $\mathcal{S}^{\textsc{tr}}_{n}=\{(X_i, y_i)\}_{i=1}^n$, logging policy $\pi_0$.\\
\textbf{Output:} logged bandit data $\mathcal{D}_n=(X_i, A_i, C_i)_{i \in [n]}$.\\
Initialize $\mathcal{D}_n =\{\}$ \\
\For{$i=1, \dots, n$}
{$A_i \sim \pi_0(\cdot | X_i)$\\
$C_i = - \mathbb{I}_{\{A_i = y_i\}}$\\
$\mathcal{D}_n \gets \mathcal{D}_n \cup \{(X_i, A_i, C_i)\}\,.$ 
}
\end{algorithm}

\begin{algorithm}
\caption{Supervised-to-bandit: testing policies}
\label{alg:es-supervised_to_bandit_test}
\textbf{Input:} image classification dataset $\mathcal{S}^{\textsc{ts}}_{n_{\textsc{ts}}}=\{(X_i, y_i)\}_{i=1}^{n_{\textsc{ts}}}$, learned policy $ \hat{\pi}_n$.\\
\textbf{Output:} reward $r$.\\
\For{$i=1, \dots, n_{\textsc{ts}}$}
{$A_i \sim \hat{\pi}_n(\cdot | X_i)$\\
$R_i = \mathbb{I}_{\{A_i = y_i\}}$}
$r = \frac{1}{n_{\textsc{ts}}} \sum_{i=1}^{n_{\textsc{ts}}} R_i\,.$
\end{algorithm}

\begin{table}[t]
\caption{Statistics of the datasets used in our experiments.}
\label{tab:stats}
\vskip 0.15in
\begin{center}
\begin{tiny}
\begin{sc}
\begin{tabular}{lcccr}
\toprule
Data set & Nbr. train samples $n$ & Nbr. test samples $n_{\textsc{ts}}$ & Nbr. actions $K$ & Dimension $d$ \\
\bottomrule
\texttt{MNIST}    & 60000 & 10000 & 10& 784 \\
\texttt{FashionMNIST} & 60000 & 10000 & 10& 784\\
\texttt{EMNIST}    & 112800 &18800 & 47& 784 \\
\texttt{CIFAR100}    & 50000 & 10000 & 100&   2048\\
\bottomrule
\end{tabular}
\end{sc}
\end{tiny}
\end{center}
\vskip -0.1in
\end{table}

Now it remains to explain the learning policies $\pi_{\mathbb{Q}}$ and the corresponding closed-form bounds using either our results or those in existing works \citep{london2019bayesian,sakhi2022pac}. 

\subsection{Policies}\label{app:policies}
Here we present the two families of policies that we use in our experiments, Gaussian and mixed-logit policies.
\subsubsection{Mixed-Logit}
Let $\mathcal{H} = \set{h_{\theta, \gamma} \, ; \theta \in \real^{dK}, \gamma \in \real^K}$ be a hypothesis space of mappings $h_{\theta, \gamma}(x) = \argmax_{a \in \cA} \phi(x)^\top \theta_a + \gamma_a$ for any $x \in \cX$. Here $\phi(x)$ outputs a $d$-dimensional representation of context $x \in \cX$. Now assume that for any $a \in \cA$, $\gamma_a$ is a standard Gumbel perturbation, $\gamma_a \sim {\rm G}(0, 1)$, then we have that 
\begin{align}\label{eq:es-app_softmax_pac_bayes}
    \pi^{\textsc{sof}}_{\theta}(a | x) &= \frac{\exp(\phi(x)^\top \theta_a)}{\sum_{a^\prime \in \cA}\exp(\phi(x)^\top  \theta_{a^\prime})}\,,\nonumber\\
    &= \E{\gamma \sim {\rm G}(0, 1)^K}{\mathbb{I}_{\{ h_{\theta, \gamma}(x) = a \}}}\,.
\end{align}
In addition, we randomize $\theta$ such as $\theta \sim \cN(\mu, \sigma^2 I_{dK})$ where $\mu \in \real^{dK}$ and $\sigma>0$. It follows that the posterior $\mathbb{Q}$ is a multivariate Gaussian $\cN(\mu, \sigma^2 I_{dK})$ over the parameters $\theta$ with standard Gumbel perturbations $\gamma \sim {\rm G}(0, 1)^K$. We denote such policies by $\pi^{\textsc{mixL}}_{\mu, \sigma}$ and they are defined as
\begin{align}\label{eq:es-app_logit_pac_bayes}
    \pi^{\textsc{mixL}}_{\mu, \sigma}(a | x) 
    &=  \E{\theta \sim \cN(\mu, \sigma^2 I_{dK})}{\frac{\exp(\phi(x)^\top \theta_a)}{\sum_{a^\prime \in \cA}\exp(\phi(x)^\top  \theta_{a^\prime})}}\,,\nonumber\\
    &=  \E{\theta \sim \cN(\mu, \sigma^2 I_{dK})}{\pi^{\textsc{sof}}_{\theta}(a | x)}\,,\nonumber\\
    &= \E{\theta \sim \cN(\mu, \sigma^2 I_{dK})\,, \gamma \sim {\rm G}(0, 1)^K}{\mathbb{I}_{\{ h_{\theta, \gamma}(x) = a \}}}\,.
\end{align}
To sample from the mixed-logit policies $\pi^{\textsc{mixL}}_{\mu, \sigma}$, we first sample $\theta \sim \cN(\mu, \sigma^2 I_{dK})$ and $\gamma \sim {\rm G}(0, 1)^K$ and then set the sampled action as $a \gets h_{\theta, \gamma}(x)$. Now we also need to compute the gradient of the expectation in \eqref{eq:es-app_logit_pac_bayes}. This needs additional care since the distribution under which we take the expectation depends on the parameters $\mu, \sigma$. Fortunately, the reparameterization trick can be used in this case. Roughly speaking, it allows us to express a gradient of the expectation in \eqref{eq:es-app_logit_pac_bayes} as an expectation of a gradient. In our case, we use the \emph{local} reparameterizaton trick \citep{kingma2015variational} which is known for reducing the variance of stochastic gradients. Precisely, we rewrite \eqref{eq:es-app_logit_pac_bayes} as
\begin{align}
      \pi^{\textsc{mixL}}_{\mu, \sigma}(a | x)  &=  \E{\epsilon \sim \cN(0, \norm{\phi(x)}^2 I_{K})}{\frac{\exp(\phi(x)^\top \mu_a + \sigma \epsilon_a)}{\sum_{a^\prime \in \cA}\exp(\phi(x)^\top  \mu_{a^\prime} + \sigma \epsilon_{a^\prime})}}\,.\nonumber\\
      &=  \E{\epsilon \sim \cN(0, I_{K})}{\frac{\exp(\phi(x)^\top \mu_a + \sigma \epsilon_a)}{\sum_{a^\prime \in \cA}\exp(\phi(x)^\top  \mu_{a^\prime} + \sigma \epsilon_{a^\prime})}}\,,\nonumber
\end{align}
where we used that $\norm{\phi(x)}^2=1$ since features are normalized. It follows that gradients read 
$$\nabla_{\mu, \sigma} \pi^{\textsc{mixL}}_{\mu, \sigma}(a | x) =  \E{\epsilon \sim \cN(0, I_{K})}{\nabla_{\mu, \sigma}\frac{\exp(\phi(x)^\top \mu_a + \sigma \epsilon_a)}{\sum_{a^\prime \in \cA}\exp(\phi(x)^\top  \mu_{a^\prime} + \sigma \epsilon_{a^\prime})}}\,.$$
Moreover, the propensities are approximated as 
\begin{align}
      &\pi^{\textsc{mixL}}_{\mu, \sigma}(a | x)  \approx \frac{1}{S} \sum_{i \in [S]}{\frac{\exp(\phi(x)^\top \mu_a + \sigma \epsilon_{i, a})}{\sum_{a^\prime \in \cA}\exp(\phi(x)^\top  \mu_{a^\prime} + \sigma \epsilon_{i, a^\prime})}}\,, & 
\epsilon_i  \sim \cN(0, I_{K})\,, \forall i \in [S]\,.
\end{align}
In all our experiments, we set $S=32$.

\subsubsection{Gaussian}
We define the hypothesis space $\mathcal{H} = \set{h_{\theta} \, ; \theta \in \real^{dK}}$ of mappings $h_{\theta}(x) = \argmax_{a \in \cA} \phi(x)^\top \theta_a$ for any $x \in \cX$. It follows that the learning policies $ \pi_{\mathbb{Q}} = \pi^{\textsc{gaus}}_{\mu, \sigma}$ read 
\begin{align}\label{eq:es-app_gaussian_pac_bayes}
   \pi^{\textsc{gaus}}_{\mu, \sigma}(a | x) = \E{\theta \sim \cN(\mu, \sigma^2 I_{dK})}{\mathbb{I}_{\{ h_{\theta}(x) = a \}}}\,.
\end{align}
To see why this can be beneficial \citep{sakhi2022pac}, let $\pi_*$ be the optimal policy. Given $x \in \cX$, $\pi_*(\cdot | x)$ should be deterministic; it chooses the best action for context $x$ with probability 1. That is, there exists $\mu_* \in \real^{dK}$ such that $\pi_* = \mathbb{I}_{\{ h_{\mu_*}(x) = a \}}$. When $\mu \rightarrow \mu_*$ and $\sigma \rightarrow 0$, the Gaussian policy in \eqref{eq:es-app_gaussian_pac_bayes} approaches $\pi_*$. In contrast, the mixed-logit policy in \eqref{eq:es-app_logit_pac_bayes} approaches $\pi^{\textsc{sof}}_{\mu_*}$. However, $\pi^{\textsc{sof}}_{\mu_*}$ is not deterministic due to the additional randomness in $\gamma$ and is equal to $\pi_*$ only if $\phi(x)^\top \mu_{*,a_*(x)} \rightarrow \infty$. This explains the choice of removing the Gumbel noise. First, \citet{sakhi2022pac} showed that \eqref{eq:es-app_gaussian_pac_bayes} can be written as 
\begin{align*}
 \pi^{\textsc{gaus}}_{\mu, \sigma}(a | x)  & =\mathbb{E}_{\epsilon \sim \cN(0, 1)}\Big[\prod_{a^{\prime} \neq a} \Phi\big(\epsilon+\frac{\phi(x)^\top\left(\mu_a-\mu_{a^{\prime}}\right)}{\sigma\|\phi(x)\|}\big)\Big]\,,
\end{align*}
where $\Phi$ is the cumulative distribution function of a standard normal variable.  But $\|\phi(x)\|=1$ in all our experiments. Thus
\begin{align*}
 \pi^{\textsc{gaus}}_{\mu, \sigma}(a | x)  & =\mathbb{E}_{\epsilon \sim \cN(0, 1)}\Big[\prod_{a^{\prime} \neq a} \Phi\big(\epsilon+\frac{\phi(x)^\top\left(\mu_a-\mu_{a^{\prime}}\right)}{\sigma}\big)\Big]\,.
\end{align*}
Then similarly to mixed-logit policies, the gradient reads 
\begin{align*}
    \nabla_{\mu, \sigma} \pi^{\textsc{gaus}}_{\mu, \sigma}(a | x) =  \mathbb{E}_{\epsilon \sim \cN(0, 1)}\Big[\nabla_{\mu, \sigma}\prod_{a^{\prime} \neq a} \Phi\big(\epsilon+\frac{\phi(x)^\top\left(\mu_a-\mu_{a^{\prime}}\right)}{\sigma}\big)\Big]\,.
\end{align*}
Moreover, the propensities are approximated as 
\begin{align}
      &\pi^{\textsc{gaus}}_{\mu, \sigma}(a | x)  \approx \frac{1}{S} \sum_{i \in [S]}{\prod_{a^{\prime} \neq a} \Phi\big(\epsilon_i+\frac{\phi(x)^\top\left(\mu_a-\mu_{a^{\prime}}\right)}{\sigma}\big)}\,, & 
\epsilon_i  \sim \cN(0, 1)\,, \forall i \in [S]\,.
\end{align}
In all our experiments, we set $S=32$.

\subsection{Baselines}\label{app:baselines}
Here we present all the methods that we use in our experiments. For each method, we state the result that holds for any learning policy $\pi$. After that, we derive the corresponding closed-form bounds for Gaussian and mixed-logit policies that we presented previously. All the baselines require computing the KL divergence between the prior $\mathbb{P}$ and the posterior $\mathbb{Q}$. Thus before presenting them, we state the following lemma that allows bounding the KL divergence between the prior $\mathbb{P}$ and the posterior $\mathbb{Q}$ in the cases of mixed-logit or Gaussian policies.

\begin{lemma}[KL divergence for Gaussian distributions with Gumbel noise]\label{lemma:es-kl_gaussian}
For distributions $\mathbb{P} = \mathcal{N}\left(\mu_0, \sigma_0^2 I_{dK}\right) \times \operatorname{G}(0,1)^K$ and $\mathbb{Q} = \mathcal{N}\left(\mu, \sigma^2 I_{dK}\right) \times \operatorname{G}(0,1)^K$, with $\mu_0, \mu \in \mathbb{R}^{dK}$ and $0<\sigma^2 \leq \sigma_0^2<\infty$,
$$
D_{\mathrm{KL}}(\mathbb{Q} \| \mathbb{P}) \leq \frac{\left\|\mu-\mu_0\right\|^2}{2 \sigma_0^2}+\frac{dK}{2} \log \frac{\sigma_0^2}{\sigma^2}\,.
$$
Moreover, this result holds when the Gumbel noise is removed. That is when $\mathbb{P} = \mathcal{N}\left(\mu_0, \sigma_0^2 I_{dK}\right)$ and $\mathbb{Q} = \mathcal{N}\left(\mu, \sigma^2 I_{dK}\right)$.
\end{lemma}

We borrow this lemma from \citet{london2019bayesian}. In particular, \cref{lemma:es-kl_gaussian} shows that the KL terms for both policies can be bounded by the same quantity. As a result, the corresponding bounds will be the same; the only difference is the space of learning policies on which we optimize. For completeness, however, we write these bounds for both types of policies although they are similar. Since existing approaches are not named, we name them as \textbf{(Author, Policy)} where \textbf{Author $\in$ \{Ours, London et al., Sakhi et al. 1, Sakhi et al. 2\} } and \textbf{Policy $\in$ \{Gaussian, Mixed-Logit\} }. Here \textbf{Ours}, \textbf{London et al.}, \textbf{Sakhi et al. 1} and\textbf{ Sakhi et al. 2} correspond to \cref{thm:es-main_result},  \citet[Theorem 1]{london2019bayesian}, \citet[Proposition 1]{sakhi2022pac}, \citet[Proposition 3]{sakhi2022pac}, respectively. For example, \citet[Theorem 1]{london2019bayesian} leads to two baselines \textbf{(London et al., Gaussian)} and \textbf{(London et al., Mixed-Logit)}. In all our experiments, the learning policies are trained using Adam \citep{kingma2014adam} with a learning rate of $0.1$ for $20$ epochs.

\subsubsection{Ours, \cref{thm:es-main_result}}

\textbf{(Ours, Gaussian)} Here we use the Gaussian policies in \eqref{eq:es-app_gaussian_pac_bayes}. Thus we only replace the term, $D_{\mathrm{KL}}(\mathbb{Q} \| \mathbb{P})$, with its closed-form bound in \cref{lemma:es-kl_gaussian}. This leads to the following objective. 
\begin{align*}
  \min_{\mu \in \real^{dK}, \sigma >0} \Big( \hat{\mathcal{L}}_n^\alpha\left(\pi^{\textsc{gaus}}_{\mu, \sigma}\right) &+ \sqrt{ \frac{\frac{\left\|\mu-\mu_0\right\|^2}{2} - \frac{dK}{2} \log \sigma^2+\log \frac{4\sqrt{n}}{\delta}}{2n} } + B_n^\alpha(\pi^{\textsc{gaus}}_{\mu, \sigma})\\&  +
\frac{\frac{\left\|\mu-\mu_0\right\|^2}{2} - \frac{dK}{2} \log \sigma^2+\log \frac{4}{\delta}}{n \lambda } + \frac{\lambda}{2}\operatorname{Var}_n^\alpha(\pi^{\textsc{gaus}}_{\mu, \sigma})\Big)\,, 
\end{align*}
where we used that $\sigma_0=1$ since our prior is $\mathbb{P}=\cN(\eta_0 \mu_0, I_{dK})$ for Gaussian policies. Moreover, we set $\lambda=\sqrt{2\frac{\frac{\left\|\mu-\mu_0\right\|^2}{2} - \frac{dK}{2} \log \sigma^2+\log \frac{4}{\delta}}{n \operatorname{Var}_n^\alpha(\pi^{\textsc{gaus}}_{\mu, \sigma}) }}$.

\textbf{(Ours, Mixed-Logit)} Here we use the mixed-logit policies in \eqref{eq:es-app_logit_pac_bayes}. Thus we only replace the terms, $D_{\mathrm{KL}}(\mathbb{Q} \| \mathbb{P})$, with their closed-form bound in \cref{lemma:es-kl_gaussian}. This leads to the following objective. 
\begin{align*}
  \min_{\mu \in \real^{dK}, \sigma >0} \Big( \hat{\mathcal{L}}_n^\alpha\left(\pi^{\textsc{mixL}}_{\mu, \sigma}\right) + & \sqrt{ \frac{\frac{\left\|\mu-\mu_0\right\|^2}{2} - \frac{dK}{2} \log \sigma^2+\log \frac{4\sqrt{n}}{\delta}}{2n} } + B_n^\alpha(\pi^{\textsc{mixL}}_{\mu, \sigma})\\&  +
\frac{\frac{\left\|\mu-\mu_0\right\|^2}{2} - \frac{dK}{2} \log \sigma^2+\log \frac{4}{\delta}}{n \lambda } 
+ \frac{\lambda}{2}\operatorname{Var}_n^\alpha(\pi^{\textsc{mixL}}_{\mu, \sigma})\Big)\,, 
\end{align*}
where we used that $\sigma_0=1$ since our prior is $\mathbb{P} = \cN(\eta_0 \mu_0, I_{dK}) \times {\rm G}(0, 1)^K$ for mixed-logit policies. Moreover, we set $\lambda=\sqrt{2\frac{\frac{\left\|\mu-\mu_0\right\|^2}{2} - \frac{dK}{2} \log \sigma^2+\log \frac{4}{\delta}}{n \operatorname{Var}_n^\alpha(\pi^{\textsc{mixL}}_{\mu, \sigma}) }}$.

\subsubsection{\citet[Theorem 1]{london2019bayesian}}

\begin{proposition}\label{prop:london}
Let $\tau \in (0, 1)$, $n \geq 1$, $\delta \in(0,1)$ and let $\mathbb{P}$ be a fixed prior on $\mathcal{H}$, then with probability at least $1-\delta$ over draws $\cD_n \sim \mu_{\pi_0}^n$, the following holds simultaneously for all posteriors, $\mathbb{Q}$, on $\mathcal{H}$ that
\begin{align}
R\left(\pi_{\mathbb{Q}}\right) & \leq \hat{\mathcal{L}}_n^\tau\left(\pi_{\mathbb{Q}}\right) +\sqrt{\frac{2\left(\hat{\mathcal{L}}_n^\tau\left(\pi_{\mathbb{Q}}\right)+\frac{1}{\tau}\right)\left(D_{\mathrm{KL}}(\mathbb{Q} \| \mathbb{P})+\log \frac{n}{\delta}\right)}{\tau(n-1)}} +\frac{2\left(D_{\mathrm{KL}}(\mathbb{Q} \| \mathbb{P})+\log \frac{n}{\delta}\right)}{\tau(n-1)}\,.
\end{align}

\end{proposition}

\textbf{Baseline 1: (London et al., Gaussian)} Here we use the Gaussian policies in \eqref{eq:es-app_gaussian_pac_bayes}. Thus we only replace the terms, $D_{\mathrm{KL}}(\mathbb{Q} \| \mathbb{P})$, with their closed-form bound in \cref{lemma:es-kl_gaussian}. This leads to the following objective. 
\begin{align*}
  \min_{\mu \in \real^{dK}, \sigma >0} \Big( \hat{\mathcal{L}}_n^\tau\left(\pi^{\textsc{gaus}}_{\mu, \sigma}\right) +\sqrt{\frac{2\left(\hat{\mathcal{L}}_n^\tau\left(\pi^{\textsc{gaus}}_{\mu, \sigma}\right)+\frac{1}{\tau}\right)\left(\frac{\left\|\mu-\mu_0\right\|^2}{2} - \frac{dK}{2} \log \sigma^2+\log \frac{n}{\delta}\right)}{\tau(n-1)}}\\ +\frac{2\left(\frac{\left\|\mu-\mu_0\right\|^2}{2} - \frac{dK}{2} \log \sigma^2+\log \frac{n}{\delta}\right)}{\tau(n-1)}\Big)\,, 
\end{align*}
where we used that $\sigma_0=1$ since our prior is $\mathbb{P}=\cN(\eta_0 \mu_0, I_{dK})$ for Gaussian policies.

\textbf{Baseline 2: (London et al., Mixed-Logit)} Here we consider the mixed-logit policies in \eqref{eq:es-app_logit_pac_bayes}. Since the additional Gumbel noise does not affect the KL divergence (\cref{lemma:es-kl_gaussian}), we have the same objective as in the Gaussian case. That is
\begin{align*}
  \min_{\mu \in \real^{dK}, \sigma >0} \Big( \hat{\mathcal{L}}_n^\tau\left(\pi^{\textsc{mixL}}_{\mu, \sigma}\right) +\sqrt{\frac{2\left(\hat{\mathcal{L}}_n^\tau\left(\pi^{\textsc{mixL}}_{\mu, \sigma}\right)+\frac{1}{\tau}\right)\left(\frac{\left\|\mu-\mu_0\right\|^2}{2} - \frac{dK}{2} \log \sigma^2+\log \frac{n}{\delta}\right)}{\tau(n-1)}}\\+\frac{2\left(\frac{\left\|\mu-\mu_0\right\|^2}{2} - \frac{dK}{2} \log \sigma^2+\log \frac{n}{\delta}\right)}{\tau(n-1)}\Big)\,,
\end{align*}
where we used that $\sigma_0=1$ since our prior is $\mathbb{P} = \cN(\eta_0 \mu_0, I_{dK}) \times {\rm G}(0, 1)^K$ for mixed-logit policies.

\subsubsection{\citet[Proposition 1]{sakhi2022pac}}

\begin{proposition}\label{prop:sakhi1}
Let $\tau \in (0, 1)$, $n \geq 1$, $\delta \in(0,1)$ and let $\mathbb{P}$ be a fixed prior on $\mathcal{H}$, then with probability at least $1-\delta$ over draws $\cD_n \sim \mu_{\pi_0}^n$, the following holds simultaneously for all posteriors, $\mathbb{Q}$, on $\mathcal{H}$ that
\begin{align}
R\left(\pi_{\mathbb{Q}}\right) & \leq \min_{\lambda >0} \frac{1}{\tau\left(e^\lambda-1\right)}\Big(1-e^{-\tau \lambda \hat{\mathcal{L}}_n^\tau\left(\pi_{\mathbb{Q}}\right)+\frac{D_{\mathrm{KL}}(\mathbb{Q} \| \mathbb{P})+\log \frac{2 \sqrt{n}}{\delta}}{n}}\Big)\,.
\end{align}
\end{proposition}

\textbf{Baseline 3: (Sakhi et al. 1, Gaussian)} Here we use the Gaussian policies in \eqref{eq:es-app_gaussian_pac_bayes}.
\begin{align}
\min_{\mu \in \real^{dK}, \sigma >0, \lambda>0} \Big(\frac{1}{\tau\left(e^\lambda-1\right)}\Big(1-e^{-\tau \lambda \hat{\mathcal{L}}_n^\tau\left(\pi^{\textsc{gaus}}_{\mu, \sigma}\right)+\frac{\frac{\left\|\mu-\mu_0\right\|^2}{2} - \frac{dK}{2} \log \sigma^2+\log \frac{2 \sqrt{n}}{\delta}}{n}}\Big)\Big)\,,
\end{align}
where we used that $\sigma_0=1$ since our prior is $\mathbb{P} = \cN(\eta_0 \mu_0, I_{dK})$ for Gaussian policies.

\textbf{Baseline 4: (Sakhi et al. 1, Mixed-Logit)} Here we consider the mixed-logit policies in \eqref{eq:es-app_logit_pac_bayes}.
\begin{align}
\min_{\mu \in \real^{dK}, \sigma >0, \lambda>0} \Big(\frac{1}{\tau\left(e^\lambda-1\right)}\Big(1-e^{-\tau \lambda \hat{\mathcal{L}}_n^\tau\left(\pi^{\textsc{mixL}}_{\mu, \sigma}\right)+\frac{\frac{\left\|\mu-\mu_0\right\|^2}{2} - \frac{dK}{2} \log \sigma^2+\log \frac{2 \sqrt{n}}{\delta}}{n}}\Big)\Big)\,.
\end{align}
where we used that $\sigma_0=1$ since our prior is $\mathbb{P} = \cN(\eta_0 \mu_0, I_{dK}) \times {\rm G}(0, 1)^K$ for mixed-logit policies. 

\subsubsection{\citet[Proposition 3]{sakhi2022pac}}
\begin{proposition}\label{prop:sakhi2}
Let $\tau \in (0, 1)$, $n \geq 1$, $\delta \in(0,1)$, let $\mathbb{P}$ be a fixed prior on $\mathcal{H}$, and let $\Lambda = \set{\lambda_i}_{i \in [n_\lambda]}$ a set of $n_\lambda$ positive scalars. Then with probability at least $1-\delta$ over draws $\cD_n \sim \mu_{\pi_0}^n$, the following holds simultaneously for all posteriors, $\mathbb{Q}$, on $\mathcal{H}$ and any $\lambda_i \in \Lambda$,
\begin{align}
\mathcal{L}\left(\pi_{\mathbb{Q}}\right) & \leq \hat{\mathcal{L}}_n^\tau\left(\pi_{\mathbb{Q}}\right) +\sqrt{\frac{D_{\mathrm{KL}}(\mathbb{Q} \| \mathbb{P})+\log \frac{4 \sqrt{n}}{\delta}}{2 n}}+\frac{D_{\mathrm{KL}}(\mathbb{Q} \| \mathbb{P})+\log \frac{2 n_\lambda}{\delta}}{\lambda} +\frac{\lambda}{n} g\left(\frac{\lambda}{ \tau n}\right) \mathcal{V}_{n}^\tau\left(\pi_{\mathbb{Q}}\right)\,,
\end{align}
where $g: u \rightarrow \frac{\exp (u)-1-u}{u^2}$ and $\mathcal{V}_{n}^\tau(\pi_{\mathbb{Q}})=\frac{1}{n} \sum_{i=1}^{n} \E{A \sim  \pi_{\mathbb{Q}}\left(\cdot | X_i\right)}{\frac{\pi_0\left(A | X_i\right)}{\max \left(\tau, \pi_0\left(A | X_i\right)\right)^2}}.$
\end{proposition}

\textbf{Baseline 5: (Sakhi et al. 2, Gaussian)} Here we consider the Gaussian policies in \eqref{eq:es-app_gaussian_pac_bayes}.

\begin{align}
\min_{\mu \in \real^{dK}, \sigma >0, \lambda \in \Lambda}\Big(\hat{\mathcal{L}}_n^\tau\left(\pi^{\textsc{gaus}}_{\mu, \sigma}\right) +\sqrt{\frac{\frac{\left\|\mu-\mu_0\right\|^2}{2} - \frac{dK}{2} \log \sigma^2+\log \frac{4 \sqrt{n}}{\delta}}{2 n}}+\frac{\frac{\left\|\mu-\mu_0\right\|^2}{2} - \frac{dK}{2} \log \sigma^2+\log \frac{2 n_\lambda}{\delta}}{\lambda} \nonumber\\ +\frac{\lambda}{n} g\left(\frac{\lambda}{ \tau n}\right) \mathcal{V}_{n}^\tau\left(\pi^{\textsc{gaus}}_{\mu, \sigma}\right)\Big)\,,
\end{align}
where we used that $\sigma_0=1$ since our prior is $\mathbb{P} = \cN(\eta_0 \mu_0, I_{dK})$ for Gaussian policies. 

\textbf{Baseline 6: (Sakhi et al. 2, Mixed-Logit)} Here we consider the mixed-logit policies in \eqref{eq:es-app_logit_pac_bayes}.

\begin{align}
\min_{\mu \in \real^{dK}, \sigma >0, \lambda \in \Lambda}\Big(\hat{\mathcal{L}}_n^\tau\left(\pi^{\textsc{mixL}}_{\mu, \sigma}\right) +\sqrt{\frac{\frac{\left\|\mu-\mu_0\right\|^2}{2} - \frac{dK}{2} \log \sigma^2+\log \frac{4 \sqrt{n}}{\delta}}{2 n}}+\frac{\frac{\left\|\mu-\mu_0\right\|^2}{2} - \frac{dK}{2} \log \sigma^2+\log \frac{2 n_\lambda}{\delta}}{\lambda} \nonumber \\ +\frac{\lambda}{n} g\left(\frac{\lambda}{ \tau n}\right) \mathcal{V}_{n}^\tau\left(\pi^{\textsc{mixL}}_{\mu, \sigma}\right)\Big)\,,
\end{align}
where we used that $\sigma_0=1$ since our prior is $\mathbb{P} = \cN(\eta_0 \mu_0, I_{dK}) \times {\rm G}(0, 1)^K$ for mixed-logit policies.

\subsection{Additional Results and Discussion} \label{app:add_results}

\begin{figure}[H]
  \centering  \includegraphics[width=\linewidth]{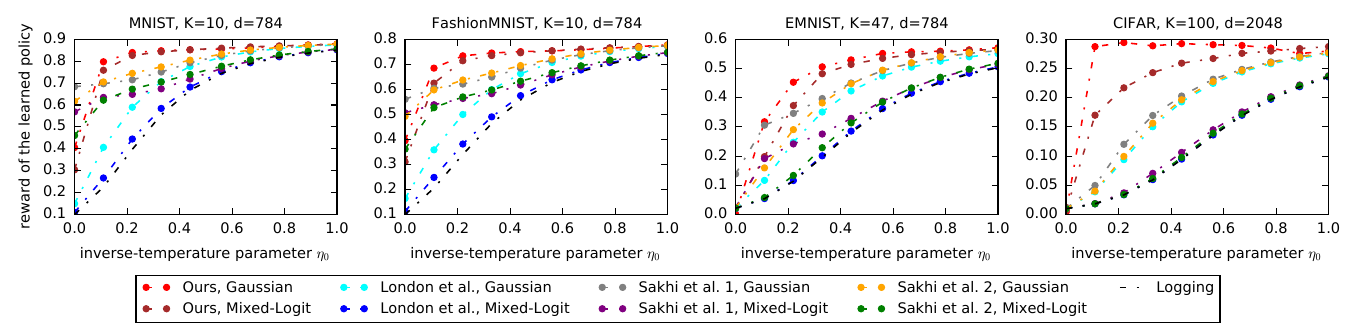}
  \caption{The reward of the learned policy for four datasets with varying quality of the logging policy $\eta_0 \in [0, 1]$.} 
  \label{fig:es-app_main_exp_results}
\end{figure}

\begin{figure}
  \centering  \includegraphics[width=0.7\linewidth]{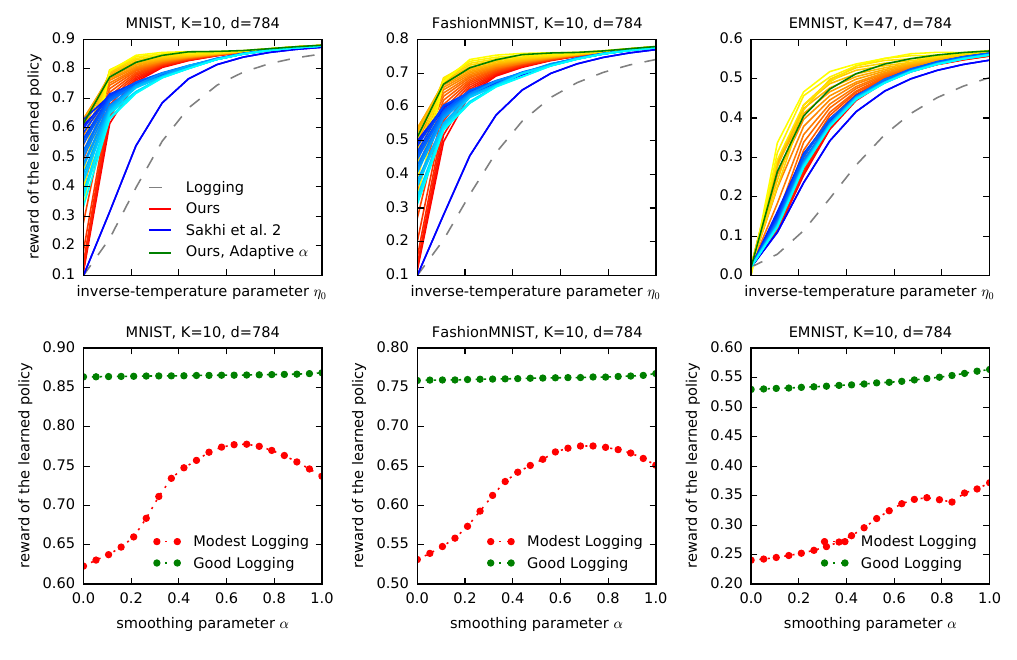}
  \caption{In the first row, we report the reward of the learned policy with 20 evenly space values of $\tau \in (0, 1)$ and $\alpha \in (0, 1)$ and varying $\eta_0 \in [0, 1]$, and for an adaptive and data-dependent $\alpha$ obtained using \eqref{eq:es-data_dependent_alpha} in \cref{subsec:es-data_dep_alpha}. The blue-to-cyan colors correspond to different values of $\tau$. The lighter the color, the higher the value of $\tau$. For instance, the cyan lines correspond to high values of $\tau$ while the blue ones correspond to very small values of $\tau$. Similarly, the red-to-yellow colors correspond to different values $\alpha$. The lighter the color, the higher the value of $\alpha$. For instance, the yellow lines correspond to high values of $\alpha$ while the red ones correspond to very small values of $\alpha$. Finally, the green curve corresponds to the reward of the learned policy using an adaptive and data-dependent $\alpha$ described in \eqref{eq:es-data_dependent_alpha} (\cref{subsec:es-data_dep_alpha}). In the second row, we report the \emph{average} reward of the learned policies using our method across the modest logging group ($\eta_0 \in [0, 0.5]$ in red) and the good logging group  ($\eta_0 \in [0.5, 1]$ in green).}
  \label{fig:es-app_varying_params}
\end{figure}

In \cref{fig:es-app_main_exp_results}, we report the reward of the learned policy using one of the considered methods. We make the following observations:
\begin{itemize}[topsep=0pt,itemsep=0pt]
\item \textbf{Choice of $\tau$ and $\alpha$:} in \cref{fig:es-app_main_exp_results}, we set $\tau= 1/\sqrt[\leftroot{-2}\uproot{2}4]{n} \approx 0.06$ and $\alpha = 1-1/\sqrt[\leftroot{-2}\uproot{2}4]{n} \approx 0.94$ so that when $n$ is large enough, both $\hat{\mathcal{L}}_n^\tau(\pi)$ and $ \hat{\mathcal{L}}_n^\alpha(\pi)$ approach $\hat{\mathcal{L}}_n^{\textsc{ips}}(\pi)$ \citep{ionides2008truncated}. This is because standard IPS should be preferred when $n \rightarrow \infty$. For completeness, we also show in \cref{fig:es-app_varying_params} that the choice of $\alpha$ and $\tau$ does not affect the conclusions that we make here. We also include in \cref{fig:es-app_varying_params} the results with an adaptive and data-dependent $\alpha$ obtained using \eqref{eq:es-data_dependent_alpha} in \cref{subsec:es-data_dep_alpha}. The results in \cref{fig:es-app_varying_params} will be discussed in detail after we finish analyzing the results in \cref{fig:es-app_main_exp_results}.
\item \textbf{Overall performance:} our method outperforms the baselines for any class of learning policies (Gaussian or mixed-logit) and any choice of logging policies. The only exception is when the logging policy is uniform.
\item \textbf{Effect of the class of learning policies:} the class of policies, Gaussian or mixed-logit, affects the performance of all the baselines. In general, Gaussian policies behave better than mixed-logit policies. However, this is less significant for our method; the performance of both Gaussian and mixed-logit policies are comparable, and in both cases, our method outperforms the baselines with Gaussian policies. Therefore, in general, Gaussian policies should be preferred over mixed-logit policies. But in case engineering constraints impose the choice of mixed-logit or softmax policies, then the performance of our method is robust to this choice. 
\item \textbf{Effect of the logging policy:} our method reaches the maximum reward even when the logging policy is not performing well. In contrast, the baselines only reach their best reward when the logging policy is already well-performing ($\eta_0 \approx 1$), in which case minor to no improvements are made. Note that the baselines have a better reward than ours when the logging policy is uniform. But our method has better reward when the logging policy is not uniform, that is when $\eta_0>0$. This is more common in practice since the logging policy is deployed in production and thus it is expected to perform better than the uniform policy.
\end{itemize}

In \cref{fig:es-app_varying_params}, we compare our method to \textbf{(Sakhi et al. 2)} with Gaussian policies since this was the best-performing baseline in our experiments in \cref{fig:es-app_main_exp_results}. Note that we did not include \texttt{CIFAR100} in \cref{fig:es-app_main_exp_results} as it was computationally heavy to run these experiments with varying $\eta_0$, $\alpha$ and $\tau$ for a very high-dimensional dataset such as \texttt{CIFAR100}. We consider $20$ varying values of $\tau$ and $\alpha$ evenly spaced in $(0, 1)$. We also include the results using the adaptive tuning procedure of $\alpha$ described in \cref{subsec:es-data_dep_alpha} (green curve). We make the following observations:

\begin{itemize}[topsep=0pt,itemsep=0pt]
\item \textbf{Adaptive and data-dependent $\alpha$:} This procedure is reliable since the performance with an adaptive $\alpha$ (green curve) is comparable with the best possible choice of $\alpha$. This is consistent for the three datasets.
\item \textbf{Effect of the choice $\alpha$:} as we observed before, the only case where the choice of $\alpha$ may lead to bad-performing policies is when the logging policy is uniform. When the logging policy is not uniform, our method outperforms the best baseline with the best $\tau$ for a wide range of values of $\alpha$. Also, note that there is no very bad choice of $\alpha$, in contrast with $\tau \approx 0$ that led to a very bad performing policy that slightly improved upon the logging policy. This attests to the robustness of our method to the choice of $\alpha$. Moreover, our bound regularizes better $\alpha$; it contains a bias-variance trade-off term for $\alpha$. Also, the bound of \textbf{(Sakhi et al. 2)} has a $1/\tau$ making it vacuous for small values of $\tau$.
\item \textbf{Best choice of $\alpha$:} To see the effect of $\alpha$ for varying problems, we consider the following experiment. We split the logging policies into two groups. The first is \emph{modest logging} which corresponds to logging policies whose $\eta_0$ is between $0$ and $0.5$. This includes uniform logging policies and other average-performing logging policies. The second is \emph{good logging} which corresponds to logging policies whose $\eta_0$ is between $0.5$ and $1$. After that, for each $\alpha$, we compute the average reward of the learned policy across either the group of modest or good logging policies. For each dataset, this leads to the two red and green curves in the second row of \cref{fig:es-app_varying_params}. Overall, we observe that $\alpha \approx 0.7$ leads to the best performance for the \emph{modest logging} group. Thus when the performance of the logging policy is average, regularizing the importance weights can be critical. In contrast, when the performance of the logging policy is already good, regularization is less needed and we can set $\alpha \approx 1$. Fortunately, one of the main strengths of this work is that our bound also holds for standard IPS recovered for $\alpha=1$. The bounds in all prior works cannot provide good performance for standard IPS due to their dependency on $1/\tau$.
\end{itemize}

\subsection{Learning Principles} \label{app:add_discussion}

\begin{figure}[H]
  \centering  \includegraphics[width=0.35\linewidth]{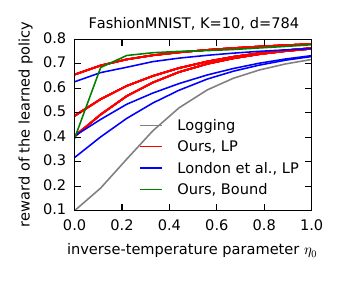}
  \caption{The reward of the learned policy using either our bound in \cref{thm:es-main_result} (referred to as \textbf{(Ours, Bound)} in green), our learning principle in \eqref{eq:es-learning_principle} (referred to as \textbf{(Ours, LP)} in red for multiple values of hyper-parameters) or the learning principle in \citet{london2019bayesian} (referred to as \textbf{(London et al., LP)} in blue) for multiple values of hyper-parameters).
} 
  \label{fig:es-app_lp_compare}
\end{figure}

Here we compare our bound in \cref{thm:es-main_result} and our learning principle in \eqref{eq:es-learning_principle} to the one in \citet{london2019bayesian}. We do not include the learning principle in \citet{swaminathan2015batch} since the one in \citet{london2019bayesian} enjoys similar performance and is far more scalable. The learning principle of \citet{london2019bayesian} is defined as 
\begin{align}\label{eq:es-lp_london}
\min_{\mu} \hat{\mathcal{L}}^\tau_n(\pi_\mu) + \lambda \norm{\mu - \mu_0}^2\,.
\end{align}
where $\lambda$ is a tunable hyper-parameters, $\pi_\mu$ is the softmax policy defined in \eqref{eq:es-app_softmax_pac_bayes} and $\mu \in \real^{dK}$ is its parameter vector. This learning principle is referred to as \textbf{(London et al., LP)}. In contrast, our learning principle is defined as 
\begin{align}\label{eq:es-our_learning_principle}
\hat{\mathcal{L}}^\alpha_n(\pi_{\mu}) + \lambda_1 \norm{\mu - \mu_0}^2  + \lambda_2  \operatorname{Var}_n^\alpha(\pi_{\mu}) + \lambda_3 B_n^\alpha(\pi_{\mu})\,,
\end{align}
where $\lambda_1, \lambda_2$ and $\lambda_3$ are tunable hyper-parameters and $\pi_{\mu}$ is the Gaussian policy in \eqref{eq:es-gaussian_pac_bayes} with a fixed $\sigma=1$. Our learning principle is referred to as \textbf{(Ours, LP)}. Finally, our bound in \cref{thm:es-main_result} with Gaussian policies is referred to as \textbf{(Ours, Bound)}. Similarly to the previous experiments, we set $\tau= 1/\sqrt[\leftroot{-2}\uproot{2}4]{n} \approx 0.06$ and $\alpha = 1-1/\sqrt[\leftroot{-2}\uproot{2}4]{n} \approx 0.94$ so that when $n$ is large enough, both $\hat{\mathcal{L}}_n^\tau(\pi)$ and $ \hat{\mathcal{L}}_n^\alpha(\pi)$ approach $\hat{\mathcal{L}}_n^{\textsc{ips}}(\pi)$ \citep{ionides2008truncated}. For the learning principles, we tried multiple values of hyper-parameters $\lambda, \lambda_1, \lambda_2$ and $\lambda_3$, all between $10^{-5}$ and $ 10^{-1}$. For instance, we found that the best hyper-parameter for \citet{london2019bayesian} is $\lambda=10^{-5}$ which matches the value they found in their \texttt{FashionMNIST} experiments. For our learning principle, the best hyper-parameters were $\lambda_1=10^{-5}, \lambda_2=10^{-5}$ and $\lambda_3=10^{-5}$. In contrast, our bound does not require hyper-parameter tuning. We report in \cref{fig:es-app_lp_compare} the reward of the learned policy on the \texttt{FashionMNIST} for all these methods with varying values of hyper-parameters. To reduce clutter, we only report the reward for good choices of hyper-parameters $\lambda, \lambda_1, \lambda_2$ and $\lambda_3$. We observe that for a wide range of hyper-parameters, our learning principle outperforms the one in \citet{london2019bayesian}. However, both learning principles are sensitive to the choice of hyper-parameters. In contrast, our bound does not require the tuning of any additional hyper-parameter and it achieves the best performance except for the uniform logging policy. In addition to being more theoretically grounded, this approach also enjoys favorable empirical performance without additional hyper-parameter tuning, an important practical consideration.

\subsection{Other Importance Weight Corrections} \label{app:other_corrections}

\begin{figure}[H]
  \centering  \includegraphics[width=0.8\linewidth]{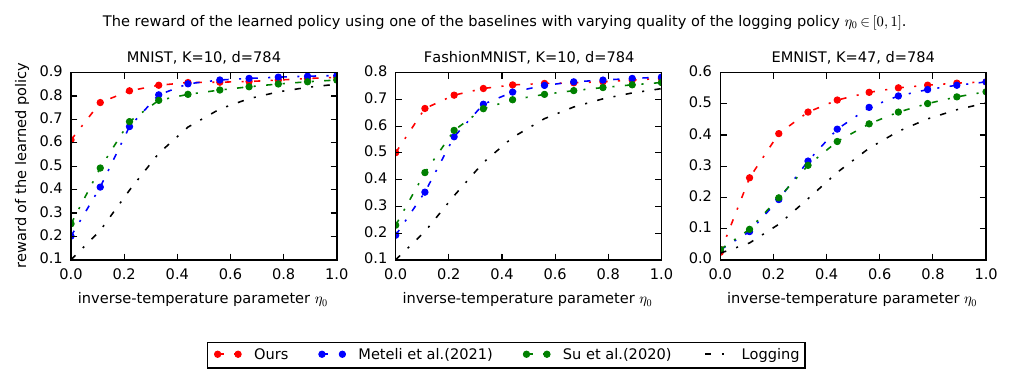}
  \caption{The reward of the learned policy with varying quality of the logging policy $\eta_0 \in [0, 1]$ using either our regularization ($\alpha$-\texttt{IPS}) or the ones in \citet{su2020doubly,metelli2021subgaussian}.} 
  \label{fig:es-app_other_corrections}
\end{figure}

\citet{su2020doubly,metelli2021subgaussian} also proposed corrections that are different from hard clipping (a detailed comparison is given in \cref{sec:es-ope}). However, they were not included in our main experiments since they do not provide generalization guarantees; they focus on OPE and only propose a heuristic for OPL in their Appendix B.2 and Section 6.1.2, respectively. Those heuristics are not based on theory, in contrast with ours which is directly derived from our generalization bound. However, for completeness, we also compare our regularization of importance weights to theirs. To make such a comparison, we use the hyper-parameters and tuning procedures provided in Section 6 and Appendix B.2 for \citet{metelli2021subgaussian} and Sections 5 and 6.1.2 for \citet{su2020doubly}. Overall, we observe in \cref{fig:es-app_other_corrections} that our method outperforms these baselines in OPL and the gap is more significant when the logging policy is not performing well.

